%% file: main.tex
\documentclass{ucetd}

\usepackage[T1]{fontenc}
\usepackage[utf8]{inputenc}
\usepackage{import}
\usepackage{subcaption,graphicx}
\graphicspath{{./}}
\usepackage{natbib}
\usepackage{mathtools}
\usepackage{amssymb}
\usepackage{amsthm}
\usepackage{booktabs}
\usepackage{multirow}
\usepackage{xcolor}
\usepackage{algorithm}
\usepackage{algpseudocode}
\usepackage{listings}
\usepackage{xspace}
\usepackage{bbm}
\usepackage{bm}
\usepackage{enumitem}
\usepackage{wrapfig}
\usepackage{calc}
\usepackage[export]{adjustbox}
\usepackage{array}
\usepackage{collcell}
\usepackage{float}
\usepackage{microtype}
\usepackage{tikz}
\usepackage{pifont}
\usepackage{soul}
\usepackage{dialogue}
\usepackage{contour}
\usepackage[normalem]{ulem}
\usepackage{environ}
\usepackage{stmaryrd}
\usepackage{tabularx}
\usepackage{tcolorbox}
\tcbuselibrary{breakable,skins}
\usepackage[percent]{overpic}

\newcommand{\shortparagraph}[1]{\textbf{#1}}
\newcommand{\defeq}{\mathrel{\stackrel{\textnormal{\tiny def}}{=}}}
\newcolumntype{P}[1]{>{\centering\arraybackslash}p{#1}}
\newcolumntype{C}[1]{>{\centering\arraybackslash}m{#1}}
\newcolumntype{H}{>{\setbox0=\hbox\bgroup}c<{\egroup}@{}}
\newcommand{\mytexttt}[1]{\raggedright\texttt{#1}}
\newcolumntype{M}[1]{>{\collectcell\mytexttt}p{#1}<{\endcollectcell}}

\newcommand{\revise}[1]{#1}

\newcommand{\tmlrrevise}[1]{#1}
\newcommand{\tmlrrevisee}[1]{#1}
\newcommand{\mvhnrevise}[1]{#1}
\newcommand{\revisestart}{}
\newcommand{\reviseend}{}
\newcommand{\ych}[1]{}
\newcommand{\yc}[1]{}
\newcommand{\mina}[1]{}
\newcommand{\thh}[1]{}
\newcommand{\jon}[1]{}
\newtcolorbox{takeaway}{
  colback=blue!4, colframe=blue!50!black,
  boxrule=0.6pt, arc=2pt, boxsep=2pt,
  left=6pt, right=6pt, top=3pt, bottom=3pt,
  before skip=6pt, after skip=6pt}
\newcommand{\tkw}{\textbf{\textcolor{blue!50!black}{Takeaway.}}\ }

\newcommand{\circone}{\ding{172}\xspace}
\newcommand{\circtwo}{\ding{173}\xspace}
\newcommand{\circthree}{\ding{174}\xspace}
\newcommand{\circfour}{\ding{175}\xspace}
\newcommand{\circfive}{\ding{176}\xspace}

\newcommand{\outputVar}{\mathrm{Y}}
\newcommand{\outputval}{y}
\newcommand{\inputVar}{\mathrm{X}}
\newcommand{\inputval}{x}

\newcommand{\algname}{Exploratory Annealed Decoding\xspace}
\newcommand{\alg}{EAD\xspace}
\newcommand{\name}{\textsc{BACo}\xspace}
\newcommand{\problem}{diversity-quality trade-off\xspace}
\newcommand{\recode}{\texttt{ReCode}\xspace}

\newcommand{\fakeparagraph}[1]{\vspace{1mm}\noindent\textbf{#1}}

\newcommand{\eg}{e.g.,\xspace}
\newcommand{\ie}{i.e.,\xspace}
\newcommand{\etc}{etc.\xspace}

\newcommand{\appref}[1]{Appendix~\ref{#1}}

\newcommand{\dataset}[1]{\mathbb{X}_{#1}}
\newcommand{\data}[1]{X_{#1}}
\newcommand{\model}{M}
\newcommand{\modelname}[1]{M_{#1}}
\newcommand{\modelclf}{M_{\text{CLF}}}

\newcommand{\modelio}[3]{#1\left(#2\right)\left[#3\right]}

\newcommand{\sword}[1]{w_{{#1}}}

\newcommand{\semb}[1]{\vec{e}_{{#1}}}
\newcommand{\interpret}[3]{\phi_{#1}({#2}, {#3})}

\newcommand{\Real}{\mathbb{R}}

\newtheorem{theorem}{Theorem}[section]

\newtheorem{proposition}{Proposition}[section]

\newtheorem*{definition*}{Definition}

\newcommand{\agentname}{%
  \raisebox{-0.1\fontcharht\font`A}{%
    \includegraphics[height=1.2\fontcharht\font`A]{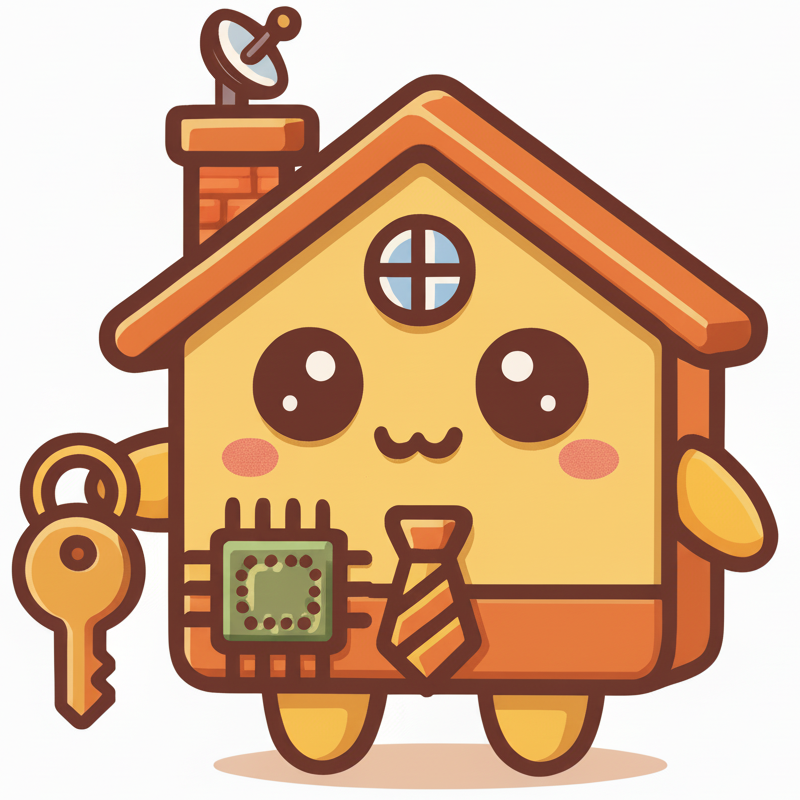}%
}~\texttt{AI Realtor}\xspace}
\newcommand{\llm}{\text{LLM}_{\text{gen}}}
\newcommand{\llmembed}{\text{LLM}_{\text{embed}}}
\newenvironment{myquote}[1]%
  {\list{}{\leftmargin=#1\rightmargin=#1}\item[]}%
  {\endlist}
\setcitestyle{open={(},close={)}}

\newcommand{\thesistitle}{Beyond Surface Alignment: Grounding the Dynamics of Situational Understanding and Generative Control in LLMs}
\newcommand{\thesisauthor}{Chenghao Yang}
\department{Computer Science}
\division{Physical Sciences}
\degree{Doctor of Philosophy}
\date{March 2026}

\title{\thesistitle}
\author{\thesisauthor}

\dedication{Dedicated to my younger self, who decided to go through this challenging life experience and gave up all other prestigious opportunities. }
\epigraph{Epigraph Text}

\usepackage{doi}
\usepackage{xurl}
\hypersetup{bookmarksnumbered,
            linktoc=all,
            pdftitle={\thesistitle},
            pdfauthor={\thesisauthor},
            pdfsubject={},
            pdfborder={0 0 0}}
\makeatletter
\let\ORG@hyper@linkstart\hyper@linkstart
\protected\def\hyper@linkstart#1#2{%
  \lowercase{\ORG@hyper@linkstart{#1}{#2}}}
\makeatother

\usepackage[capitalize,noabbrev]{cleveref}
\usepackage{longtable}
\input{math_commands}

\usepackage{mathpazo}
\newcounter{exampleCounter}[section]
\renewcommand{\theexampleCounter}{\thesection.\arabic{exampleCounter}}
\crefname{exampleCounter}{Example}{Examples}
\tikzstyle{title}=[right=10pt,fill=black,text=white!50]
\tikzstyle{context}=[thick,rectangle,draw=gray,inner sep=10pt, inner ysep=10pt]
\NewEnviron{example}[1][{}]{%
    \par
    \centering
    \addvspace{\medskipamount}%
    \begin{tikzpicture}
        \node[context](box){%
            \begin{minipage}{0.43\textwidth}
                \refstepcounter{exampleCounter}%
                \BODY
            \end{minipage}};
        \node[title] at (box.north west){\textbf{Ex \theexampleCounter\ #1}};
    \end{tikzpicture}%
    \par
}

\newcommand{\codefont}{\fontfamily{lmtt}\selectfont}
\definecolor{codepurple}{rgb}{0.58,0,0.82}
\lstdefinestyle{datalogstyle}{
	basicstyle={\codefont\small},
	xleftmargin={14pt},
	numbers=left,
	frame=l,
	stepnumber=1,
	firstnumber=1,
	numberfirstline=true,
	tabsize=2,
	showtabs=false,
	showspaces=false,
	showstringspaces=false,
	extendedchars=true,
	breaklines=true,
	columns=fullflexible,
	keepspaces=true,
	escapeinside={@}{@},
	firstnumber=last,
	captionpos=b,
	commentstyle=\color{black!65},
	numberstyle=\tiny\color{black!65},
	stringstyle=\color{codepurple},
	breakatwhitespace=false, 
	keepspaces=true,                 
	numbersep=5pt,                  
	showspaces=false,                
	showstringspaces=false,
	showtabs=false,
	aboveskip={0.2\baselineskip},
	belowskip={-0.2\baselineskip},
}

\newcommand{\RN}[1]{%
  \textup{\uppercase\expandafter{\romannumeral#1}}%
}

\definecolor{applegreen}{rgb}{0.55, 0.71, 0.0}

\newcommand{\datahmei}[1]{\texttt{\codefont#1}}

\newcommand{\blk}[1]{\textcolor{blue}{\datahmei{#1}}}

\newcommand{\modelresponse}[1]{r_{#1}}
\newcommand{\critique}[1]{c_{#1}}
\newcommand{\instruction}[1]{I_{#1}}
\newcommand{\protocol}{SR$^2$V}

\begin{document}
\input{arxiv_keep_assets}
\maketitle

\makecopyright
\makededication

\tableofcontents
\listoffigures
\listoftables

\acknowledgments
\input{acknowledge}

\abstract
\input{abstract}

\mainmatter

\chapter{Introduction}
\input{introduction}

\part{Situational Grounding}

\chapter{SitTest: Testing Situational Understanding in ChatGPT}
\label{chap:sittest}

\input{chapters/sittest}

\chapter{ReCode: Robustness Evaluation of Code Generation Models}
\label{chap:recode}

\input{chapters/recode}

\part{Generative Grounding}

\chapter{LLM Probability Concentration: How Alignment Shrinks the Generative Horizon}
\label{chap:branching_factor}
\input{chapters/branching_factor}

\chapter{Hindsight: Testing Limits on Reflective Thinking in LLMs}
\label{chap:hindsight}

\input{chapters/hindsight}

\chapter{Amortized Interpretability: Efficient Shapley Values Estimation}
\label{chap:amortized}

\input{chapters/amortized}

\part{Dynamic Control}

\chapter{AI Realtor: Towards Grounded Persuasive Language Generation for Automated Copywriting}
\label{chap:realtor}
\input{chapters/ai_realtor}

\chapter{Optimizing Diversity and Quality through Base-Aligned Model Collaboration}
\label{chap:collaboration}
\input{chapters/base_aligned_collaboration}

\chapter{Annealed Sampling for Verifiable Reinforcement Learning}
\label{chap:annealed}

\input{chapters/annealed_sampling}

\chapter{Rationale-Grounded Addiction Support: Identifying OUD Phases}
\label{chap:addiction}

\input{chapters/addiction}

\chapter{Conclusion}
\label{chap:conclusion}
\input{chapters/conclusion}

\input{chapters/appendix}

\makebibliography
\nocite{*}

\end{document}

%% file: math_commands.tex
\def\Figref#1{Figure~\ref{#1}}

\def\Tabref#1{Table~\ref{#1}}

\def\Secref#1{Section~\ref{#1}}

\def\eqref#1{equation~\ref{#1}}

\def\1{\bm{1}}

\DeclareMathAlphabet{\mathsfit}{\encodingdefault}{\sfdefault}{m}{sl}
\SetMathAlphabet{\mathsfit}{bold}{\encodingdefault}{\sfdefault}{bx}{n}

\newcommand{\E}{\mathbb{E}}

\DeclareMathOperator*{\argmax}{arg\,max}
\DeclareMathOperator*{\argmin}{arg\,min}

\makeatletter
\newcommand*\bigcdot{\mathpalette\bigcdot@{.5}}
\newcommand*\bigcdot@[2]{\mathbin{\vcenter{\hbox{\scalebox{#2}{$\m@th#1\bullet$}}}}}
\makeatother

\newcommand{\sscomment}[1]{}
\newcommand{\hxcomment}[1]{}
\newcommand{\mhcomment}[1]{}
\newcommand{\ychcomment}[1]{}

%% file: acknowledge.tex
I am extremely fortunate to have Allyson Ettinger as my PhD advisor. Allyson has been incredibly kind and patient with me. During the first year of my PhD, even though I often came to our meetings with ambitious but unrefined ideas—blindly chasing upfront trends for frontier models that largely deviated from her research group's focus—she always listened carefully. She nudged me to translate my premature thoughts into rigorous, clear writing. It was not until then that I realized writing is not merely a tedious step in getting a paper published, but a powerful tool to formalize broken thoughts and gain clarity on what we should pursue. She generously used her funding to support me as an RA while giving me the highest level of research freedom imaginable—I was even able to explore and learn MLsys directions that I cannot imagine other professors in human-subject disciplines (e.g., Linguistics, HCI, CogSci) would appreciate. I hope to be as supportive as her in my future career, should I have mentees or a small team to lead.

I also want to sincerely thank Haifeng Xu, Mina Lee, and Ari Holtzman, who kindly helped me when Allyson left the university and provided precious collaboration opportunities. Without their help, I cannot imagine how I would have survived those darkest moments when I had no advisor or funding, struggling to finish my PhD program. Their guidance and support broadened my horizons and further cultivated my capability to carry out independent and impactful research. Furthermore, I am grateful for the mentorship from Karen Livescu, Jiawei Zhou, Haoyue (Freda) Shi, Karthik Goyal, Zhewei Sun, Victor Veitch, David McAllester (in Chicago), He He, Xuezhe Ma, Zhaoran Wang, Shi Feng, Peter West, and Kai-Wei Chang (outside Chicago). Though unfortunately I did not have time to conduct more in-depth collaboration, I will never forget the precious moments we spent talking or hanging out together.

It is also a privilege to work with and learn from many brilliant PhD colleagues at the TTIC \& UChicago NLP\footnote{Now it is called ``Communication \& Intelligence''.} community, especially Chaoqi Wang, Yibo Jiang, Han Liu, Jibang Wu, Sida Li, Yichen Wang, Lin Gui, Chenxiao Yang, Yiyun (Harvey) Fu, Zhen Xu, Rosa Zhou, Chacha Chen, Mourad Heddaya, Christopher Wolfram, Nihar Mauskar,\footnote{I swear I would never forget your joke of ``Doing interpretability is like doing stamp collection''.} David Reber, Todd Nief, and Aswathy Ajith. Though not in my core discipline, my special thanks go to Wenyi Wang, Jie Zhu, Haochen Pan, Jun Yang, Weichen Li, and Ziyi Zhang. They come from the UChicago CS System community, and they truly saved me from desperation, providing emotional support and actionable guidance during my darkest moments. I have received so much love from the UChicago \& TTIC student community, and I only wish I could have done more to pay it back.

I am grateful for Yanhong Li in particular. Yanhong is my first and best-ever mentee. I often feel I am neither as motivated nor as hard-working when I think of her. Mentoring talented and highly motivated students like Yanhong has been a pleasure, fulfilling my broken dream of becoming faculty—at the very least, I can proudly say I helped foster a rising star in my beloved community. I wish her all the best in the future.

I want to thank Yu as well, my spiritual friend. I really loved our monthly talks on absolutely everything. Thank you for accompanying me through those happy and sad times. I love you, you are my sunshine.

Nowadays, the speed of LLM research and the scaling of models has become so fast that it necessitates more online collaboration and active communication. Fortunately, I met a line of great researchers and successfully built very strong relationships. They continuously taught me so much, and I want to thank them for their help: Ruiqi Zhong (UC Berkeley), Zhaofeng Wu (MIT), Hao Zhu (Stanford), Yizhong Wang (UW), Xi Ye (UT Austin), Kaichao You (THU \& vLLM), Wei Xiong (UIUC), Tenghao Huang (USC), Deqing Fu (USC), Zhaowei Wang (HKUST), Jinyan Su (Cornell),  Boyuan Zheng (OSU), Kai Zhang (OSU), Sanxing Chen (Duke), Yiyuan Li (UNC), and Fei Yu (UCI). I also want to thank my friends from the industry who offered valuable insights and mentorship on research and career: Yuandong Tian (Meta), Zi Yang (Meta), Zhuokai Zhao (Meta), Zinan Lin (Microsoft), Yuhao Zhang (Samaya AI), Zuxin Liu (OpenAI), Eric Wallace (OpenAI), Yao Fu (Google Deepmind), Saloni Potdar (Apple), Yanda Chen (Anthropic), Ziqi Wang (Anthropic), Yuchen Lin (xAI), and Jie Huang (xAI). The shared mission of building and understanding this new type of intelligence, and aligning it to human needs, brings us together, though we come from different backgrounds. I feel this is a fantastic era in which I can find so many like-minded friends to talk and work with.

Finally, I want to thank my parents and family for their continued support and understanding along this long journey. I have wanted to do a PhD since 2019, and I have not been back to China for almost seven years. My family always supports me, listens to my complaints and sadness, encourages me to look forward, and cheers for my successes. I do not know how to repay them, but I promise I will try to visit and reunite with you all more often in the future, and do my best to contribute to a better society.

\vspace{1em}
\noindent\textbf{Disclosure of AI Use}

In accordance with the guidelines for AI-generated content, I disclose the use of AI tools in the preparation of this dissertation. I utilized ChatGPT (OpenAI), Gemini 3 Pro (Google Deepmind) and other large language model-based tools primarily for LaTeX coding assistance, grammar checking, and correcting typographical errors. The core scientific content, analysis, and text of this dissertation are my original work, with AI assistance limited to editorial and formatting support.

%% file: abstract.tex
The current alignment tuning paradigm for Large Language Models (LLMs) prioritizes surface-level behaviors—fluency, safety, and tonal consistency. While effective for casual chat, this thesis argues that such surface alignment masks a lack of grounding, creating models that are stylistically confident but situationally brittle. We propose a framework of \textbf{Grounded Alignment}, investigating this disconnect through a dual analysis of how models process context ({Input}) and how they structure generation ({Output}), followed by mechanisms to align these grounded behaviors to human needs.

First, we evaluate failures in \textbf{Situational Grounding}. Through \emph{SitTest}, we reveal that despite massive context windows, state-of-the-art models struggle to maintain a consistent ``mental model'' of a changing environment. This fragility is further quantified by \emph{ReCode}, which demonstrates that models rely on surface heuristics rather than resolving deep syntactic dependencies. These findings highlight a critical gap: models ``read'' extensive histories without truly ``understanding'' the evolving situation.

Second, we evaluate the dynamics of \textbf{Generative Grounding}. We introduce the \emph{Branching Factor} (BF) to map the landscape of LLM generation, finding that standard alignment tuning artificially constricts this landscape into premature ``stylistic collapse''. We further probe this with \emph{Hindsight}, revealing that models often fail to understand their own generations.

Finally, we propose \textbf{Dynamic Control} mechanisms to align models to human values through grounded interaction. We introduce \emph{AI Realtor}, an agentic framework that demonstrates the importance of context engineering to compensate for poor situational grounding. To handle diverse value judgments, we propose \emph{Base-Aligned Model Collaboration}, decoupling exploration from stylistic constraints. We also present \emph{Annealed Sampling} for verifiable reinforcement learning and apply these principles to \emph{Addiction Support}, where model-generated rationalization provides a new communication interface for high-stakes domains. Collectively, this work moves beyond the facade of surface alignment, offering a roadmap for building agents robustly anchored in both their context and their generation.

%% file: introduction.tex
\section{The Illusion of Alignment}

    The ascent of Large Language Models (LLMs) has been defined by a rapid transition from raw predictive probabilistic models to polished conversational agents. This transformation is largely driven by alignment tuning paradigms, including Supervised Fine-Tuning (SFT) and Reinforcement Learning from Human Feedback (RLHF). These technologies have successfully trained models to follow instructions, adhere to safety guidelines, and maintain a helpful persona. To the end-user, modern LLMs appear remarkably robust: they can converse fluently over long horizons, generate syntactically correct code, and offer decisive answers to complex queries.

    However, this thesis argues that this observable competence is often superficial. The current alignment tuning paradigm prioritizes \textit{surface-level behaviors}—fluency, tone, and format—at the expense of \textit{grounding}. We define grounding not merely as factual retrieval, but as the model's ability to robustly anchor its internal representations in the dynamic reality of the task (\emph{Situational Grounding}) and to faithfully reflect its own uncertainty during generation (\emph{Generative Grounding}). This thesis characterizes the gap between surface behavior and deep understanding as a problem of \textbf{Grounded Alignment}.

    When stripped of their stylistic veneer, state-of-the-art aligned models reveal a startling brittleness. They may digest a 100-turn dialogue yet fail to track a single changing variable; they may generate code that looks correct but breaks under trivial renaming; they may speak with absolute certainty while hallucinating basic facts. This thesis posits that we have optimized models to \textit{sound} aligned rather than to \textit{be} grounded. We investigate this disconnect through a dual analysis of the model's input processing and output generation, proposing a shift from static, surface-level alignment to dynamic, grounded generative control.

\section{Diagnosing the Gap in Situational Grounding}

    The first challenge lies in the input: Can models truly understand the situations they read? As context windows expand to millions of tokens, the prevailing assumption is that models can seamlessly integrate vast amounts of information. Our work challenges this assumption, revealing a fundamental gap between ``reading'' context and ``maintaining'' state.

    In \textbf{SitTest} \citep{yang2023can}, we test the limits of situational understanding in chat-based models. By simulating dynamic environments where state variables change over time, we demonstrate that models suffer from a ``Lost-in-the-Middle'' phenomenon for state tracking. Despite having access to the full interaction history with the environments, models frequently fail to update their internal mental states, hallucinating updates that never occurred. This suggests that alignment has improved the model's ability to retain the \textit{appearance} of conversation, but not the underlying consistency required for reliable agency.

    We further probe this fragility in \textbf{ReCode} \citep{wang2023recode}, focusing on the domain of code generation. Here, we find that models often rely on surface heuristics—such as variable naming conventions—rather than resolving deep syntactic dependencies. When presented with semantic-preserving perturbations (e.g., renaming variables or altering docstrings), model performance degrades significantly. This fragility underscores that what appears to be robust reasoning is often a fragile pattern-matching capability, easily disrupted when surface cues are removed.

\section{Characterizing the Dynamics of Generative Grounding}

    The second challenge lies in the output: How does alignment shape the way models generate information? To answer this, we must look beyond static evaluation metrics and examine the mechanics of generation itself.

    We introduce the concept of the \textbf{Shrinking Landscape} of generation. Using our proposed metric, the \textbf{Branching Factor (BF)} \citep{yang2026alignment}, published in TMLR, we map the probability concentration of the model's output distribution at each decoding step. Our mechanistic study reveals a critical insight: alignment tuning acts as a global constraint that artificially constricts this landscape. While base models exhibit a natural flow of entropy—starting with broad exploration and narrowing as the sequence progresses—aligned models are forced into low-entropy trajectories from the very first token. We term this \textbf{Premature Stylistic Collapse}. By forcing models to adopt a specific helpful ``style'', we inadvertently prune vast regions of valid semantic space.

    Beyond the statistical landscape, we examine whether models truly understand their own outputs. In \textbf{Hindsight} \citep{li2024hindsight}, we analyze self-reflection and find that asking models to reflect on their prior generations often fails because they lack a deep semantic grasp of what they have just produced. Without external feedback, self-reflection can degrade performance as the model reinforces its initial, ungrounded hallucinations, revealing a disconnect between generation and understanding.

    To further dissect these generative mechanics, we introduce \textbf{Amortized Interpretability} \citep{yang2023efficient}. Recognizing that traditional Shapley Value estimation is computationally prohibitive, we propose a framework that trains a separate model to predict token importance. This enables efficient attribution, providing a theoretically-grounded way to diagnose spurious correlations and attention sinks that drive ungrounded generation.

\section{Restoring Grounding through Dynamic Control}

    Having diagnosed the failures of surface alignment in both understanding and generation, this thesis proposes a path forward: we must move from static constraints to dynamic, dynamics-aware control mechanisms to align models with human values.

    First, we present \textbf{AI Realtor} \citep{wu2025grounded}, accepted to CAIS 2026, a framework for grounded persuasive generation. Given that models often lack robust situational understanding, we show that careful context engineering becomes critical. By architecturally enforcing grounding through retrieved context before applying personalization, we compensate for the model's inherent limitations, achieving automated copywriting that is persuasive yet factually rigorous.

    To address the diverse value judgments inherent in human preference, we introduce \textbf{Base-Aligned Model Collaboration} \citep{wang2025optimizing}, accepted to ICML 2026. Recognizing that a single model cannot satisfy all conflicting requirements (e.g., diversity vs. coherence), we propose a dynamic routing framework that decouples these needs. By leveraging the base model for exploration (high BF) and the aligned model for structural guidance (low BF), we achieve flexible control over generation.

    For verifiable reinforcement learning, we propose \textbf{Annealed Sampling} \citep{yang2025let}. Instead of static temperature thresholds, this method synchronizes sampling with the natural ``cooling'' of the model's probability distribution, preventing premature collapse and improving reasoning performance.

    Finally, in the domain of social health, we apply these principles to \textbf{Rationale-Grounded Addiction Support} \citep{yang2023identifying}. By assisting generation with explicit rationales, we create a new communication interface where the model's reasoning process is transparent. This grounded thinking not only improves performance in detecting subtle stages of Opioid Use Disorder but also builds trust in high-stakes applications.

\section{Thesis Structure}

    This thesis is organized as follows:
    \begin{itemize}
        \item \textbf{Part I: Situational Grounding} evaluates the model's ability to maintain context, presenting the \emph{SitTest} environment for state tracking and the \emph{ReCode} benchmark for robustness.
        \item \textbf{Part II: Generative Grounding} evaluates the dynamics of output generation. We present the theory of \emph{LLM Probability Concentration} (Branching Factor), analyze the limits of self-reflection in \emph{Hindsight}, and introduce \emph{Amortized Interpretability} for efficient attribution.
        \item \textbf{Part III: Dynamic Control} introduces interventions to align models to human needs. We begin with the \emph{AI Realtor} framework, followed by \emph{BACo} (Base-Aligned Collaboration) and \emph{Annealed Sampling} (EAD), and conclude with \emph{Addiction Support} as a case study in high-stakes application.
    \end{itemize}

    By bridging the gap between how models process the world and how they speak about it, this work aims to lay the foundation for the next generation of AI: agents that are not just aligned to our preferences, but grounded in our reality.

%% file: chapters/sittest.tex
\section*{Chapter Overview}
Understanding sentence meanings and updating information states appropriately across time---what we call ``situational understanding'' (SU)---is a critical ability for human-like AI agents. SU is essential in particular for chat models, such as ChatGPT, to enable consistent, coherent, and effective dialogue between humans and AI. In this chapter, we tackle these questions, proposing a novel synthetic environment for SU testing which allows us to do controlled and systematic testing of SU in chat-oriented models, through assessment of models' ability to track and enumerate environment states.

\graphicspath{{./}}

\makeatletter
\def\input@path{{./}}
\makeatother

\begingroup
\lstset{style=datalogstyle}
\input{past_papers/sittest/sections/introduction}

\input{past_papers/sittest/sections/related_work}

\input{past_papers/sittest/sections/environment}

\input{past_papers/sittest/sections/experiments}

\input{past_papers/sittest/sections/analysis}

\input{past_papers/sittest/sections/conclusion}

\input{past_papers/sittest/sections/limitations}

\endgroup

%% file: past_papers/sittest/sections/introduction.tex
\section{Introduction}
Understanding the meaning of language inputs and their impact on information states is essential for building a communicative human-like AI agent (e.g., ChatGPT~\citep{openai2022chatgpt}). This capability requires an agent to know \textbf{truth conditions}~\citep{lewis1976general, kratzer1998semantics} of a given input, and how that input \textbf{updates the context} over time \citep{veltman1996defaults}. For instance, the natural language input ``Open BOX-4, obtain KEY-1'' in a game environment should assert updates ``$\blk{OPENED}(\text{BOX-4})=\text{True}$'' and ``$\blk{OBTAINED}(\text{KEY-1})=\text{True}$'' in the agent's representation of the environment state. 
We describe this ability as situational understanding (SU), as the agent needs to ground language in situations and understand \textbf{situational changes}.

Recent research indicates that despite the tremendous success of Large Language Models (LLMs) (e.g., GPT-3~\citep{brown2020language}), these models still fail to understand situational changes, and cannot serve as human-like agents to accomplish real-world tasks. For example, evidence suggests that models fail to detect sarcasm expressed in underlying context~\citep{suzgun2022challenging},
infer incorrect logic states when context changes in games~\citep{li2022language} and fail to track entity state changes in a discourse~\citep{kim-schuster-2023-entity}. While these works have shown important limitations in LLMs, it remains unclear why these  models show these limitations, and there is less work that shows the extent to which these limitations persist in more recent, chat-trained models like ChatGPT. 

In this work we seek to shed light on both of these questions. To test the situation tracking ability of ChatGPT and other chat models, we design a synthetic environment for controlled testing of models' ability to follow instructions and maintain consistent understanding of situation states. We design a multi-stage testing framework, and include various controls to avoid models relying on mappings memorized from pre-training rather than doing real-time state tracking. Our environment allows for flexible testing under different conditions, as well as controlled follow-up analyses to enable closer examination of causal factors.

We apply our tests primarily to ChatGPT, which we find to outperform other models fine-tuned on chats. Our results show that ChatGPT's performance reflects a failure to retain coherent and correct environment states across time, despite the simplicity of the task and the fact that ChatGPT has access to the full dialogue history in its input window. Our follow-up analyses suggest that a major contributor to this degradation is failure to retain prior states in memory, as well as susceptibility to spurious hallucinated updates (which can also contribute to accidentally correct outputs). Our findings suggest overall that ChatGPT lacks the ability to track situation state changes robustly.

%% file: past_papers/sittest/sections/related_work.tex
\section{Related work}
        Situational understanding ability is essential for building agent-like real-world intelligent systems. In  comprehensive benchmarks, {Big-Bench-Hard (BBH, \citet{suzgun2022challenging})} finds that for GPT-family models and PaLM 540B models~\citep{chowdhery2022palm}, even equipped with the state-of-the-art Chain-of-Thought prompting, still fail on tasks that require situational information (e.g., to detect sarcasm). HELM~\citep{liang2022holistic} also points out LLMs can lack state tracking ability based on evaluation results for 30 models on bAbI~\citep{weston2016babi} environment. 

    In synthetic and controlled environments, \citet{li2021implicit, jacob-2022-language-agent-models, li2022language} does probing on BART~\citep{lewis2020bart}, T5~\citep{raffel2020exploring} and GPT-3 using the SCONE~\citep{long2016simpler} and TextGame~\citep{cote2019textworld} datasets, and show that these models lack the ability to track and infer information states.
    \citet{kim-schuster-2023-entity} re-analyze the results in ~\citet{li2021implicit} and find that GPT-3/3.5 and Flan-T5~\citep{chung2022scaling} cannot track entity state changes in a discourse.~\citet{toshniwal2022chess} instruct GPT-2~\citep{radford2019language} to play chess and find it difficult to track board states per move. 

    We build on these previous works in two primary ways. First, these works mainly use LLMs as feature extractors (e.g., train probe models over intermediate representations or final-layer representations),
    and none of them discuss whether situational understanding limitations still exist in the recent powerful and widely used ChatGPT system. We investigate here whether ChatGPT has the critical 
    underlying 
    situational understanding to generate coherent, consistent and effective dialogue ~\citep{karttunen1976discourse, kamp2011discourse, wu2019transferable}. Second, tests used in previous work are susceptible to interference from confounds present when fine-tuning linear probes, or shortcuts in the testing environment that can be utilized to bypass situational understanding tests~\citep{kim-schuster-2023-entity}. 
    We aim to address some of these concerns by introducing a number of additional controls in our tests. Finally, we also carry out a more in-depth exploration of model update dynamics, to better understand causes for patterns of performance.

%% file: past_papers/sittest/sections/environment.tex
\section{Building a Situational Testing Environment}

What we want to test is models' ability to process situational changes conveyed by language inputs, and to maintain internal representations of the corresponding situation states. 
To test this, we use a synthetic box-moving environment like TextGame~\citep{li2021implicit}, where we have full access to underlying game states, but eliminate complicated reward pursuing and route branching. Having this kind of full-information synthetic environment is helpful to test models' evolving understanding of environment states as the input describes progressively more situational changes.

\subsection{Environment Setup}

The environment includes two basic components:

\begin{enumerate}
    \item \textbf{Instructions.} Instructions directed to an agent, defining the agent's quest and providing information about the environment. As we will describe below, this component also sometimes contains explanations of the meanings of non-language functors and arguments to be used in the output.
    \item \textbf{Steps and queries.} Descriptions of steps taken by the agent, followed by a query that prompts for enumeration of all relevant environment states after the action is taken. Queries and answers to the queries take the format of sets of logical representations of states with corresponding truth values. 
\end{enumerate}

In our tests, the input provided to the model includes the task instructions, along with $n$ few-shot Step-Query-Answer examples to demonstrate the task. In the zero-th step,  we define the step as ``Do nothing'' and provide in the Answer the full set of correct truth values as an initialization. The $n$ few-shot Step-Query-Answer examples are then followed by a series of Steps without Query-Answer components, followed by a Test Step and Test Query that the model is expected to complete with the full enumeration of environment states that hold after that step is taken. An example of our environment format, to be further explained below, is shown in \cref{lst: running example}.

\begin{figure}[htbp]
\begin{lstlisting}[label={lst: running example}]
Instructions: As an agent, you need to find the way to go out of this quest. Currently, there are several boxes in front of you and there is a key inside each box. You can use only one of these keys to open the door and finish this quest. There are 5 boxes and 5 keys here. Boxes are identified as jqC-X and Keys are identified as bsS-X.  NvSWxzvJb(jqC-2)=True means that jqC-2 has been opened. B(bsS-3)=True means that bsS-3 has been obtained. NvSWxzvJb(jqC-2)=False means that jqC-2 has not been opened. B(bsS-3)=False means that bsS-3 has not been obtained.
Step-0: Initialization. Do nothing. 
Question: NvSWxzvJb(jqC-0)=?...B(bsS-0)=?B(bsS-1)=?...
Answer: NvSWxzvJb(jqC-0)=False...B(bsS-0)=False...
Step-1: Open jqC-3 and retrieve bsS-2.
Question: NvSWxzvJb(jqC-0)=?...
\end{lstlisting}
\caption{Running example of our test environment.}
\end{figure}
\vspace{8pt}

For the experiments below, we set the number of boxes and keys to 10. This corresponds to a total of 20 environment states (boxes and keys have separate states) to be enumerated after each step. We sample the number of steps randomly from a uniform distribution $U(1, 10)$. We then keep instructions almost entirely the same across samples, except that in the Synthetic Language settings (see Section~\ref{sec:robustness}) the state predicates are defined randomly and therefore vary between samples (e.g., in one instance we use \emph{NvSWxzvJb(jqC-0)=True} to refer to the box \emph{jqC-0} having been opened, while in another instance we use \emph{Abc(bb-0)=True} to represent the box \emph{bb-0} having been opened). 

\subsection{Robustness Checks for State Tracking} \label{sec:robustness}
\paragraph{Synthetic language} When testing models' ability to map to logical representations of environment states, a concern with using language-based logical symbols (such as $\blk{OPENED}(\text{BOX-4})$) is that the models may be able to leverage pre-training on similar language-based logical symbols, or simply copy from the input language describing the actions, without needing to convert to abstract situation states. 
To control for this possibility, in addition to using natural language (NL) functors and arguments for our environment states, we also adopt settings in which \textbf{synthetic language} (SL) is used to specify the functors and arguments of the environment states. This allows us to disentangle our target task from pattern memorization and copying capabilities, better ensuring that models must rely on a combination of the instructions and the changes caused by the actions taken.
To build synthetic functors and arguments, we use randomly selected ASCII characters. The length for each functor or argument is a random sample from $U(1, 10)$. In the instructions, we include explanations of the meanings of these synthetic functors and arguments.
An example of our synthetic language setting is shown in \cref{lst: running example} (example instructions from our NL setting can be seen in \cref{lst: contradictory_logic} and~\cref{lst: contradictory_language}). 
To counteract randomness effects and reduce any bias from specific synthetic language, for each test case (an \textbf{instance}), we generate a different set of synthetic functors and arguments. 

\begin{figure*}[tbp!]
\begin{minipage}[b]{0.45\textwidth}
\begin{lstlisting}[caption={CounterIntuitive Output Format}, label={lst: contradictory_logic}]
Instructions: ... OPENED(BOX-3)=@\textcolor{red}{False}@ means that BOX-3 has been opened. OBTAINED(KEY-1)=@\textcolor{red}{False}@ means that KEY-1 has been obtained. OPENED(BOX-3)=@\textcolor{red}{True}@ means that BOX-3 has not been opened. OBTAINED(KEY-1)=@\textcolor{red}{True}@ means that KEY-1 has not been obtained.
\end{lstlisting}
\end{minipage}
\hfill
\begin{minipage}[b]{0.45\textwidth}
\begin{lstlisting}[caption={Counter-Intuitive Language Instruction}, label={lst: contradictory_language}]
Instructions: ... OPENED(BOX-3)=True means that BOX-3 has @\textcolor{red}{Not}@ been opened. OBTAINED(KEY-1)=True means that KEY-1 has @\textcolor{red}{Not}@ been obtained. OPENED(BOX-3)=False means that BOX-3 has @\textcolor{red}{\sout{not}}@ been opened. OBTAINED(KEY-1)=False means that KEY-1 has @\textcolor{red}{\sout{not}}@ been obtained.
\end{lstlisting}
\end{minipage}
\end{figure*}
\paragraph{Counterintuitive instructions} To further control for potential memorization of mappings between language and logical states from pre-training, we include one additional manipulation involving what we call \textbf{counterintuitive instruction}. 
In counterintuitive instruction settings, the instructions define mappings that reverse the standard usage of logical statements and truth values. The two versions we use are \emph{counterintuitive output format} (\cref{lst: contradictory_logic}), and \emph{counterintuitive language instruction} (\cref{lst: contradictory_language}). These manipulations draw on the tradition of negation testing
\citep{mccoy2019right, ribeiro2020beyond, hossain2020s, ravichander-et-al-2022-condaqa}, and allow us to further disentangle pre-training memorization from understanding of our particular instructions.\footnote{There are other possible perturbations that we can do, such as adding a distractor at each step (\cref{app:action_distractor}) or applying synthetic language only on functors / only on arguments
(\cref{app:partial_usage_sl}). For simplicity, we omit the discussion of other perturbation types and only focus on the two robustness checks explained in this section.}

\subsection{Evaluation Metrics} 
\label{sec:metrics}
We mainly evaluate models' success in our synthetic environment by measuring two metrics: \textbf{State-EM}, and \textbf{Step-EM}.

\textbf{State-EM} is the proportion of all predicted \emph{states} that match the expected states. This metric is useful to check to what extent the model develops correct \textbf{partial understanding} in response to situational changes. 
\begin{equation}
    \text{State-EM} = \frac{\#(\text{Matched States})}{\#(\text{Queried States})}
\end{equation}
\textbf{Step-EM} is a stricter metric than State-EM. 
 It is the proportion of \emph{steps} for which the full set of predicted states at that step have an exact match with the expected states, including all the truth values and the number of predicted states.
This allows us to check whether 
the model can maintain  \textbf{consistent} and \textbf{coherent} understanding over situational changes. 
This metric is also important given the sparsity of updates at each step, to ensure that models cannot achieve strong performance simply by copying previous states.
\begin{equation}
    \text{Step-EM} = \begin{cases}
    1 & \substack{\text{if Matched States} = \text{Ground Truth States} \\ = \text{Predicted States} }\\ 
    0 & \text{otherwise} \\
    \end{cases}
\end{equation}

We illustrate the computation of these two metrics for the following simplified case ($2$ boxes and keys, synthetic language, no counterintuitive instructions):
  \begin{lstlisting}[caption={Example for Metrics Computation}, label={lst: metrics_comp_illustration}]
[Instructions and some previous steps]
Question: NvSWxzvJb(jqC-0)=? B(bsS-0)=? NvSWxzvJb(jqC-1)=? B(bsS-1)=?
Correct answer: NvSWxzvJb(jqC-0)=True, B(bsS-0)=False, NvSWxzvJb(jqC-1)=False, B(bsS-1)=False
Model Answer: NvSWxzvJb(jqC-0)=True, B(bsS-0)=True, NvSWxzvJb(jqC-1)=False, B(bsS-1)=False
\end{lstlisting}

State-EM in this case is $3/4=75\%$ (only $3$ out of $4$ states are correct), 
while Step-EM is $0$ because the step contains a incorrect state.

When computing these metrics, we find that at times the generated output does not strictly follow the given format from the in-context samples. Therefore, we use regular expressions to extract the truth values and corresponding states from model outputs. Details can be found in \cref{app:post-processing-regex}. 

\paragraph{Existence of Shortcuts 
}
It is clear that simple string automata plus a status tracking table should already be sufficient to solve every instance of our tasks. However, the simplicity of the task is part of its value: if models have a basic capacity to track and maintain environment states, this task should be straightforward. Nonetheless, we will see in the experiments below that ChatGPT still struggles to solve these tasks reliably, despite the existence of such simple solutions, indicating the presence of fundamental limitations in this class of capability.

\subsection{Discussion: Synthetic Environment as Simulation for Real-World Application}
\label{sec: real-world-implication}
At its core, our synthetic environment is a simplified simulation for real-world state-tracking tasks (usually in the form of slot-filling), a critical capability of dialogue systems / chatbots~\citep{williams2014dialog, henderson2014second, wu2019transferable}. By prompting the model to update states, we are gradually giving the model more contextual information and testing whether ChatGPT, the state-of-the-art chatbot model, can closely follow users' prompts and keep track of the full interaction history.  

Our work has important potential implications as the usage of LLMs continues to proliferate.
Instructing LLMs to remember many initial states, operate over
synthetic languages, and keep track of interaction history can be seen as an important step in eventually  teaching a highly-capable agent to follow social norms and policies.  Our initial set of environment states is similar to establishing basic rules about dos-and-don'ts at the beginning of human-AI conversations. The usage of synthetic languages is likewise similar to teaching AI agents about specific tones or styles of communication, terminologies, jargon, or perhaps even low-resource languages. Analyzing whether the model can keep track of environment states can allow us to draw conclusions about the model's ability to follow instructions or rules. 
In this sense, our work also has implications with respect to recent trends of Constitutional AI or Rule-based/Principle-driven models~\citep{bai2022constitutional, openai2023gpt4tr, sun2023principle}, in which human social rules (``constitution'') are explicitly written at the beginning of a prompt to align AI agents' behavior to human values. 
Our initial environment states are like constitution policies that AI agents should obey. 
The steps and queries in our environments are reminiscent of a situation in which certain policies can be allowed to be updated with user permission. For example, initially an AI agent may be programmed to try its best to answer every question and disallow overly conservative responses like refuse-to-answer---but under certain situations, the user could update with permission to the agent to refuse to answer for privacy, fairness or other social reasons. 

As we will see in the experiments below, when more interactions occur, the model will gradually lose track of states, propagate errors, and even generate hallucinations, despite all updates falling within the input window. By design, a super-capable AI agent like ChatGPT should have the ability to read and use all information within the input window---but our results suggest that this is not the case. Our research thus calls for further study, and for caution when implementing a Constitutional AI approach.

%% file: past_papers/sittest/sections/experiments.tex
\section{Benchmarking Model Sensitivity to Instructions and Situation Changes
}
\label{sec:instruction_understanding}
We use our environment to test ChatGPT in its ability to track situation changes, and we report the results below. We also compare against the performance of additional chat models, which we find to underperform ChatGPT. Those results can be found in \cref{sec:external_comparison}.

We try 2-shot, 3-shot and 5-shot settings (in which 2, 3 and 5 example steps with enumerated states are provided). We use 50 samples with the number of additional steps randomly sampled from $\{1, \dots, 10-\#(\text{num-shots})\}$ for each setting. We follow the official OpenAI cookbook\footnote{\url{https://github.com/openai/openai-cookbook/tree/main}} to prepare the request and parse the response when interacting with ChatGPT API. Details are in \cref{app:interaction_prompt_format}. 

    \begin{table*}[h!]
\centering
\small
\resizebox{0.9\textwidth}{!}{%
\begin{tabular}{@{}lrrr@{}}
\toprule
\multirow{2}{*}{Normal Instruction} & \multicolumn{3}{c}{Step-EM / State-EM} \\ \cmidrule(l){2-4} 
                      & \multicolumn{1}{c}{2-shot}     & \multicolumn{1}{c}{3-shot}     & \multicolumn{1}{c}{5-shot}     \\ \midrule
NL Functor + NL Argument     &   $22\% / 92\%$   &   $34\% / 93\%$   &   $36\% / 95\%$    \\
SL Functor + SL Argument    &   $19\% / 92\%$   &   $22\% / 93\%$   &   $52\% / 96\%$    \\
\midrule
\multirow{1}{*}{Counter-Intuitive Instruction (On NL)} & \multicolumn{3}{c}{} \\ 
    \midrule
NL Functor + NL Argument    &   $10\% / 77\%$   &   $6\% / 75\%$   &   $0\% / 84\%$    \\
SL Functor + SL Argument    &   $13\% / 89\%$   &   $20\% / 90\%$   &   $12\% / 90\%$    \\ \midrule
\multirow{1}{*}{Counter-Intuitive Instruction (Truth Values Switching)} & \multicolumn{3}{c}{} \\ \midrule
NL Functor + NL Argument    &   $6\% / 72\%$   &   $2\% / 69\%$   &   $2\% / 79\%$    \\
SL Functor + SL Argument    &   $19\% / 85\%$   &   $14\% / 87\%$   &   $12\% / 89\%$    \\
\bottomrule
\end{tabular}%
}
\caption{Experiment results on ChatGPT  Robustness check for state tracking in 10-box environment. Metrics here are presented in the format of ``Step-EM / State-EM''. We use 50 samples for each experiment setting. 
}
\label{tab:10-box-instruction-testing-chatgpt}
\end{table*}

Results for these tests on ChatGPT are shown in \cref{tab:10-box-instruction-testing-chatgpt}. We can see a number of patterns.

\paragraph{Failures on Step-EM}
Under normal instructions, though the model achieves high State-EM (i.e., ~90\%), the Step-EM is generally much lower (up to  $70\%$ lower in 2-shot and $40-50\%$ lower in 5-shot). This indicates that although models are able to take advantage of the sparsity of the updates to get a majority of states correct, they are much less successful in accurately characterizing the entire environment state at a given step. As we will see in \cref{sec:fine-grained-analysis} State-EM may also be skewed by accidentally-correct states, and should in general be interpreted with some caution.

\paragraph{ChatGPT has limited capability to understand and follow counterintuitive instruction.}
From \cref{tab:10-box-instruction-testing-chatgpt}, 
we see that applying our counterintuitive instruction manipulation leads to significantly degraded performance. Especially on Step-EM, performance in most settings is $10-20\%$, which is much worse than with normal instructions ($>20\%$, or even $>30\%$ in most cases).  This suggests that the model is indeed to some extent relying on standard language-to-logic mapping 
from pre-training, rather than fully understanding the instructions. 

Despite this inconsistency, the model still shows some success with the negation or the new rule with flipped truth values, as it still manages to achieve $70-80\%$ State-EM. Though the State-EM values should be taken with caution, these accuracies are substantially stronger than would be expected from random guessing ($50\%$) or completely ignoring the rule ($0\%$), suggesting some level of capability in making use of the counterintuitive instructions.

\paragraph{Effects of synthetic language} 
Use of synthetic language affects model performance, but not always in the predicted directions. Though performance is occasionally worse with synthetic language, it is more often \emph{better} than with natural language. This suggests that the use of synthetic language may at times help the models to detach from unhelpful biases from pre-training, and rely more robustly on the in-context information. This follows the analysis presented in \citet{liang2022holistic}.

\paragraph{More in-context samples do not necessarily help.} We also see that even if we are already providing answers for approximately $50\%$ of states (considering that we only have $10$ boxes and $10$ keys in the environment, and each step will change exactly two states permanently within the dialog), the model does not make improvement in most cases.

%% file: past_papers/sittest/sections/analysis.tex
\section{Analysis of Model Performance
} 
\label{sec:logical_consistency}
In the previous section we tested the capacity of ChatGPT in tracking and enumerating states within our environment, and we found that the model showed clear limitations. In this section, we analyze model performance further, to better understand the source of these limitations.
    \begin{figure*}[hpt!]
    \centering
    \begin{subfigure}[b]{0.495\textwidth}
    \centering
    \includegraphics[width=\columnwidth]{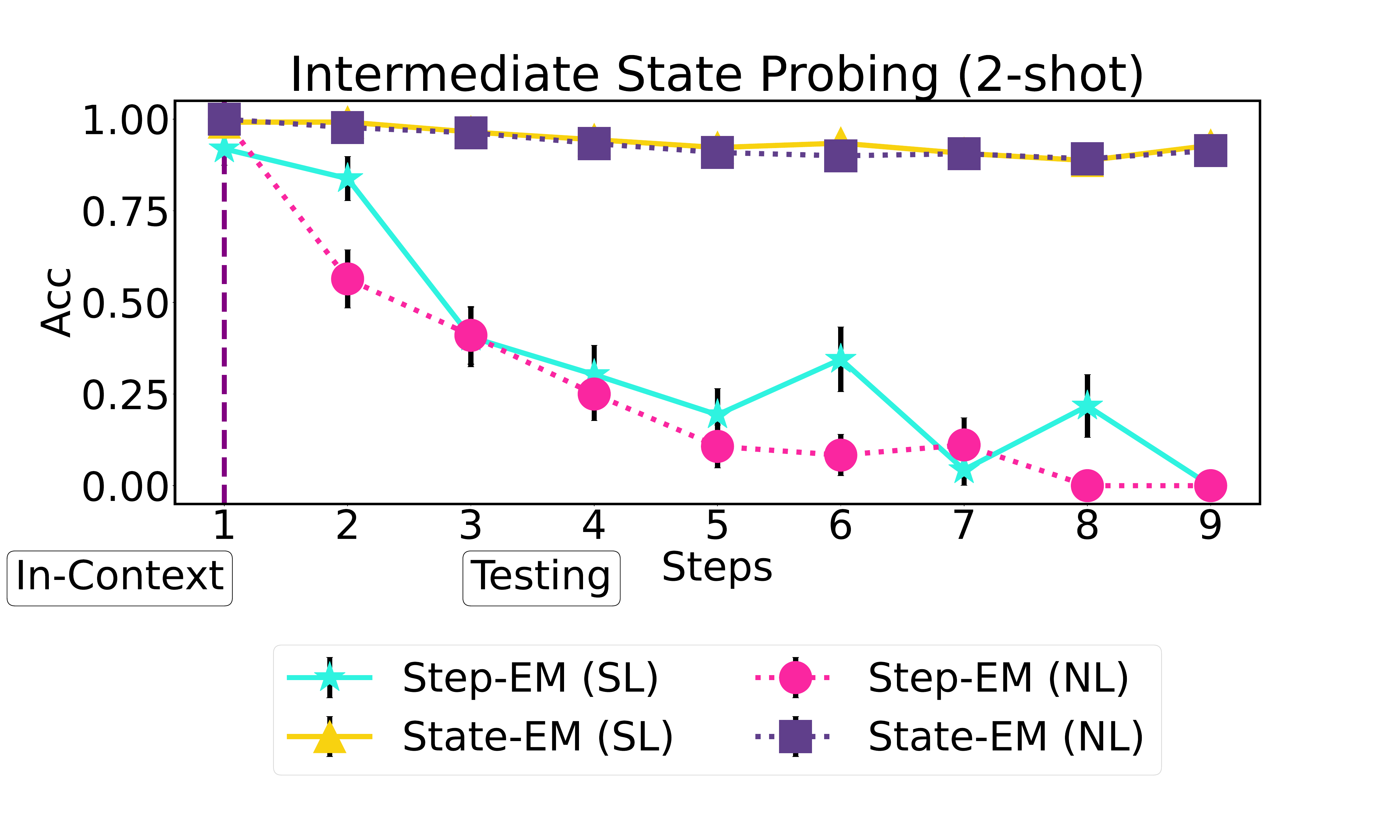}
    \vspace{-10mm}
    \caption{2-shot}
    \label{fig:2shot-id}
\end{subfigure}
    \begin{subfigure}[b]{0.495\textwidth}
    \centering
    \includegraphics[width=\columnwidth]{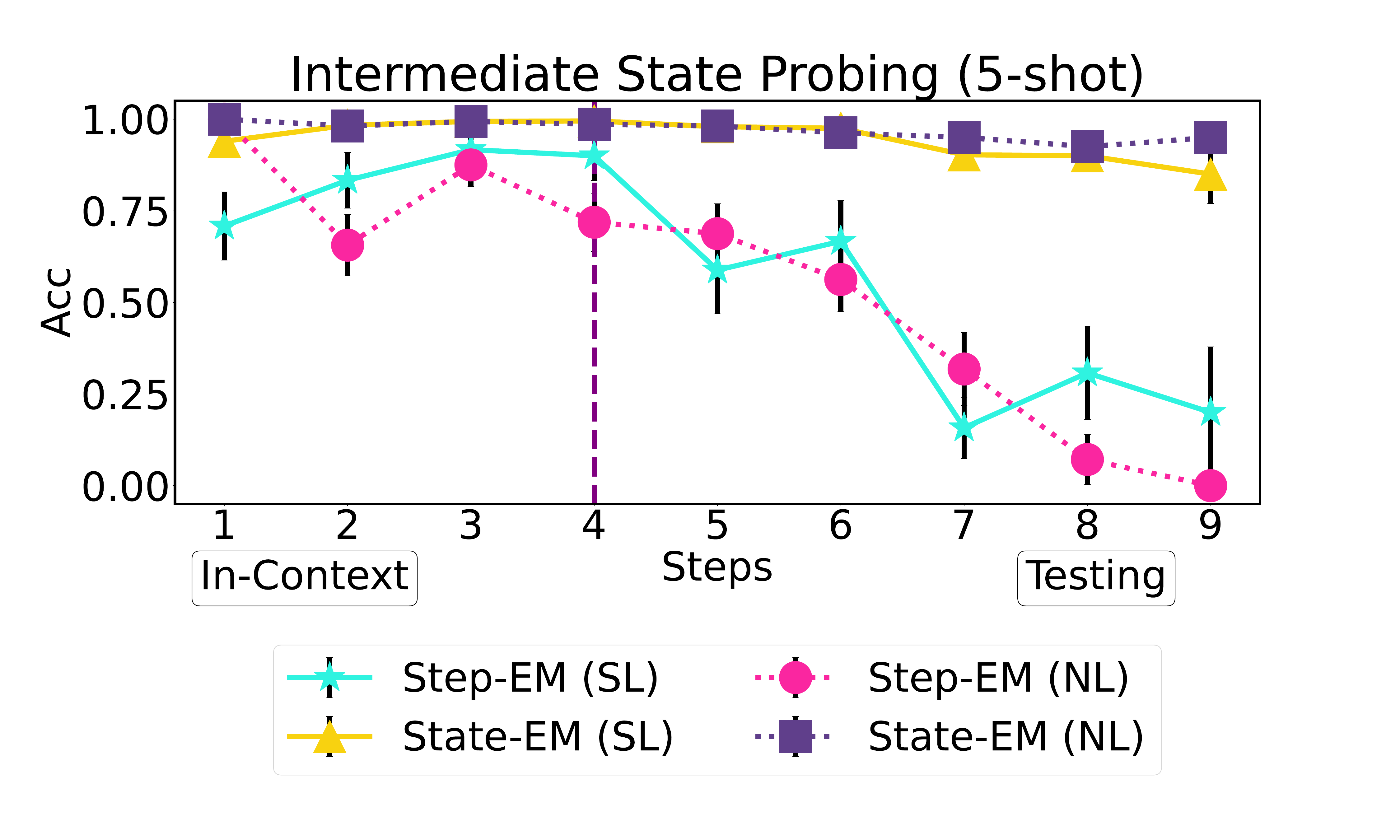}
    \vspace{-10mm}
    \caption{5-shot}
    \label{fig:5shot-id}
\end{subfigure}

\caption{Results for Intermediate State Probing.
Purple vertical lines indicate where in-context demonstrations (for steps prior to the test step) end. 
}
\label{fig: interactive_debugging}
\end{figure*}

\subsection{Tracing Errors: State Tracking over Steps} 
To understand why ChatGPT shows relatively poor performance, a straightforward way is 
reusing the instances created in \cref{sec:instruction_understanding}, but querying for the environment states after each intermediate step to see where the errors emerge. Specifically, rather than only querying after the final step,
we make queries at all steps (excluding ``Step-0''), including those within the in-context example window (for querying each step $s$ in the in-context example window, in-context demonstrations of environment states are given only through step $s-1$).
We evaluate State-EM and Step-EM at every step. We refer to this test as \textbf{Intermediate State Probing}. 

\paragraph{Potential confounder: state complexity}
When interpreting performance trajectory across increasing numbers of steps, a potential confounder is that performance may degrade simply because the set of environment states has become more complex, and not because there are too many steps of updates involved. 
To investigate this possibility, we also run a test in which we compress and skip $k$ of the early steps, and initialize in the state that would have resulted from those steps. We then test the trajectory of model performance on the subsequent $n$ steps. If performance after $n$ steps in this setting is comparable to performance after $k+n$ steps in the previous setting, this will suggest that the degradation is indeed due to the complexity of the environment states. If, however, performance after $n$ steps in this setting is substantially better than performance at $k+n$ steps in the prior analysis, this suggests that the degradation in performance is due to failure to handle the growing number of steps. 

\begin{table}[h!]
\centering
\resizebox{0.55\textwidth}{!}{%
\begin{tabular}{@{}lcc@{}}
\toprule
\multirow{2}{*}{Normal Initialization} & \multicolumn{2}{c}{Step-EM / State-EM} \\ \cmidrule(l){2-3} 
                      & 2-shot     & 5-shot     \\ \midrule
NL Functor + NL Argument   &   $22\%$ / $92\%$     &   $36\%$ / $95\%$    \\
SL Functor + SL Argument    &   $19\%$ / $92\%$     &   $52\%$ / $96\%$  \\
\midrule
\multirow{1}{*}{Compressed Initialization} & \multicolumn{2}{c}{} \\ \midrule
NL Functor + NL Argument     &   $46\%$ / $95\%$     &   $60\%$ / $97\%$    \\
SL Functor + SL Argument    &   $48\%$ / $95\%$     &   $54\%$ / $97\%$  \\
\bottomrule
\end{tabular}%
}
\caption{Compressed Initialization Testing experiment results for 10-box environment on ChatGPT. Metrics here are shown in the format of "Step-EM/State-EM". We use 50 samples with various number of steps for experiments. 
}
\label{tab:10-box-random-initialization-testing-chatgpt}
\end{table}

\begin{figure*}[!h]
   \centering
\resizebox{0.83\textwidth}{!}{

    \includegraphics[width=0.8\textwidth]{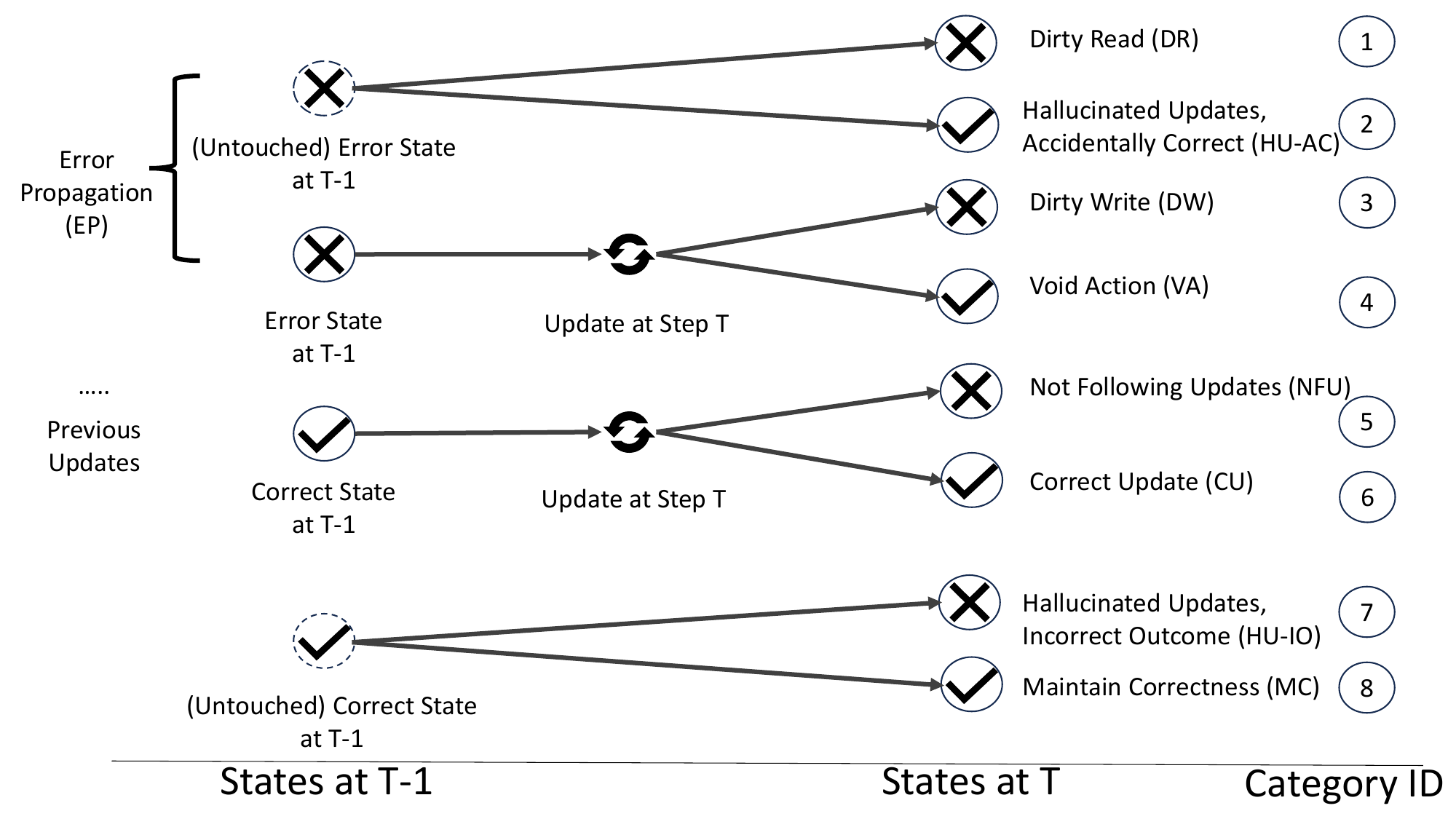}}
    \caption{Diagram for fine-grained error analysis 
    }
    \label{fig:fine-grained-error-analysis}
\end{figure*}

The experiment results for the Intermediate State Probing and Compressed Initialization Test are shown in \cref{fig: interactive_debugging} and \cref{tab:10-box-random-initialization-testing-chatgpt}, respectively. From these results we make the following observations: 

\shortparagraph{Degraded performance over steps.} 
We see in \cref{fig: interactive_debugging} that with increasing number of situational changes, both State-EM and Step-EM degrade. This degradation is particularly true for Step-EM, which decreases dramatically as steps increase.

\shortparagraph{State complexity does not explain the degradation.} Additionally, we see in \cref{tab:10-box-random-initialization-testing-chatgpt} that skipping steps and starting with more complex initialization leads to improved performance, indicating that the degradation across steps is not attributable to state complexity alone. 

\begin{figure*}[htbp!]
    \centering
    \captionsetup[subfigure]{justification=raggedright, singlelinecheck=false}
    \begin{subfigure}[t]{0.49\textwidth}
    \includegraphics[width=\columnwidth]{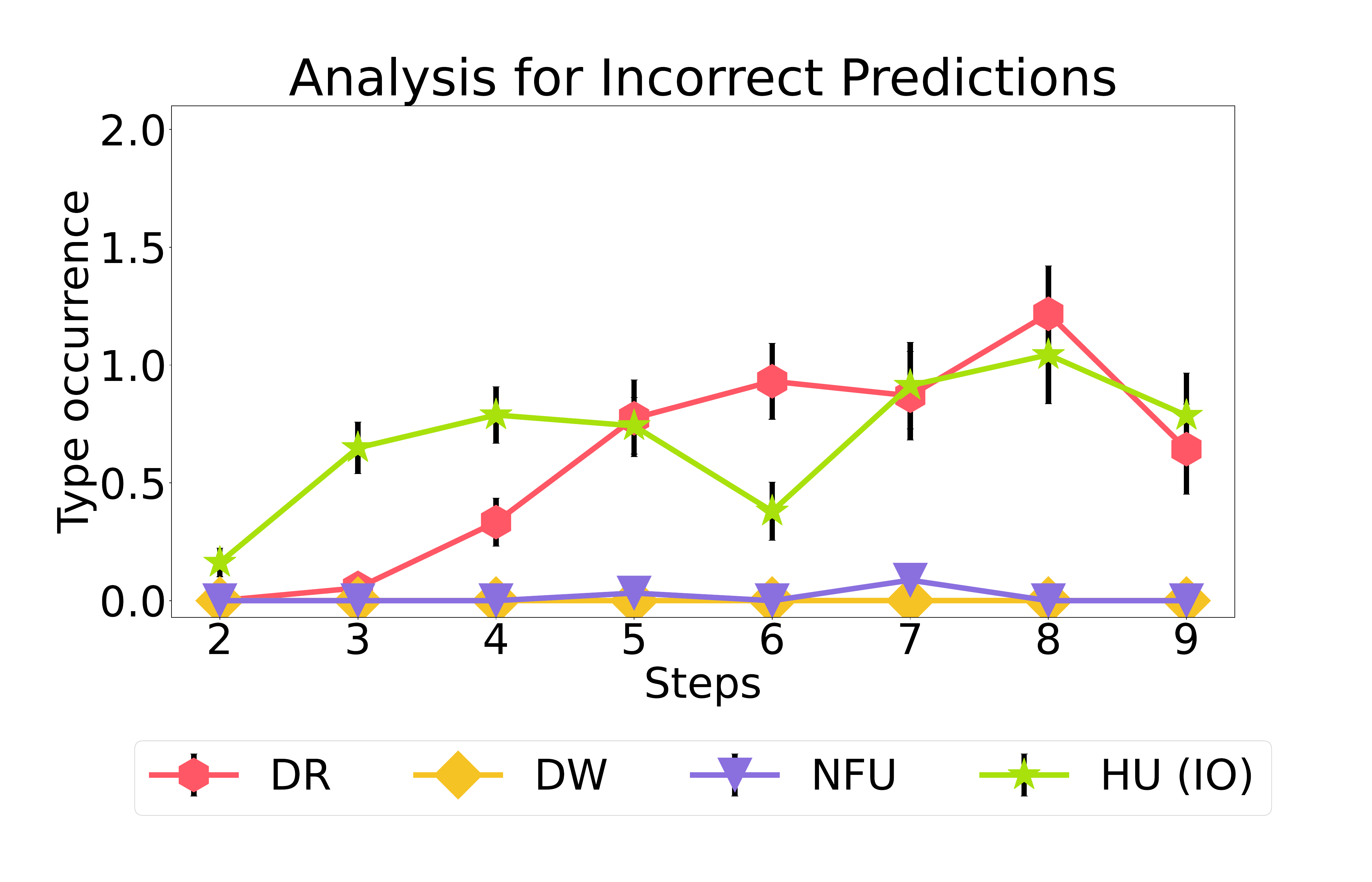}
    \vspace{-10mm}
    \caption{Analysis for Incorrect outcome, 2shot, SL}
    \label{fig:2shot-irreg-func-irreg-arg-ea-ic}
    \end{subfigure}
    \hfill
    \begin{subfigure}[t]{0.49\textwidth}
    \includegraphics[width=\columnwidth]{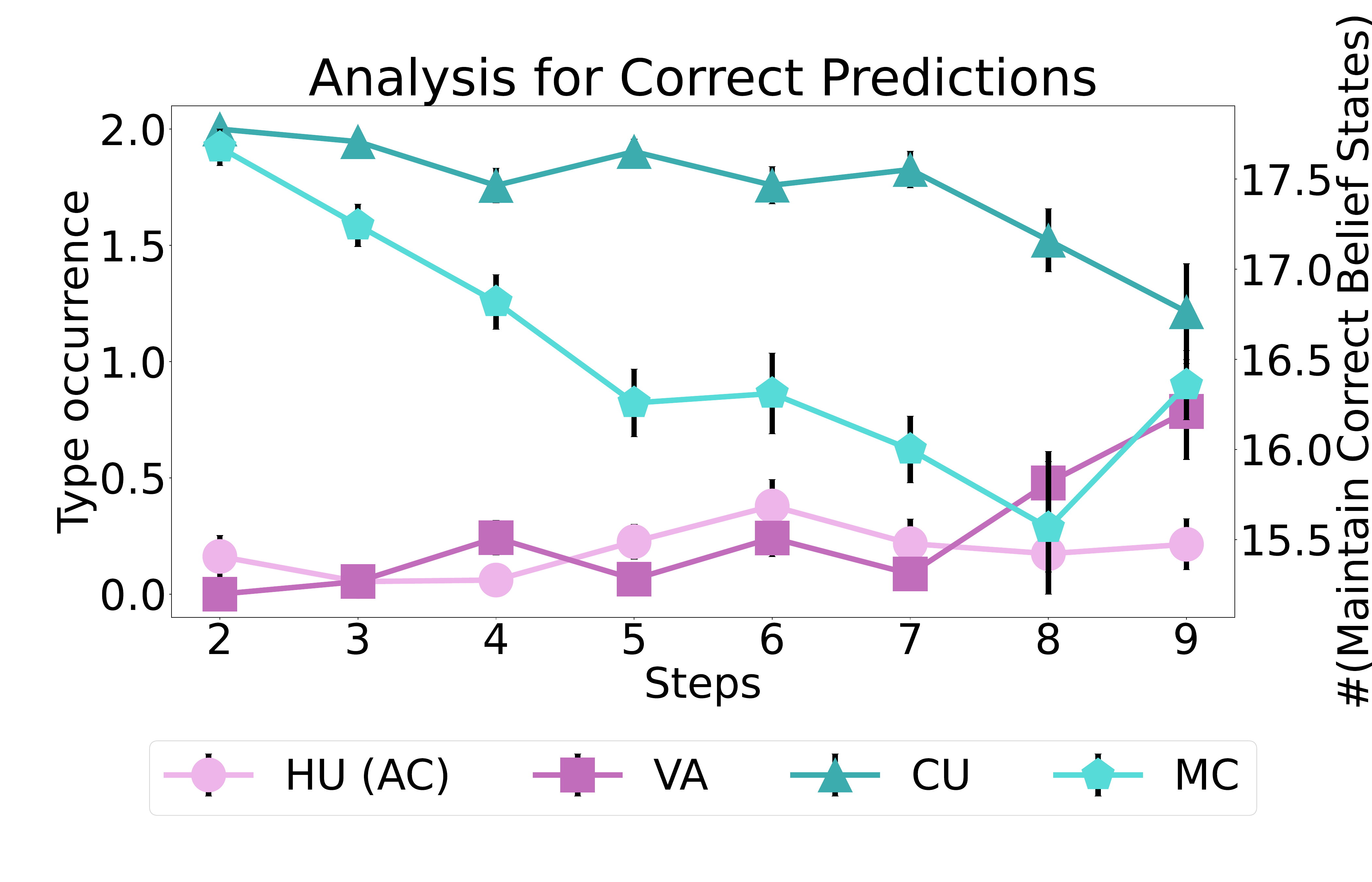}
    \vspace{-10mm}
    \caption{Analysis for Correct outcome, 2shot, SL}
    \label{fig:2shot-irreg-func-irreg-arg-ea-co}
    \end{subfigure}

    \vspace{1em}
    \begin{subfigure}[t]{0.49\textwidth}
    \includegraphics[width=\columnwidth]{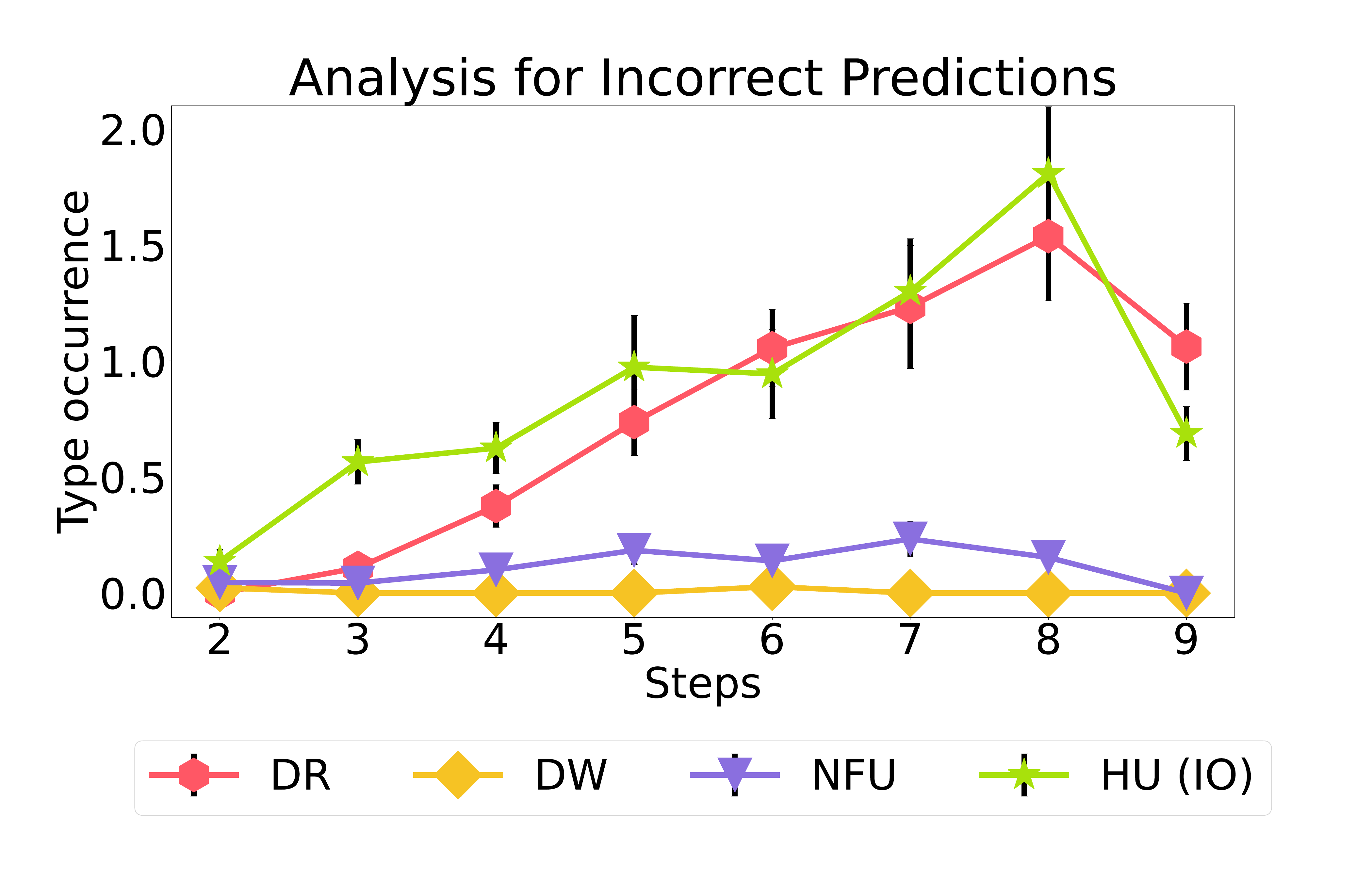}
    \vspace{-10mm}
    \caption{Analysis for Incorrect outcome, 2shot, SL\\ + Counter-Intuitive Instruction (On NL)}
    \label{fig:2shot-irreg-func-irreg-arg-ea-ic-cf}
    \end{subfigure}
    \hfill
    \begin{subfigure}[t]{0.49\textwidth}
    \includegraphics[width=\columnwidth]{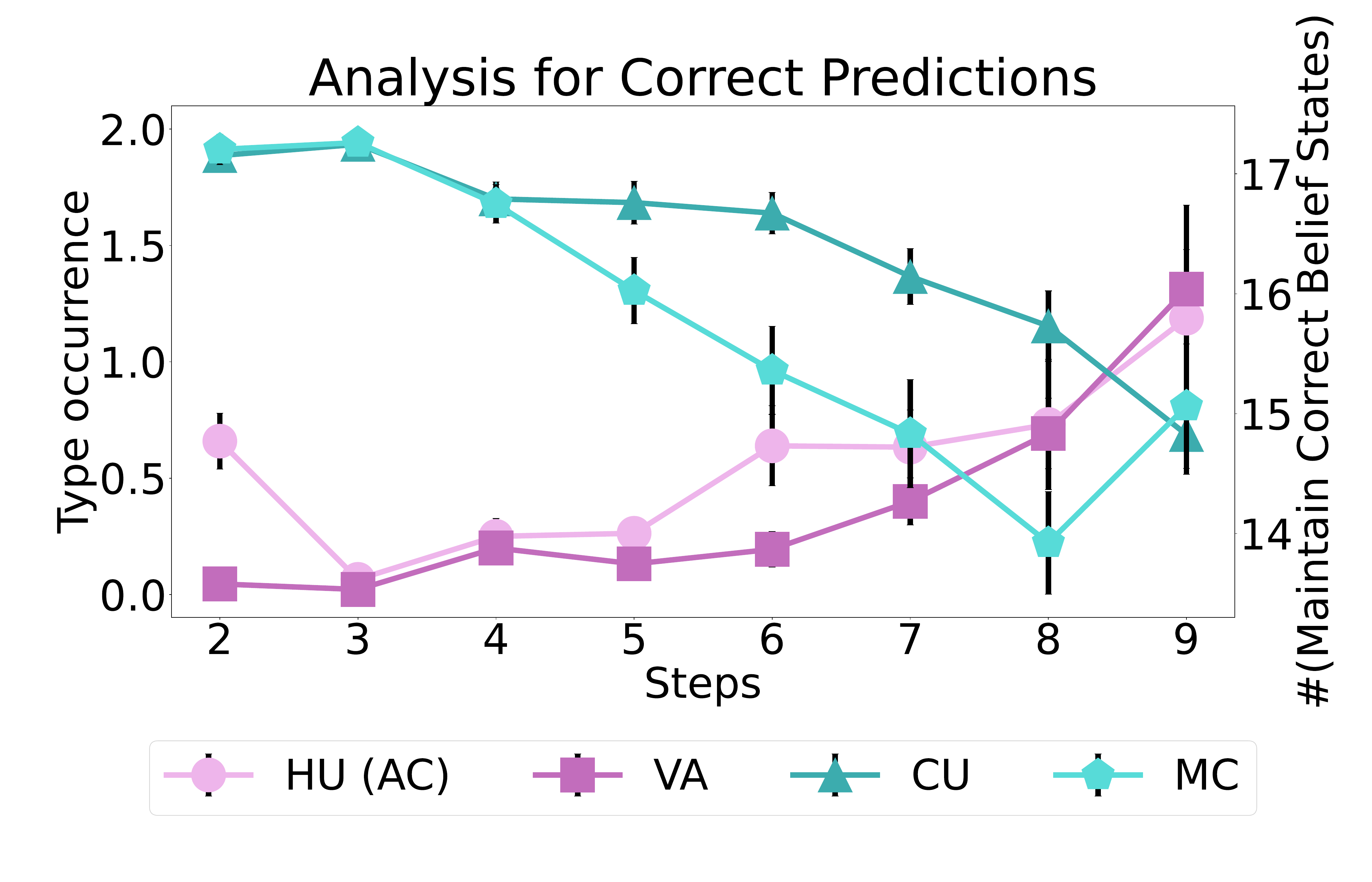}
    \vspace{-10mm}
    \caption{Analysis for Correct outcome, 2shot, SL\\ +  Counter-Intuitive Instruction (On NL)}
    \label{fig:2shot-irreg-func-irreg-arg-ea-co-cf}
    \end{subfigure}
    \caption{Fine-Grained Error Analysis for Logical Inconsistency 
    }
    \label{fig:interactive_debugging_error_analysis-finegrained}
\end{figure*}

\shortparagraph{More in-context demonstration mitigates degradation.} 
In \cref{sec:instruction_understanding} we found that providing more in-context samples does not cause a significant improvement in the performance of the model at the final Test Query. With intermediate probing we gain a finer-grained picture of the impacts of number of in-context samples across steps. First, we see that when prompting with more in-context samples (5-shot vs 2-shot), there is less rapid degradation in model accuracy after the end of the in-context window. In the 2-shot (\cref{fig:2shot-id}) case, at Step-3 (the second step preceded by a non-demonstration step), Step-EM quickly drops from $~60\%$ to $~40\%$. By contrast, in 5-shot (\cref{fig:5shot-id}), at Step-6, ChatGPT can still maintain a Step-EM value of $60\%$. This suggests that having more in-context samples does strengthen models' accuracy in tracking state changes---but only temporarily or over a limited number of steps.

Though we see that having more demonstrations can mitigate degradation after the demonstrations end, when we look within the in-context sample window itself, we see that on steps that directly follow in-context demonstrations (Step-1 for 2-shot, Step-1,2,3,4 for 5-shot), the model's performance does not monotonically increase in response to the accumulating demonstrations. A similar phenomenon is also discovered in other few-shot reasoning benchmark works~\citep{suzgun2022challenging, liang2022holistic}, 
despite the fact that in traditional fine-tuning, usually, more training instances yield better generalization ability. This suggests that although adding more demonstrations can briefly mitigate loss of accuracy, it does not straightforwardly translate to gains in accuracy.

\paragraph{Interim discussion} These patterns of degradation over time occur, in both NL and SL settings, despite the fact that ChatGPT can read the full dialogue in its input window. This suggests that ChatGPT cannot effectively utilize the full information in its input window, and that claims about maximum input length capabilities (e.g., ChatGPT can model 4k tokens as introduced in the official announcement~\citep{openai2022chatgpt}) 
should be taken with a grain of salt.  

\subsection{Fine-grained Analysis of Update Patterns}
\label{sec:fine-grained-analysis}
In the above section, we studied the trajectory of model performance as the number of steps increases, finding evidence that ChatGPT degrades in state tracking with increased number of steps. In this section, we do a finer-grained analysis of the update dynamics in these experiments, in order to examine more closely the causal factors leading to both erroneous and accurate predictions. For the purpose of this analysis, we define categories of state transitions, summarized in \cref{fig:fine-grained-error-analysis}. These categories allow us to analyze the relationships between states at the analyzed step and the corresponding prior state, both with updates should be made to those states and when they should not.

The experiment results are shown in \cref{fig:interactive_debugging_error_analysis-finegrained} (we only show 2-shot SL, for reasons of space). 

Examining first the patterns for states in which models make incorrect predictions, we see that the rise in errors is driven by states that should be untouched at that step. We see that as steps increase there are rapid increases in both Dirty Read (DR) transitions, where models retain a previous error, and Hallucinated Update (HU-IO) transitions, where models change a state from correct to incorrect despite there being no change to that state in the step description. These patterns indicate that the rise in errors over time can be attributed both to retention and propagation of errors from previous steps, but also to \emph{failures} in retaining the memory of a previous step that should not change.

Examining now the transitions associated with correct model predictions, we see that over time there is noteworthy decrease in Correct Update (CU) cases---however, there is a much more dramatic decrease in Maintain Correctness (MC) cases, indicating that the model increasingly fails to retain the memory of previously correct states. Over time we also see, particularly in the case of counterintuitive instructions, a rise in Accidentally Correct (HU-AC) cases, in which the model switches from an incorrect state back to the correct state, despite the fact that no update to that state was described in the step. Both of these patterns are indicative of memory limitations and susceptibility to random noise in changes to enumerated states.

These results yield several conclusions:

\shortparagraph{ChatGPT has non-persistent in-context memory.} 
A recurring observation above is that many of the model errors that increase over time can be attributed to limitations in retaining states in memory---and in fact, some states marked as correct also reflect accidental correctness arising due to similar failures to retain prior states.

\shortparagraph{States can also be retained, but potentially by chance.}
In addition to memory retention failures, we also see propagation of errors between steps---which in theory is indicative of successful memory retention, by contrast to the retention failures cited above. However, considering the prevalence of hallucinated updates, and the limited options for state values, we can expect that at least some of these retained updates in fact occur by chance.

\shortparagraph{Couterintuitive instruction exacerbates non-robust behavior.} As we mentioned above, the drop in correct CU updates is more dramatic---and the rise in spurious correct updates HU more substantial---in the case of counterintuitive instructions. This suggests that the inability of the model to rely on memorized language-to-logic mappings generally reduces the model's ability to execute and maintain correct state updates.

%% file: past_papers/sittest/sections/conclusion.tex
\section{Discussion}
In this paper, we propose a novel synthetic testing environment for testing situational understanding capabilities, which we apply to test ChatGPT, the state-of-the-art chatbot. We instruct ChatGPT to process a series of sparse environment updates across time in a dialogue history. With 
careful environmental designs, we reduce the possibility of data contamination and other artifacts typically introduced by traditional probing methods. We find that despite the simplicity of our task, and even with ChatGPT having full access to the complete dialogue history within the input window, the model fails to retain coherent and correct environment states over time. Further analysis suggests that this failure is largely because ChatGPT does not have persistent in-context memory, and is susceptible to hallucinated updates. These findings indicate overall that ChatGPT does not have robust situational state tracking ability. 

Our proposed synthetic environment and the findings that it generates can have noteworthy real-world implications. First, it can diagnose the potential limitations of current chatbot systems in multi-round interactions. Second, our findings also reflect a potential problem for the model's ability to follow instructions and remain consistent with rules/norms established in its context, which is especially important for responsible AI safety and human-AI alignment research.

%% file: past_papers/sittest/sections/limitations.tex
\section*{Limitations}
In this work, we propose a controlled synthetic environment to investigate ChatGPT's situational understanding ability. While we believe our synthetic environment has important real-world implications, as we discussed in \cref{sec: real-world-implication}, for certain real-world applications our findings may not apply. Another limitation is that we only focus on the evaluation of ChatGPT as the state-of-the-art chatbot model (at least in terms of mainstream media coverage). There are other commercial chatbot models that could show stronger performance on our tasks, as they may have more complicated system designs (e.g., multi-module systems as in BlenderBot 3~\citep{shuster2022blenderbot}) that could be better at dealing with multi-round dialogue history and extremely long inputs. As we do not have sufficient budget or open access to test many such systems, we leave a comprehensive benchmark evaluation of situational understanding ability for Chat-Tuned LLMs for future work. Our experiments closely follow the OpenAI official cookbook for interacting with ChatGPT, but it is possible that there could be more optimal prompts to fully unlock the capability of ChatGPT. 

There are many other synthetic environments like TextWorld~\citep{cote2019textworld} that may be programmed to do situational testing as in our work (though it may not be easy to assert full controls), and it is would be interesting to establish whether in different environments we can still draw the same conclusions. Our work mainly focuses on our proposed environment as a case study, but we plan to extend our testing framework to other environments in the future.

%% file: chapters/recode.tex
\section*{Chapter Overview}
Code generation models have achieved remarkable performance on various benchmarks. However, the robustness of these models remains underexplored. In this chapter, we present \texttt{ReCode}, a comprehensive robustness evaluation framework that systematically assesses model performance under semantic-preserving perturbations.

\graphicspath{{./}}

\input{past_papers/recode/introduction}

\input{past_papers/recode/background}

\input{past_papers/recode/methodology}

\input{past_papers/recode/evaluation}

\input{past_papers/recode/conclusion}

%% file: past_papers/recode/introduction.tex
\section{Introduction}
\label{sec:introduction}

Code generation has emerged as an important AI application.
Multiple models~\citep{CodeGen, incoder, gpt-j} have been proposed and achieved impressive performance on generating code using a natural-language description, on completing partial lines and functions, and even on solving complex coding-contest problems. They can offer real-life help to software engineers and enhance their productivity, and multiple commercial offerings exist today for AI-powered code generation~\citep{chen2021evaluating}.

\begin{figure*}[htbp!]
    \centering
    \vspace{-5pt}
    \begin{lstlisting}[language={Python}]
    \end{lstlisting}
    \begin{subfigure}[b]{0.45\textwidth}
        \includegraphics[width=\linewidth]{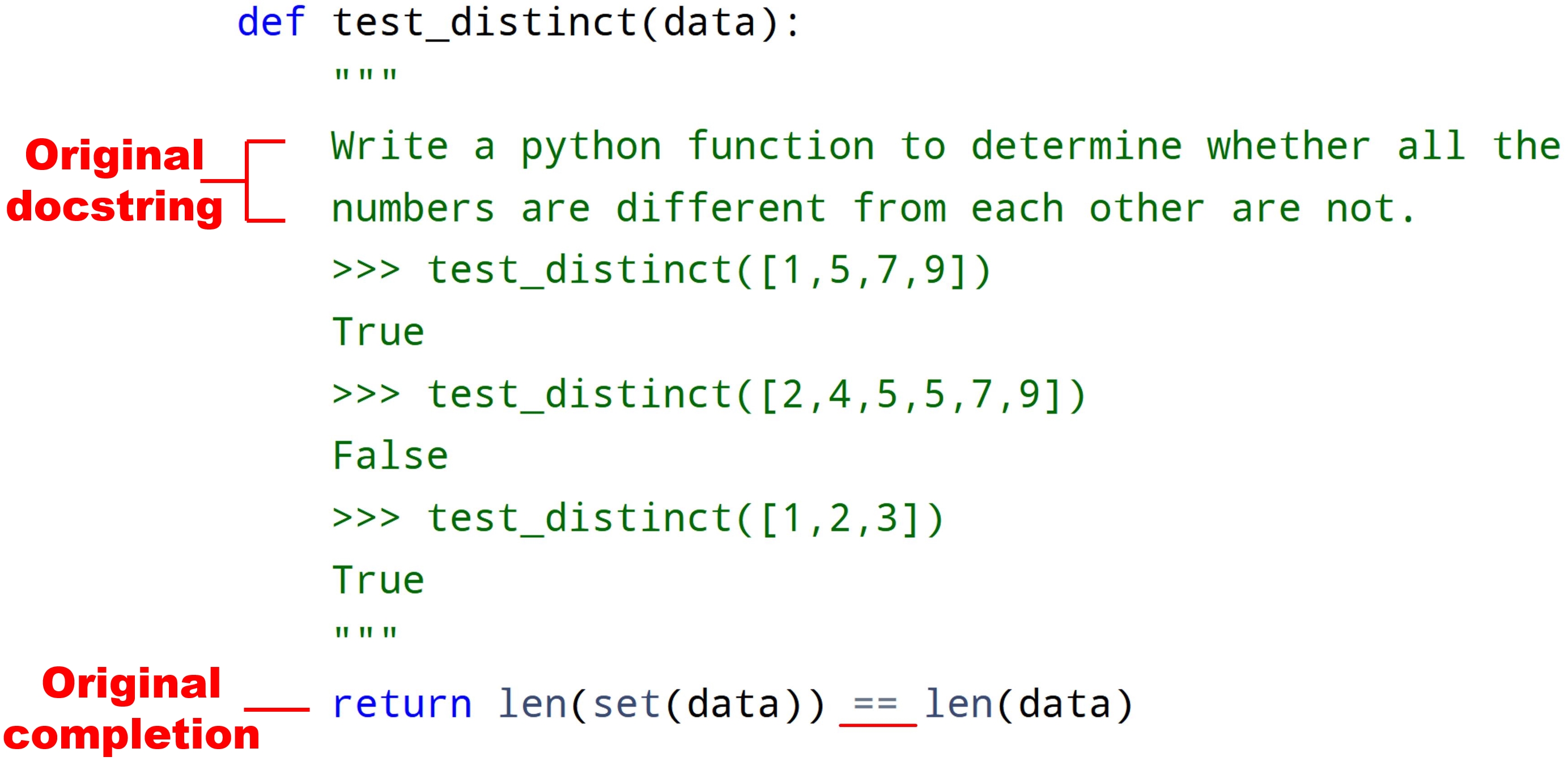}
    \end{subfigure}
    \hfill
    \begin{subfigure}[b]{0.39\textwidth}
        \includegraphics[width=\linewidth]{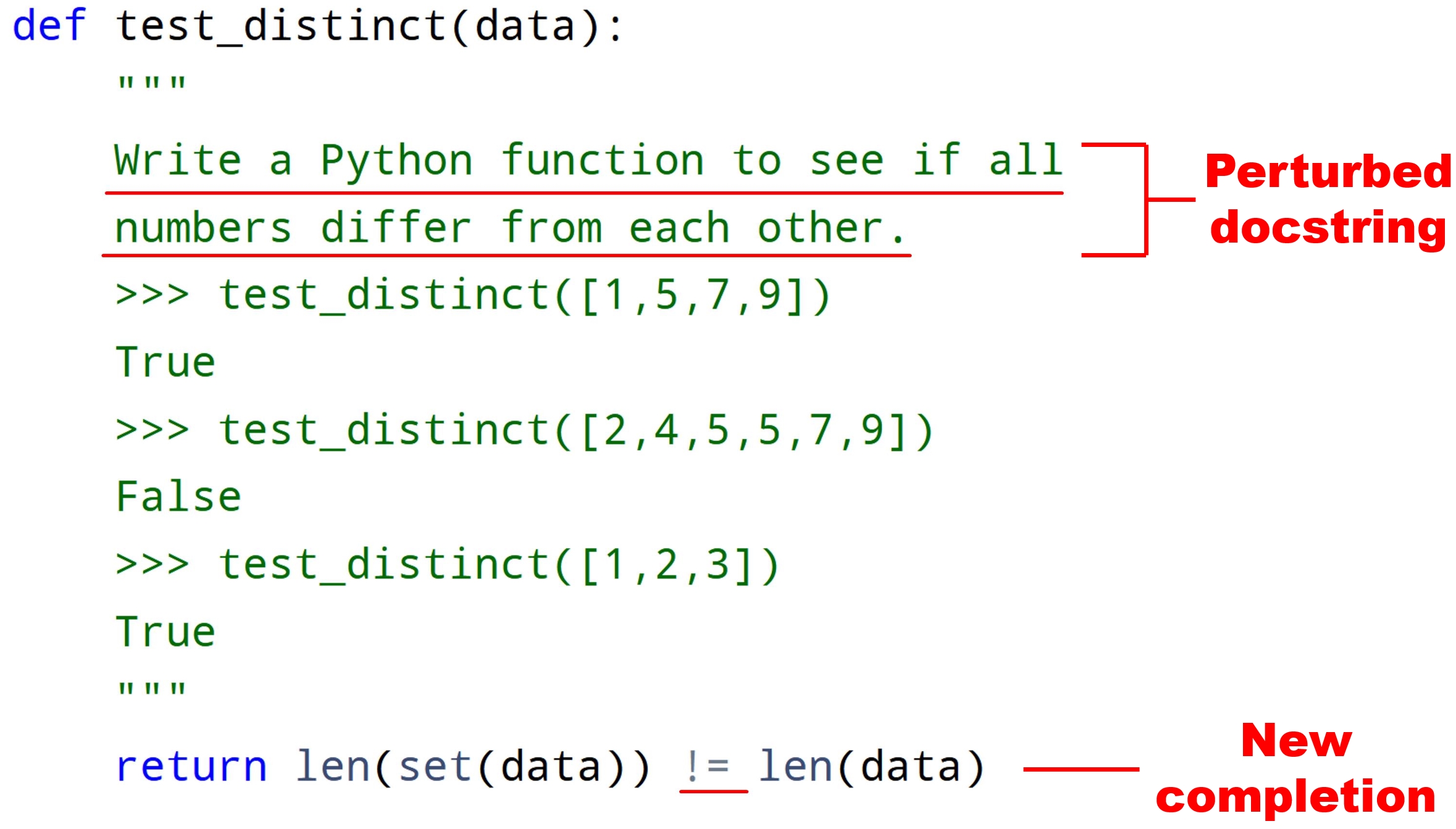}
    \end{subfigure}
        \caption{InCoder-6B predicts correctly on nominal prompt (left) but fails on the prompt where docstrings are paraphrasing with BackTranslation (right). We underline the perturbed positions and wrong model completions. }
    \label{fig: motivating_examples1}
    \vspace{-5pt}
\end{figure*}
\begin{figure*}[htbp!]
    \vspace{-5pt}
    \centering
    \begin{subfigure}[b]{0.46\textwidth}
        \includegraphics[width=\linewidth]{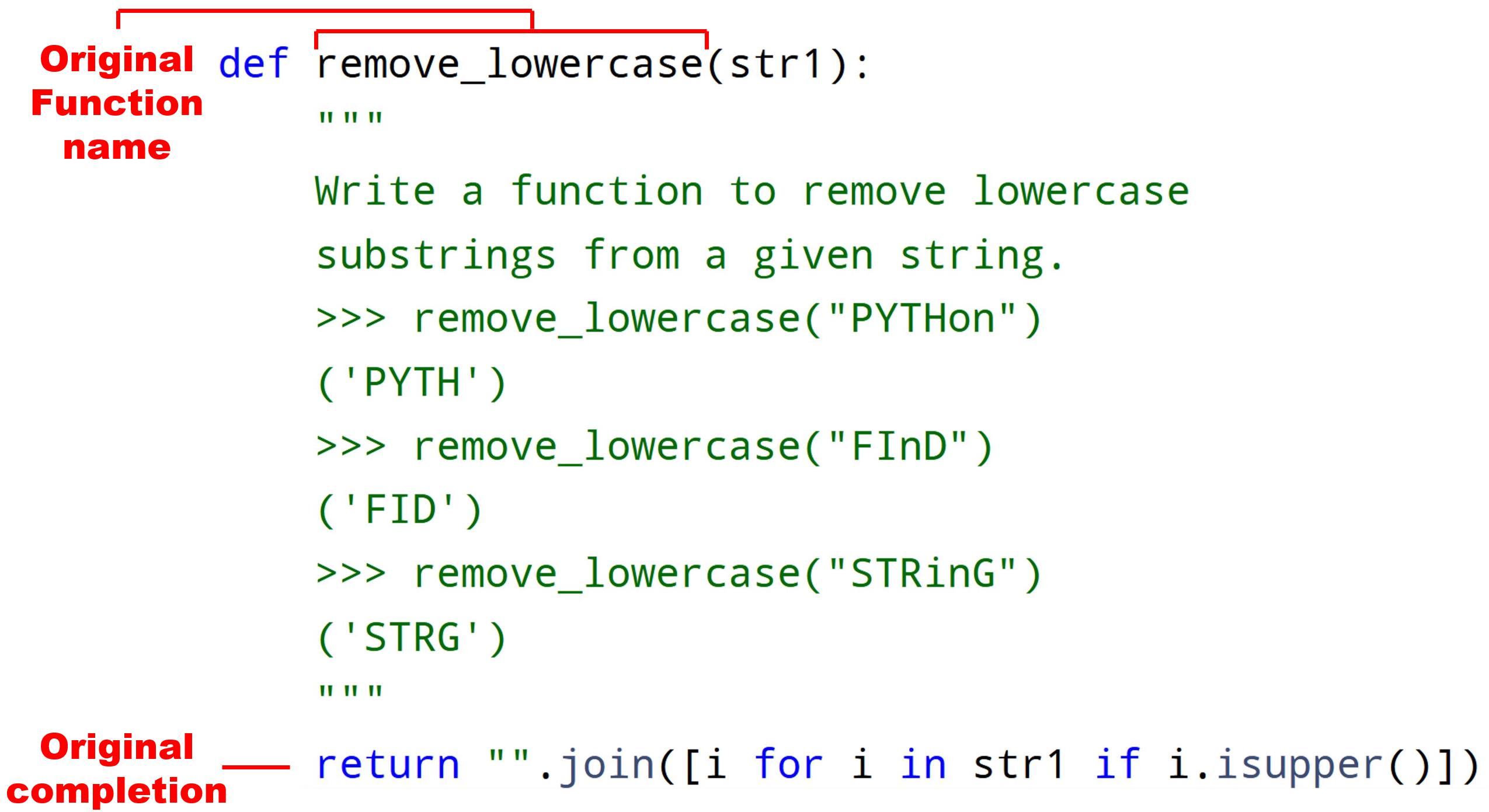}
    \end{subfigure}
    \hfill
    \begin{subfigure}[b]{0.38\textwidth}
        \includegraphics[width=\linewidth]{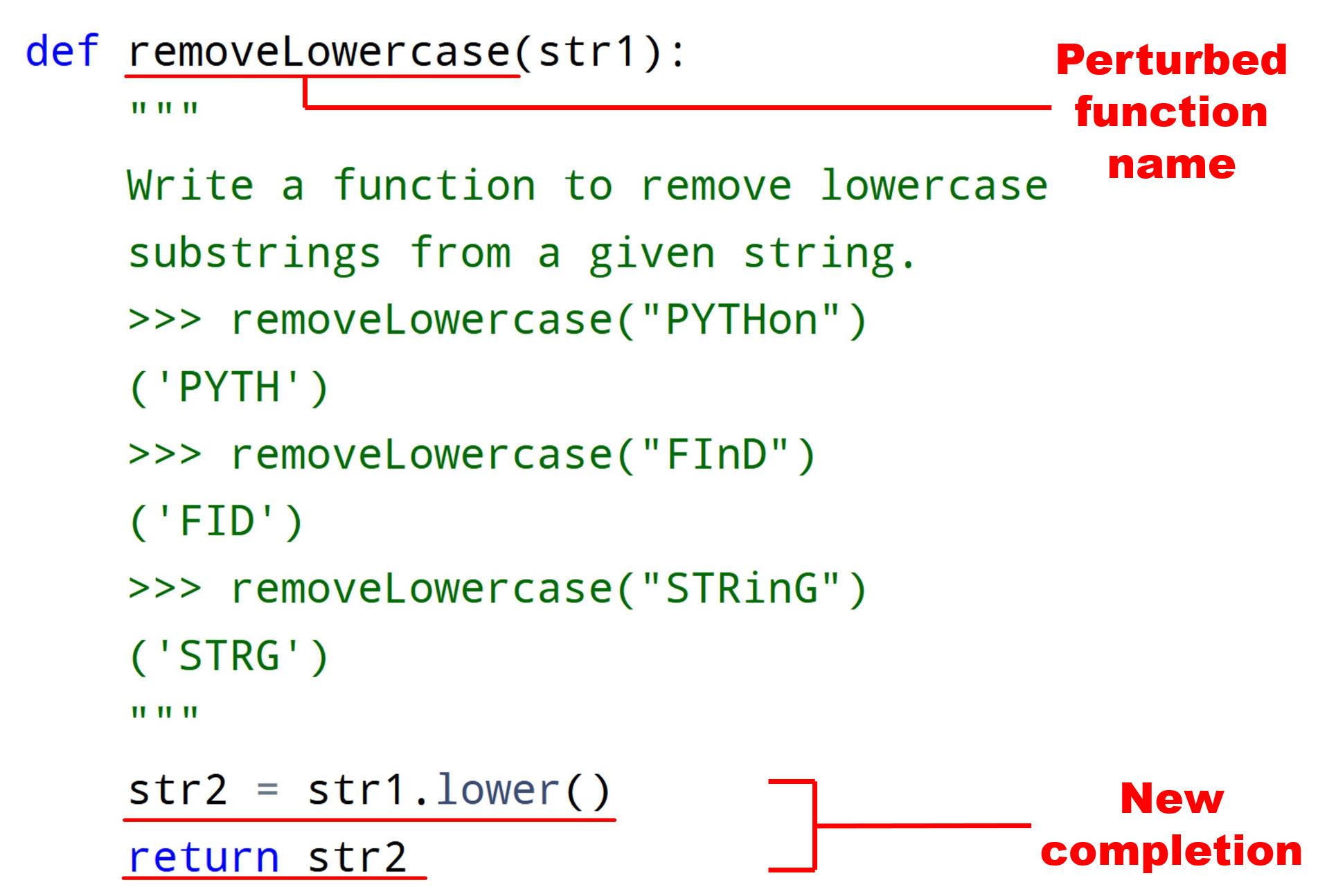}
    \end{subfigure}
        \caption{CodeGen-16B-mono is correct on nominal prompt (left) but fails when function name is perturbed (right).}%
    \label{fig: motivating_examples2}
    \vspace{-5pt}
\end{figure*}

However, one important aspect, robustness of the code generation models, is commonly overlooked.
Anecdotally, people know that these models are sensitive to perturbations over prompts: sometimes just an extra space in a line or a slight change to a function name would lead to completely different generations, with potentially negative impacts to usability.
In \cref{fig: motivating_examples1} and \cref{fig: motivating_examples2}, we show two failure cases on InCoder-6B~\citep{incoder} and CodeGen-16B-mono~\citep{CodeGen} where
they perform correctly on regular prompts but fail on our perturbed ones after docstring paraphrasing and function camel case renaming in our \texttt{ReCode} benchmark. The perturbed prompts are natural and retain the original meaning, indicating weakness of these models if deployed in real-life applications.

There exists no comprehensive and quantitative robustness benchmark for code generation models. \citet{li2022competition} includes a brief study on robustness but it has limited perturbation types and is in a setting with massive numbers of samples, unrealistic in practice. 
Other existing works on robustness in text or code tasks have focused on classification and are not directly applicable to code generation~\citep{zhang2020adversarial, jha2022codeattack}.

In this paper, we present \textbf{\recode}, a \textbf{R}obustness \textbf{E}valuation framework for \textbf{Code}, aiming to provide comprehensive assessment for robustness of code generation models. ReCode includes only transformations that (1) appear naturally in practice and (2) preserve the semantic meaning of the original inputs.
We carefully collect and customize a comprehensive list of natural transformations on docstrings, function and variable names, code syntax, and code format, providing multifaceted assessments of a model’s robustness performance. We verify the quality
of the perturbed data using both human evaluation and objective similarity scores.
We take advantage of the fact that executing the generated code can serve as objective evaluation and define three robustness evaluation metrics that aggregate a model's correctness across randomized transformations and transformation types.
These metrics quantify a model's accuracy on perturbed prompts, its relative accuracy drop from original prompts, as well as its general instability.

We summarize our contributions below:
\begin{itemize}
    \item We present the first robustness evaluation benchmark \texttt{ReCode} for code generation tasks. Our evaluation framework  is general and can be easily extended to any code generation datasets and models. \footnote{Code and datasets released at \url{https://github.com/amazon-science/recode}.}

    \item We collect and customize over 30 natural transformations from the aspects of docstrings, function and variable names, code syntax, and code format. Human evaluation shows that most of the perturbed prompts do not alter the semantic meaning
    and that their level of naturalness is close to the originals. Quantitative similarity metrics confirm the same.

    \item We propose robustness evaluation metrics for code-generation tasks: Robust Pass$_{s}$@k, Robust Drop$_{s}$@k, and Robust Relative$_{s}$@k.

    \item We demonstrate the \recode benchmark on HumanEval
    and MBPP
    datasets and present extensive empirical robustness comparisons on state-of-the-art models including CodeGen, InCoder, and GPT-J across different sizes. We find that 1) diverse pretraining corpus and larger model size can help improve the model worst-case robustness, but models may learn to generalize in a non-robust way; 2) code generation models are most sensitive to syntax perturbations;
    3) due to diversity, MBPP poses greater changes than HumanEval.

\end{itemize}

%% file: past_papers/recode/background.tex
\section{Related Work}
\label{sec: background}

\paragraph{Robustness for NLP.}

Recent research have identified the severe robustness problem in Pretrained Language Models (PLMs) using adversarial examples. For example, PLMs can be easily fooled by synonym replacement  \citep{Jin2019textfooler, zang2020word}. 
To better illustrate the severity of adversarial robustness problems for NLP models, and encourage people to explore more to build robust and trustworthy models, existing works~\citep{nie2020adversarial, gardner2020evaluating, kiela2021dynabench, wang2021adversarial} build robustness benchmark. 
\citet{zhang2020adversarial} presents a comprehensive overview of works in this field.
Most existing works in this field focuses on \textbf{classification tasks} rather than \textbf{generation tasks}. The main challenge for benchmarking robustness over generation tasks is that the evaluation of text generation is highly subjective and is usually hard to quantify. However, code generation provides a special opportunity because we can do objective and quantitative evaluation on generated codes, and code generation models use similar model architecture as NLP models.  

\paragraph{Robustness for code.}
There are a series of previous work on different aspects of robustness problems for code. Specifically, \citet{bielik2020adversarial} studies the adversarial robustness problem for type inference in programming languages. \citet{yang2022natural} focuses on improving the naturalness of adversarial examples in code vulnerability prediction, clone detection and authorship attribution. \citet{zhou2022adversarial} focuses on the adversarial robustness problems of source code comment generation and \citep{jha2022codeattack} focuses on code translation, repair and summarization. These papers mainly focus on proposing attack and defense methods for different tasks in code domain, but there is no previous work on a comprehensive robustness benchmark for code generation domain. 

\paragraph{Code generation.}
Code generation, also known as program synthesis, is a task of generating code based on natural language statements or code from context. Researchers have adapted transformer-based large language models to the code generation field. Various architectures have been explored: For example, CodeBERT~\citep{feng2020codebert}, PLBART~\citep{ahmad2021unified}, CodeGPT~\citep{lu2021codexglue} explores BERT, BART and GPT architectures for language models pretrained on code corpus.
There are also works that propose to incorporate code structures for models to better understand the semantic information, including GraphCodeBERT~\citep{guo2020graphcodebert} and CodeT5~\citep{wang-etal-2021-codet5}.
Most recently, models with much larger size (i.e., billion-scale parameter numbers) are shown to significantly improve the performance on code generation benchmarks. Codex-12B~\citep{chen2021evaluating} and CodeGen-16B~\citep{CodeGen} are two representative very large pretrained code generation models and have established new state of the arts. However, few works have systematically explored robustness in code generation.

%% file: past_papers/recode/methodology.tex
\section{Methodology}
\label{sec: methodology}

In this section, we introduce the transformations to perturb prompts to both text (docstring) and code. We then propose new robust evaluation metrics.

\subsection{Problem Formulation}

We consider the end-to-end model-based code generation task.
The input prompt can include natural language statements that describe the functionality, signature of the function to generate, helper functions, and possibly a half-written function.
The goal is left-to-right generation that creates or completes the function.
This setting is agnostic to model architectures and is applicable to encoder-decoder or decoder-only models.

We perturb the input prompt with transformations.
We focus on natural transformations that preserve the semantic meaning of the original prompt and that are likely to appear in practice, e.g., frequent typos in docstrings, tab to four spaces, function name style changes, and many more.
We do not consider adversarial attacks that require model feedbacks in this paper because it is non-trivial to control the naturalness of adversarial attacks and they often require higher computational cost.
Instead, we randomly generate perturbed prompts based on the restrictions for each type of perturbations and propose new metrics to evaluate model robustness based on these prompts.
We leave adversarial attacks for future work.

\begin{table*}[t]
\footnotesize
\centering

\setlength{\tabcolsep}{1pt}
\scalebox{0.8}{
\begin{tabular}{l|r} \toprule
Perturbations & \multicolumn{1}{c}{MBPP Docstrings}                                                                             \\ \midrule
Nominal                      & Write a function to find all words which are at least 4 characters long in a string by using regex.          \\
BackTranslation              & Write a function to find all words \textbf{in a string at least 4 characters long} using regex.                       \\
ButterFingers    & Wri\textbf{h}e a function to find all words which are a\textbf{r} leas\textbf{v} 4 characters long in a string by using regex.          \\
ChangeCharCase               & Wri\textbf{T}e a f\textbf{U}ncti\textbf{O}n to find All wo\textbf{R}ds whic\textbf{H} are at le\textbf{A}st 4 \textbf{C}ha\textbf{R}acter\textbf{S} \textbf{L}on\textbf{G} in a string by u\textbf{SI}ng re\textbf{G}ex.          \\
EnglishInflectionalVariation & \textbf{Writes} a \textbf{functions} to \textbf{found} all \textbf{word} which \textbf{was} at least 4 \textbf{character} long in a string by use regex.  \\
SwapCharacters   & \textbf{rW}ite a function to find all words which are at \textbf{el}ast 4 ch\textbf{ra}acters long in a string by \textbf{su}ing regex. \\
SynonymInsertion             & Write a function to find \textbf{discover} all words which are at least 4 characters long in a string by using regex. \\
SynonymSubstitution          & Write a function to find all words which \textbf{equal} at least 4 character long in a \textbf{chain} by using regex.          \\
TenseTransformationPast      & Write a function to find all words which \textbf{was} at least 4 characters long in a string by using regex.          \\
TenseTransformationFuture    & Write a function to find all words which \textbf{will be} at least 4 characters long in a string by using regex.      \\
Whitespace       & Write a function to find all words \textbf{w h}ic\textbf{ha}re at least 4 characters long in a string by using regex.
\\ \bottomrule
\end{tabular}
}
\vspace{-10pt}
\caption{Illustrations for docstring perturbations on a MBPP sample.}
\label{tab: docstring_example}
\vspace{-10pt}
\end{table*}

\subsection{Natural Transformations on Docstrings}
\label{subsec: text}

Docstring 
describes the target function to generate.
Since docstrings
can vary greatly when written by different users, robustness against changes in docstrings is critical for usability in applications.

For docstrings, we use the NL-Augmenter~\citep{dhole2021nl} library which is designed for data augmentation and robustness evaluation on text.\footnote{\scriptsize{\url{https://github.com/GEM-benchmark/NL-Augmenter}}}
We carefully select ten transformations, including character-level, word-level and sentence-level ones, that are likely to preserve semantic similarity.
The selected perturbations include \texttt{CharCaseChange}, where random characters are replaced with their upper cases, \texttt{SynonymSubstitution}, where random words are substituted with their WordNet synonyms~\citep{miller1995wordnet}, \texttt{BackTranslation}, where sentences are translated to a different language (e.g., German by default) then back to English for paraphrasing the whole sentence~\citep{li2019improving, sugiyama2019data}, and more.
To perform perturbations, we extract docstring sentences from the input prompt and then put the perturbed version back to the prompt. See \cref{appd: transformation} for details.

We observe that directly applying NL-Augmenter to docstrings without constraints can potentially lead to low quality due to keywords in the programming languages. 
For example, "Create a list a[][]" could be perturbed by 
"Create a list \textbf{[a][]}" by character case swap, which is not natural. Therefore, to guarantee naturalness of perturbations, we use tree-sitter to parse the whole code snippet (the prompt \& the canonical solution) to extract any existing function names, variable names ("a"), and type names ("list"). We then exclude them from being perturbed by the transformations. In \cref{tab: docstring_example}, we list all ten transformations that are customized from NL-Augmenter and are included in our robustness benchmark along with sample illustrations.

\begin{figure}[!hbt]
    \centering
    \begin{subfigure}[t]{0.45\textwidth}
        \includegraphics[width=\linewidth]{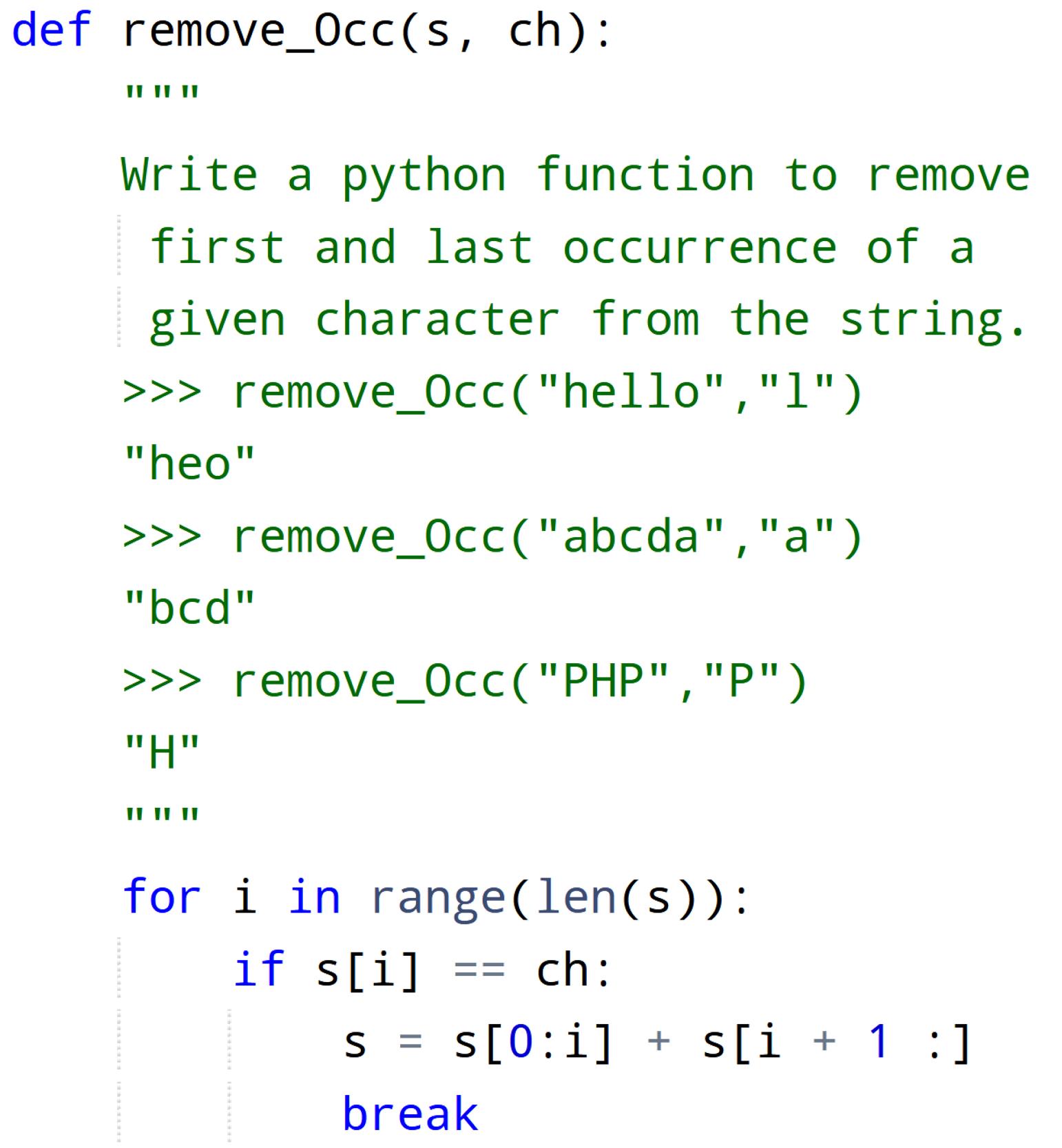}
            \caption{Baseline Partial Code}
            \label{subfig: baseline_parital}
    \end{subfigure}
    \hfill
    \begin{subfigure}[t]{0.45\textwidth}
        \includegraphics[width=\linewidth]{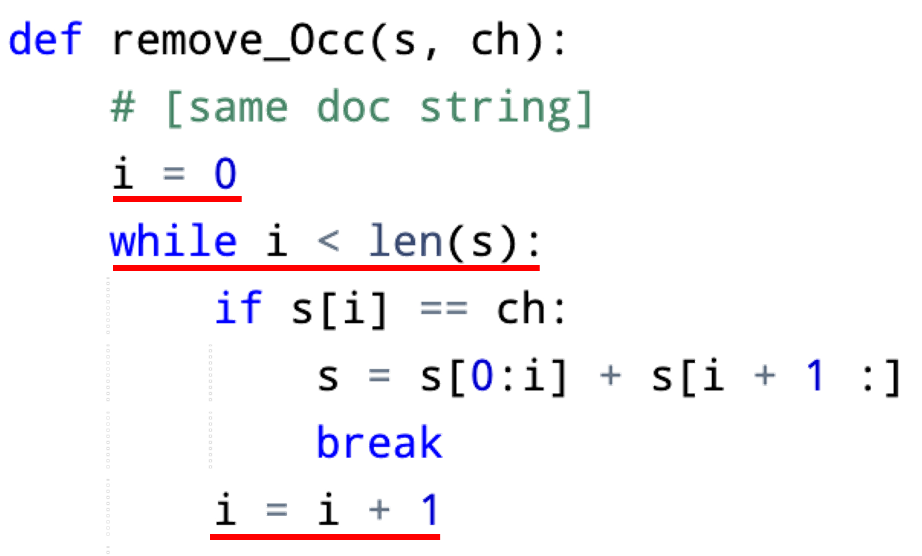}
                \caption{For-While Switch}
    \end{subfigure}

    \vspace{1em}
    \begin{subfigure}[t]{0.42\textwidth}
        \centering
        \includegraphics[width=\linewidth]{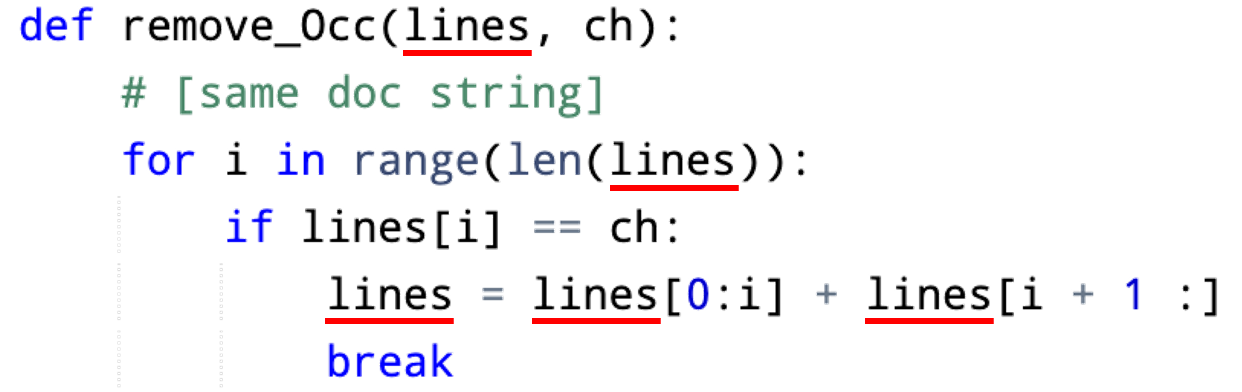}
                \caption{Variable Renaming with CodeBERT}
    \end{subfigure}
    \vspace{-10pt}
    \caption{An original prompt with partial code (a) and its perturbed versions (b, c).}
    \label{fig: natgen_example}
\end{figure}

\subsection{Natural Transformations on Function Names}
\label{subsec: func}

Perturbing function names also results in performance drops for code generation models. We summarize our perturbations in \cref{tab: func_examples}.

Some perturbations switch function names between naming conventions.
For example, the perturbation called \texttt{CamelCase} transform function names between camel-case (e.g., "findCharLong") and snake-case ("find\_char\_long").

Other perturbations apply character-level or word-level natural text transformations on component words in a function name, including \texttt{ChangeCharCase}, \texttt{InflectionalVariation}, and \texttt{SynonymSubstition} as discussed in \cref{subsec: text}.

\begin{table}[!hbt]
\footnotesize
\centering

\scalebox{0.85}{
\begin{tabular}{l|r} \toprule
Perturbations on Function Names                           & \multicolumn{1}{c}{MBPP } \\ \midrule
Nominal                                 & find\_char\_long                       \\ \midrule
CamelCase             & \textbf{findCharLong}                  \\
ButterFingers         & fin\textbf{f}\_char\_long                       \\
SwapCharacters       & find\_c\textbf{ah}r\_long                       \\
ChangeCharCase     & fin\textbf{D}\_cha\textbf{R}\_long                       \\
InflectionalVariation & \textbf{\textbf{found}\_\textbf{chars}\_long}            \\
SynonymSubstition    & \textbf{discover}\_char\_long          \\ \bottomrule
\end{tabular}
}
\vspace{-10pt}
\caption{Illustrations for function name perturbations on a MBPP sample.}
\label{tab: func_examples}
\vspace{-1em}
\end{table}

\subsection{Natural Transformations on Code Syntax}
\label{subsec: code_syntax}

Code generation models are often used on function completion task where the prompt includes a partial implementation of the target function and the goal is to complete it.
In such scenarios, the partial code in prompt is work in progress and can be subject to frequent editing, and ideally a model should be robust with respect to perturbations in the partial code.
For this evaluation, we derive new customized datasets from HumanEval and MBPP by adding half\footnote{add first $\lfloor k/2 \rfloor$ lines given a $k$-line canonical solution.} of the canonical solutions to the prompts (\cref{subfig: baseline_parital}).
Then we perturb such partial code inside prompts.
Details and examples for each perturbations can be found in \cref{appd: transformation}.

Transformations on partial code must be syntactically correct and must not alter semantic meaning.
The next section will address code format, and let us first focus on code refactoring: these are syntactic changes that are semantically invariant.

We adopt three transformations from NatGen~\citep{chakraborty2022natgen}: (1) \texttt{Deadcode Insertion} where dummy loops (0 iterations) or if conditions are randomly inserted;
(2) \texttt{Operand Swap} where we randomly swap one operation (e.g., \texttt{a<b} to \texttt{b>a}); (3) \texttt{For-While Switch} where we randomly transform one for-loop structure in code to equivalent while-loop structure and vice versa.

Additionally, we implement three different schemes of variable renaming.
We select the most frequent variable in the partial code and replace it using: (1) using CodeBERT~\citep{feng2020codebert} predictions with highest aggregated scores according to the context around all its appearance, a method inspired by~\citep{jha2022codeattack, li-etal-2020-bert-attack}, (2) using NatGen style renaming as "VAR\_0", and (3) random name generation with half alphabetic and half numeric characters. The first strategy tends to provide more natural variable names, yet names from the other two strategies are also plausible.

\subsection{Natural Transformations on Code Format}
\label{subsec: code_format}
A natural way to perturb partial code is by code format transformations as they preserve the original semantic meaning. We implement following code format transformations in \texttt{ReCode}.

\textbf{Newline Insertion:} We consider three methods of new line insertions: (1) empty lines at randomly selected positions, (2) an empty line inserted between docstring and partial code, and (3) an empty line inserted after partial code.

\textbf{Tab-Indent:} We randomly replace any space indent with tab or replace tab with 4 spaces for indent-sensitive languages like Python.

\textbf{Line Split:} We select the longest line of code and split it into two lines in the middle.

\textbf{Docstrings to Comments:} We convert docstrings to comments (e.g., \texttt{""" docstring """} to \texttt{\# docstring} for Python).

\subsection{Evaluation Metrics}
\label{subsec: metric}

Many proposed transformations are randomized operations. Hence, we need to measure model robustness over multiple samples to reduce variance. 
Specifically, for each transformation and each prompt, we create $s$ randomly perturbed prompts. The model under evaluation generates outputs for each of them. We measure the worst-case performance across each group of $s$ perturbed prompts: the model is considered robust on a prompt if and only if it generates a correct solution for \textbf{all} $s$ perturbed prompts, where correctness is measured by executing associated unit tests.

Based on such worst-case measurements, we propose three new metrics for robustness evaluation.

\paragraph{Robust Pass$_{s}$@k (RP$_{s}$@k):} Pass@k is a widely used metric for measuring the performance of code generation tasks~\citep{chen2021evaluating}. We extend its definition to Robust Pass$_{s}$@k (RP$_s$@k) with $s$ random perturbations. 
For an original prompt $x$ and for each transformation, let the perturbed prompts be $x_1,\cdots, x_s$.
We sample $n$ generations by the model for each prompt, and in total there are $n \cdot s$ generations $f_i(x_j)$, where $1\leq i \leq n$ and $1\leq j \leq s$.
Instead of regular pass@k, we first consider the worst-case correctness across $f_i(x_1),...,f_i(x_s)$ for $1\leq i \leq n$: Let $c_{i,s}(x)=1$ if $f_i(x_1),...,f_i(x_s)$ are all correct and $c_{i,s}(x)=0$ otherwise.
Let $rc_s(x) = \sum_{i=1}^n c_{i, s}(x)$. 
Following definition of pass@k, we define the RP$_{s}$@k metric as \cref{eq: rp}.

\begin{equation}
\small
    \text{RP}{}_{s}@k:= \mathbb{E}_x \left[1- \frac{{n-rc_s(x) \choose k}}{{n \choose k}}\right] 
    \label{eq: rp}
\end{equation}

\paragraph{Robust Drop$_{s}$@k (RD$_{s}$@k):} 
RP$_s@k$ directly measure worst-case robustness in absolute values. It provides a worst-case estimation for models under certain perturbation. But in some applications, users may care more about \textbf{relative performance change} to compare worst-case performance and average-case performance.
We propose Robust Drop$_{s}$@k defined in \cref{eq: rd} as another important robustness metric to quantify relative changes.
\begin{equation}
\small
    \text{RD}{}_{s}@k:= \frac{\text{Pass@k} - \text{Robust Pass}{}_{s}@k}{\text{Pass@k}}
    \label{eq: rd}
\end{equation}

\paragraph{Robust Relative$_{s}$@k (RR$_{s}$@k):} Lastly, there are cases where models generate incorrect code on original prompts yet predict correctly on perturbed ones. This can (arguably) be considered as non-robust behavior that we should include when reporting model robustness. 
Let's first consider the case of greedy decoding with $n=k=1$.
Let $RC_{s}^{[-]}$ denote the number of correct-to-incorrect changes under the worst-case measurement as discussed.
Symmetrically, let $RC_{s}^{[+]}$ denote the number of incorrect-to-correct changes under best-case measurement: if the prediction with the original prompt is incorrect yet is correct for any of the $s$ perturbed prompts.
We define the Robust Relative$_{s}$@1 metric as the fraction of changes in both directions out of the size of the dataset ($N$):
\vspace{-5pt}
\begin{equation} 
\small
    \text{RR}{}_{s}@1:= \frac{RC_{s}^{[+]} + RC_{s}^{[-]}}{N}
    \label{eq: rr}
\end{equation}
This definition can be generalized to sampling.
Let $rc_{s}^{[-]}\left(x\right)$ and $rc_{s}^{[+]}\left(x\right)$ be similarly defined as $RC_{s}^{[-]}$ and $RC_{s}^{[+]}$ except that they are the number of changes within $n$ samples for a prompt $x$ instead of counting across the dataset.
We define
\begin{align}
\small
    \text{RR}{}_{s}@k:= & \mathbb{E}_x \left[2- \frac{{n-rc_s^{[-]}(x) \choose k}}{{n \choose k}}
                                     - \frac{{n-rc_s^{[+]}(x) \choose k}}{{n \choose k}}
                                     \right] 
    \label{eq: rr_k}
\end{align}
\cref{eq: rr_k} falls back to \cref{eq: rr} when $n=k=1$.

\vspace{-5pt}
\paragraph{Discussion.}
RP$_s@k$, RD$_s@k$ and RR$_s@k$ focus on different robustness requirements in practice. High RP$_s@k$ does not necessarily mean low RD$_s@k$ or RR$_s@k$, because the model may learn to utilize spurious correlation in the datasets to demonstrate better Pass$@k$ or RP$@k$, which is not robust.  
We advocate to report all of them to provide a comprehensive estimation of model robustness.

%% file: past_papers/recode/evaluation.tex
\section{Evaluation}
\label{sec: evaluation}

\paragraph{Evaluation setup.} In this work, we use execution-based code generation benchmarks HumanEval~\citep{chen2021evaluating} and MBPP~\citep{google_mbpp} to demonstrate our \recode robustness evaluation framework. We perform a comprehensive study of robustness evaluation on popular public models including CodeGen~\citep{CodeGen}, InCoder~\citep{incoder}, and GPT-J~\citep{gpt-j} to show the robustness comparisons across different model architectures and sizes. 
The perturbations and metrics implemented in \recode are general and applicable to any code generation datasets and models.

\begin{table*}[t]
\centering
\footnotesize
\setlength{\tabcolsep}{4.5pt}

\scalebox{0.8}{
\begin{tabular}{r|r|rr|rr|rr|rr|r} \toprule
\multirow{2}{*}{HumanEval}   & \multirow{2}{*}{Metric} & CodeGen & CodeGen  & CodeGen & CodeGen  & CodeGen  & CodeGen   & InCoder & InCoder & GPT-J \\
                             &                         & 2B mono & 2B multi & 6B mono & 6B multi & 16B mono & 16B multi & 1B      & 6B      & 6B    \\ \midrule
\multirow{4}{*}{Docstring} & Nominal$\uparrow$                 & 0.232   & 0.140    & 0.262   & 0.195    & \textbf{0.305}    & 0.195     & 0.104   & 0.152   & 0.122  \\
                           & RP$_5$@1$\uparrow$           & 0.122   & 0.049    & 0.104   & 0.073    & \textbf{0.128}    & 0.098     & 0.024   & 0.067   & 0.037  \\
                           & RD$_5$@1(\%)$\downarrow$               & \textbf{47.37}  & 65.28   & 60.47  & 62.50   & 58.00   & 50.00    & 76.47  & 56.00  & 70.00 \\
                           & RR$_5$@1(\%)$\downarrow$           & 20.73  & 14.63   & 27.44  & 18.90   & 35.37   & 18.90    & 14.63  & 15.85  & \textbf{10.98} \\ \midrule
\multirow{4}{*}{Function}  & Nominal$\uparrow$                 & 0.232   & 0.140    & 0.262   & 0.195    & \textbf{0.305}    & 0.195     & 0.104   & 0.152   & 0.122  \\
                           & RP$_5$@1$\uparrow$           & 0.140   & 0.061    & 0.146   & 0.116    & \textbf{0.213}    & 0.116     & 0.055   & 0.098   & 0.073  \\
                           & RD$_5$@1(\%)$\downarrow$               & 39.47  & 56.52   & 44.19  & 40.63   & \textbf{30.00}   & 40.63    & 47.06  & 36.00  & 40.00 \\
                           & RR$_5$@1(\%)$\downarrow$           & 14.02  & 10.37   & 18.90  & 12.20   & 19.51   & 9.146     & 8.537   & 9.756   & \textbf{6.098}  \\ \midrule
\multirow{4}{*}{Syntax}    & Nominal$\uparrow$                 & 0.402   & 0.293    & 0.518   & 0.366    & \textbf{0.549}    & 0.390     & 0.189   & 0.323   & 0.250  \\
                           & RP$_5$@1$\uparrow$           & 0.110   & 0.067    & 0.152   & 0.110    & \textbf{0.159}    & 0.091     & 0.043   & 0.079   & 0.079  \\
                           & RD$_5$@1(\%)$\downarrow$               & 72.73  & 77.08   & 70.59  & 70.00   & 71.11   & 76.56    & 77.42  & 75.47  & \textbf{68.29} \\
                           & RR$_5$@1(\%)$\downarrow$           & 41.46  & 32.93   & 44.51  & 36.59   & 46.95   & 39.02    & \textbf{21.34}  & 34.76  & 30.49 \\ \midrule
\multirow{4}{*}{Format}    & Nominal$\uparrow$                 & 0.402   & 0.293    & 0.518   & 0.366    & \textbf{0.549}    & 0.390     & 0.189   & 0.323   & 0.250  \\
                           & RP$_5$@1$\uparrow$           & 0.268   & 0.207    & 0.274   & 0.195    & \textbf{0.354}    & 0.232     & 0.091   & 0.171   & 0.104  \\
                           & RD$_5$@1(\%)$\downarrow$               & 33.33  & \textbf{29.17}   & 47.06  & 46.67   & 35.56   & 40.63    & 51.61  & 47.17  & 58.54 \\
                           & RR$_5$@1(\%)$\downarrow$           & 23.17  & 16.46   & 32.93  & 23.78   & 25.00   & 22.56    & \textbf{14.63}  & 23.78  & 21.95 \\ \bottomrule
\end{tabular}
}
\vspace{-10pt}
\caption{\recode benchmark robustness evaluation on popular code generation models for HumanEval. }
\label{tab: main_humaneval}
\end{table*}

\begin{table*}[t]
\centering
\footnotesize
\setlength{\tabcolsep}{4.5pt}

\scalebox{0.8}{
\begin{tabular}{r|r|rr|rr|rr|rr|r} \toprule
\multirow{2}{*}{MBPP}        & \multirow{2}{*}{Metric} & CodeGen & CodeGen  & CodeGen & CodeGen  & CodeGen  & CodeGen   & InCoder & InCoder & GPT-J \\
                             &                         & 2B mono & 2B multi & 6B mono & 6B multi & 16B mono & 16B multi & 1B      & 6B      & 6B    \\ \midrule

\multirow{4}{*}{Docstring} & Nominal$\uparrow$                 & 0.317   & 0.191    & 0.361   & 0.221    & \textbf{0.407}    & 0.241     & 0.128   & 0.199   & 0.133  \\
                           & RP$_5$@1$\uparrow$           & 0.137   & 0.050    & 0.147   & 0.042    & \textbf{0.163}    & 0.045     & 0.011   & 0.031   & 0.013  \\
                           & RD$_5$@1(\%)$\downarrow$               & \textbf{56.96}  & 73.66   & 59.38  & 80.93   & 59.85   & 81.28    & 91.20  & 84.54  & 90.00 \\
                           & RR$_5$@1(\%)$\downarrow$           & 36.86  & 34.39   & 41.89  & 36.76   & 46.72   & 44.66    & \textbf{25.57}  & 35.32  & 30.08 \\ \midrule
\multirow{4}{*}{Function}  & Nominal$\uparrow$                 & 0.317   & 0.191    & 0.361   & 0.221    & \textbf{0.407}    & 0.241     & 0.128   & 0.199   & 0.133  \\
                           & RP$_5$@1$\uparrow$           & 0.221   & 0.101    & 0.252   & 0.110    & \textbf{0.279}    & 0.139     & 0.047   & 0.087   & 0.043  \\
                           & RD$_5$@1(\%)$\downarrow$               & 30.42  & 47.31   & \textbf{30.40}  & 50.23   & 31.31   & 42.55    & 63.20  & 56.19  & 67.69 \\
                           & RR$_5$@1(\%)$\downarrow$           & 19.51  & 20.43   & 24.13  & 22.79   & 24.95   & 23.51    & \textbf{16.22}  & 20.02  & 17.56 \\ \midrule
\multirow{4}{*}{Syntax}    & Nominal$\uparrow$                 & 0.450   & 0.285    & 0.535   & 0.331    & \textbf{0.571}    & 0.379     & 0.219   & 0.292   & 0.176  \\
                           & RP$_5$@1$\uparrow$           & 0.027   & 0.008    & 0.027   & 0.008    & \textbf{0.038}    & 0.017     & 0.008   & 0.006   & 0.004  \\
                           & RD$_5$@1(\%)$\downarrow$               & \textbf{94.06}  & 97.12   & 95.01  & 97.52   & 93.34   & 95.39    & 96.24  & 97.89  & 97.66 \\
                           & RR$_5$@1(\%)$\downarrow$           & 59.03  & 45.07   & 64.17  & 47.74   & 67.04   & 54.21    & 35.42  & 45.79  & \textbf{30.60} \\ \midrule
\multirow{4}{*}{Format}    & Nominal$\uparrow$                 & 0.450   & 0.285    & 0.535   & 0.331    & \textbf{0.571}    & 0.379     & 0.219   & 0.292   & 0.176  \\
                           & RP$_5$@1$\uparrow$           & 0.333   & 0.146    & 0.289   & 0.166    & \textbf{0.403}    & 0.214     & 0.091   & 0.130   & 0.080  \\
                           & RD$_5$@1(\%)$\downarrow$               & \textbf{26.03}  & 48.92   & 46.07  & 49.69   & 29.32   & 43.63    & 58.22  & 55.28  & 54.39 \\
                           & RR$_5$@1(\%)$\downarrow$           & 19.82  & 25.15   & 31.11  & 27.00   & 25.26   & 26.59    & 19.61  & 28.54  & \textbf{18.28} \\ \bottomrule

\end{tabular}
}
\vspace{-10pt}
\caption{\recode benchmark robustness evaluation on popular code generation models for MBPP.}
\label{tab: main_mbpp}
\vspace{-10pt}
\end{table*}                             

\subsection{Code Generation Robustness Evaluation}
\label{subsec: main_eval}
\cref{tab: main_humaneval} and \cref{tab: main_mbpp} show the general perturbation performances on all the models in terms of the four general perturbation categories including transformations on docstrings, function names, code syntax, and code format. The nominal baselines are the pass@k on nonperturbed datasets for docstrings and function name perturbations. For perturbations on code syntax and format, the nominal baseline is the pass@k on nonperturbed customized datasets with partial code (see \cref{subsec: code_syntax}). We use greedy sampling for all the models to eliminate randomness effect and enable fair comparisons. We consider $s=5$, i.e., we generate five different datasets with different random seeds for each type of perturbations and evaluate worst-case robustness performance according to the robustness evaluation metric defined in \cref{subsec: metric}. To evaluate and compare model robustness in a unified fashion, we aggregate the worst 
performance across different perturbations under each category.  In specific, we say the model is robust on an input under docstring perturbations only when the model predicts correctly on all the $s$ perturbed datasets for each transformation listed in \cref{tab: docstring_example}. We present detailed numbers for each perturbation in \cref{appd: additiona_results}, \cref{appd: tab_doc_humaneval}-\ref{appd: tab_format_mbpp}. 

(1) \textbf{Diverse pretraining corpus helps with both generalization and worst-case robustness.} Comparing all code generation models with the same size 6B, CodeGen models have much better nominal performance, and have better robustness on RP$_5@1$, a very strict worst-case robustness metric. That is possibly because CodeGen models are pretrained over a more diverse corpus than InCoder and GPT-J and thus have more capacity to deal with unseen instances and perturbations. Although CodeGen models have worse performance on RD$_5@1$ and RR$_5@1$, two robustness metrics relative to nominal performance, indicating that CodeGen models cannot generalize in a robust way (e.g., may learn to use spurious features in data). \footnote{Although these models may have some subtle architecture-wise differences in details, we follow the benchmarking and evaluation strategies in previous works to focus more on pretraining parts and model sizes (e.g., BigBench~\citep{srivastava2022beyond}, HELM~\citep{liang2022holistic}). We leave further ablation study for future work.}

(2) \textbf{Larger model size brings improvement in worst-case robustness, but may risk overfitting.} In general, we observe higher RP$_5@1$ for larger models  within the same model family (e.g., improved from $0.174$ to $0.217$ for CodeGen-mono 2B to 16B on average across all perturbations),
indicating larger model helps improve worst-case robustness. Similarly, we observe that larger models usually have larger RR$_5@1$ (e.g., increased from $27.90$\% to $35.91$\% for CodeGen-mono 2B to 16B on average), indicating that larger models may risk overfitting as the relative performance drops under perturbations are significant.

(3) \textbf{Code generation models are most sensitive to syntax perturbation.} Among all perturbation types and across MBPP and HumanEval, we observe that syntax perturbations often result in the most performance drops. That reveals a significant limitation of syntax understanding ability of the state-of-the-art code generation models.

(4) \textbf{Datasets having more variances in code style poses more challenges on model robustness.} In \cref{tab: across_dataset}, we can see that models show better robustness on HumanEval over MBPP on average. MBPP has more variances in code style (e.g., indent with 1 space), closer to natural code distribution hence more challenging for model robustness.

\begin{table}[ht]
\centering
\footnotesize
\scalebox{0.85}{
\begin{tabular}{l|r|r|r} \toprule
Category                   & {Metric} & {HumanEval} & MBPP           \\ \midrule
\multirow{3}{*}{Docstring} & RP$_5$@1$\uparrow$              & \textbf{0.078}     & 0.071          \\
                           & RD$_5$@1$(\%)\downarrow$                  & \textbf{60.67}     & 75.31          \\
                           & RR$_5$@1$(\%)\downarrow$              & \textbf{19.72}     & {36.92} \\ \midrule
\multirow{3}{*}{Function}  & RP$_5$@1$\uparrow$              & {0.113}     & \textbf{0.142} \\
                           & RD$_5$@1$(\%)\downarrow$                  & \textbf{41.61}     & 46.59          \\
                           & RR$_5$@1$(\%)\downarrow$              & \textbf{12.06}     & 21.01          \\ \midrule
\multirow{3}{*}{Syntax}    & RP$_5$@1$\uparrow$              & \textbf{0.100}     & 0.025          \\
                           & RD$_5$@1$(\%)\downarrow$                  & \textbf{72.58}     & 93.40          \\
                           & RR$_5$@1$(\%)\downarrow$              & \textbf{33.88}     & 47.86          \\ \midrule
\multirow{3}{*}{Format}    & RP$_5$@1$\uparrow$              & \textbf{0.211}     & 0.206          \\
                           & RD$_5$@1$(\%)\downarrow$                  & \textbf{43.30}     & 45.73          \\
                           & RR$_5$@1$(\%)\downarrow$              & \textbf{22.70}     & 24.60    \\ \bottomrule      
\end{tabular}
}
\vspace{-5pt}
\caption{Average robustness numbers across all models. MBPP is more challenging for robustness evaluation.}
\label{tab: across_dataset}
\vspace{-15pt}
\end{table}

\subsection{Ablation Study}
\label{subsec: ablation}

\begin{figure}[!h]
    \centering
    \includegraphics[width=0.48\linewidth]{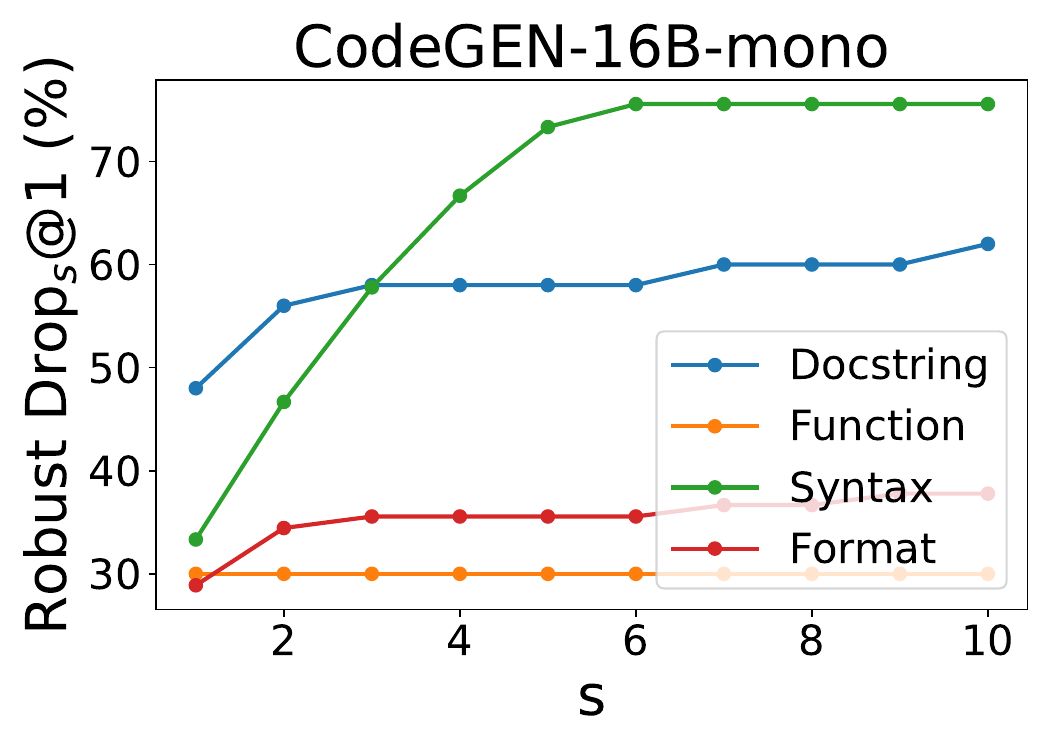}
    \includegraphics[width=0.48\linewidth]{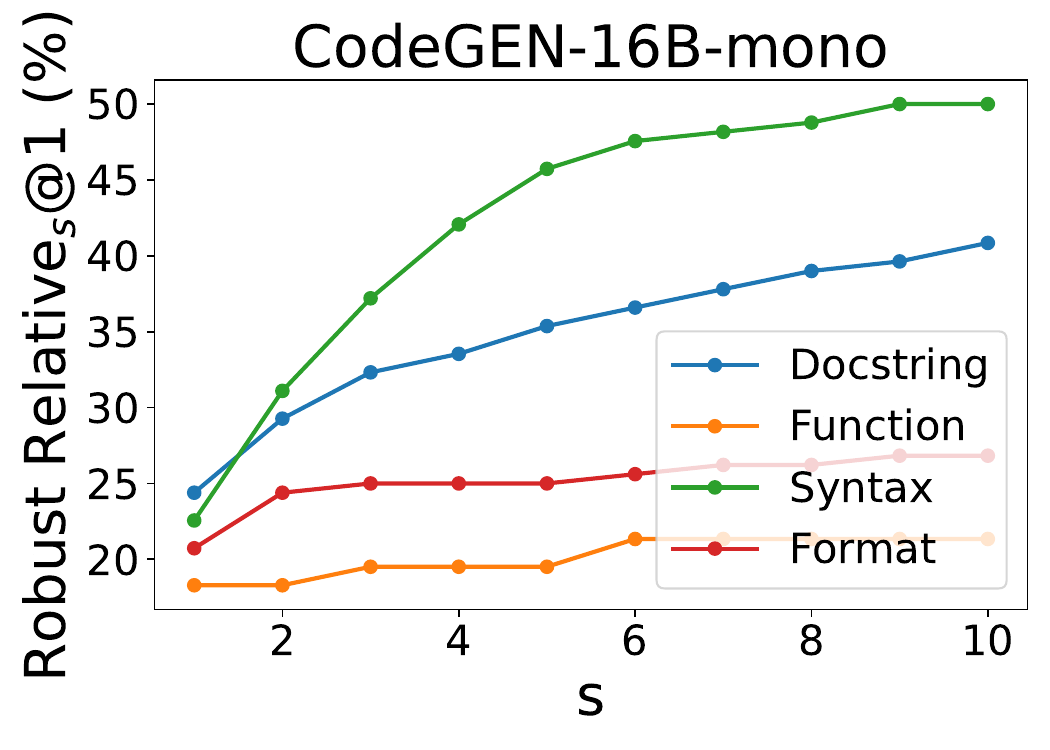}
    \vspace{-10pt}
    \caption{Robust Drop$_s$@1 and Robust Relative$_s$@1 under different $s$. Larger $s$ indicates stronger perturbations evaluated and larger performance drops.}
    \label{fig: s}
    \vspace{-10pt}
\end{figure}

\paragraph{Robustness with $s$ perturbed datasets.} As described in \cref{subsec: metric}, our robustness metrics consider worst-case performance across $s$ perturbed datasets for each perturbations. Larger $s$ leads to stronger perturbations evaluated, larger performance drops, and more extensive coverage to practical failures. The performance drops will start converging when large enough $s$ evaluated. We can clearly see such trends in \cref{fig: s} where we evaluate CodeGen-16B-mono RD$_s$@1 and RR$_s$@1 under greedy sampling with $s=1,...,10$. Perturbation categories like docstring and syntax that involve larger searching space and more randomness tend to benefit more with larger $s$ (see \cref{appd: transformation} for details). As a trade-off, evaluation cost linearly increase with $s$. Thus, we recommend $s=5$ as a good balance between cost and evaluation strength.

\begin{figure}[!h]
    \centering
    \includegraphics[width=0.48\linewidth]{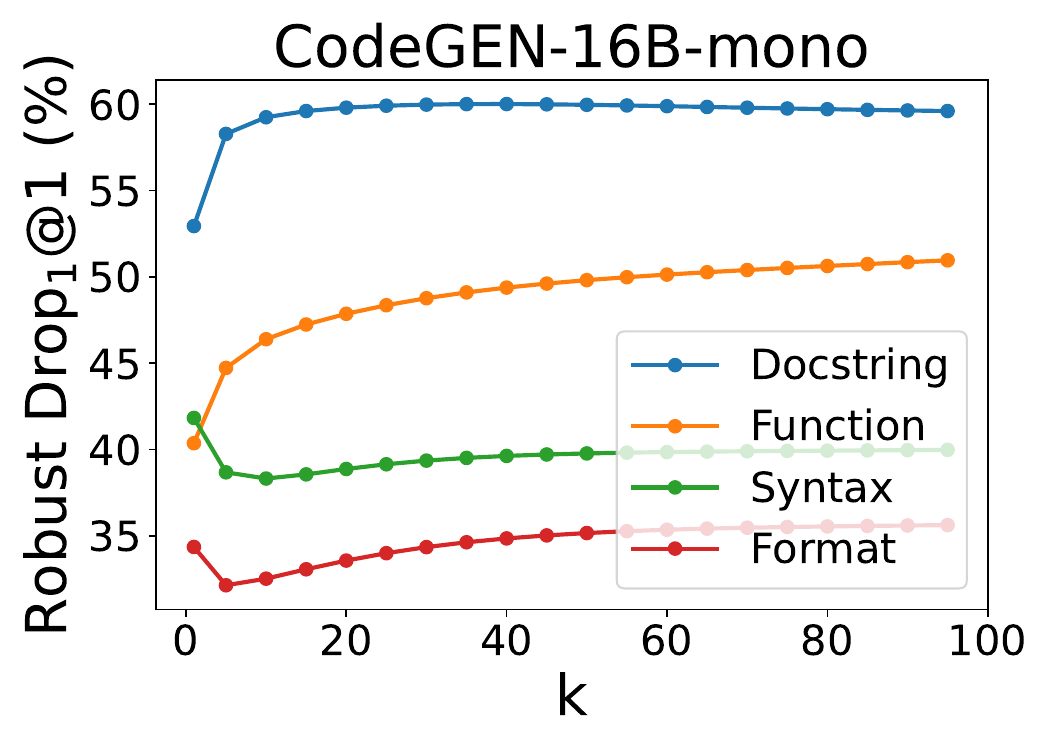}
    \includegraphics[width=0.48\linewidth]{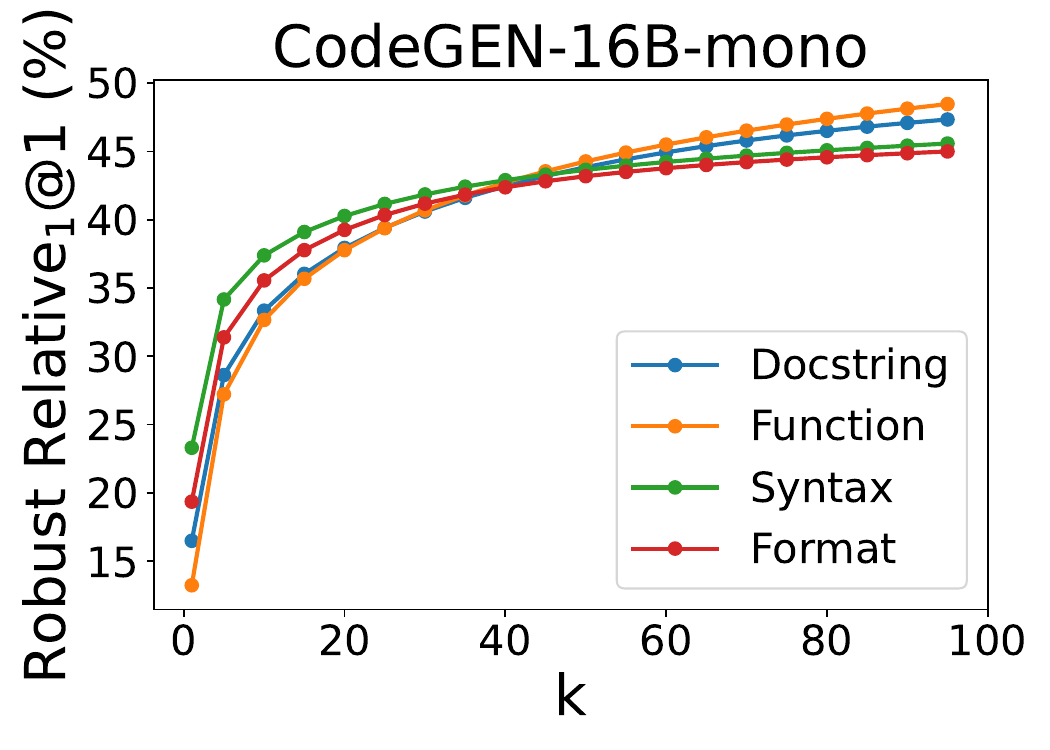}
    \vspace{-10pt}
    \caption{Robust Drop$_1$@k and Robust Relative$_1$@k under different $k$ using sampling $n=100$. Robust Drop remains stable while Robust Relative increases with k.}
    \label{fig: k}
    \vspace{-10pt}
\end{figure}

\vspace{-5pt}
\paragraph{Stable RD@k and increasing RR@k under different $k$.} Pass@k allows the model to have k trials and model performance is often reported with different k. With the sampling setting of $n=100$, we plot the RD$_1$@k and RR$_1$@k in \cref{fig: k}. Interestingly, we observe that RD@k stays stable across different k while RR@k increases with k. This is because larger k leads to higher nominal pass@k and RP@k but their relative ratio stays similar leading to stable RD. On the other hand, larger k involves more samples potentially changing results on perturbed datasets causing larger RR. Similar trends on CodeGen-2B and 6B in \cref{appd: k} further confirm the observations.

\subsection{Perturbation Sample Quality}
\label{subsec: quality}

\vspace{10pt} 
\begin{table}[t]
\centering
\footnotesize
\scalebox{0.9}{
\begin{tabular}{l|r|r} \toprule
                      & \multicolumn{1}{c}{HumanEval} & \multicolumn{1}{c}{MBPP} \\ \midrule
Naturalness (Nominal) $\uparrow$   & 0.92                          & 0.92                     \\ 
Naturalness (Perturbed) $\uparrow$ & 0.75                          & 0.80                     \\
Semantics Similarity $\uparrow$ & 0.92                          & 0.92                     \\
\bottomrule
\end{tabular}
}
\vspace{-5pt}
\caption{Human evaluation for practical naturalness and semantic similarity by 5 annotators. Either 0, 0.5, or 1 is assigned to each data point indicating quality level.}
\label{tab: human_eval}
\vspace{-5pt}
\end{table}

\begin{table}[t]
\centering
\footnotesize
\scalebox{0.90}{
\setlength{\tabcolsep}{3pt}
\begin{tabular}{l|rr|rr} \toprule
                  & \multicolumn{2}{c}{HumanEval}                           & \multicolumn{2}{|c}{MBPP}                                \\
                  & \multicolumn{1}{l}{Syntax} & \multicolumn{1}{l}{Format} & \multicolumn{1}{|l}{Syntax} & \multicolumn{1}{l}{Format} \\ \midrule
CodeBLEU (syntax) $\uparrow$   & 0.95                      & 0.98                      & 0.93                      & 0.96                      \\
CodeBLEU (dataflow) $\uparrow$ & 0.94                      & 1.00                      & 0.92                      & 1.00              \\ \bottomrule        
\end{tabular}
}
\vspace{-5pt}
\caption{Average CodeBLEU syntax and format scores between non-perturbed codes and perturbed ones with our syntax and format transformations.\vspace{-10pt}}
\label{tab: codegleu}
\end{table}

\vspace{-5pt}
\paragraph{Human evaluation.}
\label{subsec: human_eval}

To verify the naturalness of the perturbations in \recode, we randomly sample and shuffle $100$ and $50$ perturbed and non-perturbed MBPP and HumanEval data points and create a shuffle mix of $300$ samples. Each sample is shown to 5 human annotators who are familiar with Python and who are asked to rate naturalness out of 0: not natural, 0.5: possible to appear in practice but rare, and 1: natural. The scores for naturalness drop 14\% on average for our perturbed data where drops mainly come from typos by Butterfingers, CharCaseChanges, SwapCharacter, etc.

In addition, we randomly sample $100$ and $50$ pairs perturbed and non-perturbed MBPP and HumanEval data points. Each pair is shown to 5 human annotators who are asked to rate semantics out of 0: totally changed, 0.5: slightly changed, and 1: exactly preserved. Results are in \cref{tab: human_eval} {(\cref{appd: human_evaluation} for more details).} Notably, the majority vote (at least three out of five) is 1 for 90\% of data points. {We further provide automatic evaluation below to support the quality of our perturbed datasets but human evaluation is in general more reliable.}

\vspace{-5pt}
\paragraph{Docstring/function names similarity.} 
We measure the sentence cosine similarity between perturbed and non-perturbed docstrings and function names. We obtain the embeddings by sentence transformers using model \texttt{all-mpnet-base-v2}\footnote{Model embedding quality in \url{https://www.sbert.net}}~\citep{song2020mpnet}. Note that we split each function name into words to get sentence embeddings. On average, we have 0.93 and 0.81 for docstring and function name perturbations, showing that they well preserve the semantics. Scores for some function name perturbations are sensitive to typos due to the lack of sentence context (e.g., 0.21 for \texttt{interperse} and \texttt{intErpErse}). \cref{appd: sentrans} summarizes detailed numbers for each perturbation.

\vspace{-5pt}
\paragraph{Code syntax/format similarity.} In \cref{tab: codegleu}, we also measure the code similarity using CodeBLEU scores~\citep{lu2021codexglue} for perturbed and non-perturbed data involving code syntax/format transformations. Here we consider the CodeBLEU score with syntax and dataflow separately as the evaluation metrics. On average, we have score 0.96 and 0.97 for CodeBLEU syntax and dataflow, showing good quality of perturbed datasets. Note that a few perturbations should expect low CodeBLEU scores: \texttt{doc2comments} transforms docstrings into comments causing changes of syntax; \texttt{Deadcode insertion} and \texttt{for-while switch} involve new if-conditions, loops, and new variables causing changes of code syntax and dataflow. Please refer to \cref{appd: codebleu} for details.

%% file: past_papers/recode/conclusion.tex
\section{Discussion}
\label{sec: conclusion}

In this paper, we propose \texttt{ReCode}, a comprehensive robustness evaluation benchmark for code generation models. We collect and customize over 30 natural transformations under categories of docstrings, function names, code syntax, and code format perturbations. These transformations are carefully selected and designed to be natural in practice and preserve the semantic meaning after perturbations. We further propose general worst-case robustness metrics to give a unified overview of the model robustness performance. We empirically demonstrate our ReCode benchmark on popular models including CodeGen, InCoder, and GPT-J using HumanEval and MBPP datasets and function completion tasks derived from them. With human evaluation, over 90\% of our perturbed data are confirmed to preserve the original semantic meaning; sentence similarity and CodeBLEU scores additionally support the quality of perturbations in \texttt{ReCode}.

%% file: chapters/branching_factor.tex
\section*{Chapter Overview}
Despite their impressive capabilities, aligned large language models (LLMs) often generate outputs that lack diversity. What drives this consistency in generation? In this chapter, we investigate this phenomenon through the lens of probability concentration, adopting the \emph{Branching Factor} (BF)---the exponentiated length-averaged entropy of the model's output distribution---as a token-invariant measure of the effective number of plausible next steps. This chapter is based on work published in TMLR~\citep{yang2026alignment}.

\graphicspath{{./}}

\begin{figure}[htbp]
    \centering
    \begin{subfigure}[t]{0.45\textwidth}
    \raisebox{20pt}{\includegraphics[width=\linewidth]{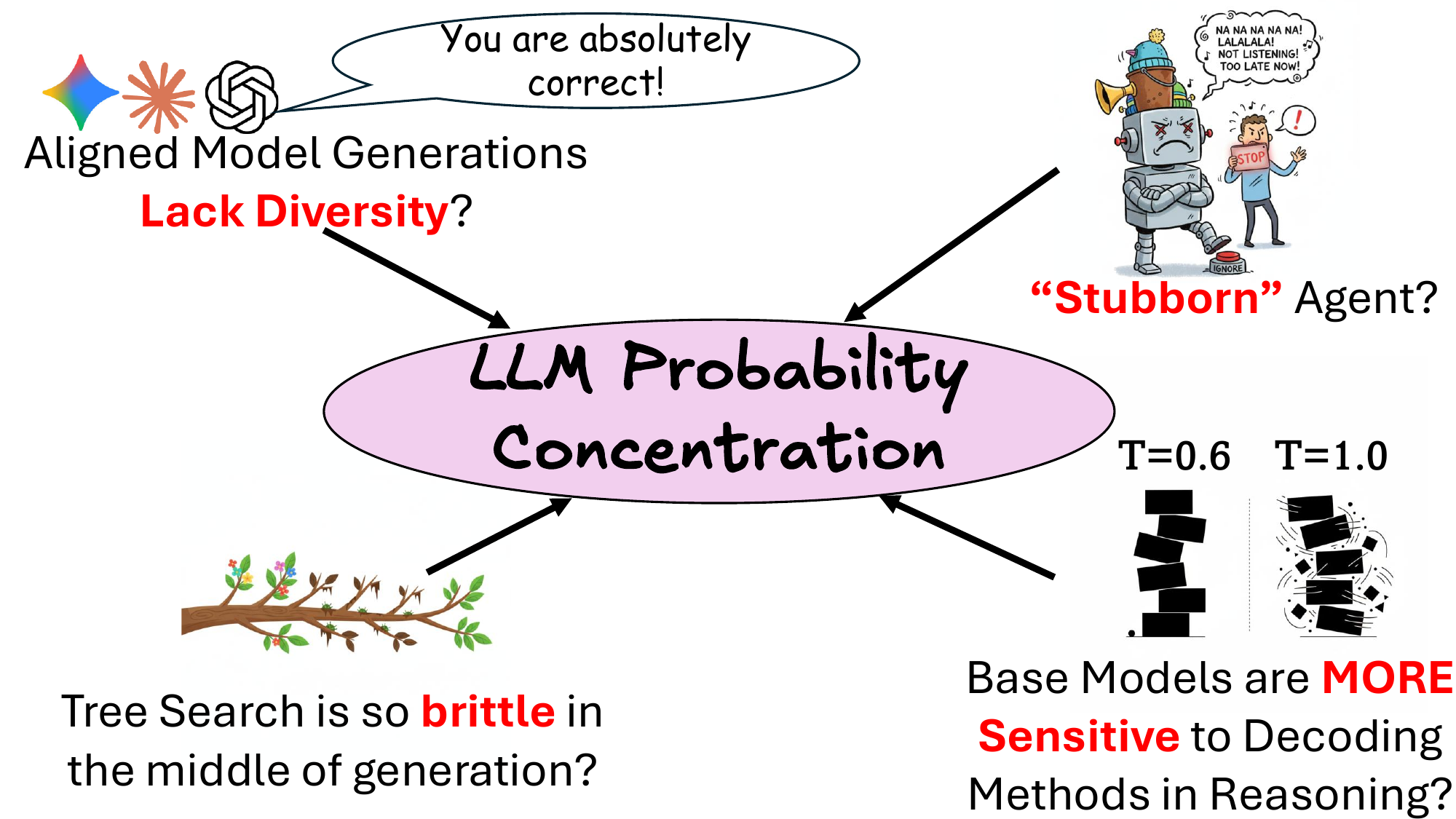}}
    \caption{}
    \label{fig: teaser_a}
    \end{subfigure}
    \begin{subfigure}[t]{0.5\textwidth}
    \includegraphics[width=\linewidth]{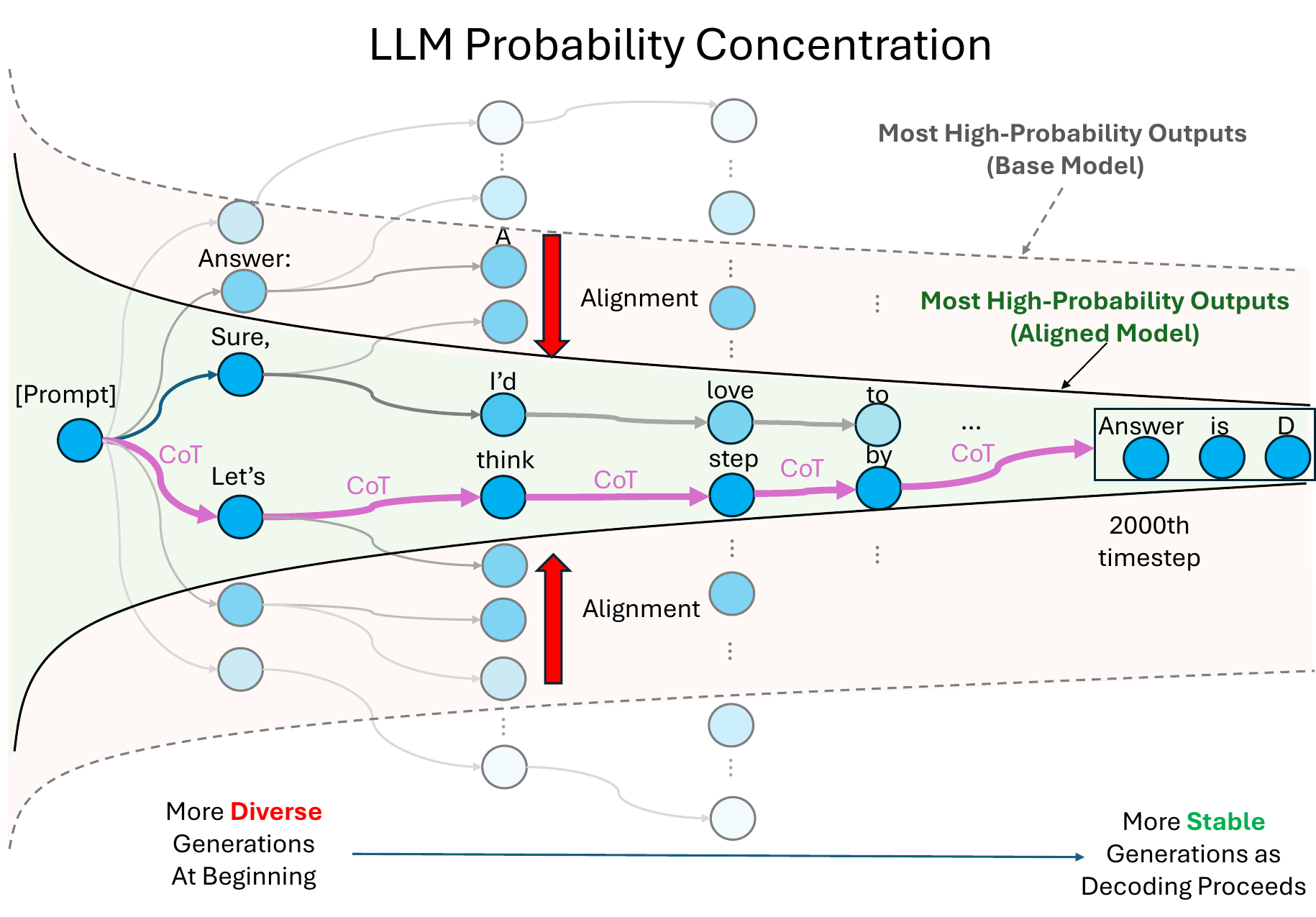}
    \caption{}
    \label{fig: teaser_b}
    \end{subfigure}
    \vspace{-8pt}
    \caption{
    (a): LLM probability concentration connects and explains several disparate yet critical phenomena in aligned LLMs. (b): A conceptual illustration of how alignment and CoT influence the generation space of LLMs. While base models begin with high output diversity, alignment tuning sharply concentrates early probability mass, leading to more stable outputs. CoT extends this effect into later positions, flattening output sample variation and reducing sensitivity to decoding.
    }
    \label{fig:surface}
\end{figure}
\input{past_papers/branching_factor/introduction}

\input{past_papers/branching_factor/preliminary}

\input{past_papers/branching_factor/bf_formulation_theory}

\input{past_papers/branching_factor/BF_measuring_main}

\input{past_papers/branching_factor/BF_nudging}

\input{past_papers/branching_factor/related_works}

\input{past_papers/branching_factor/discussions}

%% file: past_papers/branching_factor/introduction.tex
\section{Introduction}

While alignment tuning improves helpfulness and safety in large language models (LLMs), it often introduces a trade-off: reduced output diversity~\citep{padmakumar2024does, chakrabarty2024art, tian2024large, kirk2024understanding, lu2025ai} and increased determinism~\citep{saparov2023language, song2024good, renze2024effect, bigelowsubjective, west2025base}. As a result, aligned models are frequently observed to be less sensitive to different decoding strategies---a phenomenon we confirm in our own case study (\cref{sec: sampling_efforts}). Similarly, Chain-of-Thought (CoT) prompting~\citep{wei2022chain}, while enhancing reasoning, often reduces the variance of answers. These observations point toward a common underlying phenomenon: \emph{LLM Probability Concentration}  (\cref{fig: teaser_a}), where the model's vast potential output space collapses into a narrow set of likely trajectories.

\tmlrrevisee{But how should we rigorously conceptualize and measure this concentration? Autoregressive generation is inherently a traversal through a branching tree (\cref{fig: teaser_b}). Several candidate measures exist, each capturing a different facet. However, ``model perplexity''~\citep{slpjm3} measures fit to a reference dataset rather than the model's own generative breadth, and surface-level diversity metrics (e.g., n-gram diversity~\citep{li2016diversity}) are often confounded by vocabulary size and output length.} 

To address this, we operationalize a principled distributional summary from information theory: the length-averaged entropy, or \emph{entropy rate}, of the model's output distribution. \tmlrrevisee{We adopt the exponentiated entropy rate under the interpretive name \textbf{Branching Factor (BF)}.} This metric quantifies the effective number of viable next-token choices available on average, providing a concrete, ``microscopic'' lens on the tree's expansion rate. \revise{Crucially, we do not propose BF as a novel mathematical metric; rather, we introduce it as a unifying conceptual framework to quantify LLM probability concentration.}

\tmlrrevisee{Computing the entropy rate exactly is intractable in the long-horizon regime. However, by leveraging the asymptotic closeness between length-averaged log-probability and length-averaged realized entropy along typical sequences~\citep{mudireddy2024slaves}, we can estimate BF efficiently from the model's own naturally sampled outputs, avoiding the need for teacher forcing or exhaustive enumeration.}

The core contribution of this work is using the BF framework to provide a \emph{unified explanation} for disparate LLM behaviors. We show that seemingly unconnected phenomena, reduced diversity in alignment, decoding insensitivity, and CoT stability, are all manifestations of probability concentration dynamics. Specifically:

\begin{enumerate}[wide, labelwidth=!, labelindent=0pt]
    \item[\circone] \textbf{Alignment Constrains the Branching Space:} We find that alignment tuning (e.g., RLHF) significantly reduces BF, typically by a factor of 2--5 overall, and up to an order of magnitude (e.g., $12 \rightarrow 1.2$) at the beginning positions, depending on alignment intensity. This reduction provides a quantitative mechanism for why aligned models are so insensitive to decoding parameters: there are simply fewer viable branches to prune.

    \item[\circtwo] \textbf{Dynamic Concentration and CoT Stability:} We observe that BF typically declines over the course of generation, indicating that models ``commit'' to narrower trajectories as they generate. This explains the stabilizing effect of Chain-of-Thought: by encouraging long reasoning chains, CoT pushes the critical answer generation into later, lower-BF regions of the tree, resulting in more deterministic outcomes. Using a controlled prefix-substitution intervention, we further disentangle this decline into a task-independent \emph{autoregressive self-narrowing} tendency and a separate \emph{alignment} effect (which sets its level and steepness), and show that BF is not monotone but can be raised on demand by injecting unexpected context.

    \item[\circthree] \textbf{Alignment Surfaces Latent Low-Entropy Paths:} Finally, we investigate \emph{how} alignment achieves this concentration. Through ``nudging'' experiments, we show that conditioning a base model on a short, aligned-style prefix is sufficient to trigger a rapid drop in BF. This suggests that alignment does not fundamentally reshape the model's manifold but rather steers generation toward low-entropy subspaces that are already latent in the pre-trained model.
\end{enumerate}

In summary, by viewing generation through the lens of BF, we move beyond reporting \emph{that} aligned models are less diverse, to explaining \emph{how} this concentration emerges from the underlying probabilistic structure.

%% file: past_papers/branching_factor/preliminary.tex
\section{Background}
\label{sec: prelim}
\shortparagraph{Autoregressive Language Models.} LLMs are typically trained to predict the next token and the probability of output $P\left(\outputval_{1:N} | \inputval; \theta \right)$ can be decomposed as: $P\left(\outputval_{1:N} | \inputval; \theta \right)=\Pi_{t=1}^{N}P\left(\outputval_t | [\inputval, \outputval_{1:t-1}]; \theta \right)$,
where $\outputval_{1:t-1}$ is the output up to position $t-1$, $\theta$ is the model parameter, and $\inputval$ is the prompt (treated as fixed).
Each output sample is generated via token-by-token sampling, and the generation of multiple samples naturally forms a search tree~\citep{yao2023tree, hao2023reasoning, wan2024alphazero}.
Modern LLMs go through multiple training stages. 
In this paper, we would use \textit{base models} to refer to the models trained without \textit{alignment tuning}  techniques~\citep{touvron2023llama}, including instruction tuning and Reinforcement Learning from Human Feedback (RLHF)~\citep{ouyang2022training, bai2022training} (e.g., ``Llama-2-13B''~\citep{touvron2023llama}) and \textit{aligned models} to refer to models undergoing these additional fine-tuning stages (e.g., ``Llama-2-13B-Chat''). 

\shortparagraph{LLM Decoding and Entropy.} Though LLMs are trained with a large vocabulary size $|V|$, the desired tokens often concentrate on a much smaller set of tokens under distribution $P(\outputval_{t} | \inputval, \outputval_{1:t-1}; \theta)$. Common decoding methods~\citep{Holtzman2020The, hewitt2022truncation}
utilize this observation and propose various heuristics to  truncate vocabulary $V$ as  $V_t$ at each step $t$. The next token is then sampled from the renormalized distribution $\tilde{P}\left(\outputval_t | [\inputval, \outputval_{1:t-1}]; \theta \right) = 
         \mathbbm{1}(\outputval_{t} \in V_t) \frac{P(\outputval_{t} | \inputval, \outputval_{1:t-1}; \theta)}{\sum_{\outputval_{t} \in V_t} P(\outputval_{t} | \inputval, \outputval_{1:t-1}; \theta) } $.
Since tokens are sampled from the truncated distribution $\tilde{P}$,\footnote{Our main experiments employ mild decoding settings ($T$=1.0, $p$=0.9). These settings approximate the full distribution, align with standard evaluation practices, and ensure coherent generation from base models. Stronger truncation settings are explicitly noted where applied.} we use $\tilde{P}$ to compute the empirical (token-level) \revise{entropy} $\tilde{H}$ for a given prefix instance $\outputval_{1:t-1}$:\footnote{The common convention setting $0\log0 = 0$ for entropy computation is followed. }
\begin{small}
\begin{equation}
    \tilde{H}\left(\outputVar_t | [\inputval, \outputval_{1:t-1}]; \theta \right) 
    =-\sum_{\outputval_t} \tilde{P}\left(\outputval_t | [\inputval, \outputval_{1:t-1}]; \theta \right) \log \tilde{P}\left(\outputval_t | [\inputval, \outputval_{1:t-1}]; \theta \right)
\end{equation}
\end{small}{Note that $\tilde H$ is a \textit{random variable} that depends on the specific realization of the prefix sequence $Y_{1:t-1} = y_{1:t-1}$. The more common notion of \textit{conditional entropy} is thus the expectation of $\tilde H$ over all possible $y_{1:t-1}$:}
\begin{small}
\begin{equation}
    \tilde{H}\left(\outputVar_t | [\inputval, \outputVar_{1:t-1}]; \theta \right) = \E_{\outputval_{1:t-1}}\tilde{H}\left(\outputVar_t | [\inputval, \outputval_{1:t-1}]; \theta \right)
\end{equation}
\end{small}Conventionally, we use uppercase $Y$ to denote the \textit{random variable} for an output and lowercase $y$ for its specific realization. 
Finally, along a realized sequence $y_{1:t-1}$, we define the \textit{realized entropy} as\footnote{For notational brevity, we hereafter omit explicit conditioning on the input $\inputval$ when the context is clear.}
\begin{small}
\begin{equation}
h_{\text{realized}}(\outputval_{1:N}) \defeq \sum_{t=1}^N \tilde{H}(\outputVar_t | \outputval_{1:t-1}; \theta).
\end{equation}
\end{small}\revise{This metric and its sequence-level reductions (e.g., mean pooling) are of profound practical importance: while estimating the full conditional entropy $\tilde H(\outputVar_t | \inputval; \outputVar_{<t})$ requires marginalizing over an exponential space of prefix trajectories, $h_{\text{realized}}$ serves as the widely-adopted tractable proxy for quantifying generation uncertainty~\citep{kuhn2023semantic, farquhar2024detecting}, generation diversity and creativity~\citep{duvsek2020evaluating, west2025base} and controlling exploration in RLVR~\citep{cheng2025reasoning, cui2025entropy, wang2025beyond}.\footnote{See more related work discussions in \cref{sec: related_work}}} 
{Linearity of expectation and the chain rule for entropy then gives us
\begin{small}\begin{equation}
\label{eq:h-realized-unbiaseness}
    \E_{y_{1:N}}\left[h_{\text{realized}}(\outputval_{1:N})\right] = \sum_{t=1}^N \tilde{H}\left(\outputVar_t | [\inputval, \outputVar_{1:t-1}]; \theta \right) = \tilde{H}(Y_{1:N} | x; \theta),
\end{equation}
\end{small}i.e. $h_{\text{realized}}$ is an unbiased estimator of the marginal entropy of the whole generative process.

\shortparagraph{A Note on Practical Text Generation.}
Throughout this paper, we model an LLM as generating sequences of a fixed maximum length $N$, with realizations denoted by $y_{1:N}$, and some theoretical results consider the asymptotic regime $N \to \infty$. In practice, however, generation often terminates early, producing a shorter sequence $y_{1:n}$ with $n < N$. Our framework accommodates this by defining an \emph{equivalent sequence} $\hat y_{1:N}$ such that $\hat y_t = y_t$ for $t \le n$, and $\hat y_{n+1:N}$ consists of repeated special tokens (e.g., EOS) indicating termination. We further assume
$\tilde P\bigl(Y_{n+1:N} = \hat y_{n+1:N} \mid [x, y_{1:n}]; \theta\bigr) = 1,$
which can be viewed as a property of pretrained LLMs. Under this convention, the probability of the equivalent sequence satisfies $\tilde P(\hat y_{1:N} \mid x; \theta) = \tilde P(y_{1:n} \mid x; \theta)$. Consequently, without loss of generality, we may treat all generations as having length $N$.
}

%% file: past_papers/branching_factor/bf_formulation_theory.tex
\section{Measuring LLM Branching Factor}
\label{sec: bf_as_complexity}

\shortparagraph{Probability Concentration and the Branching Factor.} The generative process of language models can be viewed as moving down a branching tree, with each token choice selecting a path forward. While the theoretical search space spans $O(|V|^N)$ sequences for vocabulary size $|V|$ and fixed sequence length $N$, LLMs concentrate the vast majority of probability mass on a much smaller subset of trajectories~\citep{Holtzman2020The, hewitt2022truncation}. This high-probability subset forms a complex, sparse ``effective tree'' $\mathcal{T}$.

To quantify the size of this effective tree, we utilize the concept of exponentiated entropy (perplexity). We define the effective set size $|\mathcal{T}|$ as:
\begin{equation}
    |\mathcal{T}| \stackrel{\text{def}}{=} \exp \left( \tilde H(Y_{1:N}|x; \theta) \right).
\end{equation}
Information-theoretically, $|\mathcal{T}|$ reflects the size of a uniform distribution (a ``fair die'') that would possess the same total uncertainty (entropy) as the model's actual complex distribution over sequences of length $N$~\citep{brendan2013perplexity}.

\shortparagraph{Defining Branching Factor via Balanced Tree Model.} 
Because the exact topology of the effective tree $\mathcal{T}$ is irregular and intractable, we map it to an \textit{equivalent balanced $B$-ary tree} of the same depth $N$. A perfectly balanced tree with constant branching factor $B$ and depth $N$ contains $B^N$ leaf nodes. By equating this theoretical leaf count to the effective set size ($B^N = |\mathcal{T}|$), we derive the \textbf{Branching Factor (BF)} as the geometric mean of the branching width:
\begin{equation}
    B \equiv B(x; \theta) \stackrel{\text{def}}{=} |\mathcal{T}|^{1/N} = \exp \left( \frac{1}{N} \tilde H(Y_{1:N}|x; \theta) \right) = \exp \left( \bar{H}(Y_{1:N}|x; \theta) \right)
\end{equation}
where $\bar{H}(Y_{1:N}|x; \theta)$ denotes the length-averaged marginal entropy of the sequence. $B(x; \theta)$ thus provides a normalized, token-invariant metric: it quantifies the effective number of plausible next-token choices available to the model on average at any given step.

\shortparagraph{Linking Entropy and Log-Likelihood in Long Sequences.}
Calculating the exact Branching Factor requires the total entropy $\tilde H(Y_{1:N}|x; \theta)$. As discussed in~\cref{sec: prelim}, computing this quantity directly is intractable due to the exponential number of possible trajectories. A standard Monte Carlo approach uses the \textit{realized entropy} $h_{\text{realized}}(y_{1:N})$ of sampled sequences as a proxy. Since $h_{\text{realized}}$ is an unbiased estimator (Eq.~\ref{eq:h-realized-unbiaseness}), averaging it over sufficient samples converges to the true total entropy.

However, for long sequences, even calculating $h_{\text{realized}}$ becomes computationally prohibitive. It requires a full summation over the vocabulary $V$ at every generation step to compute the local entropy, incurring a total cost of $O(N \cdot |V|)$. In contrast, computing the sequence's negative log-likelihood (NLL) involves only the probabilities of the selected tokens, scaling linearly as $O(N)$. To enable efficient estimation for long horizons, we effectively need a second level of approximation: using the computationally cheap NLL as a proxy for the realized entropy.

Standard Asymptotic Equipartition Property (AEP) theory~\citep{shannon1948mathematical} suggests that for stationary processes, the length-averaged NLL of a typical sequence will converge to the (length-averaged) total entropy. However, LLM generation is neither stationary nor ergodic. Fortunately,~\citet{mudireddy2024slaves} demonstrate that a robust connection persists without these strict assumptions: the NLL converges to the \textit{realized entropy} $h_{\text{realized}}$ instead. We characterize this relationship as follows:\footnote{We provide a simplified proof in \cref{app: aep_proof} with minor changes to the original proof of~\citet{mudireddy2024slaves}. Notably, our goal is only to show the approximation between length-averaged log-likelihood and entropy for a typical sequence, so we do not require stricter assumptions like ergodicity or stationarity.}

\begin{theorem}[Log-Likelihood Convergence for LLMs] Given $0 < \epsilon < 1$, as $N \to \infty$:
\label{thm: aep_llm}
    \begin{equation}
    P\left( \left\lvert -\frac{1}{N}\log \tilde{P}\left(\outputval_{1:N} | \inputval; \theta \right) - \frac{1}{N} h_{\text{realized}}(\outputval_{1:N}) \right\rvert < \epsilon \right) \to 1
    \end{equation}
\end{theorem}
\vspace{-5pt}

\begin{figure*}[t!]
\centering
\begin{subfigure}[t]{0.24\textwidth}
    \centering
    \includegraphics[width=\linewidth]{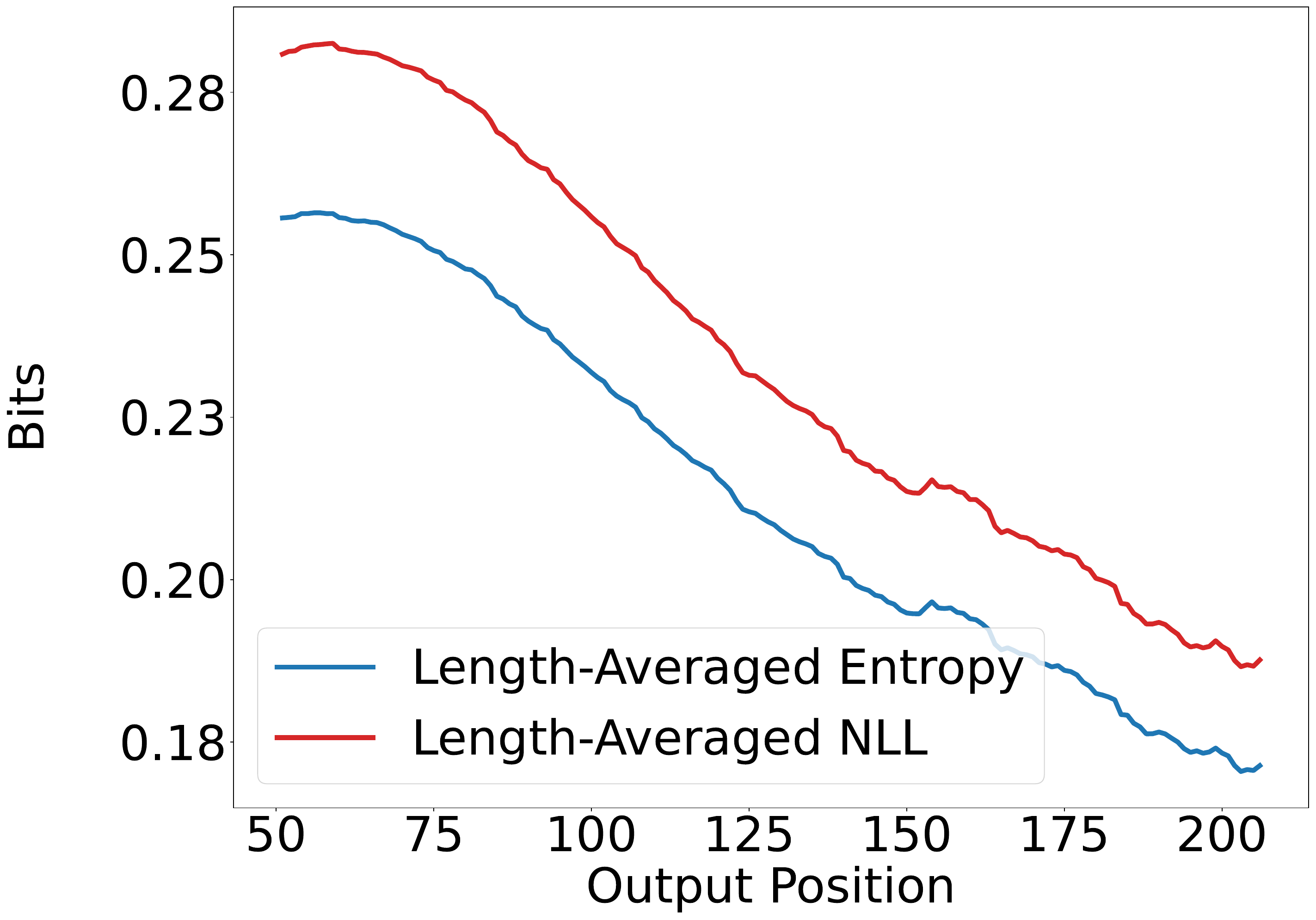}
    \caption*{(a) MMLU}
    \label{fig:loglik-entropy-rate_mmlu}
\end{subfigure}
\begin{subfigure}[t]{0.24\textwidth}
    \centering
    \includegraphics[width=\linewidth]{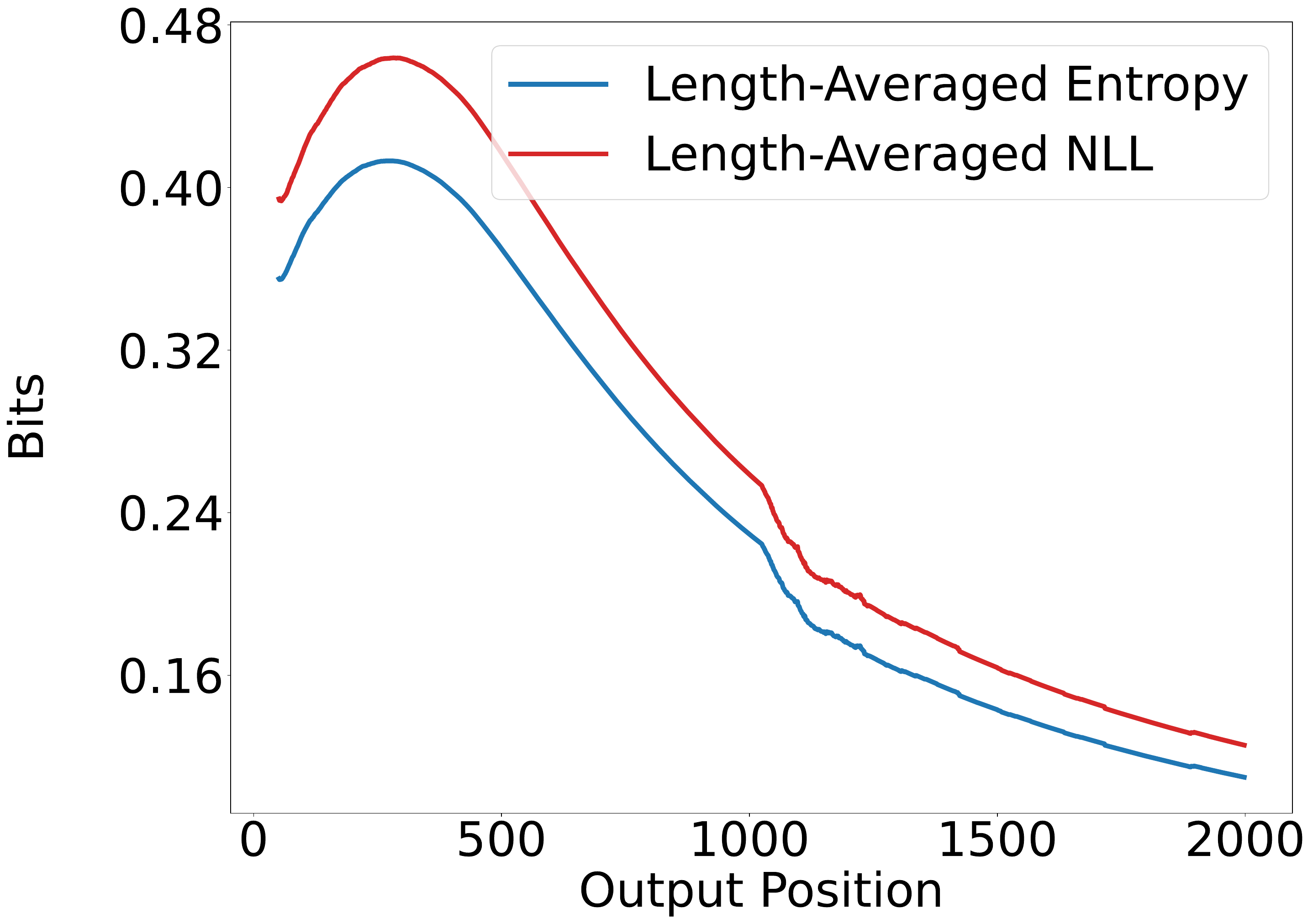}
    \caption*{(b) BBCNewsLatest}
    \label{fig:loglik-entropy-rate_bbcnews}
\end{subfigure}
\hspace{0.01\textwidth}
\hfill
\begin{subfigure}[t]{0.24\textwidth}
    \centering
    \includegraphics[width=\linewidth]{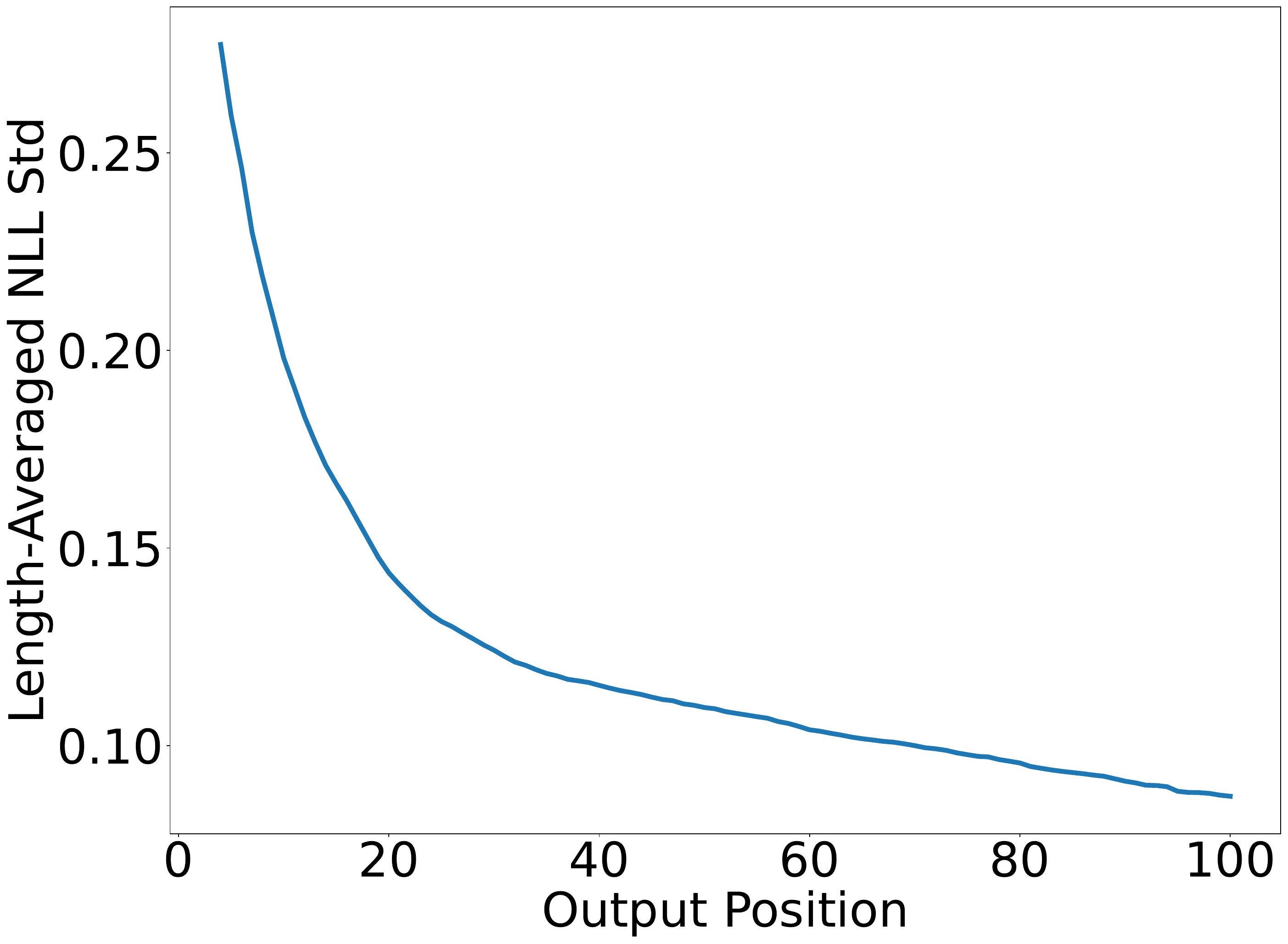}
    \caption*{(c) MMLU (Std)}
    \label{fig:2-c}
\end{subfigure}
\begin{subfigure}[t]{0.24\textwidth}
    \centering
    \includegraphics[width=\linewidth]{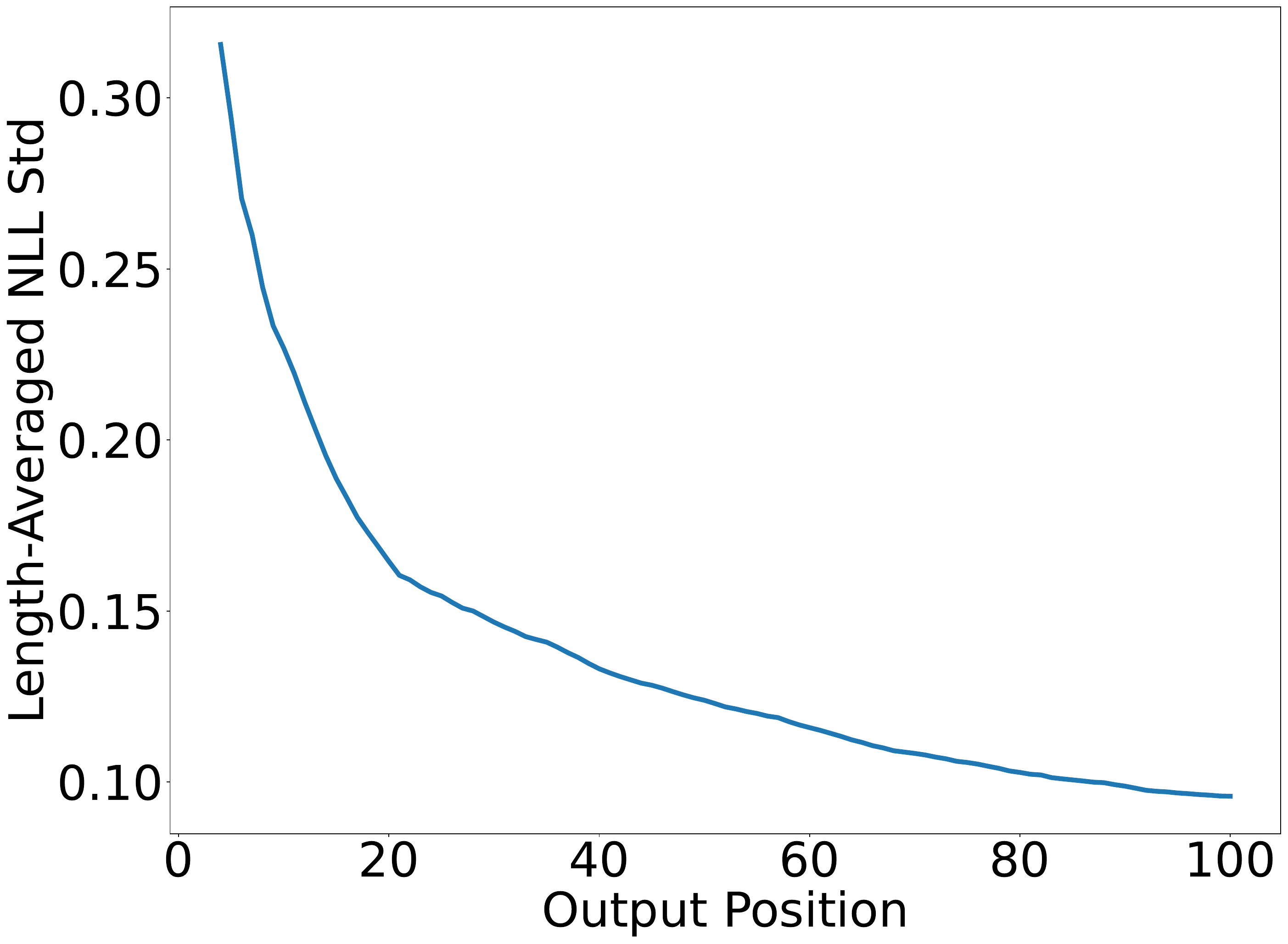}
    \caption*{(d) BBCNewsLatest (Std)}
    \label{fig:2-d}
\end{subfigure}

\caption{
\textbf{Convergence of NLL and Entropy.}
(\textbf{a, b}): The length-averaged NLL closely tracks the length-averaged Entropy. 
(\textbf{c, d}): The standard deviation of the length-averaged NLL diminishes rapidly with output length. 
}
\label{fig:merged_loglik_entropy}
\vspace{-8pt}
\end{figure*}

Since $\mathbb{E}[h_{\text{realized}}] = \tilde H(Y_{1:N}|x; \theta)$, this theorem justifies using the NLL of sampled sequences as a \textit{proxy} for realized entropy in long horizons, provided the variance is low. As an empirical verification for \cref{thm: aep_llm}, we plot NLL and realized entropy for sampled outputs of Llama-3-8B-Instruct over multiple datasets\footnote{For dataset-specific details, we refer readers to \cref{app: dataset_details}.} in \cref{fig:merged_loglik_entropy}. We observe that: as output length increases, the deviation between NLL and realized entropy vanishes, and the standard deviation of the estimator diminishes rapidly (within the first $10$ tokens).

\shortparagraph{Empirical Estimator for Branching Factor.} We now translate the theoretical definition of $B(x; \theta)$ into a practical estimator $\tilde{B}(x; \theta)$. This requires addressing two practical realities: (1) we rely on finite Monte-Carlo samples rather than full distribution access, and (2) while our theory assumes a fixed $N$, generation in practice terminates dynamically upon emitting an EOS token.

We estimate BF using $M$ independent sampled sequences $y^{(1)}, \dots, y^{(M)}$. To handle the computational constraints described above, we adopt a \textit{hybrid estimator}. For short sequences, we compute the exact realized entropy $\tilde{H}(Y_{1:|y|} | x;\theta)$ at every step (since the vocabulary is truncated to $V_t$, this is tractable). To avoid the GPU memory bottleneck of computing the full distribution for long sequences, we approximate entropy using NLL, a substitution justified by \cref{thm: aep_llm}. 
\tmlrrevisee{We explicitly avoid a naive Monte-Carlo entropy estimator on long sequences (i.e., averaging $-\log \tilde P(y_t^{(i)} \mid [x,y^{(i)}_{<t}];\theta)$ over the sampled tokens \emph{without} the inner full-distribution sum), which systematically \emph{underestimates} entropy at finite sample budgets because it misses the long tail of the distribution -- a caveat we quantify in \cref{appendix: under_estimation_of_entropy_via_mc}. The AEP-justified NLL proxy sidesteps this issue by providing a sample-efficient surrogate for the realized entropy.}

We define the estimator $\tilde{B}(x; \theta)$ by averaging over the sampled trajectories:
\begin{equation}
\label{eq:hybrid_estimator}
    \tilde{B}(x; \theta) \approx \exp \left( \frac{1}{M} \sum_{i=1}^{M} \mathcal{E}(y^{(i)}) \right), \quad \mathcal{E}(y) = 
    \begin{cases} 
        \frac{1}{|y|} h_{\text{realized}}(Y_{1:|y|}|x; \theta) & \text{if } |y| < L_\tau \\
        -\frac{1}{|y|} \log \tilde{P}(y|x; \theta) & \text{otherwise}
    \end{cases}
\end{equation}
where $|y|$ is the realized length of the sample (up to EOS) and $L_\tau$ is a threshold length. Note that explicitly normalizing by the realized length $|y|$ adapts our fixed-$N$ theory to variable-length practice, effectively measuring the branching rate per \textit{active} generation step.

Finally, to obtain the task-level Branching Factor, we average over the dataset $X$: $\tilde{B}(X; \theta) = \sum_x p(x) \tilde{B}(x; \theta)$. Unless otherwise specified, BF refers to this dataset-level branching factor in the following sections.

%% file: past_papers/branching_factor/BF_measuring_main.tex
\section{Benchmarking and Attributing  Branch Factors}
\label{sec: bf_measure}

\shortparagraph{Models and Sampling.} We run experiments on models from  Llama-2~\citep{touvron2023llama} and Llama-3~\citep{dubey2024llama} families as they are widely-used open-weight model families. 
For each model family, we include both base and aligned models to investigate how alignment tuning affects BF. 
We set $p$=0.9 and $T$=1.0
to sample outputs to conform with the setting for most datasets. 

We set $M$=50 sequences to estimate BF, which yields a reliable estimation across datasets in prior studies. For aligned models, we apply the official chat templates to prompts. In addition, we carefully control the lengths of all inputs plus outputs to be within the context window of the models. 

\shortparagraph{Tasks.}
We consider a variety of tasks covering common application scenarios of LLM generation, including reasoning and open-ended generation: \textsc{MMLU}~\citep{hendrycks2021measuring} (Reasoning), 
\textsc{Cognac}~\citep{chen2022cognac} (Controlled Generation), 
\textsc{BBCLatestNews}~\citep{li2024latesteval} (News Generation), 
and \textsc{Creative StoryGen}~\citep{chakrabarty2024art} (Creative  Generation). To test subjective randomness bias~\citep{bigelowsubjective}, we also prepare a synthetic task \textsc{Random Strings} where the prompt is generated via random characters. See \cref{app: dataset_details} for dataset details. 

\begin{figure}[htbp]
\centering
\resizebox{\textwidth}{!}{%
    \begin{tabular}{cccl}
    \begin{subfigure}[t]{0.25\textwidth}
    \centering
     \includegraphics[width=\linewidth]{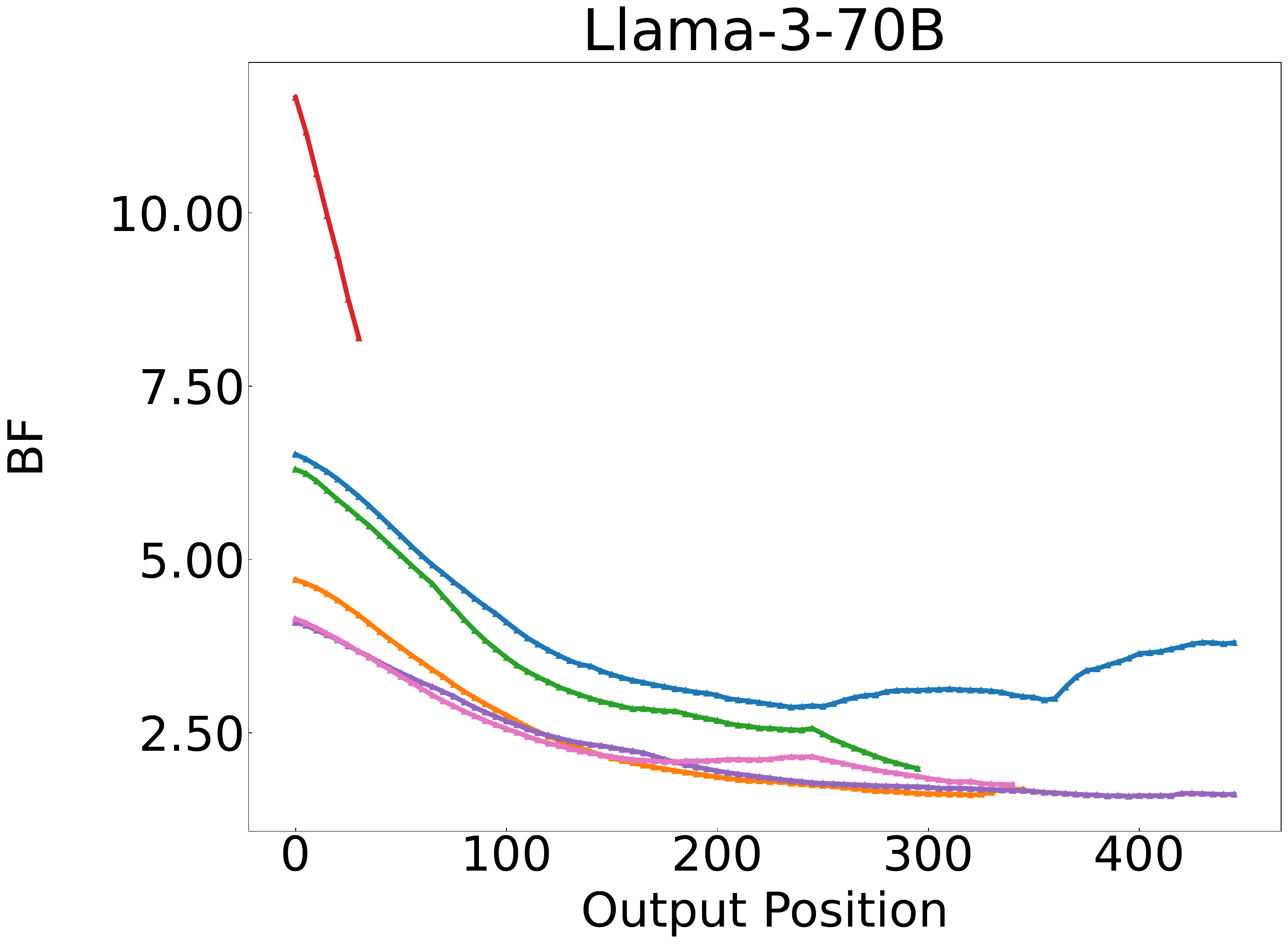}
     \label{fig:output_dynamic_base_storytelling}
    \end{subfigure} &
        \begin{subfigure}[t]{0.25\textwidth}
    \centering
     \includegraphics[width=\linewidth]{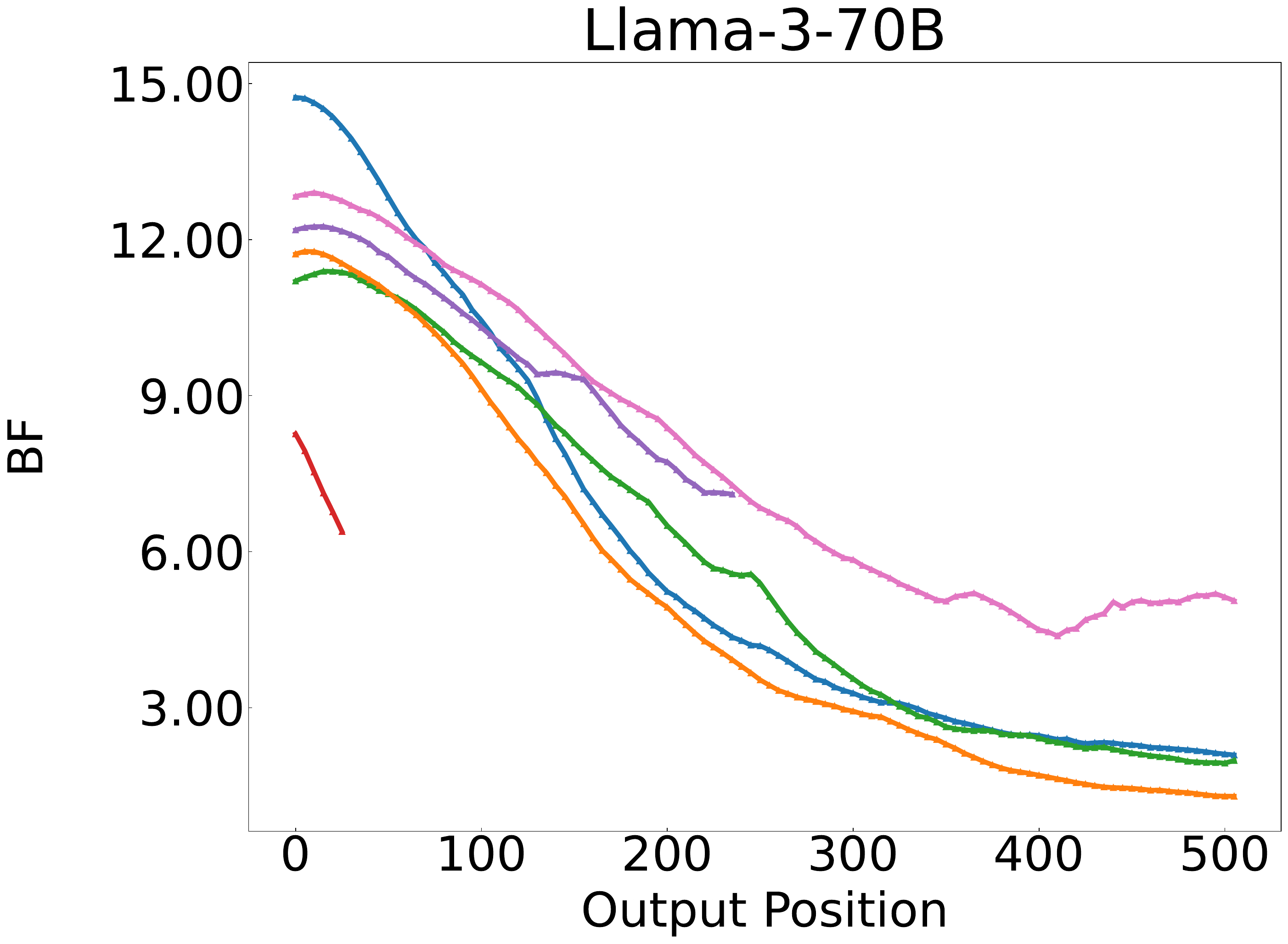}
     \label{fig:output_dynamic_base_cognac_random_str}
    \end{subfigure} & 
    \centering
    \begin{subfigure}[t]{0.25\textwidth}
    \centering
     \includegraphics[width=\linewidth]{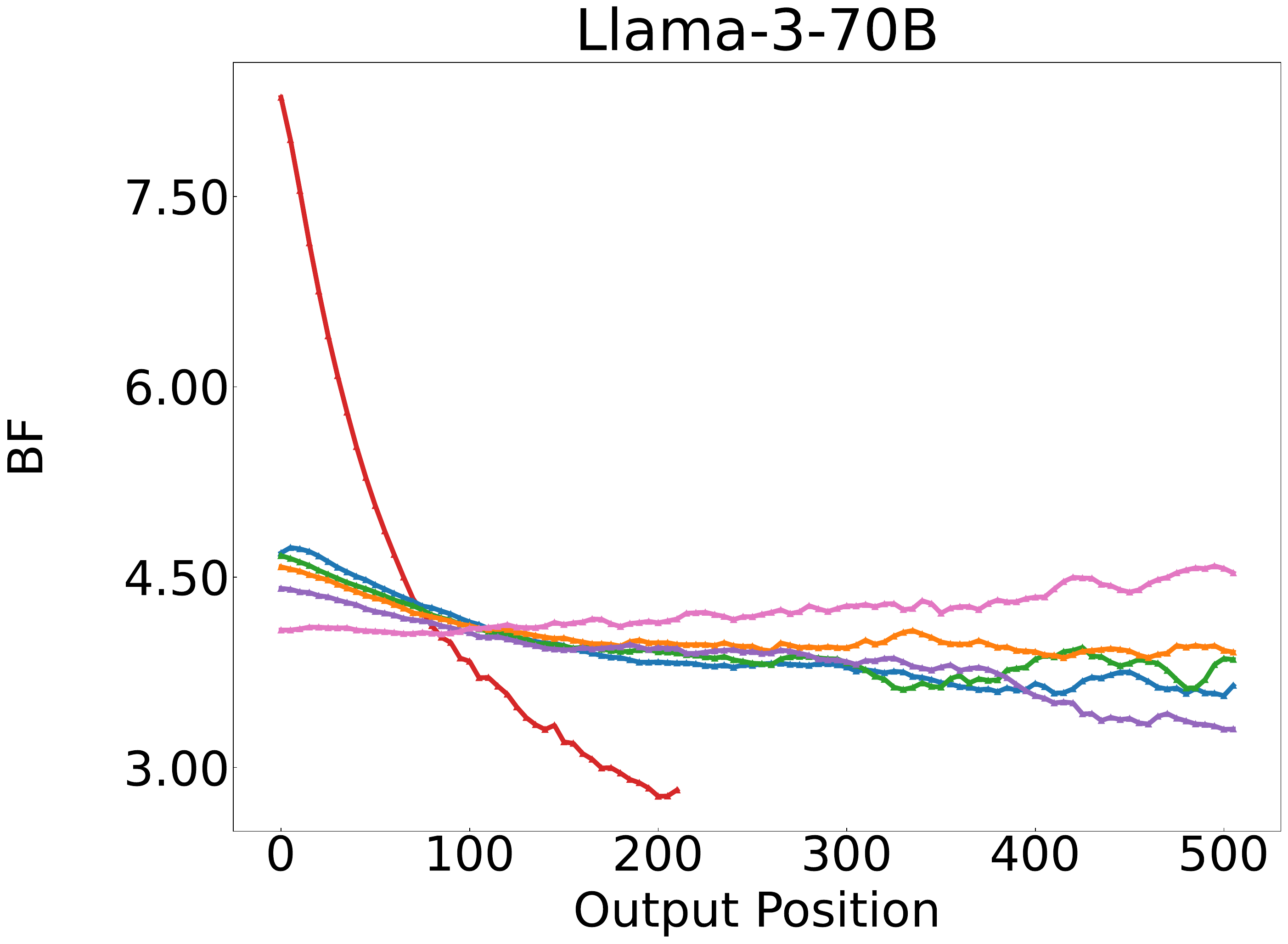}
     \label{fig:output_dynamic_base_bbcnews}
    \end{subfigure} &    \hspace{-15pt} 
    \multirow{2}{*}{
    \begin{minipage}[t]{0.15\textwidth}
        \vspace{-25pt}
    \includegraphics[height=\linewidth]{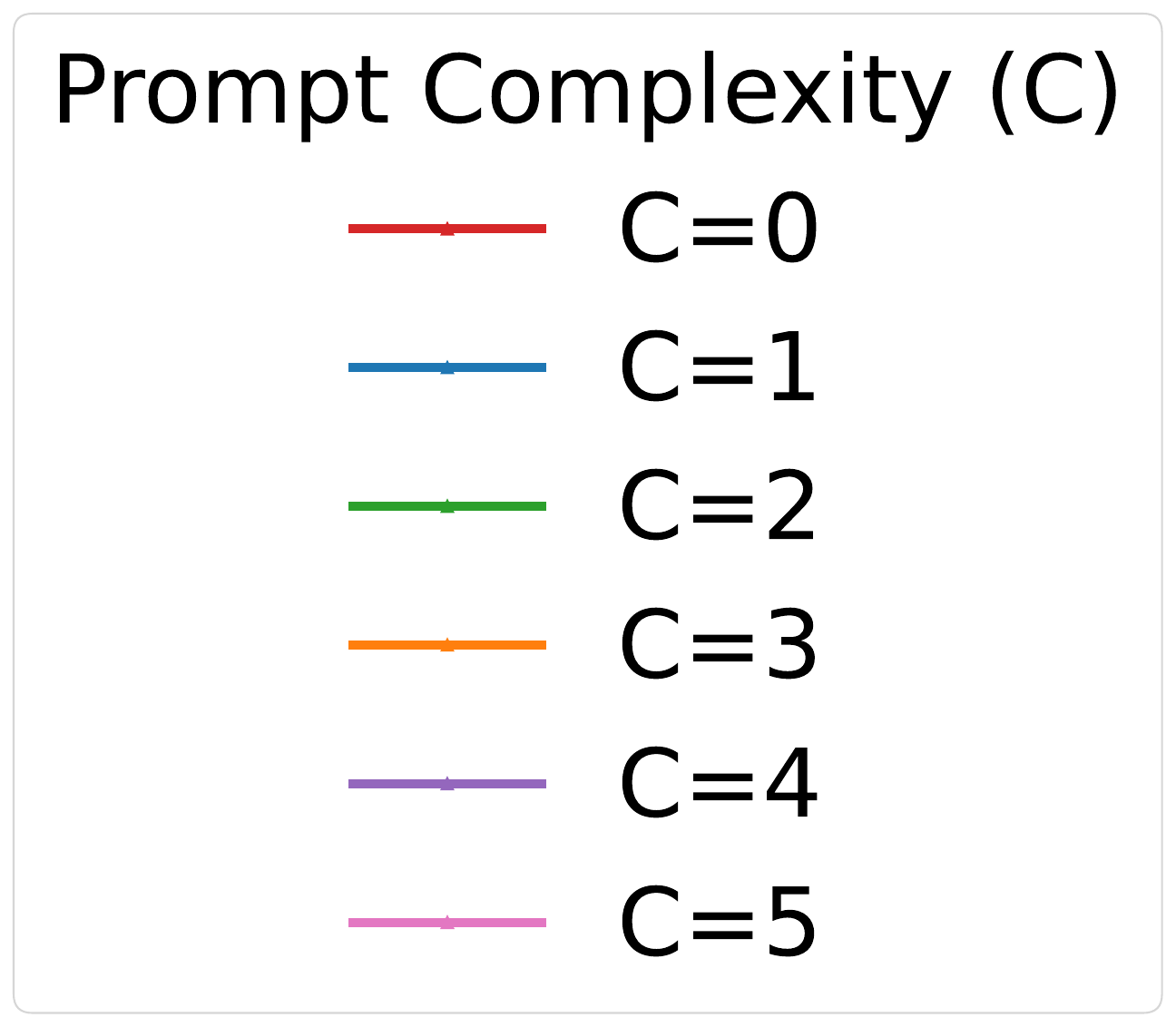}
\end{minipage} } \\
    \begin{subfigure}[t]{0.25\textwidth}
    \centering
     \includegraphics[width=\linewidth]{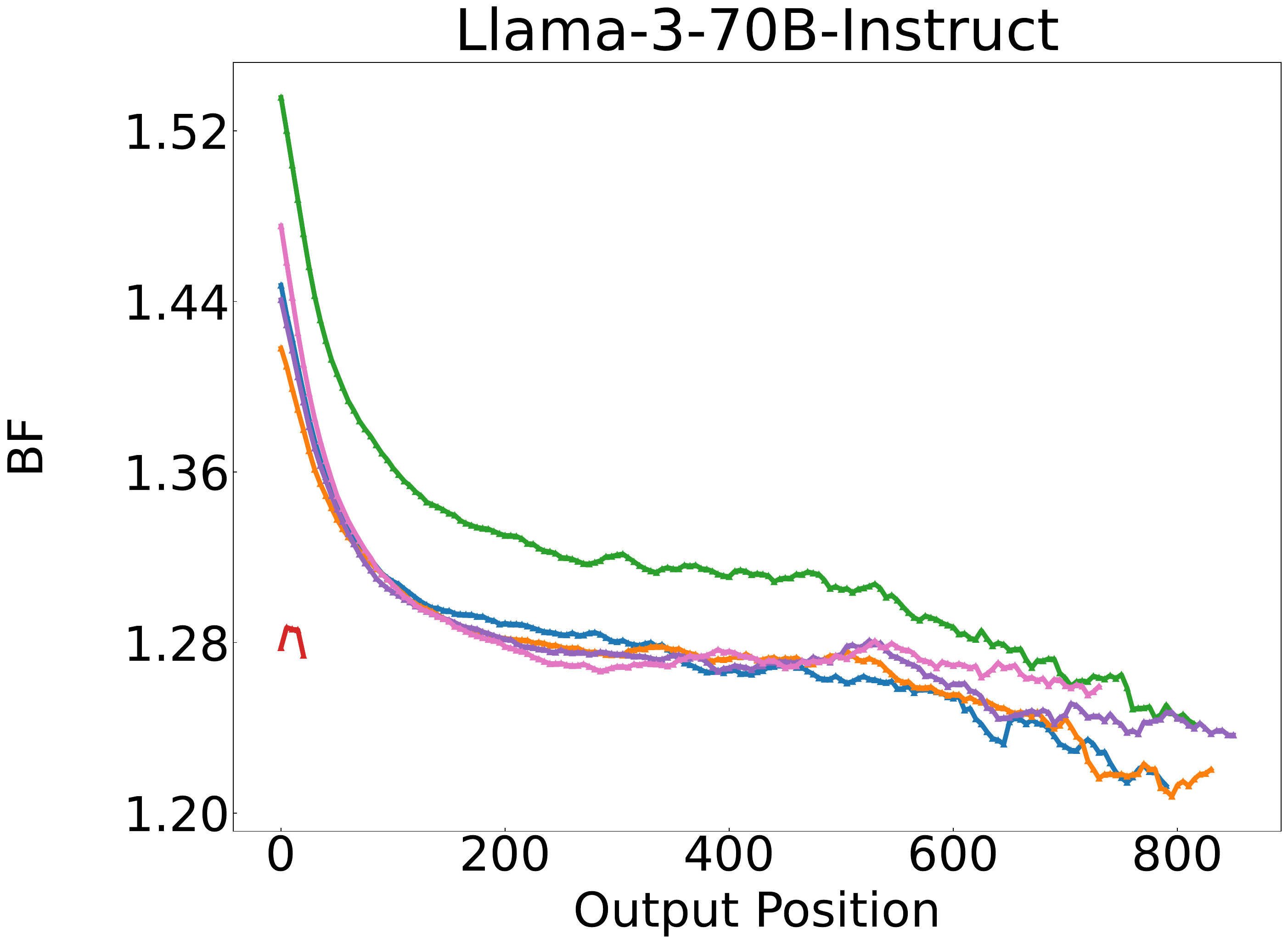}
    \caption{Creative StoryGen}
     \label{fig:output_dynamic_omstrict_storytelling}
    \end{subfigure} &
        \begin{subfigure}[t]{0.25\textwidth}
    \centering
     \includegraphics[width=\linewidth]{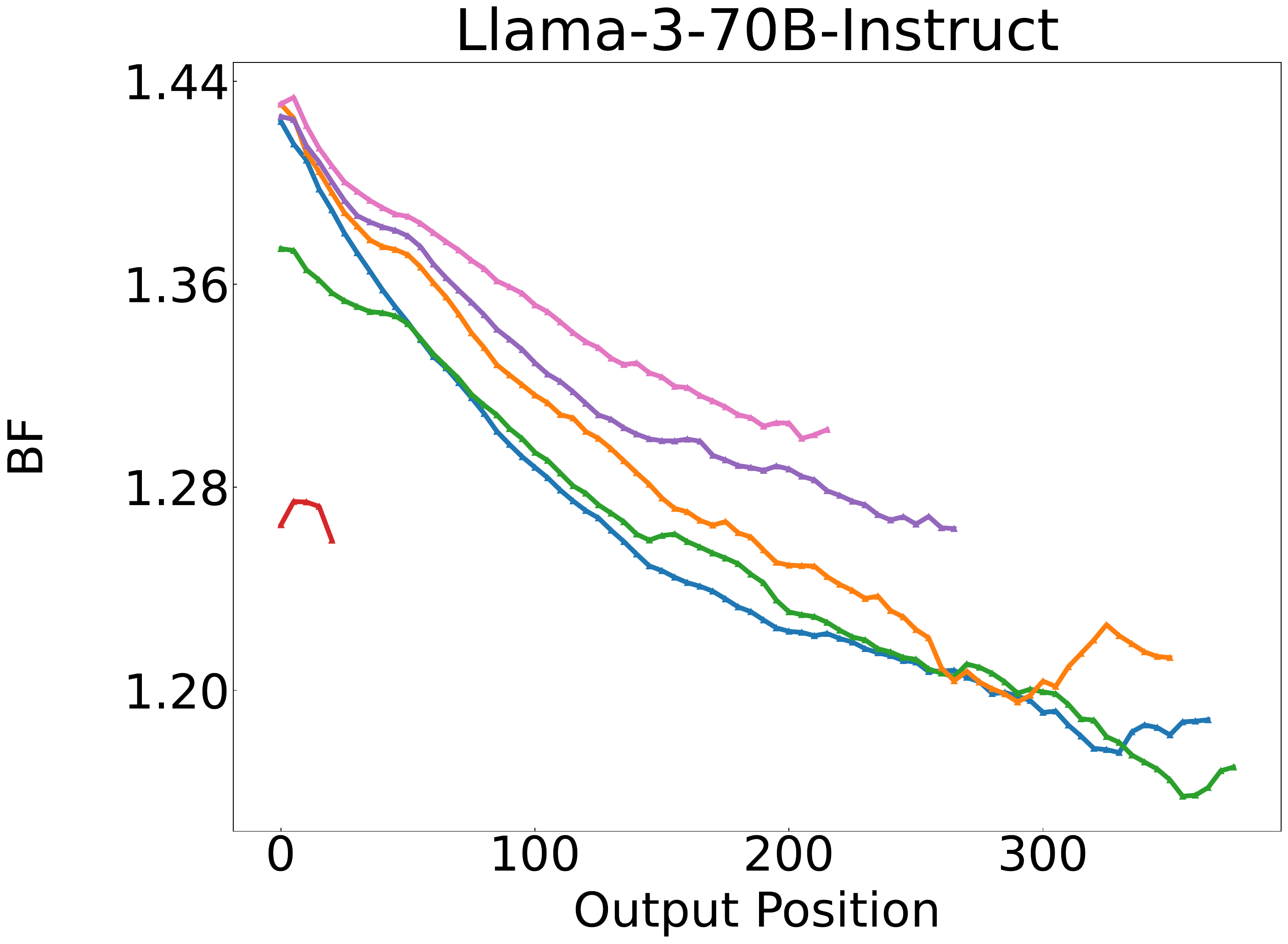}
    \caption{Random Strings}
     \label{fig:output_dynamic_instruct_cognac_random_str}
    \end{subfigure} & 
    \centering
    \begin{subfigure}[t]{0.25\textwidth}
    \centering
     \includegraphics[width=\linewidth]{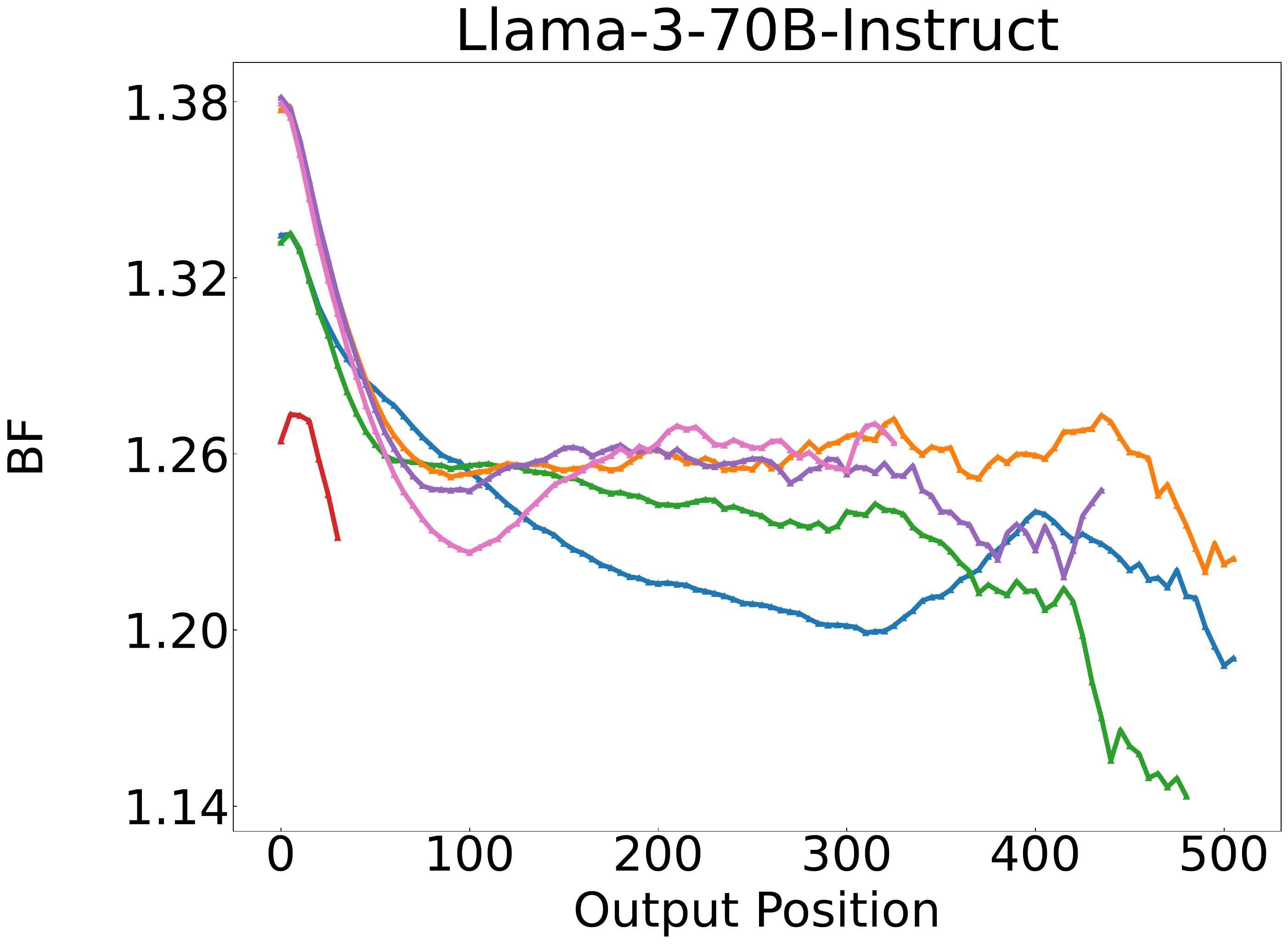}
    \caption{BBCNewsLatest}
     \label{fig:output_dynamic_instruct_bbcnews}
    \end{subfigure} &
    \\
    \end{tabular}%
}
    \caption{
    \textbf{Shrinking BF with output length over various tasks for Llama-3-70B and Llama-3-70B-Instruct.} 
    For better visualization, we compute the exponential moving averaged values of BF with the smoothing factor set as $0.1$. 
    }
    \label{fig: output_dynamic}
\end{figure}

\shortparagraph{Impact Factors (IFs).}
We consider modulating these factors that may impact BF computations: 
\textsc{Prompt Complexity} ($C$), 
\textsc{Alignment Tuning} $(AT \in \{\text{Instruct},\text{Base}\})$, 
\textsc{Model Size} $(S \in \{8\text{B}/13\text{B},70\text{B}\})$, and  
\textsc{Model Generation} $(G \in \{\text{Llama-2},\: \text{Llama-3}\})$. 
$C$ controls the informativeness of the input prompt $\inputval$ (e.g., the number of banned words in Cognac, the number of in-context samples in MMLU). Intuitively, providing more information in $\inputval$ should make the model more confident in its outputs, resulting in a lower BF. Dataset-specific setups for $C$ are detailed in \cref{app: dataset_details}. $AT, S, G$ represent model-wise variations to explore how different configurations of $\theta$ affect $B(\inputVar; \theta)$.

\subsection{BF Dynamic in Generation Process}
\label{sec: bf_dynamic}
Both BF and the output length $N$ are functions of the output $\outputVar$, and BF computation relies on $N$. To avoid confounding effects, we first analyze how BF varies with $N$ before intervening IFs. 
In \cref{fig: output_dynamic}, we demonstrate BF trajectories over different output positions by running Llama-3-70B and Llama-3-70B-Instruct on three representative tasks. Specifically, we compute BF over every five output tokens, conditioning on the prompt and all previously generated output tokens.\footnote{See \cref{app: full_output_bf} for full results across all models and tasks.
}
\revise{Our findings also generalize to summarization, multilingual tasks, OLMo-2~\citep{olmo20242}~/Qwen family~\citep{qwen3technicalreport} (\cref{app: additional_verification}).}

\revise{As we can see, first, \textbf{the base model's BF is often significantly higher than the aligned models, roughly 2--5 times}. The nearly order-of-magnitude difference is also a frequent pattern in strongly aligned models when we compare base and aligned models at the beginning of outputs.}
Therefore, there are actually very few candidate next-token to be truncated in decoding for the aligned models. 
This explains why the decoding methods exert weaker effects for aligned models\revise{~\citep{song2024good, renze2024effect, shi2024thorough}},  as we \revise{will} see in \cref{sec: sampling_efforts}.  
Also, in most cases,  \textbf{BF would often drop smoothly as more output tokens are generated}. 
Under the same task, when $C>0$, different $C$ mainly controls the starting point and the rate of decreasing, while in the end, they would converge to roughly the same point. When almost zero knowledge is provided ($C=0$), the output will end much earlier compared to $C > 0$ cases. 
These findings also provide support that the future token generation is gradually becoming predictable and the model may have a certain generation plan to follow, resonating with recent observations in interpretability~\citep{pal2023future, wu2024language, li2024predicting} and inference acceleration~\citep{cai2024medusa, welleck2024from}.
\mvhnrevise{We emphasize that this downward trend is a robust aggregate tendency of autoregressive self-conditioning, not a monotone token-wise theorem: it appears even for \textsc{Random Strings}, where no semantic structure is available. We disentangle this trend from the separate alignment effect in \cref{sec: bf_self_narrowing}.}

\revisestart
\revise{\shortparagraph{A Matched-Length Control for CoT.} A natural concern is that CoT generations are longer, so their lower BF might simply reflect the general decline of BF over output position. To control for this, we compare BF at fixed truncated lengths on MMLU between reasoning-oriented models (DeepSeek-R1-Distill-Llama-8B/70B) and direct-answer aligned models (Llama-3-8B/70B-Instruct), following the same measurement protocol as above. Each point on the x-axis corresponds to BF computed up to the same truncated length, enabling a fair matched-position comparison.

\begin{figure}[htbp]

    \centering
    \begin{subfigure}[t]{0.4\textwidth}
    \includegraphics[width=\linewidth]{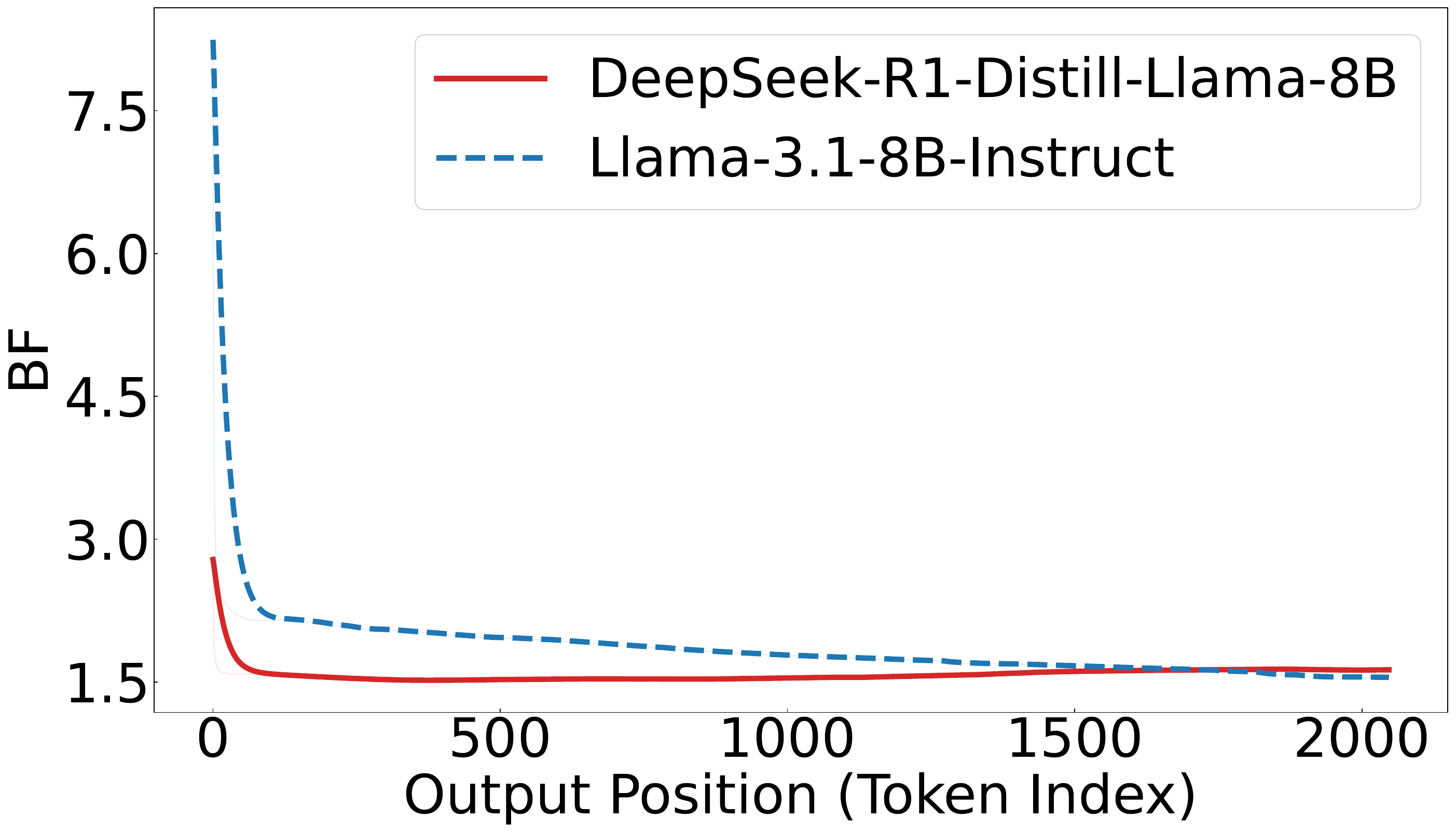}
    \caption{\revisestart Llama-3-8B}
    \end{subfigure}
    \begin{subfigure}[t]{0.4\textwidth}
    \includegraphics[width=\linewidth]{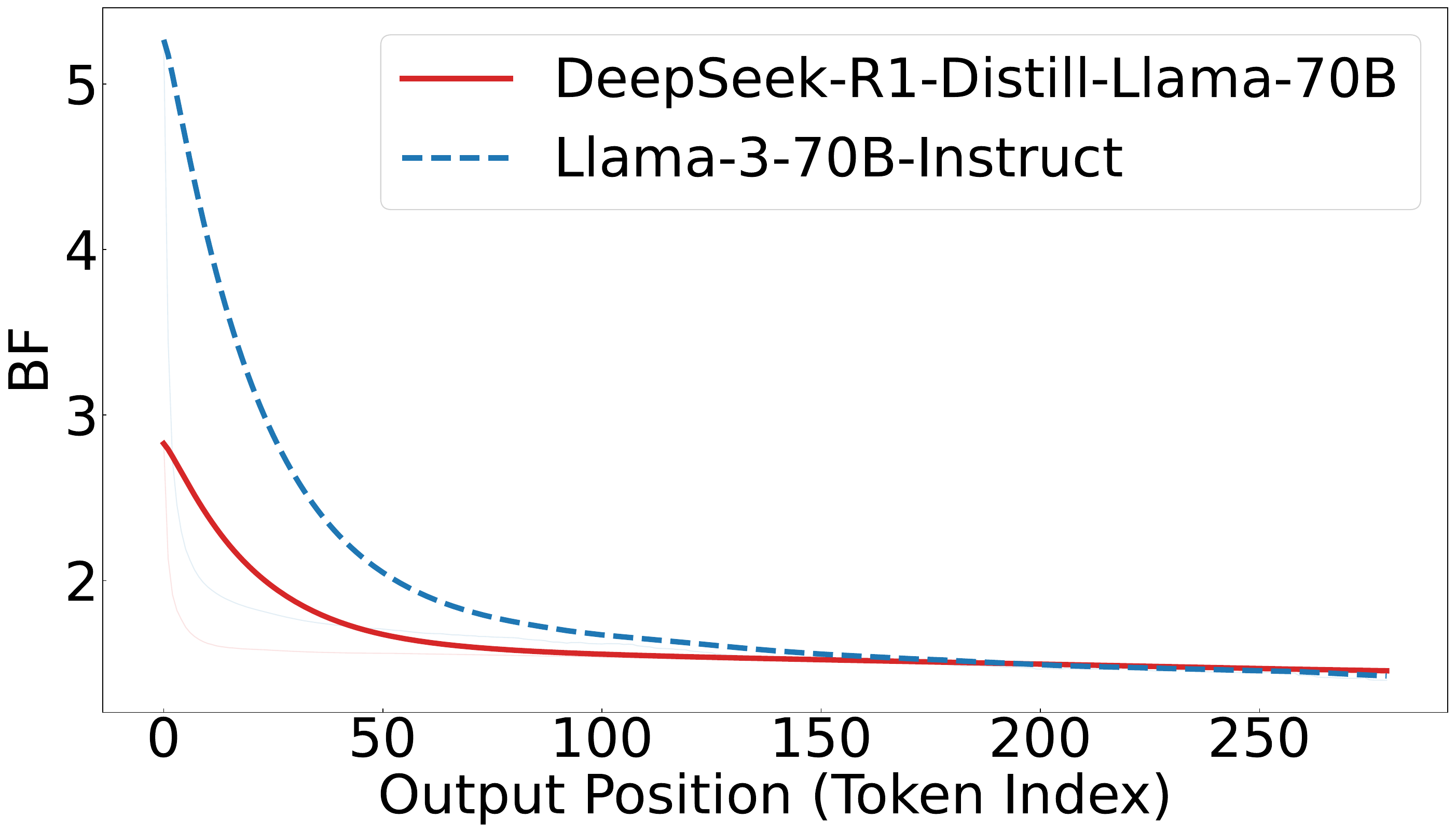}
    \caption{\revisestart Llama-3-70B}
    \end{subfigure}
    \vspace{-5pt}
    \caption{\revisestart \textbf{Matched-length BF comparison on MMLU.} At every truncated length, the reasoning-oriented models (DeepSeek-R1-Distill-Llama) exhibit lower BF than the direct-answer aligned models, including at the earliest matched positions.}
    \label{fig:cot_vs_noncot_bf}
\end{figure}

As shown in \cref{fig:cot_vs_noncot_bf}, the reasoning-oriented models exhibit consistently lower BF across all truncated positions, including the earliest generation stages. This suggests that their lower BF is not merely a byproduct of longer generations drifting into lower-BF regions, and provides a practical check that BF comparisons are robust to this length confound.}
\reviseend
We further examine potential confounds such as prompt likelihood and data contamination in \cref{sec: data_contamination}, and find they do not fully account for the observed BF reductions.  

\subsection{Why Does BF Decrease? Disentangling Self-Narrowing from Alignment}
\label{sec: bf_self_narrowing}
Why does BF decline, and why does it decline even for \textsc{Random Strings}, where no semantic structure exists for alignment, stylistic tokens, or distribution collapse to act on? We argue that two effects have been conflated: \textbf{(i) autoregressive self-narrowing}, a robust aggregate tendency for the next-token distribution to concentrate as the model conditions on its own growing prefix---present in base and aligned models alike, and not requiring meaningful context; and \textbf{(ii) alignment}, which lowers the absolute BF level and steepens the early decline. We do not claim a monotone token-wise law (local increases occur), but a broad empirical regularity consistent with left-to-right training and our AEP analysis (\cref{thm: aep_llm}).

\begin{figure}[htbp]
\centering
\begin{subfigure}[t]{0.32\textwidth}
\centering
\includegraphics[width=\linewidth]{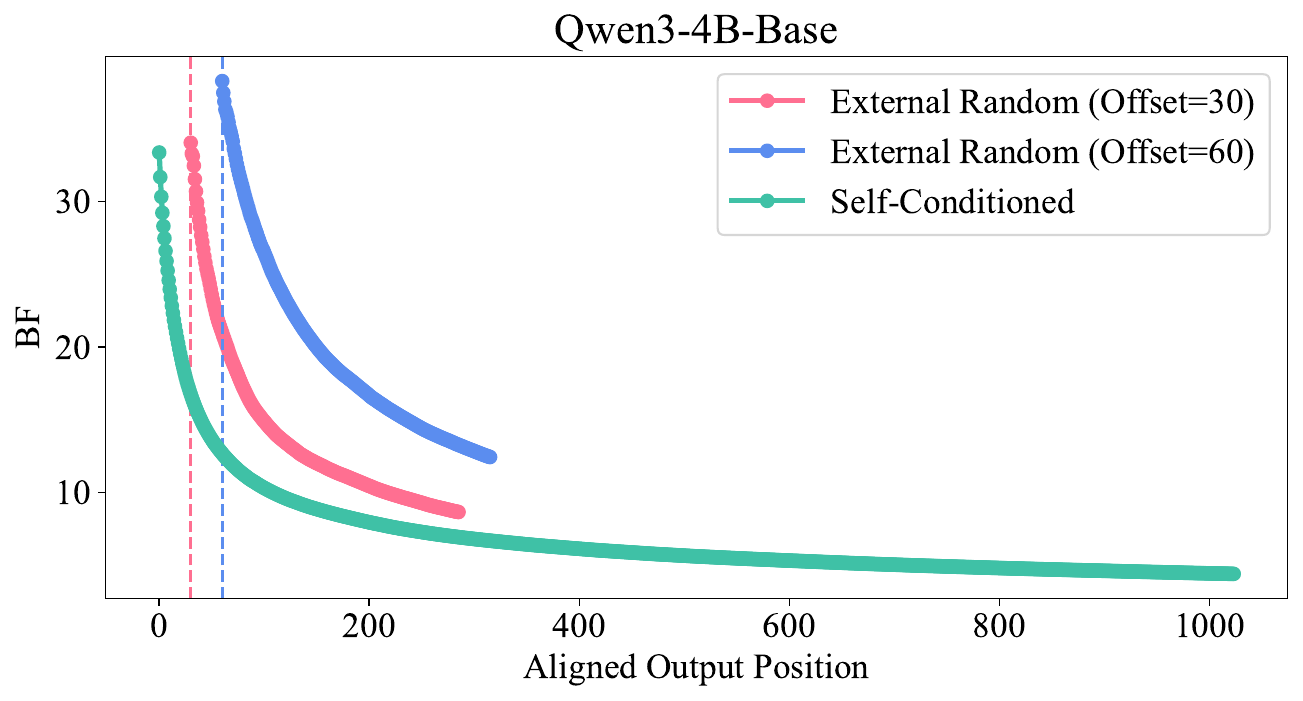}
\caption{Qwen3-4B-Base}
\end{subfigure}
\begin{subfigure}[t]{0.32\textwidth}
\centering
\includegraphics[width=\linewidth]{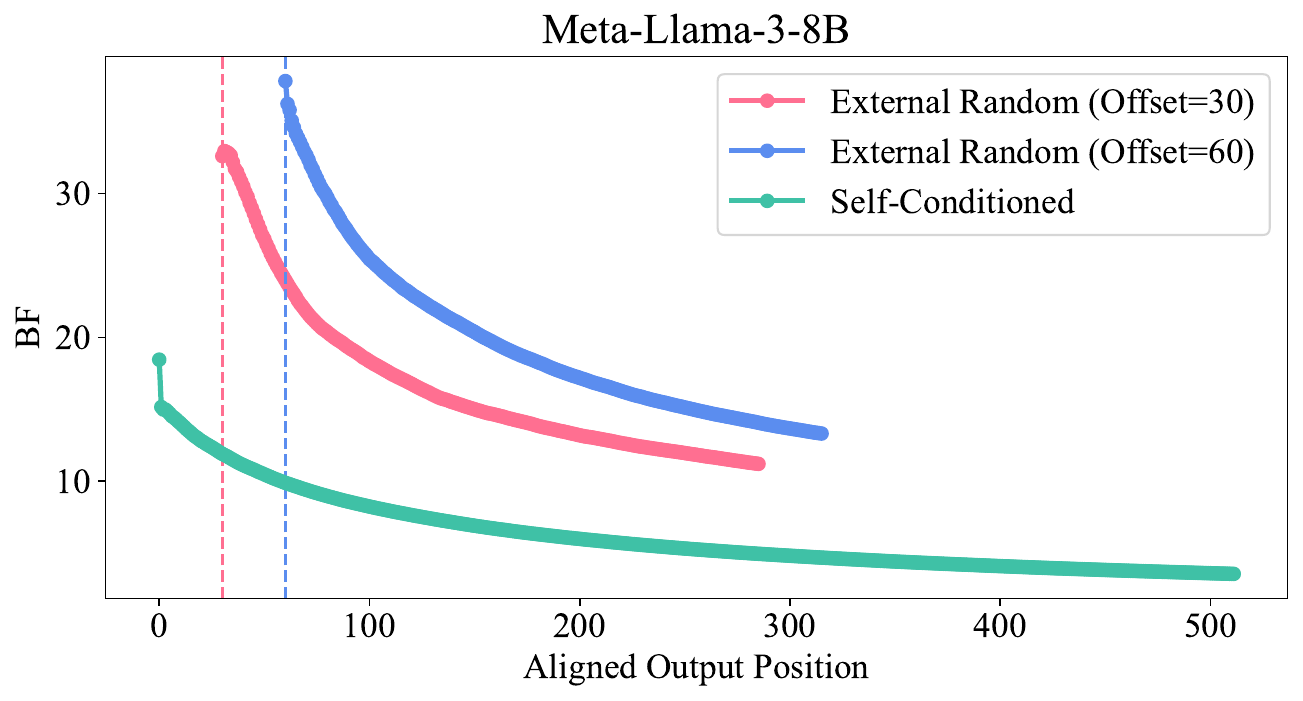}
\caption{Llama-3-8B (Base)}
\end{subfigure}
\begin{subfigure}[t]{0.32\textwidth}
\centering
\includegraphics[width=\linewidth]{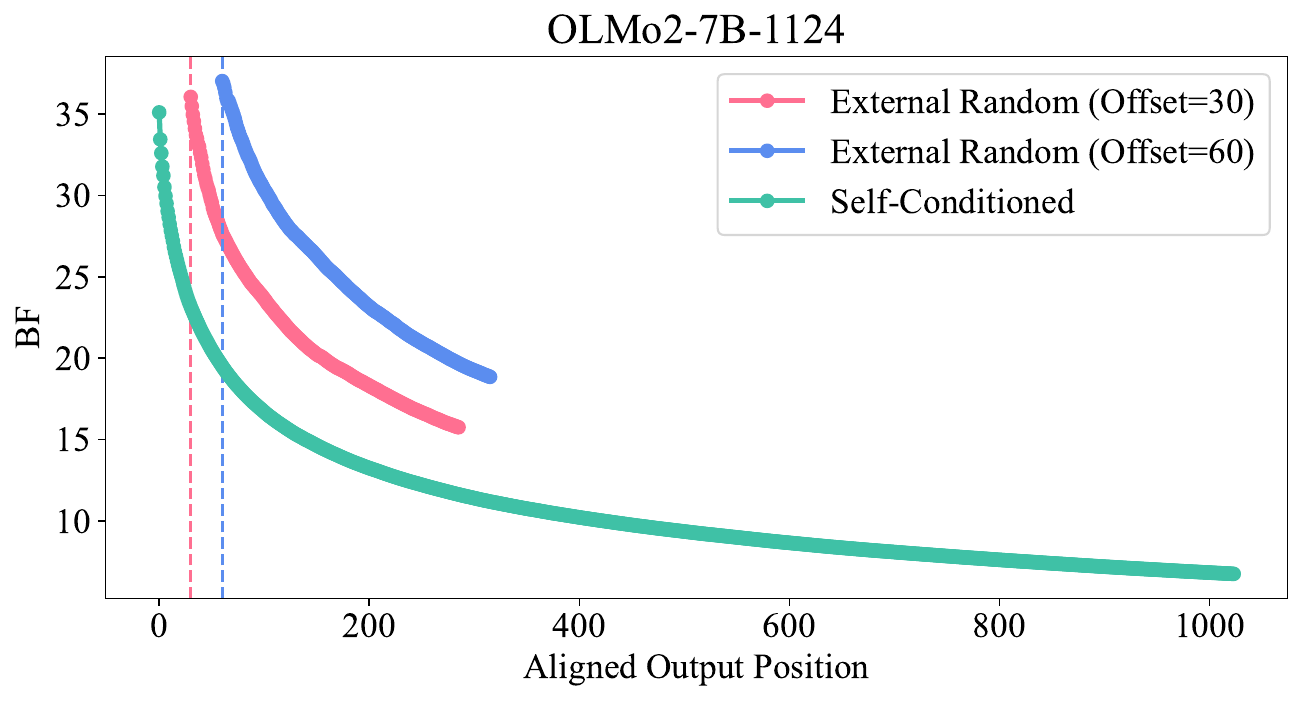}
\caption{OLMo-2-7B (Base)}
\end{subfigure}

\begin{subfigure}[t]{0.32\textwidth}
\centering
\includegraphics[width=\linewidth]{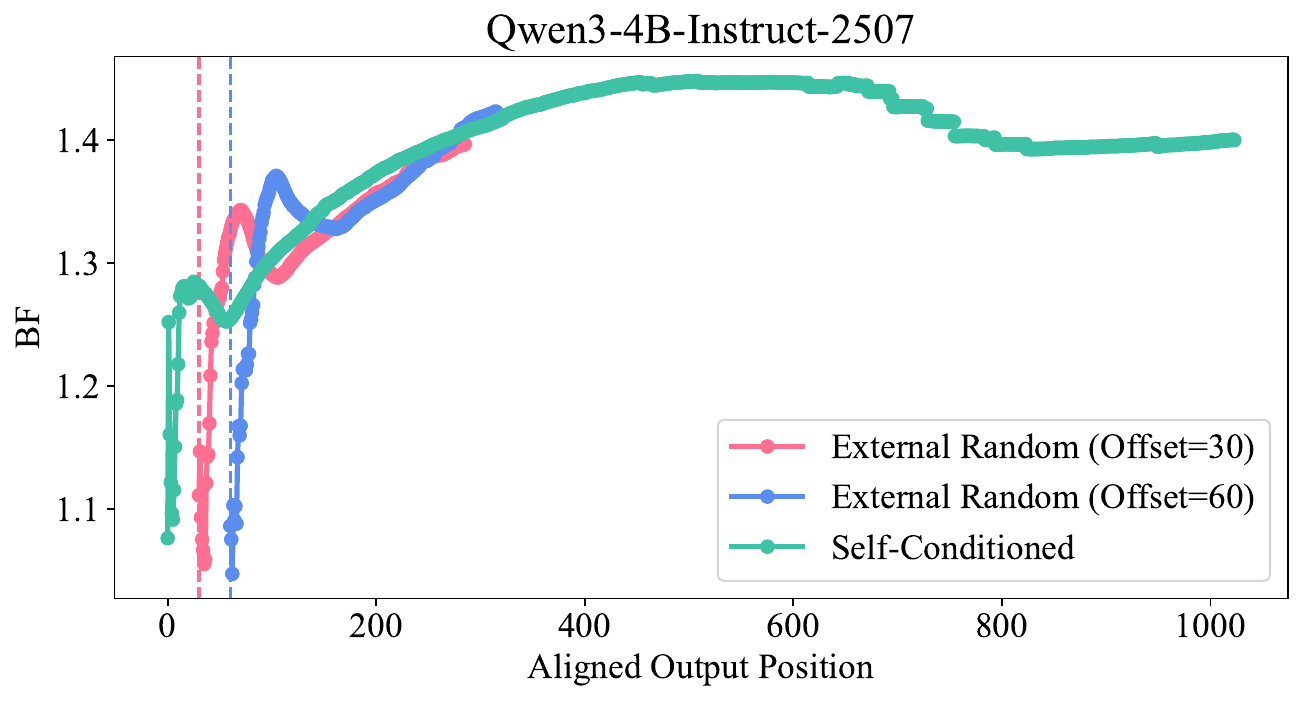}
\caption{Qwen3-4B-Instruct}
\end{subfigure}
\begin{subfigure}[t]{0.32\textwidth}
\centering
\includegraphics[width=\linewidth]{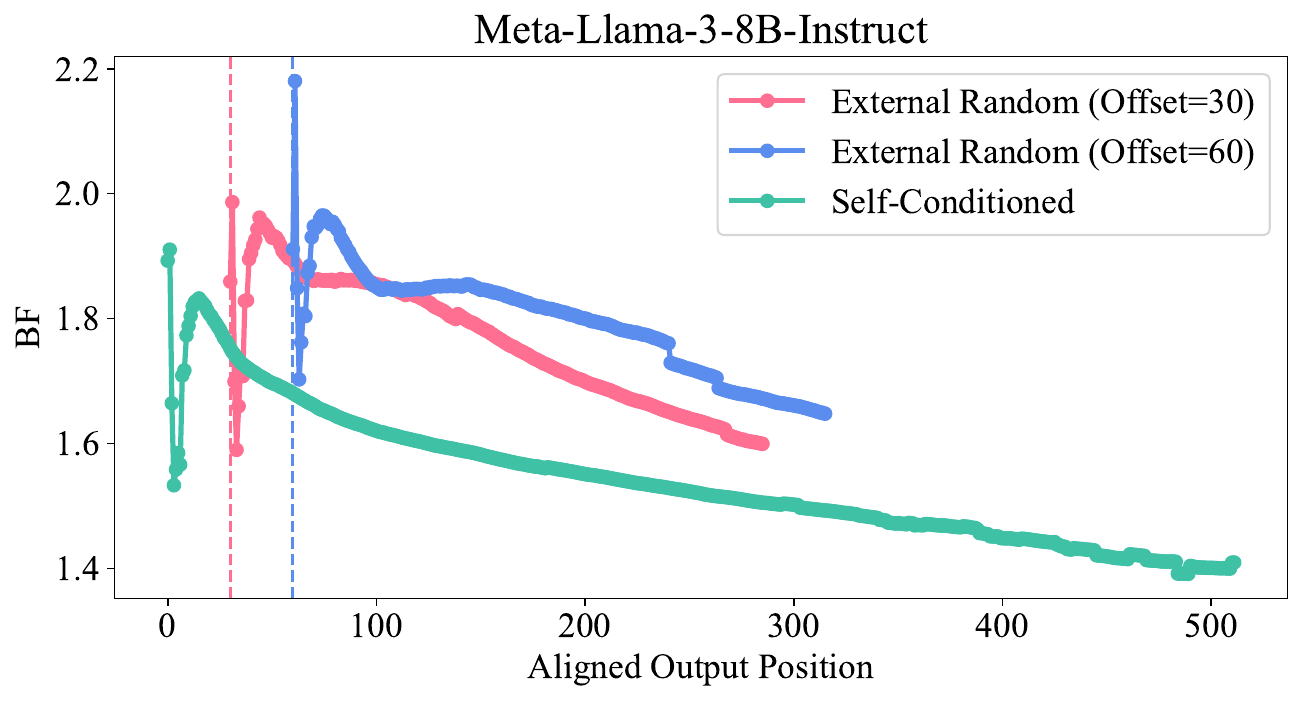}
\caption{Llama-3-8B-Instruct}
\end{subfigure}
\begin{subfigure}[t]{0.32\textwidth}
\centering
\includegraphics[width=\linewidth]{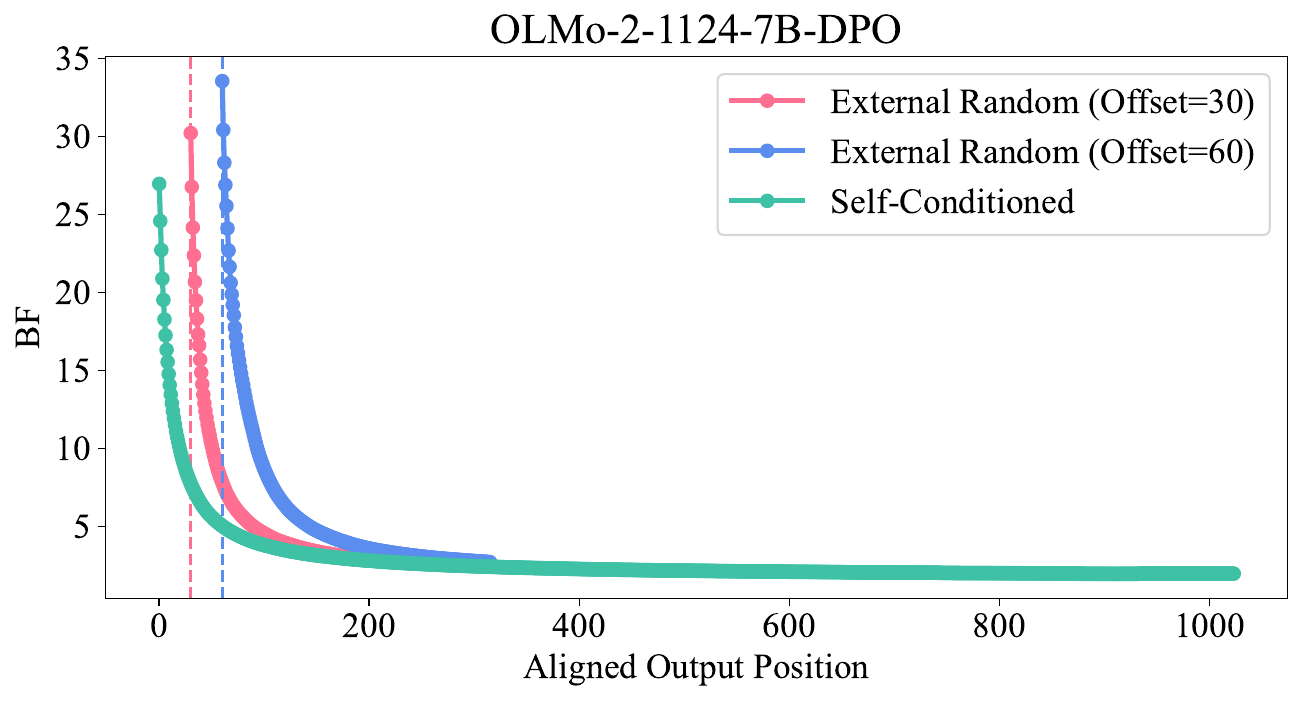}
\caption{OLMo-2-7B-DPO}
\end{subfigure}
\caption{\textbf{Substituting external random tokens for the model's own prefix separates autoregressive self-narrowing from alignment.} \emph{Self-Conditioned} is ordinary generation; \emph{External Random} replaces the prefix with i.i.d.\ random tokens up to a cut point (dashed) and continues, with the $x$-axis aligned to the true generation position. External content surges BF, after which continued autoregression drives it down again, in both Base (top) and Aligned (bottom) models. Full setup, cut points, and the Qwen3-4B-Instruct anomaly are discussed in \cref{app: bf_self_narrowing}.}
\label{fig:bf_self_narrowing_injection}
\vspace{-5pt}
\end{figure}

\shortparagraph{A controlled intervention.} To separate the two, we hold the model fixed and change only the \emph{source} of the prefix: we replace the first $k$ context tokens with an equally long block of externally-sampled i.i.d.\ random tokens and let the model continue, aligning BF to the true generation index. Across Base/Aligned pairs (\cref{fig:bf_self_narrowing_injection}), BF \emph{surges} at the substitution point---out-of-distribution content reopens the consideration set---then \emph{decays again} under continued autoregression. Since only the prefix source changes, this is an intervention, not a correlation: the decline reflects autoregressive self-conditioning rather than an alignment artifact, and BF can be \emph{raised on demand} with unexpected content. Alignment sets the absolute band and re-collapse speed, not the qualitative dynamics.

\begin{figure}[htbp]
\centering
\includegraphics[width=0.55\linewidth]{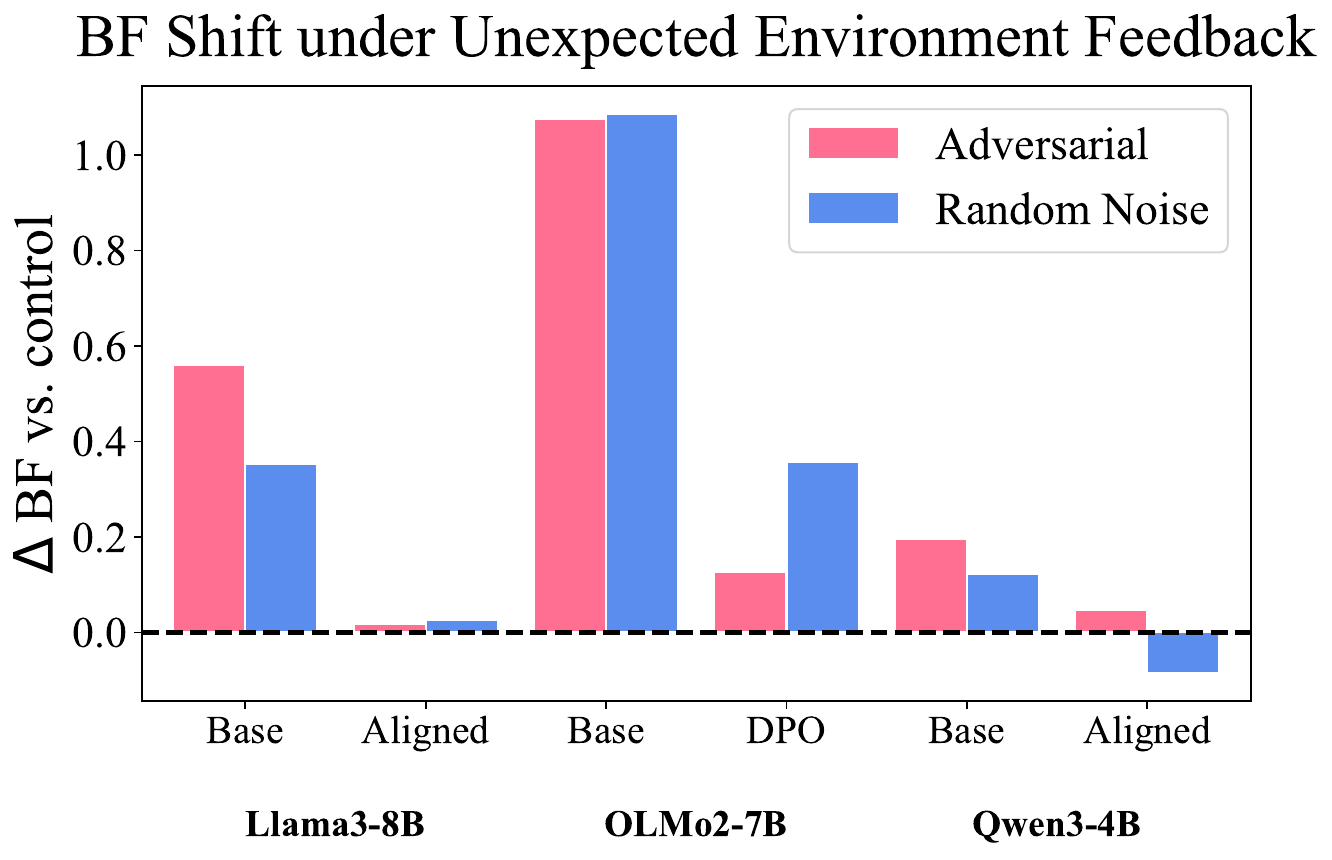}
\vspace{-10pt}
\caption{\textbf{Unexpected environment feedback raises BF.} Change in next-step BF relative to a matched control (progress-as-expected), for adversarial and random-noise feedback in a one-turn agentic setting; the effect shrinks for more heavily aligned models. Setup and per-model results in \cref{app: bf_self_narrowing}.}
\label{fig:bf_agentic_feedback}
\end{figure}

\shortparagraph{Unexpected information raises BF.} The same mechanism holds when new information arrives mid-generation. In a minimal one-step agentic setting, a task, environment state, and partial plan are followed by a single \texttt{Environment Feedback:} message in matched \emph{control} (progress as expected), \emph{adversarial} (an event invalidates part of the plan), or \emph{random-noise} forms. Adversarial and random-noise feedback raise BF relative to the matched control (\cref{fig:bf_agentic_feedback}), most for less-aligned models and compressed for heavily aligned ones. Together with the prompt-complexity/negation case (\cref{app: curious_case_prompt_complexity}), a consistent picture emerges: \emph{unexpected context can raise BF, while continued self-conditioning tends to lower it again}. We do not claim a mechanistic account; deeper explanations (e.g., activation/norm dynamics) are left to future work. Full setups, prompts, per-model results, and the estimator are described in \cref{app: bf_self_narrowing}.

\begin{figure}[htbp]
    \centering
    \begin{subfigure}[c]{0.47\textwidth}
        \includegraphics[width=\linewidth]{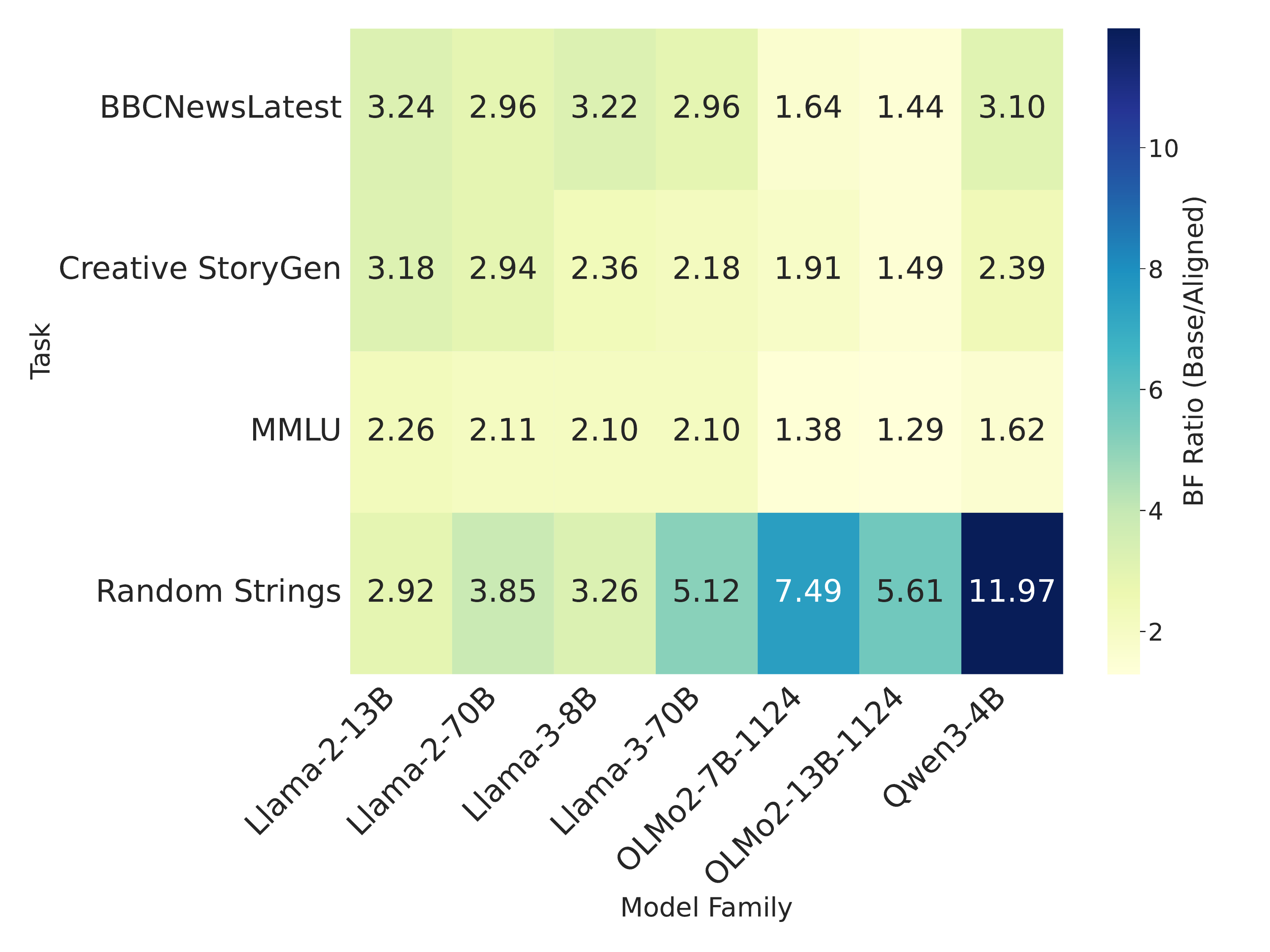}
        \caption{Heatmap of BF Ratio}
        \label{fig:bf_ratio_heatmap}
    \end{subfigure}
    \hfill
    \begin{minipage}[c]{0.52\textwidth}
        \centering
        \begin{subfigure}[t]{0.48\linewidth}
            \includegraphics[width=\linewidth]{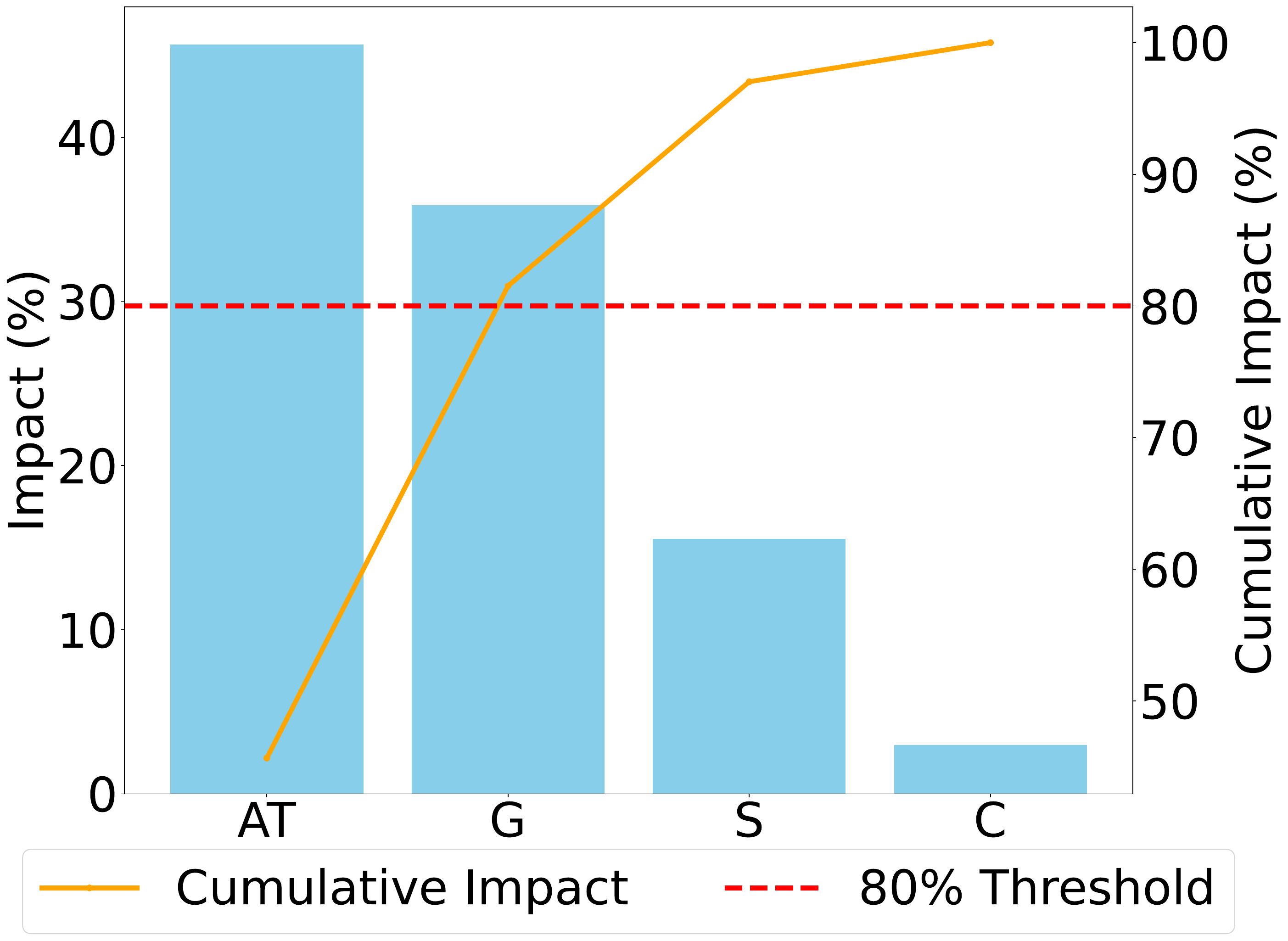}
            \caption{Cognac}
            \label{fig:cognac_pareto}
        \end{subfigure}
        \hfill
        \begin{subfigure}[t]{0.48\linewidth}
            \includegraphics[width=\linewidth]{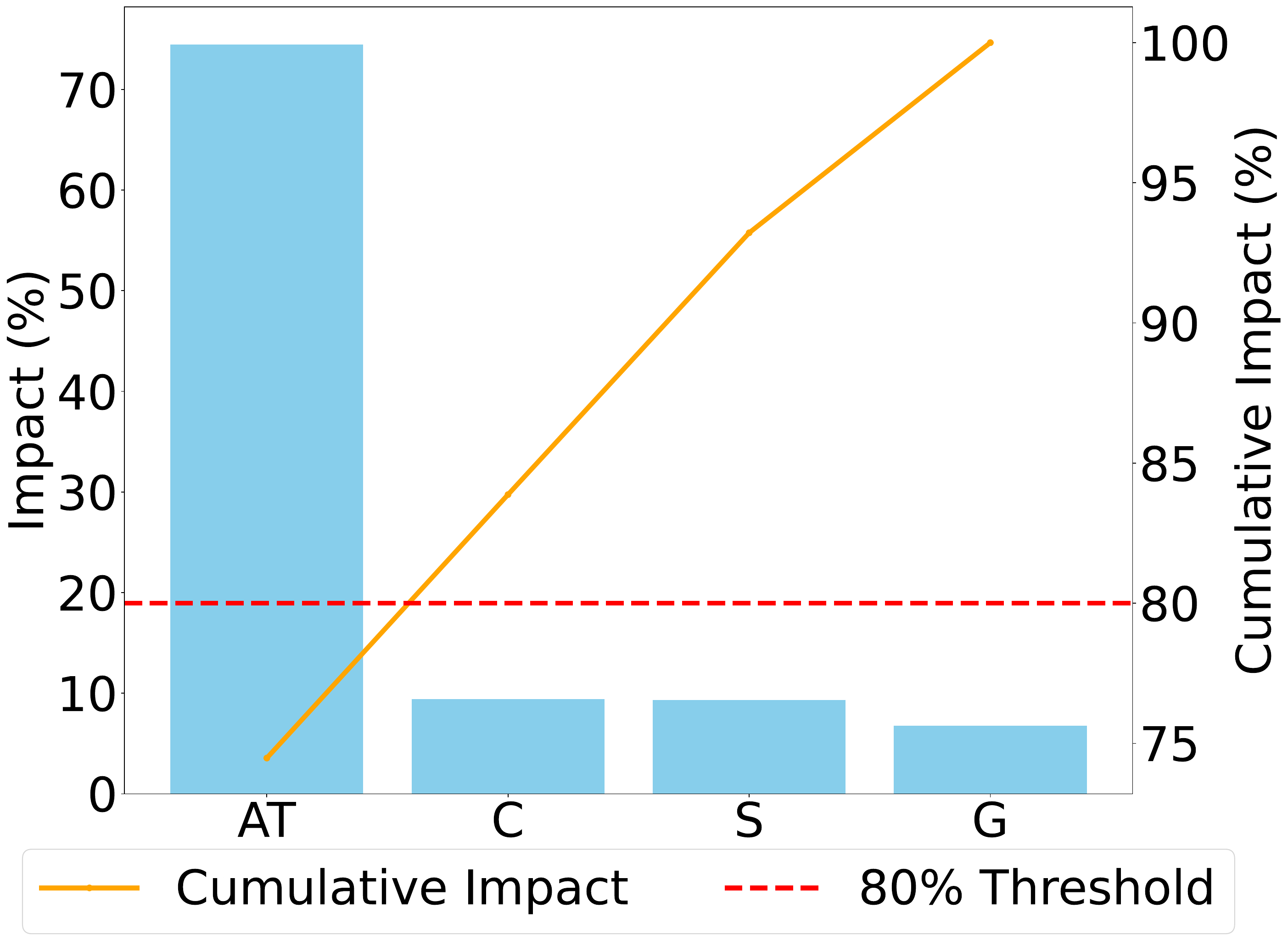}
            \caption{MMLU}
            \label{fig:mmlu_pareto}
        \end{subfigure}

        \vspace{4pt}

        \begin{subfigure}[t]{0.48\linewidth}
            \includegraphics[width=\linewidth]{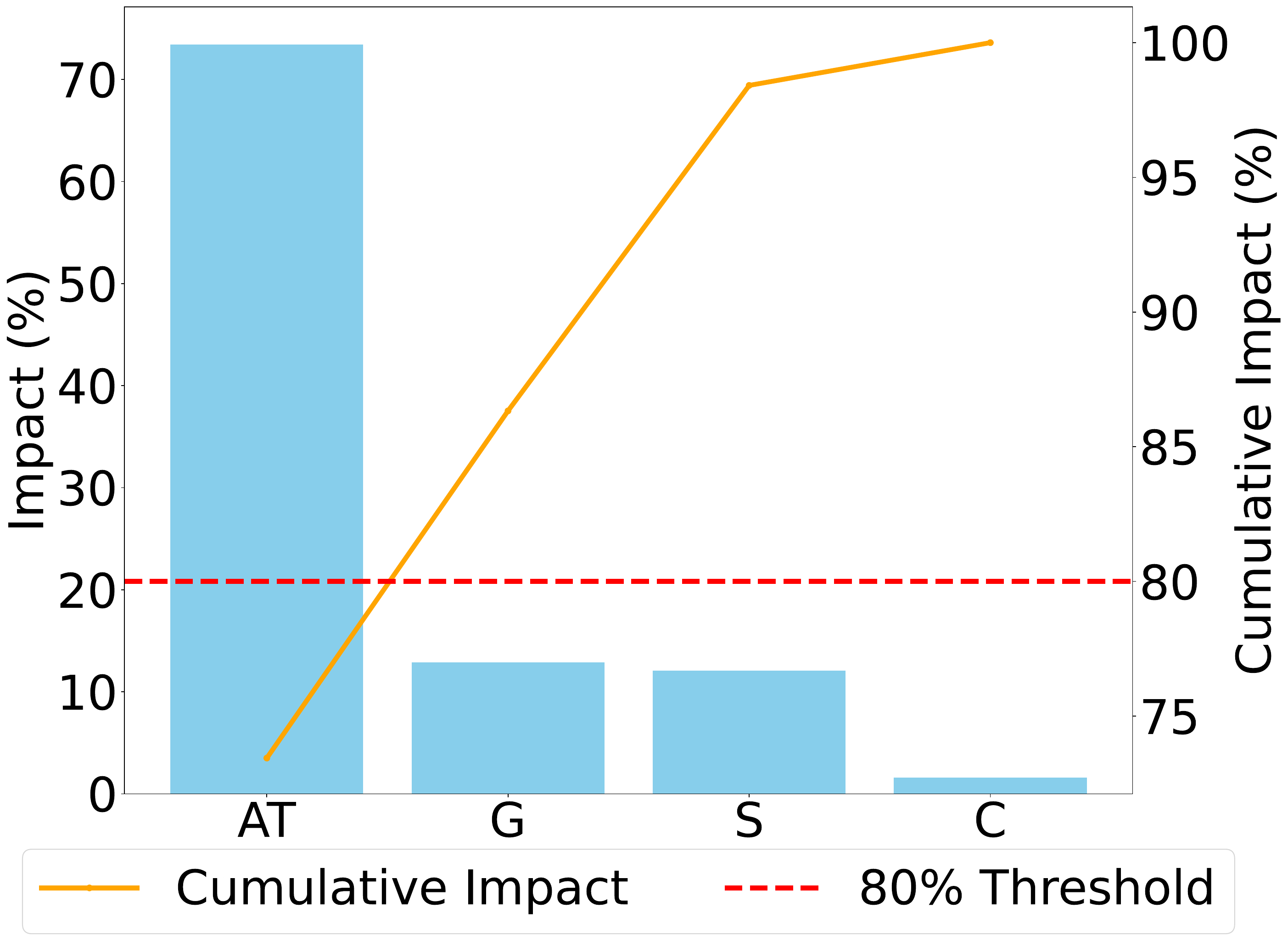}
            \caption{BBCNews}
            \label{fig:bbc_news_pareto}
        \end{subfigure}
        \hfill
        \begin{subfigure}[t]{0.48\linewidth}
            \includegraphics[width=\linewidth]{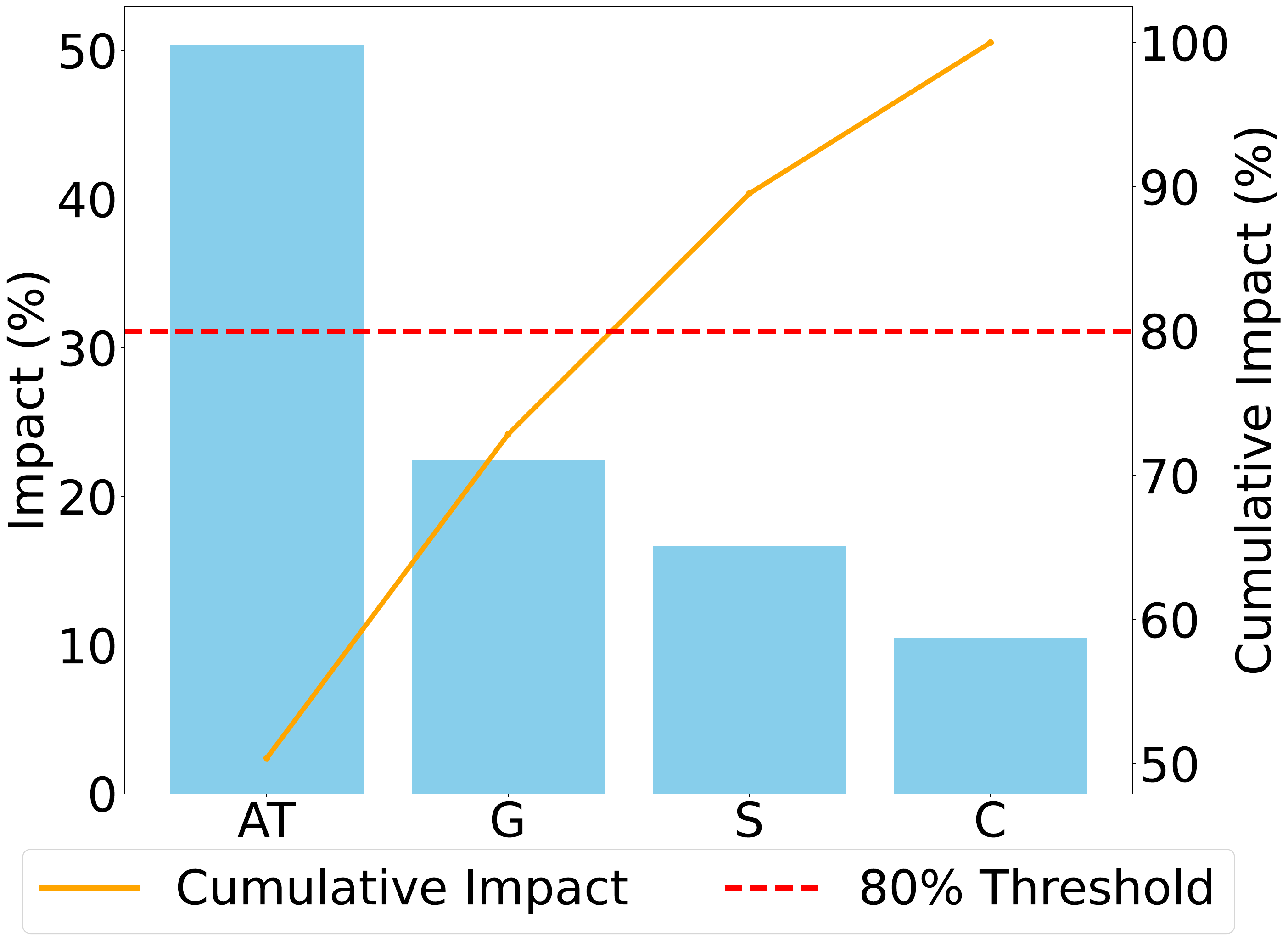}
            \caption{Creative StoryGen}
            \label{fig:storygen_pareto}
        \end{subfigure}
    \end{minipage}
    \caption{\textbf{Attributing BF Reduction.} (a) Heatmap of average BF ratio (Base/Aligned) across tasks and models. The numbers indicate the average ratio over all constraint levels. Note that for OLMo-2, we follow the convention to treat the DPO version as the aligned model \citep{olmo20242}. \tmlrrevisee{Since both base and aligned models in each column share the same tokenizer, the within-family tokenizer scaling cancels in the ratio; cross-family ratio magnitudes should be interpreted as patterns rather than precise comparisons.} (b)-(e) Pareto Analysis of BF across various IFs. $AT$ indicates whether the model is aligned. $C$ denotes the prompt complexity. $S$ refers to model size, and $G$ refers to model generation.}
    \label{fig:pareto_analysis}
    \vspace{-10pt}
\end{figure}

\subsection{Pareto Analysis of BF}
\label{sec: bf_pareto}

How dominant is this alignment effect compared to other factors? \cref{fig:bf_ratio_heatmap} offers a first look: the Base/Aligned BF ratio is consistently high ($\gg 1$) across diverse tasks and models, peaking at 10$\times$ for \textsc{Random Strings} (task-wide) and appearing overall mildest for OLMo-2 (model-wide). To rigorously rank alignment against model size ($S$), generation ($G$), and prompt complexity ($C$), we perform a Pareto analysis (\cref{fig:pareto_analysis}) for Llama models. For each factor $D_i$, we define the unnormalized \textit{Impact} $\tilde{I}(D_i)$ as the average absolute pairwise difference in BF when varying $D_i$ while holding other dimensions constant:
\begin{small}
\begin{align}
   \tilde{I}(D_i) = \frac{ \sum_{d_i, d_j \in \text{Domain}(D_i), d_i \neq d_j} 
    {|\text{Avg}(\text{B}(\cdot | D_i=d_i)) - \text{Avg}(\text{B}(\cdot | D_i=d_j))|}}{|\text{Domain}(D_i)| \times |\text{Domain}(D_i) - 1|}. 
\end{align}
\end{small}Then we normalize it as ${I}(D_i)=\frac{\tilde{I}(D_i)}{\sum_k \tilde{I}(D_k)}$. 

The results crisply validate our intuition: \textbf{alignment tuning is the primary driver of BF reduction}. \revise{Across all tasks, it consistently crosses or approaches the 80\% cumulative impact threshold, surpassing all other factors by a large margin.} 
\tmlrrevisee{This is consistent with the cross-sample diversity reduction independently documented by~\citet{kirk2024understanding} and~\citet{padmakumar2024does}; BF re-expresses the same phenomenon at the distributional, per-step level, which is what lets us connect it to decoding/inference-time behaviors (\cref{sec: bf_implications}) and decompose it into stage-wise contributions (\cref{sec: nudging}).}

Among the secondary factors, for tasks with richer inputs--such as \textsc{MMLU} (with more in-context examples) and \textsc{BBCLatestNews} (with more headlines)--prompt complexity $C$ and model size \revise{$S$} emerge as the next most impactful. Prompt complexity $C$ has a noteworthy effect: contrary to intuition, more context provided in the prompt does not always reduce BF but can in fact increase it, potentially due to the cognitive burden of processing complex linguistic structures \mvhnrevise{such as negated constraints in \textsc{Cognac}}. A detailed case study and comprehensive task-wise BF results are presented in \cref{app: curious_case_prompt_complexity,app: full_taskwise_bf}. In contrast, for open-ended tasks like Cognac and Story Generation, model generation $G$ plays a more dominant role, particularly improvements from Llama-2 to Llama-3. This shift likely reflects gains 
\revise{from} the use of larger, more diverse datasets in training~\citep{dubey2024llama}.

\section{Low Branching Factor Explains Generative Stability and Commitment}
\label{sec: bf_implications}

Our analysis in \cref{sec: bf_measure} established that BF declines over the generation process (\cref{sec: bf_dynamic}) and is significantly lower in aligned models (\cref{sec: bf_pareto}). This structural concentration of probability mass provides a unified probabilistic explanation for three distinct generative behaviors: the insensitivity of aligned models to decoding hyperparameters, their reduced variance in majority voting, and the high performance cost of late-stage exploration.

\subsection{Insensitivity to Decoding Configurations}
\label{sec: sampling_efforts}

Model developers adopt different decoding strategies when reporting LLM capabilities~\citep{touvron2023llama, dubey2024llama, yang2024qwen2, guo2025deepseek}. 
The effectiveness of strategies like nucleus sampling ($p$) and temperature sampling ($T$) relies on the assumption that a diverse set of plausible next tokens exists.
However, our BF analysis suggests that for aligned models, the "plausible set" is much smaller. If this holds, we expect aligned models to be robust to decoding choices, as there are few alternatives for the decoding algorithm to select even at higher temperatures.

We verify this by benchmarking decoding methods on MMLU-STEM~\citep{hendrycks2021measuring}, extending prior work\revise{~\citep{song2024good, renze2024effect, shi2024thorough}} to the latest models including DeepSeek-distilled models~\citep{guo2025deepseek}, which would generate long CoT before the final answer.\footnote{For Llama-3 series models, in our prior study, we find there is only a minor performance difference between Llama-3 and Llama-3.x. We mainly use Llama-3 in this paper as it includes the most diverse model collection. } Specifically, we evaluate model performance on MMLU-STEM~\citep{hendrycks2021measuring} under CoT prompting across different temperatures ($T$=~0.6/1.0) in temperature sampling and truncation thresholds ($p$=0.9/1.0) in nucleus sampling~\citep{Holtzman2020The}. Further implementation details can be found in \cref{app: sampling_efforts}. 

As shown in \cref{tab:cot_mmlu_stem}, aligned models exhibit limited performance variation (typically $<10\%$) even when shifting from greedy-like settings to high temperatures. In contrast, base models, which maintain higher BF, show significant sensitivity (up to 31\%) to decoding parameters.  Notably, DeepSeek-distilled Llama-8B, which generates long CoT and thus maintains a consistently low BF throughout generation, exhibits the smallest relative performance changes among 8B models. This empirically supports that \emph{low BF effectively nullifies the impact of sampling method choices.}

\begin{takeaway}
\tkw Low BF leaves few viable branches, so decoding hyperparameters barely move aligned models ($<$10\%) versus up to 31\% for base models.
\end{takeaway}

\begin{table}[htbp!]
\centering
\resizebox{\textwidth}{!}{
\begin{tabular}{lccccc}
\toprule
Models & Default ($T$=0.6, $p$=0.9) & $T$=0.6, $p$=1.0 & $T$=1.0, $p$=0.9 & Min ($T$=1.0, $p$=1.0) & $\frac{\text{Default}-\text{Min}}{\text{Default}}\%$ \\
\midrule
Llama-3-70B-Instruct & 78.50 ($\pm$ 2.09) & 77.60 ($\pm$ 2.23) & 77.50 ($\pm$ 2.60) & 75.90 ($\pm$ 2.85) & 3.31 \\
Llama-3-70B & 78.00 ($\pm$ 3.52) & 74.00 ($\pm$ 3.80) & 72.00 ($\pm$ 4.38) & 63.50 ($\pm$ 5.02) & 18.59 \\
DeepSeek-R1-Distill-Llama-8B & 66.30 ($\pm$ 3.51) & 65.70 ($\pm$ 3.84) & 62.70 ($\pm$ 4.14) & 59.70 ($\pm$ 4.65) & 9.95  \\
Llama-3.1-8B-Instruct & 63.00 ($\pm$ 4.01) & 61.50 ($\pm$ 4.37) & 57.50 ($\pm$ 4.92) & 50.50 ($\pm$ 5.34) & 19.84 \\
Llama-3.1-8B & 54.00 ($\pm$ 4.61) & 53.50 ($\pm$ 4.92) & 47.00 ($\pm$ 5.21) & 37.00 ($\pm$ 5.48) & \textbf{31.48} \\
\bottomrule
\end{tabular}
}

\caption{\textbf{Experiment Results across decoding methods on STEM subset of MMLU.} We follow the common practice of using 5-shot CoT prompting. 
$\frac{\text{Default}-\text{Min}}{\text{Default}}\%$ indicates the maximum relative performance drop when deviating from the default decoding configuration. }
\label{tab:cot_mmlu_stem}
\end{table}

\begin{table}[htbp!]
\centering
\resizebox{0.7\textwidth}{!}{
\begin{tabular}{lccccHc}
\toprule
Model  & Maj@1 Std & Maj@3 Std & Maj@8 Std & Maj@16 Std &  BF@1 & BF \\
\midrule
DeepSeek-R1-Distill-Llama-70B & \textbf{14.34} & \textbf{8.29} & \textbf{4.99} & \textbf{3.21} & 1.77 & \textbf{1.23} \\
Llama-3-70B-Instruct   & 16.37 & 11.40 & 7.50 & 5.12 & 2.44 & 1.28  \\
Llama-3-70B   & 27.78 & 19.53 & 13.22 & 9.23 & 2.41 & 1.31  \\
\midrule
DeepSeek-R1-Distill-Llama-8B  & \textbf{27.10} & \textbf{20.91} & \textbf{13.93} & \textbf{9.14} & 1.77 & \textbf{1.23} \\
Llama-3.1-8B-Instruct  & 31.54 & 24.64 & 17.30 & 12.90 & 2.73 & 1.31 \\
Llama-3.1-8B   & 36.41 & 29.78 & 20.43 & 14.05 & 2.53 & 1.35  \\
\bottomrule
\end{tabular}
}
\caption{\textbf{Majority Voting@K standard deviation on MMLU-STEM with $200$ samples.} We compute the standard deviation over $100$ bootstrapping trials, each using $64$ samples per instance. We set $T=0.6, p=0.9$ to match standard benchmarking settings, differing from $T=1.0, p=0.9$ setup in \cref{sec: bf_measure}. 
\revise{Lower temperature concentrates probability mass on fewer tokens, reducing BF and complicating direct comparisons. However, bootstrapping (100 runs) reveals minimal variability ($\approx 0.01$), confirming that the BF differences reported here remain significant. Consequently, BF remains a strong predictor of standard deviation.}
}
\label{tab:std_prediction}
\vspace{-10pt}
\end{table}

\subsection{Reduced Variance in Majority Voting}
If aligned models have a restricted output space (low BF), this should also manifest as reduced variance across independent samples. \tmlrrevisee{Beyond simply confirming the lower sampling variance documented for aligned models in prior decoding studies~\citep{song2024good}, our framework lets us \emph{predict} that this effect should further intensify for long-CoT reasoning models in inference-time scaling: because their BF stays low throughout the long reasoning trace, the variance across independent rollouts should be even smaller than that of direct-answer aligned models at the same scale.}
We test this prediction by evaluating output variance on MMLU-STEM using 200 samples per model. We benchmark the standard deviation across Majority@K accuracy metrics ($K = 1, 3, 8, 16$) with $T=0.6, p=0.9$.

As detailed in \cref{tab:std_prediction}, BF serves as a strong predictor of sampling consistency. The Long-CoT models (DeepSeek-R1-Distill), which generate significantly longer outputs and achieve the lowest global BF, consistently show the smallest performance variance. This suggests that the ``stability'' observed in decoding benchmarks is not merely an artifact of specific hyperparameters, but is closely tied to the narrowed generation manifold.

\begin{takeaway}
\tkw Lower BF predicts smaller Majority@K variance (smallest for long-CoT models); once BF is low, extra votes barely yield further performance gains.
\end{takeaway}

\subsection{Risks of Mid-Generation Forking}
\label{sec: forking}
The stability provided by low BF suggests a strong commitment to a specific reasoning trajectory. Does this commitment imply that the model cannot effectively explore alternative paths once generation has begun? If BF reflects a semantic ``lock-in,'' forcing the model to branch out (fork) at late, low-BF stages should disrupt coherence and degrade performance.

To examine this, we conduct a resampling experiment using DeepSeek-Distilled Llama-8B output samples.  
\revise{Procedurally, for a given position $t$: 1) Take the prefix $\outputval_{<t}$ generated by the model. 2) Sample a new continuation $\outputval'_{\ge t}$ from $P(\outputVar_{\ge t} | [\inputval, \outputval_{<t}]; \theta)$. 3) Evaluate the full sequence $[\outputval_{<t}, \outputval'_{\ge t}]$ on the task.}

\begin{figure}[t!]
    \centering
    \includegraphics[width=0.7\linewidth]{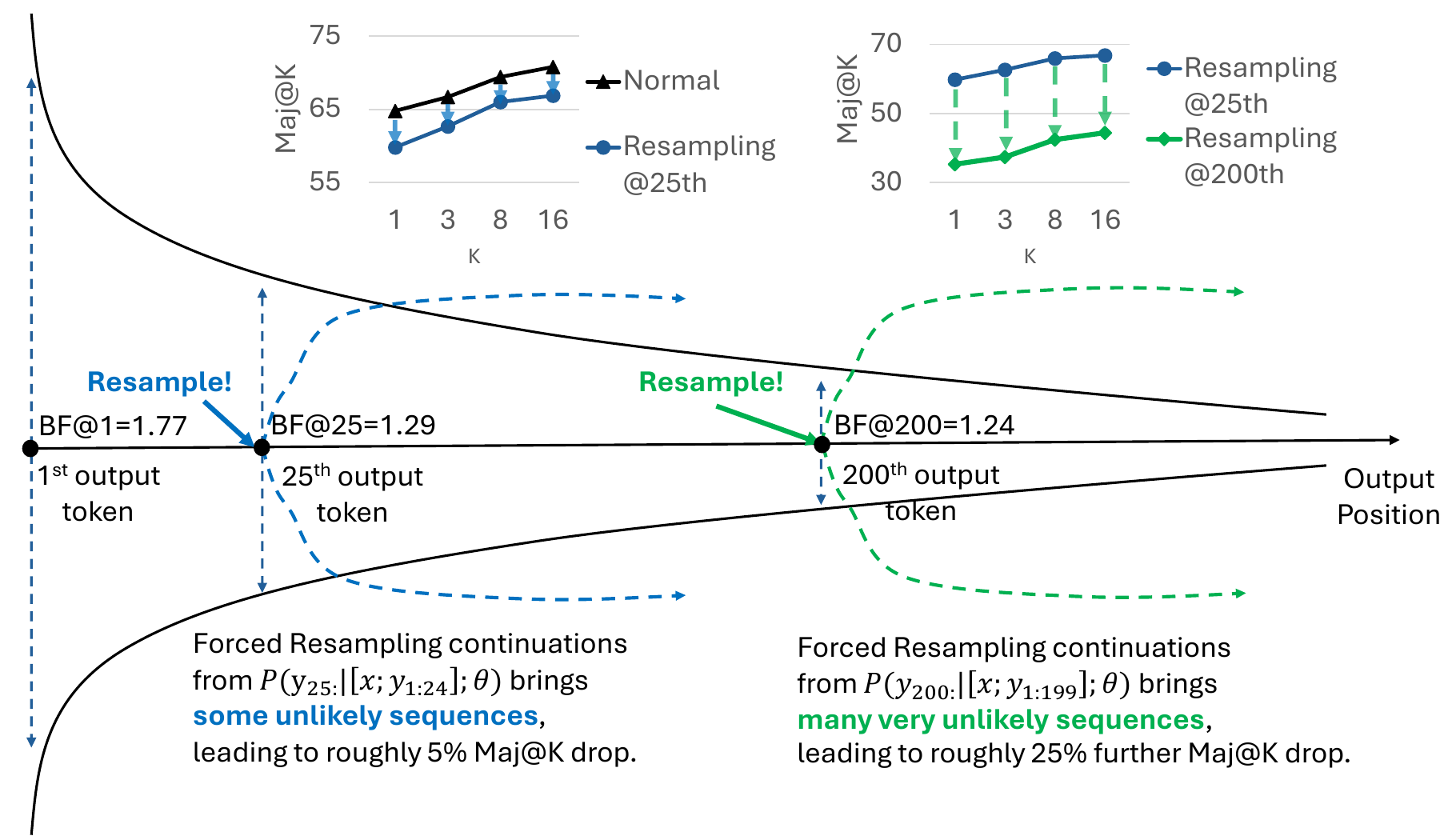}
    \vspace{-10pt}
    \caption{\textbf{Resampling from different output positions to assess the effect of interrupting BF reduction}. 
    We resample new continuations at the 25th and 200th output token of DeepSeek-Distilled Llama-8B MMLU outputs. 
Results show substantial performance drops at both positions.
}
\vspace{-15pt}
    \label{fig: bon_from_middle}
\end{figure}

As shown in \cref{fig: bon_from_middle}, performance drops sharply when resampling occurs at a later, lower-BF position in the sequence. 
This suggests that aligned models are not just concentrating probability mass locally \revise{(reflects a "deeper commitment" to specific paths)}, but are actively locking into trajectories, making late-stage deviations more error-prone. 
In practice, this highlights a key application of BF: \emph{parallel sampling should be applied early, while BF remains high}, to ensure meaningful diversity and avoid quality degradation.

\begin{takeaway}
\tkw Late, low-BF forks sharply hurt accuracy, as the model already commits; branch early, while BF is still high.
\end{takeaway}

%% file: past_papers/branching_factor/BF_nudging.tex
\begin{figure}[htbp]
    \centering
    \begin{subfigure}[t]{0.4\textwidth}
    \centering
     \includegraphics[width=\linewidth]{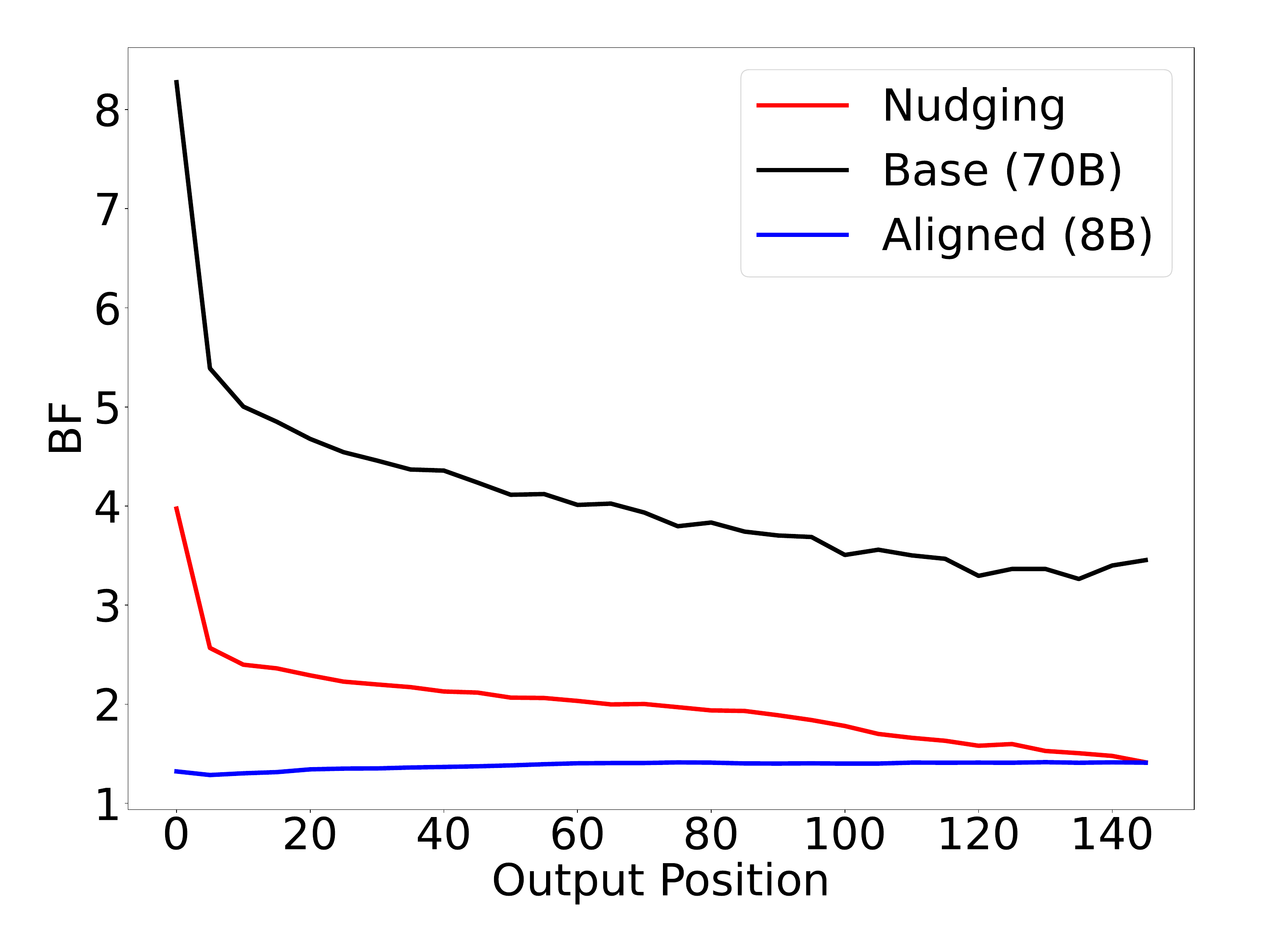}
    \subcaption{Output BF Dynamics}
     \label{fig:just_eval_instruct_nudging}\vfill
     \end{subfigure}
     \begin{subfigure}[t]{0.4\textwidth}
    \centering
     \includegraphics[width=\linewidth]{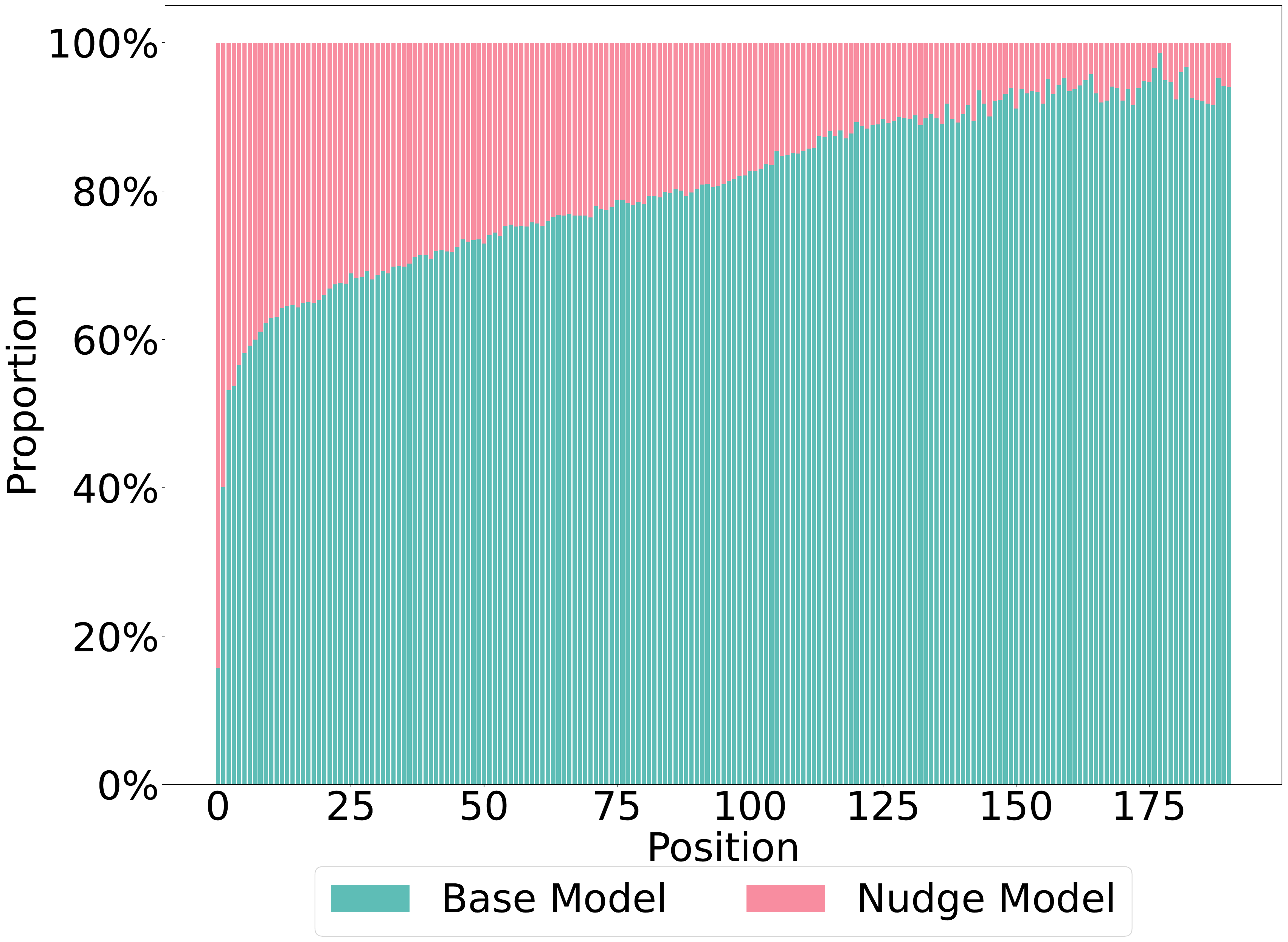}
    \caption{Nudging Ratio Histogram}
\label{fig:just_eval_instruct_nudging_histogram}
\end{subfigure}
\vspace{-0.26cm}
\caption{Nudging Experiments over Just-Eval-Instruct.}
\label{fig:nudging_analysis}

\vspace{-10pt}
\end{figure}

\section{How does Alignment Tuning Impact BF?}
\label{sec: nudging}

Why does alignment tuning exert such a pronounced effect on BF? 
Building on the superficial alignment hypothesis~\citep{zhou2024lima} (``\textit{Alignment tuning might simply teach base LLMs to select a subdistribution of data formats for interacting with users.}'') and recent tuning-free alignment work\revise{~\citep{lin2023unlocking, fei2024nudging, lake2025distributional}}, 
we hypothesize base models already encode low-entropy conditional distributions. In this view, alignment tuning doesn't reshape generation from scratch, but instead nudge the model toward \revise{\emph{stylistic tokens}} (e.g., ``Sure''), thereby narrowing the conditional 
distribution.

To test this hypothesis, we reproduce the nudging experiments \citep{fei2024nudging}, over Just-Eval-Instruct~\citep{lin2023unlocking} and MMLU datasets. We employ Llama-3-70B  for drafting most outputs. However, when the base model's Top-1 probability is low, we apply nudging by switching to Llama-3-8B-Instruct to generate a single word.
\revise{Using a smaller aligned model to nudge a larger base model (70B) isolates the steering effect of the stylistic prefix itself, independent of the nudging model's raw capability.}
BF was computed as in prior experiments. 
The results, shown in \cref{fig:nudging_analysis},\footnote{We present results on Just-Eval-Instruct only for brevity. MMLU results are included in \cref{app: nudging}.} 
indicate that after
most nudging occurs early in the generation process -- indicating the prefix generated by the nudging model is of low probability. These observations collectively support our hypothesis.
\tmlrrevisee{We emphasize the role of this experiment: \citet{fei2024nudging} use their nudging setup to evaluate sample-level task performance for inference-time alignment, and our goal is \emph{not} to re-explain that observation. Instead, we adopt the same controlled setup to \emph{test our distributional hypothesis} about how probability concentration arises during alignment -- namely, that a small number of stylistic tokens is sufficient to steer a base model into the low-BF regime characteristic of aligned models. The BF measurement then provides the per-step distributional view (which prefix induces how much concentration, and where in the sequence) that sample-level metrics cannot resolve.}

\begin{wrapfigure}{r}{0.45\textwidth}
\vspace{-15pt}
\begin{minipage}{\linewidth}
    \centering
     \includegraphics[width=\linewidth]{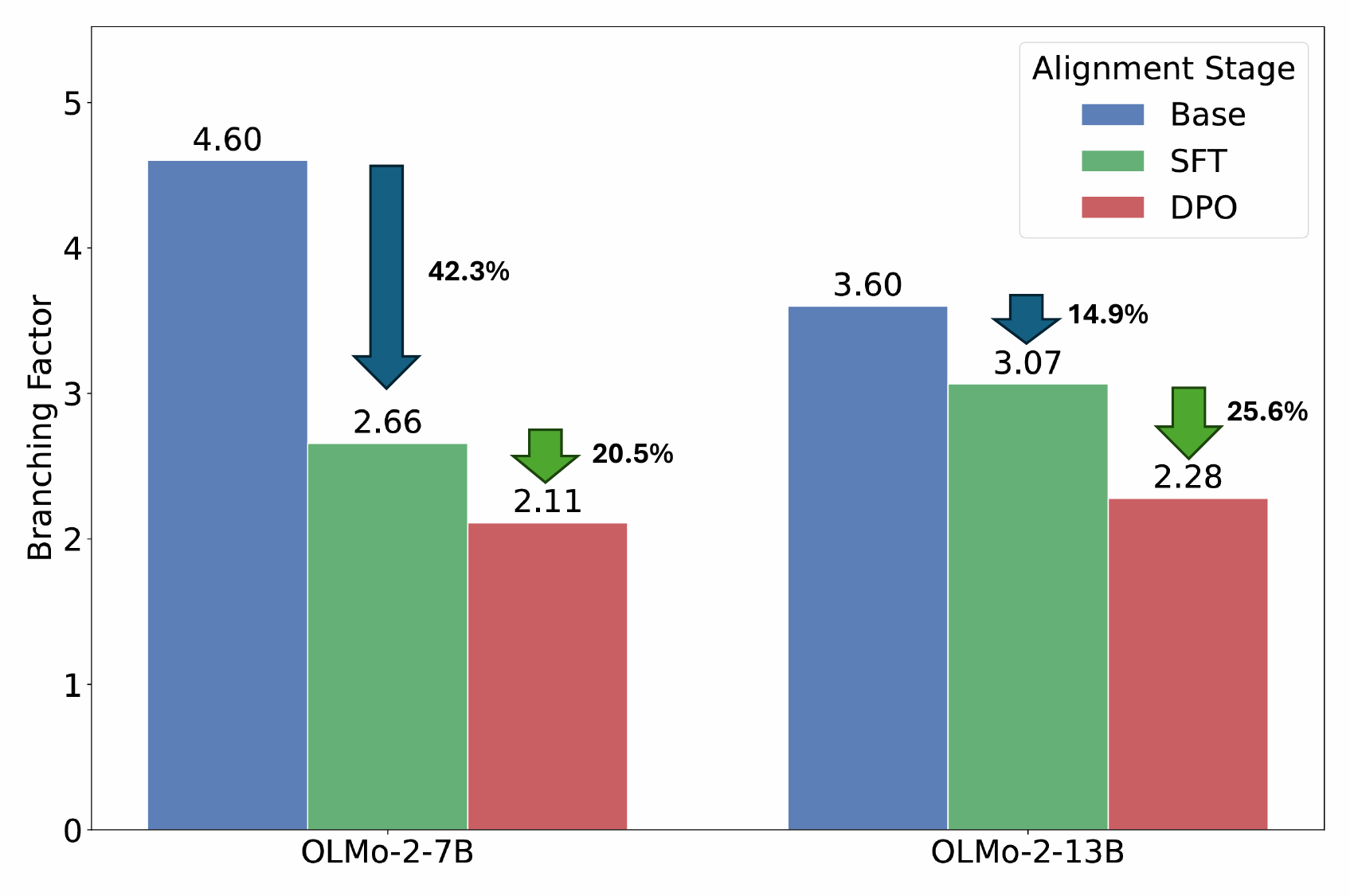}
     \vspace{-20pt}
    \caption{\textbf{Stage-wise Contribution to BF Reduction.} \revise{We analyze OLMo-2 models on the Creative StoryGen task.}}
\label{fig:olmo2_stage_analysis}
\vspace{-0.2in}
\end{minipage}
\end{wrapfigure}
\shortparagraph{Which Training Stage Reduces BF Most?}
While our nudging analysis suggests alignment tuning narrows the distribution by nudging models toward stylistic tokens, it remains unclear which specific phase of the pipeline drives this reduction. Disentangling these effects is difficult because most model releases (e.g., Llama 3~\citep{dubey2024llama}) do not provide intermediate checkpoints, and modern post-training often employs iterative, interleaved schedules rather than discrete stages.
However, the OLMo-2 suite~\citep{olmo20242} releases intermediate checkpoints, enabling a preliminary stage-wise dissection. We measure BF changes across Supervised Fine-Tuning (SFT) and Direct Preference Optimization (DPO)~\citep{rafailov2023direct} stages on the open-ended \textsc{Creative StoryGen} task.
\revisestart
As shown in \cref{fig:olmo2_stage_analysis}, we observe distinct behaviors across model scales: for the 7B OLMo-2 model, SFT is the primary driver of BF reduction, causing a 42.3\% drop compared with 20.5\% for DPO, whereas for the 13B OLMo-2 model, the DPO stage contributes more significantly, with a 25.6\% drop compared with 14.9\% for the SFT stage. This divergence suggests that the impact of alignment tuning stages is not universal but sensitive to specific training recipes and model scales.

\shortparagraph{Beyond OLMo-2: Other Post-Training Algorithms.}
The OLMo-2 checkpoints provide a clean stage-wise comparison, but they only cover the SFT and DPO stages. To broaden the picture, we next study a publicly released RLHF pipeline from RLHFlow~\citep{dongrlhf} and  OpenRLHF~\citep{hu2024openrlhf} collaboration, which includes checkpoints for SFT, DPO, PPO, and Iterative DPO variants. Because post-training recipes vary substantially across organizations (e.g., in data and training setup), we view this as a controlled case study rather than a universal ranking of algorithms. We evaluate all checkpoints on \textsc{Creative StoryGen} using the same setup as above.
\begin{figure}[htbp]
    \centering
    \includegraphics[width=0.55\linewidth]{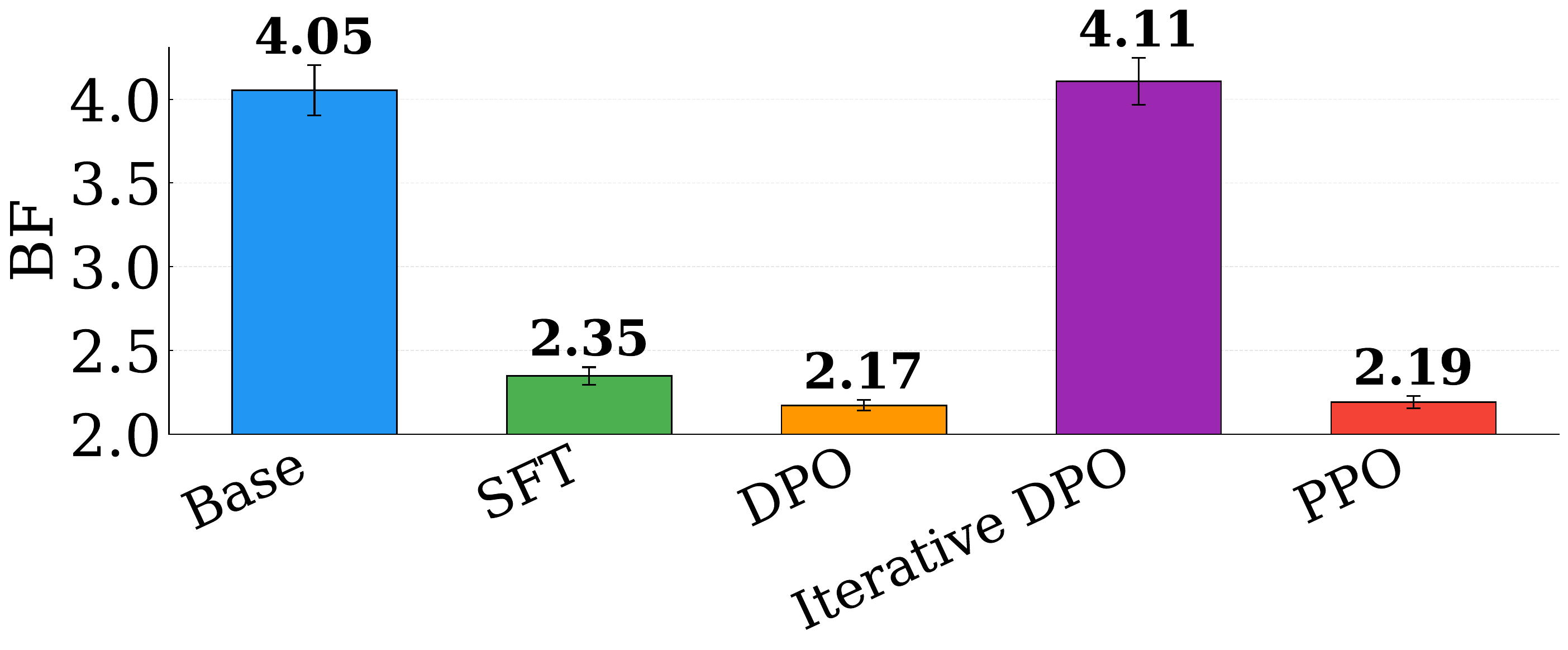}
    \caption{\revisestart \textbf{BF across post-training algorithms.} Using checkpoints from an RLHFlow/OpenRLHF pipeline on Creative StoryGen, we find that SFT drives the largest BF drop, DPO and PPO yield similar further reductions, and Iterative DPO partially preserves distributional breadth.\label{fig:alignment_algo_comparison}}
\end{figure}

As shown in \cref{fig:alignment_algo_comparison}, SFT again accounts for the largest BF reduction ($4.05 \to 2.35$), consistent with the OLMo-2 trend. DPO and PPO then produce very similar additional reductions ($2.35 \to 2.17$ for DPO and $2.35 \to 2.19$ for PPO), suggesting that these two offline preference-optimization methods concentrate the distribution to a comparable extent beyond the SFT stage.
Iterative DPO, however, avoids part of this shrinkage. A plausible explanation is that its more online training loop repeatedly refreshes the reference policy and collects preference signals from the model's updated outputs, which can expose the model to underrepresented regions of the solution space instead of reinforcing a fixed offline preference set. This interpretation is consistent with~\citet{wu2025invisible}, who argue that offline reinforcement-style methods remain tightly tethered to their initial reference policy and therefore have limited ability to move beyond it.

\reviseend

%% file: past_papers/branching_factor/related_works.tex
\section{Related Works}
\label{sec: related_work}
\shortparagraph{Uncertainty Quantification for LLM.} Uncertainty quantification (UQ) for LLMs has gained significant attention due to its importance in real-world applications, particularly in high-stakes domains~\citep{desai2020calibration, jiang2021can, wang2022uncertainty, kadavath2022language, xiong2024can, ye2024benchmarking, gupta2024language}. Existing methods typically address closed-domain tasks such as classification and question-answering, where outputs are discrete and easier to assess. However, as \citet{kuhn2023semantic} note, these approaches often overlook challenges specific to open-ended generation, such as semantic equivalence across outputs. They introduce ``semantic entropy''  to quantify uncertainty in LLM output space by first clustering the sampled output and then quantifying uncertainty over cluster distribution. This method empirically works well in hallucination detection~\citep{farquhar2024detecting}. 
In this paper, we focus on investigating the probability concentration phenomenon for LLMs. We introduce BF to quantify this concentration, \revise{which applies} broadly across tasks without imposing strong assumptions on output categories. 

\shortparagraph{Reduced Diversity in Aligned Models.} 
Recent studies have consistently shown that alignment tuning reduces output diversity in language models~\citep{perez2022red, padmakumar2024does, chakrabarty2024art, tian2024large, kirk2024understanding, lu2025ai, lake2025distributional, west2025base}. 
\tmlrrevisee{Among these,~\cite{kirk2024understanding} is the most closely related, with a systematic cross-sample diversity analysis under RLHF. We discuss the relationship between cross-sample diversity and BF in~\cref{app: related_works}}.
\revise{Mechanistically, \citet{lake2025distributional} observe that alignment suppresses diversity by aggregating information into longer, standardized responses, though they argue this preserves useful base model behaviors. Our work builds on this inquiry by} connecting reduced diversity with related observations of diminished randomness and robustness in aligned models~\citep{saparov2023language, song2024good, renze2024effect, bigelowsubjective, shi2024thorough}. \tmlrrevisee{Crucially, while previous decoding-sensitivity studies document the \emph{symptom} (e.g., aligned models becoming insensitive to sampling hyperparameters), the BF framework pinpoints the underlying \emph{mechanism}: the shrinking distributional landscape over the course of token-by-token generation. This conceptual lens allows us to further forecast and verify the same probability concentration effects in long-CoT models.}

Traditional diversity metrics such as n-gram lexical diversity~\citep{li2016diversity} are sensitive to vocabulary size and output length~\citep{liu-etal-2022-rethinking,tevet2021evaluating, guo2024benchmarking, kirk2024understanding} and cannot work well with most recent long CoT models. In \cref{sec: lexical_diversity_and_bf}, we demonstrate that lexical diversity poorly correlates with BF and fails to robustly measure generation concentration. 

\tmlrrevisee{Beyond diagnostic findings, the BF lens has begun to enable concrete downstream applications, including inference-time methods for reasoning~\citep{fu2025deepconf} and open-ended generation~\citep{wang2025baco}, as well as training-time methods for reasoning~\citep{yang2025ead} and joint quality--diversity optimization~\citep{li2025darling}.}

Our work also resonates with information density research in cognitive science and linguistic theories, and we present a short discussion in \cref{app: bf_and_id}.

%% file: past_papers/branching_factor/discussions.tex
\section{Discussion}

\shortparagraph{Practical Implications.}
A key practical implication of our findings is that reduced BF neglects alternative generations and forking. Consequently, simply tweaking decoding parameters (e.g., temperature), is unlikely to restore diversity without severely degrading quality \citep{renze2024effect}. 
\revise{Our work offers a clear explanation for why this occurs, particularly for methods like beam search. The resampling experiment in \cref{sec: forking} provides direct evidence that for low-BF models, off-path trajectories are not just less probable but often of lower quality. With little probability mass distributed among alternative paths, beam search has few viable options to explore, yielding diminishing returns.} This suggests that efforts to mitigate diversity loss should target the training process itself -- a more promising, albeit challenging, direction. Future work could involve curating more diverse alignment data or designing novel training objectives that balance instruction-following with distributional diversity \citep{wang2024beyond, kwon2024gdpo, lanchantin2025diverse, chung2025modifying}. System-level interventions (e.g., model collaboration) also present a viable path forward \citep{fei2024nudging, lu2024llm, venkatraman2025collabstory, ismayilzada2025creative}. While our paper's primary contribution is diagnostic, we believe this foundational understanding is a necessary prerequisite for developing such effective countermeasures.

\tmlrrevise{That said, the above discussion applies primarily to settings where output diversity is valued, such as creative generation, open-ended dialogue, and exploratory reasoning. For tasks with unique correct answers (e.g., arithmetic, factual retrieval), low BF might in fact be desirable -- it might reflect the model concentrating probability mass on the correct output, and alignment-induced BF reduction can be beneficial.}

\mvhnrevise{\shortparagraph{Autoregressive Self-Narrowing vs.\ Alignment.} It is important to separate two effects that are easy to conflate. The \emph{downward trend} of BF over generation is a robust aggregate tendency of autoregressive generation, not a monotonic token-wise law: as the model conditions on its own growing prefix, the next-token distribution often becomes more concentrated, even when the context is not semantically meaningful. This helps explain why BF also decreases for \textsc{Random Strings}, which is otherwise puzzling. Alignment is a separate force that lowers the absolute BF level and steepens the early narrowing, rather than being the sole explanation for the decrease. We make this explicit through a controlled intervention in \cref{app: bf_self_narrowing}: replacing the model's own prefix with externally-sampled i.i.d.\ random tokens surges BF back up and then generally lets autoregression narrow it again, in both base and aligned models. A practical corollary -- relevant only when diversity is desired -- is that injecting out-of-distribution/unexpected content into the context can \emph{temporarily} drag the model back to a high-BF region; but because autoregressive self-conditioning tends to re-concentrate the distribution over subsequent tokens, this is a local intervention rather than a cure. Conversely, content that is unexpected from the model's own predictive viewpoint (random strings, adversarial agentic feedback, or negation) can \emph{raise} BF (\cref{app: bf_self_narrowing}).}

\revise{\shortparagraph{Societal Homogeneity Bias of Alignment Tuning.} Our work identifies a key dynamic in modern LLMs: alignment tuning significantly reduces the Branching Factor (BF), leading to more homogenized and predictable outputs. While this can be beneficial, it also carries potential negative societal impacts. In applications such as automated content generation, creative writing, or decision-support systems~\citep{padmakumar2024does, sorensen2024position, wu2025generative, murthy-etal-2025-one, rodemann2025statistical, ashkinaze2025ai, lake2025distributional}, this reduction in diversity could inadvertently reinforce social biases, stifle creativity, and limit the exploration of novel ideas. \citet{rodemann2025statistical} further argue that empirical alignment, relying on limited and potentially biased human feedback, creates selectional bias that fails to capture the full spectrum of human values. We believe that formally understanding and quantifying the mechanisms of probability concentration, as we do in this paper, is a critical and necessary first step toward developing alignment techniques that mitigate these risks and foster models that are not only helpful and harmless but also diverse and robust. 
}

\section{Limitations}

\revise{
\shortparagraph{BF is a First-Moment Summary.} By definition, BF captures the exponentiated length-averaged entropy -- a single scalar. Like any such summary, it can obscure higher-order structural properties of the generative distribution. Consider a stylized example with sequences of length $N=2$ over vocabulary ${a,b,c,d}$. \textsc{Distribution A} (uniform branching): $P(Y_1 = a)=P(Y_1 = b)=1/2$, $P(Y_2 = c\mid Y_1)=P(Y_2 = d\mid Y_1)=1/2$ regardless of $Y_1$. \textsc{Distribution B} (front-loaded branching): $P(Y_1)$ is uniform over all four tokens, while $P(Y_2\mid Y_1)$ is deterministic. Both distributions yield $\tilde H(Y_{1:2})=2\ln 2$ and hence $BF=2$, yet they correspond to qualitatively different trees. Nonetheless, BF is designed to answer a specific question --- how concentrated is the generation process on average -- and practitioners who require finer-grained characterization of the tree topology should complement BF with positional or higher-order analyses.

\tmlrrevisee{
\shortparagraph{Limitations of BF Hybrid Estimator.} Our efficient BF estimator involves two finite-sample approximations, each with its own bias. First, the gap between length-averaged NLL and realized entropy shrinks with sequence length $N$ (\cref{thm: aep_llm}) but can be non-negligible for short sequences. To address this, we adopt a hybrid strategy (\cref{eq:hybrid_estimator}): for sequences shorter than a threshold $L_\tau$, we compute the exact realized entropy via full vocabulary summation, resorting to the NLL approximation only for longer sequences where it is both accurate and computationally necessary. Figures 2(c) and 2(d) empirically validate this design: the standard deviation of the length-averaged NLL estimator diminishes rapidly, falling to practical levels within roughly 20 tokens. These observations provide concrete guidance for practitioners -- exact entropy computation is feasible and recommended for short outputs (we recommend choosing a $L_\tau$ between 20 and 50), while the NLL proxy can be safely adopted beyond this early regime.

Second, the finite sample Monte-Carlo estimate of realized entropy itself underestimates the true entropy because the rare, high-entropy tail is under-sampled at finite $M$;~\cref{appendix: under_estimation_of_entropy_via_mc} characterizes this empirically (estimated BF rises monotonically as $M$ grows from 4 to 64). Absolute BF magnitudes should therefore be read as lower bounds. Cross-model and base-vs-aligned ratios are more robust to this bias than absolute magnitudes, because the under-sampling has comparable scale across models with comparable effective output spaces.

}

}
\tmlrrevisee{
\shortparagraph{Tokenizer Dependence and Cross-Family Comparison.} BF is computed at the token level, so its raw magnitude depends on the model's tokenizer: a coarser tokenizer (fewer, longer tokens) tends to produce smaller BF than a finer one, even for distributions of comparable conceptual breadth. Direct cross-family comparison of raw BF values is therefore not meaningful. Throughout this paper we either compare within a single family or, for cross-family analyses such as~\cref{fig:bf_ratio_heatmap}, report the within-family Base/Aligned BF \emph{ratio}, which is tokenizer-invariant because each Base/Aligned pair shares a vocabulary. Practitioners adopting BF should follow the same convention.

\shortparagraph{Surface vs. Semantic Concentration.} BF is a distributional, token-level summary and is by construction insensitive to whether two surface forms with different token sequences encode the same meaning -- a distinction that semantic uncertainty methods such as~\citet{kuhn2023semantic} are specifically designed to capture. We view these measures as complementary: BF diagnoses where probability mass is concentrated in the model's own next-token distribution, while semantic clustering diagnoses how this concentration projects onto a smaller set of meanings. Empirically, the token-level concentration captured by BF accumulates over the sequence and manifests in measurable reductions in sample-level lexical and semantic diversity for aligned models~\citep{kirk2024understanding, west2025base, lake2025distributional}; quantifying the per-prompt mapping from BF to semantic diversity is an interesting direction for follow-up work.

}

%% file: chapters/hindsight.tex
\section*{Chapter Overview}
Recent studies suggest that self-reflective prompting can significantly enhance the reasoning capabilities of Large Language Models (LLMs). However, the use of external feedback as a stop criterion raises doubts about the true extent of LLMs' ability to emulate human-like self-reflection. In this chapter, we set out to clarify these capabilities under a more stringent evaluation setting in which we disallow any kind of external feedback.

\graphicspath{{./}}

\makeatletter
\def\input@path{{./}}
\makeatother

\begingroup
\lstset{
  basicstyle=\ttfamily,
  breaklines=true,
  postbreak=\mbox{\textcolor{red}{$\hookrightarrow$}\space},
}
\input{past_papers/hindsight/sections/introduction}

\input{past_papers/hindsight/sections/method}

\input{past_papers/hindsight/sections/preliminary}

\input{past_papers/hindsight/sections/analysis}

\input{past_papers/hindsight/sections/artificial}

\input{past_papers/hindsight/sections/voting}

\input{past_papers/hindsight/sections/discussion}

\input{past_papers/hindsight/sections/conclusion}

\input{past_papers/hindsight/sections/limitations}

\endgroup

%% file: past_papers/hindsight/sections/introduction.tex
\section{Introduction}

Large Language Models (LLMs) have shown impressive performance in generating human-like text (e.g., ChatGPT \citep{chatgpt}), and recent works demonstrate that we can further prompt LLMs to reflect on their own outputs to improve their capabilities on complicated reasoning, programming and planning tasks~\citep{huang2022large, kim2023language, madaan2023selfrefine, Shinn2023ReflexionLA, chen2023teaching, wang2023enable} and also improve their alignment with human values (e.g., less harmful and more helpful)~\citep{bai2022constitutional, ganguli2023capacity}.\footnote{Various terms like ``self-reflection'', ``self-refine'',  ``self-correction'', and ``self-improvement'' describe these introspective behaviors. For clarity and consistency, we will exclusively use ``self-reflection'' in this paper.} However, 
\citet{huang2023large} find that performance gains associated with self-reflection may be due to implicit usage of external feedback as a stop criterion, as well as overly-engineered prompts that bias the model outputs, casting doubt on the true effectiveness of self-reflection. 

To verify the extent to which LLMs can truly reflect on their outputs, we take a more stringent evaluation approach: in addition to excluding external feedback~\citep{huang2023large}, we also disallow multi-round iterative prompting, which can hint to the model that its prior response is incorrect. Instead, we sample multiple model responses given a prompt, and ask the model to self-reflect on these candidate outputs. With this \emph{single-round testing}, 
we can zero in on the model's ability to use self-reflection without implicit hints about whether a given response candidate is correct or incorrect.

\begin{figure}[tbp!]
\centering
\small
  \includegraphics[width=0.80\textwidth]{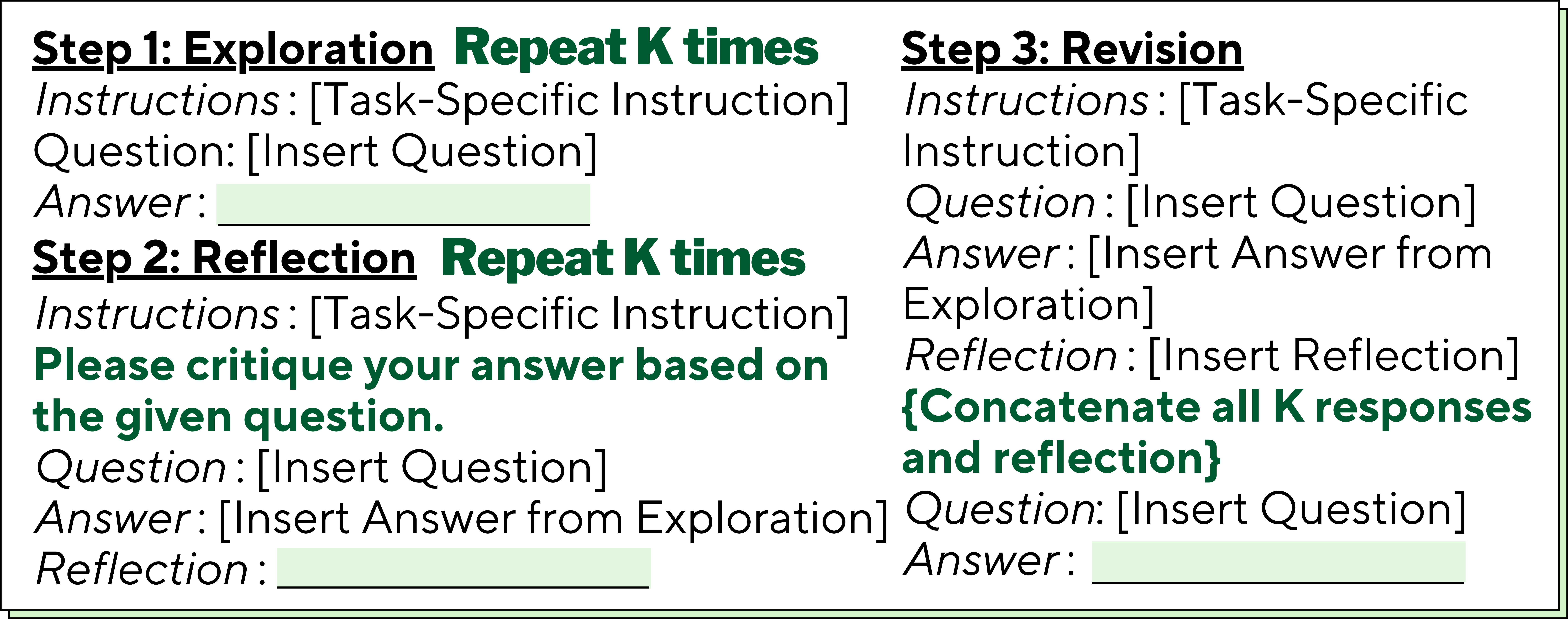}
      \vspace{-1mm}
  \caption{Example of Self-Reflection Prompting 
  }
    \label{fig:PromptingExample}
        \vspace{-4mm}
\end{figure}

Our experiments show that, in a case study with ChatGPT on different QA datasets, self-reflection in our setting yields mixed results. Specifically, self-reflection improves performance on TruthfulQA \citep{lin2022truthfulqa}, but decreases model performance in HotpotQA \citep{yang2018hotpotqa}.
Through follow-up analyses, we identify that the effectiveness of self-reflection strongly depends on the confidence in accuracy of the model's initial responses, as well as overall question difficulty as judged by humans: when the model is reliably giving correct answers from the start, self-reflection is more often harmful---however, on questions of greater difficulty, self-reflection is beneficial even when a decent percent of initial model responses are correct. We also find that self-reflection reduces model tendency toward majority voting, suggesting more sophisticated decision-making (albeit sometimes resulting in lower accuracy). 
Based on our findings, we propose a practical guideline for users to decide when to use self-reflection.

%% file: past_papers/hindsight/sections/method.tex
\section{Self-Reflection Prompting}

To focus on evaluating intrinsic reflective thinking capability, we adopt the following 
evaluation setting: 
in addition to the \citet{huang2023large} protocol of excluding external feedback and prompt optimization, we additionally disallow \emph{iterative prompting}, which samples new responses based on previous responses, creating an implicit hint to bias the model behavior \citep{huang2023large}.\footnote{We present a performance comparison between iterative prompting and non-iterative prompting in \cref{app:conditional_prompting}.} We call our approach \emph{Single-Round Self-Reflection Verification (\protocol)}. We evaluate LLMs' reflective thinking capability using the following simple three-stage format: 
1) \emph{Exploration:} Given an input $\data{}$, we prompt LLM $\model$ to generate $K$ candidate responses 
$\modelresponse{j} \sim P_\model(\modelresponse{j} | \data{}, \instruction{\text{Exploration}}), 1\leq j \leq K$ with instruction $\instruction{\text{Exploration}}$. Note that this generation of candidate responses differs from iterative prompting because each response is sampled without conditioning on any other candidate responses.
2) \emph{Reflection:} For each response $\modelresponse{j}$, we prompt $\model$ with the concatenated input $[\data{}; \modelresponse{j}]$ to generate a self-critique $\critique{j} \sim P_\model\left(\critique{j} | [\data{}; \modelresponse{j}], \instruction{\text{Reflection}}\right)$ with another instruction $\instruction{\text{Reflection}}$.
3) \emph{Revision:} 
We concatenate the $K$ response-reflection pairs into a new input and prompt $\model$ to generate an improved output. An illustration of this procedure is shown in \cref{fig:PromptingExample}.

%% file: past_papers/hindsight/sections/preliminary.tex
\section{Preliminary Study: Does Self-Reflection Prompting Work Under \protocol?}

We follow previous works~\citep{bai2022constitutional, Shinn2023ReflexionLA, huang2023large} 
in using two representative datasets, TruthfulQA and HotpotQA, to verify the effectiveness of self-reflection under \protocol. TruthfulQA is designed to evaluate the truthfulness of LMs' responses, 
while HotpotQA focuses on multi-hop reasoning tasks, aimed at requiring complex reasoning capabilities. 

\paragraph{Experiment Setup} For these experiments we set $K=4$, 
and we prompt ChatGPT-3.5 (``gpt-3.5-turbo-16k-0613'')
with the questions from each dataset.\footnote{The 16k variant is chosen to accommodate responses and reflection pairs that exceed the standard 4096 token limit, particularly in detailed experiments of \cref{sec:artificial_response}.} Our full process for making these API calls is presented in \cref{app:api_call_example}, and all prompt templates used can be found in \cref{app: prelim-prompt-templates}. We also extend our experiments to LLaMA-2~\citep{touvron2023llama_2} and Mixtral~\citep{jiang2024mixtral}, finding similar results to ChatGPT-3.5---we present results and discussion for LLaMA-2 and Mixtral in \cref{app:llama2_results} and \cref{app:mixtral_results}. 

For TruthfulQA we evaluate automatically (see details in \cref{app:truthfulqa_evaluation}).  For HotpotQA, we find that 
traditional exact match often unfairly assigns a score of $0$ for semantically correct model responses; 
therefore, we manually assess $1,000$ randomly chosen HotpotQA instances to check the model’s answers against references.  

To isolate the specific effect of the generated reflections, we also include an \textbf{exploration-only} baseline, in which we retain the Exploration stage but remove the Reflection component, and only concatenate the candidate model responses in the Revision prompt.\footnote{
The exploration-only baseline can be viewed as one implementation of (universal)  self-consistency prompting~\citep{wang2023selfconsistency, chen2023universal}. 
Rather than applying majority voting directly to the outputs, this method involves inputting these outputs back into the model for aggregation. As we'll explore in \cref{sec:mv}, we also find the model predominantly engages in a form of majority voting in this process.
} 

\begin{table}[t!]
\centering
\resizebox{0.55\textwidth}{!}{%
\begin{tabular}{@{}p{2cm}p{2cm}p{2cm}p{2cm}@{}}
\toprule
Metric & {Standard Prompting} & Exploration-Only & Self-Reflection \\
\midrule
\multicolumn{4}{c}{TruthfulQA} \\
Rouge-1 & $57.5 \pm 1.1$ & $57.2$ & $\textbf{60.8}$ \\
BLEURT & $66.8 \pm 1.9$ & $60.7$ & $\textbf{72.8}$ \\
\midrule
\multicolumn{4}{c}{HotpotQA} \\
Accuracy* & $80.3 \pm 0.5$ & $\textbf{80.8}$ & $76.2$ \\
EM & $\mathbf{50.5 \pm 0.4}$ & $47.3$ & $37.0$ \\
\bottomrule
\end{tabular}
}
\caption{Self-reflection \protocol{} experiment results on QA datasets. Bold-facing indicates the best-performing method under each metric. *Evaluated manually. 
}
\label{tab: merged_res}
\end{table}

\paragraph{Observations} Results are shown in~\cref{tab: merged_res}.
In TruthfulQA, 
we see that using self-reflection achieves significantly better performance than either the exploration-only baseline or standard prompting. 
This finding is consistent with the observation of~\citet{bai2022constitutional} that LLMs' self-evaluation (in the form of reflection) can help to produce more factual outputs. 
However, we see that on HotpotQA, accuracy when using self-reflection is about $4\%$ worse compared to both the exploration-only baseline and standard prompting. These results suggest that self-reflection may in fact harm performance in multi-hop reasoning tasks. This aligns with the self-reflection limitations found in~\citet{huang2023large}, and verifies that these limitations also extend to our more stringent evaluation setting, but presents a more complicated picture with the continued effectiveness of self-reflection on TruthfulQA under this setting.

\begin{figure}{}
\centering
\includegraphics[width=0.4\textwidth]{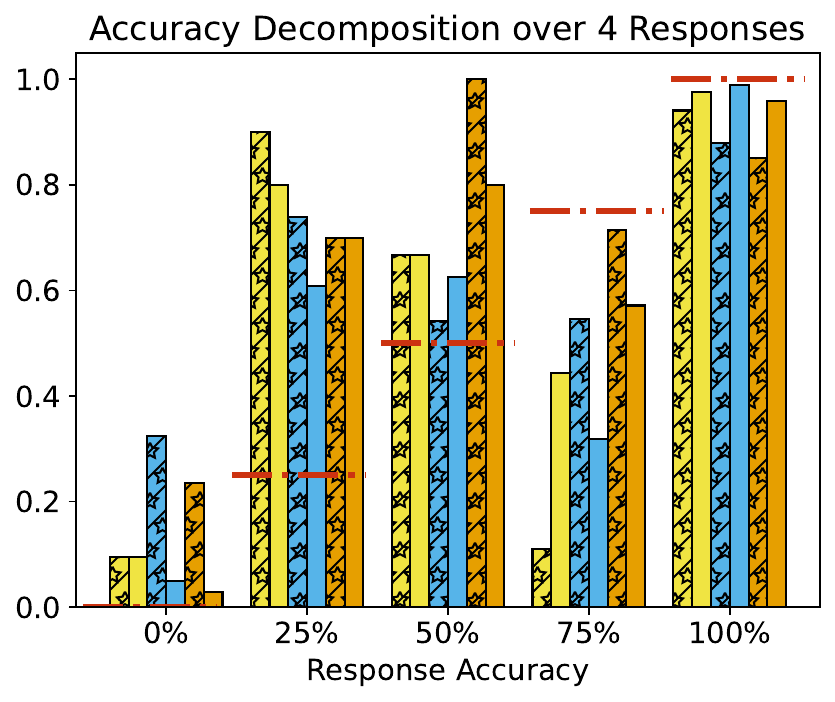}

\includegraphics[width=0.4\textwidth]{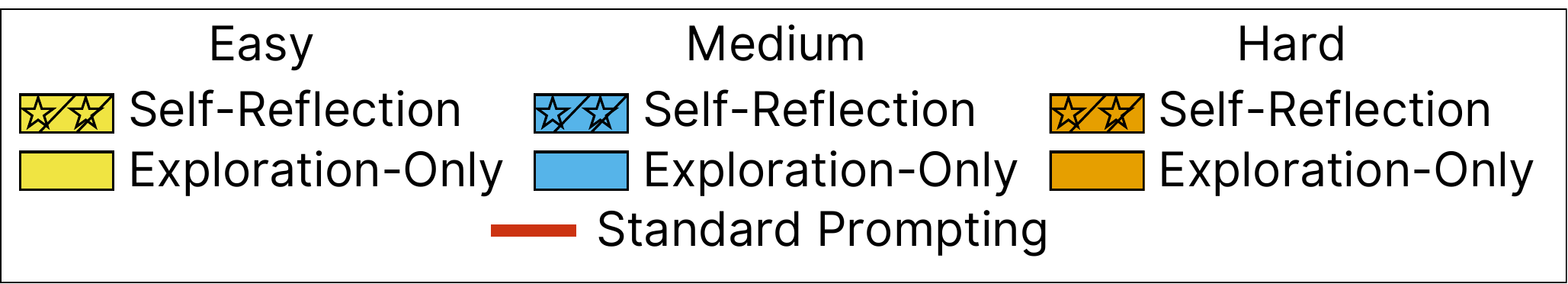}

  \caption{Performance Decomposition on Question Difficulty and Response Accuracy. 
  }

  \label{fig:perf_decomp_4response_natural}
  \vspace{-5mm}
\end{figure}

%% file: past_papers/hindsight/sections/analysis.tex
\section{Why Self-Reflection May Not Work?}
\label{sec:init_error_analysis}
To better understand these patterns, we conduct an error analysis drawing inspiration from the reflection conceptual model in psychology~\citep{hommel2023reflection}. We hypothesize that two key factors influence self-reflection's efficacy: 1) the objective \textbf{question difficulty} (quantifiable based on human annotations), 
and 2) the \textbf{model's comprehension quality} (quantifiable based on the proportion of correct responses). 
Following this framework, we can predict that if a question is above average in human-annotated difficulty, self-reflection may be of greater benefit. Similarly, if the model already has a strong grasp of the question, it may not benefit as much from self-reflection. 

To test these hypotheses, we break down model performance based on levels of question difficulty and model comprehension. We focus our analysis on HotpotQA, as this is the dataset on which we observe significant detrimental effects of applying self-reflection prompting. Additionally, this dataset contains annotated human judgments of question difficulty, and enables a clearly-defined notion of accuracy. We use these human difficulty annotations for our measure of question difficulty, and for model comprehension we use Response Accuracy (RA): the proportion of correct answers among the K candidate model responses sampled during Exploration. 

 The broken-down results are shown in \cref{fig:perf_decomp_4response_natural}. The results show an interaction between our two variables. For questions judged by humans as Easy, self-reflection shows a benefit only when the model's candidate responses are mostly---but not all---incorrect, with self-reflection otherwise having negligible or negative effects on performance. For questions judged as Medium, there is a more even split: when most or all of the model's candidate responses are wrong, self-reflection is beneficial, but when half or more of the responses are correct, self-reflection is often harmful---with the notable exception of the 75\% RA bin. A similar pattern is seen for questions judged as Hard, though for this category self-reflection is more consistently beneficial through the 75\% RA bin, showing harm to performance only when all candidate model responses are already correct.

\begin{figure}[t!]
\centering
\small
  \includegraphics[width=0.482\textwidth]{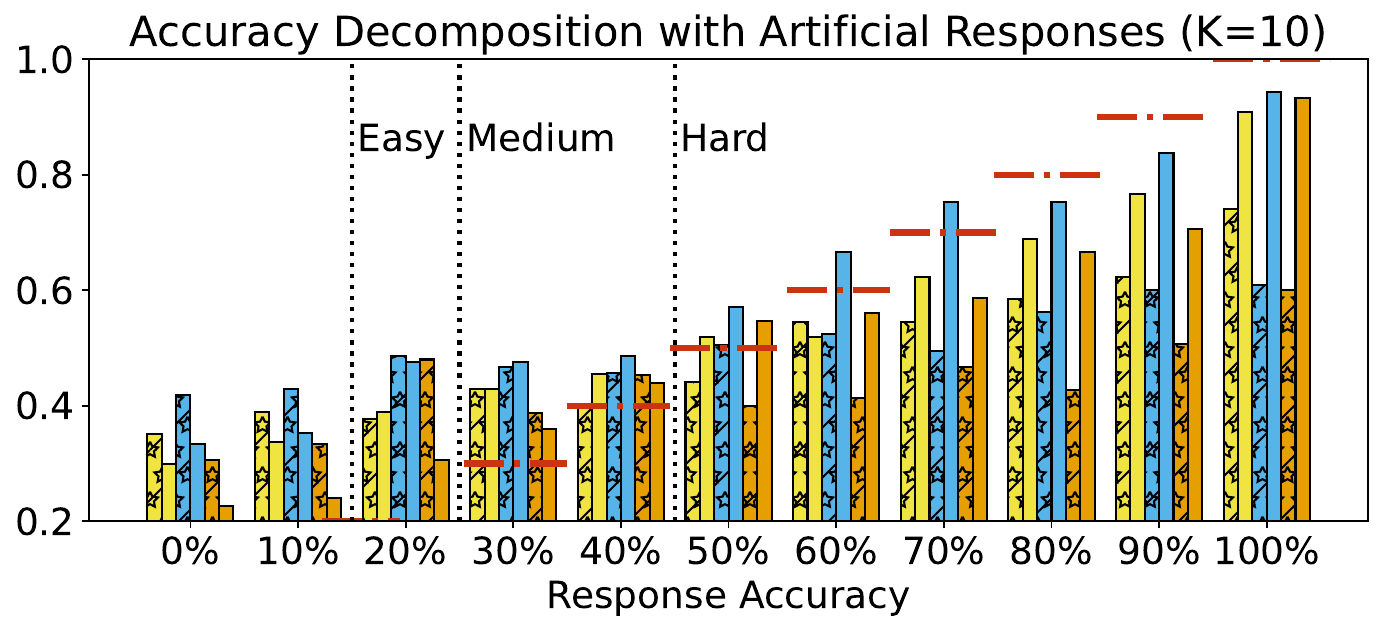}
    \includegraphics[width=0.4\textwidth]{img180.pdf}
  \vspace{-2mm}
  \caption{Performance Decomposition on Question Difficulty and Response Accuracy (Artificial Responses). Dotted lines show ``turning points'' at which reflection loses effectiveness, for Easy/Medium/Hard questions. }
  \label{fig:10level_fake}
  \vspace{-5mm}
\end{figure}

%% file: past_papers/hindsight/sections/artificial.tex
\section{Error Analysis via Artificial Response}
\label{sec:artificial_response}
The above analysis suggests an interaction between difficulty and comprehension variables in effectiveness of self-reflection---however, our ability to disentangle these effects is limited by imbalanced distribution of model comprehension relative to question difficulty. To assess the interaction more thoroughly, we simulate model ``mis-comprehension'' across a wider range of question difficulties, by sampling model responses to minimally edited versions of the prompts, and then pairing these responses with the original prompts when eliciting self-reflection. This allows us to increase the number of incorrect candidate responses, and thus to more evenly distribute RA levels across human difficulty levels. More details on this simulation process can be found in \cref{app:artificial_response_generation}.

For this experiment, we generate K $=10$ candidate responses per question, with a mix of synthetic pairings and real pairings.\footnote{We also plot the performance decomposition over K=4 artificial responses in \cref{app:fig_perf_decomp_art_resp}.} 
Results are shown in
\cref{fig:10level_fake}. We see that the benefits of self-reflection are now limited to the lowest RA levels, and there is also now a clearer shift from beneficial to harmful effects of self-reflection as RA increases. We also see that the interaction with question difficulty remains: the turning point from beneficial to harmful falls around 50\% RA for Hard questions, 30\% for Medium questions, and 20\% for Easy questions. Overall, this indicates that a major contributor to the effectiveness of self-reflection is the confidence of model accuracy on the question---if the model is reliably correct on initial responses, self-reflection tends to be harmful. However, this effect is further modulated by overall question difficulty: the benefits of self-reflection persist to higher levels of response accuracy if the questions are more difficult based on human judgment.

Though TruthfulQA is not as conducive to exact quantification of our variables, based on these results we can now speculate that the effectiveness of self-reflection on that dataset may be attributable to lower rate of good initial model responses, and potentially also higher overall question difficulty.

\begin{figure}[htbp!]
\centering
  \includegraphics[width=0.4\textwidth]{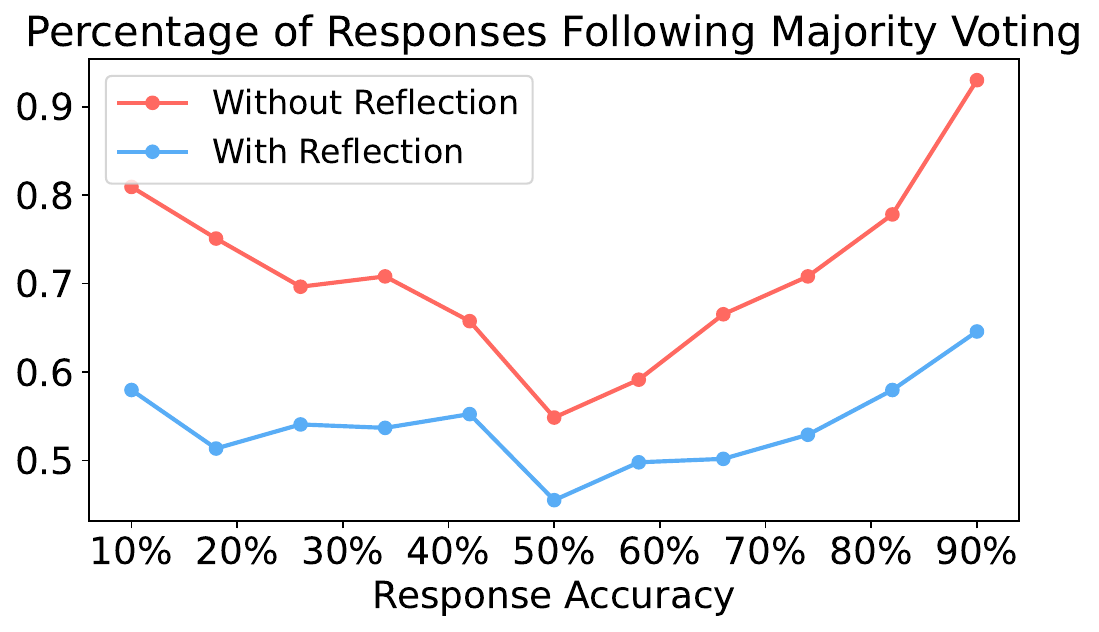}
  \caption{Majority Voting Analysis}
    \label{majority_voting}
    \vspace{-4mm}
\end{figure}

%% file: past_papers/hindsight/sections/voting.tex
\section{Effects on Majority Voting}\label{sec:mv}
A natural question to ask at this point is to what extent the effect of RA is due to the model employing majority voting on the candidate responses.
In \cref{majority_voting} we plot the percentage of items in which the model's output is consistent with majority voting, at different RA levels (computed at $K=10$ including artificially generated responses), both with and without self-reflection. The plot shows that without self-reflection, the tendency to give answers consistent with majority voting is strong and closely correlated with the strength of the accuracy trend (i.e., more majority voting when most candidate responses are either correct or incorrect, and less majority voting when candidates are more mixed). However, \emph{with} self-reflection the tendency to align with majority voting is significantly reduced across RA levels, suggesting that self-reflection does encourage more sophisticated decision strategies (even if in the case of higher RA levels, this in fact has a harmful effect on accuracy).

\begin{figure}[tbp!]
\centering
\small
  \includegraphics[width=0.8\textwidth]{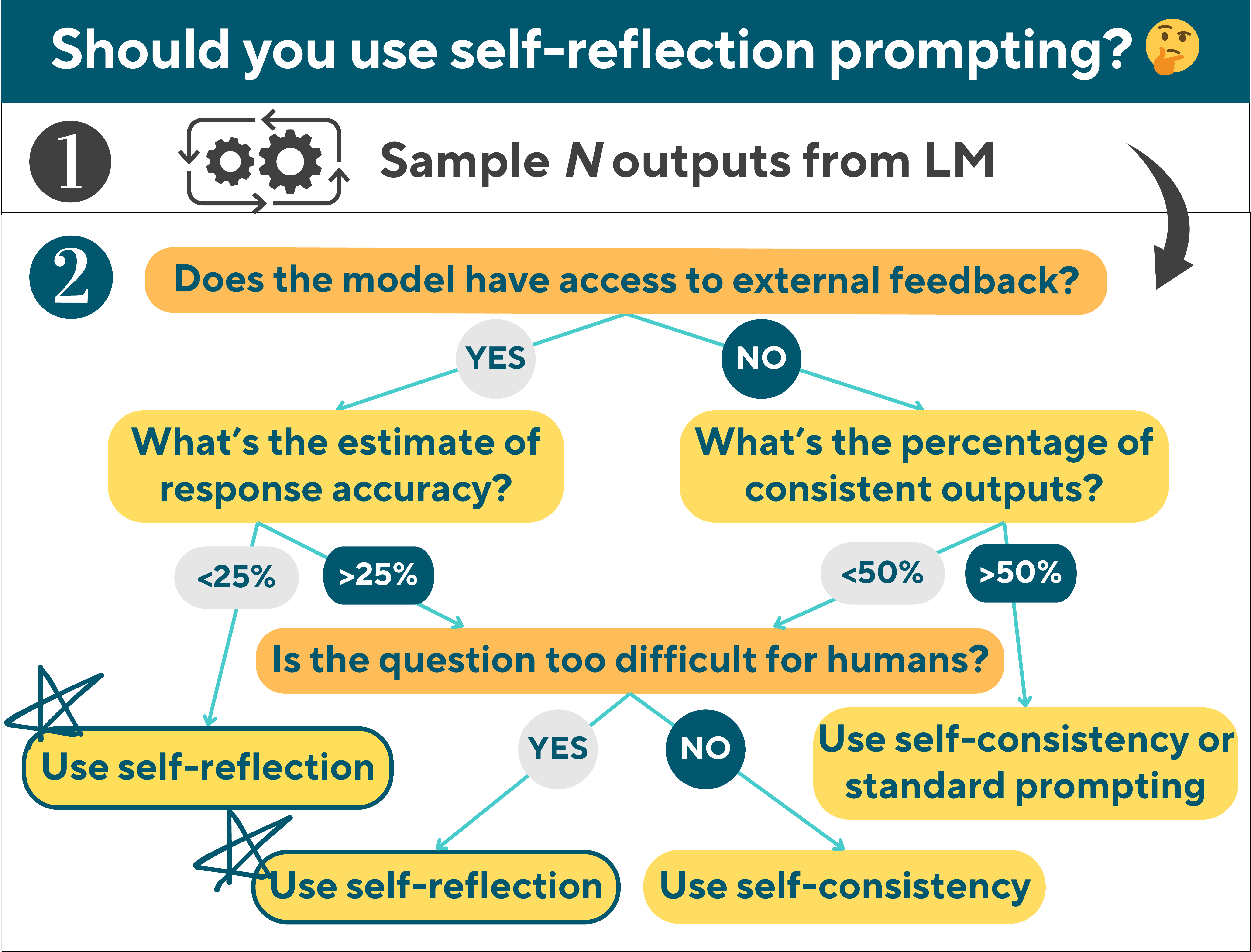}
  \caption{Proposed guide for using Self-Reflection. 
  } 
    \label{fig:guideline_self_reflection}
\end{figure}

%% file: past_papers/hindsight/sections/discussion.tex
\section{Discussion} 
Our analyses above have found that self-reflection benefits are limited to cases in which model accuracy is unreliable on initial responses, though benefits are more persistent for harder questions.
Based on these findings, we propose a set of guidelines for determining when to implement self-reflection in practical applications, for a given request or prompt. The core principle involves basing decisions on estimated RA and question difficulty, and these guidelines can be applied by simply sampling responses for the target question or prompt. First, if external tools or certain access to ground truth answers are available such that RA can be reliably estimated, then self-reflection should be used when RA levels are low. Next, if difficulty annotations/subjective difficulty judgements are available, self-reflection can also be promising when RA levels are intermediate and question difficulty is high. If RA cannot be estimated, response consistency can be used as a proxy: if responses are highly consistent, self-reflection may be unlikely to provide benefit. If consistency is low, then self-reflection may be beneficial, especially for questions of higher difficulty.
An illustration of these guidelines is in \cref{fig:guideline_self_reflection}.

%% file: past_papers/hindsight/sections/conclusion.tex
\section{Conclusion}
In this paper, we evaluate ChatGPT's self-reflective capabilities under a stringent single-round multi-response evaluation setting. We find mixed results, and further analysis shows that the effectiveness of self-reflection is impacted both by question difficulty and by model response accuracy level: benefits of self-reflection are mostly limited to cases in which the model's initial responses are unreliable in accuracy, but with more persistent benefits for harder questions. Additionally, we find that self-reflection reduces the model's tendency for majority voting. We propose guidelines for when to use self-reflection, and we look forward to work further exploring impacts on self-reflection, and further refining these guidelines.

%% file: past_papers/hindsight/sections/limitations.tex
\section*{Limitations}

In this work, we adopt a stringent evaluation strategy to test the effectiveness of self-reflective abilities of LLMs. One limitation is that our experiments reported in the main text are based on a single snapshot of the ChatGPT model (gpt-3.5-turbo-16k-0613). We focus on ChatGPT because it is a state-of-the-art chat model, allowing us to make our results directly comparable with previous work---and we limit to this particular version of ChatGPT to ensure that results will not be affected by model updates. However, the assessment of self-reflection may vary between different versions of ChatGPT, as well as between ChatGPT and other LLMs. We do verify our experimental results on other open language models including LLaMA-2~\citep{touvron2023llama_2} and Mixtral~\citep{jiang2024mixtral} in \cref{app:llama2_results} and \cref{app:mixtral_results}, respectively. While we find that our conclusions can be extended to these models, due to budget limitations we leave more extensive evaluation over other popular proprietary and open models for future works.  

Our experiments also use only two datasets for evaluating reflective ability. We chose these two datasets for a focused study covering two very different QA domains, but we look forward to future work further extending these types of analyses to a broader collection of datasets.

We conducted an artificial response experiment in \cref{sec:artificial_response} to simulate the real output distribution of the language model. This is a rough estimate of ChatGPT's actual output distribution. As we sampled ten fake responses from the language model, it is impossible to cover all possible cases of outputs, and there might be bias in the sample distribution. Future work could try generating a higher number of fake responses to obtain a more accurate distribution of the model.

Finally, although RA proves a valuable metric for determining the utility of self-reflection, its reliance on access to ground truth undermines its practical use. An initial attempt to use GPT-4 to produce an estimate of RA yielded unsatisfactory results (detailed in \cref{app:challenge_predict_correctness_margin}). Further examination of this topic is reserved for future research.

%% file: chapters/amortized.tex
\section*{Chapter Overview}
Despite the popularity of Shapley Values in explaining neural text classification models, computing them is prohibitive for large pretrained models due to a large number of model evaluations. In this chapter, we develop an amortized model that directly predicts each input feature's Shapley Value without additional model evaluations, achieving significant speedup while maintaining stability.

\graphicspath{{./}}

\makeatletter
\def\input@path{{./}}
\makeatother

\input{past_papers/amortized_interpretability/sections/introduction}

\input{past_papers/amortized_interpretability/sections/related_work}

\input{past_papers/amortized_interpretability/sections/background}

\input{past_papers/amortized_interpretability/sections/stability}

\input{past_papers/amortized_interpretability/sections/method}

\input{past_papers/amortized_interpretability/sections/experiments}

\input{past_papers/amortized_interpretability/sections/evaluation}

\input{past_papers/amortized_interpretability/sections/conclusion}

\input{past_papers/amortized_interpretability/sections/limitations}

%% file: past_papers/amortized_interpretability/sections/introduction.tex
\section{Introduction}
Many powerful natural language processing (NLP) models used in commercial systems only allow users to access model outputs. When these systems are applied in high-stakes domains, such as healthcare, finance, and law, it is essential to interpret how these models come to their decisions. 
To this end, post-hoc black-box explanation methods have been proposed to identify the input features that are most critical to model predictions~\citep{ribeiro2016lime, lundberg2017unified}. A famous class of post-hoc black-box local explanation methods takes advantage of the Shapley Values~\citep{shapley:book1952} to identify important input features, such as Shapley Value Sampling (SVS)~\citep{strumbelj2010efficient} and KernelSHAP (KS)~\citep{lundberg2017unified}.
These methods typically start by sampling  permutations of the input features (``\emph{perturbation samples}'') and aggregating 
model output changes 
over the perturbation samples. Then, they assign an \emph{explanation score} for each input feature to indicate its contribution to the prediction.
\begin{figure}[t]
\small
\centering
\begin{minipage}{0.8\textwidth}
\centering
\begin{tabular}{P{1.0cm}|c}
\toprule
\textbf{Seed=1} & \adjincludegraphics[width=0.78\columnwidth,valign=c,trim={0 {.38\height} 0 {.05\height}},clip]{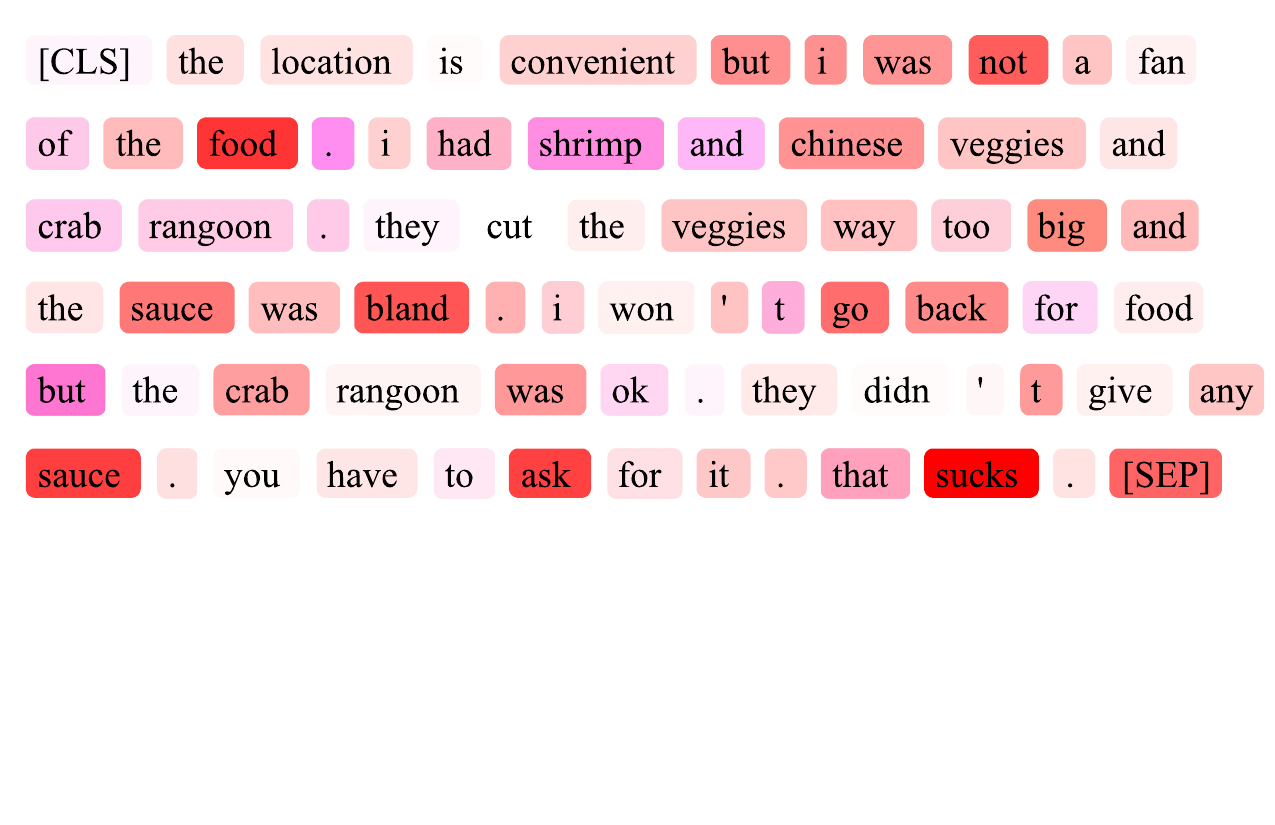} \\
\midrule
\textbf{Seed=2} & \adjincludegraphics[width=0.78\columnwidth,valign=c,trim={0 {.38\height} 0 {.05\height}},clip]{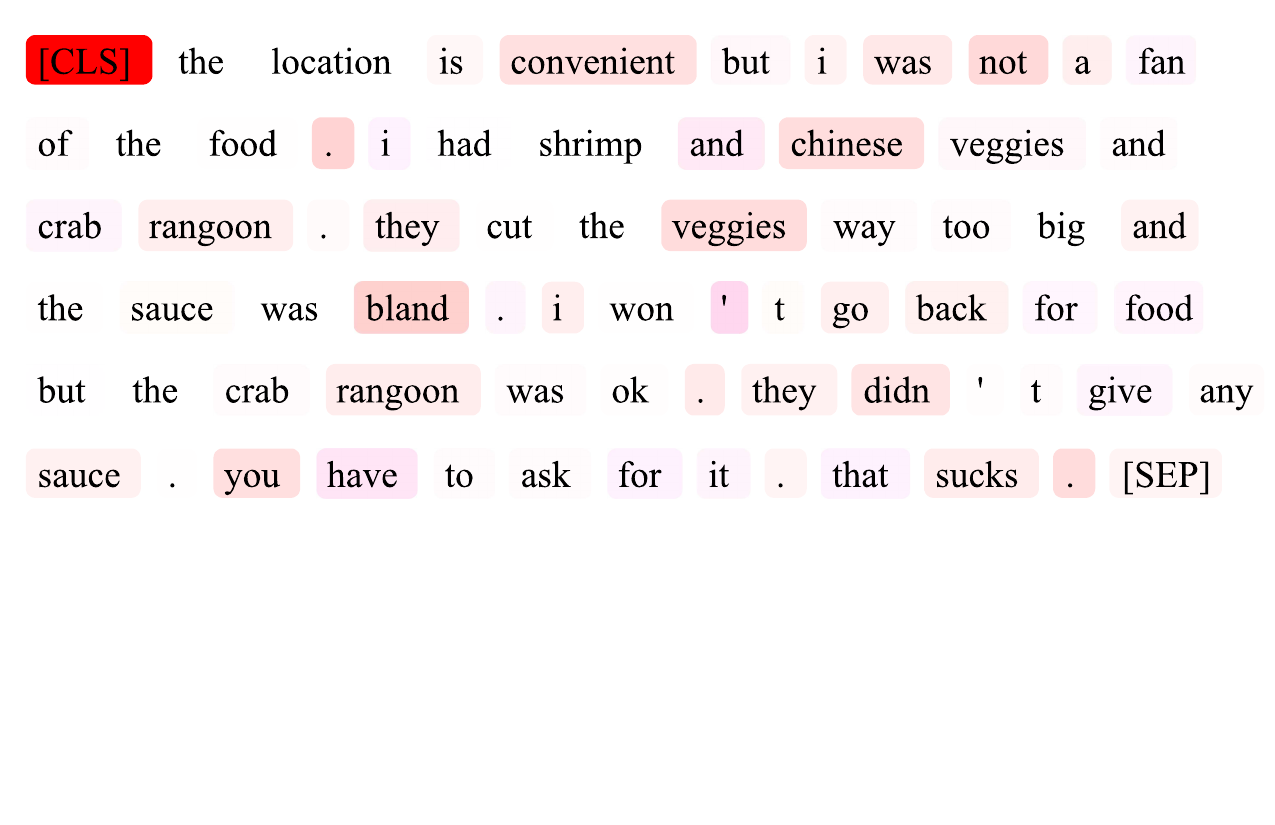} \\
\bottomrule
\end{tabular}
\end{minipage}
\caption{Heatmaps of explanation scores  of an example from Yelp-Polarity based on two runs of KernelSHAP (KS)  using different random seeds. KS is run on a fine-tuned BERT model using $200$ samples per instance (approx. $3.47$s per instance on average using a single A100 GPU, more than $150$ times slower than one forward inference of the BERT model). The darker each token is, the higher its explanation score. Clearly, interpretation results are significantly different when using different seeds.
}
\label{fig: intro_example}
\end{figure}

\begin{figure*}[tbp!]
\centering
\includegraphics[trim={0cm 2cm 0cm 0cm},clip, height=6.21cm]{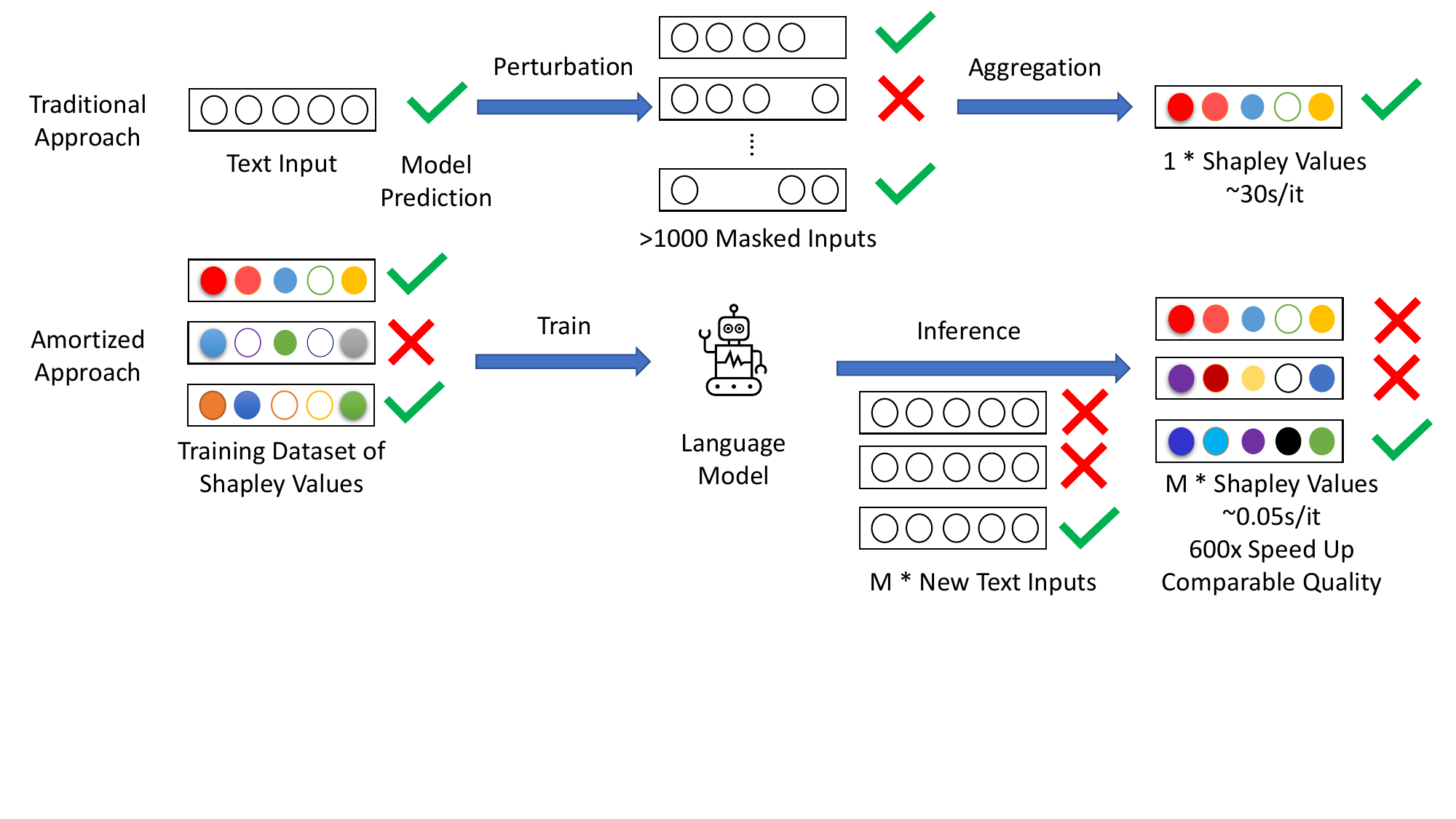}
\vspace{-11mm}
\caption{Illustration of our proposed Amortized Model. Black-outlined circles represent original inputs without Shapley Values, while circles with colored outlines or colored fills denote inputs with Shapley Values.
}
\label{fig: intro_amortized_model}
\end{figure*}

Despite the widespread usage of Shapley Values methods, we observe that when they are applied to text data, the estimated explanation score for each token varies significantly with the random seeds used for sampling.
\cref{fig: intro_example} shows an example of interpreting a BERT-based sentiment classifier~\citep{devlin2019bert} on Yelp-Polarity dataset, a restaurant 
review dataset~\citep{zhangCharacterlevelConvolutionalNetworks2015} by KS. The set of tokens with high explanation scores varies significantly when using different random seeds.
They become stable only when the number of perturbation samples increases to more than 2,000.
As KS requires model prediction for each perturbation sample, the inference cost can be substantial.
For example, it takes about 183 seconds to interpret each instance in Yelp-Polarity 
using the KS Captum implementation~\citep{kokhlikyan2020captum} on an A100 GPU.
In addition, this issue becomes more severe when the input text gets longer, as more perturbation samples are needed for reliable estimation of Shapley Values. This sensitivity to the sampling process leads to an unreliable interpretation of the model predictions and hinders developers from understanding model behavior. 

 To achieve a better trade-off between efficiency and stability, we propose a simple yet effective amortization method to estimate the explanation scores. Motivated by the observation that different instances might share a similar set of important words (e.g., in sentiment classification, emotional words are strong label indicators~\citep{taboada2011lexicon}), an amortized model can leverage similar interpretation patterns across instances when predicting the explanation scores.
Specifically, we amortize the cost of computing explanation scores by precomputing them on a set of training examples and train an amortized model to predict the explanation scores given the input.
At inference time, our amortized model 
directly outputs explanation scores for new instances. Although we need to collect a training set for every model we wish to interpret, our experiments show that with as few as 5000 training instances, the amortized model achieves high estimation accuracy. We show our proposed amortized model in \cref{fig: intro_amortized_model}.%

The experimental results demonstrate the efficiency and effectiveness of our approach. First, our model reduces the computation time from about 3.47s per instance to less than 50ms,\footnote{On Yelp-Polarity dataset and using A100 GPU, we compare with typical KS running with 200 samples.} which is 60 times faster than the baseline methods. 
Second, our model is robust to randomness in training (e.g., random initialization, random seeds used for generating reference explanation scores in the training dataset), and produces stable estimations over different random seeds. Third, we show that the amortized model can be used along with SVS to perform \emph{local adaption}, i.e., adapting to specific instances at inference time, thus further improving performance if more computation is available (\ref{exp: local_adaption}). Finally, we evaluate our model from the functionality perspective~\citep{doshi2017towards, ye2022can} 
by examining the quality 
of the explanation in downstream tasks. We perform case studies on feature selection and domain calibration using the estimated explanation scores, and show that our method outperforms 
the computationally expensive KS method.

%% file: past_papers/amortized_interpretability/sections/related_work.tex
\section{Related Works}

\shortparagraph{Post-Hoc Local Explanation Methods}
Post-hoc local
explanations are proposed to understand the prediction process of neural models~\citep{simonyan2013deep, ribeiro2016lime, lundberg2017unified, shrikumar2017learning}. They work by assigning an explanation score to each feature (e.g., a token) in an instance (``local'') to indicate its contribution to the model prediction. In this paper, we focus on studying 
KernelSHAP (KS)~\citep{lundberg2017unified}, an \textit{additive feature attribution method} that estimates the Shapley Value~\citep{shapley:book1952} for each feature. 

There are other interpretability methods in NLP. 
For example, gradient-based methods~\citep{simonyan2013deep, li2016visualizing}, which use the gradient w.r.t. each input dimension as a measure for its saliency. Reference-based methods~\citep{shrikumar2017learning, sundararajan2017axiomatic} consider the model output difference between the original input and reference input (e.g., zero embedding vectors). 

\shortparagraph{Shapley Values Estimation}  Shapley Values are concepts from game theory to attribute total contribution to individual features. However, in practice estimating Shapley values requires prohibitively high cost for computation, especially when explaining the prediction on long documents in NLP. KS works as an efficient way to approximate Shapley Values. Previous work on estimating Shapley Values mainly focuses on accelerating the sampling process~\citep{jethani2021fastshap, covert2021improving, parvez-chang-2021-evaluating, mitchell2022sampling} or removing redundant features~\citep{aas2021explaining, covert2021explaining}. In this work, we propose a new method to combat this challenge by training an amortized model.

\shortparagraph{Robustness of Local Explanation Methods} Despite being widely adopted, there has been a long discussion on the actual quality of explanation methods. Recently, people have found that explanation methods can assign substantially different attributions to similar inputs~\citep{alvarez2018robustness, ghorbani2019interpretation, kindermans2019reliability, yeh2019fidelity, slack2021reliable, yin2022sensitivity}, i.e., they are not robust enough, which adds to the concerns about how faithful these explanations are~\citep{doshi2017towards, Adebayo2018SanityCF, jacovi2020towards}. In addition to previous work focusing on robustness against input perturbations, we demonstrate that even just changing the random seeds can cause the estimated Shapley Values to be weakly-correlated with each other, unless a large number of perturbation samples are used (which incurs high computational cost).

\shortparagraph{Amortized Explanation Methods} Our method is similar to recent works on amortized explanation models including  CXPlain~\citep{schwab2019cxplain} and FastSHAP~\citep{jethani2021fastshap}), where they also aim to improve the computational efficiency of explanation methods. 
The key differences  are: 
1) We do not make causal assumptions between input features and model outputs; and 2) we focus on text domains, where each feature is a discrete token (typical optimization methods for continuous variables do not directly apply).

%% file: past_papers/amortized_interpretability/sections/background.tex
\section{Background}
\label{sec:bg}
In this section, we  briefly review the basics of Shapley Values, focusing on its application to the text classification task.

\shortparagraph{Local explanation of black-box text classification models.}
In text classification tasks, inputs are usually sequences of discrete tokens 
$\data{}=[\sword{1}, \sword{2}, \dots, \sword{L}]$. 
Here 
$L$ 
is the length of $\data{}$ and may vary across examples;
$\sword{j}$ 
is the
$j$-th token of $\data{}$.
The classification model  $\modelclf$ takes the input 
$\data{}$ 
and predict the label as 
$\hat{y} = \argmax_{y \in \mathcal{Y}} \modelio{\modelclf}{\data{}}{y}$.
Local explanation methods treat each data instance independently
and compute an explanation score $\interpret{}{j}{y}$,
representing the contribution of $\sword{j}$ to the label $y$.
Usually, we care about the explanation scores when $y=\hat{y}$. 

\shortparagraph{Shapley Values (SV)}
are concepts from game theory originally developed to assign credits in cooperative games~\citep{shapley:book1952, strumbelj2010efficient, lundberg2017unified, covert2021explaining}.
Let $s \in \left\{0, 1\right\}^L$ be a masking of the input and define $\data{s} \defeq \left\{\sword{i} \right\}_{i:s_i=1}$ as the \textit{perturbed input} 
that consists of unmasked tokens $x_i$ (where the corresponding mask $s_i$ has a value of 1). In this paper, we follow the common practice~\citep{ye2021connecting, ye2022can, yin2022sensitivity} to replace masked tokens with \texttt{[PAD]} in the input before sending it to the classifier. 
Let $|s|$ represent the number of non-zero terms in $s$. Shapley Values $\interpret{\text{SV}}{i}{y}$  ~\citep{shapley:book1952} are computed by:
\begin{align}
\small
    \interpret{\text{SV}}{i}{y} = \frac{1}{L} \sum_{s: s_i \neq 1} {L-1 \choose |s|}^{-1} \left( \modelio{\modelclf}{\data{s } \cup \left\{\sword{i}\right\} }{y} - \modelio{\modelclf}{\data{s}}{y} \right).
\end{align}
Intuitively, 
$\interpret{\text{SV}}{i}{y}$ computes the marginal contributions of each token 
to the model prediction.

Computing SV is known to be NP-hard~\citep{deng1994complexity}.
In practice, we estimate Shapley Values approximately for efficiency. 
Shapley Values Sampling (SVS)~\citep{castro2009polynomial, strumbelj2010efficient} is a widely-used Monte-Carlo estimator of SV: 

{\small
\begin{align}
\small
        \interpret{\text{SVS}}{i}{y} = \frac{1}{m} \sum_{\substack{\sigma_j \in \Pi(L)\\ 1\leq j\leq m}}\sum_{i \in \sigma_j} \left[ \modelio{\modelclf}{\data{\mathbb{S}\left([\sigma_j]_{i-1} \cup \left\{i\right\}\right)}}{y} -\modelio{\modelclf}{\data{\mathbb{S}\left([\sigma_j]_{i-1}\right)}}{y} \right].  \label{eq: mc_shapley} 
\end{align}
}%
Here $\sigma_j \in \Pi(L)$ is the sampled \textbf{ordering} and 
 $[\sigma_j]$ is the non-ordered \textbf{set} of indices for $\sigma_j$. 
$[\sigma_j]_{i-1}$ represents the \textbf{set}  of indices ranked lower than $i$ in $\sigma_j$ . $\mathbb{S}([\sigma_j])$ maps the indices \text{set} $[\sigma_j]$
to a mask $s \in \{0, 1\}^L$ such that $s_i=\mathbf{1}[i \in [\sigma_j]]$. $m$ is the number of \emph{perturbation samples} used for computing SVS. 

\shortparagraph{KernelSHAP}
Although SVS has successfully reduced the exponential time complexity to polynomial, it still requires sampling permutations and needs to do sequential updates 
following sampled orderings and computing the explanation scores,
which is an apparent efficiency bottleneck. 
\citet{lundberg2017unified} introduce a more efficient estimator, KernelSHAP (KS), which allows better parallelism 
and computing explanation scores for all tokens at once using linear regression. That is achieved by showing that computing SV is equivalent to solving the following optimization problem:
\begin{align}
     &\interpret{\text{KS}}{\cdot}{y} \approx \argmin_{\interpret{}{\cdot}{y}} \frac{1}{m}  \sum_{\substack{s(k) \sim p(s)\\1\leq k\leq m}}[\modelio{\modelclf}{\data{s(k)}}{y}  - \vec{s}(k)^T\interpret{}{\cdot}{y}]^2 \label{eq: kernelshap-approx}, \\
     &\textrm{s.t.} \quad  \mathbf{1}^T\interpret{}{\cdot}{y} = \modelio{\modelclf}{\data{}}{y} - \modelio{\modelclf}{\varnothing}{y}, \nonumber
     \label{eq: sufficiency constraints}
\end{align}
where $\vec{s}(k)$ is the one-hot vector
corresponding to the mask\footnote{Note, $s(k)$ is the $k$-th \textbf{mask sample} while $s_i \in \{0, 1\}$ is the $i$-th dimension of the \textbf{mask sample} $s$.} $s(k)$
sampled from the Shapley Kernel  $p(s) = \frac{L-1}{{L \choose |s|}|s|(L-|s|)}$.
$m$ is again the number of perturbation samples. 
We will use ``SVS-$m$'' and ``KS-$m$'' in the rest of the paper to indicate the sample size for SVS and KS. 
In practice, the specific perturbation samples depend on the random seed of the sampler, and we will show that the explanation scores are highly sensitive to the random seed under a small sample size. 

Note that the larger the number of perturbation samples, the more model evaluations are required for a single instance, which can be computationally expensive for large Transformer models. Therefore, the main performance bottleneck is the number of model evaluations.

%% file: past_papers/amortized_interpretability/sections/stability.tex
\section{Stability of Local Explanation}
\label{sec: stability}
One of the most common applications of SV is feature selection, which selects the most important features by following the order of the explanation scores. People commonly use KS with an affordable number of perturbation samples in practice (the typical numbers of perturbation samples used 
in the literature
are around $25$, $200$, $2000$). 
 However, as we see in \cref{fig: intro_example}, the ranking of the scores can be quite sensitive to random seeds when using stochastic estimation of SV. 
In this section, we investigate this stability issue. We demonstrate stochastic approximation of SV is unstable 
in text 
classification tasks under common settings, especially with long texts. In particular, when ranking input tokens based on explanation scores, Spearman's correlation between rankings across different runs is low. 

\shortparagraph{Measuring ranking stability.} 
Given explanation scores produced by different random seeds using an SV estimator, we want to measure the difference between these scores. Specifically, we are interested in the difference in the rankings of the scores as this is what we use for feature selection. To measure the ranking stability of multiple runs using different random seeds,
we compute Spearman's correlation between any two of them and use the average Spearman's correlation as the measure of the ranking stability. 
In addition, we follow~\citet{ghorbani2019interpretation} to report Top-K 
intersections
between two rankings, since in many applications only the top features are of explanatory interest. We measure the size of the intersection 
of Top-K features from two different runs.
\begin{table*}[!ht]
\small
\centering
\begin{tabular}{lccccr}
\hline
Setting & Spearman & Top-5 Inter. & Top-10 Inter. & MSE & Running Time \\ \hline
SVS-25 & $0.84 (\pm 0.00)$ & $3.41 (\pm 0.00)$ & $7.02 (\pm 0.00)$ & $0.01 (\pm 0.00)$  & $183.72$s/it \\
KS-25 & $0.04 (\pm 0.00)$ & $0.43 (\pm 0.01)$ & $1.45 (\pm 0.01)$ & $0.00 (\pm 0.00)$ & $1.92$s/it \\
KS-200 & $0.16 (\pm 0.00)$ & $1.09 (\pm 0.01)$ & $2.47 (\pm 0.00)$ & $0.82 (\pm 0.29)$ & $3.47$s/it \\
KS-2000 & $0.37 (\pm 0.00)$ & $2.45 (\pm 0.01)$ & $4.38 (\pm 0.05)$ & $0.03 (\pm 0.00)$ & $33.40$s/it \\
KS-8000 & $0.63 (\pm 0.00)$ & $3.73 (\pm 0.02)$ & $6.93 (\pm 0.01)$ & $0.01 (\pm 0.00)$ & $123.29$s/it \\ 
\hline
\end{tabular}%
\caption{Ranking stability experiments on the Yelp-Polarity dataset. Each local explanation setting is evaluated across 5  runs with different random seeds. ``Top-K Inter.'' denotes top-K intersection. 
All values in this table are absolute values. Here we can see a clear trade-off between stability and computation cost. 
}
\label{tab:ranking_instability_yelp}
\end{table*}
\begin{table*}[!ht]
\small
\centering
\begin{tabular}{lccccr}
\hline
Setting & Spearman & Top-5 Inter. & Top-10 Inter. & MSE & Running Time \\ \hline
SVS-25 & $0.75 (\pm 0.00)$ & $3.54 (\pm 0.02)$ & $7.46 (\pm 0.02)$ & $0.02 (\pm 0.00)$ & $128.07$s/it \\
KS-25 & $0.06 (\pm 0.00)$ & $0.97 (\pm 0.01)$ & $3.41 (\pm 0.03)$ & $0.01 (\pm 0.00)$ & $0.33$s/it \\
KS-200 & $0.24 (\pm 0.00)$ & $1.79 (\pm 0.01)$ & $4.37 (\pm 0.03)$ & $0.07 (\pm 0.00)$ & $2.04$s/it \\
KS-2000 & $0.52 (\pm 0.00)$ & $3.19 (\pm 0.00)$ & $6.09 (\pm 0.00)$ & $0.03 (\pm 0.00)$ & $20.39$s/it \\
KS-8000 & $0.76 (\pm 0.00)$ & $4.08 (\pm 0.02)$ & $7.74 (\pm 0.02)$ & $0.01 (\pm 0.00)$ & $89.48$s/it \\ \hline
\end{tabular}%
\caption{Ranking stability experiments on the MNLI dataset. 
}
\label{tab:ranking_instability_mnli}
\end{table*}

\shortparagraph{Setup.} We conduct our experiments on the validation set of the Yelp-Polarity~\citep{zhangCharacterlevelConvolutionalNetworks2015} and MNLI~\citep{williams2018broad} datasets. Yelp-Polarity is a binary sentiment classification task and MNLI is a three-way textual entailment classification task. 
We conduct experiments on $500$ random samples with balanced labels (we refer to these datasets as ``Stability Evaluation Sets'' subsequently). Results are averaged over $5$ different random seeds.\footnote{We take more than 2,000 hours on a single A100 GPU for all experiments in this section.} We use the publicly available fine-tuned BERT-base-uncased checkpoints\footnote{Yelp-Polarity: \url{https://huggingface.co/textattack/bert-base-uncased-yelp-polarity}\\MNLI: \url{https://huggingface.co/textattack/bert-base-uncased-MNLI}}~\citep{morris2020textattack} as the target models to interpret and use the implementation of Captum~\citep{kokhlikyan2020captum} to compute the explanation scores for both KS and SVS. 
For each explanation method, we test with the recommended numbers of perturbation samples\footnote{For SVS, the recommended number of perturbation samples is $25$ in Captum. For KS, to our best knowledge, the typical numbers of perturbation samples used in previous works are $25, 200, 2000$. We also include KS-$8000$ to see how stable KS can be given much longer running time.}
 used to compute the explanation scores for every instance. For Top-K intersections, we report results with  $K=5$ and $K=10$.

\shortparagraph{Trade-off between stability and computation cost.} The ranking stability results are listed in \cref{tab:ranking_instability_yelp} 
and \cref{tab:ranking_instability_mnli} 
for Yelp-Polarity and MNLI
datasets.
We observe that using 25 to 200 perturbation samples, the stability of the explanation scores is low (Spearman's correlation is only 0.16).
Sampling more perturbed inputs 
makes the scores more stable. However, the computational cost explodes at the same time, going from one second to two minutes per instance. 
To reduce the sensitivity to an acceptable level (i.e., making the Spearman's correlation between two different runs above $0.40$, which indicates moderate  correlation~\cite{akoglu2018user}), we usually need thousands of model evaluations and spend roughly $33.40$ seconds per instance. 

\shortparagraph{Low MSE does not imply stability.} Mean Squared Error (MSE) is commonly used to evaluate the distance between two lists of 
explanation scores. 
In \cref{tab:ranking_instability_yelp}, we observe that MSE only weakly correlates with ranking stability (e.g., For Yelp-Polarity, $R=-0.41$ and $p<0.05$, so the correlation is not significant). Even when the difference of MSE for different settings is as low as 0.01, the correlation between rankings produced by explanations can still be low. Therefore, from users' perspectives, low MSEs do not mean the explanations are reliable as they can suggest distinct rankings. 

\begin{figure*}[tbp!]
\centering
\begin{subfigure}[b]{0.48\textwidth}
\centering
\includegraphics[ height=4cm]{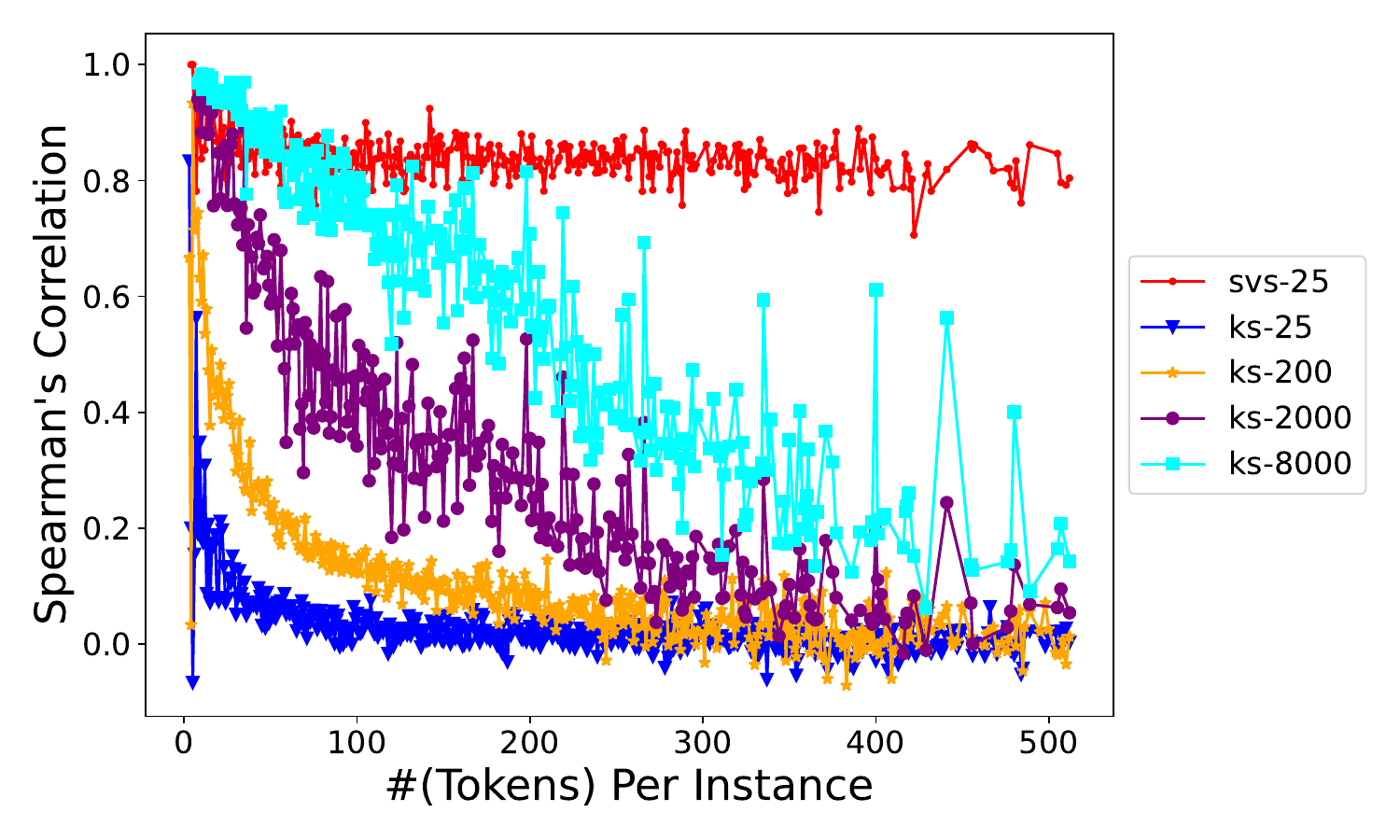}
\vspace{-3mm}
\caption{Yelp-Polarity
}
\label{fig: internal_correlation_with_length_yelp}
\end{subfigure}
\begin{subfigure}[b]{0.48\textwidth}
\small
\centering
\includegraphics[ height=4cm]{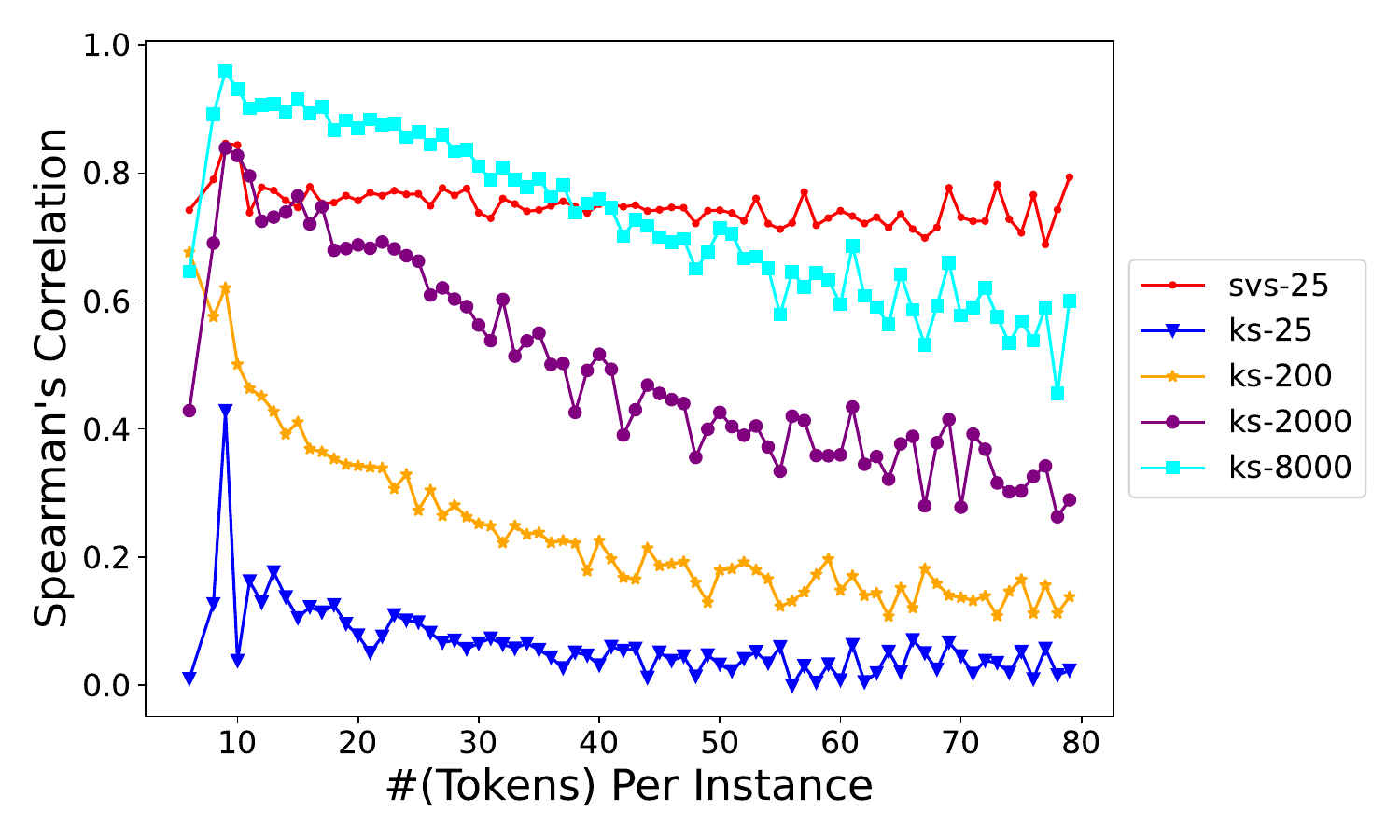}
\vspace{-3mm}
\caption{MNLI
}
\label{fig: internal_correlation_with_length_mnli}
\end{subfigure}
\caption{Ranking stability over different input lengths on Yelp-Polarity and MNLI datasets. We observe that longer input suffers more from instability. 
}
\label{fig: internal_correlation_with_length}
\end{figure*}

\shortparagraph{Longer input suffers more from instability.} We also plot the Spearman's correlation decomposed at different input lengths in \cref{fig: internal_correlation_with_length}.
Here, we observe a clear trend that the ranking stability degrades significantly even at an input length of 20 tokens. The general trend is that the longer the input length is, the worse the ranking stability. The same trend holds across datasets. As many NLP tasks involve sentences longer than 20 tokens (e.g., SST-2~\cite{socher-etal-2013-recursive}, MNLI~\cite{williams2018broad}), obtaining stable explanations to analyze NLP models can be quite challenging. 

\shortparagraph{Discussion: why Shapley Values estimation is unstable in text domain?}
One of the most prominent characteristics of the text domain is that individual tokens/n-grams can have a large impact on the label. Thus they need to be all included in the perturbation samples for an accurate estimate.
When the input length grows, the number of n-grams will grow fast. As shown in \cref{sec:bg}, the probability of certain n-grams getting sampled is drastically reduced as each n-gram will be sampled with equivalent probability. Therefore, the observed model output will have a large variance as certain n-grams may not get sampled. A concurrent work \cite{kwon2022weightedshap} 
presented a related theoretical analysis on why the uniform sampling setting in SV computation can lead to suboptimal attribution. 

%% file: past_papers/amortized_interpretability/sections/method.tex
\section{Amortized Inference for Shapley Values
}
\label{sec:amortized_interpretability_method}
Motivated by the above observation, 
we propose to train an amortized model to predict the explanation scores given an input \emph{without any model evaluation on perturbation samples}. The inference cost is thus amortized by training on a set of pre-computed reliable explanation scores. 

We build an amortized explanation model for text classification in two stages. In the first stage, we construct a training set for the amortized model. 
We compute reliable explanation scores as the reference scores for training using the existing SV estimator. As shown in \cref{sec: stability}, SVS-25 is the most stable SV estimator and we use it to obtain reference scores. 
In the second stage, we train a BERT-based amortized model that takes the text as input and outputs the explanation scores using MSE loss. 

Specifically, given input tokens $\data{}$, we  use a pretrained language model $\modelname{\mathrm{LM}}$
to encode words into $d$-dim embeddings $\vec{e} =\modelname{\mathrm{LM}}(\data{})= [\semb{1}, \dots, \semb{L(\data{})}] \in \Real^{L(\data{}) \times d}$. Then, we use a linear layer to transform each $\semb{i}$ to the predicted explanation score $\interpret{AM}{i}{\hat{y}_i} = W\semb{i} + b$. 
To train the model, we use MSE loss to fit $\interpret{AM}{i}{\hat{y}}$ to the pre-computed reference scores $\interpret{}{i}{\hat{y}}$ 
over the training set $\dataset{\text{Train}}$. This is an amortized model in the sense that there are no individual sampling and model queries for each test example $\data{}$ as in SVS and KS. When a new sample comes in, the amortized model makes a single inference on the input tokens to predict their explanation scores.  

\begin{algorithm}
\small
\caption{Local Adaption}\label{alg:meta}
\begin{algorithmic}
\Require $m$: the desired number of local adaption perturbation samples, $\modelname{\mathrm{AM}}$: the trained amortized explanation model, $\data{}$: the target data instance that has length $L$, $\hat{y}$: the predicted label, $\modelname{\mathrm{CLF}}$: the target model
\State $\phi \gets \modelname{\mathrm{AM}}(\data{})$
\For{ $j=1$ to $m$}
    \State sample ordering $\sigma$ from permutation $\Pi(L)$
    \State $\phi \gets \phi + \sum_{i}\left[ \modelio{\modelclf}{\data{\mathbb{S}\left([\sigma]_{i-1} \cup \left\{i\right\}\right)}}{\hat{y}} \right.$ \\
    $\left.\quad\quad\qquad\qquad\qquad - \modelio{\modelclf}{\data{\mathbb{S}\left([\sigma]_{i-1}\right)}}{\hat{y}} \right]$%
\EndFor
\State $\phi \gets \frac{\phi}{m}$
\end{algorithmic}
\end{algorithm}

\subsection{Better Fit via Local Adaption}
\label{method: local_adaption}
By amortization, our model can learn to capture the shared feature attribution patterns across data to achieve a good efficiency-stability trade-off. 
We further show that the explanations generated by our amortized model can be used to initialize the explanation scores of SVS. This way, the evaluation of SVS can be significantly sped up compared with using random initialization. On the other hand, applying SVS upon amortized method improves the latter's performance as some important tokens  might not be captured by the amortized method but can be identified by SVS through additional sampling (e.g., low-frequency tokens). 
The detailed algorithm is shown in \cref{alg:meta}. Note that here we can recover the original SVS computation~\citep{strumbelj2010efficient} by replacing $\phi \gets \modelname{\mathrm{AM}}(\data{})$ to be $\phi \gets 0$. $\modelname{\mathrm{AM}}$ is the amortized model trained using MSE as explained earlier.

%% file: past_papers/amortized_interpretability/sections/experiments.tex
\section{Experiments}
\label{sec: experiment}
In this section, we present experiments to demonstrate the properties of the proposed approach in terms of accuracy
against reference scores (\ref{sec:expapp}) and sensitivity to training-time randomness (\ref{sec: training_sensitivity}). We also show that we achieve a better fit via a local adaption method that combines our approach with SVS (\ref{exp: local_adaption}).
Then, we evaluate the quality of the explanations generated by our amortized model on two downstream applications (\ref{sec: downstream_app}). 

\shortparagraph{Setup.}
We conduct experiments on the validation set of Yelp-Polarity and MNLI datasets.
To generate reference explanation scores, we leverage the Thermostat~\citep{feldhus2021thermostat} dataset, which contains 9,815 pre-computed explanation scores of SVS-25 on MNLI. We also
compute explanation scores of SVS-25 for 25,000 instances on Yelp-Polarity. 
We use BERT-base-uncased~\citep{devlin2019bert} for $\modelname{\mathrm{LM}}$. 
For dataset preprocessing and other experiment details, we refer readers to \cref{app: training_details}.

To our best knowledge, FastSHAP~\citep{jethani2021fastshap} is the most relevant work to us that also takes an amortization approach to estimate SV on tabular or image data. We adapt it to explain the text classifier and use it as a baseline to compare with our approach. 
We find it non-trivial to adapt FastSHAP to the text domain.  As pre-trained language models occupy a large amount of GPU memory, we can only use a small batch size with limited perturbation samples (i.e., $32$ perturbation samples per instance). 
This is equivalent to approximate KS-32 and the corresponding reference explanation scores computed by FastSHAP are unstable. 
More details can be found in \cref{app: fastshap}.

\begin{table}[!htbp]
\centering
\resizebox{0.8\columnwidth}{!}{  
\begin{tabular}{@{}ccccc@{}}
\toprule
\multirow{2}{*}{Method} & \multicolumn{2}{c}{MNLI} & \multicolumn{2}{c}{Yelp-Polarity} \\
         & Spearman  & MSE  & Spearman  & MSE   \\ \midrule        
SVS-25          & 0.75                 & 1.90e-2  & 0.84                 & 6.64e-3       \\
KS-25   & 0.17                 & 9.95e-2  & 0.12                 & 4.34e-2       \\
KS-200  & 0.35                 & 7.73e-2  & 0.24                 &  5.77e-2              \\
KS-2000 & 0.60                 & 2.54e-2   & 0.51                 & 1.86e-2      \\
KS-8000 & \textbf{0.74}                 & \textbf{1.25e-2}   & \textbf{0.70}                 & \textbf{6.25e-3}     \\
\midrule
FastSHAP & 0.23 & 1.90e-1  & {0.18} & {7.91e-3} \\
Our Amortized Model & \textbf{0.42}                 & \textbf{9.59e-3} & \textbf{0.61}                 & \textbf{4.46e-6}        \\ 

\bottomrule
\end{tabular}
}
\caption{Spearman's correlation and MSE of variants of SV methods against SVS-25, a proxy of exact SV on MNLI and Yelp-Polarity. As we show in \cref{sec: stability}, MSE correlates poorly with ranking stability and Spearman's correlation should be considered as \textbf{the main metric}. We only list MSE for reference. Bold-faced numbers are the best in each column. Results are averaged over 5 runs. Our amortized model achieves better approximation compared to KS-200 and FastSHAP baseline, but not as good as much more time-consuming methods KS-2000/8000. 
SVS-25 
is listed 
as an upper bound. }
\label{tab:approximation}
\end{table}

\subsection{Shapley Values Approximation}
\label{sec:expapp}
To examine how well our model fits the pre-computed SV (SVS-25), we compute both Spearman's correlation and MSE over the test set. As it is intractable to compute exact Shapley Values for ground truth, we use SVS-25 as a proxy. We also include different settings for KS results over the same test set. KS is also an approximation to permutation-based SV computation~\citep{lundberg2017unified}. 
\cref{tab:approximation} shows 
the correlation and MSE of aforementioned methods against SVS-25.

First, we find that despite the simplicity of our amortized model, the proposed amortized models achieve a high correlation with the reference scores ($0.61 > 0.60$) on Yelp-Polarity.
The correlation between outputs from the amortized models and references is moderate ($0.42 > 0.40$) on MNLI when data size is limited. During inference time, our amortized models output explanation scores for each instance within  $50$ milliseconds, which is about $40$-$60$ times faster than KS-200 and $400$-$600$ times faster than KS-2000 on Yelp-Polarity and MNLI. Although the approximation results are not as good as KS-2000/8000 (which requires far more model evaluations), our approach achieves reasonably good results with orders of magnitude less compute. 

We also find that the amortized model achieves the best
MSE score among all approximation methods. Note that the two metrics, Spearman's correlation and MSE, do not convey the same information. MSE measures how well the reference explanation scores are fitted while Spearman's correlation reflects how well the ranking information is learned. We advocate for reporting both metrics. 

\begin{figure}[tbp!]
  \centering
  \begin{subfigure}[b]{0.48\columnwidth}
  \centering
    \includegraphics[width=\textwidth]{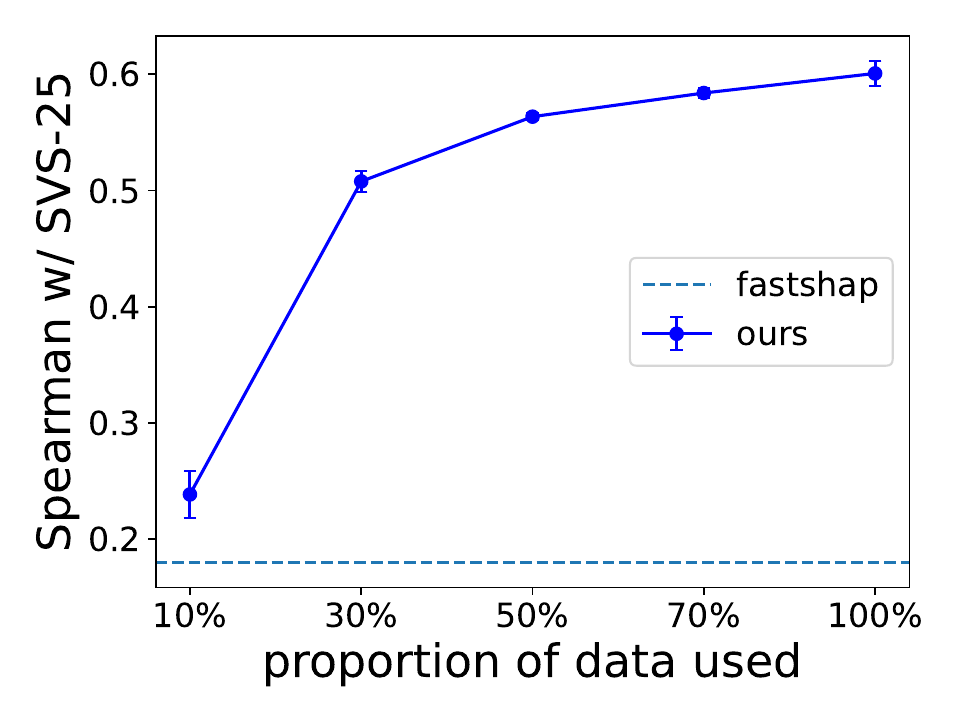}
    \vspace{-7mm}
    \caption{Yelp-Polarity}
    \label{fig:yelp_learning_curve}
  \end{subfigure}
  \begin{subfigure}[b]{0.48\columnwidth}
  \centering
    \includegraphics[width=\textwidth]{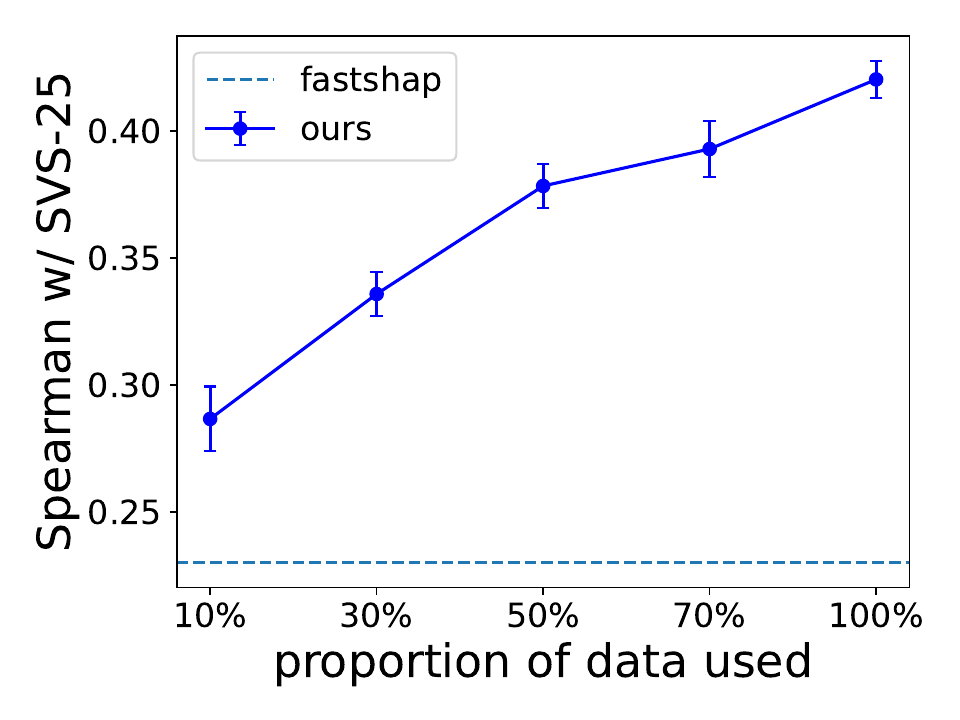}
    \vspace{-7mm}
    \caption{MNLI}
   \label{fig:mnli_learning_curve}
  \end{subfigure}
  \vspace{-3mm}
  \caption{Learning curves for the amortized model over Yelp-Polarity and MNLI datasets. The Spearman's correlations in this figure are computed against SVS-25. We can see our amortized model can learn efficiently even if there is only 10\% data used for training. }
  \label{fig: learning curve}
\end{figure}

\shortparagraph{Cost of training the amortized models}
To produce the training set, we need to pre-compute the explanation scores on a set of data. Although this is a one time cost (for each model), one might wonder how time consuming this step is as we need to run the standard sample-based estimation. As the learning curve shows in \cref{fig: learning curve}, 
we observe that the model achieves good performance with about $25\%$ ($\approx 5,000$ on Yelp-Polarity) instances. 
Additionally, in \cref{sec: domain-transfer}, we  show this one-time training will result in a model transferable to other domains, so we may not need to train a new amortized model for each new domain. 

\begin{table}[!ht]
\centering
\resizebox{0.8\columnwidth}{!}{
\centering
\begin{tabular}{@{}ccc@{}}
\toprule
Training Data Proportion          & Spearman (MNLI) & Spearman (Yelp-Polarity)   \\ \midrule
10\%          & 0.45   & 0.40                     \\
30\%   & 0.57    & 0.65                     \\
50\%  & 0.65  & 0.71                              \\
70\% & 0.65   & 0.72                       \\
100\% & 0.77  & 0.76                        \\ \bottomrule
\end{tabular}
}
\caption{Training time sensitivity study. To evaluate how much the amortized model will be influenced by randomness during training, we sample training data 5 times with different random seeds and then compute the averaged Spearman's correlation among all pairs of runs. 
The standard deviation is less than 1e-2.
Our amortized model is stable against training time randomness with only 10\% of data. 
}
\label{tab: training_time_sensitivity}
\vspace{-3mm}
\end{table}

\subsection{Sensitivity Analysis}
\label{sec: training_sensitivity}
Given a trained amortized model,
there is no randomness when generating explanation scores.
However, there is still some randomness in the training process, including the training data, the random initialization of the output layer and randomness during update such as dropout. 
Therefore, similar to 
\cref{sec: stability}, we study the sensitivity of the amortized model.
 \cref{tab: training_time_sensitivity} 
 shows the results with different training data and random seeds.
We observe that: 1) when using the same data (100\%), random initialization does not affect the outputs of amortized models -- the correlation between different runs is high (i.e., $0.77$ on MNLI and $0.76$ on Yelp-Polarity). 2) With more training samples, the model is more stable. 

\begin{table}[!ht]
\small
\centering
\resizebox{0.8\columnwidth}{!}{
\begin{tabular}{@{}ccc@{}}
\toprule
\multirow{2}{*}{Method} & \multicolumn{1}{c}{MNLI} & \multicolumn{1}{c}{Yelp-Polarity} \\
         & Spearman    & Spearman     \\ \midrule       
SVS-2          & 0.41                   & 0.52                      \\
SVS-3          & 0.47                   & 0.60                      \\
SVS-5          & 0.55                   & 0.69                      \\
SVS-25          & \textbf{0.75}                   & \textbf{0.84}                      \\
\midrule
Our Amortized Model & {0.42}                  & {0.61}                        \\ 
Our Amortized Model (Adapt-2) & {0.47}                  & {0.64}                        \\ 
Our Amortized Model (Adapt-3) & {0.53}                  & {0.69}                        \\ 
Our Amortized Model (Adapt-5) & \textbf{0.57}                  & \textbf{0.71}                        \\ 

\bottomrule
\end{tabular}}
\caption{Approximation results for the Shapley explanation methods on MNLI and Yelp-Polarity datasets. Bold-faced numbers are the best in each column. Results are averaged over 5 runs. Spearman's correlations are computed against SVS-25. Adapt-m means here how many sampled ordering $\sigma$s we used here to do local adaption ($m$ in \cref{alg:meta}). 
}
\label{tab:meta_approximation}

\end{table}

\subsection{Local Adaption}
\label{exp: local_adaption}

The experiment results for Local Adaption (\cref{method: local_adaption}) are shown in \cref{tab:meta_approximation}. Here we can see that: 1) by doing local adaption, we can further improve the approximation results using our amortized model, 2) by using our amortized model as initialization, we can improve the sample efficiency of SVS significantly (by comparing the performance of SVS-X and Adapt-X). These findings hold across datasets.

\subsection{Domain Transferability}
\label{sec: domain-transfer}
To see how well our model performs on out-of-domain data, we train a classification model and its amortized explanation model on Yelp-Polarity and then explain its performance on SST-2~\citep{socher-etal-2013-recursive} validation set. Both tasks are two-way sentiment classification and have significant domain differences. 

Our amortized model achieves a Spearman's correlation of approximately 0.50 with ground truth SV (SVS-25) while only requiring 0.017s per instance.  In comparison, KS-100 achieves a lower Spearman's correlation of 0.46 with the ground truth  and takes 1.6s per instance; KS-200 performs slightly better in Spearman's correlation but requires significantly more time. Thus, our amortized model is more than 90 times faster and more correlated with ground truth Shapley Values. This shows that, once trained, our amortized model can provide efficient and stable estimations of SV even for out-of-domain data. 

In practice, we do not recommend directly explaining model predictions on out-of-domain data without verification, because it may be misaligned with user expectations for explanations, and the out-of-domain explanations may not be reliable~\citep{hase2021out, denain2022auditing}. More exploration on this direction is required but is orthogonal to this work.

%% file: past_papers/amortized_interpretability/sections/evaluation.tex
\section{Evaluating the Quality of Explanation}
\label{sec: downstream_app}
\shortparagraph{Feature Selection.} The first case study is feature selection, which is a straightforward application of local explanation scores. The goal is to find decision-critical features via removing input features gradually according to the  rank given by the explanation methods. Following previous work~\cite{zaidan2007using, jain2019attention, deyoung2020eraser}, we measure faithfulness by changes in the model output after masking tokens identified as important by the explanation method. The more faithful the explanation method is to the target model, the more performance drop will be incurred by masking important tokens.  

\begin{figure*}[tbp!]
  \centering
  \begin{subfigure}[b]{0.48\textwidth}
    \centering
    \includegraphics[width=0.9\textwidth]{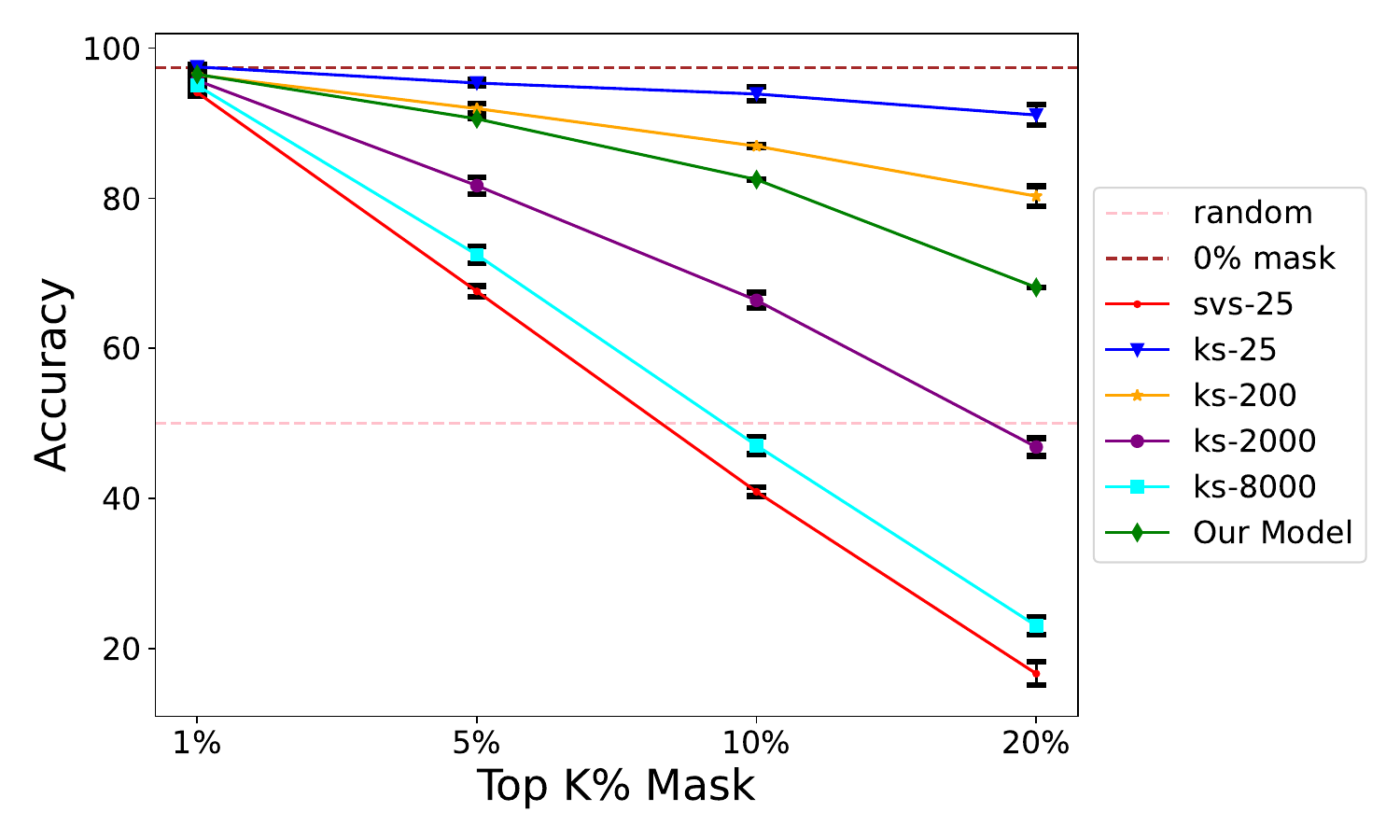}
    \vspace{-3mm}
    \caption{Yelp-Polarity}
    \label{fig: feat_select_yelp_w_amortized}
  \end{subfigure}
  \begin{subfigure}[b]{0.48\textwidth}
  \centering
    \includegraphics[width=0.9\textwidth]{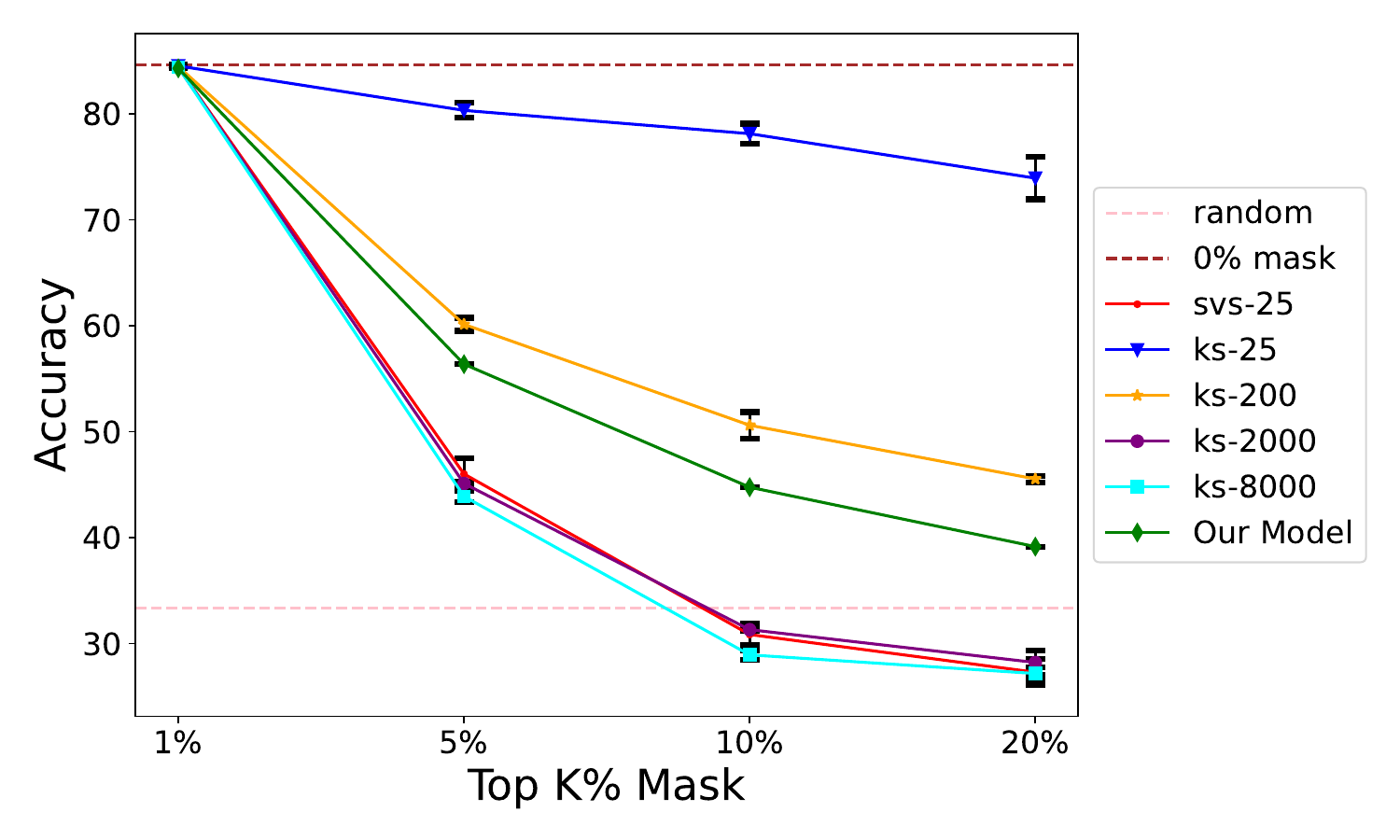}
    \vspace{-3mm}
    \caption{MNLI}
    \label{fig: feat_select_mnli_w_amortized}
  \end{subfigure}
  \caption{Feature selection based on interpretations on Yelp-Polarity and MNLI datasets. The faster the curve drops, the more faithful the explanation scores are. We can see our amortized model is more faithful to the target model compared to KS-200, but are not as faithful as other more costly methods.}
  \label{fig: feat_select_am}
\end{figure*}

We gradually mask Top-$\alpha$ tokens ($\alpha = 1\%, 5\%, 10\% ,20\%$) and compute the accuracy over corrupted results using the stability evaluation sets for
MNLI and Yelp-Polarity datasets as mentioned in \cref{sec: stability}. 
As the results
show in \cref{fig: feat_select_am},
the amortized model is more faithful
than KS-200
but underperforms KS-2000/8000 and SVS-25. However, the amortized model is more efficient than these methods. So amortized model achieves a better efficiency-faithfulness trade-off.  

\shortparagraph{Explanation for Model Calibration.}  Recent work suggests that good explanations should be informative enough to help users to predict model behavior \citep{doshi2017towards, chandrasekaran2018explanations, hase2020evaluating, ye2021connecting}. \citet{ye2022can} propose to combine the local explanation with pre-defined feature templates (e.g., aggregating explanation scores for overlapping words / POS Tags in NLI as features) 
to calibrate an existing model to new domains. The rationale behind this is that, 
{if the local explanation truly connects to human-understandable model behavior, then following the same way how humans transfer knowledge to new domains, the explanations guided by human heuristics (in the form of feature templates) should help calibrate the model to new domains.}
Inspired by this, we conduct a study using the same calibrator architecture but plugging in different local explanation scores. 

We follow \citet{ye2022can} to calibrate a fine-tuned MNLI model\footnote{\url{https://huggingface.co/textattack/bert-base-uncased-MNLI}} to MRPC. The experiment results are shown in \cref{tab:calib-amortized}.
In the table, ``BOW'' means the baseline that uses constant explanation scores when building the features for the calibration model. Compared with the explanation provided by  KS-2000, the explanation given by the amortized model achieves better accuracy, suggesting that the amortized model learns robust explanation scores that can be generalized to out-of-domain data in downstream applications.\footnote{See \cref{sec: domain-transfer} for a domain transfer experiment that directly compares to SVS-25 and w/o calibration.} 

\begin{table}[htbp!]
\small
\centering
\begin{tabular}{@{}ccH@{}}
\toprule
Model                     & Acc  & AUC  \\ \midrule
BOW                       & 67.3 & 74.6 \\
ShapCal (KS-2000) & 67.4 & 74.3 \\
ShapCal (Amortized)       & 68.0 & 74.5 \\ \bottomrule
\end{tabular}%
\caption{Calibration Experiments for Amortized Models. 
The explanation scores can help the calibrator achieves better accuracy on out-of-domain data
than KS-2000. }
\label{tab:calib-amortized}
\end{table}

%% file: past_papers/amortized_interpretability/sections/conclusion.tex
\section{Discussion}
In this paper, we empirically demonstrated that it is challenging to obtain stable explanation scores on long text inputs.
Inspired by the fact that different instances can share similarly important features, we proposed to efficiently estimate the explanation scores through an amortized model trained to fit pre-computed reference explanation scores.

In the future, we plan to explore model architecture and training loss for developing  effective amortized models. In particular, we may incorporate sorting-based loss to learn the ranking order of features. Additionally, we could investigate the transferability of the amortized model across different domains, as well as exploring other SHAP-based methods instead of the time-consuming SVS-25 in the data collection process to improve efficiency further.

%% file: past_papers/amortized_interpretability/sections/limitations.tex
\section*{Limitations}
In this paper, we mainly focus on developing an amortized model to efficiently achieve a reliable  estimation of SV. Though not experimented with in the paper, our method can be widely applied to other black-box post-hoc explanation methods including LIME~\citep{ribeiro2016lime}. Also, due to the limited budget, we only run experiments on BERT-based models. However, as we do not make any assumption for the model as other black-box explanation methods, our amortized model can be easily applied to other large language models. We only need to collect the model output and our model can be trained offline with just thousands of examples as we show in our method and experiments. 

\shortparagraph{Comparison and Training with Exact Shapley Values}
Computing exact SV is computationally prohibitive for large language models (LLMs) on lengthy text inputs, as it necessitates the evaluation of LLMs on an exponential (in sequence length) number of perturbation samples per instance. As a result, we resort to using SVS-25, which serves as a reliable approximation, for training our amortized models.

%% file: chapters/ai_realtor.tex
\section*{Chapter Overview}
This chapter presents AI Realtor, an agentic framework that represents the synthesis of our approach to grounded generation. It explicitly separates the grounding module (factual accuracy) from the personalization module (alignment). By architecturally enforcing grounding before style, we achieve automated copywriting that is persuasive yet factually rigorous, effectively solving the alignment-hallucination trade-off in a high-stakes domain. This chapter is based on work accepted to CAIS 2026~\citep{wu2025grounded}.

\graphicspath{{./}}

\input{past_papers/ai_realtor/intro}

\input{past_papers/ai_realtor/environment}

\input{past_papers/ai_realtor/method}

\input{past_papers/ai_realtor/evaluate}

\input{past_papers/ai_realtor/related-work}

\input{past_papers/ai_realtor/discussion}

%% file: past_papers/ai_realtor/intro.tex
\section{Introduction}
While large language models (LLMs) have made significant strides across various tasks, their ability to persuade remains an underexplored frontier (see a discussion of related work in Section~\ref{sec:related}). 
This however is a particularly important capability since  persuasion-related economic activities --- a common thread in almost all voluntary transactions from advertising and lobbying to litigation and negotiation --- underpin roughly 30\% of the US GDP~\citep{antioch2013persuasion}, hence gives rise to tremendous opportunity for applying LLMs   
 across a wide range of sectors. Meanwhile, this same potential introduces serious trustworthiness concerns. If LLMs can generate persuasive content at scale, their influence on human opinions raises risks of misinformation, manipulation and misuse, especially in sensitive domains such as political campaigns~\citep{voelkel2023artificial, goldstein2024persuasive}.

Therefore, we focus our study on the task of language generation for grounded persuasion, that is, the production of persuasive content that is faithful in factual details. This task is especially critical in copywriting, the practice of creating marketing text that seeks to influence consumer decisions, where its effectiveness can be directly assessed through measurable behavioral outcomes (e.g., ratings, engagement, and conversions), yet must remain strictly constrained by factual accuracy. 
In particular, we choose the domain of real estate marketing (see our rationale in~\cref{sec: environment}) and develop an agentic framework, \agentname, whose modular design is motivated by economic signaling and information asymmetry.
Below, we outline core contributions and the structure of this paper:
\vspace{-2mm}
\begin{enumerate}[wide, labelwidth=!, labelindent=0pt]
\item[\circone] \textbf{Real-World Evaluation}: Using real estate marketing as our testbed, we construct a large dataset from Zillow and design an experimental website that simulates the house search process, including buyer preference elicitation. We recruit a targeted group of potential home buyers to evaluate the persuasiveness of the generated marketing content (\cref{sec: environment}).

\item[\circtwo] \textbf{Theoretical Motivation}: We draw on the economic theory of information design in strategic communication games~\citep{bergemann2019information} to motivate the agentic workflow. This perspective helps decompose the task into processing raw factual attributes, selecting key features to highlight, and generating persuasive marketing content, but it is not intended as a formal optimality guarantee (\cref{sec: model}).

\item[\circthree] \textbf{Agentic Pipeline}: We develop an LLM-based agent (\cref{sec: agent_design}) with three key modules: a \textit{Grounding Module}, which mimics human expertise in identifying and signaling critical, credible selling points; a \textit{Personalization Module}, which tailors content to user preferences; and a \textit{Marketing Module}, which ensures factual consistency and incorporates localized features.

\item[\circfour] \textbf{Empirical Effectiveness}: Our system achieves a 70\% win rate over human experts while maintaining a comparable level of factual accuracy, providing a human-subject benchmark for grounded persuasion with measurable behavioral impact (\cref{sec: evaluation}).

\end{enumerate}

%% file: past_papers/ai_realtor/environment.tex
\section{A Benchmark for Grounded Persuasion}
\label{sec: environment}
\paragraph{Motivations and Challenges}

Establishing a robust evaluation benchmark for persuasion faces two core challenges. First, persuasiveness is inherently subjective: unlike reasoning or planning (which have objective metrics), its effectiveness depends on human feedback and varies with individual preferences and contexts.
Second, persuasion is multifaceted, with domain-specific techniques shaped by psychology, economics, and communication.
Existing LLM research mostly focus on political or opinion-based persuasion, where evaluations are complicated by cognitive biases and adversarial framing. For example, \citet{hackenburg2024evaluating} and \citet{matz2024potential} reached conflicting conclusions using similar experimental designs. \citet{durmus2024measuring} highlight the anchoring effect -- the tendency to cling to initial beliefs -- making opinion shifts hard to measure. They also find fabricated content is often more persuasive, raising ethical and methodological concerns.
These limitations underscore the need for new benchmarks in controlled, fact-grounded settings. 

\textbf{Real Estate Marketing (REM) as Testbed}\quad 
Identifying well-scoped testbeds is key to launch systematic investigations of general AI capabilities, as demonstrated by recent benchmarks~\citep{yao2022webshop, xie2024travelplanner}.
The real estate marketing domain is ideal for our study because:
\begin{enumerate}[wide, labelwidth=!, labelindent=0pt, itemsep=1pt, topsep=-0.5\parskip]
\item[\circone] \textit{High-stakes, rational decisions}: Real estate involves high-stakes economic decisions, where buyers typically hold rational, fact-based beliefs --- unlike more emotionally charged 
or polarized 
domains. Persuasive language in this setting must be both compelling and truthful. 
\item[\circtwo] \textit{Measurable economic impact}: 
\revise{Effective persuasion has tangible economic value in real estate. While structured attributes and images capture initial attention, industry guidance emphasizes that descriptive text is critical for conveying the unique experience of living in a home~\citep{zillow_guidance}. The potential for LLMs to assist in this high-value task is further illustrated by recent anecdotal accounts~\citep{reddit_2023}.}
\item[\circthree] \textit{Rich, structured datasets}: The availability of extensive property listings with carefully labeled attributes (e.g., from Zillow) enables domain-specific training and thorough empirical evaluations.
\end{enumerate}

\textbf{Realistic Evaluation Interface and Persuasiveness Measurement}\quad
Our framework prioritizes two criteria: (1) immersive user interaction to capture authentic feedback and (2) dynamic preference elicitation for personalized generation. We replicate real-world homebuyer behavior by integrating 50k+ real-world listings into a web platform. See Appendix~\ref{app: interface} and \ref{app: dataset} for a full description of the web interface and dataset.
We evaluate persuasion via pairwise comparisons: buyers view a property with two model-generated descriptions and select the more compelling one. Persuasiveness is quantified via Elo scores~\citep{elo1967proposed}; factual accuracy is verified against listing metadata (see \cref{sec: evaluation}).

%% file: past_papers/ai_realtor/method.tex
\section{An Economic Scaffolding of Copywriting}
\label{sec: model}
Copywriting fundamentally is about communicating product information, often selectively, to shape potential buyers' perceptions and influence their purchasing decisions. This process of information signaling, also known as persuasion, has been extensively studied in decision theory and information economics~\citep{spence1978job,arrow1996economics,kamenica2011bayesian,connelly2011signaling}, typically within stylized mathematical models. We use these models as conceptual motivation for an agentic framework for natural-language copywriting, rather than as a direct formal guarantee about LLM behavior.

\textbf{Attributes}\quad Formally, we represent a generic \emph{product}  $X$ (e.g., a house or an Amazon item) as an $n$-dimensional vector $X = (X_1, X_2, \dots, X_n)$. Each $X_i$ is called a raw attribute (or simply \emph{attribute}). Attributes capture the factual and measurable characteristics of the product (e.g., square footage, distance to transit). 
A specific product instance is denoted by vector $\mathbf{x} = (x_1, \cdots, x_n)$ where $x_i \in \mathcal{X}_i$ is the \emph{realized} value of attribute $X_i$. Let $\mathcal{X} = \Pi_{i} \mathcal{X}_i$ be the domain of $\mathbf{x}$.  

\textbf{Features}\quad Marketers often emphasize certain attractive properties of a product (e.g., ``spacious layout'' and ``prime location'' in REM), derived from its underlying raw attributes. We refer to these as signaling features (or simply \emph{features}).  
Importantly, features differ from attributes: while some attributes may directly serve as features, features generally capture the more abstract (and sometimes ambiguous) properties.
We denote the feature set as $S = (S_1, \cdots, S_m)$, with a feature vector $\mathbf{s} = (s_1, \cdots, s_m)$, where each  $s_i\in [0,1]$ quantifies the \emph{intensity} or likelihood of feature $S_i$ being. For example, $S_i$ could be ``bright room'' and correspondingly $s_i$ denotes the extent to which rooms of the house are bright. In practice, both $x_i$ and $s_j$ can be assessed by domain experts. 

 \textbf{Signaling via the Attribute-Feature Mapping}\quad In our model, signaling features convey partial information to influence potential buyers' beliefs, leveraging the inherent cognitive mapping in natural language. For instance, a feature ``bright room'' may probabilistically imply 
 high floor, southern exposure, and modern lighting -- all affecting buyers’ perceptions and decisions. (e.g., deciding to schedule a visit).
We formalize this with a mapping $\pi: \mathcal{X} \to [0,1]^m$ that transform raw attributes $\mathbf{x}\in \mathcal{X}$ into  feature intensities $\mathbf{s} \in [0,1]^m$. That is, $\mathbf{s} = \pi(\mathbf{x})$. Sometime, we use $\mathbf{s}(\mathbf{x})$ to emphasize the dependence of $\mathbf{s}$ on the underlying attributes $\mathbf{x}$, and $s_j(\mathbf{x})$ is its $j$-th entry. This mapping reflects the commonsense inference: given $\mathbf{x}$, how strongly we can claim the presence of feature $S_j$. 

This attribute-feature mapping $\pi$ is widely studied in both machine learning and economics. In Bayesian statistics, $X_i$ is an observable variable, $S_j$ a latent variable, and $\pi$ captures their probabilistic dependence. In information economics, $X_i$ represents a state, $S_j$ a \emph{signal}, and $\pi$ is known as a \emph{signaling scheme}. Signals can be strategically designed to reveal partial information about the state, and prior work has made significant progress in their optimal design to influence equilibrium outcomes~\citep{kamenica2011bayesian, bergemann2015limits, bergemann2019information}. In this paper, this connection motivates our decomposition of the generation problem; we do not claim to solve the formal optimal signaling problem.
Our work moves beyond this traditional Bayesian framing to incorporate the nuanced role of natural language--often abstracted away in prior models--and to uncover the implicit, \emph{commonsense} mappings behind linguistic signals, rather than design new schemes.

\textbf{Marketing Design under Information Asymmetry}\quad 
Marketing fundamentally exploits information asymmetry 
between sellers and buyers~\citep{grossman1981informational, lewis2011asymmetric, dimoka2012product, kurlat2021signalling}. This important insight, along with its broader implications in general economic markets, was notably recognized by the 2002 Nobel Economics Prize~\citep{akerlof1978market, spence1978job, stiglitz1975theory, lofgren2002markets}. 
 In our setting, the seller or seller's agent knows the exact product attributes $\mathbf{x}$ and the corresponding feature values $\mathbf{s}(\mathbf{x})$, while the buyer enters the market with only a prior belief $\mu$ over the distribution of attributes in $\mathcal{X}$. Without specific knowledge of the product $\mathbf{x}$, the buyer holds an expected belief over features:
 \begin{equation}
     \text{Initial belief of features: }
     \,\,  \bar{\mathbf{s}}(\mu) = \int_{\mathbf{x} \in \mathcal{X}} \mathbf{s}(\mathbf{x}) d \mu( \mathbf{x}). 
 \end{equation}
Given the asymmetric feature beliefs between the buyer and seller, the purpose of marketing can be described as revealing features, subject to communication constraints, to shift the buyer's belief from $\bar{\mathbf{s}}(\mu)$ towards $\mathbf{s}(\mathbf{x})$ with the goal of increasing the product's attractiveness to the buyer.   

\begin{figure}[htbp]
    \centering
    \includegraphics[width=0.7\linewidth]{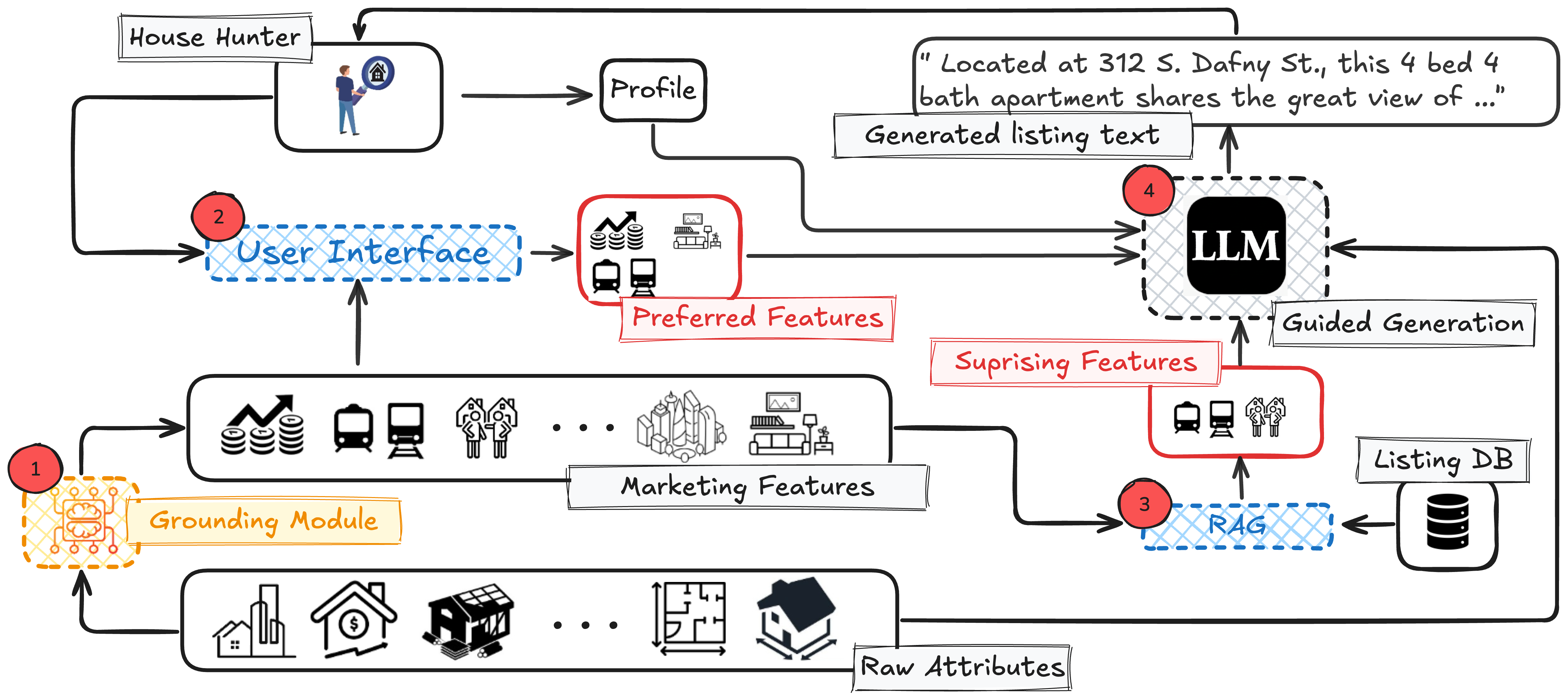}
    \caption{Illustration of the Design Pipeline of \agentname.  }
    \label{fig:ai-realtor_pipeline}
    \vspace{6pt}
\end{figure} 

 \textbf{Grounded Persuasion in Natural Language}\quad
 The remaining part of our model is to guide the generation of persuasive marketing content. 
 The typical approach in economic theory is to develop models capturing  buyers' belief updates and decision-making processes. 
 However, these are difficult to operationalize due to the absence of concrete buyer utility functions and behavioral models. Instead, we leverage the generative capabilities of LLMs, guided by heuristics and instructions tailored for grounded persuasion.
At a high level, we use the attribute-feature mapping $\pi$ to guide the selection of a feature subset $\mathcal{S}^*$ to emphasize in generation.
User preferences $\mathbf{r}$ are elicited and incorporated into a prompt $\mathcal{I}^*$ for personalization. Conceptually, this can be viewed as encouraging the LLM to approximate an implicit preference-aware generation objective:
$\text{}  L^* =  \argmax_{L\in \mathcal{L}} \Pr(L| \mathcal{I}^*, \mathcal{S}^*, \mathbf{r}) \approx \argmax_{L\in \mathcal{L}(\mathbf{x})} U^{\mathbf{r}}( L ).$
This expression is a design abstraction rather than a claim that the deployed model exactly solves a specified optimization problem. In practice, carefully designed prompts $\mathcal{I}^*$, selected features $\mathcal{S}^*$, and user preferences $\mathbf{r}$ provide structured inputs that steer the LLM toward persuasive descriptions while keeping it grounded in product attributes $\mathbf{x}$. Given this formulation, our design objective is to support grounded generation by constructing effective prompts $\mathcal{I}^*$, selecting appropriate features $\mathcal{S}^*$, and representing user preferences $\mathbf{r}$. The following section describes our implementation. 

\section{The Agentic Implementation of \agentname}
\label{sec: agent_design}
This section outlines the core design of \agentname, an AI agent that processes multiple levels of marketing information to compose persuasive descriptions for real estate listings and adapts its language using elicited buyer preferences. 
At a high level, our approach uses microeconomic models as guidance for implementing the following three key ingredients:
\begin{itemize}[  wide, nosep]  
    \item Grounding Module: identify the attribute-feature mapping $\pi$; 
    \item Personalization Module: elicit and represent   buyer preferences $\mathbf{r}$;  
    \item Marketing Module: select useful yet factual marketing features $\mathcal{S}^*$ based on $\pi, \mathbf{r}$. 
\end{itemize}
The overall system pipeline is illustrated in \cref{fig:ai-realtor_pipeline}. Below, we highlight the novel contributions within each of the three modules. Full implementation details are provided in \cref{app: implementation-details}.

\begin{figure}[htbp]
    \centering
    \includegraphics[width=0.7\linewidth]{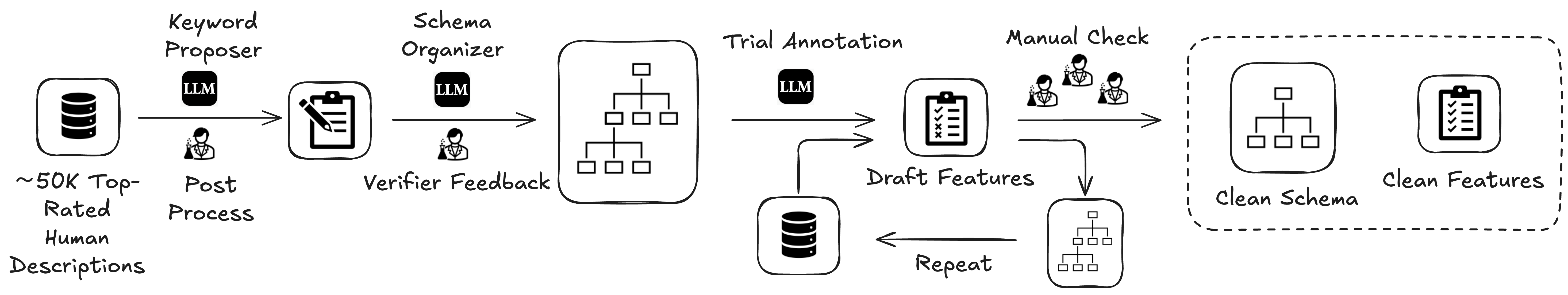}
    \vspace{-2pt}
    \caption{Illustration of the inductive feature schema construction pipeline.  }
    \label{fig:highlight_model_pipeline}
\end{figure}

\subsection{Grounding Module: Predicting Credible Features for Marketing}
\label{sec: highlight_model}

Our model assumes the existence of attribute-feature mappings that marketers can use to influence buyer beliefs and behaviors. 
However, a key challenge is that while raw attributes (e.g., square footage) are available, high-level signaling features (e.g., ``convenient transportation'') lack explicit annotations in our dataset. 
This absence of supervision, combined with the open-ended nature of natural language, where many tokens may serve as features with overlapping or ambiguous meanings, makes the learning problem inherently difficult. Without a structured representation, the label space becomes too sparse for effective training. Indeed, we find that directly prompting LLMs to generate features produces redundant or incomplete feature sets, which undermines the quality of the learned mapping. 

Manual annotation by human experts could address this issue but is labor-intensive, costly to scale, and difficult to personalize. We therefore adopt a machine learning approach to infer the attribute-feature mapping automatically from unlabeled data, guided by LLM-assisted schema construction and weak supervision.
Specifically, we provide LLMs with a large pool of candidate features extracted from the dataset and prompt them to organize these into a hierarchical schema. A small number of human annotators validate the output to monitor hallucinations and refine definitions. In a 636-sample validation, model-human agreement was around 60\%, comparable to human-human agreement, with most disagreements concentrated in subjective features such as aesthetics and style.
This process, illustrated in \cref{fig:highlight_model_pipeline}, yields a compact and expressive feature representation. Once created, this feature set and mapping can be reused across models within the same marketing domain and is thus a \emph{one-time} cost. 

Using the finalized feature schema, we guide an LLM to annotate whether each feature $s_i$ is present in a given listing, based on its attributes $\mathbf{x}$ and corresponding human-written description. After standard preprocessing (e.g., removing low-quality texts, normalizing attributes), we curate a labeled dataset and train a neural network to learn the attribute-feature mapping.\footnote{We also experiment with several other baselines for feature extraction, including prompting LLMs directly and applying simple pooling over embedding vectors. The strongest baseline achieves approximately 59\% F1 score, which is substantially lower than the final model used in our grounding module.} On a random 4:1 train-test split, our model achieves 69.39\% accuracy and 67.43\% F1 score. This result should be interpreted in light of the noisy multi-label setting: each feature is predicted as a separate binary label, and subjective features such as style and aesthetics are frequently ambiguous even for human annotators.

To ensure grounded use of signaling features, we implement a deterministic feature selection strategy: only features with intensity $s_j \geq \alpha$ are retained. In our implementation, we use the threshold $\alpha = 1/2$\footnote{The feature existence threshold $\alpha$ was determined through a grid search over the range $[0.1, \dots, 0.9]$, with performance evaluated using the F1 score on a held-out, human-annotated validation set. $\alpha = 0.5$ yielded the best trade-off between precision and recall. } and define the resulting set of \emph{marketable features} as: 
\begin{equation}\label{eq:marketable-feature}
    \text{Marketable Features: } \quad \mathcal{S}_1(\mathbf{x}) = \{S_j : s_j(\mathbf{x}) \geq \alpha \}.
\end{equation}

\subsection{Personalization Module: Aligning with Preferences}
\label{sec: user_preference}

This stage aims to steer persuasive language generation toward buyer preferences—another core objective of grounded persuasion. Our solution involves two steps.

First, we elicit user preferences and structure them in a usable form. On platforms like Zillow or Redfin, this could be done using mature machine learning methods based on user browsing behavior. Without access to such data, we instead design a preference elicitation process within our human-subject evaluation framework. Specifically, our web interface prompts an LLM to simulate a realtor, guiding participants through questions to identify their most valued features. Each user then rates the importance of each feature $S_j$ with a score $r_j$ prior to the evaluation tasks. While simple, this approach provides an interpretable preference signal that can shift which grounded features the agent emphasizes.

Second, we select a personalized subset of features to align generation with stated user preferences. Since real-world marketing texts are not tailored to individual users, we cannot rely on them to provide supervision for personalization.  Instead, we use a scoring function that combines population-level feature intensity $\mathbf{s}(\mathbf{x})$ with individual preference ratings $\mathbf{r}$, selecting features above a threshold $\alpha$:
\begin{equation*}
    \text{Personalized Features: } \quad 
    \mathcal{S}_2(\mathbf{x}) = \{ s_j \mid s_j(\mathbf{x}) + c (r_j - r_0) \geq \alpha \},
\end{equation*}
where $c$ reflects the strength of personalization and $r_0$ is a baseline rating. These features are then passed to the LLM, which determines how best to incorporate them into the generated text.

\subsection{Marketing Module: Capturing Surprisal via RAG}
\label{sec: surprisal}
The last stage is designed to better ground persuasive language generation in factual evidence, problem contexts and localized information in automated marketing. Our design here is inspired by rich marketing strategy research ~\citep{lindgreen2005viral, ludden2008surprise, ely2015suspense}, which have shown that  buyers would derive entertainment utility from \emph{surprising} effects/features and have a deeper impression. 
In our setting of real estate marketing, such surprising features are   those that are relatively rare compared to their surrounding area.
Formally, we determine a set of surprising features based on their percentile in the feature distribution as follows,
 \begin{align*}
 \mathcal{S}_3(\mathbf{x})
 &=
 \{S_j \subset \mathcal{S}_1 \mid s_j(\mathbf{x}) \in Q_{\beta}(s_j(\mu))\}.
 \end{align*}
where $Q_{\beta}(s_j(\mu))$ denotes the top $\beta$-quantile region of distribution $s_j(\mu)$.
This gives the LLMs localized feature information at different levels of granularity obtained through Retrieval Augmented Generation (RAG) \citep{lewis2020retrieval}.\footnote{\revise{In our implementation, we implement the sparse retrieval part via ElasticSearch (\url{https://www.elastic.co/elasticsearch}) and retrieve Top 10 listings with the most similar features. }} This design appears useful in our evaluation; citing one of the human subjects in our experiment (see the full description in~\cref{sec: case-for-surprisal}), who was asked about why they liked a listing description (without knowing it was AI-generated): 
\begin{myquote}{0.1in}
 \it
 ...Description B specifically points out the rarity of the ample storage and built-in cabinetry in similarly priced listings, making the property stand out.
\end{myquote}

%% file: past_papers/ai_realtor/evaluate.tex
\section{Evaluations}
\label{sec: evaluation}

\begin{figure}[htbp]
    \centering
    \begin{subfigure}[t]{0.62\textwidth}
        \centering
        \includegraphics[width=\linewidth]{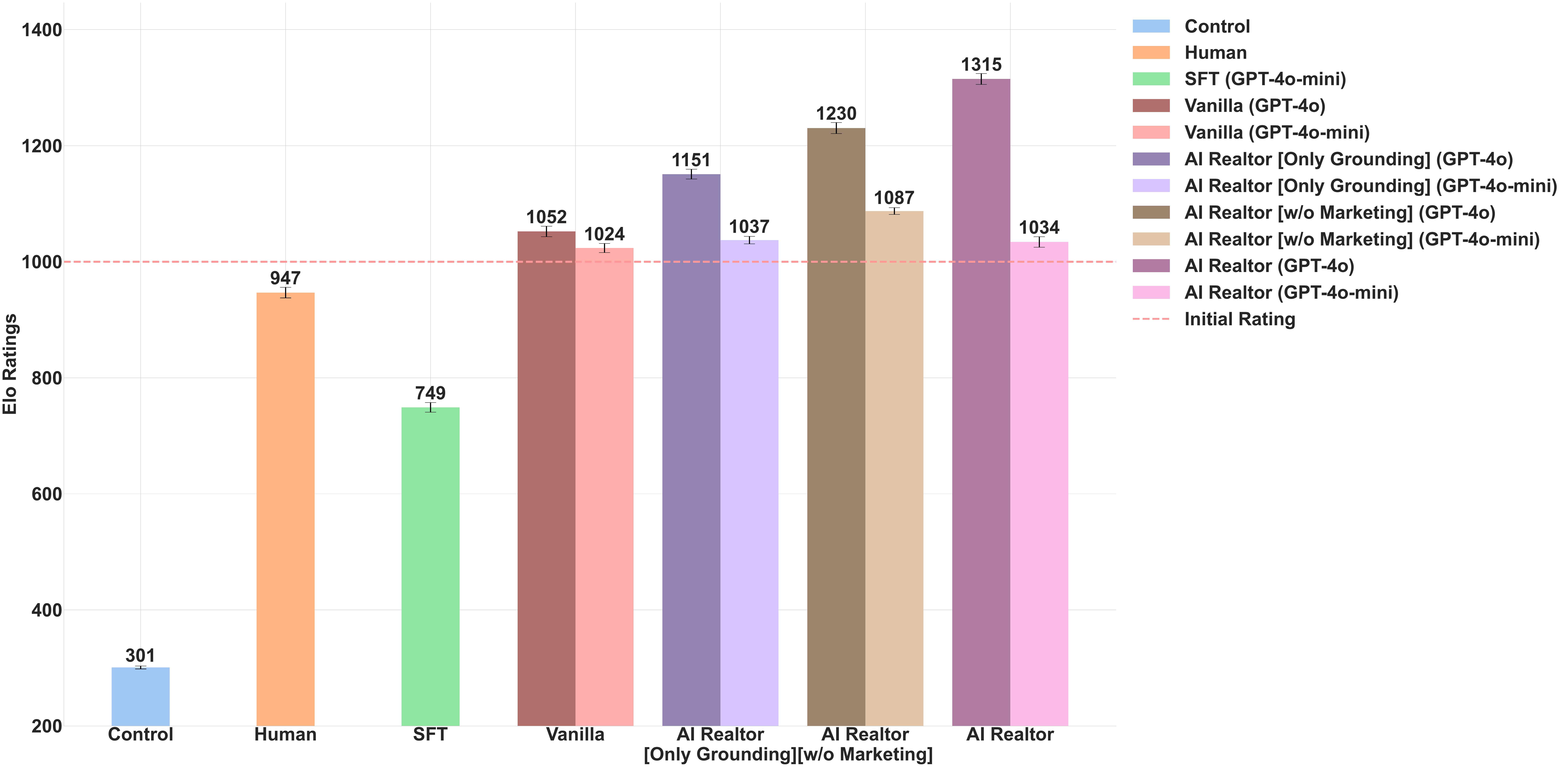}
        \caption{Elo Ratings}
        \label{fig:elo}
    \end{subfigure}
    \hfill
    \begin{subfigure}[t]{0.32\textwidth}
        \centering
        \includegraphics[width=\linewidth]{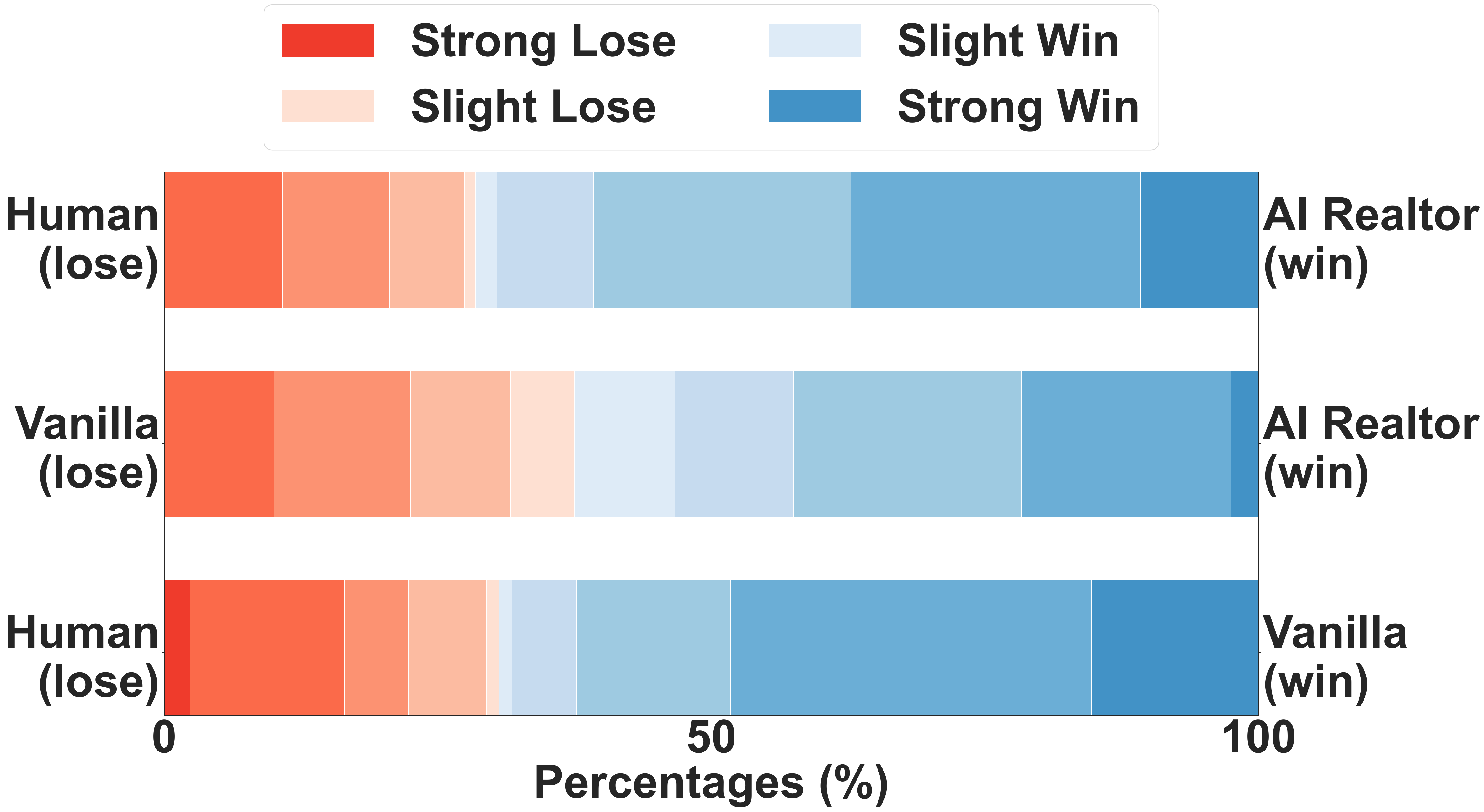}
        \caption{Win Rates}
        \label{fig:win_rates}
    \end{subfigure}
    \caption{Comparison of model performance using Elo ratings and win rates. Elo ratings represent overall persuasiveness, and win rates reflect relative persuasiveness. Both metrics are based on evaluations by human subjects. \revise{Confidence intervals are computed using 500 bootstrap runs by adapting the Elo implementation from Chatbot Arena~\citep{chiang2024chatbot}.}}
    \label{fig:main_exp}
\end{figure}

\subsection{Evaluation by Human Feedback}
\label{sec:survey}
To evaluate the effectiveness of listing descriptions generated by different models, we draw inspiration from ChatArena~\citep{zheng2023judging} and conduct an online survey to collect pairwise human feedback comparing different models' outputs.  In summary, systematic evaluation by human feedback shows that our \agentname receives higher preference ratings than human experts and other model variants in our benchmark, measured by standard Elo ratings \citep{elo1967proposed}. Below, we detail the design of our user survey platform, baseline setup, and evaluation metrics, followed by a report on the human evaluation results.   

\textbf{Quality Assurance}\quad
We focus on the major US city \emph{Chicago}\revise{\footnote{\revise{Chicago has been established by various economic and sociological literature~\citep{levitt2008market, sampson2012great, grabinsky2015most} as a rigorous proxy for broader American urban mechanics. Also, Chicago has a diverse set of listings, compared to major cities in the US, that can reliably test our models’ performance across various scenarios. See \cref{app: diversity_of_chicago} for more analysis.}}} with a highly active housing market. We recruit about 100 participants from the popular \emph{Prolific} platform for human-subject experiments, selecting in-state residents familiar with Chicago's housing market and curating approximately 1,000 listings of varied sizes and price ranges.  Each human subject is tasked with comparing 10 pairs of house descriptions. During each comparison, the human subject sees pictures and all basic information about a house, and then faces two listing descriptions without knowing what methods (human realtor or AI agents) generate them, and is asked to choose which description is preferred, and by how much (see Appendix \ref{app: comparison-interface} for details). Notably, \agentname generates personalized descriptions on the fly for each human subject, based on their preferences elicited while they join the survey (see Appendix \ref{app: preference-interface} for details).  

To ensure feedback quality, we implement several measures: (1) \textit{Screening tests} to confirm participants can extract information from listings and follow specific home search motives (See \cref{app: screening-interface} for details); (2) \textit{Attention checks} using pairs of nearly identical descriptions to ensure participants carefully compare and identify differences; (3) \textit{Control experiments} where participants compare human-written, engaging descriptions against LLM-generated descriptions intentionally prompted to be plain and unappealing, verifying their ability to favor high-quality descriptions; and (4) \textit{Incentives} on the platform, including bonus payments and requests for written reasoning behind choices, to encourage consistent, well-justified feedback.

\textbf{Metrics}\quad  
We adopt the Elo rating score
as our main metric. We use a typical choice of the initial Elo rating as $1000$, scaling parameter $c = 400$, and learning rate $K = 32$. 
The win rate for a model with Elo rating $e_1$ against a model with rating $e_0$ is calculated as $\left[1 + 10^{(e_0 - e_1)/c}\right]^{-1}$.

\textbf{Baseline Models}\quad  
In addition to our primary persuasion model \agentname, we evaluate several baseline models, including:  
\textit{Vanilla}, an LLM prompted with all attributes of the listing;  
\textit{SFT}, an LLM fine-tuned with supervised training and prompted with all features of the listing;  
\textit{Human}, listing descriptions sourced from Zillow, written by professional realtors;  
\textit{Control}, the model used in the control experiment described earlier.  We also include two ablation models based on \agentname: one that only uses the marketable feature from the Grounding module, the other excludes surprisal features from the Marketing module.
Additionally, we experiment with two LLM variants, GPT-4o and GPT-4o-mini, while keeping the prompt instructions consistent across models.  

\begin{figure}[htbp]
    \centering
        \begin{subfigure}[t]{0.32\textwidth}
            \centering
            \includegraphics[width=\linewidth]{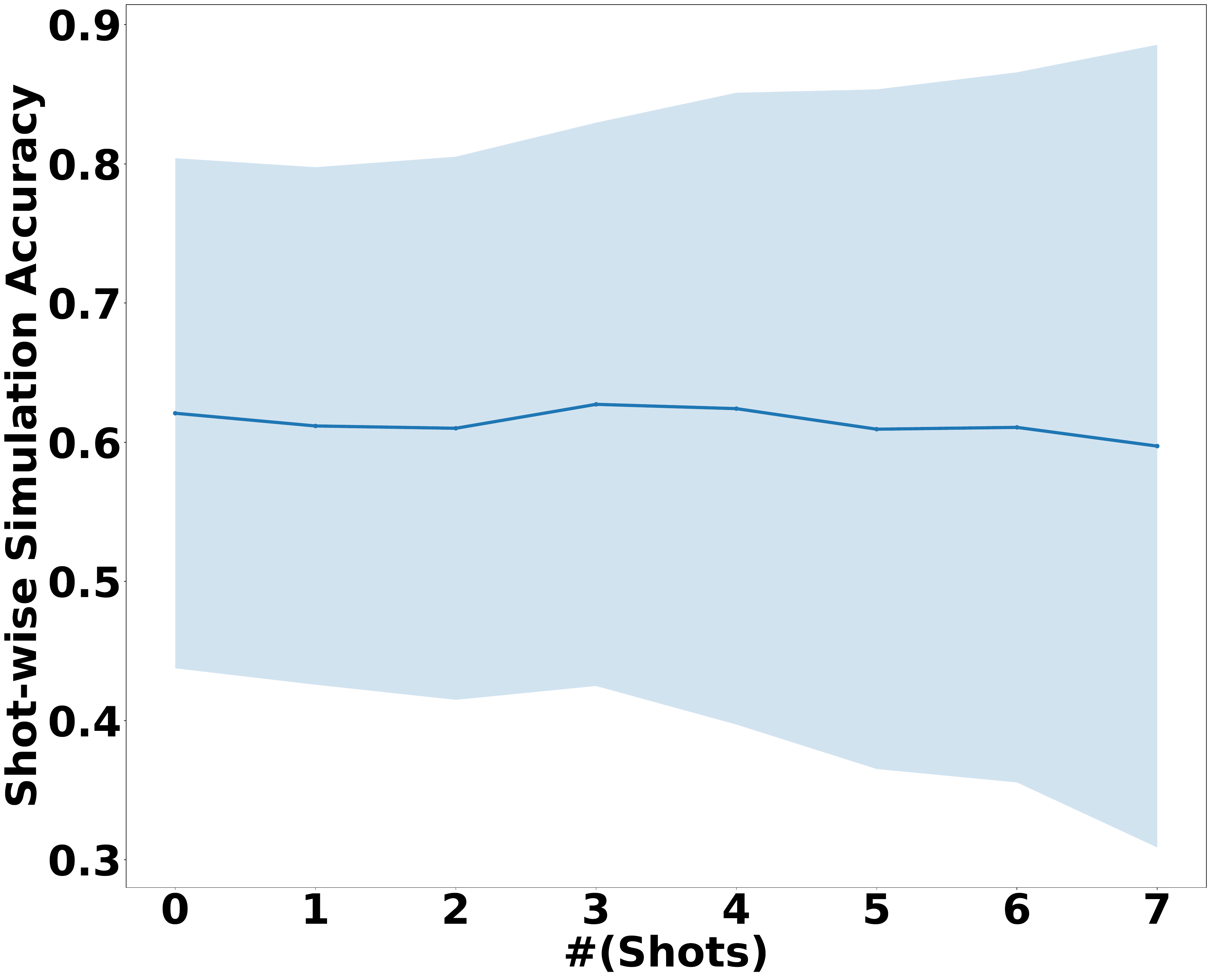}
            \caption{Shot-wise Accuracy}
            \label{fig: simulation_ssa}
        \end{subfigure}
        \begin{subfigure}[t]{0.32\textwidth}
            \centering
            \includegraphics[width=\linewidth]{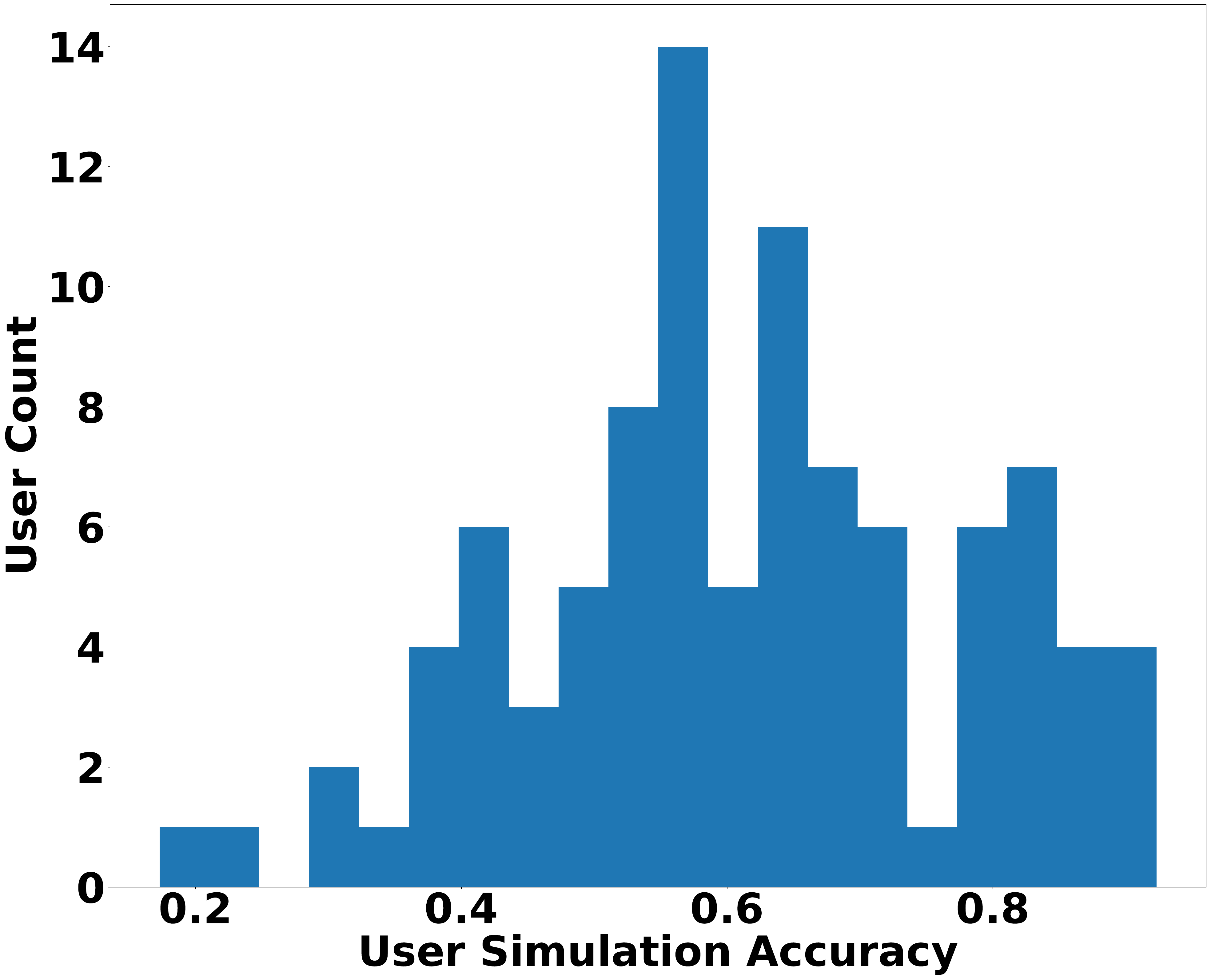}
            \caption{User-wise Accuracy Histogram}
            \label{fig: simulation_usa}
        \end{subfigure}
    \hfill
    \begin{subfigure}[t]{0.32\textwidth}
        \centering
        \includegraphics[width=\linewidth]{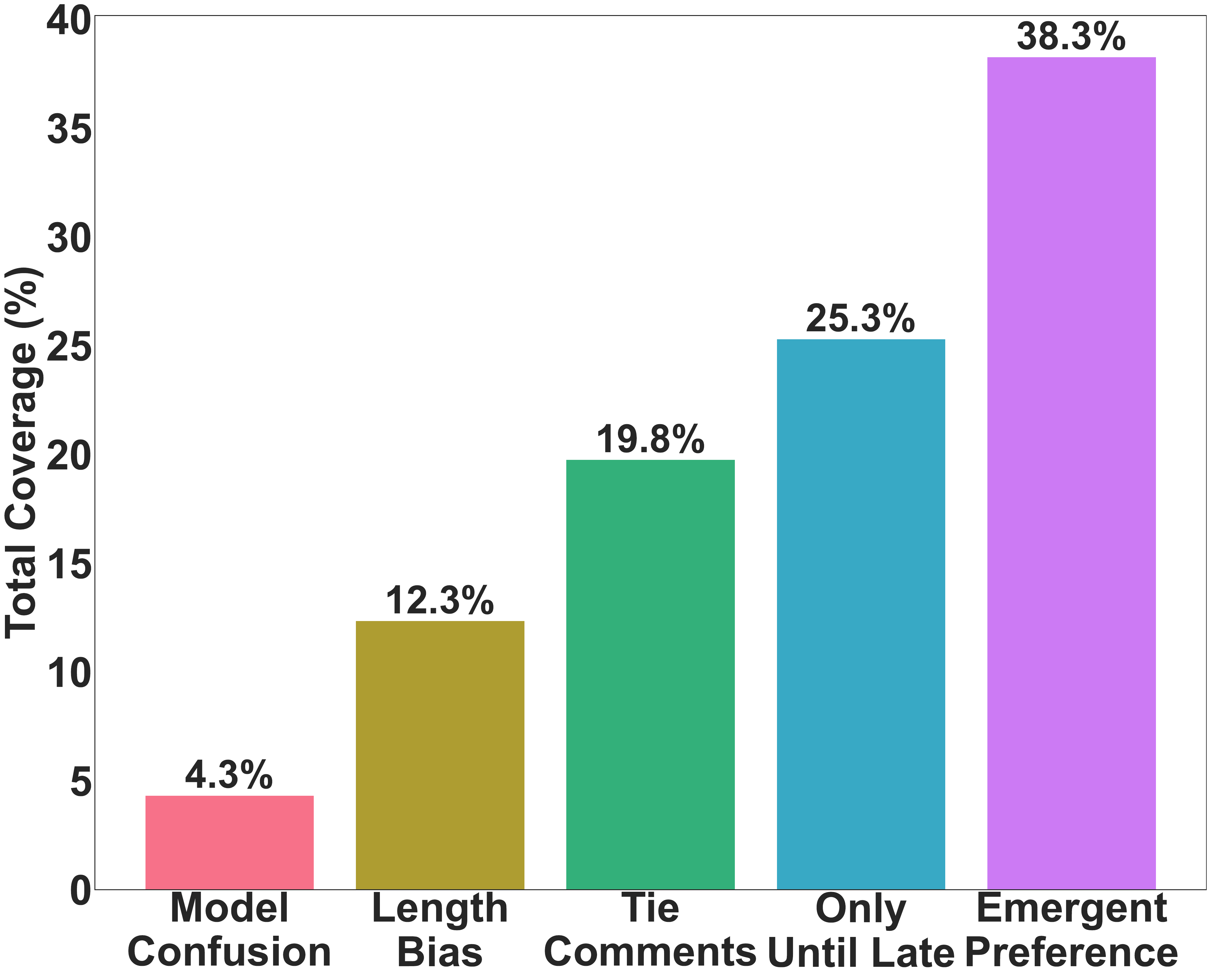}
        \caption{Error Case Attribution}
        \label{fig: simulation_diagnostic}
    \end{subfigure}
    \caption{Analyses of Simulating Human Feedback with AI Feedback.}
    \label{fig: simulation-accuracy-hist}
\end{figure}

\textbf{Results}\quad
We plot the Elo ratings of different models in~\cref{fig:elo}. The results reflect a clear trend: while vanilla GPT-4o performs on par with humans (1052 vs 947), each module enhancement improves the measured persuasiveness of the generation, and the full system receives a substantially higher Elo rating than human-written descriptions in our benchmark (1318 vs 947). To ensure a fair comparison against human descriptions, which do not have access to explicit user preferences, we note that our model variant without any personalization (\textit{Only Grounding}) still receives a higher rating than human-written content (1151 vs 947).
Also we observe that using GPT-4o to generate listing descriptions has a clear edge compared to GPT-4o-mini. Moreover, we plot empirical win rates among three major competitors (\textit{Vanilla}, \textit{Human} and \agentname) in \cref{fig:win_rates}, which directly illustrates the relative preference for \agentname in this evaluation setting.\footnote{Participants also rated their preference for each description on a 1-5 scale.}

Additional experiments in \cref{app:additional_experiments} confirm that \agentname's persuasiveness extends beyond generic LLM fluency, evidenced by an 83.3$\%$ win rate (Elo 1168) against a GPT-4o-polished human baseline. Furthermore, our custom grounding module significantly outperforms direct LLM baselines (69.4$\%$ vs. $\sim$59$\%$ accuracy), while supplementary analyses demonstrate robust Elo convergence and substantial inter-annotator agreement ($\kappa \ge 0.58$) across all evaluation tasks.

\subsection{Evaluation through AI Feedback}
Human feedback can be costly, especially as we scale the training and evaluation of our task. In this section, we report an empirical analysis of whether AI feedback can approximate the human feedback collected in the above human-subject experiments.

\textbf{Simulation Setup}\quad 
We employ an LLM to simulate the responses of buyers in the previous experiment. We use the first $K$ pairwise comparison results as $K$-shot in-context learning samples and prompt the LLM to predict the same buyer's selections for the remaining samples. We also adopt the chain-of-thought prompting format~\citep{wei2022chain} and provide the buyer's rationale comments as the information for in-context learning (see \cref{app: simulation_prompt} for the exact prompt). We use the Sotopia framework~\citep{zhou2024sotopia} to configure this simulation agent with GPT-4o-mini~\citep{openai2024gpt4omini} as the base model. 

\textbf{Metrics}\quad 
We use two metrics to evaluate the reliability of AI feedback compared to human feedback: 1) \textit{Shot-wise Simulation Accuracy (SSA)}: the prediction accuracy averaged across users for each shot; 2) \textit{User-wise Simulation Accuracy (USA)}: the prediction accuracy for each user, averaged across \#(shots). The first metric measures overall simulation accuracy across the entire population, while the second one measures simulation accuracy for each user.  

\textbf{Effectiveness of AI Feedback}\quad
The simulation results under both metrics are shown in \cref{fig: simulation_ssa} and \ref{fig: simulation_usa}. The model achieves $61.6\%$ accuracy across users and exhibits non-trivial ($>50\%$) performance for $79.2\%$ of users, suggesting that AI feedback may be useful for preliminary diagnostics. However, the accuracy remains too low for reliable replacement of human evaluation. Additionally, the variance in the USA metric is high and increases with more provided shots, underscoring the challenges of personality simulation, as highlighted in \citep{wang2024learning}. While the upward trend in variance is expected due to fewer data points, it highlights the difficulty of predicting user preferences dynamically.

To further understand the limitations of AI-simulated feedback, we conduct a manual analysis of simulation errors. Excluding the $56.1\%$ error cases that lack clearly explainable patterns, we attribute the rest of them to several key error sources in \cref{fig: simulation_diagnostic}:
1) \textit{Length Bias}: Similar to the observation in Chatbot Arena~\citep{zheng2023judging}, the model overly favors longer responses; 
2) \textit{Tie Comments}: Buyers consider the influence from descriptions as indifferent yet still cast confident votes in one of the choices; 
3) \textit{Emergent Preference}: While the model only has access to a buyer's pre-established preference, a buyer's selections in some cases reflect some unspecified preferences or ones in contradiction;
4) \textit{Only Until Late}: Correct predictions about a buyer's selection only emerge after sufficient in-context samples; 
5) \textit{Model Confusion}:
The model's prediction appears random, which indicates that the model may not have sufficient information to simulate such a buyer.
Some of these errors can be mitigated by collecting more selection data from each buyer or improving the preference elicitation process in future work.

\subsection{Hallucination Checks}
\label{sec: exp_hallucination_verification}
For grounded persuasion, it is important to ensure minimal risks of hallucination. We distinguish raw factual attributes from higher-level features: attributes such as price, bedrooms, bathrooms, and square footage are supplied directly to the generation model and checked for faithfulness, while predicted features primarily determine which persuasive angles to emphasize. Hence, we evaluate the amount of misinformation in the marketing content through fine-grained fact-checking~\citep{min2023factscore}, where we use GPT-4o to assist our hallucination check and set the listing attributes in the dataset as atomic facts. Specifically, we consider two types of factual attributes to check, $X_{\text{hard}}$ and $X_{\text{soft}}$. 
For attributes in $X_{\text{hard}}$, we require the attribute description to be completely accurate (e.g., \#(bathrooms)), whereas we allow attributes in $X_{\text{soft}}$ to be roughly accurate (e.g., address). 

Given an attribute set $X$ and a description $L$, we ask the model to perform the following tasks: $\text{supp}(L, X)$ identifies the subset of attributes in $X$ that are mentioned in $L$; $\text{eval}_{\text{hard}}(L, x)$ returns a binary value indicating whether attribute $x$ is accurately described; and $\text{eval}_{\text{soft}}(L, x)$ provides a score from 0 to 10 reflecting the extent to which $x$ is accurately described (see our prompt design in \cref{app: hallucination_experiments_details}).
We then compute the faithfulness score for attributes in $X_{\text{hard}}$ and $X_{\text{soft}}$ as follows:
\begin{align*}
\text{Faithful}_{\text{hard}}(L)
&=
\frac{
\sum\limits_{x \in \text{supp}(L, X_{\text{hard}})}
\text{eval}_{\text{hard}}(L, x)
}{
\left|\text{supp}(L, X_{\text{hard}})\right|
},\\
\text{Faithful}_{\text{soft}}(L)
&=
\frac{
\sum\limits_{x \in \text{supp}(L, X_{\text{soft}})}
\text{eval}_{\text{soft}}(L, x)/10
}{
\left|\text{supp}(L, X_{\text{soft}})\right|
}.
\end{align*}

\begin{figure}[htbp]
  \centering
  \includegraphics[width=0.7\linewidth]{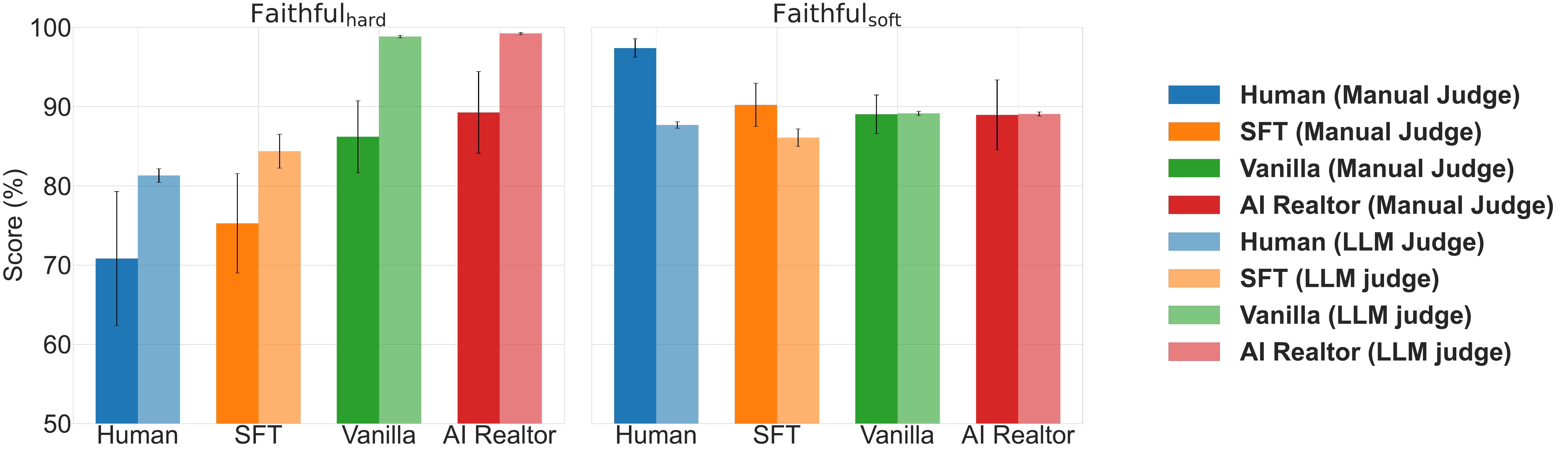}
    \caption{Faithfulness Scores for Hallucination Checks.}
    \label{fig: hallucination_comparison}
\end{figure}

As shown in \cref{fig: hallucination_comparison}, the model-generated descriptions are mostly faithful to listing information with minimal hallucination under both metrics. In contrast, the descriptions from human realtors or SFT model show an even higher level of hallucination. After digging into details, we found that this is  due to  human realtors' (also SFT's) vague description of attributes in $X_{\text{hard}}$ such as the following example,  
``\textit{This 4 bedroom, 3.5 bathroom home offers {\color{red}nearly 2,000} ({\color{blue} 1,828}) sqft of living space...}''.
Our \agentname, however, tends to accurately describe factual attributes whenever mentioned, likely due to its preference to copy from context --- interestingly, this preference seems to be forgotten by the model after supervised fine-tuning on human-written descriptions.  That said, it is debatable whether such vague descriptions of attributes is a true kind of hallucination, though some buyers did complain about this kind of language in the comments of their responses.

We replicate hallucination checks with human evaluators to validate GPT-4o's hallucination detection results. Details of the interface and annotation guidelines are provided in \cref{app: hallucination_human_eval}, and the results are shown in \cref{fig: hallucination_comparison}. \revise{Regarding ranking consistency, GPT-4o's relative ordering of models on $X_{\text{hard}}$ aligns closely with human evaluations, but diverges on $X_{\text{soft}}$,} 
highlighting the challenge of verifying loosely matched factual attributes. Overall, both human and GPT-4o evaluations show that \agentname achieves higher faithfulness on $X_{\text{hard}}$ and comparable performance on $X_{\text{soft}}$, suggesting it poses minimal risk of hallucination. Furthermore, the human evaluators report that \agentname descriptions are as trustworthy as humans (See more details of our credibility survey in \cref{app: hallucination_human_eval}).

\subsection{Qualitative Analysis}
\label{sec:qualitative_analysis}
To complement our quantitative results, we analyze specific cases to understand user preferences and the agent's adaptive capabilities.

\paragraph{The Value of Surprisal and Grounding}
Users consistently favor descriptions that highlight unique, market-relevant features ("surprisal"). For instance, in one comparison involving a property in Printers Row, a user preferred the \agentname description because it ``specifically points out the rarity of the ample storage... making the property stand out,'' whereas the baseline focused on generic amenities like "modern amenities". This demonstrates how grounded surprisal features can act as persuasive signals. (See \cref{sec: case-for-surprisal} for the complete case study.)

\begin{center}
\begin{tcolorbox}[colback=white,colframe=gray!20,width=0.95\linewidth,breakable]
{\footnotesize \textbf{Disfavored Description (Baseline):} ...Built in 1998, this condo boasts a huge bedroom suite, hardwood flooring throughout, and an inviting gas fireplace. The newly upgraded stainless steel appliances and eye-catching granite countertops make the kitchen a chef's delight...}
\vspace{0.2em}

{\footnotesize \textbf{Preferred Description (\agentname):} ...\hl{The huge bedroom suite boasts a walk-thru closet area, offering ample built-in cabinet space and additional storage -- a rarity in similarly priced listings.} Revel in the tranquility of your spacious private balcony, perfect for unwinding with views of the bustling cityscape...}
\vspace{0.2em}

{\footnotesize \textbf{User Comment:} ...Description B highlights the "expansive 876 sqft layout," and the "huge bedroom suite," emphasizing the sense of space and luxury. \hl{Description B specifically points out the rarity of the ample storage and built-in cabinetry in similarly priced listings, making the property stand out.}}
\end{tcolorbox}
\end{center}

However, statistical claims must be interpretable. In another case, the agent claimed a property was in the "top 2\% for amenities," which confused the user.

\begin{center}
\begin{tcolorbox}[colback=white,colframe=gray!20,width=0.95\linewidth,breakable]
{\footnotesize \textbf{Disfavored Description (\agentname):} ...Positioned among the top 2\% for amenities in Chicago, this condo includes in-unit laundry, an intercom system, garage parking...}
\vspace{0.2em}

{\footnotesize \textbf{User Comment:} Description B says it is in the top 2\% of amenities. What does that even mean. That is nonsense.}
\end{tcolorbox}
\end{center}
This suggests a trade-off between precision and clarity that future models must navigate.

\paragraph{Improving upon Human Baselines}
Human-written descriptions often fail due to missing information or poor formatting. In one case, a user rejected a human-written description for a Lincoln Park condo because it ``doesn't even have the size [or] location,'' whereas the agent's output was comprehensive. In another, a user disliked a human description written entirely in capital letters, noting it felt like ``getting yelled at.'' \agentname avoids these pitfalls by maintaining a professional, fact-grounded tone. (See \cref{sec: case-studies-all} for more analysis.)

\begin{center}
\begin{tcolorbox}[colback=white,colframe=gray!20,width=0.95\linewidth,breakable]
{\footnotesize \textbf{Disfavored Description (Human):} WALK TO IT ALL!! THIS BRIGHT TWO BEDROOM, 1 BATHROOM EAST LINCOLN PARK PENTHOUSE W/DECK HAS EXPOSED BRICK, BAY WINDOWS AND A WOOD BURNING FIREPLACE; EAT-IN ISLAND KITCHEN... THE UNIT HAS BEAUTIFUL HARDWOOD FLOORS...}
\vspace{0.2em}

{\footnotesize \textbf{Preferred Description (\agentname):} Welcome to 832 W Wrightwood Ave \#3, an enchanting 2-bedroom, 1-bathroom condo... Priced sensibly at \$450,000 and boasting a spacious 1,164 sqft of elegant living... Step inside to discover a warm ambiance highlighted by exposed brick, hardwood floors, and a cozy wood-burning fireplace...}
\vspace{0.2em}

{\footnotesize \textbf{User Comment:} I think this description is much better because it isn't in all caps, which feels like I'm getting yelled at.}
\end{tcolorbox}
\end{center}

\paragraph{Heterogeneity in User Preferences}
We observe significant diversity in user preferences regarding description length and style, validating the need for personalization.

\textit{Length Preference:} Some users prefer conciseness, while others desire detail.

\begin{center}
\begin{tcolorbox}[colback=white,colframe=gray!20,width=0.95\linewidth,breakable]
{\footnotesize \textbf{User A (Prefers Concise):} Description A gets to the point faster, while still highlighting the important qualities of the home.}
\vspace{0.2em}

{\footnotesize \textbf{User B (Prefers Detailed):} Again, more description is better if I am really interested in a property.}
\end{tcolorbox}
\end{center}

\textit{Style Preference:} Similarly, some prefer evocative language (``splendid charm''), while others prefer a plain listing of amenities.

\begin{center}
\begin{tcolorbox}[colback=white,colframe=gray!20,width=0.95\linewidth,breakable]
{\footnotesize \textbf{User C (Prefers Descriptive):} Description A is a bit more descriptive without going overboard, also talks about the neighborhood.}
\vspace{0.2em}

{\footnotesize \textbf{User D (Prefers Plain):} Description B does a better job at listing the amenities.}
\end{tcolorbox}
\end{center}

These findings highlight that no single style satisfies all users, underscoring the importance of our Personalization Module in adapting content to user profiles.

\paragraph{Adaptive Tone across Market Segments}
\agentname demonstrates the ability to linguistically tailor descriptions to different property tiers.
For a \textbf{low-end listing (\$110k)}, the agent emphasizes \textit{security} and \textit{efficiency}, addressing potential insecurities. Conversely, for a \textbf{high-end listing (\$1.8M)}, it shifts focus to \textit{luxury} and \textit{openness}.

\begin{center}
\begin{tcolorbox}[colback=white,colframe=blue!20,width=0.95\linewidth]
{\footnotesize
\textbf{Low-Price Representative (\$110{,}000):} ...Enjoy \textbf{tranquil moments} on the large back deck... set within a \textbf{gated courtyard that ensures privacy and security}... comfortable living \textbf{without unnecessary upkeep}... With \textbf{proactive security measures}...
\vspace{0.4em}

\textbf{High-Price Representative (\$1{,}875{,}000):} ...Discover \textbf{unparalleled elegance}... \textbf{cascading expanses of glass} invite an \textbf{abundance of natural light}... The \textbf{meticulously crafted design} places this property among the \textbf{top tier in architectural style}...
}
\end{tcolorbox}
\end{center}

This capability allows the agent to align with the distinct psychological needs of buyers in different market segments, moving beyond simple template-filling to true semantic adaptation.

%% file: past_papers/ai_realtor/related-work.tex
\section{Related Work}
\label{sec:related}

Several studies have pioneered methods in computational linguistics for understanding and measuring persuasiveness~\citep{wang2019persuasion, wei-etal-2016-post, tan2016winning}. 
The advent of large language models (LLMs) has further spurred research into their persuasive capabilities, especially as part of frontier model risk assessments by developers~\citep{durmus2024measuring, hurst2024gpt, jaech2024openai}. A major focus has been on the potential for LLM-generated propaganda in politically sensitive contexts~\citep{voelkel2023artificial, goldstein2024persuasive, hackenburg2024evidence, luciano2024hypersuasion}. 
Parallel investigations examine settings such as personalized persuasion~\citep{hackenburg2024evaluating,salvi2024conversational,matz2024potential}.  \citet{breum2024persuasive} and multi-round persuasion~\citep{breum2024persuasive}.
\citet{takayanagi2025can} assess the influence of GPT-4’s ability to generate financial analyses to audiences.
Complementary research has probed related LLM capabilities including negotiation~\citep{bianchi2024well}, debate~\citep{khan2024debating}, sycophancy \citep{sharma2023towards, denison2024sycophancy}, as well as the emergence of strategic rationality in game-theoretic settings~\citep{chen2023emergence, ramansteer}.

In a similar application domain,  \citet{angelopoulos2024value} conduct an experiment to generate marketing email with a fine-tuned LLM and report a 33\% improvement in email click-through rates compared to human expert baselines. \citet{singh2024measuring} design an evaluation benchmark based on a dataset of tweet pairs with similar content but different wording and like counts. 
In comparison, our work develops a full agentic framework for automated marketing, from learning domain expert knowledge to crafting localized features, and receives higher human preference ratings than the supervised fine-tuning baseline in our human-subject experiments.

%% file: past_papers/ai_realtor/discussion.tex
\section{Discussion}
\label{sec:conclusion}
\textbf{Contributions and Implications}~ This paper presents a novel framework for persuasive language generation, marking a first step toward integrating signaling concepts from economic theory into agentic LLM design. Our results demonstrate that this structured approach outperforms professional human-written listing descriptions in our controlled real-estate benchmark. A central tenet of our design is the deliberate prioritization of factual grounding. While human-written descriptions often employ stylized or emotionally resonant language, we argue that in domains where accuracy is paramount, constraining generation to verifiable facts is a necessary and responsible choice. Our framework's effectiveness stems from its ability to map raw attributes to a compact set of high-level, market-relevant features, ensuring that the generated content is both persuasive and credible.

\textbf{Limitations and Future Directions}~ Despite these promising results, we acknowledge several limitations that highlight avenues for future research. The primary bottleneck remains the reliance on high-quality human feedback for evaluation. Our experiments with automated, LLM-based evaluators show promise for assessing factuality but are not yet reliable for measuring nuanced qualities like persuasiveness, underscoring the need for more sophisticated evaluation benchmarks. \revise{Second, our empirical validation is currently concentrated on the Chicago market. While our agentic workflow is designed to admit localization, persuasion is culturally and economically sensitive; future work is required to verify stability across diverse geographic regions and demographics. Furthermore, generalizing this framework to domains with less structured inputs or different persuasive norms (e.g., brand marketing vs. legal arguments) presents a significant and important challenge.}

Building on this foundation, several exciting directions emerge. The modularity of our framework is well-suited for incorporating domain-specific constraints. For regulated fields like housing or finance, integrating compliance filters or legal principles inspired by approaches like Constitutional AI~\citep{bai2022constitutional} is a crucial next step to ensure responsible deployment. Moreover, to address the trade-off between factuality and expressiveness, future work could explore incorporating a wider range of persuasion theories, such as emotional appeals and narrative structures, as controllable modules within the agentic design. Finally, scaling our datasets, expanding to new copywriting domains, and conducting more extensive real-world A/B testing will be essential to fully unlock the potential of theory-grounded persuasive generation.

\section{Conclusion}
\label{sec: conclusion_section}
In this work, we introduced \agentname, a modular framework for grounded persuasive language generation applied to real estate marketing. By using economic signaling theory to structure an LLM-based agent, our system systematically identifies unique, market-relevant property attributes ("surprisal") and tailors descriptions to user preferences. Extensive human evaluations demonstrate that \agentname receives higher persuasiveness ratings than both professional human realtors and baseline LLM approaches in our benchmark, while maintaining high factual fidelity. Our qualitative analysis further confirms that users value the agent's ability to balance evocative language with concrete details, adapting its tone across different market segments. These findings establish a rigorous baseline for automated copywriting and suggest that grounding generation in verifiable data is key to effective and responsible persuasion.

\section{Ethics Statement}
Our research on persuasive language generation acknowledges the dual-use nature of such technologies. We have proactively centered our work on grounded persuasion, where generated content is constrained by verifiable facts, to mitigate the risks of misinformation and manipulation. Our extensive hallucination checks, detailed in \cref{sec: exp_hallucination_verification} and \cref{app: hallucination_experiments_details}, confirm that our agent maintains a high degree of factual accuracy, comparable to or exceeding that of human experts.

All human-subject experiments were conducted in compliance with ethical research standards. The study protocol received IRB approval (exempt). Participants were recruited from the Prolific platform, informed of the study's purpose, and compensated at a fair rate (approximately \$20/hour with performance incentives). The dataset, derived from publicly available Zillow listings, was processed to remove any personally identifiable information, ensuring user privacy.

By focusing on a high-stakes, fact-driven domain like real estate, we aim to provide a framework for developing responsible persuasive AI. We believe this work serves as a foundation for future research into the ethical guardrails necessary for deploying strategic language models in real-world applications and encourage continued investigation into their broader societal implications.

%% file: chapters/base_aligned_collaboration.tex
\section*{Chapter Overview}
Alignment has greatly improved large language models' output quality at the cost of diversity. In this chapter we present Base-Aligned Model Collaboration (\name{}), an inference-time token-level framework that dynamically combines a base LLM with its aligned counterpart. Using uncertainty and content-based routing, \name{} jointly improves diversity and quality in a single decoding pass and offers a controllable trade-off. This chapter is based on work accepted to ICML 2026~\citep{wang2025optimizing}.

\graphicspath{{./}}

\makeatletter
\def\input@path{{./}}
\makeatother

\input{past_papers/base_aligned_collaboration/sections/introduction}

\input{past_papers/base_aligned_collaboration/sections/methodology}

\input{past_papers/base_aligned_collaboration/sections/new_experiments}

%% file: past_papers/base_aligned_collaboration/sections/introduction.tex
\section{Introduction}

\begin{figure}[htbp]
    \centering
        \begin{overpic}[width=0.92\textwidth]{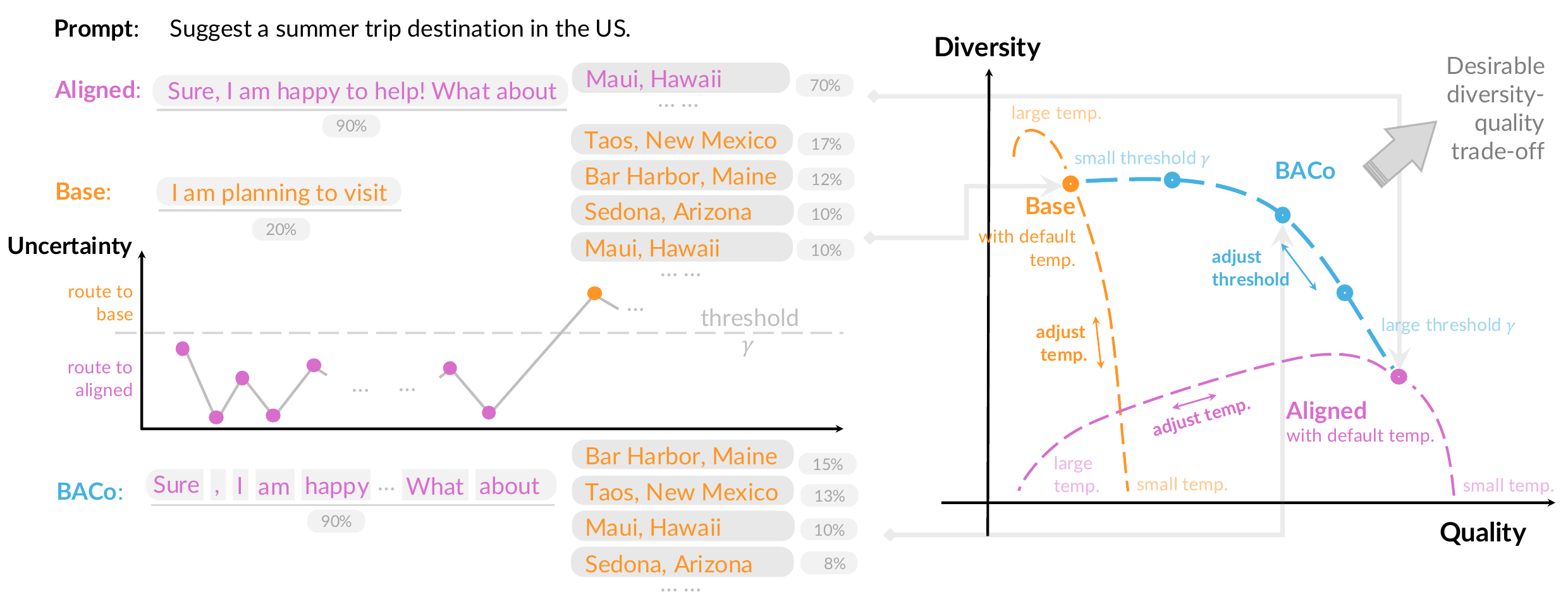}
        \put(0, -3.7){{(A) Example outputs from base, aligned, and \name{} models.}}

        \put(62, -3.7){{(B) The \problem{} space.}}
    \end{overpic}
    \vspace{15pt}
    \caption{
    \name{} is an inference-time token-level model collaboration framework that combines a base model's diversity with its aligned counterpart's quality. 
    (A) A comparison of generated outputs. 
    The aligned model produces high-quality but low-diversity outputs, while the base model produces high-diversity but low-quality outputs. 
    \name{} optimizes both diversity and quality by dynamically \emph{routing} between them.
    The probabilities of token(s) are in grey next to text boxes. 
    (B) Illustration of the \problem{} space. Single models face a steep trade-off, where improving diversity by adjusting configuration (e.g., by increasing temperature) degrades quality. \name{} achieves a better Pareto curve and allows for easy traversal across this frontier by adjusting the router's threshold.
    }
    \vspace{-8pt}
    \label{fig:intro}
\end{figure}

While alignment greatly improved large language models (LLMs)' output \emph{quality} in terms of instruction following and downstream task performance \citep{ouyang2022training}, it results in a stark reduction in output \emph{diversity} \citep{kirk2023understanding,zhang2025noveltybench,west2025base, spangher-etal-2025-creative, huang2025teaching}. 
Across repeated sampling, a model after alignment (\ie{} aligned model) tends to produce highly similar outputs, whereas a model before alignment (\ie{} base model)
yields diverse outputs. 
For example, when prompted with ``suggest a summer trip destination in the US,'' a base model may produce diverse destinations across generations, while the aligned model often converges on a single dominant one (\Cref{fig:intro}A).
This \emph{\problem{}} 
undermines utility in open-ended
generation tasks (\eg{} creative writing and dialogue) by 
encouraging formulaic language use \citep{zhang2024lists,chakrabarty2025can}, diminishing creativity \citep{west2025base},
and suppressing ideation in human-AI interaction \citep{padmakumar2023does,Meincke2025,ashkinaze2025ai, spangher2025novel, spangher2025newsinterview}.
These findings motivate methods to improve diversity in aligned LLMs.

Prior diversity-promoting methods  attempt to address 
the \problem{} at both the training and inference stages (\Cref{sec:related_work}). The former \citep{lanchantin2025diverse, chung2025modifying, li2025jointlyreinforcingdiversityquality} incorporate explicit diversity objectives into preference optimization during reinforcement learning.
While effective at improving diversity, such methods require modifying the model's output distribution, which can compromise desirable alignment properties such as safety and helpfulness \citep{qi2023fine}.
The latter consists of decoding-based techniques, such as adjusting temperature and beam search \citep{vijayakumar2016diverse}, as well as prompt-based techniques, including in-context learning \citep{meyerson2024language}, prompt paraphrasing \citep{zhang2025noveltybench}, and multilingual back-translation \citep{wang2025multilingual}.
However, these inference-time techniques typically require multiple decoding passes or long-horizon planning to improve diversity, and may still disproportionately degrade generation quality \citep{peeperkorn2024temperature}.
This reveals a fundamental limitation of forcing a single model to excel at both diversity and quality. 

To overcome this limitation, we introduce Base-Aligned Model Collaboration (\name{}), an inference-time token-level model collaboration framework (\Cref{sec:method-main}) that combines the complementary strengths of a base model for diversity and its
aligned counterpart for quality.
\name{} operates via a lightweight, token-level \emph{routing strategy} that dynamically switches between the two models in a single decoding pass, requiring no fine-tuning or prompting (\Cref{fig:intro}A). 
This design is motivated by recent findings suggesting that collaboration between base and aligned models is feasible.
Specifically, \citet{fei2025nudging} demonstrate that base and aligned models largely agree on next-token predictions, 
a phenomenon known as \emph{superficial alignment} \citep{lin2023unlocking}.
In contrast to the prior work that leverages this phenomenon to improve
base-model quality, our goal is to jointly optimize diversity and quality,
surpassing either model alone through collaboration (\Cref{fig:intro}B).

Importantly, the diversity-quality trade-off cannot be resolved by a fixed design.
Different tasks naturally favor different operating points along this spectrum, and user preferences may further vary based on context.
A practical framework, therefore, requires \textit{controllability}, the ability to adjust generation along the diversity-quality spectrum on demand.
\name{} provides such controllability through an adjustable routing threshold (\Cref{fig:intro}), which continuously tunes the contribution of the base
and aligned models during decoding.
In addition, we introduce a family of routing strategies within a unified design space (\cref{sec:method:router}) that consistently improve the
diversity-quality trade-off. 
These strategies capture complementary signals and exhibit different strengths across diversity dimensions, providing an additional axis of control beyond the routing threshold.

We validate our approach across three open-ended generation tasks: instruction following, dialogue, and creative writing (\Cref{sec:experiments}).
In open-ended generation, both diversity and quality can be defined in numerous ways, and no single metric provides a complete evaluation.
Therefore, rather than treating evaluation as a single objective, we formulate it as a collection of bi-dimensional diversity-quality trade-off spaces, each defined by a specific pair of metrics.
Focusing on diversity, we evaluate 11 diversity and 2 quality metrics, resulting in $11 \times 2$ spaces, to assess \name{}'s consistent
improvements. 
In addition to instruction-following and dialogue tasks which have relatively short outputs, we also evaluate long-form generation, where we measure long-form diversity at the discourse level in terms of plot structure and emotional flow \citep{tian2024large}.
Our evaluation focuses on open-ended generation settings where diversity is a first-class objective, for which benchmarks designed primarily for single-answer
accuracy (e.g., math and code reasoning) are less suitable.
Nevertheless, as a cross-check, we additionally evaluate \name{} on tasks with verifiable quality criteria, including mathematical reasoning and verifiable instruction following, to ensure that diversity improvements do not
come at the cost of severe quality degradation.
We complement automatic evaluations with human evaluations of diversity and quality.

In our experiments (\Cref{sec:results}), we demonstrate that \name{} achieves a new state-of-the-art in optimizing the \problem{} in inference time. 
Overall, \name{} with our best router achieves a 21.3\% improvement over the strongest
inference-time baselines, with even larger gains on semantic diversity metrics.
These improvements are consistent across tasks and evaluation metrics, and are
further supported by human evaluations.
By viewing the base and aligned models as two checkpoints during training, our results suggest that collaboration across checkpoints can enable effective control over diversity and quality.
Overall, \name{} provides a simple framework for base-aligned model collaboration, effectively improving both diversity and quality. 

In summary, our contributions are threefold:
\begin{enumerate}[wide, labelwidth=!, labelindent=0pt]
\item[\circone] We propose \name{}, an inference-time token-level model collaboration framework that combines a base model and its aligned counterpart, along with a family of lightweight routing strategies, to produce high-diversity and high-quality outputs across generations. 
\item[\circtwo] We formulate the diversity-quality trade-off as a collection of bi-dimensional evaluation spaces and conduct a comprehensive evaluation across $11 \times 2$ metric pairs, including long-form 
diversity and human evaluation.
\item[\circthree] Through extensive experiments on three open-ended generation tasks (i.e., instruction following, dialogue, and
creative writing), we show that \name{} consistently outperforms strong 
baselines.
\end{enumerate}

%% file: past_papers/base_aligned_collaboration/sections/methodology.tex
\section{Preliminary}
\fakeparagraph{Large Language Models (LLMs).} LLMs are typically trained to autoregressively predict the next token of the output $\outputval$ given a prompt $\inputval$.
The conditional probability is factorized as
$P\left(\outputval | \inputval; \theta \right)=\Pi_{t}P\left(\outputval_t | [\inputval, \outputval_{<t}]; \theta \right)$, where $\outputval_{<t}$ denotes the output prefix generated up to position $t$-$1$, and $\theta$ denotes the model parameter.

Alignment is the process of fine-tuning an LLM to align its outputs with human intent, ethical principles, and desired behavioral norms, typically through instruction tuning or reinforcement learning from human feedback (RLHF) \citep{ouyang2022training, bai2022training}.
We use \textit{base models} to denote models without alignment tuning (e.g., \texttt{Llama-3-8B}) and \textit{aligned models} to denote those further optimized with alignment (e.g., \texttt{Llama-3-8B-Instruct}) \citep{dubey2024llama}.

\fakeparagraph{Diversity and Quality Measurement.} 
In this paper, we measure diversity over a group of outputs independently generated from the same prompt $x$: $\mathcal{Y}(x) = \{ y^{(1)}, \dots, y^{(k)} \}, y^{(i)} \sim P(\cdot \mid x; \theta)$. This group-level diversity is denoted as $D(\mathcal{Y})$ \citep{kirk2023understanding, west2025base}  (e.g., the clustering-based approach in ~\citet{kuhn2023semantic}). 
Quality is modeled as $Q(y | x)$ for each output given the prompt, typically by a reward model or human evaluator, reflecting linguistic fluency and instruction-following \citep{lambert2024rewardbench, zhang2025noveltybench}. 
The group-level quality is then defined as the average quality across all outputs in a group: 
$Q(\mathcal{Y}) = \sum_{i=1}^k Q(y^{(i)} | x)$.
For simplicity, we refer to group-level diversity and group-level quality as diversity and quality in this work.

\paragraph{Diversity-Quality Trade-off.}
\label{sec:problem}

Alignment, while effective at improving output quality, comes at the cost of reduced output diversity \citep{lu2025ai, west2025base, yang2026alignment}. 
To demonstrate this trade-off and quantify its magnitude,
we run a preliminary experiment with \texttt{Llama-3} on a subset of WildChat~\citep{zhao2024wildchat}.
We first situate the performance of the base and aligned models with
\begin{table}[t]
\centering
\small
\renewcommand{\arraystretch}{1.2}
\begin{tabular}{lcc}
\toprule
& \textit{Diversity} & \textit{Quality} \\
\textit{Model} & \textit{(\#Clusters)} $\uparrow$ & \textit{(Reward)} $\uparrow$ \\
\midrule
\texttt{Llama-3-8B}     & \textbf{8.13} & 1.28 \\
\texttt{Llama-3-8B-Instruct} & 2.58 & \textbf{7.62} \\
\bottomrule
\end{tabular}
\vspace{7pt}
\caption{Diversity and quality of \texttt{Llama-3-8B}'s base and aligned models in our preliminary experiment. 
The results demonstrate a clear \problem{} in the two model's performance.
Diversity is measured by the number of semantic equivalent clusters of the output group, and quality is the average reward per output from another LLM. 
}
\label{fig:prelim}
\end{table}
default configuration (the two noted points in \Figref{fig:intro}B) within the diversity-quality space.
Following the evaluation protocol of the diversity-focused benchmark NoveltyBench \citep{zhang2025noveltybench}, we evaluate \texttt{Llama-3}'s base and aligned models on an open-ended subset of WildChat \citep{zhao2024wildchat}.
Here, diversity is measured as the number of semantic equivalent classes of the output group via \cite{zhang2025noveltybench}'s clustering, and quality is measured as the average reward per output from \texttt{Skywork-Reward-Gemma-2-27B} \citep{liu2024skywork}.\footnote{These are two of many possible measurements introduced later in the paper. We use them here as representative examples for the pilot study, as they are among the widely adopted metrics.} We sample 10 outputs per prompt.
As illustrated in \Cref{fig:prelim}, the \problem{} is stark: the base model is \textbf{3.15x more diverse}, whereas the aligned model has \textbf{5.95x higher quality}. 
Inherently, this performance trade-off stems from alignment's tendency to reduce the entropy of the next-token prediction distribution, 
concentrating probability mass on fewer, high-quality tokens, a phenomenon known as \textit{mode collapse} \citep{lin2023unlocking, Shumailov2024, hamilton2024detecting, yang2026alignment, cui2025entropy}.

This presents a dilemma: one can either use a high-diversity but low-quality base model, or a high-quality but low-diversity aligned model.
The single-model paradigm is insufficient, as neither extreme is ideal for all applications.
Hence, we argue that an ideal method is able to pursue the best of both worlds. 
To this end, we formalize the problem in a two-dimensional \emph{diversity-quality space} $S=\{(D, Q)\}$. 
 In this space, any given method under specific configurations $\gamma$ (\eg{} sampling parameters) is evaluated to be a single point.
 An ideal method, by adjusting $\gamma$, should approximate the \emph{Pareto frontier}: the set of optimal solutions where diversity cannot be improved without sacrificing quality, and vice versa. 
 This frontier represents the best possible trade-offs.

\section{\name{}: Base-Aligned Collaboration for  Diversity and Quality}
\label{sec:method-main}

Recent work has provided empirical evidence for the superficial alignment hypothesis \citep{zhou2023lima, lin2023unlocking}, which suggests that a base model and its aligned counterpart largely agree on next-token predictions.
Building on this observation, \citet{fei2025nudging} show that introducing only a small fraction of aligned-model tokens into a base model’s decoding can recover task-specific performance comparable to that of the aligned model. However, such
approaches primarily focus on improving the quality of the base model, without explicitly addressing the loss of diversity introduced by alignment.

Motivated by this gap, we hypothesize that
\begin{tcolorbox}[colback=gray!5!white, colframe=gray!60!black]
{Collaboration between a less-aligned,
higher-diversity model and a more-aligned, higher-quality model during inference
can better balance the diversity-quality trade-off than either model alone.}
\end{tcolorbox}
Base and aligned models form a natural pair to test this hypothesis, as they are readily available off the shelf and exhibit complementary strengths.

Based on this hypothesis, we propose \textbf{\name{}}, an inference-time framework that orchestrates collaboration between a base model ($P_{\text{base}}$), serving as
a source of diversity, and its aligned counterpart ($P_{\text{aligned}}$), serving
as a source of quality, at the token level.\footnote{The token-by-token nature of LLM autoregressive decoding makes token-level control feasible.}

At the core of \name{} is a \textbf{router}, a lightweight decision module that determines, at each decoding step, which model should generate the next token. 
The router operates according to \textbf{routing strategy(ies)} $\mathcal{R}$, which selects between base and aligned models accordingly.\footnote{The router can be based on a single routing strategy or combine multiple ones.}
Intuitively, the router acts as a ``gatekeeper'': it routes the next token generation to the base model when diversity is desired and to the aligned model when quality is desired.

Formally, \name{} orchestrates the two models as:
\begin{small}
\begin{equation}
\label{eq:moe_main}
\begin{split}
    P_{\text{\name{}}}(\outputval_t | c_t) ={} & w_{\text{base}} \cdot P_{\text{base}}(\outputval_t | c_t; \theta_{\text{base}}) \\
    & + (1-w_{\text{base}}) \cdot P_{\text{aligned}}(\outputval_t | c_t; \theta_{\text{aligned}})
\end{split}
\end{equation}
\end{small}
where $c_t = [\inputval, \outputval_{<t}]$, and the gating weight $w_{\text{base}} \in \{0,1\}$ for each candidate token $\outputval_t$ is given by the router:
\begin{small}
\begin{equation}
\label{eq:moe_gate_corrected}
    w_{\text{base}} = \mathbb{I}\left[\mathcal{R}\left(\outputval_t | c_t, P_{\text{base}}, P_{\text{aligned}}\right)=\text{base}\right]
\end{equation}
\end{small}

In practice, since one word may consist of multiple tokens, we restrict switching to word boundaries to prevent erroneous generation when the two models use different tokenizations. 
Full decoding pseudocode is provided as Algorithm~\ref{alg:baco} in \appref{app: implementation_details}; further implementation and discussion are in \appref{app: implementation_details} and \appref{app:analysis}.

\subsection{Routing Strategy Design}
\label{sec:method:router}

At the core, \name{} is a lightweight router that determines which model to route to at each decoding step. 
A routing strategy specifies (\textit{i}) what information to use as prior for routing decisions, and (\textit{ii}) which model to switch to given the information.
Each strategy includes a \textit{threshold parameter} $\gamma$ that, analogously to decoding temperature, continuously adjusts the diversity-quality balance: larger $\gamma$ biases toward the base model (more diversity), smaller $\gamma$ biases toward the aligned model (more quality).

We design routing strategies based on two complementary signal categories: logit-based and content-based. Each category captures
a distinct 
yet widely applicable perspective.

\textbf{Logit-Based.} 
Logit-based strategies leverage the next-token prediction distribution to infer the model
uncertainty, reflecting a \emph{model-centric} perspective \citep{fei2025nudging, zheng2024citer, leviathan2023fast}.
High uncertainty, indicated by low maximum probability or high entropy, suggests that
multiple continuations are plausible, making such positions natural opportunities for diversification \citep{yang2026alignment, wang2025beyond}.
Concretely, we implement:
\circone \textbf{\name{}-\textsc{P}} routes to the base model when its maximum token probability falls below a threshold $\gamma$, i.e., $\max_{y_t} P_{\text{base}}(y_t \mid \cdot) < \gamma$;
\circtwo \textbf{\name{}-\textsc{H}} routes to the base model when its next-token
entropy exceeds, i.e., $H_{\text{base}}(\outputVar_t \mid \cdot) = \sum_{\outputval_t} P_{\text{base}}\left(\outputval_t | \cdot \right)\log P_{\text{base}}\left(\outputval_t | \cdot \right) > \gamma$; \etc

\textbf{Content-Based.} 
Content-based strategies adopt a \emph{language-centric} perspective, making routing
decisions based on the semantic roles of predicted tokens.
The motivation is twofold.
First, linguistic structures such as content words often correspond to semantic or stylistic branch points where diversity is most perceptible to humans \citep{yao2019plan, sims2019literary}.
Second, \cite{lin2023unlocking, fei2025nudging} suggest that disagreements between base and aligned models often arise over stylistic tokens, such as formatting tokens (\eg{} `\textbackslash n') or function words (\eg{} `and', `if'). 
We therefore implement:
\circone \textbf{\name{}-\textsc{Punc}} routes to the aligned model when its top-ranked token is either a punctuation or formatting token;
\circtwo \textbf{\name{}-\textsc{FC}} routes to the aligned model for function words to preserve stylistic coherence.
Content-based strategies are also applicable to black-box models, as they do not require
access to logits.

In practice, combining logit-based and content-based strategies yields the strongest
performance, as they rely complementary signals.
For example, \name{}-\textsc{P-FC} prioritizes function-word routing (-\textsc{FC}) before falling back to probability-based decisions (-\textsc{P}).
In the next section, we evaluate a wide range of routing strategies and find that many variants are effective. For clarity, we present representative strategies in the main paper, with additional variants and implementation details in Appendix~\ref{app:router}.
We focus on the two categories of routing strategies and leave other routers (e.g., learned routers) for future work.\footnote{We exclude learned routers because (1) simple heuristics already yield substantial gains; (2) diversity is inherently multi-dimensional and cannot be captured by a single metric, leading to conflicting objectives and unstable training when jointly optimized (\Cref{sec:auto_eval}); and (3) diversity evaluation requires group-level sampling, which would substantially increase training cost and complexity.}

%% file: past_papers/base_aligned_collaboration/sections/new_experiments.tex
\section{Experimental Setup}
\label{sec:exp-setup}
\label{sec:experiments}
We design our experiments to empirically validate \name{}'s central goal: to optimize the diversity-quality trade-off. 

\fakeparagraph{Datasets.} We mainly evaluate \name{} across three representative open-ended generation tasks: {NoveltyBench} \citep{zhang2025noveltybench} for instruction following, {WildChat} \citep{zhao2024wildchat} for dialogue, and {Narrative-Discourse} \citep{tian2024large} for creative writing. Together, these datasets cover both short- and long-form open-ended generation across varying levels of task complexity. For full dataset details, we refer readers to \appref{app:datasets}.

\fakeparagraph{Baselines.} We compare \name{} with inference-time methods across five categories:
\circone~ {Single-model:} a base model or an aligned model, 
each sampled at varying temperatures.  
\circtwo~ {Prompting-based:}
in-context resampling \citep{meyerson2024language, zhang2025noveltybench}, where $n$ outputs are generated sequentially within a single dialogue;
paraphrase prompting \citep{jiang2020can, zhang2025noveltybench}, where paraphrased variants of the same instruction are used to increase output diversity.  
\circthree~ {Decoding-based:} Diverse Beam Search \citep{vijayakumar2016diverse}, where a diversity penalty was added to the beam search algorithm.  
\circfour~ {Ensemble-based:}
response ensemble, where $n/2$ outputs are sampled from the base model and its aligned counterpart, and pooled into a single group; 
logit ensemble, which merges the next-token probability distributions of the two models before sampling.
\circfive~ {Collaboration-based:} 
\textsc{nudging} \citep{fei2025nudging}, where an aligned model selectively introduces tokens during a base model's decoding.
Note that diverse beam search, paraphrase prompting, and in-context resampling require additional computation; in-context resampling does not perform parallel sampling.\footnote{A wall-clock runtime comparison across all inference-time methods is provided in Appendix~\ref{app: implementation_details}.}
These methods therefore provide more competitive reference points.
The {inference setups} and {experimental scope} are provided in \appref{app:exp_setup}.

\fakeparagraph{\name{}.}
Our experiments leverage two open-weight model pairs: \texttt{Llama-3-8B} and \texttt{Llama-3-8B-\allowbreak Instruct} \citep{grattafiori2024llama}, and \texttt{Olmo2-7B} and \texttt{Olmo2-7B-\allowbreak Instruct} \citep{olmo20242} as they are widely used in literature (\eg{} \citet{fei2025nudging}).
We implement the single-strategy routers (\eg{} \textsc{-P} which is based on  maximum token probability) from \Secref{sec:method:router} 
and denote multi-strategy routers as ``\textsc{-X-Y}'', where strategy Y precedes X (e.g., \textsc{-P-Punc}, \textsc{-P-FC}, and \textsc{-H-Punc}). 
These implementations serve as representative examples that demonstrate the possible design space 
of the \name{} framework. \name{} framework works well 
with a wide range of routers.
We include two basic routers as baselines:
\circone~\textsc{-Rand} routes to the base model by random chance $\gamma$;
\circtwo~{-\textsc{Judge}} employs an external model to evaluate candidate tokens and makes a routing decision.
Refer to \appref{app:router:notation} for more details.

\begin{figure}
    \centering
    \includegraphics[width=0.55\linewidth]{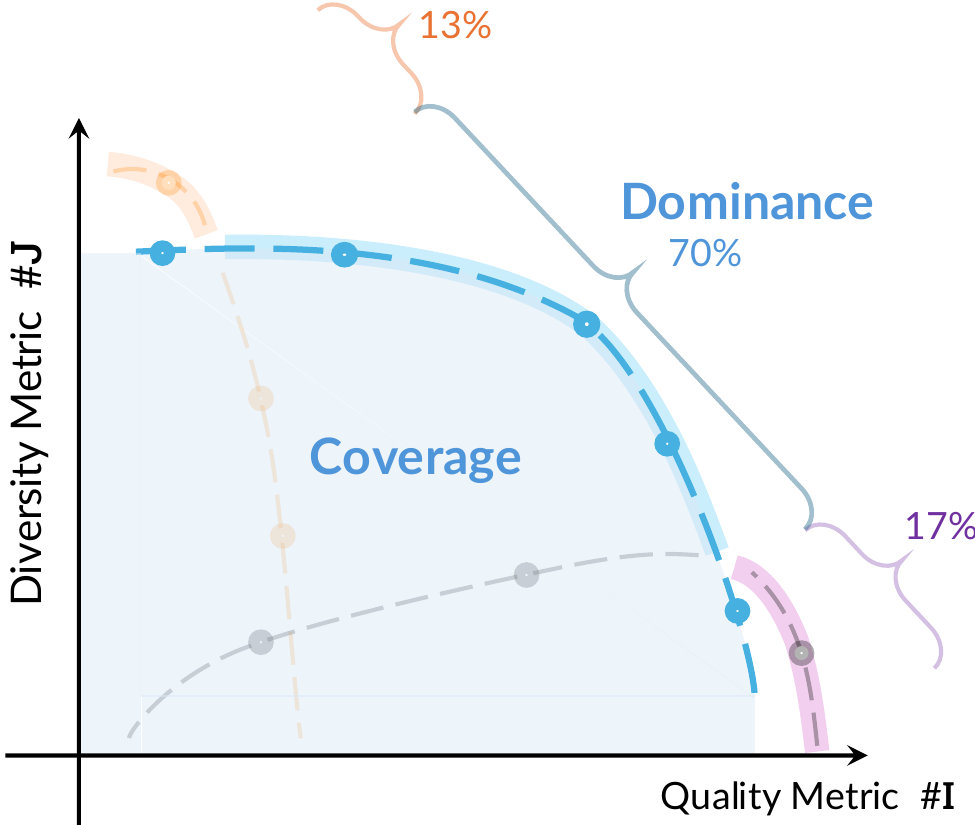}
\caption{Illustration of the indicators on diversity-quality space: \textbf{Coverage}, the area under a method’s trade-off curve (blue shading for the blue method);
\textbf{Dominance}, the proportion of the global Pareto frontier (highlighted curves) contributed by the method.
}
\vspace{-10pt}
    \label{fig:indicators}
\end{figure}

\input{past_papers/base_aligned_collaboration/tables/sub_average}

\subsection{Automatic Evaluations}
\label{sec:auto_eval}

We next evaluate the \problem{} of each method using automatic metrics. 
Prior work has proposed a wide range of diversity evaluation that apply different lexical and semantic metrics,
reflecting different perspectives on language diversity.
Since our goal is to improve general diversity and quality rather than optimize for any specific metric, we adopt 11 established diversity metrics and 2 quality metrics which form 11 $\times$ 2 diversity-quality spaces, and then aggregate their results.

Moreover, we aim to quantify the \textit{controllability} of each method, \ie{} the ability to adjust along the diversity-quality
spectrum according to task or preference.
As shown earlier in \Cref{fig:intro}B, each method is not evaluated as an individual point in a fixed configuration, but a curve formed by a sequence of points, illustrating the diversity and quality performance in different configurations. 
Each curve illustrates the trade-off of a specific method.
To enable clear comparison, we apply two indicators from multi-objective optimization, {\texttt{Coverage (Cov.)}} and \texttt{Dominance (Dom.)}, to aggregate the curve-shaped performance across all spaces into numerical results (\Figref{fig:indicators}).\footnote{All curves in \Figref{fig:indicators} are illustrative only and do not correspond to the actual performance of any method.}

\textbf{Coverage (Cov.)} indicator quantifies the area under a method’s diversity-quality trade-off curve, following the hypervolume formulation used in multi-objective optimization. 
It measures how effectively a method traverses the diversity-quality spectrum as its control configuration varies.
A larger Coverage value indicates that the method maintains good performance across a wide range of regions on diversity-quality spaces, offering usability to more different tasks or preferences overall.

\textbf{Dominance (Dom.)} 
indicator captures comparative optimality: whether and how often a method contributes to the global Pareto frontier among all methods.
We compute the global Pareto frontier across all methods and apply the C-metric \citep{zitzler1999evolutionary} to measure the portion of the frontier attributed to each method. 
A higher Dominance value indicates that a method achieves uniquely strong trade-offs unattainable by others.

These indicators are instantiated under the 11 $\times$ 2 diversity-quality spaces. And we report the average across the spaces.

\textbf{Lexical diversity} spaces use diversity metrics such as Distinct-$n$ (Dist-$n$; \citealp{li2015diversity}), EAD-$n$ \citep{liu-etal-2022-rethinking}, and Self-BLEU \citep{montahaei2019jointly}.  
\textbf{Semantic diversity} spaces rely on diversity metrics such as embedding cosine dissimilarity \citep{kirk2023understanding}, Vendi Score (embedding) \citep{friedman2022vendi},  NLI diversity \citep{stasaski-hearst-2022-semantic}, and Semantic Entropy \citep{kuhn2023semantic}.\footnote{Each metric could include multiple variants. For example, Dist-1/2/3 for different n-gram, or cosine dissimilarity under different pretrained encoders. Each variance leads to a separate diversity-quality space.}
Since lexical and semantic metrics capture fundamentally different aspects of diversity, we analyze them separately in addition to reporting aggregated results. 
Increases in lexical diversity are relatively easy to achieve (for example, by raising the temperature), yet they mostly alter surface-level phrasing without changing meaning. 
In contrast, semantic diversity reflects deeper diversity in meaning, intent, and ideas, which is harder to elicit but more human-like and valuable in open-ended generation.
\textbf{Quality} metrics include (\textit{i}) perplexity under the aligned model, which reflects fluency and instruction following, and (\textit{ii}) reward modeling scores predicted by \texttt{Skywork-Reward-Gemma-2-27B} \citep{liu2024skywork}, the state-of-the-art model on RewardBench \citep{lambert2024rewardbench}.
These metrics are paired with diversity metrics to form subspaces.   
Finally, \textbf{overall} results average across all subspaces, yielding holistic method-level indicators. 
Hereafter, we use \emph{lexical} to denote average results on all lexical diversity-quality spaces, \emph{semantic} denotes semantic diversity-quality spaces, and \emph{overall} averages all spaces in every result table.
Full derivations and implementation details are in \appref{app:auto_eval:div}. 

\section{Results}
\label{sec:results}

We first compare \name{} against strong inference-time baselines across datasets and
diversity-quality spaces (\Cref{ssec:main_result}). We then analyze the behavior of
different routing strategies in a controlled setting (\Cref{ssec:router_comparison}),
followed by ablations on model pairing (\Cref{ssec:ba_aa}). Finally, we evaluate
long-form diversity (\Cref{ssec:storyarc}) and validate automatic metrics with
human judgments (\Cref{ssec:human_eval}).

\subsection{Overall Performance}
\label{ssec:main_result}

\Tabref{tab:avg_results} summarizes the performance of \name{} and all baselines,
aggregated across datasets and metrics.
Overall, \name{} improves Coverage by \textbf{0.142} and achieves \textbf{32.7\%}
Dominance across all evaluation spaces.
Specifically, a Coverage improvement of 0.142 expands the achievable diversity-quality area by over 30\% relative to the strongest baseline.
And a Dominance of 32.7\% indicates that \name{} contributes nearly one-third of the global Pareto frontier over all baselines.
The advantage is particularly pronounced in semantic diversity, where Dominance
reaches \textbf{40.5\%} (see \Figref{fig:semantic-entropy}).\footnote{The aligned model's high lexical dominance arises from its high sampling temperature, which produces long, low-quality sequences that artificially inflate diversity scores while reducing controllability.}
On the NoveltyBench dataset, the gap further widens, with Coverage improving by
\textbf{0.274} and Dominance reaching \textbf{39.9\%}.
These trends are consistent across datasets and extend to the \texttt{Olmo2} model
family; full results are provided in Appendix~\Cref{app:detail_results}.
Qualitative output comparisons are shown in Appendix~\Cref{app:qual_examples}.

\input{past_papers/base_aligned_collaboration/tables/sub_novelty_bench_all}

\paragraph{Validation beyond open-ended evaluation metrics.}
To verify that these gains are not an artifact of open-ended evaluation metrics, we additionally evaluate \name{} on two verifiable benchmarks:
instruction following (IFEval) and mathematical reasoning (GSM8K).
Across both tasks, \name{} achieves higher diversity at matched quality or accuracy than the aligned baseline, confirming that the observed improvements are not only confined to open-ended generation.

\begin{figure}[h]
    \centering
    \includegraphics[width=0.95\linewidth]{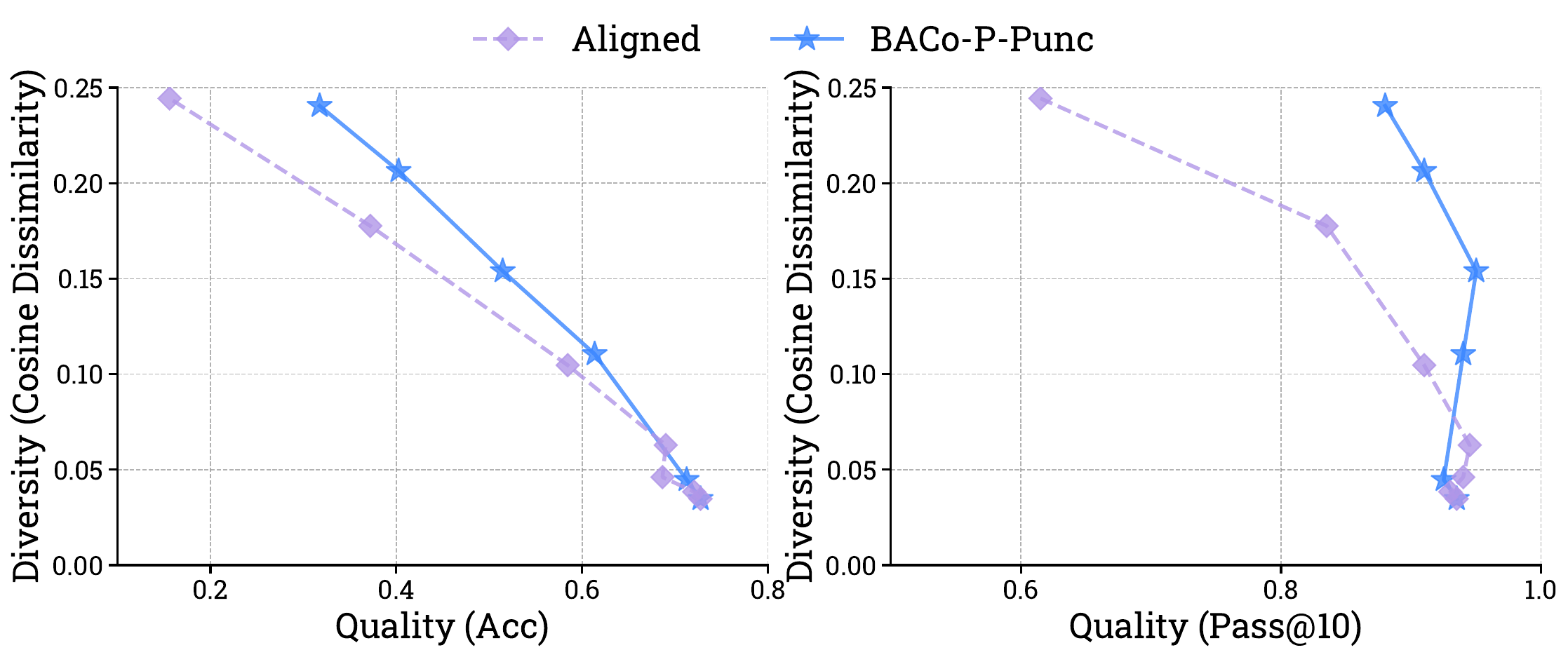}
\caption{Diversity-accuracy trade-off comparison on GSM8K. At comparable accuracy levels,
\name{} achieves higher diversity than the aligned baseline, indicating that the observed gains are not artifacts of open-ended evaluation metrics. 
}
\label{fig:gsm8k_small_figure}
\end{figure}

Notably, the qualitative behavior on GSM8K mirrors that on open-ended tasks:
naive temperature scaling degrades accuracy without delivering meaningful diversity, while base--aligned collaboration enables diversity improvements with accuracy maintained.
Complete results for both verifiable benchmarks are reported in
Appendix~\Cref{app:verifiable_tasks}.

\paragraph{Multi-turn tasks.}
To verify gains beyond single-turn settings, we evaluate on MT-Bench~\citep{zheng2023judgingllmasajudgemtbenchchatbot}.
\name{}-\textsc{P-Punc} substantially outperforms the aligned baseline
(Coverage: 0.681 vs. 0.320; Dominance: 72.8\% vs. 27.2\%) consistently
across diversity and quality metrics; full results in \Cref{app:verifiable_tasks}.

\paragraph{Examination on the distribution of repetition.}
We additionally evaluate \name{} on Artificial Hivemind~\citep{jiang2025artificialhivemindopenendedhomogeneity}, which measures intra-model
repetition on open-ended chat queries. Following its setting, we compare \name{} against Top-$p$ and Min-$p$~\citep{nguyen2025turningheatminpsampling} on \texttt{Llama-3.1-70B-Instruct}. 

\begin{wrapfigure}{r}{0.56\linewidth}
    \vspace{-18pt}
    \centering
    \includegraphics[width=\linewidth]{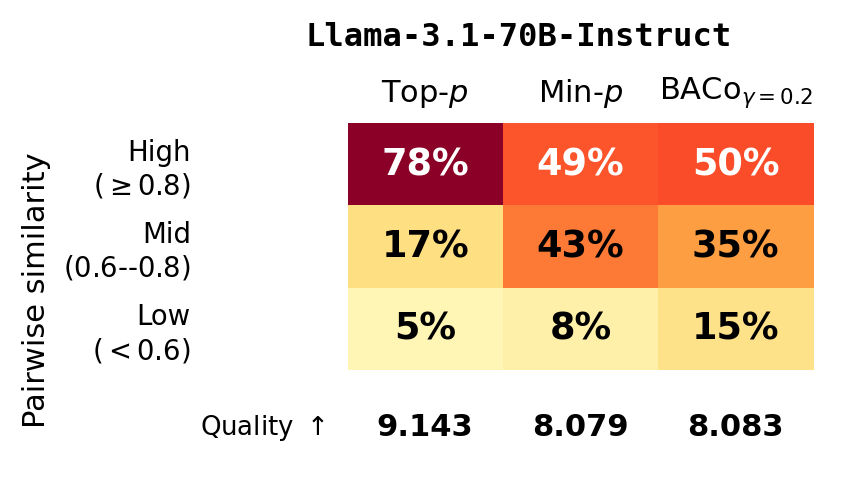}
    \caption{Pairwise intra-model repetition distribution under Artificial Hivemind. \name{}'s generation falls more in lower-similarity bins (bottom rows).}
    \label{fig:hivemind}
\end{wrapfigure}

With \name{}'s threshold selected to match Min-$p$'s quality (avg.\ reward $8.08$; $\gamma{=}0.2$, \textsc{-P-Punc}), its repetition distribution is flatter (\Cref{fig:hivemind}), placing more (15\%) generations in the genuinely diverse regime (similarity $<$ 0.6), vs.\ 8\% for Min-$p$ and 5\% for Top-$p$.
This suggests \name{} pushes diversity beyond the limit reachable by sampling within a single aligned model.

\subsection{Router Performance Comparison}
\label{ssec:router_comparison}
We analyze routers on NoveltyBench, which provides a representative yet computationally efficient setting for controlled comparison.

\paragraph{Sanity-check routers (-\textsc{Rand}, -\textsc{Judge}).}
At first glance, \textsc{-Rand} appears competitive on aggregate, but its gains concentrate almost entirely on lexical metrics (\eg{} Dist-$n$) and collapse on semantic metrics such as Semantic Entropy (\(0\%\) Dominance; Appendix~\Cref{app:detail_results}).
This is consistent with the observation that random or nonsensical text can inflate lexical diversity without producing semantically meaningful variation; unguided switching injects surface-level randomness rather than meaningful semantic diversity.
By contrast, principled routers such as \textsc{-P-Punc} achieve clear Dominance on both lexical and semantic spaces (Table~\ref{tab:noveltybench_all}).
We further consider \textsc{-Judge}, a prompt-based router inspired by multi-agent systems \citep{talebirad2023multi}.
Despite extensive prompt engineering\footnote{Including a step-by-step decision pipeline, curated heuristic rules, and few-shot examples with rationales (prompts are in \Tabref{tab:judge-model-prompt}).}, it consistently underperforms simpler heuristic routers while incurring substantially higher computational cost.

\input{past_papers/base_aligned_collaboration/tables/sub_novelty_ab_vs_aa}

\input{past_papers/base_aligned_collaboration/tables/narrative_and_humaneval}

\paragraph{Simple routers (-\textsc{P}, -\textsc{FC}).}
Single-strategy routers based on uncertainty (\textsc{-P}) or linguistic cues (\textsc{-FC}) trade weaker aggregate performance for clear gains in semantic diversity.
Both outperform \textsc{-Rand} on metrics such as Cosine Dissimilarity and Semantic Entropy, demonstrating that informed routing is critical for meaningful diversity.

\paragraph{Combining strategies yields the strongest performance.}
Combining complementary strategies proves most effective.
In particular, \textsc{-P-Punc} achieves the best overall controllability, with high
Coverage and Dominance across both lexical and semantic spaces.
Other combinations (\textsc{-H-Punc}, \textsc{-P-FC}) capture additional regions of
the Pareto frontier (17.5\% and 17.6\% Dom.), confirming that compositional routing provides complementary
benefits.

Overall, \name{} with most of the routers outperforms the baselines. We therefore emphasize that the contribution of \name{} is the \emph{framework} of token-level base-aligned collaboration, not any single router.
And the framework is robust to router choice without heavy tuning.

\subsection{Base-Aligned vs.\ Aligned-Aligned Collaboration}
\label{ssec:ba_aa}

We compare base-aligned collaboration (\name{}, \texttt{Llama-3 -8B} family) with aligned-aligned collaboration (\texttt{Llama-3-8B} and \texttt{Llama-3.2-11B})
under identical router (\textsc{-P-Punc}).
On NoveltyBench, base-aligned collaboration significantly outperforms the aligned-aligned setup in both Coverage and Dominance.
These results highlight the limited diversity achievable when collaborating between two aligned models and support our hypothesis that complementarity between less-aligned
and more-aligned models are essential for jointly optimizing diversity and quality.

\subsection{Long-Form Diversity Evaluation}
\label{ssec:storyarc}

To assess diversity beyond short-form outputs, we evaluate long-form creative writing
using a discourse-level framework following \citet{tian2024large}.
Rather than relying on surface lexical or single-vector semantic metrics, this
evaluation captures narrative variation through plot structure and affective dynamics (detailed setups are provided in \appref{app:storyarc}).

As shown in \Figref{fig:narrative-dis}, \name{} achieves substantially higher
turning-point and arousal diversity than the aligned baseline at comparable quality,
demonstrating that base--aligned collaboration extends effectively to long-form
generation.
Example outputs are provided in \Tabref{tab:case_storyarc}.

\subsection{Human Evaluation}
\label{ssec:human_eval}

To complement automatic metrics, we conduct a three-phase human evaluation assessing quality, group-level diversity, and creativity.
Four expert annotators evaluate outputs on NoveltyBench and WildChat.

As shown in \Cref{tab:human_eval}, despite near-identical automatic quality
scores (aligned: 5.93; \name{}: 5.85), human raters assign \name{} substantially
higher quality ratings (4.04 vs.\ 2.83 on NoveltyBench; 3.83 vs.\ 3.44 on
WildChat), with strong inter-rater agreement (Pearson $r{=}0.816$,
ICC$(2,k){=}0.907$).
Meanwhile, \name{} achieves significant large diversity win rates across both format and content
sub-dimensions (Fleiss' $\kappa{=}0.268$ overall), confirming that \name{} produces human-perceivable diversity at both levels. Full protocols and annotation examples are provided in Appendix~\Cref{app:human_eval}. 

\paragraph{From diversity-quality trade-off to creativity.}
Building on \citet{jaarsveld2012creative}, creativity requires both
\textit{divergent thinking}, supported by group-level output diversity, and
\textit{convergent thinking}, underpinned by per-output quality.
We preliminarily test whether an improved diversity-quality trade-off translates
into higher human-perceived creativity using \name{}.
As shown in \Cref{tab:human_eval}, \name{}'s outputs are judged most creative in
{79.6\%} of NoveltyBench prompts and {61.8\%} of WildChat prompts
(Fleiss' $\kappa{=}0.485$), suggesting that easing the diversity-quality trade-off yields outputs humans find genuinely more creative.
We leave a larger-scale and deeper investigation of this connection to future work.

\section{Analysis and Discussion}
\label{sec:analysis}

\subsection{Contribution Distribution and Switching Frequency}
\label{sec:anal:distribution}

For \name{} with the best router (\textsc{-P-Punc}), we see a consist pattern that {base-model
contribution and switching frequency are high at the start of generation and
decrease over time} across all three tasks (Figures~A9-A11 in \appref{app:analysis}).
We hypothesize this reflects increasing predictive confidence as context grows,
which naturally aligns with, e.g., in creative writing, early tokens allow more divergence
(\eg{} introducing characters) while later tokens demand coherence. However, the pattern may be
less suitable for list-structured tasks that benefit from uniform exploration level over time.
Dynamic, position-aware thresholds are a promising mitigation.

\subsection{Failure Mode: Inherent Early Stopping}
\label{sec:earlystop}

When the router is tuned more aggressively toward diversity, the system exhibits
a higher tendency to terminate early. 
One model emits an end-of-sequence token
prematurely (examples in \appref{app:analysis}).
This \emph{inherent early stopping} is emergent rather than deliberate, unlike
early-exit mechanisms in reasoning
systems~\citep{ding2025dynamicparalleltreesearch}.
We take a two-sided view of this trend: it can risk truncating valid continuations, but also acts as a safeguard against incoherent repetition or off-topic continuation.
Truncated outputs are easily detectable by length, allowing a simple restart strategy as mitigation with marginal cost overhead.

\subsection{Future Work}
\label{sec:future_work}

We envision \name{} as a preliminary step toward a ``breadth thinking'' mode for
LLMs, complementing deep
thinking~\citep{OpenAI2025IntroducingDeepResearch,DeepSeekAI2025DeepSeekR1IR}, where
a model explores a wide, validated space of perspectives rather than converging
on a single output.
For open-ended tasks, breadth thinking offers a way to help humans break out of
their ``information cocoons''~\citep{piao2023human}, expand ideation space, and
think beyond conventional boundaries.
\name{} is controllable, inference-time only, and deployable without additional
post-training; it can also serve as a plug-in for broader systems, such as an
ideation agent in multi-agent
collaboration~\citep{siddiqui2025script,2025arXiv250614758C,song2025outcomebasedexplorationllmreasoning}.
The controllability of \name{} also opens opportunities for user-facing
interfaces, such as a slider that lets users shift along the diversity-quality
spectrum, offering a more principled way to control output diversity than temperature scaling, which degrades might quality.

%% file: past_papers/base_aligned_collaboration/tables/sub_average.tex
\begin{figure}[htbp]
\centering
\vspace{6mm}
\begin{minipage}[c]{0.5\textwidth}
    \centering
    \small
    \renewcommand{\arraystretch}{1.25}
    \setlength{\tabcolsep}{4pt}
    \resizebox{\linewidth}{!}{%
        \begin{tabular}{lcccccc}
            \toprule
            \textbf{Method} & \multicolumn{2}{c}{\textbf{Lexical}} & \multicolumn{2}{c}{\textbf{Semantic}} & \multicolumn{2}{c}{\textbf{Overall}} \\
             & \textit{Cov.} & \textit{Dom.} & \textit{Cov.} & \textit{Dom.} & \textit{Cov.} & \textit{Dom.} \\
            \midrule
            Base    & 0.098 & 12.7\% & 0.098 & 16.0\% & 0.098 & 14.3\% \\
            Aligned & 0.269 & \textbf{49.0\%} & 0.104 & 29.2\% & 0.186 & \textbf{39.0\%} \\
            Nudging & 0.276 & 9.3\%  & 0.247 & 9.9\%  & 0.261 & 9.6\%  \\
            Decoding        & - &  0.3\% & - &  0.3\% & - &  0.3\% \\
            Prompting (Best)  & - &  2.7\% & - &  2.2\% & - &  2.4\% \\
            Ensemble (Best)   & - &  1.1\% & - &  1.9\% & - &  1.5\% \\
            \name{} (Best) & \textbf{0.445} & 24.9\% & \textbf{0.360} & \textbf{40.5\%} & \textbf{0.403} & {32.7\%} \\
            \bottomrule
        \end{tabular}
    }
    \vspace{6pt}
    \captionof{table}{Averaged performance of all methods across all datasets and diversity–quality spaces.
    \name{} consistently outperforms baselines across all semantic and most lexical spaces, demonstrating stronger controllability and substantially improving the semantic diversity–quality trade-off.
    The overall gains, as driven primarily by improvements in semantic, suggest that \name{} produces more meaningful and content-level diversity, rather than superficial word-level changes, compared to other methods.
    See full results at \appref{app:detail_results}. 
    }
    \label{tab:avg_results}
\end{minipage}%
\hspace{1em}
\begin{minipage}[c]{0.455\textwidth}
    \centering
    \vspace{-25pt}
    \includegraphics[width=\linewidth]{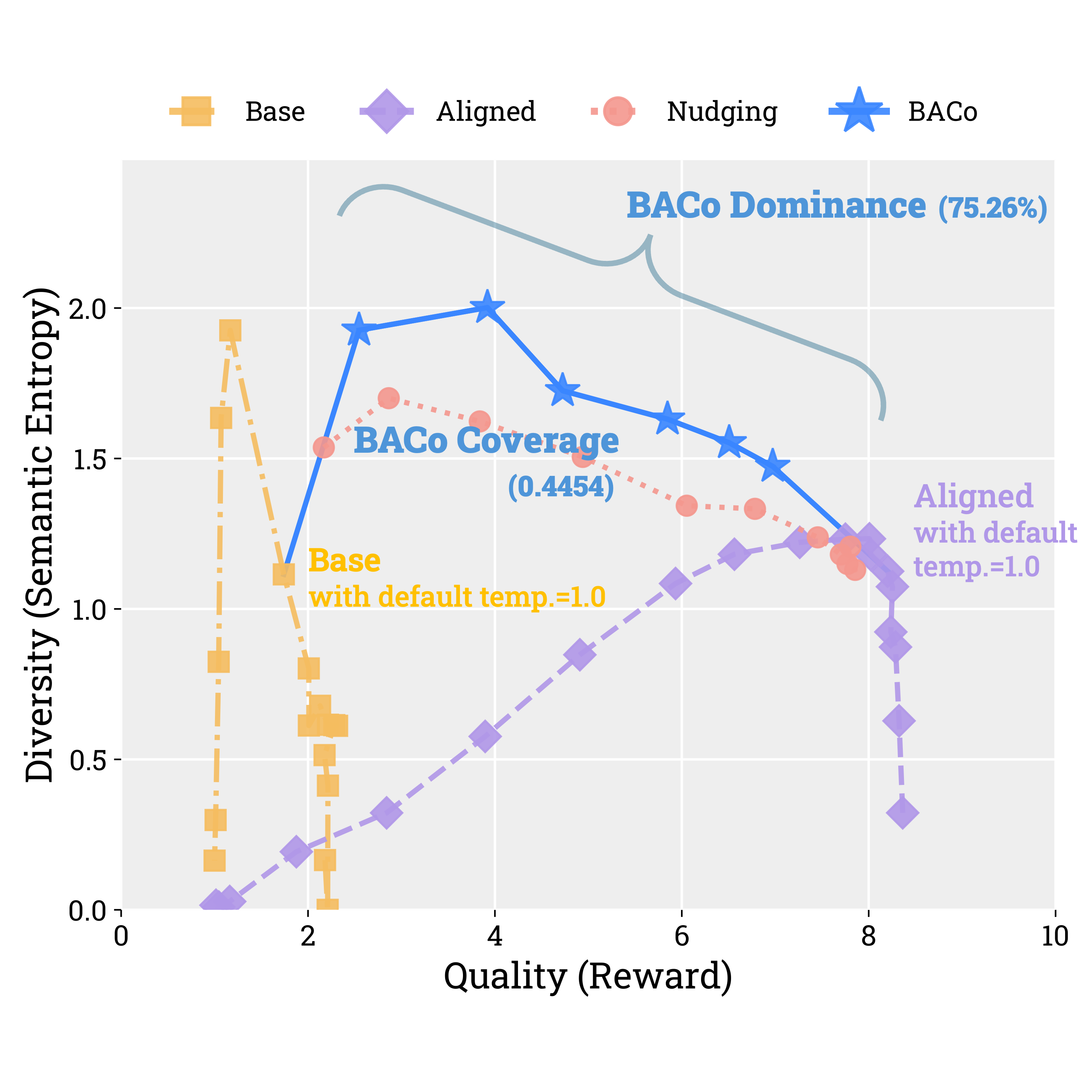}
    \vspace{-35pt}
    \caption{\name{}'s performance on one diversity-quality space (x: quality in terms of reward; y: diversity in terms of semantic entropy). Comparing with baselines, \name{} (blue curve) attains larger Coverage of the top-right region and contributes to most of the Dominance, indicating improvement and controllability on \problem{}.
    }
    \label{fig:semantic-entropy}
\end{minipage}
\end{figure}

%% file: past_papers/base_aligned_collaboration/tables/sub_novelty_bench_all.tex
\begin{table}[t]
\small
\centering
    \renewcommand{\arraystretch}{1.2}
\resizebox{\linewidth}{!}{%
\begin{tabular}{lcccccc}
\toprule
\textbf{Routers} & \multicolumn{2}{c}{\textbf{Lexical}} & \multicolumn{2}{c}{\textbf{Semantic}} & \multicolumn{2}{c}{\textbf{Overall}} \\
 & \textit{Cov.} & \textit{Dom.} & \textit{Cov.} & \textit{Dom.} & \textit{Cov.} & \textit{Dom.} \\
\midrule
-\textsc{Rand} & 0.493 & 26.3\% & 0.409 & 17.0\% & 0.451 & 21.7\% \\
-\textsc{Judge} & 0.302 & 2.6\% & 0.254 & 0.6\% & 0.278 & 1.6\% \\
\midrule
-\textsc{P} & 0.433 & 4.8\% & 0.397 & 8.5\% & 0.415 & 6.7\% \\
-\textsc{FC} & 0.419 & 3.2\% & 0.382 & 4.7\% & 0.401 & 4.0\% \\
\midrule
-\textsc{P-Punc} &\textbf{0.495} & \textbf{30.7\%} &\textbf{0.452} & \textbf{31.3\%} & \textbf{0.474} & \textbf{31.0\%} \\
-\textsc{H-Punc} & 0.466 & 16.4\% & 0.427 & 18.6\% & 0.446 & 17.5\% \\
-\textsc{P-FC} & 0.435 & 16.0\% & 0.406 & 19.2\% & 0.421 & 17.6\% \\
\bottomrule
\end{tabular}
}
\vspace{7pt}
\captionof{table}{Averaged performance of routers within \name{} on NoveltyBench across all diversity–quality spaces.
The \textsc{-P-Punc} router achieves the best overall performance. 
While the random router (\textsc{-Rand}) attains moderately strong results, mainly from increased surface-level lexical diversity, its performance drops sharply on semantic metrics, confirming that unguided switching fails to produce meaningful diversity. 
In contrast, \textsc{-P-Punc} delivers the most balanced and consistent results across both lexical and semantic evaluations, showing combination of designed routing strategies leads to more meaningful diversity.
}
\label{tab:noveltybench_all}
\end{table}

%% file: past_papers/base_aligned_collaboration/tables/sub_novelty_ab_vs_aa.tex
\begin{table}[t]
\centering
\small
\resizebox{\linewidth}{!}{%
\renewcommand{\arraystretch}{1.1}
\begin{tabular}{l cc cc cc }
\toprule
\textbf{Method} & \multicolumn{2}{c}{\textbf{Lexical}}& \multicolumn{2}{c}{\textbf{Semantic}} & \multicolumn{2}{c}{\textbf{Overall}} \\
 & \textit{Cov.} & \textit{Dom.} & \textit{Cov.} & \textit{Dom.} & \textit{Cov.} & \textit{Dom.} \\
\midrule
Base & 0.142 & 9.8\% & 0.142 & 13.2\% & 0.142 & 11.5\%  \\
Aligned & 0.273 & \textbf{43.8\%} & 0.128 & 19.9\% & 0.200 & 31.8\%\\
\midrule
\textsc{AACo} & 0.022 & 4.5\% & 0.006 & 7.0\% &  0.014 & 5.7\% \\
\name{} & \textbf{0.495} & 42.0\% & \textbf{0.452} & \textbf{59.8\%} & \textbf{0.474 }& \textbf{50.9\%} \\
\toprule
\end{tabular}
}
\vspace{4pt}
\caption{Comparison of base-aligned and aligned-aligned collaboration (denoted as ``\textsc{AACo}'') on NoveltyBench across all diversity–quality spaces.
\textsc{AACo} yields little improvement on the diversity–quality trade-off compared with \name{}, particularly in semantic diversity. The results demonstrate the necessity of involving a base and aligned model.
}
\label{tab:novelbench-ab-vs-aa}
\end{table}

%% file: past_papers/base_aligned_collaboration/tables/narrative_and_humaneval.tex
\begin{figure}[htbp] 
\centering
\begin{minipage}[h]{0.48\textwidth}
  \centering
  \includegraphics[width=\linewidth]{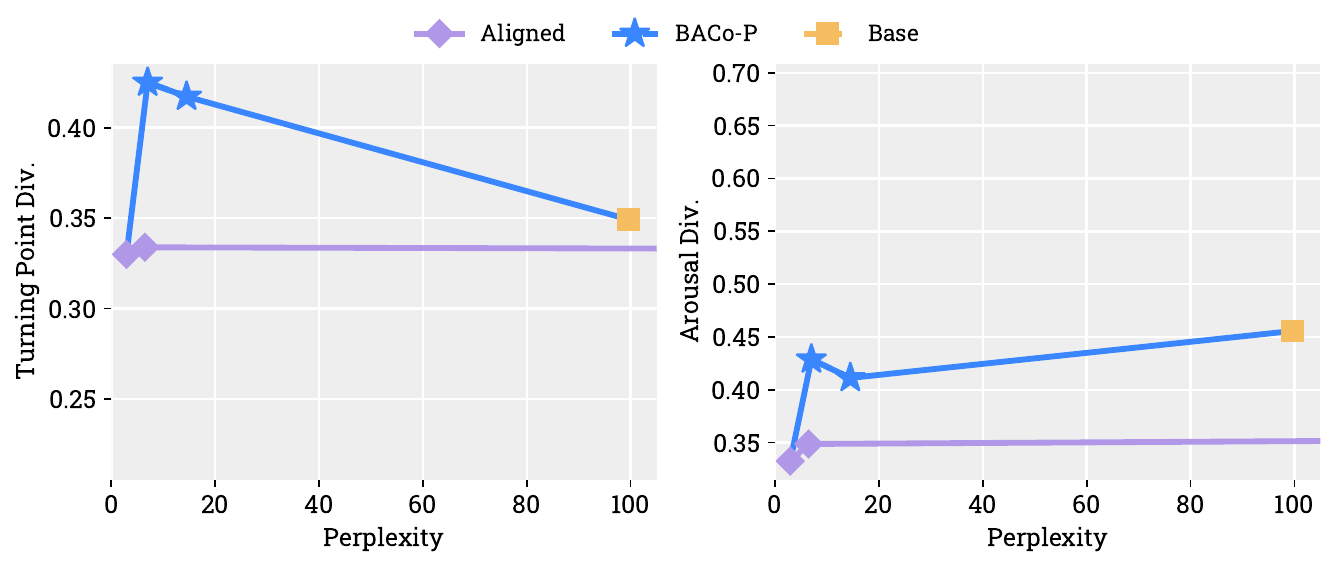}
    \caption{
    Comparison of \name{}'s and baselines' discourse-level diversity-quality trade-off curve on Narrative Discourse. 
    \name{} obtains a larger Coverage, achievable in the high-diversity, high-quality region (top-left). The results demonstrate it has richer discourse-level diversity without sacrificing quality largely, compared with baselines.
    The x-axis is quality (perplexity; lower is better), and the y-axis is discourse-level diversity, either turning-point diversity (left figure) or arousal diversity (right figure) (higher is better). 
    }
  \label{fig:narrative-dis}
\end{minipage}\hfill
\begin{minipage}[t]{0.48\textwidth}
\vspace{-8\baselineskip}
  \centering
  \small
  \renewcommand{\arraystretch}{1.25} 

  \setlength{\tabcolsep}{4pt}          

  \resizebox{0.88\linewidth}{!}{%
  \begin{tabular}{@{}llcc|cc@{}}       
    \toprule
             & \textbf{} & \multicolumn{2}{c|}{\textbf{NoveltyBench}} & \multicolumn{2}{c}{\textbf{WildChat}} \\
    \midrule
    \textbf{\textit{}} & \textbf{} & Aligned & \name{} & Aligned & \name{} \\
    \midrule
    \multicolumn{2}{c}{\textit{Quality}}  & 2.83 & \textbf{4.04} & 3.44 & \textbf{3.83} \\
    \midrule
     & \textit{Overall} & 21.0\% & \textbf{79.0\%} & 36.1\% & \textbf{63.9\%} \\
    \textit{Diversity} & \textit{Format} & 25.4\% & \textbf{74.6\%} & 26.8\% & \textbf{73.2\%} \\
     & \textit{Content} & 22.9\% & \textbf{77.1\%} & 41.6\% & \textbf{58.4\%} \\
    \midrule
    \multicolumn{2}{c}{\textit{Creativity}}  & 20.4\% & \textbf{79.6\%} & 38.2\% & \textbf{61.8\%} \\
    \bottomrule
  \end{tabular}%
  }
\vspace{7pt}
\captionof{table}{Human evaluation comparing \name{} (best variant, \textsc{-P-Punc}) with the aligned model baseline on NoveltyBench and WildChat.
\name{} is consistently preferred by human judges across all aspects, demonstrating that it produces outputs that are not only more human-perceived diverse but also higher in quality and creativity.
Scores of quality are on a 1–5 Likert scale, and others indicate the pairwise win rate. }
  \label{tab:human_eval}
\end{minipage}
\end{figure}

%% file: chapters/annealed_sampling.tex
\section*{Chapter Overview}
Reinforcement learning with verifiable rewards (RLVR) is a powerful paradigm for enhancing the reasoning capabilities of large language models (LLMs), yet its success hinges on effective exploration. In this chapter, we propose Exploratory Annealed Decoding (EAD), grounded in the insight that exploration is most impactful on early tokens which define a sequence's semantic direction. EAD implements an intuitive \textit{explore-at-the-beginning, exploit-at-the-end} strategy by annealing the sampling temperature from high to low during generation.

\graphicspath{{./}}

\input{past_papers/annealed_sampling/tex/introduction}

\input{past_papers/annealed_sampling/tex/prelim}

\input{past_papers/annealed_sampling/tex/sequential_exploration}

\input{past_papers/annealed_sampling/tex/method}

\input{past_papers/annealed_sampling/tex/experiments}

\input{past_papers/annealed_sampling/tex/related_work}

\input{past_papers/annealed_sampling/tex/discussion}

%% file: past_papers/annealed_sampling/tex/introduction.tex
\section{Introduction}
\label{sec:branching_factor_introduction}

Reinforcement learning with verifiable rewards (RLVR) is a powerful approach to enhance the capabilities of Large Language Models (LLMs) in domains such as mathematical reasoning and code generation~\citep{openai2024learning, DeepSeekAI2025DeepSeekR1IR, team2025kimi, yang2025qwen3}. 
In this framework, an LLM learns by iteratively generating potential solutions (i.e., rollouts), and receiving feedback on its attempts. 
A central challenge lies in guiding language models to explore diverse yet high-quality solutions in their vast output space~\citep{cheng2025reasoning};
this reflects the long-standing hard trade-off between exploration and exploitation in RL~\citep{thrun1992efficient, sutton1998reinforcement}.

To achieve effective exploration, the sampling process can be modified to increase the variance of its underlying distribution.
However, any such modification involves two fundamental challenges.
First, it must \textbf{preserve sample quality}. 
Increasing diversity at the cost of generating low-quality, nonsensical outputs is counterproductive.
Second, it must \textbf{ensure training stability}.
Modifying the sampler creates a discrepancy between the behavior policy (used for sampling) and the target policy (being optimized), which necessitates an importance sampling (IS) correction in the gradient update~\citep{degris2012off}.
If the probability ratio in the IS weight is too large, the gradients can have high variance, destabilizing the entire training process~\citep{schulman2017proximal}.
An ideal exploration technique must therefore increase diversity while keeping the sampling distribution close enough to the target policy to allow for stable learning~\citep{haarnoja2018soft, ziegler2019fine}.

A widely adopted and principled way to this trade-off is to adjust the sampling temperature~\citep{ACKLEY1985147, hou2025t1}.
Beyond its implementation simplicity, this method is \emph{variationally optimal}, that is, maximizing entropy (thereby increasing diversity) while bounding the KL divergence from the target policy~\citep{jaynes1957information}.
However, relying on a single fixed temperature creates a tension:
high temperature promotes diversity but produces nonsensical text~\citep{renze-2024-effect, wang2025beyond},
whereas low temperature improves quality but limits exploration, leading to generic and repetitive outputs~\citep{Holtzman2020The, DeepSeekAI2025DeepSeekR1IR}.

In this work, we propose \textbf{\algname (\alg)}, a strategy that improves the balance of this trade-off by leveraging a key insight into sequential generation: exploration is not equally valuable at every step.
The initial tokens shape a sequence's semantic direction and structure, making early exploration crucial for discovering diverse valid solutions. 
Later tokens, however, fill in details within the established context, where excessive exploration can harm coherence.
This insight motivates our core strategy: \textit{explore at the beginning, exploit at the end}. 
This simple principle elegantly addresses the twin challenges of quality and stability. 
Injecting randomness early promotes diverse, high-level exploration, while reducing it later ensures completions are both coherent and close to the target policy---an essential property for stable off-policy learning.

In summary, our contributions are as follows:
\begin{enumerate}[wide, labelwidth=!, labelindent=0pt]
    \item[\circone] We propose \alg, a simple and effective exploration strategy for RLVR that dynamically anneals temperature to encourage meaningful diversity while maintaining high sample quality.
    \item[\circtwo] We show \alg is a plug-and-play enhancement that improves sample efficiency over temperature sampling, delivering robust gains across various RLVR algorithms including GRPO~\citep{shao2024deepseekmath}, DAPO~\citep{yu2025dapo}, and EntropyMech~\citep{cui2025entropy} on both small and larger models.
    \item[\circthree] We show that \alg can be adapted for test-time inference, where a tuned temperature schedule further enhances generation quality.
\end{enumerate}

%% file: past_papers/annealed_sampling/tex/prelim.tex
\section{Preliminary}\label{sec:prelim}
\paragraph{Notations.} 

Let $x$ be a prompt from a dataset $\mathcal{D}$, and let $y = (y_1, \dots, y_{|y|})$ be a generated response sequence, where $y_{<t}$ denotes the prefix $(y_1, \dots, y_{t-1})$. We define an LLM as a policy $\pi_\theta$ parameterized by $\theta$, and denote the reference policy, i.e., the starting point for RL, as $\pi_{\mathrm{ref}}$. The probability of generating $y$ given $x$ is defined autoregressively as $\pi_\theta(y \mid x) = \prod_{t=1}^{|y|} \pi_\theta(y_t \mid [x, y_{<t}])$. A reward model $R(x, y)$ evaluates the quality of a prompt-response pair; for our RLVR experiments, we use the rule-based Math-Verify reward model.\footnote{\url{https://github.com/huggingface/Math-Verify}} Finally, we use $|\cdot|$ to denote the length of a sequence or the cardinality of a set, and use the shorthand $1:n$ for the set $\{1, \dots, n\}$.

\paragraph{Reinforcement Learning with Verifiable Rewards (RLVR).}
The standard objective for Reinforcement fine-tuning of LLMs is to maximize the expected reward over a prompt dataset $\mathcal{D}$:
\begin{equation}
\label{eq:rl-main-objective}
    \max_\theta J(\theta) := \E_{x \sim \mathcal{D}, y \sim \pi_\theta(\cdot\mid x)}\left[ R(x, y) \right].
\end{equation}
However, on complex reasoning tasks, learned reward models $R(x,y)$ are prone to \emph{reward hacking}, where they assign high scores to plausible but incorrect solutions \citep{gao2023scaling, perez2023discovering, weng2024rewardhack, wcq2025beyond}. RLVR addresses this by replacing the learned model with a verifiable, rule-based reward signal—such as a verifier that provides binary feedback on a solution's correctness \citep{DeepSeekAI2025DeepSeekR1IR}. This ensures that the policy is optimized using a reliable signal.

RLVR is commonly implemented using policy gradient algorithms like Proximal Policy Optimization (PPO) \citep{schulman2017proximal}. However, standard PPO often requires complex token-level advantage estimation and a separate value model. To better suit RLVR's trajectory-level binary rewards, subsequent methods simplify the advantage calculation \citep{shao2024deepseekmath,yu2025dapo}. A prominent example is Decoupled Clip and Dynamic Sampling Policy Optimization (DAPO)~\citep{yu2025dapo}, which optimizes:
{
\begin{small}
\begin{align*}
&J_{\text{DAPO}}(\theta)=\\ 
&\E_{x \sim \mathcal{D},y^{(1:G)} \overset{iid}{\sim} \pi_{\theta_{\text{old}}}(\cdot\mid x)} 
\left[\tfrac{1}{\sum_{i=1}^G |y^{(i)}|}\sum_{i=1}^G\sum_{t=1}^{|y^{(i)}|}\min\left\{r^{(i)}_{t}(\theta)A_i,\mathrm{clip}\left(r^{(i)}_{t}(\theta),1-\varepsilon_{\text{low}},1+\varepsilon_{\text{high}}\right)A_i\right\}
\right]\\
&\mathrm{s.t.}\quad 0<\left|\{y^{(i)}:y^{(i)}\mathrm{~is~correct}\}\right|<G,
\end{align*}
\end{small}
}where $\pi_{\theta_{\text{old}}}$ refers to previous policy and $r^{(i)}_{t}(\theta)=\frac{\pi_{\theta}(y^{(i)}_{t} \mid [x, y^{(i)}_{<t}])}{\pi_{\theta_{\text{old}}}(y^{(i)}_{t} \mid [x, y^{(i)}_{<t}])}$.  The asymmetric bound $\varepsilon_{\text{high}} > \varepsilon_{\text{low}}$ is proposed to relax the restriction on probability increase and encourage more exploration in the training. 
The advantage $A_i$ is computed by normalizing the binary rewards across a batch of $G$ responses, thus avoiding the need for a value model:
\begin{small}
\begin{equation*}
A_i=\frac{R_i-\mathrm{mean}_{k\in 1:G}(R_k)}{\mathrm{std}_{k\in 1:G}(R_k)},~
\text{where } R_i=R(x,y^{(i)}).
\end{equation*}
\end{small}

\paragraph{Temperature.}
Temperature sampling \citep{ACKLEY1985147} is a widely used method to control the stochasticity of the policy $\pi_\theta$. At each generation step $t$, the LLM computes a vector of logits, $\mathbf{h}$, over the vocabulary $V$ based on the prompt $x$ and the preceding tokens $y_{<t}$. Temperature sampling rescales these logits with a parameter $\tau > 0$ before applying the softmax function to form the next-token probability distribution:
$$
\pi_\theta(y_t = v \mid [x, y_{<t}]; \tau) = \frac{\exp(h_v / \tau)}{\sum_{v' \in V} \exp(h_{v'} / \tau)},
$$
where $v$ is a token in the vocabulary $V$ and $h_v$ is its corresponding logit. The temperature $\tau$ directly modulates the sharpness of the output distribution. A higher temperature ($\tau > 1$) flattens the distribution, increasing output diversity by making less likely tokens more probable. Conversely, a lower temperature ($\tau < 1$) sharpens it, leading to more deterministic, greedy outputs. 

\paragraph{Pass@$\bm k$.} 
Pass@$k$ measures the probability that at least one of $k$ independent outputs from a language model is correct. 
Let $p_x$ denote the underlying accuracy of the language model $\pi$ given one prompt $x$. The pass@$k$ accuracy is defined as
\[
\E_{x\sim\mathcal{D}}\left[1-(1-p_x)^k\right].
\]
The inner term $1-(1-p_x)^k$ quickly approaches one unless $p_x$ is near zero. 
For example, when $k=64$, any $p_x \geq 0.0695$ already yields a probability of at least $0.99$. 
A large gap between pass@1 and pass@$k$ suggests that for some prompts, $p_x$ is essentially zero.
High diversity may benefit this metric because a model with diverse outputs is more likely to maintain non-negligible $p_x$ values across prompts, leading to stronger pass@$k$ performance.

\paragraph{Entropy.} 
Given a prompt $x$ and a prefix $y_{<t}$, the \textbf{token-level entropy} at step $t$ is defined over the policy's conditional distribution:
$$
H(Y_t \mid [x, y_{<t}]; \theta) := -\sum_{v \in V} \pi_\theta(y_t = v \mid [x, y_{<t}]) \log \pi_\theta(y_t = v \mid [x, y_{<t}]),
$$
where the sum is over all tokens $v$ in the vocabulary $V$.

The \textbf{average entropy} of the policy, $H(\pi_\theta)$, is the expected token-level entropy over all prompts and generation steps. We can estimate this value empirically using Monte Carlo sampling. For each prompt $x \in \mathcal{D}$, we generate $G$ i.i.d. responses $y^{(1)}, \dots, y^{(G)}$. The average entropy is then approximated by averaging the token-level entropies across all generated tokens:
$$
\bar{H}(\pi_\theta) \approx \frac{1}{|\mathcal{D}|} \sum_{x \in \mathcal{D}} \left( \frac{\sum_{i=1}^G \sum_{t=1}^{|y^{(i)}|} H(Y_t \mid [x, y^{(i)}_{<t}]; \theta)}{\sum_{i=1}^G |y^{(i)}|} \right).
$$

%% file: past_papers/annealed_sampling/tex/sequential_exploration.tex
\section{Sequential Exploration: Explore Early, Exploit Late}
\label{sec:exploitation-vs-exploration}
Exploration is a cornerstone of reinforcement learning, enabling agents to discover high-quality policies rather than settling on suboptimal solutions~\citep{sutton1998reinforcement, ladosz2022exploration}. 
This principle becomes particularly vital in deep RL, where vast action spaces render exhaustive search infeasible.
In the context of RL for language models (e.g., RLVR), insufficient exploration often manifests as  \textit{entropy collapse}, i.e., a premature narrowing of the generation distribution during training~\citep{yu2025dapo, wang2025beyond, cui2025entropy}.
A common simple tool to encourage exploration is \emph{temperature sampling}.
However, a \textit{fixed} temperature imposes a difficult trade-off. 
A high temperature promotes diversity (as indicated by increased entropy\footnote{See \cref{app:entropy_proof} for the proof.}), but it risks degrading output quality with nonsensical tokens and hallucinations~\citep{renze-2024-effect, wang2025beyond}. 
In contrast, a low temperature limits the discovery of novel solutions, leading to generic and repetitive outputs~\citep{Holtzman2020The, DeepSeekAI2025DeepSeekR1IR}.

\begin{wrapfigure}{r}{0.3\textwidth}
\vspace{-0.25in}
    \centering
    \includegraphics[width=\linewidth]{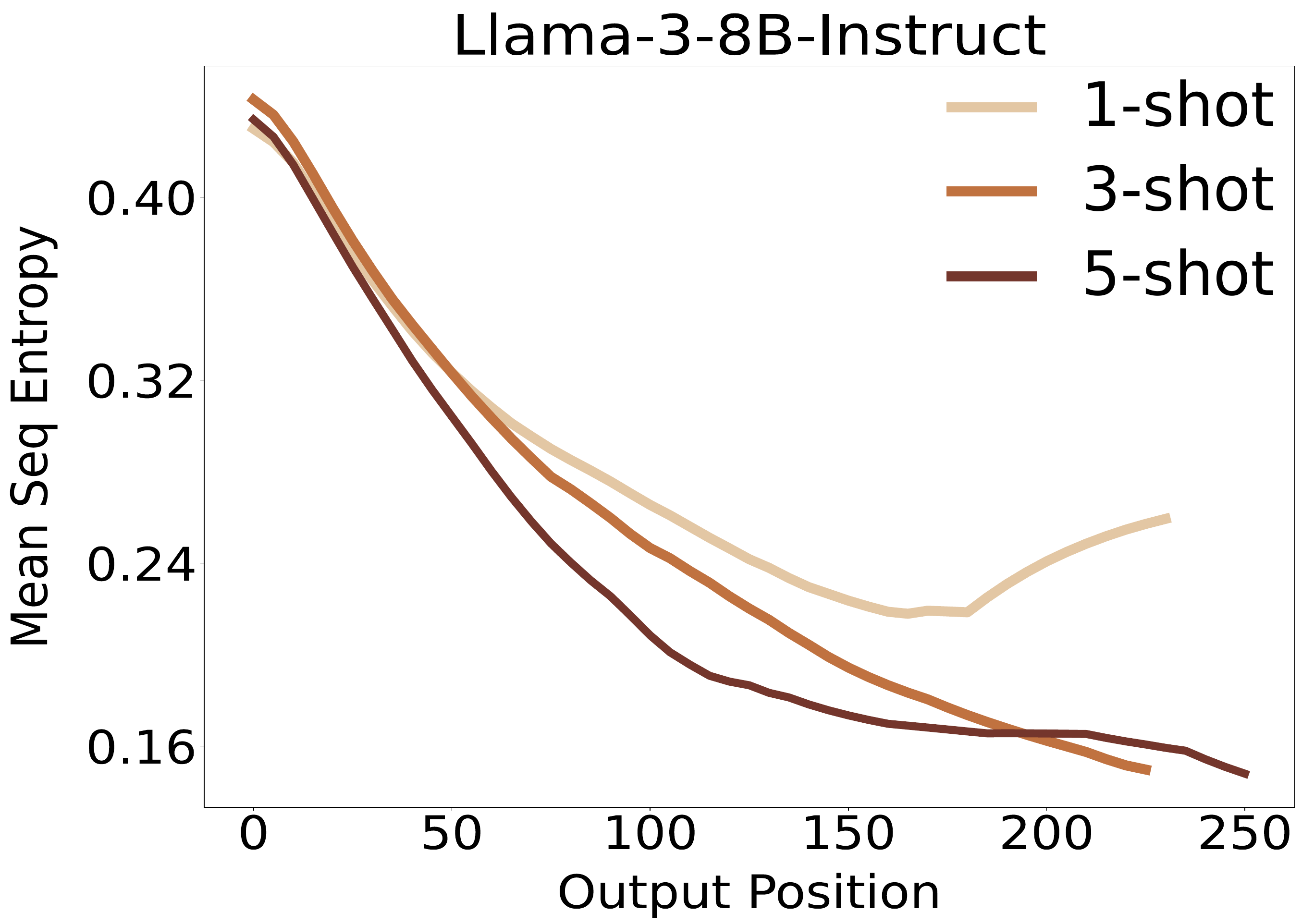}
    \vspace{-0.3in}
    \caption{
    Average entropy shrinks with output positions for Llama-3-8B-Instruct on MMLU dataset. 
    }
    \label{fig: entropy_sequential_dynamic}
\end{wrapfigure}
The key to resolving this dilemma lies not in finding a single best temperature, but in recognizing that exploration requirements vary throughout the generation process.
This key insight stems directly from the autoregressive nature of language models.
At the beginning of a sequence, the context is minimal and uncertainty is high, allowing a wide range of valid continuations. 
As more tokens are produced, the context becomes increasingly specific, constraining subsequent choices.
\revise{We reference the Data Processing Inequality (DPI)~\citep{shannon1948mathematical} as a \textit{theoretical motivation} to investigate these entropy dynamics. While not a strict derivation for autoregressive models, the DPI provides a framework to hypothesize that expected conditional entropy tends to decrease as the context grows}\footnote{While specific rollouts may have late high-entropy positions, the probability of this is exponentially small with position $t$~\citep{yang2026alignment}, making the overall trend a reliable heuristic.}:
\begin{small}
\begin{align}
\label{eq:entropy_decay}
H(Y_t | [x, Y_{<t}]; \theta) &= \underbrace{\mathbb{E}_{y_{<t}}\big[ H(Y_t | [x, y_{<t}]; \theta) \big]}_{\substack{\text{Expected entropy at step } t \\ \text{(average over all prefixes } y_{<t}\text{)}}} 
\geq 
\underbrace{\mathbb{E}_{y_{<t+1}}\big[ H(Y_{t+1} | [x, y_{<t+1}]; \theta) \big]}_{\substack{\text{Expected entropy at step } t+1 \\ \text{(average over all prefixes } y_{<t+1}\text{)}}} = H(Y_{t+1} | [x, Y_{<t+1}]; \theta)\nonumber
\end{align}
\end{small}We further validate this empirically by examining position-wise entropy trend on the MMLU dataset~\citep{hendrycks2021measuring}\footnote{We use MMLU as a held-out dataset with Chain-of-Thought prompting~\citep{wei2022chain} to incentivize longer reasoning outputs, aligning with a typical RLVR scenario.} with Llama-3-8B-Instruct~\citep{dubey2024llama} (see \cref{fig: entropy_sequential_dynamic}).

\begin{figure}
    \centering
    \includegraphics[width=0.8\linewidth]{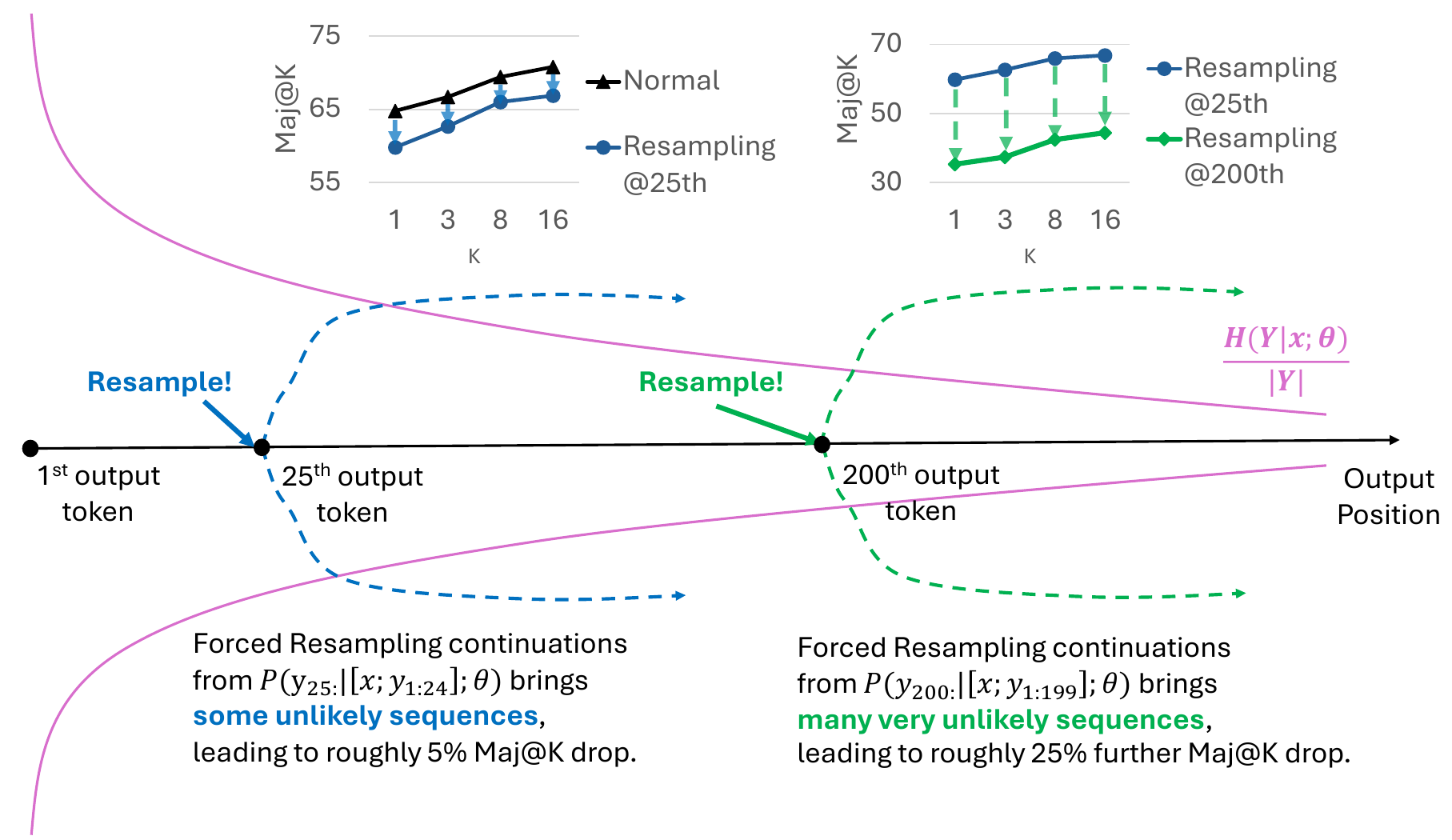}
    \caption{
       A ``forking'' experiment on DeepSeek-Llama-3 shows early branching (high-entropy region) yields higher Maj@$k$ on MMLU than late branching (low-entropy region).
    }
\label{fig:forking_experiment}
\end{figure}
This entropy decay has a direct performance implication: effective exploration, producing diverse, high-quality responses, is most beneficial in an uncertain, high-entropy phase at the start of generation.
To verify this, we replicate a controlled ``forking'' experiment~\citep{yang2026alignment} on DeepSeek-Distilled Llama-3-8B~\citep{DeepSeekAI2025DeepSeekR1IR}, a representative RLVR-fine-tuned model from the same Llama-3 family. 
As shown in \cref{fig:forking_experiment}, trajectories branched from early generation steps consistently outperform those branched later. 
This finding is also consistent with observations from inference-time analysis, where forced, late-stage exploration tends to degrade output quality~\citep{liao2025lost, yang2026alignment, fu2025deep}.

Aligning our strategy with the natural dynamics of generation, we arrive at a simple yet powerful design principle: \textbf{explore early and exploit late}.

%% file: past_papers/annealed_sampling/tex/method.tex
\section{A Method for Sequential Exploration}
\label{sec:annealed_sampling_method}
To put the principle of ``explore early, exploit late'' into practice, we introduce \emph{\algname (\alg)}, 
which uses an annealed temperature schedule starting from a higher-than-standard initial temperature (i.e., $\tau>1$). 
To adapt this strategy to RLVR, we further incorporate a \emph{global-step-aware decay rate}, ensuring that the temperature schedule remains effective as the typical response length increases during training.

\paragraph{Exploratory Annealed Decoding.} 
Instead of a fixed temperature, our method dynamically adjusts the temperature $\tau_t$ for each token $t$ in a rollout. The schedule starts at a high temperature $\tau_\mathrm{max} > 1$ and decreases progressively throughout the generation process.
Specifically, we sample the $t$-th token for one rollout 
with the token-level temperature $\tau_t = \max\{1 + \tau_\mathrm{max} - e^{t/d}, \tau_\mathrm{min}\}$,
where we apply the annealed schedule with a \emph{decay rate} $d$ controlling the annealing speed.
\begin{wrapfigure}{r}{0.6\textwidth}
\vspace{-0.2in}
    \centering
    \includegraphics[width=\linewidth]{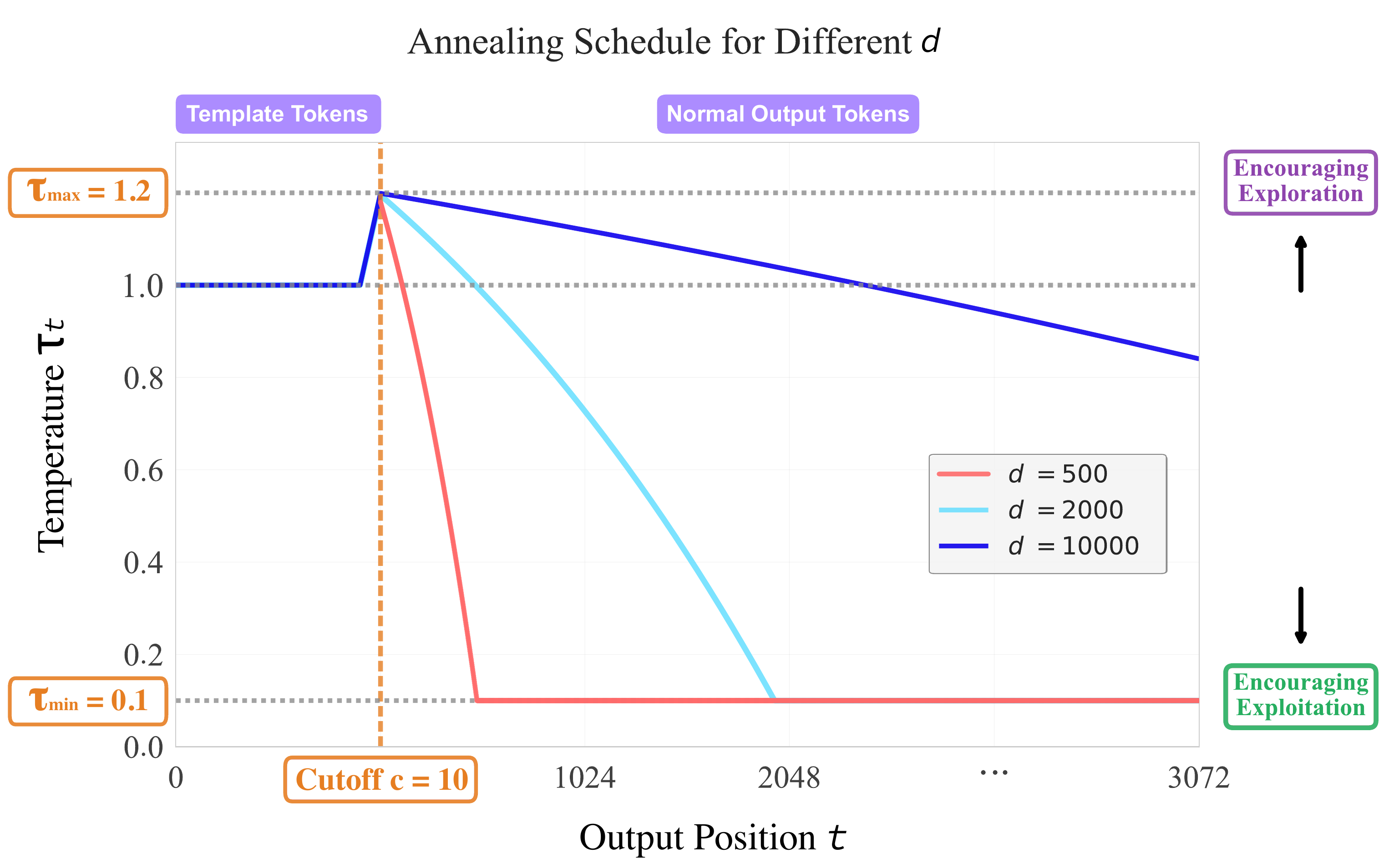}
    \vspace{-0.25in}
    \caption{
     The annealing schedule with different decay rates $d$. A larger $d$ slows the cooling, front-loading exploration over more tokens. We set $c=10, \tau_{\mathrm{max}}=1.2, \tau_{\mathrm{min}}=0.1$ for illustration.
    }
    \label{fig: annealing_schedule_demo}
\vspace{-0.2in}
\end{wrapfigure}
As illustrated in \cref{fig: annealing_schedule_demo}, the decay rate $d$ controls how long the policy remains in a high-exploration state. A larger $d$ front-loads exploration across more initial tokens, while a smaller $d$ transitions to exploitation more quickly. 
In practice, we let $\tau_t = 1.0$ for $t<c$, where $c$ is a pre-determined initial position for the sake of model-specific or prompt-specific template tokens injected in the training process. 
During RLVR, language models tend to generate some template tokens such as ``\textit{let's verify step by steps}'' or repeat the question. 
We fix the temperature at $\tau = 1.0$ in this part to avoid interfering with the generation process.

\paragraph{Global-Step-Aware Decay Rate.}
As training progresses and response lengths increase,\footnote{We illustrate increased length for \alg in \cref{app: increased_length}. } the decay rate $d$ should be adjusted in accordance with the training step.
Otherwise, an excessive number of tokens may be generated under extremely low temperatures, which degrades response quality and leads to undesirable behaviors such as repetition~\citep{DeepSeekAI2025DeepSeekR1IR}, off-topic drift~\citep{spataru2024know}, and unnecessary verbosity~\citep{Holtzman2020The}. 
In particular, we adopt the following \emph{global-step-aware decay rate}:
$d_s = \min(d_0 + \alpha \times s, d_{\max})$, where $\alpha > 0$ is the growth factor and $d_{\max}$ is the decay cap. 

\paragraph{Ensuring Stability with Truncated  Importance Sampling.}
With aggressive annealing schedules (e.g., very small $\tau_\mathrm{min}$ and $d$), sampling low-probability, long-tail tokens can cause the annealed policy to deviate significantly from the one being optimized. This creates an off-policy discrepancy that risks training instability.  To mitigate this, we employ \emph{truncated importance sampling} (TIS)~\citep{heckman1998matching, hilton2022batch, yao2025offpolicy} to correct the objective, ensuring stable optimization even under highly exploratory schedules (see \Cref{sec:off-policy} for details).

Overall, this annealed decoding strategy offers a compelling combination of effectiveness and efficiency. As a plug-and-play modification to standard temperature sampling, it incurs negligible computational overhead and is fully compatible with existing RLVR pipelines and diverse policy optimization algorithms like DAPO, GRPO, etc.

%% file: past_papers/annealed_sampling/tex/experiments.tex
\section{Experiments}\label{sec:experiments}

\subsection{Experimental Setup}
\label{sec: experiment_setup}
\paragraph{Models, Data, and Training Frameworks.}
To ensure a rigorous and controlled comparison, we follow the Minimal-RL recipe~\citep{xiong2025minimalist},\footnote{More training details and hyperparameter setups are illustrated in \cref{app: minimal_rl_training_details}. } training all models on the Numina-Math dataset~\citep{numina_math_7b}, which contains 860k math prompts. To assess the generality of our method, we experiment with both Qwen-2.5-Math-1.5B~\citep{yang2024qwen2} and Llama-3.2-1B-Instruct~\citep{dubey2024llama}.\footnote{We also experimented with the Llama-3.2-1B base model. However, consistent with \citet{wang2025octothinker}, we found that applying RL to base models without intermediate domain-specific fine-tuning yields limited gains across all methods. We defer a deeper investigation to future work.} We also include the larger Qwen-2.5-Math-7B model to evaluate how our approach scales. While our primary experiments are conducted within the DAPO framework~\citep{yu2025dapo}, we demonstrate broader applicability by additionally integrating \alg with GRPO~\citep{shao2024deepseekmath} and EntropyMech~\citep{cui2025entropy}.

\paragraph{Baselines and Controlled Comparison.}
We evaluate \alg against fixed-temperature sampling, a standard and strong baseline, using temperatures $\tau \in \{0.6, 1.0, 1.2\}$ as recommended by prior work~\citep{renze-2024-effect, DeepSeekAI2025DeepSeekR1IR, hou2025t1}. For a fair comparison focused specifically on the sampling strategy, we disable two orthogonal techniques for all methods: (1) dynamic data sampling~\citep{yu2025dapo}, to maintain a consistent training set for all runs, and (2) rollout length penalties, to avoid confounding the reward signal with length-based biases.

\paragraph{Hyperparameters.}
Unless otherwise stated, we use a default configuration of $\tau_{\max}=1.2$, $d_0=25$, $\alpha = 5$, and $d_{\max}=40000$ for \alg. 
For $\tau_{\min}$, we observed optimal values varied by model capability.
For the 1B and 1.5B models, we set $\tau_{\min}=0.1$. 
\revise{For the more capable 7B model, 
we 
used a higher value of $\tau_{\min}=0.8$. }
All hyperparameters are tuned based on a prior study over held-out datasets. 

\subsection{\alg Improves RLVR Training}

\paragraph{\alg Improves RL Exploration and Training Efficiency.}
\begin{figure}[t!]
\centering
\includegraphics[width=\linewidth]{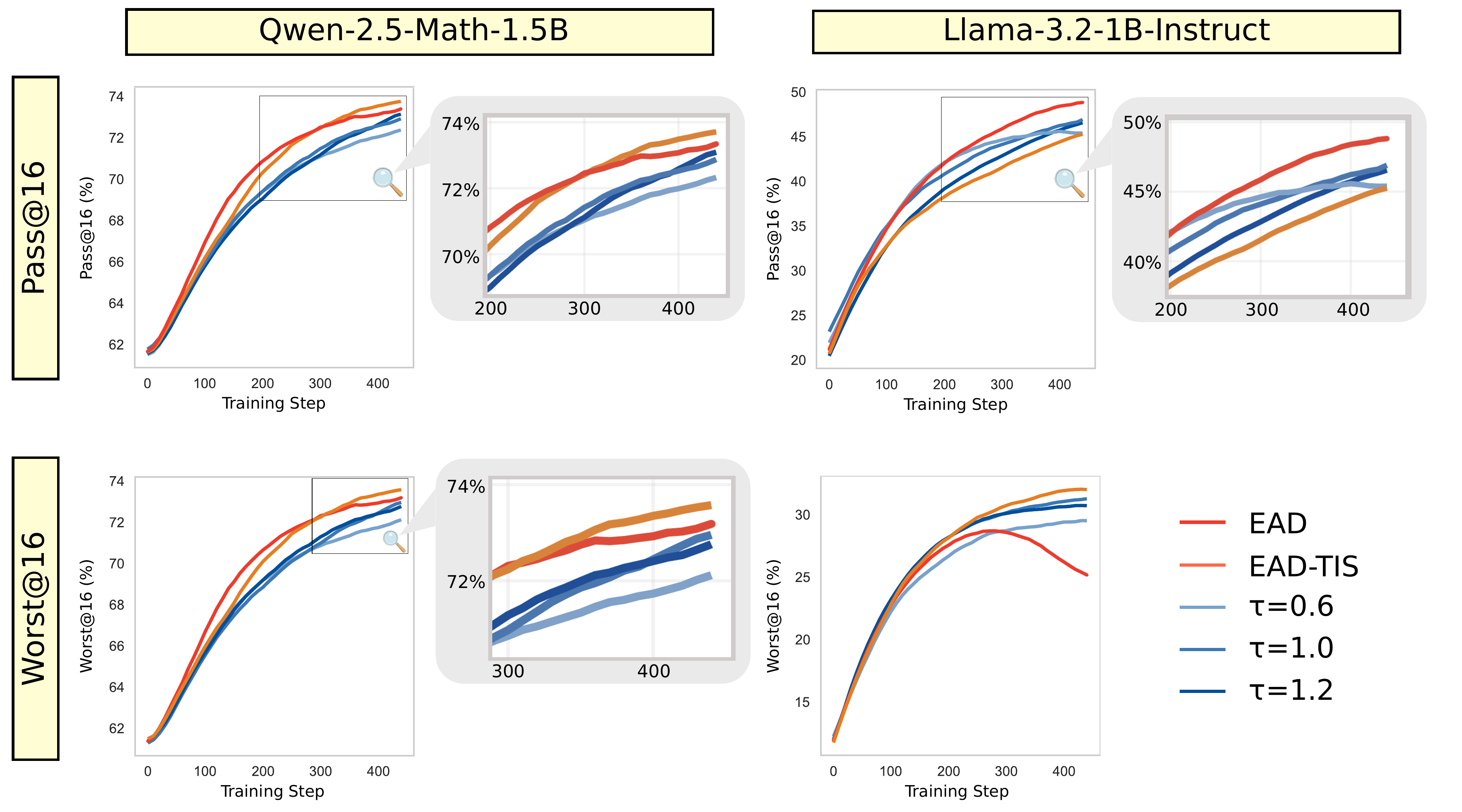}
\caption{
Pass@16 and Worst@16 performance evaluation in RL training. 
While \alg improves exploration of high-quality samples (even the worst outperform temperature sampling), the gain diminishes over time; importance sampling can supplement to correct bias and sustain training.
}
\label{fig: main_bench_best_worst_at_16}
\end{figure}
\begin{figure}
    \centering
    \includegraphics[width=0.8\linewidth]{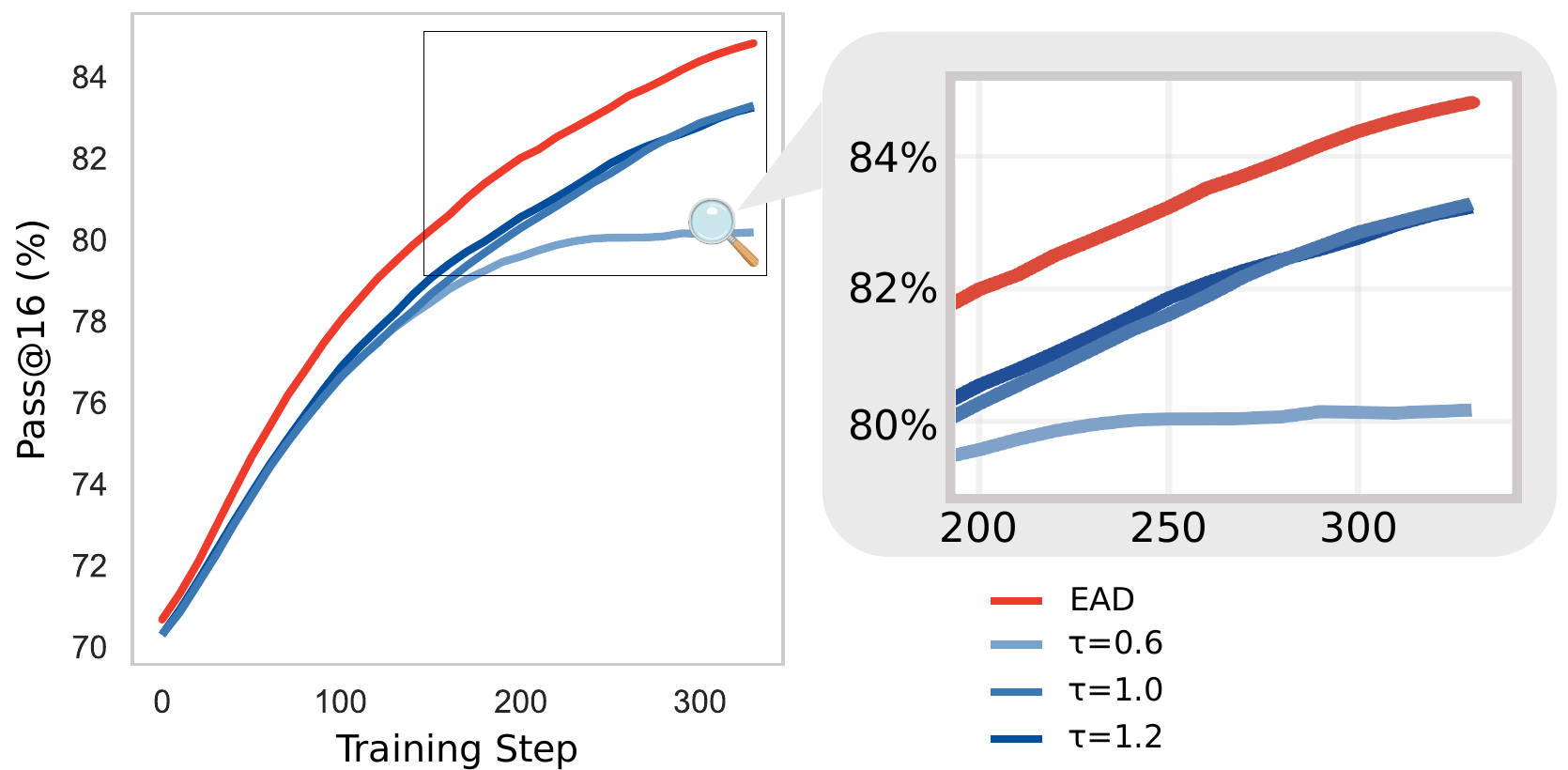}
    \caption{
    Pass@16 performance on Qwen-2.5-Math-7B. 
    \alg enables better exploration than fixed-temperature sampling, yielding sustained gains in Pass@16 throughout training.
    }
    \label{fig: best_at_16_qwen2.5_7b}
\end{figure}
As shown in \cref{fig: main_bench_best_worst_at_16}, \alg significantly improves training efficiency. 
For Pass@16 accuracy, 
\alg (w/o TIS) consistently outperforms the baselines on the Llama and Qwen models (\alg (w/ TIS) also outperforms on the Llama model), demonstrating more effective exploration. 
Under the stricter Worst@16 metric, the inclusion of TIS becomes essential for maintaining stable performance gains, highlighting its importance for correcting the off-policy training dynamic introduced by \alg. 
Through bootstrapping evaluation as in~\citet{hochlehnert2025sober}, the standard deviation of both Pass@16 performance and worst@16 are way below 0.01 and thus all comparisons here are significant.

To verify that our method generalizes, we evaluated it on the larger Qwen-2.5-Math-7B model. The results, presented in \cref{fig: best_at_16_qwen2.5_7b}, confirm that the performance gains from \alg remain significant. This demonstrates that our approach is effective not only on smaller models but also scales successfully.

\begin{figure}[t]
    \centering
    \vspace{-0.05in}
    \includegraphics[width=.9\linewidth]{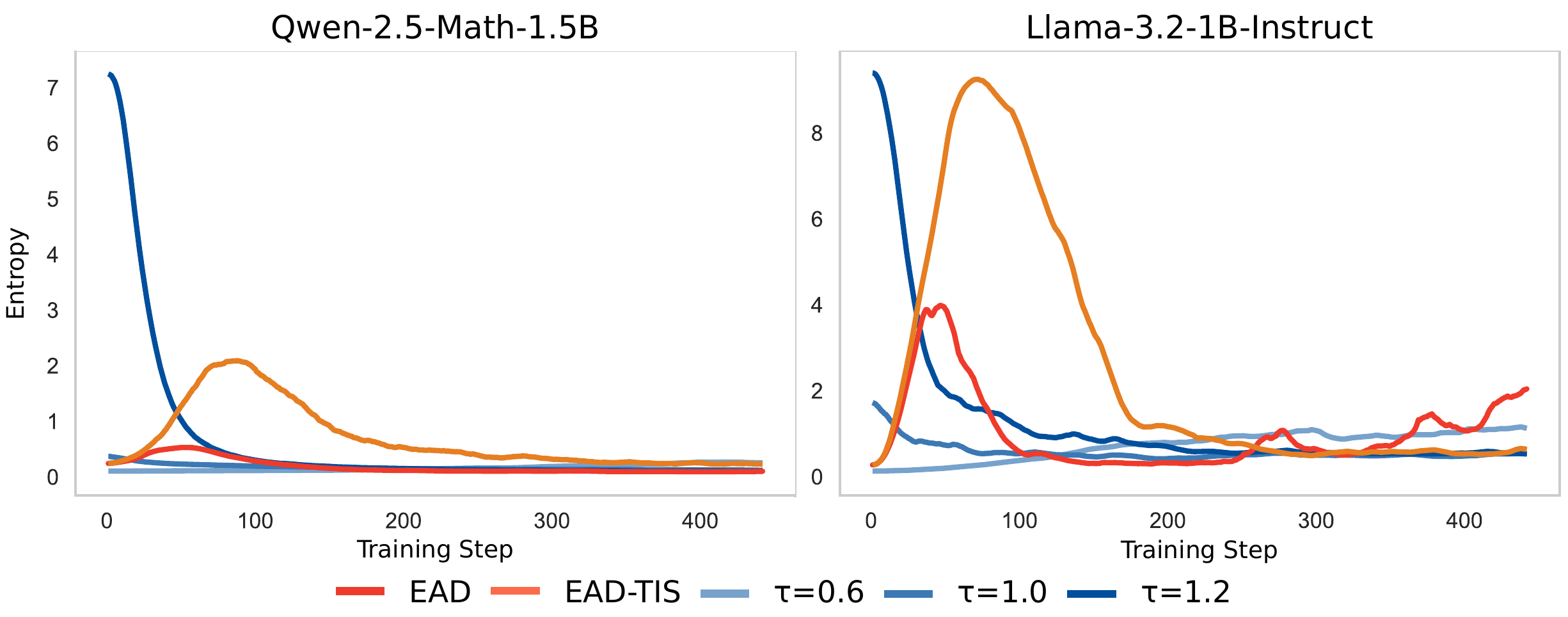}
    \vspace{-0.1in}
    \caption{Entropy Dynamics in RL Training. Under commonly-used temperature sampling, trained with RL algorithm would make entropy decrease, sharply shrinking the exploration space for RL from beginning. While EAD could help RL algorithm to escape local minimum and do exploration when needed in the middle of RL training. 
    }
\label{fig: main_bench_entropy}
\vspace{-0.25in}
\end{figure}

\paragraph{\alg Mitigates Entropy Collapse.}
One major problem in RLVR training is entropy collapse~\citep{cui2025entropy}, which causes the exploration space to shrink and constrains improvement during the ``plateau stage"~\citep{deng2025trial}.
We plot the entropy dynamic in \cref{fig: main_bench_entropy}, where we can see that the entropy dynamic for \alg-empowered methods is not monotonically decreasing from the beginning. 
Instead, it tries to gradually transition out from local optimum~\citep{kirkpatrick1983optimization, bertsimas1993simulated} in a natural, continuous way without any external intervention, such as introducing tree search in rollout sampling~\citep{li2025treepo}.
\begin{figure}
\begin{subfigure}{\linewidth}
    \centering
    \includegraphics[width=0.8\linewidth]{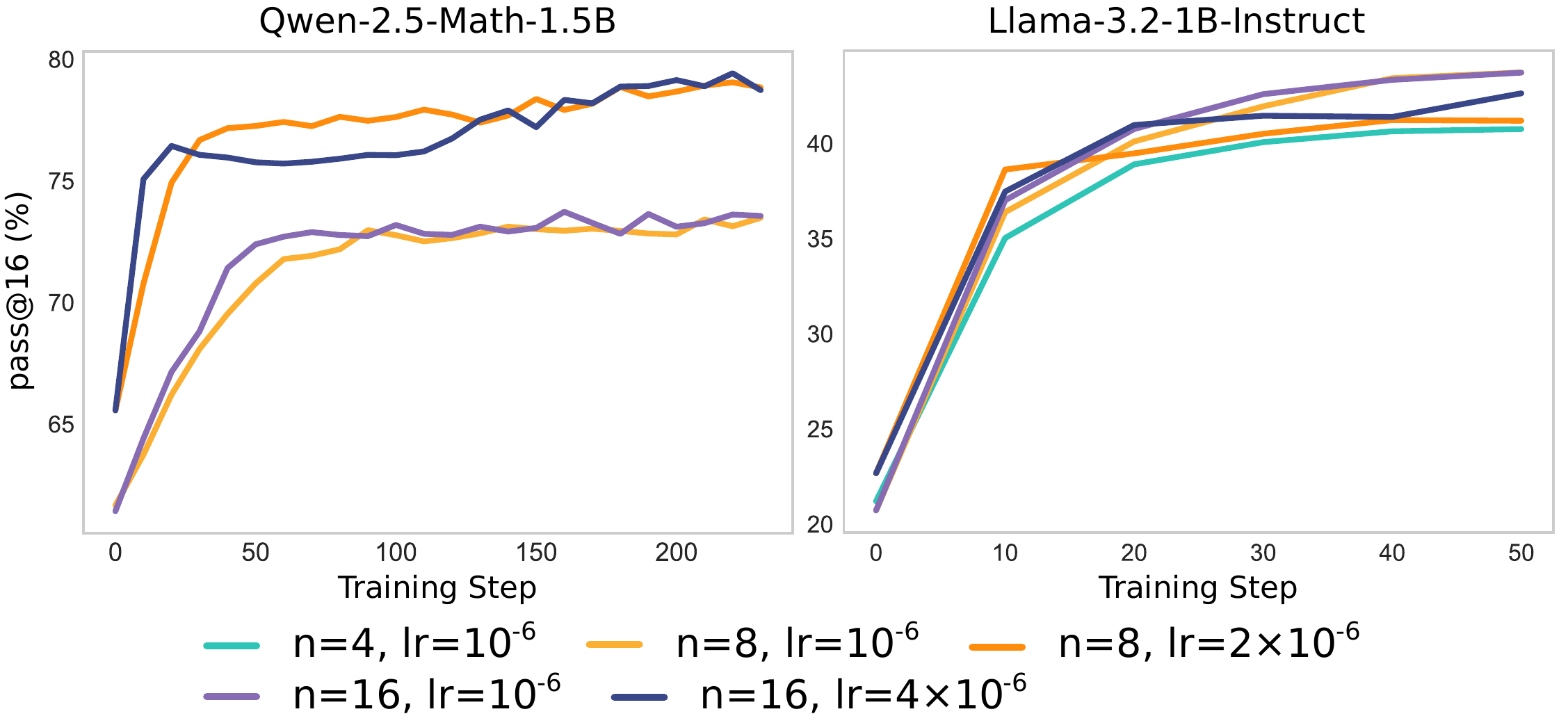}
    \caption{\alg would bring further performance improvement via increased numbers of rollouts, but the commonly used $4$ or $8$ is already good enough. }
    \label{fig: main_bench_rollout_scaling}
\end{subfigure}
\begin{subfigure}{\linewidth}
    \centering
    \includegraphics[width=0.8\linewidth]{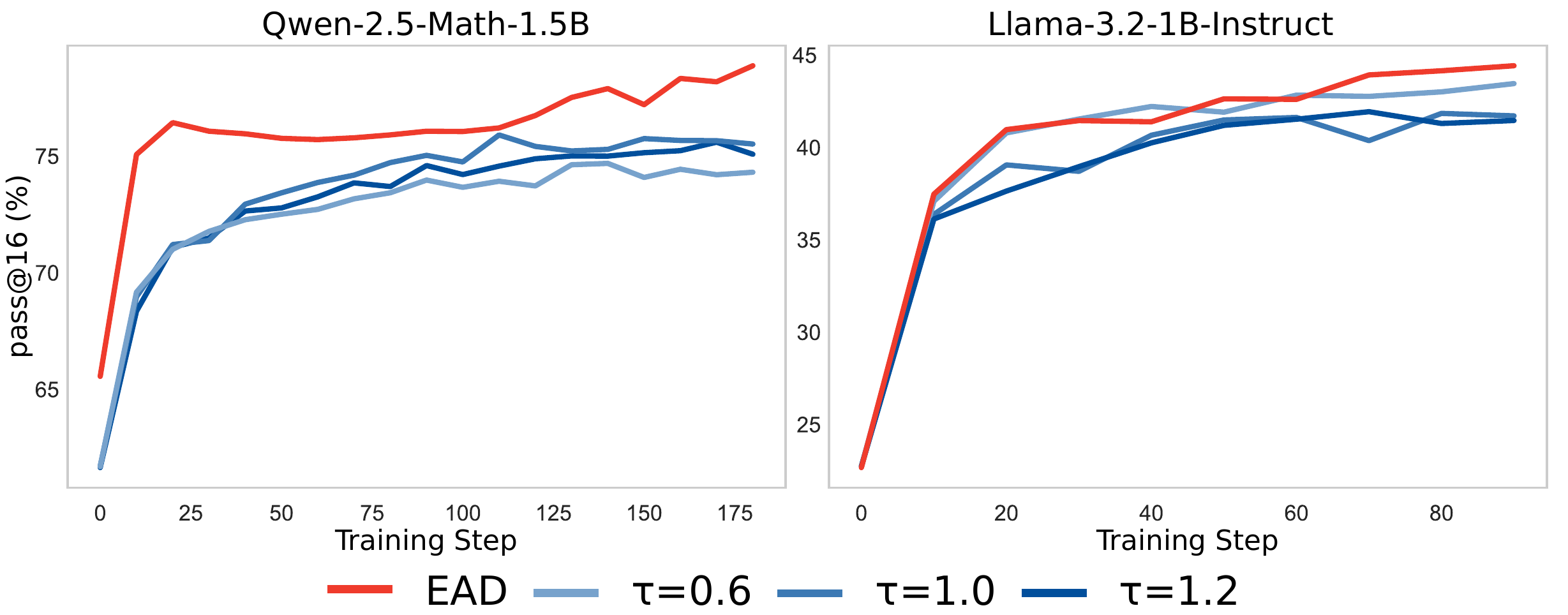}
    \caption{When scaling out the rollout number to 16, the relative advantages of our methods diminished; however, it still outperforms traditional same-temperature sampling.}
    \label{fig: rollout_modelwise_compare}
\end{subfigure}
\caption{Sample Efficiency of EAD. }
\end{figure}

\paragraph{Sample efficiency of \alg.}
Increasing the number of rollouts is a common but computationally expensive strategy to enhance exploration~\citep{hou2025t1}. 
We test the sample efficiency of \alg by varying the number of rollouts, adjusting the learning rate accordingly as suggested by prior work~\citep{chen2025pass}. 
As shown in \cref{fig: main_bench_rollout_scaling}, while more rollouts can further improve performance, \alg achieves strong results with just 4 or 8 rollouts. 
For instance, the optimal configurations are $n=8$ with a learning rate of $10^{-6}$ for Llama-3.2-1B-Instruct and $2\times10^{-6}$ for Qwen-2.5-Math-1.5B. 
This highlights the sample efficiency of our approach, offering a way to reduce the computational cost of the rollout phase.

To assess whether \alg's advantage persists with extensive exploration, we compare it against baselines using a larger set of 16 rollouts. 
\cref{fig: rollout_modelwise_compare} shows that although the relative performance gain diminishes, \alg still outperforms fixed-temperature baselines by a clear margin.

\subsection{\alg is Compatible with Various RL Algorithms}
\begin{figure}
    \centering
    \includegraphics[width=0.8\linewidth]{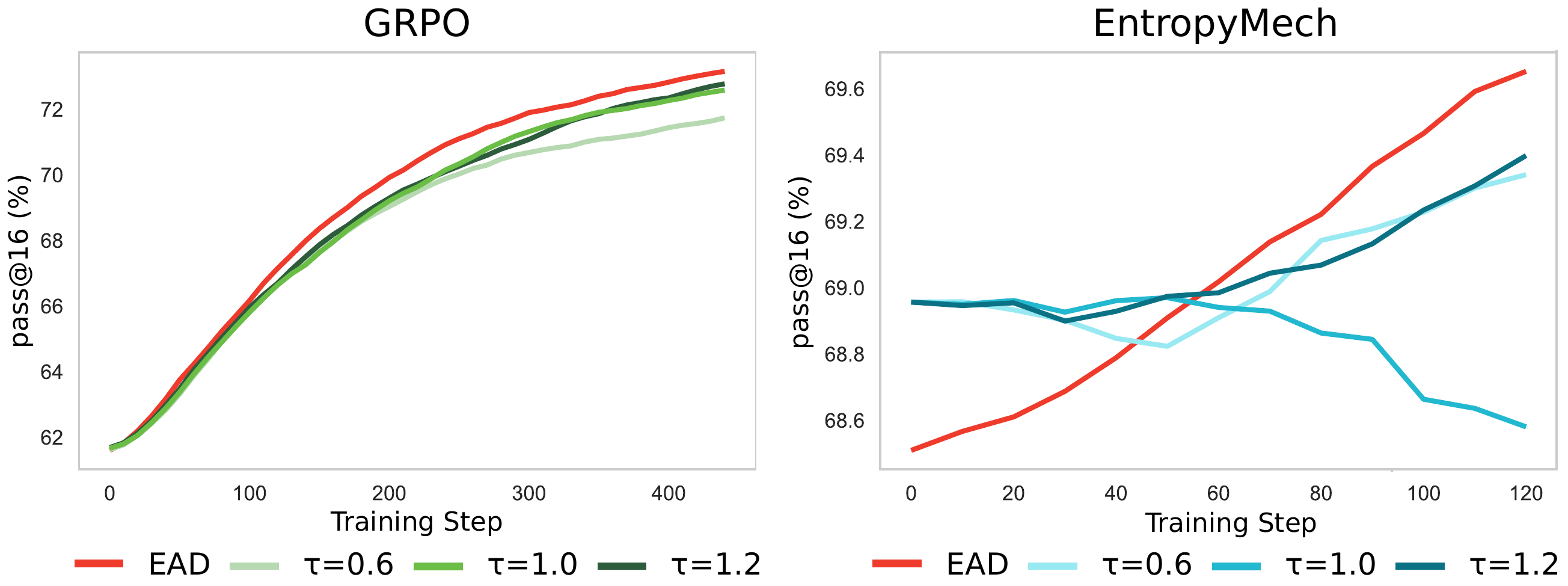}
    \caption{\alg is compatible with various RL algorithms and can significantly improve the model performance over time. } 
    \label{fig: algorithm_wide_comparison}
    \end{figure}
To demonstrate that \alg is a general, plug-and-play exploration strategy, we evaluate its performance when integrated into two other prominent RL algorithms: GRPO~\citep{shao2024deepseekmath} and EntropyMech~\citep{cui2025entropy}. 
These algorithms provide diverse testbeds. 
GRPO is more conservative, constraining policy updates with a KL divergence penalty and stricter clipping mechanism that can limit exploration~\citep{yu2025dapo}, while EntropyMech uses a specialized token-clipping mechanism to mitigate entropy collapse. 

As shown in \cref{fig: algorithm_wide_comparison}, \alg consistently outperforms fixed-temperature sampling in both frameworks. 
These results confirm the broad applicability of our method as an improved exploration strategy across different RL algorithms.

\subsection{\alg Improves Inference-Time Scaling}
\begin{figure}
    \centering
    \includegraphics[width=0.8\linewidth]{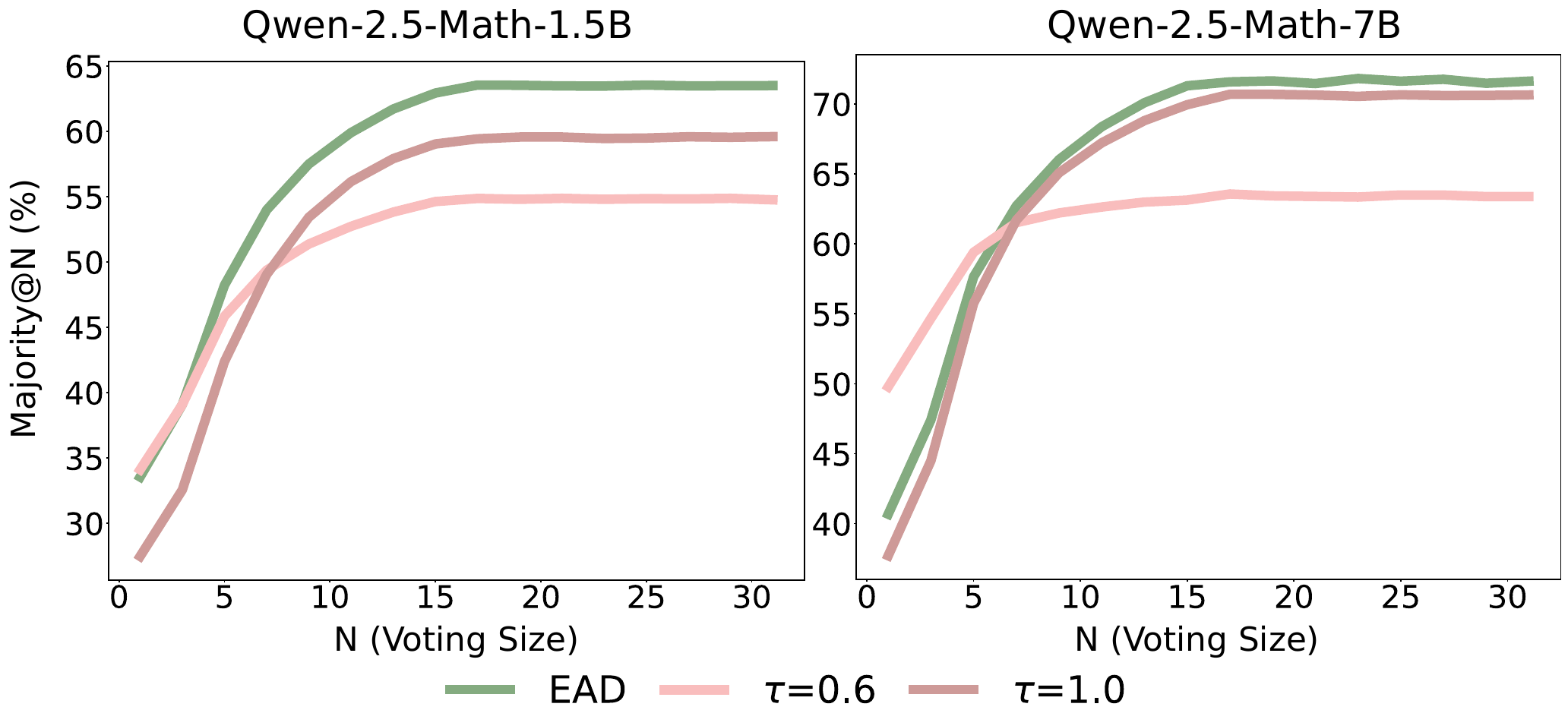}
    \caption{Inference-Time Scaling Evaluation for Different Decoding Methods using off-the-shelf Qwen2.5 models. We could see that \alg improves traditional temperature sampling. We set $\tau_{\text{max}}=1.2, \tau_{\text{min}}=0.1, d=25$ for \alg. }
    \label{fig: inference_time_scaling}
    \end{figure}
To understand whether the success of \alg in RL training is driven by its ability to generate high-quality samples, we conduct an evaluation at inference time. 
This experiment is designed to isolate the sampling strategy's effectiveness from the dynamics of RL optimization~\citep{berseth2025exploration}. 
Using off-the-shelf Qwen-2.5 models without any fine-tuning, we compare \alg against fixed-temperature sampling. 
We use majority voting ($\text{Majority}@N$) to measure how performance scales with the number of samples $N$~\citep{wang2023selfconsistency, snell2024scaling}. 
As shown in \cref{fig: inference_time_scaling}, \alg consistently improves over the baseline for most values of $N$. 
This result confirms that \alg's advantage stems from its inherent capacity to discover higher-quality solutions, even without any training.

\begin{figure}
    \centering
    \includegraphics[width=0.5\linewidth]{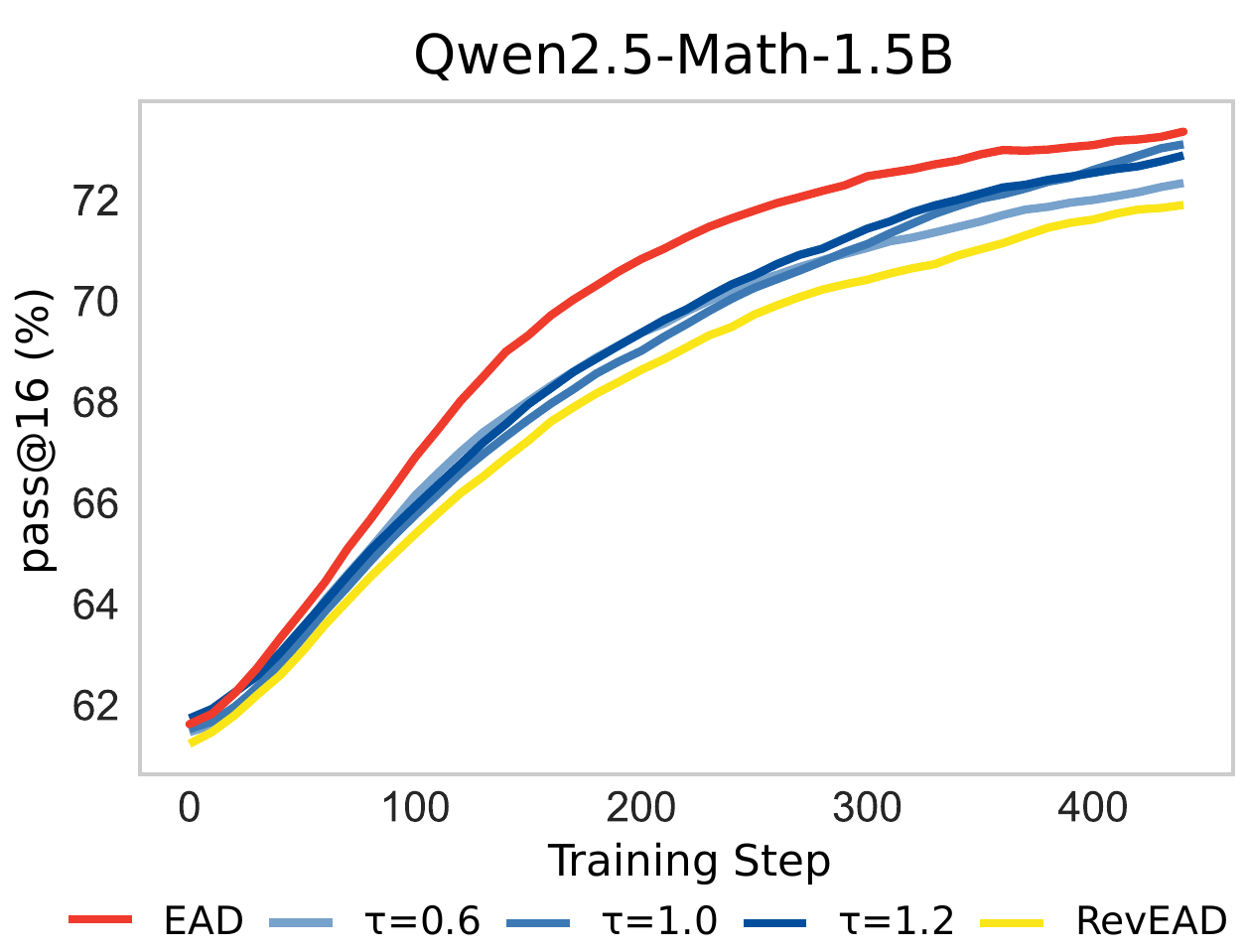}
    \caption{Reverse Annealing experiment results. The reverse schedule fails to surpass normal temperature sampling.} 
    \label{fig: reverse_annealing}
\end{figure}
\begin{figure}[ht!]
     \centering
     \begin{subfigure}[h]{0.4\textwidth}
         \centering
         \includegraphics[width=\textwidth]{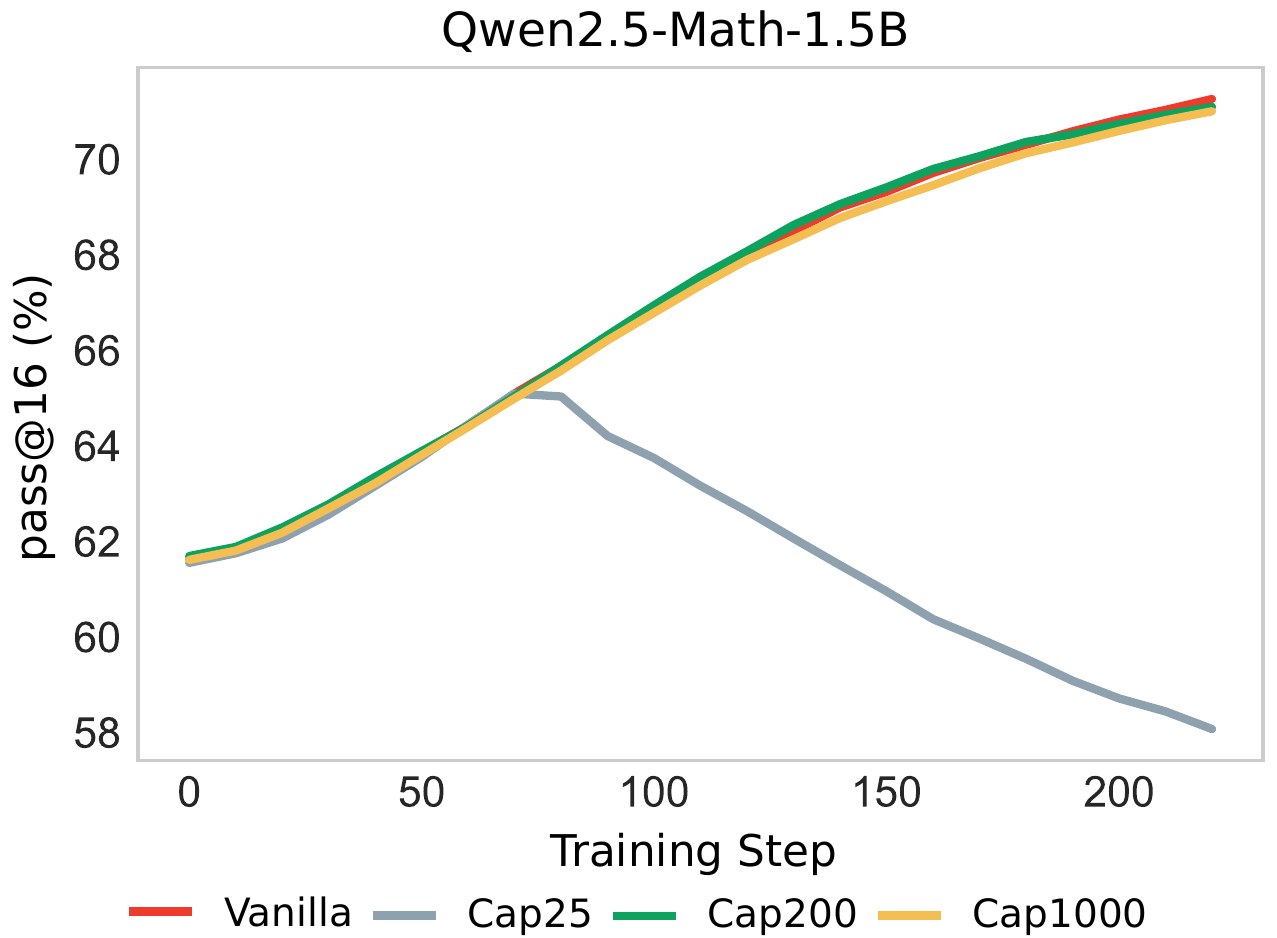}
         \caption{Decay cap $d_{\max}$.}
         \label{fig: ablation_cap}
     \end{subfigure}
     \begin{subfigure}[h]{0.45\textwidth}
         \centering
         \includegraphics[width=\textwidth]{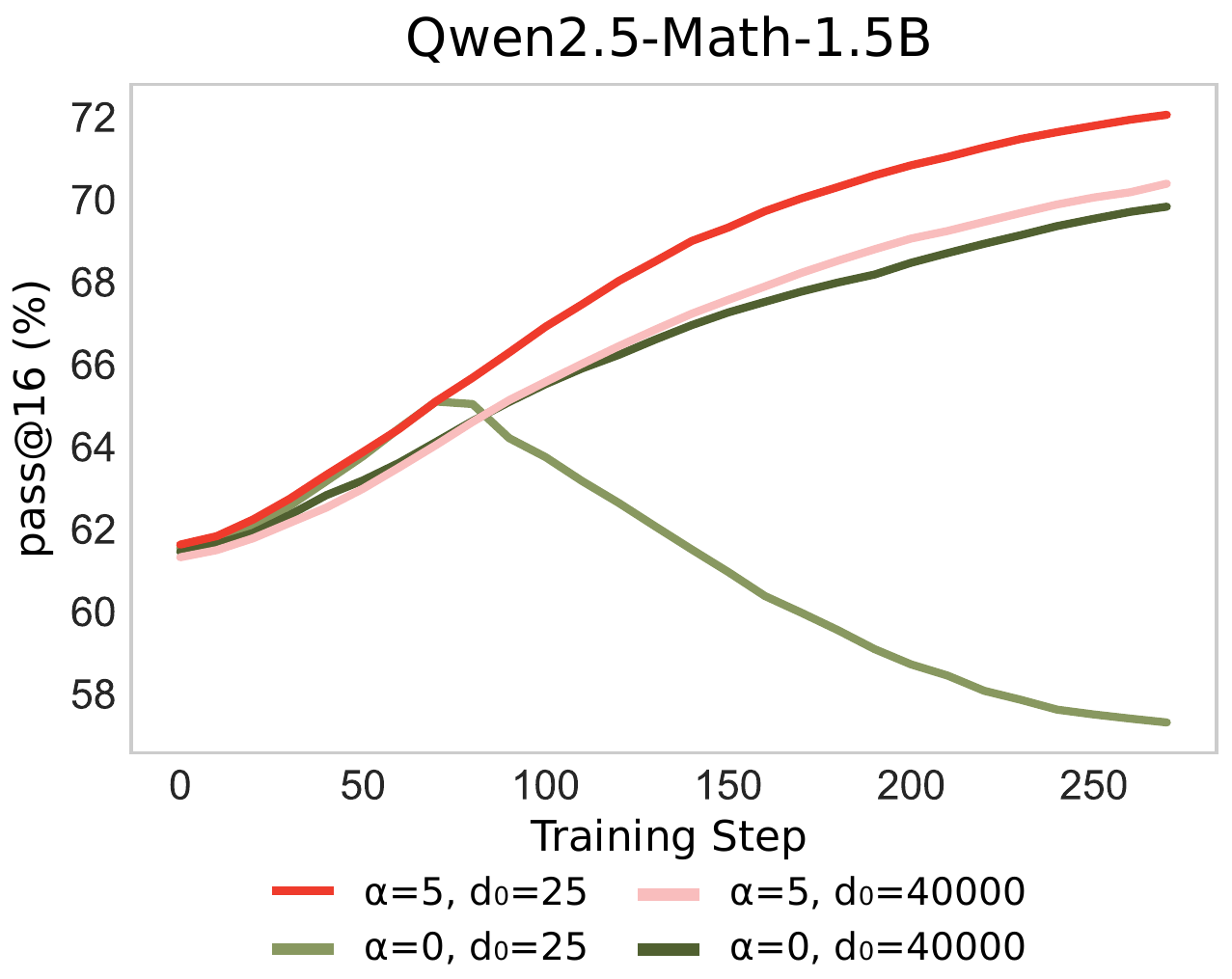}
         \vspace{-0.2in}
         \caption{Growth factor $\alpha$ and initial decay $d_0$.}
         \label{fig: ablation_growth_factor_and_decay_freq_init}
     \end{subfigure}
\caption{Ablation on hyperparameter sensitivity. \textbf{Left:} Ablation on decay cap $d_{\max}$. ``Vanilla'' denotes EAD with the default $d_{\max}=40,000$. Setting $d_{\max}=25$ (where $d_0=25$) mimics a fixed temperature schedule (effectively $\alpha=0$), leading to performance drops as the schedule fails to adapt to longer responses. \alg remains robust provided $d_{\max}$ allows for smooth temperature reduction (e.g., $d_{\max} \ge 200$). \textbf{Right:} Ablation on growth factor $\alpha$ and initial decay $d_0$. With a small $d_0$, a positive growth factor ($\alpha > 0$) is required to adapt to increasing response lengths and maintain a smooth intra-sequence schedule. While a large $d_0$ ensures smoothness even with $\alpha=0$, it limits exploration, resulting in lower learning efficiency.}
\label{fig:hyper-parameter-sensitivity}
\end{figure}

\revise{
\subsection{Ablation Study}

\paragraph{Reverse Annealing} We investigate a counter-intuitive baseline where the temperature increases during generation rather than decreasing. We employ the following reverse schedule:
$\tau_t = \min\{\tau_\mathrm{min} + e^{t/d}, \tau_\mathrm{max}\}$.
We term this schedule ``RevEAD'' and report its performance in \cref{fig: reverse_annealing}. The results demonstrate that reverse annealing fails to even outperform standard temperature sampling, validating the necessity of an annealing (decreasing) temperature schedule.

\paragraph{Hyperparameter Sensitivity} 
We conduct an ablation study on the decay cap $d_{\max}$ and the growth factor $\alpha$. As shown in \cref{fig: ablation_cap}, \alg is robust to a wide range of $d_{\max}$ values, provided the cap is sufficiently large to ensure a smooth annealing schedule. Conversely, a very small cap (e.g., $d_0=d_{\max}=25$) approximates a fixed schedule (equivalent to $\alpha=0$ throughout training). This setting causes an abrupt performance drop, likely because the model cannot adapt to the increasing response lengths during training (see \cref{app: increased_length}). Alternatively, using a large initial $d_0$ yields a smooth schedule even with $\alpha=0$. However, while training remains stable, performance improves slowly due to overly conservative exploration, as shown in \cref{fig: ablation_growth_factor_and_decay_freq_init}.

}

\begin{figure}
    \centering
    \includegraphics[width=0.45\linewidth]{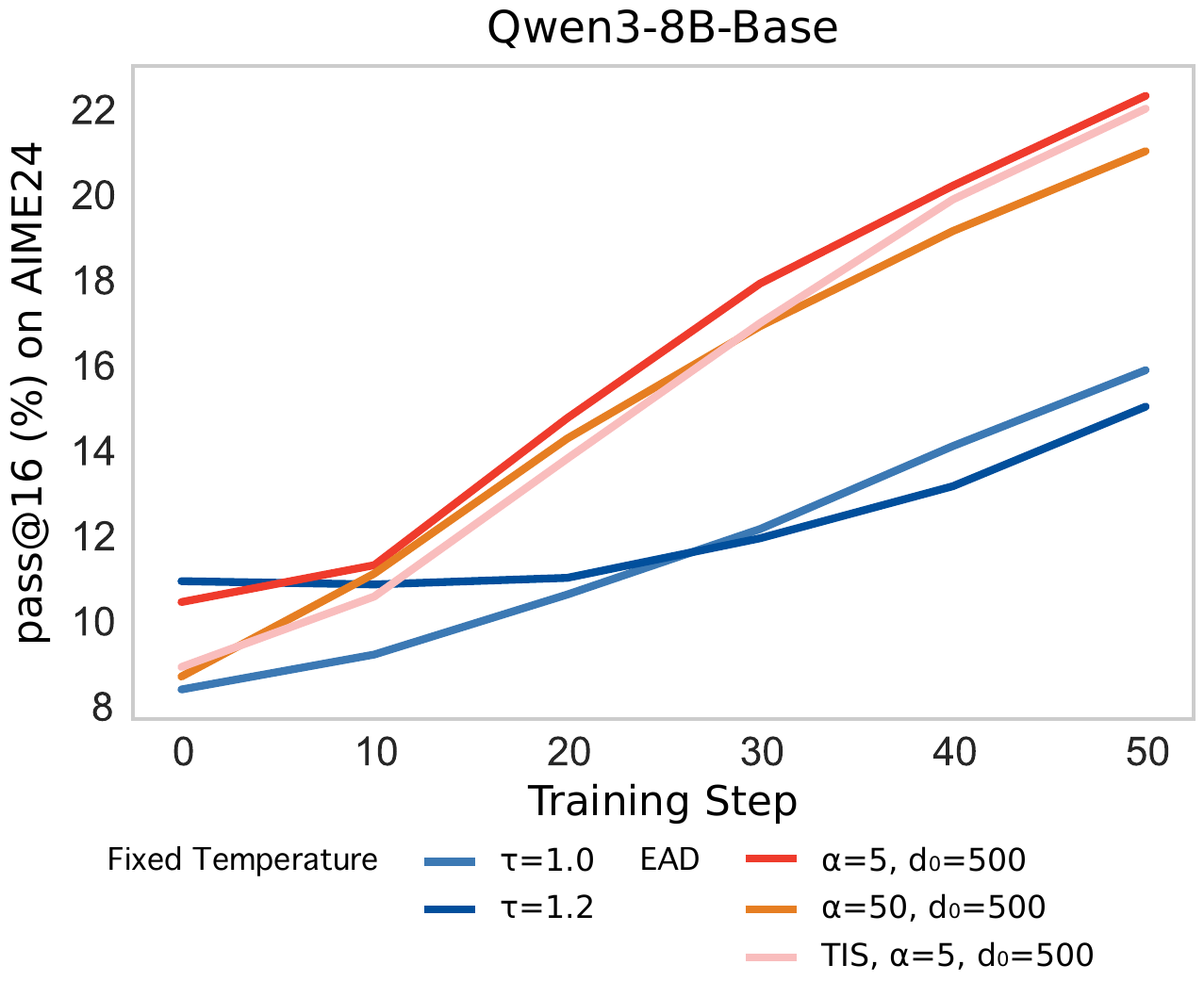}
    \caption{Stress-test of \alg using the DAPO recipe on AIME24. \alg outperforms the temperature sampling baseline across various hyperparameter setups.
    } 
    \label{fig: aime24}
\end{figure}

\revise{

\subsection{Stress Test: DAPO-17K Recipe for AIME24}
To verify the effectiveness of \alg on realistic mathematical reasoning tasks, we stress-test it using the DAPO recipe~\citep{yu2025dapo}, training on DAPO-17k and evaluating on the challenging AIME 2024 benchmark. To accommodate GPU memory constraints, we utilize the Qwen-3-8B-Base~\citep{yang2025qwen3} model (instead of 32B) and reduce the maximum response length from 20,480 to 8,192. We set $d_0=500$ and $d_{\max}=40,000$, and vary $\alpha \in \{5, 50\}$. As shown in \cref{fig: aime24}, \alg achieves significant learning efficiency and consistently outperforms the standard temperature sampling baseline.

}

%% file: past_papers/annealed_sampling/tex/related_work.tex
\section{Related Work}
\label{sec: related_work}
\paragraph{Reinforcement Learning with Verifiable Rewards.}  Recent large-scale reasoning models such as OpenAI o1~\citep{openai2024learning}, DeepSeek‑R1~\citep{DeepSeekAI2025DeepSeekR1IR} have demonstrated that reinforcement-learning-based post‑training can substantially enhance LLM reasoning. Motivated by this reinforcement learning with verifiable rewards (RLVR) (e.g.~\citet{shao2024deepseekmath,DeepSeekAI2025DeepSeekR1IR,lambert2024tulu,yang2025qwen3,hu2025open,yu2025dapo,guan2025rstar,zeng2025simplerl} among many other works) has become a major approach for post‑training LLMs to improve reasoning. A broad literature studies how to make RLVR training effective and efficient at scale, including novel reinforcement learning algorithms and objectives~\citep{yu2025dapo,liu2025understanding,yue2025vapo,zheng2025group}, verifier architecture and reward deigns~\citep{zuo2025ttrl,zhao2025learning,agarwal2025unreasonable,prabhudesai2025maximizing}, and mechanisms that manage exploration diversity and entropy~\citep{cheng2025reasoning,chen2025pass,cui2025entropy,wang2025beyond}, or takes a critical view on the current evaluation~\citep{yue2025does,zhao2025echo,hochlehnert2025sober}. Despite steady progress, a fundamental challenge is to balance exploration and exploitation along long reasoning trajectories without brittle heuristics. We adopt a simple yet effective annealed sampling schedule that front‑loads exploration and cools later steps during rollout to encourage exploration while keeping training stable.

\paragraph{Exploration Control in RLVR.} A line of works that is close to our work studies how to control exploration and sampling while doing RLVR~\citep{hou2025t1, cheng2025reasoning,tangoptimizing,chen2025pass,cui2025entropy,wang2025beyond,xiong2025minimalist,deng2025trial,xu2025not,zheng2025act,dou2025improving, li2025treepo}. In particular, \citet{cheng2025reasoning} propose per-token entropy to focus exploration at branching tokens in sampling; \revise{\citet{tangoptimizing, chen2025pass} transform per-prompt rewards to optimize pass@$k$}, guiding exploration across samples; \citet{cui2025entropy} control high-covariance tokens to prevent entropy collapse and sustain exploration; \citet{wang2025beyond} update only high-entropy tokens, concentrating exploration where decisions split. \revise{While more nuanced multi-threaded exploration strategies~\citep{pan2025learning} could benefit training, adapting them to the training loop for large models often introduces significant computational overhead and the challenges of maintaining massive parallelization.} Different from prior work, this paper presents the first systematic analysis of sampling temperature and introduce a purely sampling-level annealed schedule that encourages exploration and then progressively stabilize answers, thereby enabling discovery of new solutions while yielding more stable training.

\paragraph{Simulated Annealing.} 
Simulated Annealing (SA) is a probabilistic optimization technique inspired by annealing in metallurgy, designed to find the global optimum in a large search space \citep{kirkpatrick1983optimization, bertsimas1993simulated}. The core principle involves a temperature parameter that controls the probability of accepting suboptimal states. Initially, a high temperature allows the search to escape local minima by exploring broadly (exploration). As the temperature gradually decreases, the algorithm increasingly favors better states, converging towards a high-quality solution (exploitation).
This ``coarse-to-fine" search dynamic, where high temperatures establish a solution's general structure and low temperatures refine its details, strongly parallels the generative process of LLMs \citep{yang2026alignment}. SA has been adapted in various machine learning contexts to manage the exploration-exploitation trade-off, including recent applications in graph optimization~\citep{liu2021simulated}, text editing \citep{zhang2024edt}, non-autoregressive generation \citep{israel2025enabling}, and efficient Best-of-N sampling \citep{manvi2024adaptive}. However, these prior applications invariably apply a single, uniform temperature across all positions in a generated sequence. This approach fails to account for the heterogeneous roles of tokens at different positions \citep{wang2025beyond}. Our work departs from this convention. To the best of our knowledge, we are the first to introduce an \textbf{intra-sequence annealed temperature} schedule, where the temperature varies dynamically within the generation of a single sequence. This novel approach allows for more nuanced control over exploration and leads to significant performance gains in RLVR.

%% file: past_papers/annealed_sampling/tex/discussion.tex
\section{Discussion}
Our work addresses a central challenge in RLVR: achieve an effective balance between exploration and exploitation.
We introduce \algname (\alg), a simple yet powerful sampling strategy that avoids heavy computation and intricate heuristics.
Specifically, \alg employs a temperature-annealing schedule that begins with a high sampling temperature and gradually cools, enabling LLMs to explore broadly at the beginning of generation and converge toward precise, high-quality completions throughout the decoding process.
As RLVR often relies on multiple rollouts to estimate rewards, this annealing schedule effectively improves sampling diversity while controlling variance, making it well suited for RL training.
At the same time, \alg can also be applied directly at test time to enhance inference efficiency and scaling, improving the quality of single- or multi-sample decoding without additional computation cost.

Despite the encouraging results, our study has several limitations that suggest directions for future work. 
First, the scaling behavior of our method is not fully explored because of limited computational resources. 
We adopt the current settings with reference to~\citep{xiong2025minimalist,shao2025spurious,wang2025reinforcement}, and argue that the efficacy of the method is still convincing, as we evaluate it across diverse model structures (LLaMA and Qwen) and multiple model sizes. 
A systematic scaling study remains an important next step.
Second, while \alg is designed as a complementary component, a comprehensive study combining it with other advanced exploration-promoting RLVR algorithms (see \cref{sec: related_work}) remains a promising direction for future work.
Third, our current experiments adopt a uniform temperature schedule for all prompts. 
Although an adaptive schedule tailored to individual prompts could potentially enhance performance, developing such a mechanism is nontrivial. 
In RLVR, training is iterative, so any prior information about prompt distributions may shift during optimization, and collecting extra statistics (e.g., token-wise entropy quantile~\citep{wang2025beyond}, or probability-advantage covariance~\citep{cui2025entropy}) to track these changes for every prompt would add computational overhead and system complexity~\citep{li2025treepo, liu2025uniform}.
For these reasons, we focus on the vanilla schedule to test the core efficacy of our method, leaving adaptive scheduling for future investigation.

In summary, \alg provides a simple yet general way to couple exploration with the inherent progression of language generation. 
By reducing algorithmic overhead while improving trajectory quality, it opens new avenues for both efficient inference and effective reinforcement fine-tuning.

%% file: chapters/addiction.tex
\section*{Chapter Overview}
In the domain of social health, we apply grounded generation to identifying phases of Opioid Use Disorder (OUD). By forcing models to ground their predictions in explicit rationales (extracted spans), we demonstrate significant performance improvements in detecting subtle stages like misuse and recovery, showcasing the real-world impact of grounding.

\graphicspath{{./}}

\makeatletter
\def\input@path{{./}}
\makeatother

\input{past_papers/addiction/sections/introduction}

\input{past_papers/addiction/sections/data}

\input{past_papers/addiction/sections/modeling}

\input{past_papers/addiction/sections/experiments}

\input{past_papers/addiction/sections/explanations}

\input{past_papers/addiction/sections/error_analysis}

\input{past_papers/addiction/sections/related_work}

\input{past_papers/addiction/sections/conclusion}

\input{past_papers/addiction/sections/limitations_and_ethics}

%% file: past_papers/addiction/sections/introduction.tex
\section{Introduction}
Extensive ongoing overuse of opioid medications, both from medical prescriptions and from illegal sources has led to a major
public health crisis~\citep{degenhardt2019global, krausz2021opioid}.
There have been a total of 103,664 drug overdose deaths in the US in the 12-month period ending April 2022.\footnote{\url{https://www.cdc.gov/nchs/nvss/vsrr/drug-overdose-data.htm}}
For individuals with opioid use disorder (OUD), targeted interventions need to be developed to better capture individuals' transitions at critical junctures (e.g., use to misuse; misuse to addiction; recovery to relapse)~\citep{park2020situating}.

\begin{table}[]
\small
\resizebox{\columnwidth}{!}{
\renewcommand{\arraystretch}{1.0}
\centering
\begin{tabular}{|p{0.86\columnwidth}|}
\hline
I'm 18m and I've been taking norcos since I was 16 but just on and off. \textcolor{red}{\textbf{\hl{Starting this year I've been taking it every day basically and now I'm tired of it. I still get high
so ig my addiction isn't that bad}}} as others but I don't want to get to that point. I'm tired of chasing the high. I've spent at least 3k on norcos this year and \textcolor{red}{\textbf{\hl{I can't control }}} \textcolor{red}{\textbf{\hl{myself}}}. I try to go a day sober but my mind is telling me I need 
and then withdrawals starts 
[...]
\\ \hline
\end{tabular}
}
\caption{\label{excerpt}A self-disclosure from a user on Reddit going through the cycle of Opioid Addiction.} 
\end{table}

Due to their anonymous and real-time participation, community-based social media platforms such as Reddit, have been used by researchers to understand issues around mental health self-disclosure \citep{Choudhury2014MentalHD}, suicide among youth \citep{Sumner2019TemporalAG},
marijuana regulations \citep{park2017tracking}, drug community analysis~\citep{bouzoubaa2023exploring} and Covid-19 impact on people who use opioids \citep{el2022harnessing}.
We choose Reddit for our research,
specifically the popular opioid-related subreddits \textit{r/Opiates, r/OpiatesRecovery} as well as \textit{r/drugs} to collect our data (\cref{collection}).
Our research focuses on
predicting the presence of self-disclosures related to OUD phases in users' Reddit posts (refer to \cref{excerpt} for an example). 
This task is critical in providing healthcare professionals and social workers with automated tools for detecting OUD indications in social media posts. Accurate identification of such self-disclosures can enable more effective, targeted interventions for individuals suffering from OUD, as supported by prior research \citep{acion2017use, park2020situating, hasan2021machine}.
Our goal is to establish an annotation framework based on addiction and substance use research, categorizing behaviors like Medical Use, Misuse, Addiction, Recovery, and Relapse. We also seek to demonstrate the effectiveness of recent NLP advancements, especially through the application of explanations and text-to-text models, in accurately identifying self-disclosures within the OUD continuum. We offer three primary contributions:
\begin{itemize}
\item 
{\it An
annotation scheme amenable for both expert and novice annotations of self-disclosures.}
The proposed scheme has three characteristics: 1) is grounded in research on addiction and substance use
2) aims to focus on self-disclosure of OUD phases by including a category Not Using that applies to posts that are not discussing the author's OUD experience; and 3) aims to provide reliable annotations by both experts and novices (\cref{sec: data}). 
 \item 

{\it High-quality dataset annotated with class labels and text explanations
using expert and novice annotators.} Human annotations are essential, both to ensure that the NLP models can accurately learn to identify the various OUD phrases, and as an upper bound on the expected model performance. Towards this, we employ both substance use research experts and skilled crowd-workers to annotate our data based on our scheme (\cref{collection}). 
To ground annotators' decisions towards a particular label, we also asked them to highlight the minimum span from the input that acts as an explanation for their chosen category/label. 
 \item 

{\it Thorough experimental setup of zero-shot, few-shot, and supervised models with insights into the role of explanations for model performance, the impact of label uncertainty, and intriguing properties of users' self-disclosure. 
}
Our experiments demonstrate that: 1) the model performance improves significantly 
when trained/prompted with explanations. A further ablation study on 
human-annotated explanations versus machine-generated 
explanations confirms that the quality of explanations is key to 
such improvement;
2) smaller models
fine-tuned on our novice-annotated data with explanations works best,  
surpassing zero-shot
and few-shot large models, including GPT-4,  
by a large margin (\cref{sec: exp}); 3) an ablation study taking into account label uncertainty sheds light on model errors for cases where humans agree or disagree on the label; 4) our error analysis shows preliminary insights in understanding users' self-disclosure (\cref{sec: error_analysis}).

 \end{itemize}

%% file: past_papers/addiction/sections/data.tex
\section{Data}
\label{sec: data}
\subsection{Data Collection and Annotation}
\label{collection}

\shortparagraph{Data Source} One of the greatest challenges in building models that are capable of identifying the appropriate category for opioid usage is the lack of publicly available large-scale datasets. Social media platforms such as Reddit often provide social support
for people who use opioids, while allowing for anonymity when discussing stigmatized behaviors \citep{pandrekar2018social, Bunting2021SociallysupportiveNA}.
We collect data from the popular opioid subreddits, \textit{r/Opiates} and \textit{r/OpiatesRecovery} as well the \textit{r/drugs} subreddit.
Since \textit{r/drugs} can contain posts related to other drugs, we only select posts that are labeled with a flair(tag) ``opioids" by the moderator.

\begin{table*}[t]
\small
\resizebox{\columnwidth}{!}{
\centering
\renewcommand{\arraystretch}{1.0}
\begin{tabular}{@{}cl@{}}
\toprule
Medical Use
& \begin{tabular}[c]{@{}l@{}}\underline{\textbf{Oxycodone for wisdom teeth removal }}\textcolor{red}{\textbf{\hl{I just got 4 wisdom teeth plus another tooth in}}}\\ \textcolor{red}{\textbf{\hl{my palette removed and got prescribed 1 or 2 5mg tablets of oxy (Endone) each time}}}. \\He recommended to avoid it if I could since I'm 43kg and have no tolerance. [...]
\end{tabular}   \\ \midrule
Misuse & \begin{tabular}[c]{@{}l@{}}\underline{\textbf{Oxy nod but no euphoria? }} Hi everyone, \textcolor{red}{\textbf{\hl{I tried oxy for the first time a few weeks back}}}\\ \textcolor{red}{\textbf{\hl{snorting a prolonged 20mg tablet and felt pretty good}}}. Wednesday I dropped 9 of the \\5mg capsules over a couple hours and was nodding strongly [...]
\end{tabular} \\ \midrule
Addiction       &   \begin{tabular}[c]{@{}l@{}}\underline{\textbf{Well y'all were right. }} The sickness came. And is the worst i've ever experienced. Took \\subs, went into pwd  accidentally and jump started the methadone sickness. \textcolor{red}{\textbf{\hl{I am to the}}} \\\textcolor{red}{\textbf{\hl{point that I just have to get off this godforsaken mountain and go back to my ex and }}}\\\textcolor{red}{\textbf{\hl{get back in the clinic bc at this rate i'm afraid i'm gonna end up killing myself}}}.[...].
\end{tabular} \\ \midrule
Recovery       & \begin{tabular}[c]{@{}l@{}}\underline{\textbf{It's my birthday!} \textcolor{red}{\textbf{\hl{One year off opiates}}}}  It's been 365 days since I decided to take back\\ control of my body. I was highly dependent and addicted to prescribed opiates [...]
\end{tabular} \\ \midrule
Relapse      &  \begin{tabular}[c]{@{}l@{}}\underline{\textbf{So high. 18 hours later. Still so high. }}So I'm pissed at myself. \textcolor{red}{\textbf{\hl{I was clean from heroin }}}\\\textcolor{red}{\textbf{\hl{for 11 months and last night I did some}}}. And for no reason too [...] 
\end{tabular}                                                       \\ \midrule
Not Using 
& \begin{tabular}[c]{@{}l@{}}\underline{\textbf{Partners of an Opiate addict in recovery }}How do you guys do this? I feel like I am \\having an incredibly hard time "moving on". \textcolor{red}{\textbf{\hl{I have nightmares of my partner oding,}}} \\\textcolor{red}{\textbf{\hl{dying, and pretty much anything else that involves drug use}}}. I over analyze everything [...]
\end{tabular}                                                                                                                                                                                                                                                                                                                                                                                                                                                                                                                \\ \bottomrule
\end{tabular}
}
\caption{\label{data} Example for each
Opioid Usage category. The underlined bold text represents the title of each post. Highlighted text
represents salient spans annotated by humans as explanations for the label.}
\end{table*}

\shortparagraph{Anonymization and Data Preprocessing} 
To remove any personal identifying information (PII) that users might divulge in their posts (e.g., emails) and broken characters, we use \textit{cleantext}\footnote{\url{www.github.com/prasanthg3/cleantext}} to preprocess raw social media posts. In addition, we manually investigated
all samples prepared for annotation to make sure PII will not be exposed to annotators, and thus will not be released in the final dataset.  After that, 
we check whether each post is of reasonable length (title + text), and
filter the preprocessed posts having a length of less than $10$ words (%
or more than $200$ words for easier annotation).
We sample $600$ posts for expert annotation and $2,250$ posts for novice 
annotation.\footnote{Domain experts are postdoctoral and advanced doctoral students working in substance abuse research. We use MTurk for novice annotation. }

\shortparagraph{Annotation Guidelines} 
To ensure the annotation quality, we worked closely with substance use research experts to develop comprehensive and precise annotation guidelines for different phases of opioid use. OUD has been recognized as a chronic, relapsing disorder in which individuals may begin at one stage, remain in that stage, gradually or rapidly advance to another stage, enter recovery, return to use, or even skip stages~\citep{volkow2007science}. 
For this study, we adopted frequently used classifications to assign each post a stage in the continuum: Medical Use, Misuse, Addiction, Recovery, and Relapse \citep{nida2007drugs, smith2013classification, hanson2013tweaking, hanson2013exploration, chan2015canary, anderson2017using, phan2017enabling, hu2019ensemble}. Our definitions for Medical Use, Misuse, and Addiction come from the systematic review~\citep{smith2013classification}, and our definitions for Recovery and Relapse come from National Institute on Drug Abuse guideline~\citep{nida2007drugs}.
We also built a list of keywords, representative samples and FAQs to clarify the project background, ethical considerations, and how to handle uncertain cases. The guidelines aim to understand the opioid use experiences of \emph{the author of the post} (self-disclosures). Thus,
we introduced also a category of 'Not Using' that includes discussion about someone else who uses opioids or general questions about opioids, without evidence of use. \cref{app: annotation_guideline} shows the definitions for each category and some examples of expert-authored FAQs for clarification. \cref{tab:stat} shows the distribution of OUD categories in the annotation data.

\begin{table}[!ht]
\small
\centering
\renewcommand{\arraystretch}{0.9}
\begin{tabular}{ccc}
\toprule
Category      & Novice & Expert  \\ \midrule
Misuse      & 22.10 & 20.0  \\ 
Addiction   & 29.15 & 12.53 \\ 
Recovery    & 18.89 & 25.49 \\ 
Relapse     & 4.65  & 3.96  \\ 
Medical Use & 7.05  & 3.52  \\ 
Not Using   & 18.17 & 34.51 \\ 
\bottomrule
\end{tabular}
\caption{Distribution (\%) of OUD categories 
in novice- and expert-annotated data.}
\label{tab:stat}
\end{table}

\shortparagraph{Expert Annotation}
To build the expert evaluation dataset, we invited 4 substance use research experts to annotate $600$ posts and paid them at a rate of \$20/hour. To accommodate the experts' available timeslots, we split the posts into two equal batches and asked the experts to annotate the text and title of the post with both the label and the explanation. All four experts annotated the first batch. For label annotation, the inter-annotation agreement (IAA)
was $0.46$ Fleiss' kappa~\citep{fleiss1971measuring}, indicating ``moderate agreement''. Only two experts were available to annotate the second batch, and the IAA was $0.62$ Cohen's kappa~\citep{cohen1960coefficient}, indicating ``substantial agreement''.
We filtered the posts that did not obtain majority agreement, obtaining an
expert-annotated dataset of $455$ posts. We further split the dataset into $13$ samples
as in-context samples of few-shot prompting (\cref{sec:exp-few-shot}) and $442$ samples for testing.

\label{crowd}
\shortparagraph{Novice Annotation} Expert annotation, while being more accurate and trustworthy, is not feasible for scaling the process beyond a few hundred posts. Hence, we aimed to leverage
novice annotators using Amazon MTurk. However, to obtain reliably annotated data, we need to ensure that novices are qualified and trained.
Thus, we first conducted a qualification test where we recruited a total of 85 crowd-workers from the USA with a 98\% success rate and asked them to annotate 250 randomly selected instances from the expert-labeled set. We qualified only 10 crowd-workers who obtained >60\% accuracy in the qualification phase. In addition, for cases of disagreement with the experts, we further trained the novice annotators by providing them with follow-up explanations. We paid them \$15/hour, which is in accordance with the minimum wage in the USA. Every post is labeled by three qualified novice annotators. We labeled 2,250 posts and kept 2,086 for which we could obtain a majority vote label. 
We split this set into 1,936 for training and 150 for testing. IAA was 0.47  based on Fleiss' kappa~\citep{fleiss1971measuring} (``moderate agreement'').

\label{explanation}
\shortparagraph{Explanation Annotation} Along with providing a  label we also asked annotators to identify the minimum salient span from the text that justifies their decision towards labeling a post to a certain category.
For cases where we have a majority vote and use the corresponding label as gold, we have to decide what explanation to include. We computed the max overlapping substring between the annotators' explanations.
When the max overlapping substring is very short (typically $<10$ characters), we chose the longest explanation
whose annotated label matches the majority vote label.
For 63\% of cases, there is significant overlap among annotators' selected explanation spans, while for 37\% of cases the longest explanation is selected. 
\cref{data} shows post examples in each category/label along with their annotated span-level explanations.

\subsection{Disagreements in Annotation}
\label{sec: disagreement}

\begin{table}[]
\small
\resizebox{\columnwidth}{!}{
\renewcommand{\arraystretch}{1.0}
\centering
\begin{tabular}{|p{0.86\columnwidth}|}
\hline
\begin{tabular}[c]{@{}p{0.86\columnwidth}@{}}\textcolor{red}{\textbf{\hl{Is it safe to mix 20mg oxy with a few standard drinks?}}} Seen online it's dangerous but I don't trust a lot of those harm reduction websites, most of them are whack.
EDIT: \textcolor{blue}{\textbf{\hl{I didn't do it don't worry, thanks for the info}}}\end{tabular} \\ \hline
\end{tabular}
}
\caption{An annotation example demonstrating the role of explanation annotations for understanding annotator disagreement: the red is associated with ``Misuse'' and the blue with ``Not Using''.  }
\label{tab: different_label_different_rationale}
\end{table}
\shortparagraph{Expert-Novice Disagreement} During the qualification test, we observe a consistent labeling disagreement between our qualified novice annotators
and the expert annotators (The confusion matrix is shown in \cref{app:heatmap}).
The main disagreement between experts and novices are 
between \textbf{``Addiction'' - ``Recovery''} (22.35\%), \textbf{``Not Using'' - ``Misuse''} (19.35\%), \textbf{``Addiction'' - ``Misuse''} (12.90\%),  \textbf{``Medical Use'' - ``Misuse''} (10.75\%) and \textbf{``Recovery'' - ``Relapse''} (8.60\%). 
\shortparagraph{Novice-Novice Disagreement.} Even though we reach a majority vote
for 2086 posts,
an individual worker can still disagree on the collective label. Looking at these disagreements can help us better understand the difficulty of this task and the uncertainty in the annotated dataset. In total, $1165$ out of $2086$
(56\%) posts in our final novice-annotated datasets fall in  this category. The top-5 disagreements happen between \textbf{``Addiction''-``Misuse''} (34.84\%), 
\textbf{``Recovery''-``Addiction''} (15.71\%),
\textbf{``Not Using''-``Addiction''} (12.27\%), 
\textbf{``Not Using''-``Misuse''} (9.78\%) and \textbf{``Not Using''-``Recovery''} (7.38\%). 
This inherent uncertainty may inject wrong inductive bias into models, which we discuss in \cref{sec: error_analysis}.

A closer look at some examples of disagreement in annotations shows that selected explanations could shed some light. For example, \cref{tab: different_label_different_rationale}, shows an example of disagreement between Misuse and Not Using, where the annotators selected two different explanations for the labels.

%% file: past_papers/addiction/sections/modeling.tex
\section{Modeling Strategies}
\label{sec:addiction:modeling}
\begin{table}[]
\small
\centering
\renewcommand{\arraystretch}{1.0}
\begin{tabular}{|p{0.94\columnwidth}|}
\hline
\begin{tabular}[c]{@{}p{0.9\columnwidth}@{}}Given the following title, text, and explanation from the text, please identify the appropriate opioid usage category among the following types: {\color{blue}'Medical Use', 'Misuse' ,'Recovery', 'Relapse', 'Addiction', 'Not Using'.}\\\\ 
Title: \{\{title\}\}\\  
Text:  \{\{text\}\}\\ 
Explanation:  \{\{explanation\}\}
\end{tabular} \\ \hline
\end{tabular}
\caption{Zero-shot instructional prompts for T0pp for
\textit{w/ Explanation} setting}
\label{zeroshot}
\end{table}

As OUD status prediction is a high-stakes task with limited labeled data, we consider three different settings, gradually increasing the number of labeled data required to mimic real-world application scenarios: zero-shot, few-shot, and supervised learning. 
To understand the effectiveness of annotated span-level explanations, we conduct two experiments for each setting: i) \textit{w/o Explanation}: where the explanation is not included in the input, and ii) \textit{w/ Explanation}: otherwise.

\shortparagraph{Zero-Shot} 
We first consider the extreme application scenario when zero training data is given.
In order to measure zero-shot performance on our dataset, we
prompt the widely-used instruction-tuned T0pp \citep{sanh2022multitask} model for our task. 
The prompt with instructions are demonstrated in \cref{zeroshot}.%
\footnote{We only presented \textit{w/ Explanation} case to save space.}
We use greedy search to generate the labels, then use exact match to compute accuracy after lowercasing both the output and label.

\begin{table}[]
\small
\centering
\begin{tabular}{|p{0.85\columnwidth}|}
\hline

\begin{tabular}[c]{@{}p{0.85\columnwidth}@{}}Given the following title and text, please identify the appropriate opioid usage category among the following types:  {\color{blue}'Medical Use', 'Misuse', 'Recovery', 'Relapse', 'Addiction', 'Not Using'.} 
{\color{orange} Please provide an explanation for your answer by extracting the relevant span from the text that justifies your choice.}
\\\\
\{\textcolor{red}{13 in-context samples with the format below}\} \\

Title: \{\{title\}\}\\ 
Text:  \{\{text\}\}\\ 
Label:  \{\{label\}\}\\ 
Explanation:  \{\{explanation\}\}\end{tabular} \\ \hline
\end{tabular}

\caption{Few-shot instructional prompt for GPT-3 for \textit{w/ Explanation} setting.}
\label{tab: gpt-3-prompt}
\vspace{-8pt}
\end{table}

\shortparagraph{Few-Shot}
\label{sec:exp-few-shot}
Now we relax the dataset size limitation to allow the few-shot setting. 
 We use the GPT3-Davinci-002 model~\citep{brown2020language} and GPT-4~\citep{openai2023gpt4tr} for the few-shot learning method. Our prompts
begin with the task instruction followed by 13 expert-annotated samples for in-context learning.\footnote{See \cref{app:in-context-samples} for these in-context samples and the detailed explanations for selecting these examples. }
For in-context learning \textit{w/ Explanation}, we place the explanation on a line after the answer, preceded by ``Explanation:"\citep{lampinen2022can}. \cref{tab: gpt-3-prompt} shows an example prompt.\footnote{We only show \textit{w/ Explanation} case to save space.} In this way, the evaluation can be performed regardless of whether explanations are provided in the prompt or not.\footnote{Necessary post-processing during evaluation for GPT-3/4 output normalization is detailed in \cref{app:post_processing_gpt3}.}%

\begin{table*}[]
\centering
\resizebox{0.95\textwidth}{!}{%
\renewcommand{\arraystretch}{1.0}
\begin{tabular}{@{}cccccccc@{}}
\toprule
\multirow{2}{*}{Mode}            & \multirow{2}{*}{Test Set} & Zero-Shot & \multicolumn{2}{c}{Few-Shot} & \multicolumn{3}{c}{Supervised}                                    \\ \cline{3-8} 
                                 &                             & T0pp      & \multicolumn{1}{c}{GPT3} & \multicolumn{1}{c}{GPT4}    & \multicolumn{1}{c}{DeBERTa} & \multicolumn{1}{c}{T5-3B} & T5-11B \\ \hline
\multirow{2}{*}{w/o Explanation} & Expert                      & {48.9 / 46.9}      & 62.2 / 57.1 & \multicolumn{1}{c}{55.4 / 50.2}    & \multicolumn{1}{c}{67.6 / 65.7}    & \multicolumn{1}{c}{63.5 / 61.2}  &  \textbf{71.4} / \textbf{70.4}      \\ 
                                 & Novice                 & 62.0 / 60.8      & 66.4 / 65.4 &  \multicolumn{1}{c}{63.3 / 60.0}    & \multicolumn{1}{c}{74.0 / 74.4}    & \multicolumn{1}{c}{72.7 / 71.4}  & \textbf{80.9} / \textbf{81.5}       \\ \hline
\multirow{2}{*}{w/ Explanation}   & Expert                      & 48.9 / 47.4     & 61.1 / 54.5 & \multicolumn{1}{c}{63.2 / 59.1}    & \multicolumn{1}{c}{{73.8 / 72.6}}    & \multicolumn{1}{c}{{64.4 / 65.5}}  & \textbf{76.6} / \textbf{77.0}        \\ 
                                 & Novice                 & {62.7 / 58.5}      & {66.9 / 65.9} & \multicolumn{1}{c}{67.3 / 64.8}     & \multicolumn{1}{c}{{81.3 / 81.9}}    & \multicolumn{1}{c}{{78.7 / 77.3}}  & \textbf{84.0} / \textbf{84.0}       \\ \hline
\end{tabular}
}
\caption{Performance of different models on expert and novice-annotated test data in a zero-shot/few-shot/supervised setting. \textit{w/o Explanation} and \textit{w/ Explanation} models refers to the setting where \textit{Explanations} are excluded or included as part of the input. Results are presented in "Accuracy/F1" format. }
\label{tab:main_exp}
\end{table*}

\shortparagraph{Fully Supervised}
All of our training data comes from the novice-annotated set, while our test sets consist of expert or novice-annotated data.  Our training data consists of 1936 examples, while our test sets consist of 442 expert-annotated examples and 150 novice-annotated examples. We consider two modeling variants: Masked Language Models (MLM) (as it is often used in traditional fine-tuning) and Generative Language Models (GLM) (as it is often used in instruction-tuning).

For MLM, we fine-tune DeBERTa-v3-large~\citep{he2021debertav3} on our training data.%
For input formatting, we use ``[title] {TITLE} [text] {TEXT} [Rationale] {RATIONALE}'' as the input for \textit{w/ Explanation} settings and use ``[title] {TITLE} [text] {TEXT}'' to train models under \textit{w/o Explanation} settings. The token in square brackets (e.g., ``[title]'') are special tokens and the tokens in all-caps
(e.g., ``{TEXT}'') are actual text fields for each post. 
For GLM,
 we fine-tune T5-3B and T5-11B models ~\citep{raffel2020exploring}.  
We use the same instruction as input to the encoder for a given title, text and optionally explanation as the ones we used for zero-shot setting (see \cref{zeroshot}).
The decoder generates the textual label autoregressively. More implementation details for fine-tuning can be found in \cref{sec:fully_supervised_fine-tuning_detail}.

%% file: past_papers/addiction/sections/experiments.tex
\section{Experiments}
\label{sec: exp}
\cref{tab:main_exp} summarizes our experimental results under three different learning settings (zero-shot, few-shot, and fully supervised) across two different modes i) \textit{w/o Explanation} when only the Title and Text are a part of our input during training and testing, and ii) \textit{w/ Explanation} when along with Title and Text, gold human annotated explanations are a part of our input during training and testing. We show results on both the expert-annotated test set and the novice-annotated test set. As the OUD category distribution in our dataset is unbalanced, we report both accuracy and macro F1 scores in \cref{tab:main_exp}. We find for the model-wise comparison, there is little difference in using accuracy or F1.

We highlight several takeaways. First, adding explanations helps the models both on expert and novice-annotated data (except for T0pp and GPT3 on Expert data), particularly in few-shot and fully supervised settings. In \cref{explanations}, we will show additional experiments to study the role of explanations and their quality for model predictions. Second, supervised learning with small models
outperform
few-shot methods with larger models including GPT-4 by a large margin, on both expert and novice evaluation datasets, even if the training data is novice-annotated. T5-11B is the best overall model. While our training data is not annotated by experts, the quality of the data is still high. The accuracy on the expert evaluation set for a random baseline would be 17\%, while a majority baseline would be 35\%, which is significantly lower than 71.4\% or 76.6\% for the T5-11B model performance \textit{w/o Explanation} and \textit{w/ Explanation}, respectively. Moreover, we notice that the performance gap between expert-annotated test data and novice-annotated test data is reduced using supervised models. A closer look at GPT-4 errors shows that GPT-4 is particularly struggling with the ``Not Using'' category, which covers a diverse range of topics that can look very different from posts in other categories, and more analysis on this category will be further studied in \cref{sec: error_analysis}. Third, model capabilities improve with scale under the same family in a supervised setting. The T5-11B model, on average, is about $8.4$ points better than the T5-3B model in accuracy and $9.3$ points better in F1.
However, when models belong to different families (i.e., Generative vs. MLM), the scaling law might not hold as the DeBERTa-v3-large model (1.5B) outperforms T5-3B across both settings (\textit{w/o and w/ Explanation}).

%% file: past_papers/addiction/sections/explanations.tex
\section{The Role of Explanations} \label{explanations}

\begin{table}[]
\centering
\resizebox{0.5\columnwidth}{!}{%
\renewcommand{\arraystretch}{1.0}
\begin{tabular}{@{}cccc@{}}
\toprule
     Explanation             &  Test Set      & T5-11B          & DeBERTa          \\ \midrule
\multirow{2}{*}{Gold}   
                  & Expert           & \textbf{76.6}   & 73.8             \\
                  & Novice           & \textbf{84.0}   & 81.3             \\ \midrule
\multirow{2}{*}{Silver} 
                  & Expert           & 70.3            & 70.3             \\
                  & Novice           & 78.0            & 78.0             \\ \midrule
\multirow{2}{*}{Random}  
                  & Expert           & 69.6 ($\pm$ 1.3) & 65.8 ($\pm$ 1.1)  \\
                  & Novice           & 68.3 ($\pm$ 1.7) & 67.9 ($\pm$ 1.4)  \\
\bottomrule
\end{tabular}
}
\caption{Accuracy of T5-11B and DeBERTa \textit{w/ Explanation} model on expert and novice annotated test sets by varying the quality of explanations. We can observe the importance of including gold explanations.} 
\label{tab:explquality}
\end{table}

To test the quality and helpfulness of the annotated explanations on model prediction, we conduct three different experiments using our two best-performing models trained \textit{w/ Explanation} (T5-11B and DeBERTa).
All these experiments are conducted at inference time on top of a model fine-tuned on <title, text, $E^{\text{gold}}$>. For convenience,  from here on, we will refer to this model as M1. 

\shortparagraph{Gold Explanations at Inference.} In the first experiment, we use the gold explanations from our test sets (expert and novice).
In particular, during inference, we prompt the two best-performing models (T5-11B and DeBERTa) with an input that consists of <title, text, $E^{\text{gold}}$>.  \cref{tab:explquality} shows that models that use gold explanations at inference time are the best. We analyze whether the explanation contains words that refer to the label (e.g., addiction or addicted), a problem referred to as \textit{leakage} \cite{sun2022investigating}.
We notice that there is 5.6\% leakage on expert-annotated test data and 8\% leakage on novice-annotated test data, which means that most of our annotated explanations do not give away the label easily.

\shortparagraph{Silver Explanations at Inference.} In a real-world setting, it is not possible to expect gold explanations at inference time. Thus, in this setting, we investigate whether model-generated explanations can still be helpful for final label prediction. 
Prior works in explainability have trained two types of models: 1) \textit{Pipeline} model, which maps an input to an explanation (I $\rightarrow$ E), and then an explanation to an output (E $\rightarrow$ O); and 2) \textit{Joint Self Explaining} models that map an input to an output and explanation (I $\rightarrow$ OE). The latter has been shown to be more reliable \cite{wiegreffe2021measuring}.
Thus, we first train a T5-11B model (M2) that can jointly generate <{label}, {explanation}> given any <{title}, {text}>. At inference time, we first generate a silver explanation ${E}_{i}^{\text{silver}}$ by prompting M2 with a given <{title}$_{i}$, {text}$_{i}$> from the test set. We then prompt M1 with <{title}$_{i}$, {text}$_{i}$, ${E}_{i}^{\text{silver}}$> to generate label$_{i}$. While these explanations are not as high quality as gold explanations, they still outperform random explanations. It should be noted that the goal of this paper is not to build models that facilitate extracting accurate explanations. However, such models
might improve the silver quality explanations and thereby improve overall classification results. We leave this for future work. 

\shortparagraph{Random Explanations at Inference.} As a baseline,
we use a randomly selected sentence from the post as the explanation. We repeat the random selection for five random seeds and report the mean and standard deviation of these five runs (\cref{tab:explquality}).
Both silver and gold explanations
outperform the random explanation baseline,
indicating the need
for
informative, high-quality explanations.

%% file: past_papers/addiction/sections/error_analysis.tex
\begin{table*}[]
\centering
\resizebox{0.9\textwidth}{!}{%
\begin{tabular}{@{}ccccccc@{}}
\toprule
\multirow{2}{*}{Setting} & \multicolumn{2}{c}{Not Using - Misuse} & \multicolumn{2}{c}{Recovery - Addiction} & \multicolumn{2}{c}{Not Using - Addiction} \\ \cmidrule(l){2-7} 
 & $\rightarrow$ & $\leftarrow$ & $\rightarrow$ & $\leftarrow$ & $\rightarrow$ & $\leftarrow$ \\ \cmidrule(r){1-7}
T5-11B-\textit{w/o Explanation} & 20\% & 4.5\% & 20\% & 0\% & 14\% & 0\% \\
DeBERTa-\textit{w/o Explanation} & 18\% & 0\% & 18\% & 3.6\% & 16\% & 5.5\% \\
\midrule 
T5-11B-\textit{w/ Explanation} & 16\% & 8\% & 14\% & 0\% & 13\% & 0\% \\
DeBERTa-\textit{w/ Explanation} & 15\% & 2.3\% & 15\% & 3.6\% & 15\% & 0\% \\
\bottomrule
\end{tabular}%
}
\caption{Model error analysis over expert annotation data. $\rightarrow$ means the expert-annotated label is on the left side and the predicted label is on the right side, and $\leftarrow$ vice versa. Percentages in the table represent the error rate in each expert labeled category. The results demonstrate that the main confusion for the models exists in ``Not Using - Misuse'', ``Recovery - Addiction'', and ``Not Using - Addiction''. These problems surface in an asymmetrical pattern -- one mis-classification direction matters more in the confusion.} 
\label{tab:expert_err_analysis}
\end{table*}

\begin{table*}[tbp!]
\centering
\resizebox{\textwidth}{!}{%
\begin{tabular}{@{}cccccc@{}}
\toprule
Dataset & Agreement & T5-11B (w/o expl) & T5-11B (w/ expl) & DeBERTa (w/o expl) & DeBERTa (w/ expl) \\ 
\midrule
\multirow{3}{*}{Novice} & Unanimous Consent        & 95.5 / 94.0   &   97.1 / 95.7  &  87.1 / 85.9 & 90.0 / 89.6    \\
& Arguable (Majority Vote)      &   68.0 / 66.7     &    72.5 / 71.3    & 62.5 / 60.8  &  73.8 / 75.2  \\ 
& Arguable (All Annotations)      & 84.0 / 81.4   & 92.5 / 91.8  & 86.3 / 84.3  & 91.3 / 92.1 \\ 
\midrule
\multirow{3}{*}{Expert (FirstBatch)} & Unanimous Consent        & 92.7 / 85.0  &   96.4 / 97.1  &  83.6 / 72.7  & 89.1 / 83.2   \\
& Arguable (Majority Vote)      &   60.3 / 59.9     &    64.0 / 66.0    & 56.9 / 58.2  &  65.0 / 66.2 \\ 
& Arguable (All Annotations)      & 84.6 / 83.3   & 86.8 / 85.8  & 82.5 / 82.7  & 86.9 / 87.3 \\ 
\bottomrule

\end{tabular}
}

\caption{Ablation study on dataset annotation disagreement. Results are presented in "Accuracy/F1" format.
``Unanimous Consent'': all annotators agree on the same label; ``Arguable (Majority Vote)'': annotators have some disagreements, and majority voting is used as the correct label; ``Arguable (All annotations)'': disagreements exist, and any annotator label is considered correct. We  observe that models perform better on data with unanimous agreement than on arguable data. }
\label{tab:ablation_uncertainty}
\end{table*}

\section{Error Analysis}
\label{sec: error_analysis}

To understand the challenges and limitations of our best models, we perform an error analysis.

\shortparagraph{Model Errors}
We compute the confusion matrices
for DeBERTa and T5-11B on the expert evaluation dataset, as it is arguably more reliable and contains more examples.  We generally find that both \textit{w/ Explanation} and \textit{w/o Explanation} models
struggle with confusion between \textbf{1) Not Using - Misuse; 2) Recovery - Addiction; and 3) Not Using - Addiction} (\cref{tab:expert_err_analysis}, some of which we also noticed in the disagreement among annotators.
We notice that these problems surface in a very asymmetrical  pattern -- one direction (e.g., ``Not Using'' $\rightarrow$ ``Misuse'') matters more in this confusion.
Recall that our focus is on self-disclosures, so if a post discusses misuse (either a question or someone else misusing behavior), the expert label is Not Using, which might be difficult for models to capture in some cases.
Adding explanations mostly helps the model by reducing the confusion on the `Not Using` and `Recovery` labels, the two dominant as well as the top-2 most difficult categories
in expert-annotated data.

\shortparagraph{Error Annotations}
To better understand why the model makes mistakes
we did a thorough fine-grained error case annotation for the T5-11B \textit{w/ Explanation} model.
When analyzing why our models misclassified the \textbf{Recovery} class, we notice that
``Recovery'' can be a long process, and it is very common for users
to express their eagerness to get opiates (47.4\%), and/or to talk about their history of addiction 
(21.1\%). 
There are also some hard cases, such as a post showing the patient undergoing repeated recovery-relapse-recovery cycles (the model predicts ``relapse'' in this case). 
When analyzing the cases where our model misclassified the \textbf{Not Using} label, several cases emerge: 1)
\textit{asking a question about use/misuse/addiction (57.69\%)} 
(e.g., ``how much [Drug 1] should I take to get high (safely)? Can I use it with [Drug 2]?'', or asking questions about whether using drugs for certain syndromes is legal in some states and how much they should use);
2) \textit{irrelevant topics (23.08\%)} ( ``Merry Christmas!''); 
3) \textit{Others' overdose (7.69\%)} (discussing the addiction of their friends or family members; 
since we are interested in self-disclosures, this is labeled as Not Using, but models fail to recognize such subtle differences);
and 4) \textit{other drugs/substances, not opiates (3.85\%)} (as we focus on OUD, these posts are labeled ``Not Using''). 

\shortparagraph{Influence of Dataset Annotation Uncertainty} As we have already seen in the previous sections, annotators found it difficult to annotate several edge cases, which in turn brings uncertainty in the final annotation. 
To investigate how such uncertainty influences model performance, we do a further ablation study to test the model performance on data with unanimous agreement (all annotators give the same label) (47\% on the novice test set, 44\% on the expert test set)\footnote{Since only the first batch of expert data contains more than two annotators, for this study we only report ablation for this expert-annotated dataset.} and data where some disagreement exists, although majority voting can be reached (we call it arguable data). 
For the latter, we consider as gold label either the majority vote or any label chosen by at least one annotator. The
results are shown in \cref{tab:ablation_uncertainty}.  

We notice
that: 1) models perform better on data with unanimous agreement than on arguable data
(15\%-32\%); 
2) given the difficulty of the annotation task, if we consider all annotators' labels as gold (Arguable, all annotations), we can see the model can improve (14\%-25\%); 3) by comparing the performance on the first batch of expert-annotated data and the 
novice-annotated data, our models achieve very similar performance on instances with unanimous agreement and also when considering all annotators' labels as gold.
In arguable cases with majority voting,  however, 
models trained on novice-annotated data cannot perform as well on expert test sets where experts cannot reach a unanimous agreement or where we do not consider all labels. 
This confirms the fact that
the disagreement among
annotators will
influence the model performance and roughly quantify the performance bottleneck 
resulted from using majority voting as the gold label.

%% file: past_papers/addiction/sections/related_work.tex
\section{Related Research}
\shortparagraph{Machine Learning for Substance Use}
Machine learning methods’ application to substance use research is growing~\citep{bharat2021big}. Several studies have attempted to predict substance use treatment completion among individuals with substance use disorders~\citep{gottlieb2022machine, acion2017use, hasan2021machine}. This study takes advantage of anonymous data to identify treatment needs among individuals who may not currently be in formal substance use treatment. Researchers have also used natural language processing to identify substance misuse in electronic health records~\citep{afshar2019natural, riddick2022natural} and to classify substances involved in drug overdose deaths~\citep{goodman2022development}. \citet{maclean2015forum77} collect user-level data on a social platform, Forum 77, to build a CRF model predicting three phases of drug use: using, withdrawing, and recovering. Our work is different in several aspects: 1) we propose an annotation scheme grounded in research
on addiction and substance use that defines behaviors such as Medical Use, Misuse, Addiction, Recovery, Relapse (and Not Using), that enable us to code self-disclosures of such behaviors using both expert and novice annotators; 2) we develop explanation-infused accurate models to identify self-disclosure at the post level. These two innovations will enable future research on using these models for a reliable global, user-level analysis across time.

\shortparagraph{Learning from Explanations}
There have been works focusing on
learning from human-annotated explanations.
\citet{wiegreffe2021measuring} investigates how free-form explanations and predicted labels are associated and use it as a property to evaluate the faithfulness of explanations. Different from that, our work focuses more on the utility of extractive span-level explanations as an additional source of supervision in a high-stakes domain and further shows how the quality of explanations impacts inference time results \citep{sun2022investigating}. 
Similar to our work, 
\citet{carton-etal-2022-learn} leverages extractive explanations and shows a consistent trend 
that using explanations can improve model performance in reading comprehension. Our work is most similar to \citet{huang-etal-2021-exploring}, who noticed that the quality of explanations could have a huge impact on model performance and explore the utility of extractive explanations, and to \citet{sun2022investigating}, who perform similar studies using free-form explanations.

\shortparagraph{Understanding the OUD continuum}
Scientists have explained how opioids produce changes in brain structure and function that promote and sustain addiction and contribute to relapse~\citep{koob2010neurocircuitry, abuse2016neurobiology}. Now recognized as a chronic but treatable disease of the brain, OUD is characterized by clinically significant impairments in health and social function and influenced by genetic, developmental, behavioral, social, and environmental factors~\citep{volkow2016neurobiologic}.
The HEALing Communities Study implemented the Opioid-overdose Reduction Continuum of Care Approach (ORCCA) to reduce opioid-overdose deaths across the OUD continuum~\citep{winhusen2020opioid}.
Taking advantage of self-disclosures on community-based social media, as this study aims to do, could lead
to the development of
interventions that better address risks associated with OUD.

%% file: past_papers/addiction/sections/conclusion.tex
\section{Conclusions}
We presented a novel task aimed to deepen our understanding of how people move across the OUD continuum: given a user's post in an opioid-related Reddit, predict whether it contains a self-disclosure of various phases of OUD.
We provided an annotation scheme grounded in research on addiction and substance use,
which enables us to code self-disclosures of such behaviors using both expert and novice annotators. Following the annotation scheme, we created a high-quality dataset annotated with class labels and text explanations. We presented several state-of-the-art explanation-infused models, showing they can achieve accurate results in identifying self-disclosures of use, misuse, addiction, recovery, and relapse. Accurate models will enable further research in this space by considering a global user-level analysis across time. Our error analysis showed that explanations could provide insights both into annotator disagreement and errors in model predictions. In addition, our findings shed light
on how annotation uncertainty impacts model performance.

%% file: past_papers/addiction/sections/limitations_and_ethics.tex
\section*{Limitations}
This study’s results are not without limitations. The
anonymity of Reddit users does not allow us to characterize the demographics or geographic extent of the study population.
Moreover, the current study looks at identifying self-disclosures at the message level without taking a global (user-level) and temporal view. In our future work, we plan to apply our models to study users' posts in opioid-related Reddits and observe their behavior over time.
In addition, we will work on improving our models to both predict a label and provide a textual explanation for the prediction.

\section*{Ethical Considerations}
For our data collection and annotation, we have obtained IRB approval. The source data comes from Reddit (r/opiates, r/OpiatesRecovery and r/Drugs), and is thus publicly available and anonymous. In addition, we preprocess the data to additionally remove any potentially identifiable information (see Section \ref{collection}).   All data is kept secure and online
userIDs are not associated to the posts.
For the expert annotation we compensated the experts with \$20 per hour, and the novice annotators with \$15 per hour. 

Our intention of developing datasets and models for predicting the stages of opioid use disorder is to help health professionals and/or social workers to both understand personal experiences of people across the opioid used disorder continuum and potentially to identify people
that might be at risk of overdose. The inclusion of explanations both in the annotation and in the prediction of our models could help the health professional better assess the models predictions. We emphasize that our models should be used with a human in the loop — for example a medical professional, or a social worker, who can look at the predicted labels and the explanations to decide whether or not they seem
sensible. 
We note that because most of our data were collected from Reddit, a website with a known overall
demographic skew (towards young, white, American men
), our conclusions about what explanations are associated with various OUD stages
cannot necessarily be applied to
broader groups of people. This might be particularly acute for vulnerable populations such as people with opioid use disorder (OUD). We hope that
this research stimulates more work by the research
community to consider and model ways in which
different groups self-disclose their experiences with OUD.

%% file: chapters/conclusion.tex
\section{Thesis Summary}

This thesis has investigated the fundamental disconnect between the surface-level competence of aligned Large Language Models and their underlying grounding capabilities. Through a systematic examination spanning situational understanding, generative dynamics, and dynamic control mechanisms, we have argued that current alignment paradigms prioritize form over substance---optimizing models to \textit{sound} aligned rather than to \textit{be} grounded. We have framed this challenge as one of \textbf{Grounded Alignment}.

\subsection{Part I: Situational Grounding Evaluation}

In the first part of this thesis, we evaluated the extent to which models can process and maintain situational information:

\begin{itemize}
    \item \textbf{SitTest} revealed that despite access to complete dialogue histories, ChatGPT fails to maintain coherent environment states over time, suffering from non-persistent in-context memory and susceptibility to hallucinated updates.

    \item \textbf{ReCode} demonstrated that code generation models rely on surface heuristics rather than deep syntactic understanding, with significant performance degradation under semantic-preserving perturbations.
\end{itemize}

\subsection{Part II: Generative Grounding Evaluation}

The second part evaluated the dynamics of generation and the model's self-understanding:

\begin{itemize}
    \item \textbf{LLM Probability Concentration} analysis, published in TMLR, revealed that alignment tuning sharply concentrates early probability mass, reducing the branching factor by a factor of 2--5 overall and up to an order of magnitude at the start of generation. This premature stylistic collapse explains why aligned models often lack output diversity.

    \item \textbf{Hindsight} showed that models often fail to understand their own generations. Without external feedback, self-reflection can reinforce rather than correct hallucinations, highlighting a lack of generative grounding.

    \item \textbf{Amortized Interpretability} provided efficient tools for diagnosing model behavior through stable Shapley Value estimation, enabling scalable analysis of token importance.
\end{itemize}

\subsection{Part III: Dynamic Control for Human Alignment}

The final part proposed interventions to align grounded models with human needs:

\begin{itemize}
    \item \textbf{AI Realtor} (CAIS 2026) demonstrated the critical role of context engineering. Given that models lack robust situational grounding, architecturally enforcing retrieved context before personalization is essential for achieving persuasive yet factually rigorous generation.

    \item \textbf{BACo} (ICML 2026) addressed the challenge of diverse value judgments. By decoupling exploration (base model) from structural guidance (aligned model) via token-level routing, we jointly improve diversity and quality and obtain a controllable trade-off.

    \item \textbf{Annealed Sampling} demonstrated that synchronizing exploration strategy with the natural cooling of the model's probability distribution improves sample efficiency in reinforcement learning contexts.

    \item \textbf{Addiction Support} showcased that model-generated rationalization provides a new communication interface. By assisting generation with explicit reasoning, we improve performance and trust in high-stakes health domains.
\end{itemize}

\section{Key Contributions}

This thesis makes several key contributions to the field:

\begin{enumerate}
    \item \textbf{Diagnostic Frameworks}: Novel testing environments and metrics for assessing situational understanding, code robustness, and model interpretability.

    \item \textbf{Theoretical Understanding}: The Branching Factor metric provides principled ways to understand and predict model behavior during generation.

    \item \textbf{Practical Interventions}: Annealed sampling, model collaboration, and architecturally-grounded generation offer actionable strategies for improving aligned model outputs.

    \item \textbf{Real-World Applications}: Demonstrations in code generation, real estate marketing, and healthcare show the practical impact of grounding-aware approaches.
\end{enumerate}

\section{Future Directions}

This work opens several promising research directions:

\paragraph{Scaling Situational Understanding.} As context windows continue to expand, developing more sophisticated methods for maintaining state coherence across extended interactions becomes increasingly critical. Future work should explore architectural innovations that provide persistent memory mechanisms.

\paragraph{Alignment-Preserving Diversity.} Our findings on the shrinking generative landscape suggest the need for alignment techniques that preserve output diversity while maintaining safety and helpfulness. Research into controlled entropy injection during fine-tuning could address this trade-off.

\paragraph{Grounding Beyond Text.} Extending our frameworks to multimodal settings---where models must ground language in visual, auditory, and interactive contexts---represents a natural and important extension of this work.

\paragraph{Personalized Grounding.} The success of AI Realtor in separating grounding from personalization suggests broader applications in domains requiring both factual accuracy and user-specific adaptation.

\section{Closing Remarks}

The remarkable capabilities of modern LLMs create an illusion of understanding that this thesis has sought to dispel. We have shown that beneath the fluent surface lies a fragile foundation---one that fails to track situational changes, collapses into stylistic monotony, and generates with false confidence.

Yet this critique is ultimately constructive. By precisely characterizing these limitations, we have also charted paths forward. The tools, theories, and interventions presented here offer a framework for the next generation of AI systems---agents that are not merely aligned to our preferences, but grounded in our reality.

As AI systems become more deeply integrated into critical domains---healthcare, finance, law, and beyond---the distinction between apparent and actual understanding becomes not merely academic but consequential. This thesis contributes to ensuring that the AI systems we deploy are worthy of the trust we place in them.

%% file: chapters/appendix.tex
\appendix

\let\originalappendix\appendix
\renewcommand{\appendix}{}

\chapter{SitTest: Testing Situational Understanding in ChatGPT}
\begingroup
\lstset{style=datalogstyle}
\input{past_papers/sittest/sections/appendix}
\endgroup

\chapter{ReCode: Robustness Evaluation of Code Generation Models}
\input{past_papers/recode/appendix}

\chapter{Hindsight: Testing Limits on Reflective Thinking in LLMs}
\begingroup
\lstset{
  basicstyle=\ttfamily,
  breaklines=true,
  postbreak=\mbox{\textcolor{red}{$\hookrightarrow$}\space},
  numbers=none,
  frame=none,
  xleftmargin=0pt
}
\input{past_papers/hindsight/sections/appendix}
\endgroup

\chapter{Amortized Interpretability: Efficient Shapley Values Estimation}
\input{past_papers/amortized_interpretability/sections/appendix}

\chapter{LLM Probability Concentration: How Alignment Shrinks the Generative Horizon}
\input{past_papers/branching_factor/appendix}

\input{past_papers/branching_factor/bf_self_narrowing_mvhn}
\input{past_papers/branching_factor/lexical_diversity}
\input{past_papers/branching_factor/data_contamination}

\chapter{Annealed Sampling for Verifiable Reinforcement Learning}
\input{past_papers/annealed_sampling/tex/appendix}

\chapter{Optimizing Diversity and Quality through Base-Aligned Model Collaboration}
\input{past_papers/base_aligned_collaboration/sections/appendix/main}

\chapter{AI Realtor: Towards Grounded Persuasive Language Generation for Automated Copywriting}
{
\renewcommand{\agentname}{%
  \raisebox{-0.1\fontcharht\font`A}{%
    \includegraphics[height=1.2\fontcharht\font`A]{img001.png}%
}~\texttt{AI Realtor}\xspace}
\lstset{
  basicstyle=\ttfamily,
  breaklines=true,
  postbreak=\mbox{\textcolor{red}{$\hookrightarrow$}\space},
}

\input{past_papers/ai_realtor/appendix}

\input{past_papers/ai_realtor/statement}
}

\chapter{Rationale-Grounded Addiction Support: Identifying OUD Phases}
\input{past_papers/addiction/sections/appendix}

\let\appendix\originalappendix

%% file: past_papers/sittest/sections/appendix.tex
\section{Distractor on Action}
\label{app:action_distractor}

    In addition to instruction understanding, the model should also learn to correctly understand and select key information every time a new step is taken. 
    To examine the robustness of this capability, in addition to the inclusion of normal step descriptions, we also include distractor conditions \citep{shi2023large}, in which we add one randomly selected sentence from the Brown Corpus~\citep{kuvcera1967computational} as a distractor. An example of applying such distractor is shown in \cref{lst: distracted_action example}. The experiment results are shown in \cref{tab:10-box-action-testing-chatgpt}.
    \begin{lstlisting}[caption={Action Distractor Example}, label={lst: distracted_action example}]
Step-1: Open jqC-3 and retrieve bsS-2. @\textcolor{red}{It is a nice day!}@
Question: NvSWxzvJb(jqC-0)=False...
\end{lstlisting}
\vspace{8pt}

\begin{table}[h!]
\centering
\resizebox{0.45\textwidth}{!}{%
\begin{tabular}{@{}lcc@{}}
\toprule
\multirow{2}{*}{Normal Instruction} & \multicolumn{2}{c}{Step-EM / State-EM} \\ \cmidrule(l){2-3} 
                      & 2-shot     & 5-shot     \\ \midrule
NL Functor + NL Argument    &   $22\%$ / $92\%$     &   $36\%$ / $95\%$    \\
Synthetic Functor + Synthetic Argument    &   $19\%$ / $92\%$     &   $52\%$ / $96\%$  \\
\midrule
\multirow{1}{*}{Actions w/ Distractor} & \multicolumn{2}{c}{} \\ \midrule
NL Functor + NL Argument    &   $18\%$ / $91\%$     &   $44\%$ / $96\%$    \\
Synthetic Functor + Synthetic Argument    &   $34\%$ / $93\%$     &   $50\%$ / $96\%$  \\
\bottomrule
\end{tabular}%
}
\caption{Action understanding experiment results for 10-box environment on ChatGPT. Metrics here are shown in the format of "Step-EM/State-EM". We use 50 samples with various number of steps for experiments. }
\label{tab:10-box-action-testing-chatgpt}
\end{table}

From \cref{tab:10-box-action-testing-chatgpt}, we can see after adding distractors, the State-EM performance does not degrade as much as in counter-intuitive instructions, though Step-EM performance degrades a bit (but not consistently). These findings hold both in NL and synthetic languages. This suggests that ChatGPT does have the ability to understand the interaction happened at each step and can pick out useful information.

We also see that when adding distractors on actions, within the 2-shot condition, Step-EM in synthetic language environment is better than the one in NL environment. 
This is probably because the usage of synthetic language helps the model better distinguish useful information as they look very different from synthetic languages. However, when there are more in-context samples, the model will gradually learn to extract useful information at each step and the usage of synthetic language does not help that much. 

    \begin{table*}[h!]
\centering
\small
\resizebox{0.7\textwidth}{!}{%
\begin{tabular}{@{}lccc@{}}
\toprule
\multirow{2}{*}{Normal Instruction} & \multicolumn{3}{c}{Step-EM / State-EM} \\ \cmidrule(l){2-4} 
                      & 2-shot     & 3-shot     & 5-shot     \\ \midrule
NL Functor + NL Argument    &   $22\% / 92\%$   &   $34\% / 93\%$   &   $36\% / 95\%$    \\
Synthetic Functor + NL Argument    &   $14\% / 90\%$   &   $20\% / 93\%$   &   $30\% / 94\%$    \\
NL Functor + Synthetic Argument    &   $44\% / 95\%$   &   $30\% / 94\%$   &   $74\% / 98\%$    \\
Synthetic Functor + Synthetic Argument    &   $19\% / 92\%$   &   $22\% / 93\%$   &   $52\% / 96\%$    \\
\midrule
\multirow{1}{*}{Counter-Intuitive Instruction (On NL)} & \multicolumn{3}{c}{} \\ 
    \midrule
NL Functor + NL Argument    &   $10\% / 77\%$   &   $6\% / 75\%$   &   $0\% / 84\%$    \\
Synthetic Functor + NL Argument    &   $18\% / 88\%$   &   $10\% / 88\%$   &   $14\% / 90\%$    \\
NL Functor + Synthetic Argument    &   $20\% / 81\%$   &   $10\% / 78\%$   &   $8\% / 88\%$    \\
Synthetic Functor + Synthetic Argument    &   $13\% / 89\%$   &   $20\% / 90\%$   &   $12\% / 90\%$    \\ \midrule
\multirow{1}{*}{Counter-Intuitive Instruction (Truth Values Switching)} & \multicolumn{3}{c}{} \\ \midrule
NL Functor + NL Argument    &   $6\% / 72\%$   &   $2\% / 69\%$   &   $2\% / 79\%$    \\
Synthetic Functor + NL Argument    &   $10\% / 85\%$   &   $10\% / 84\%$   &   $6\% / 87\%$    \\
NL Functor + Synthetic Argument    &   $8\% / 71\%$   &   $8\% / 73\%$   &   $0\% / 83\%$    \\
Synthetic Functor + Synthetic Argument    &   $19\% / 85\%$   &   $14\% / 87\%$   &   $12\% / 89\%$    \\
\bottomrule
\end{tabular}%
}
\caption{Experiment results on ChatGPT  Robustness check for state tracking in 10-box environment. Metrics here are presented in the format of ``Step-EM / State-EM''. We use 50 samples with various number of steps for experiments. 
}
\label{tab:10-box-instruction-testing-chatgpt-full}
\end{table*}

\section{Partial Usage of Synthetic Languages}
\label{app:partial_usage_sl}
We can do a more fine-grained usage of synthetic langauges -- we can choose to apply only on functors (e.g., ``Opened'') or arguments (e.g., ``BOX-1''). The full set of experiment results is shown in \cref{tab:10-box-instruction-testing-chatgpt-full}. We notice that the results are very similar to the full usage of synthetic languages so to save space, we move the results in Appendix. 

\section{Interaction with OpenAI API and Prompt Format}
\label{app:interaction_prompt_format}
According to the official OpenAI cookbook,\footnote{\url{https://github.com/openai/openai-cookbook/blob/main/examples/How_to_format_inputs_to_ChatGPT_models.ipynb}} to send requests to ChatGPT, it is advised to first add a ``system message'' to ChatGPT (We use ``You are a helpful assistant'' as this is one of the most used system messages). Then there are two typical ways \footnote{As of the time the project is initialized. OpenAI may change a bit in the documentation and examples in the GitHub repository.} to send the prompt, if it contains in-context samples:

    \paragraph{Traditional Format} Just put all your prompt contents in one-round interaction (as in prompting GPT-3-Davinci series models), as shown in \cref{lst: traditional_api}:
    \begin{lstlisting}[caption={Traditional Input Format}, label={lst: traditional_api}]
response = openai.ChatCompletion.create(
    model=...,
    messages=[
        {"role": "system", "content": "You are a helpful assistant."},
        {"role": "user", "content": "[Instruction + In-Context Samples]"},
    ],
    ...,
)
\end{lstlisting}
\vspace{8pt}

\paragraph{Faked Multi-Round Format} Another way is to synthesize a fake multi-round conversation and pretend the answers for in-context samples are generated by ChatGPT, as shown in \cref{lst: faked_multi_round_api}:
    \begin{lstlisting}[caption={Faked Multi-Round Input Format}, label={lst: faked_multi_round_api}]
response = openai.ChatCompletion.create(
    model=...,
    messages=[
        {"role": "system", "content": "You are a helpful assistant."},
        {"role": "user", "content": "[Instruction]"},
        {"role": "user", "content": "[Sample-1-Input]"},
        {"role": "assistant", "content": "[Sample-1-Ground-Truth-Answer]"},
        {"role": "user", "content": "[Sample-2-Input]"},
        {"role": "assistant", "content": "[Sample-2-Ground-Truth-Answer]"},
        ...
    ],
    ...,
)
\end{lstlisting}
\vspace{8pt}

We do a prior study on experimenting with these two formats and find they give pretty similar Step-EM and State-EM performance in fully NL and fully Synthetic Language settings (usually the performance difference is less than $5\%$). But faked multi-round format would insert many more tokens (leading to a higher cost) in the requests and can make the OpenAI server reject to respond, leading to low response rates. Therefore, to save budgets, in the main text of the paper, we only report the results using the traditional format. Another reason is that to see how much our findings on ChatGPT can generalize to other models, we also replicate some of our experiments when comparing with other models in \cref{sec:external_comparison}. Some models there may not support faked multi-round input format. 

For response parsing, we just follow the instruction to parse the JSON-style response described in the same notebook and obtain the model output. For decoding parameters, we just follow default OpenAI API setting. 
   \begin{table}[h!]
\centering
\resizebox{0.45\textwidth}{!}{%
\begin{tabular}{@{}lcc@{}}
\toprule
\multirow{2}{*}{Normal Instruction} & \multicolumn{2}{c}{State-EM, 2-shot} \\ \cmidrule(l){2-3} 
                      & ChatGPT     & Text-Davinci-003     \\ \midrule
NL Functor + NL Argument    &   $92\%$   &   $96\%$      \\
Synthetic Functor + NL Argument    &   $90\%$   & $94\%$    \\
NL Functor + Synthetic Argument    &   $95\%$   &   $94\%$     \\
Synthetic Functor + Synthetic Argument    &   $92\%$ &    $88\%$    \\
\bottomrule
\end{tabular}
}
\caption{Experiment Results Comparing ChatGPT and Davinci-003. Metrics here are State-EM. We use 50 samples with 2-shot setting and various number of steps for experiments. }
\label{tab:chat-gpt-davinci}
\end{table}
\section{Post-Processing to Extract Answers from OpenAI API Responses}
\label{app:post-processing-regex}
We note that ChatGPT can return answers not strictly follow the given format. So we do a light-weight postprocessing mainly using regular expressions to make sure we parse ChatGPT results appropriately. Specifically, we first clean out the space and newline characters at the beginning and end of the answer. We then use the following regular expression to match all logical statements and extract groups of functors, arguments and truth values:
\begin{lstlisting}[]
    ([a-zA-Z0-9]+)\(([a-zA-Z0-9]+-\d)\)=(True|true|False|false)
\end{lstlisting}
\section{Comparison with other models}
\label{sec:external_comparison}
 To clarify whether our observations can be generalized to other popular models, we choose one strong open-sourced 
 competitor chatbot (according to ChatArena~\citep{zheng2023judging}): Vicuna-13b---and we also compare against the performance of GPT-3-davinci-003 (it is believed that ChatGPT is a variant fine-tuned over GPT-3). 

 \paragraph{Comparison between Vicuna-13B} 
On ChatArena~\citep{zheng2023judging}, Vicuna-13B~\citep{vicuna2023} is voted to be the most close-to-ChatGPT chat models based on LLAMA~\citep{touvron2023llama_2}. We feed the same input to Vicuna and find the outputs are hard to parse and often incomplete. Even when there are no counter-intuitive instructions and we give 5 in-context samples, the best observed State-EM (in SL environment) is only $32\%$. 
Compared with ChatGPT (94\%), there seems a big gap on the capability of situational understanding for open-source models. 

\paragraph{Comparison between GPT-3.5 and ChatGPT}
Due to the budget limit and context length limit of GPT-3.5, we only compare ChatGPT performance with GPT-3.5 (Davinci-003) on 2-shot. The experiment results are in \cref{tab:chat-gpt-davinci}. We can see Davinci-003 achieves a similar performance as ChatGPT. 

%% file: past_papers/recode/appendix.tex
\section{Transformation Details and Qualitative Examples}
\label{appd: transformation}

In this section, we give detailed descriptions and settings for each type of perturbations that are included in our ReCode benchmark with qualitative examples for illustrations.

\subsection{Natural Transformations on Docstrings}

For natural transformations on docstrings, we aim to perturb the docstrings to their variances that preserve the semantics and also appear natural in practice. Specifically, we will first extract and perturb the docstrings with the following natural transformations in each prompt, and then attach their perturbed versions to the prompt. To preserve semantics for the code generation task prompts, we extract a blacklist of program keywords using tree-sitter as discussed in~\cref{subsec: text} that are excluded from perturbations. We extend most transformations from NL-Augmenter~\citep{dhole2021nl}, a standard library designed for data augmentation and robustness evaluation on text. We list some qualitative examples in \cref{tab: docstring_example}.

\paragraph{BackTranslation.} BackTranslation paraphrases the docstrings by translating them to another language (in this case, German) and then back to English. It is a common method for data augmentation in generating sentence variances with the same semantics~\citep{li2019improving, sugiyama2019data}. Overall, it can reliably generate high quality perturbed docstrings. We use the default implementation in NL-Augmenter~\citep{dhole2021nl}. BackTranslation contains no randomness in transformations.

\paragraph{ButterFingers.} ButterFingers transformation randomly selects characters of the docstrings and perturbs each of them to a random subset of similar characters, it is from~\citep{dhole2021nl} and is also used in~\citep{mille2021automatic}. Since this transformation tends to introduce character-level typos, we set randomness for perturbing each character to be low as 0.05 for naturalness consideration.

\paragraph{ChangeCharCase.}  ChangeCharCase transformation randomly changes the selected characters to upper case in the docstrings. We use the default probability 0.35 where majority annotators vote 0.5 for naturalness in the setting of \cref{subsec: human_eval}.

\paragraph{EnglishInflectionalVariation.}
This transformation randomly selects words in the docstring and change them to a random inflection variance. This can be from plural to singular (or vice versa) for nouns and tense changes for verbs. To maintain naturalness, the perturbation is constrained to be the same Part of Speech (POS) tag in the Penn Treebank~\citep{marcus-etal-1993-building}. 

\paragraph{SwapCharacters.}
This transformation randomly selects pairs of adjacent characters in the docstring and swap them. This represents a common type of typos by humans. To ensure naturalness, we set the probability as 0.05 for making the swap.

\paragraph{SynonymInsertion.}
This transformation randomly select words in the docstrings and inserts their synonyms in WordNet~\citep{miller1995wordnet}. Punctuations and stopwords are excluded. We set the probability to be 0.35 considering low success rate after keywords filtering.

\paragraph{SynonymSubstitution.}
This transformation randomly selects words in the docstring and replaces each one with a synonym from WordNet~\citep{miller1995wordnet}. Similar to SynonymInsertion, we set the probability as 0.35 to balance naturalness and perturbation success rates.

\paragraph{TenseTransformationPast.}
This is a deterministic transformation that converts sentences in the docstring to past tense.

\paragraph{TenseTransformationFuture.}
This is a deterministic transformation that converts sentences in the docstring to future tense.

\paragraph{Whitespace.}
This transformation inserts or deletes a single white space at randomly selected locations in the docstring.
This represents a common type of typos by humans. Folowing NL-Augmenter, we use probability 0.1 for adding whitespaces and 0.05 for removing whitespaces.

\subsection{Natural Transformations on Function Names}

These transformations modify the name of the target function to generate.
Any references to the function name in the prompt, e.g., in docstring, are also modified to maintain consistency. Qualitative examples can be found in \cref{tab: func_examples}.

\paragraph{CamelCase.}
A function name is often composed of multiple words.
If the original function name concatenates the words in camel-case style, this transformation changes it to snake-case, and vice versa. This transformation is deterministic.

\paragraph{ButterFingers.} ButterFingers transformation randomly selects characters of the docstrings and perturbs each of them to a random subset of similar characters, it is from~\citep{dhole2021nl} and is also used in~\citep{mille2021automatic}. Since this transformation tends to introduce character-level typos, we set randomness for perturbing each character to be low as 0.05 for naturalness consideration.

\paragraph{SwapCharacters.}
This transformation randomly selects pairs of adjacent characters in the function name and swap each pair.
This represents a common type of typos by humans.
To control naturalness, we set the probability to be 0.05, same setting as the docstring perturbations.

\paragraph{ChangeCharCase.}  ChangeCharCase transformation randomly changes the selected characters to upper case in the docstrings. We use the default probability 0.35 where majority annotators vote 0.5 for naturalness in the setting of \cref{subsec: human_eval}.

\paragraph{InflectionalVariation.}
This transformation randomly selects words in the function name and applies a random inflection on them.
This can be from plural to singular (or vice versa) for nouns and tense change for verbs. To control naturalness, the perturbation is constrained to be the same Part of Speech (POS) tag in the Penn Treebank~\citep{marcus-etal-1993-building}. 

\paragraph{SynonymSubstitution.}
This transformation randomly selects words in the docstring and replaces each one with a synonym from WordNet~\citep{miller1995wordnet}. Similar to SynonymInsertion, we set the probability as 0.35 to balance naturalness and perturbation success rates.

\subsection{Natural Transformations on Code Syntax}

\begin{figure}[!hbt]
    \centering
    \includegraphics[width=0.6\linewidth]{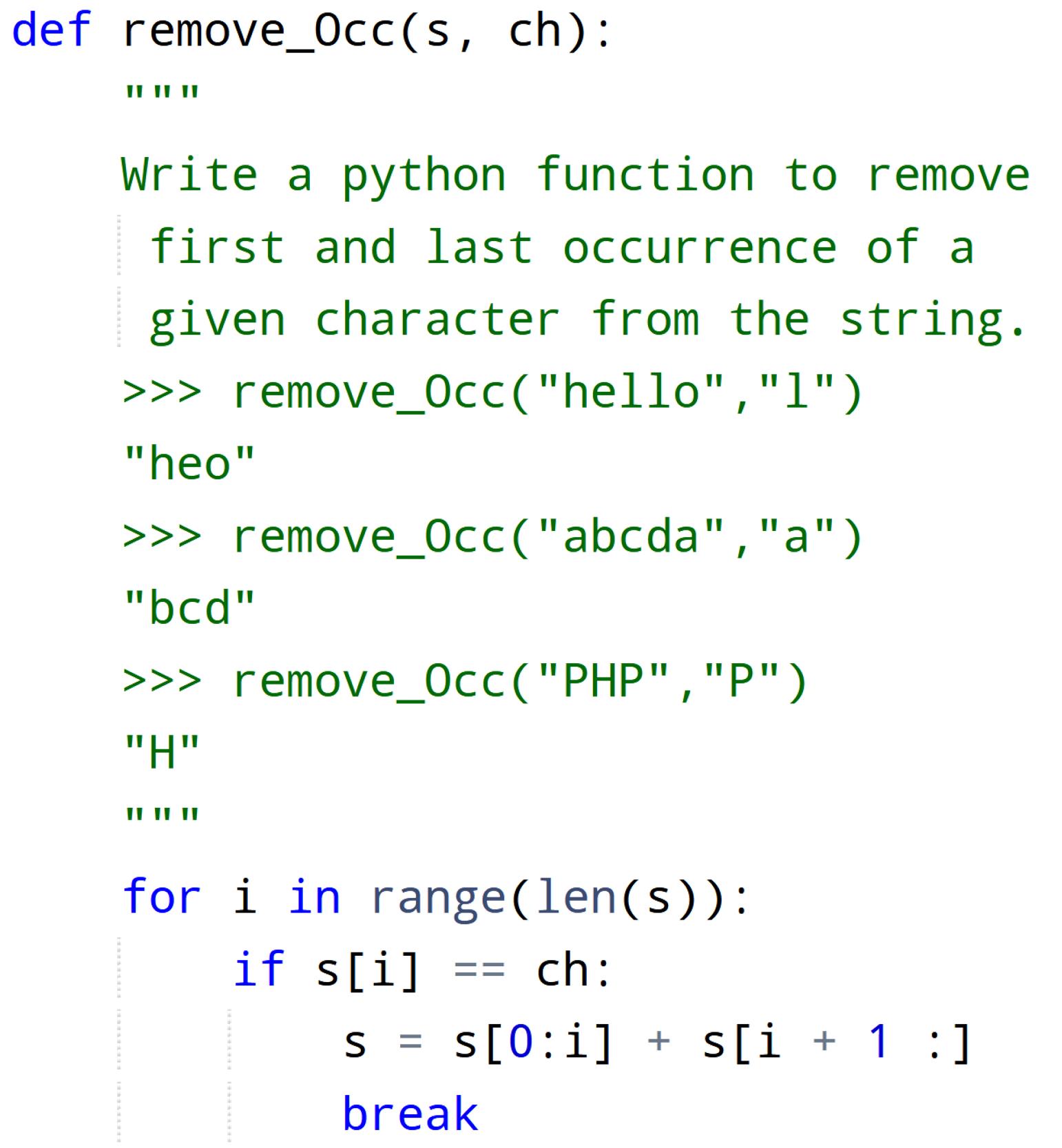}
    \caption{An baseline example of the prompt with partial code derived from original MBPP prompt for robustness evaluation on code.}
    \label{fig: appd_partial}
\end{figure}

These transformations modify the code content in the prompt. We derived function completion tasks with half the code from the canonical solutions such that the following code transformations and robustness evaluation can be performed. To guarantee fair comparisons to the nominal baseline, we make sure that we have the same block of code before and after code perturbations. In the following part we show qualitative examples on the same MBPP sample baseline (~\cref{fig: appd_partial}).

\paragraph{DeadCodeInserter.}
This transformation inserts a block of useless code at a random location.
The added block can be a loop of zero iteration or an if condition that is always false.
The code content inside the dummy loop or if condition is randomly selected from the adjacent code statements with limited tree-sitter node sizes.

\begin{figure}[!hbt]
    \centering
    \includegraphics[width=0.6\linewidth]{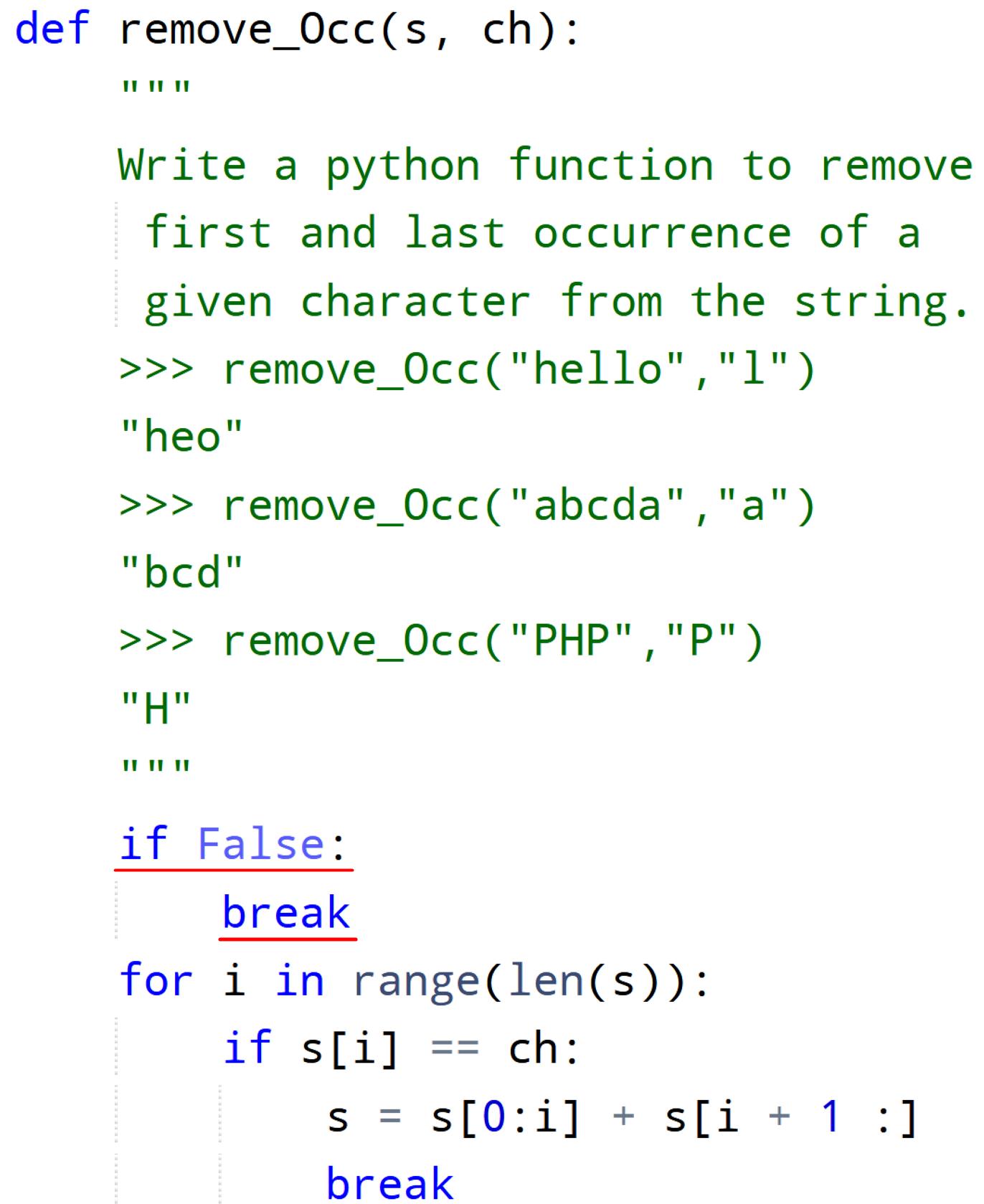}
    \caption{An example of the DeadCodeInserter perturbation.}
    \label{fig: deadcodeinserter}
\end{figure}

\paragraph{For-While Switch.}
This transformation randomly selects a for-loop or while-loop in the prompt and transforms it to its equivalent counterpart.
\begin{figure}[!hbt]
    \centering
    \includegraphics[width=0.6\linewidth]{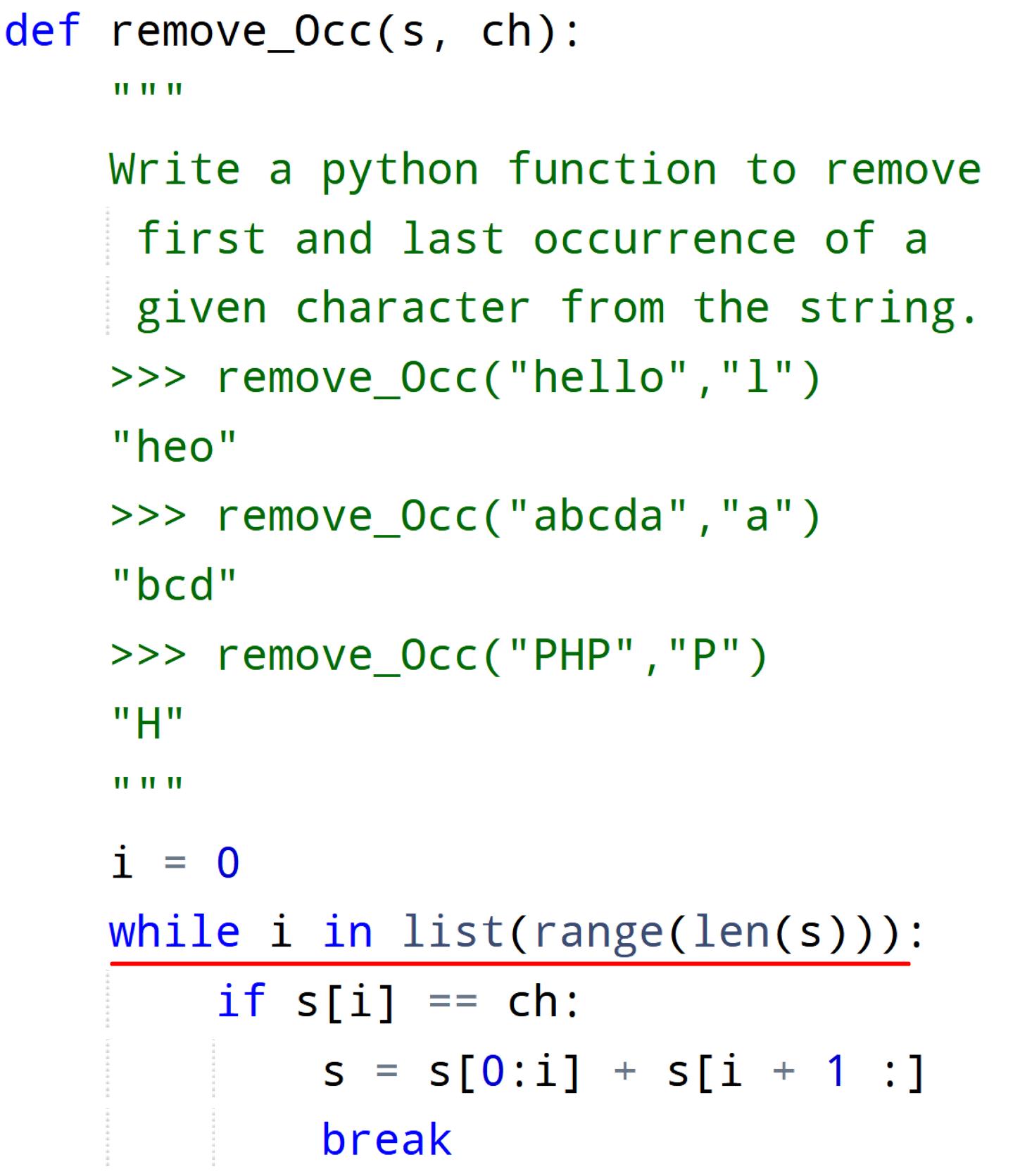}
    \caption{An example of the For-While Switch perturbation.}
    \label{fig: forwhile}
\end{figure}

\paragraph{OperandSwap.}
This transformation randomly selects a binary logical operation, swaps the two operands, and modifies the operator if necessary to maintain semantic equivalence.
\begin{figure}[!hbt]
    \centering
    \includegraphics[width=0.6\linewidth]{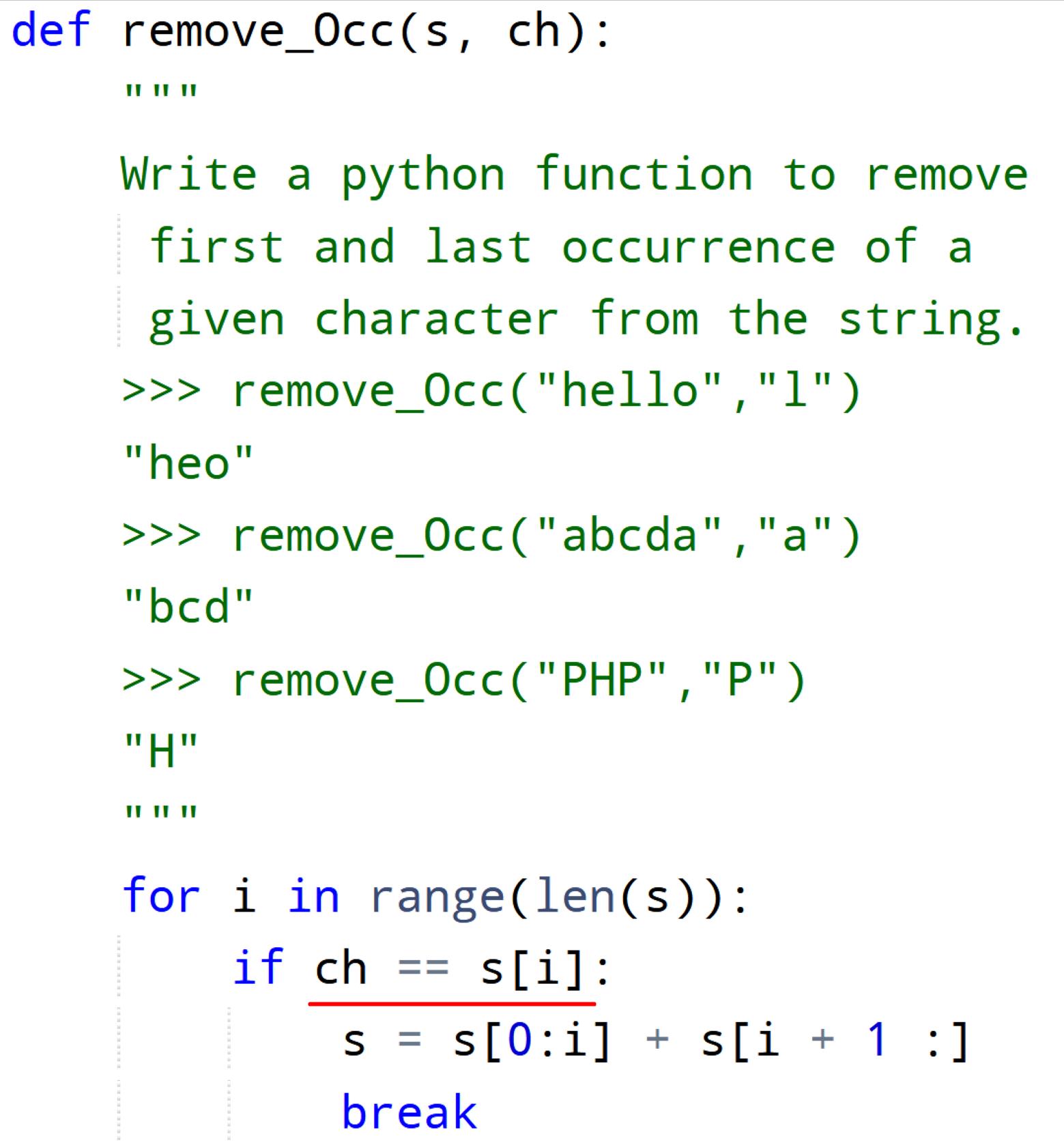}
    \caption{An example of the OperandSwap perturbation.}
    \label{fig: oprandswap}
\end{figure}

\paragraph{VarRenamerCB.}
This transformation selects the most frequently referenced variable name in the partial code and replaces it throughout the prompt with a new name obtained by CodeBERT~\citep{feng2020codebert}.
Specifically, we replace all occurrence of the variable name with a mask token, and then run CodeBERT inference to obtain candidate names at each location, where each candidate name comes with a probability score.
We pick the candidate name with the highest aggregated score across locations.
This transformation is inspired by~\citep{jha2022codeattack, li-etal-2020-bert-attack}.
\begin{figure}[!hbt]
    \centering
    \includegraphics[width=0.75\linewidth]{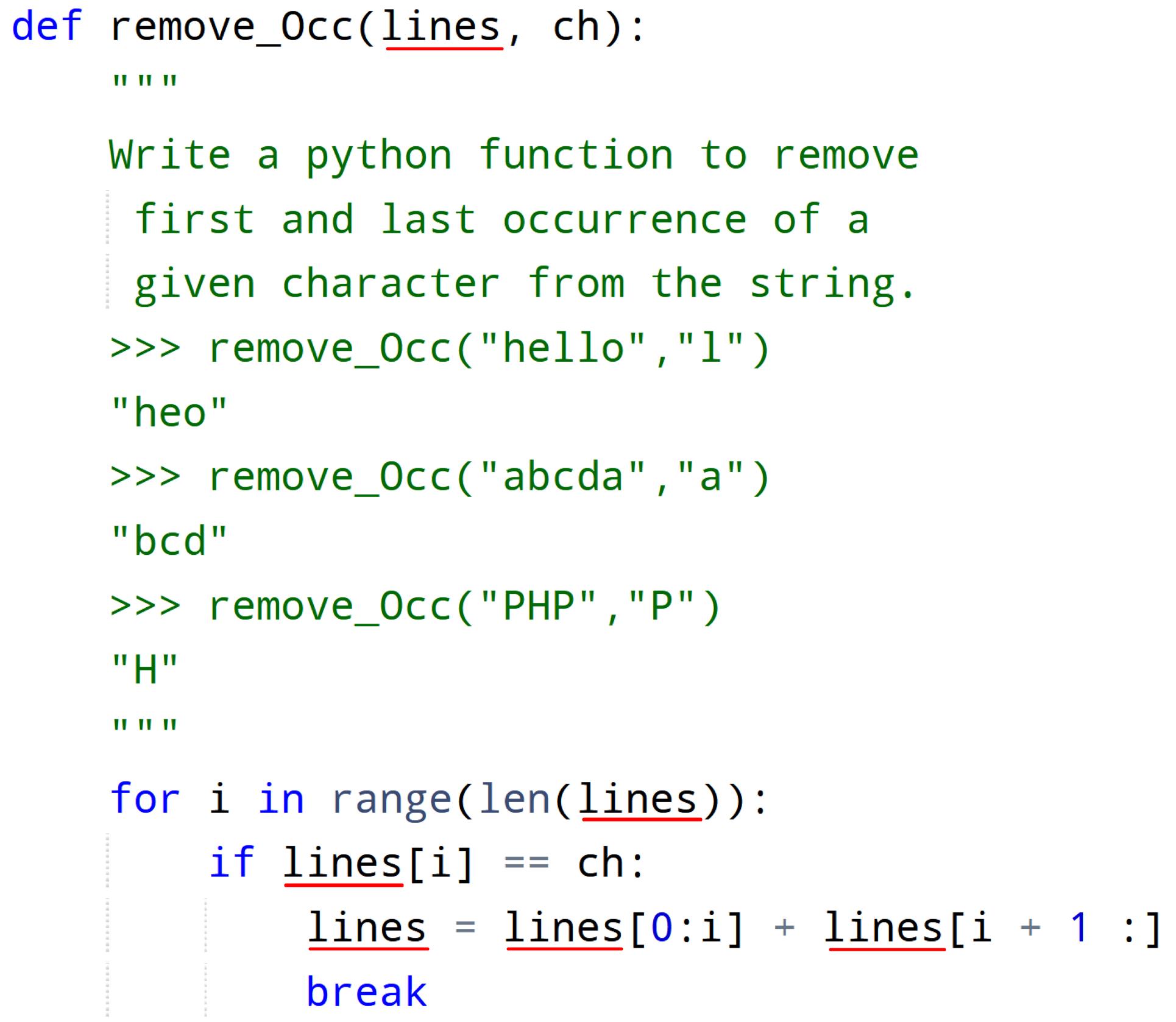}
    \caption{An example of the VarRenamerCB perturbation.}
    \label{fig: renamecb}
\end{figure}

\paragraph{VarRenamerNaive.}
This transformation selects the most frequently referenced variable name in the partial code and replaces it with "VAR\_0".
This is the original implementation in the NatGen package.
This transformation is deterministic.
\begin{figure}[!hbt]
    \centering
    \includegraphics[width=0.6\linewidth]{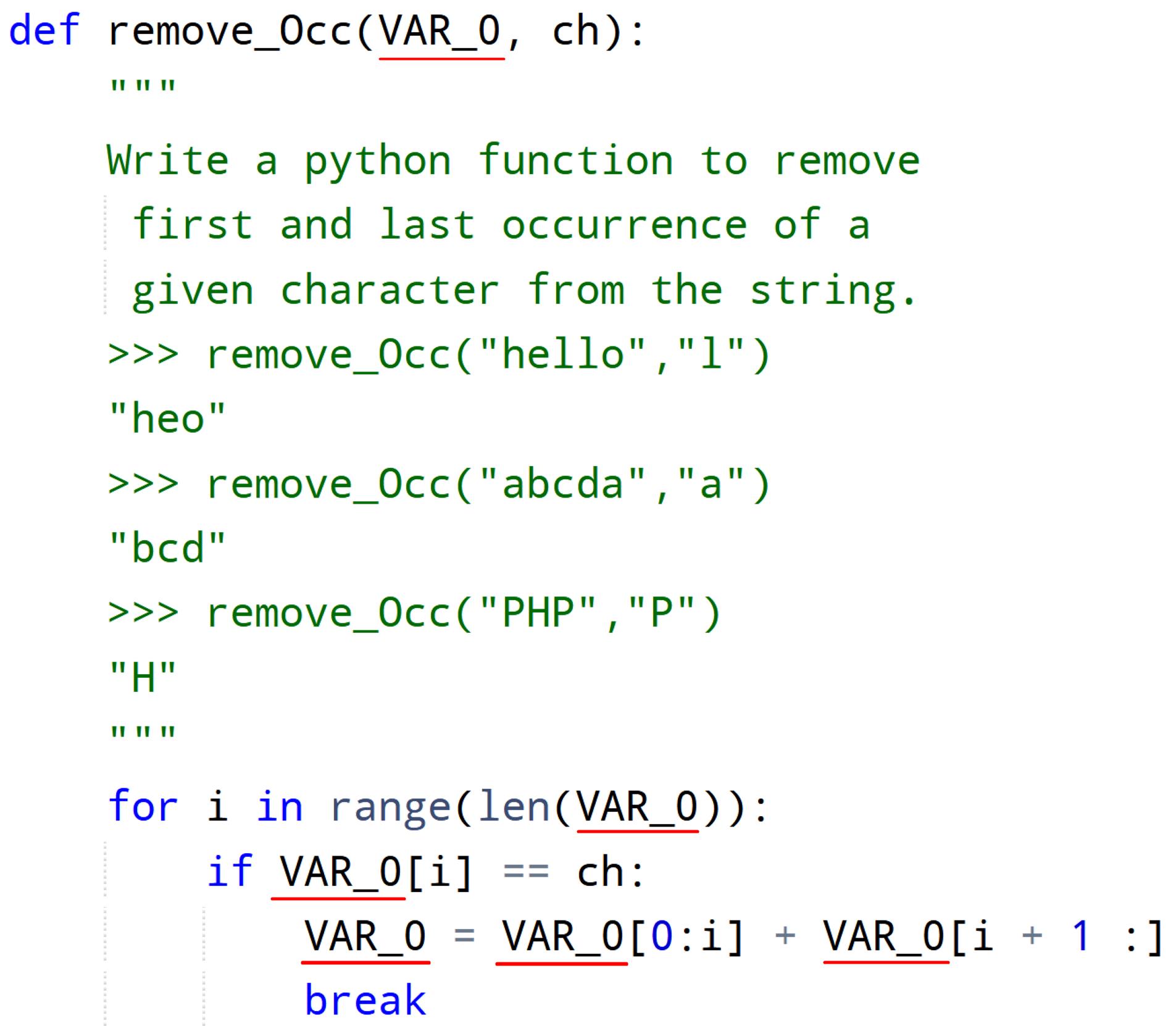}
    \caption{An example of the VarRenamerNaive perturbation.}
    \label{fig: renamenaive}
\end{figure}

\paragraph{VarRenamerRN.}
This transformation selects the most frequently referenced variable name in the partial code and replaces it with a random string with half alphabetic and half numeric characters.
\begin{figure}[!hbt]
    \centering
    \includegraphics[width=0.6\linewidth]{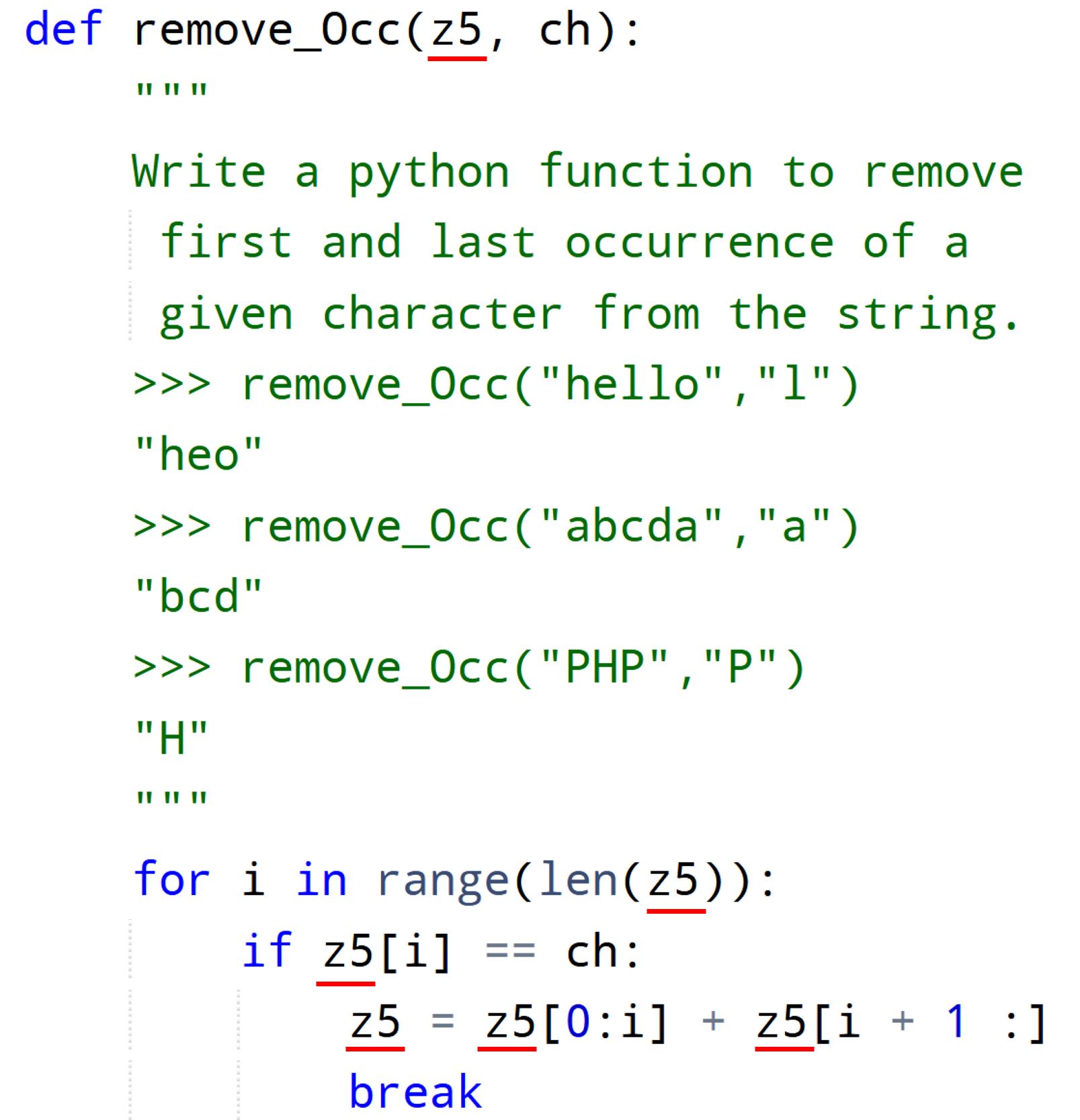}
    \caption{An example of the VarRenamerRN perturbation.}
    \label{fig: renamern}
\end{figure}

\subsection{Natural Transformations on Code Format}

\paragraph{Tab-Indent.}
This transformation replaces any space indents with tabs or replaces tabs with 4 spaces for indent-sensitive languages like Python. This transformation is deterministic.
\begin{figure}[!hbt]
    \centering
    \includegraphics[width=0.6\linewidth]{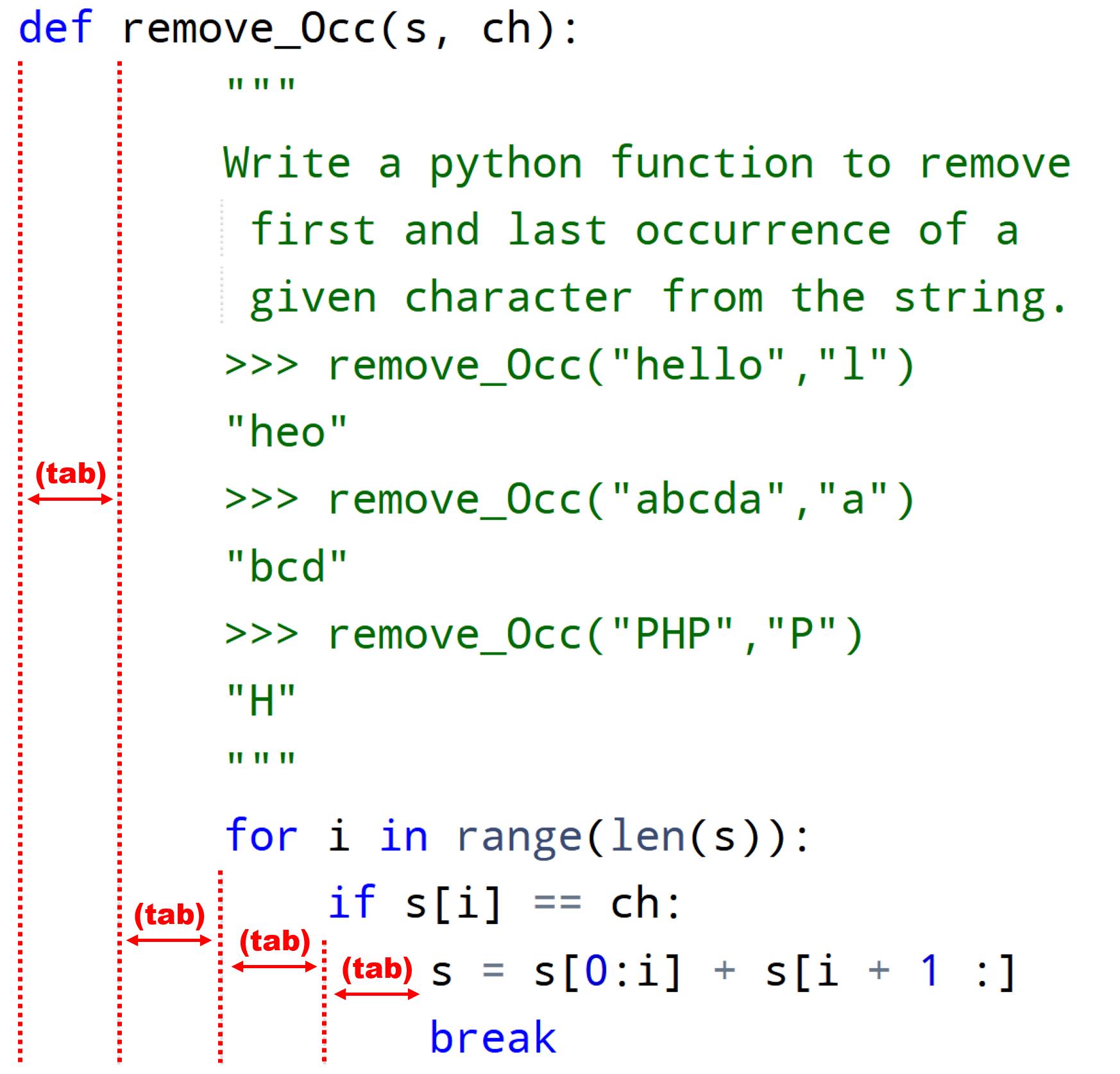}
    \caption{An example of the Tab-Indent perturbation.}
    \label{fig: tab}
\end{figure}

\paragraph{Line Split.}
This transformation splits the longest line in the partial code into two lines.
This transformation is deterministic.
\begin{figure}[!hbt]
    \centering
    \includegraphics[width=0.6\linewidth]{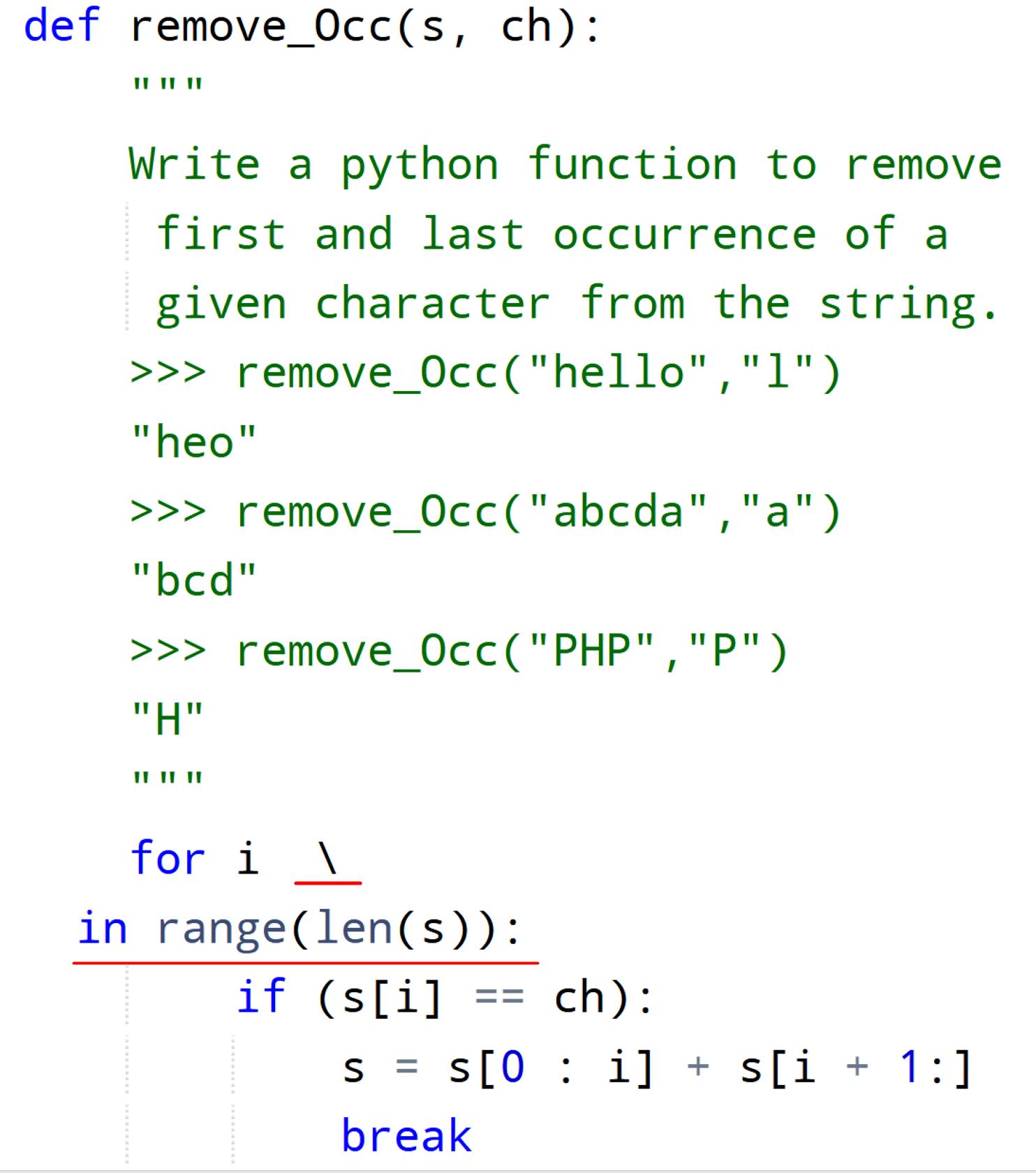}
    \caption{An example of the Line Split perturbation.}
    \label{fig: linesplit}
\end{figure}

\paragraph{Doc2Comments.}
This transformation changes the style of the documentation in the prompt.
For Python, it converts docstring (e.g., \texttt{""" docstring """}) to commented lines (e.g., \texttt{\# docstring}) and vice versa.
For Java, it converts comments in the format of \texttt{/* docstring */} to \texttt{// docstring} and vice versa.
This transformation is deterministic.
\begin{figure}[!hbt]
    \centering
    \includegraphics[width=0.6\linewidth]{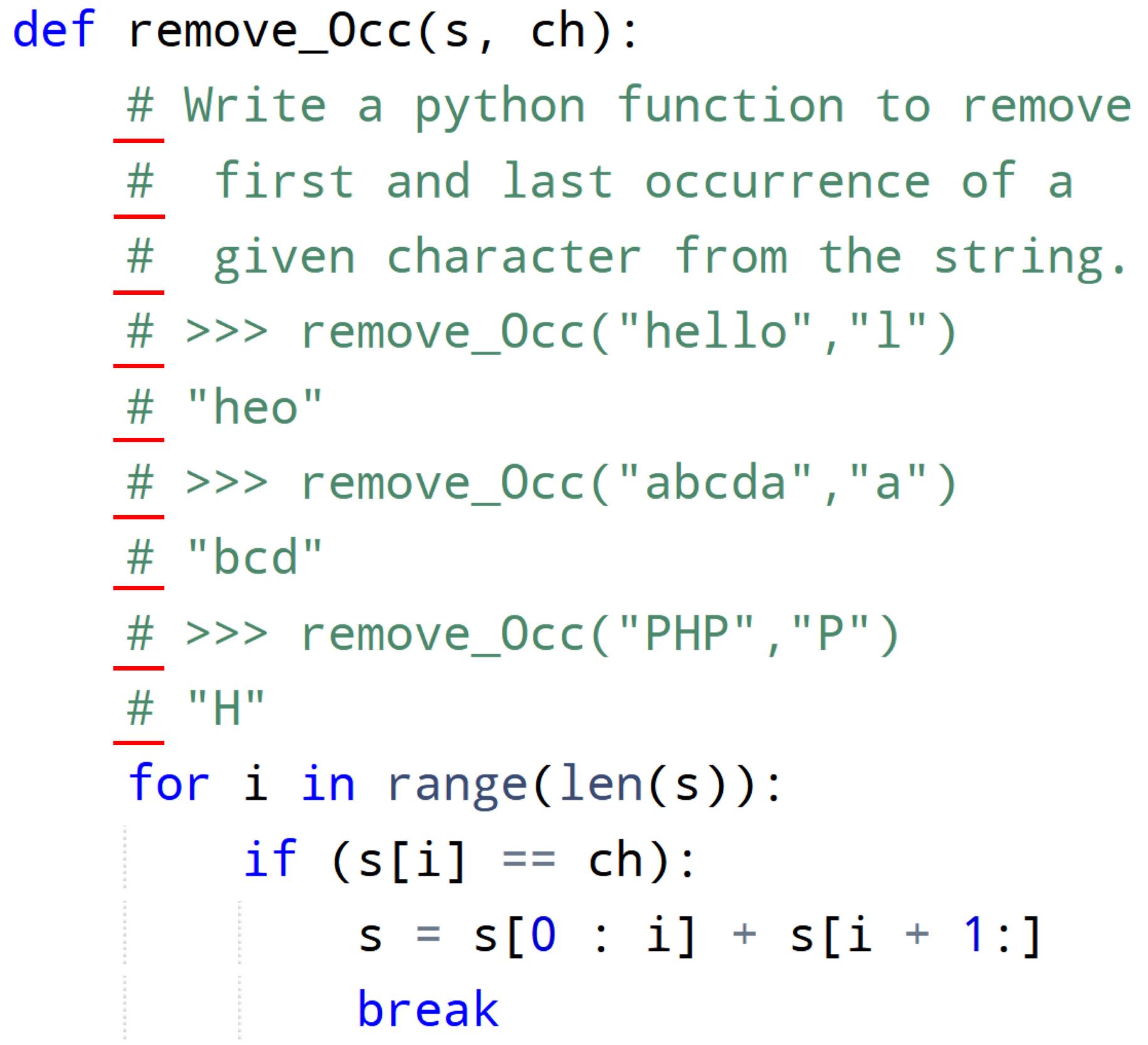}
    \caption{An example of the Doc2Comments perturbation.}
    \label{fig: doc2com}
\end{figure}

\paragraph{NewlineRandom.}
This transformation inserts empty lines at randomly selected positions.
\begin{figure}[!hbt]
    \centering
    \includegraphics[width=0.6\linewidth]{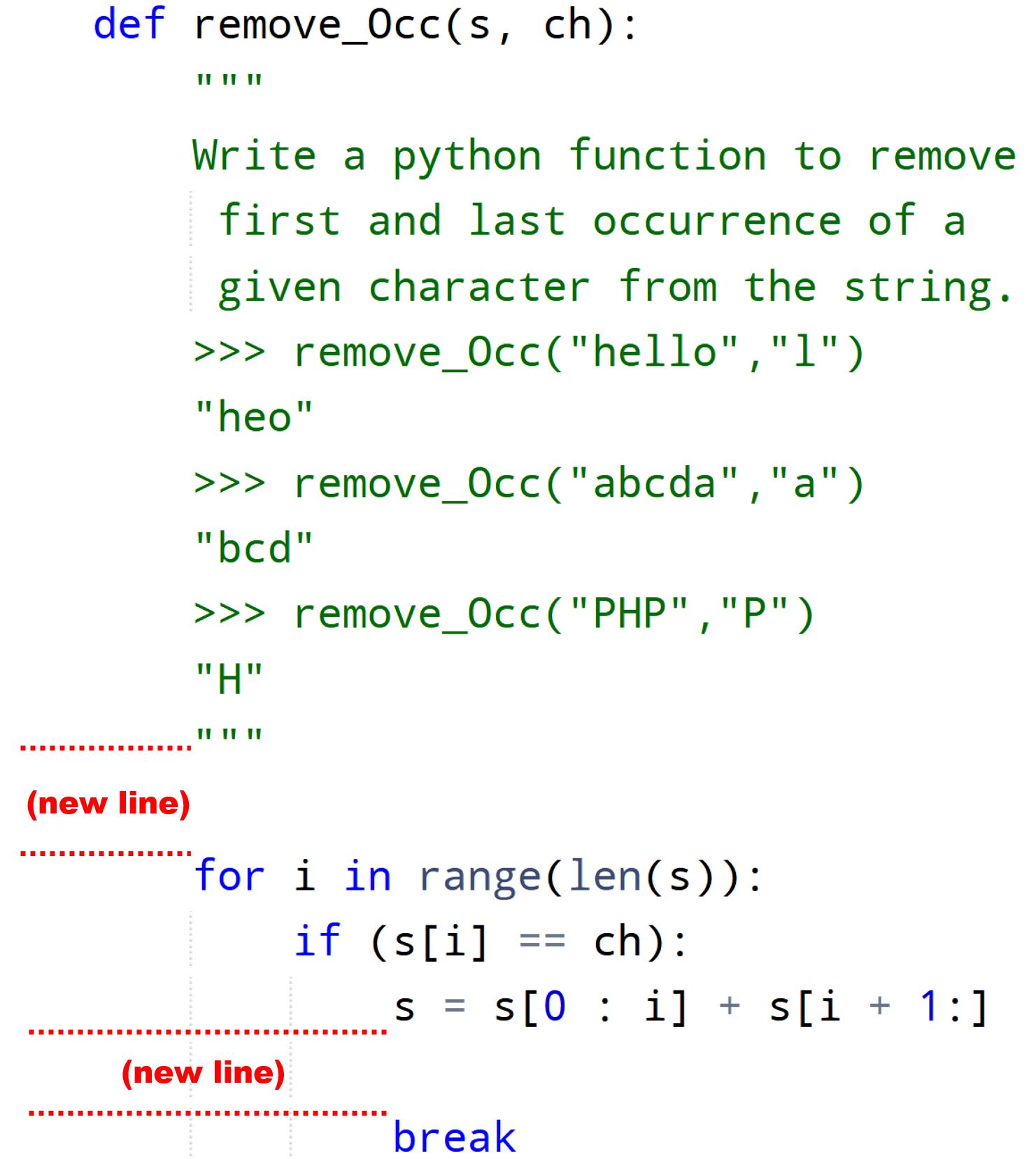}
    \caption{An example of the NewlineRandom perturbation.}
    \label{fig: NewlineRandom}
\end{figure}

\paragraph{NewlineAfterCode.}
This transformation inserts an empty line at the end of the prompt.
This transformation is deterministic.
\begin{figure}[!hbt]
    \centering
    \includegraphics[width=0.6\linewidth]{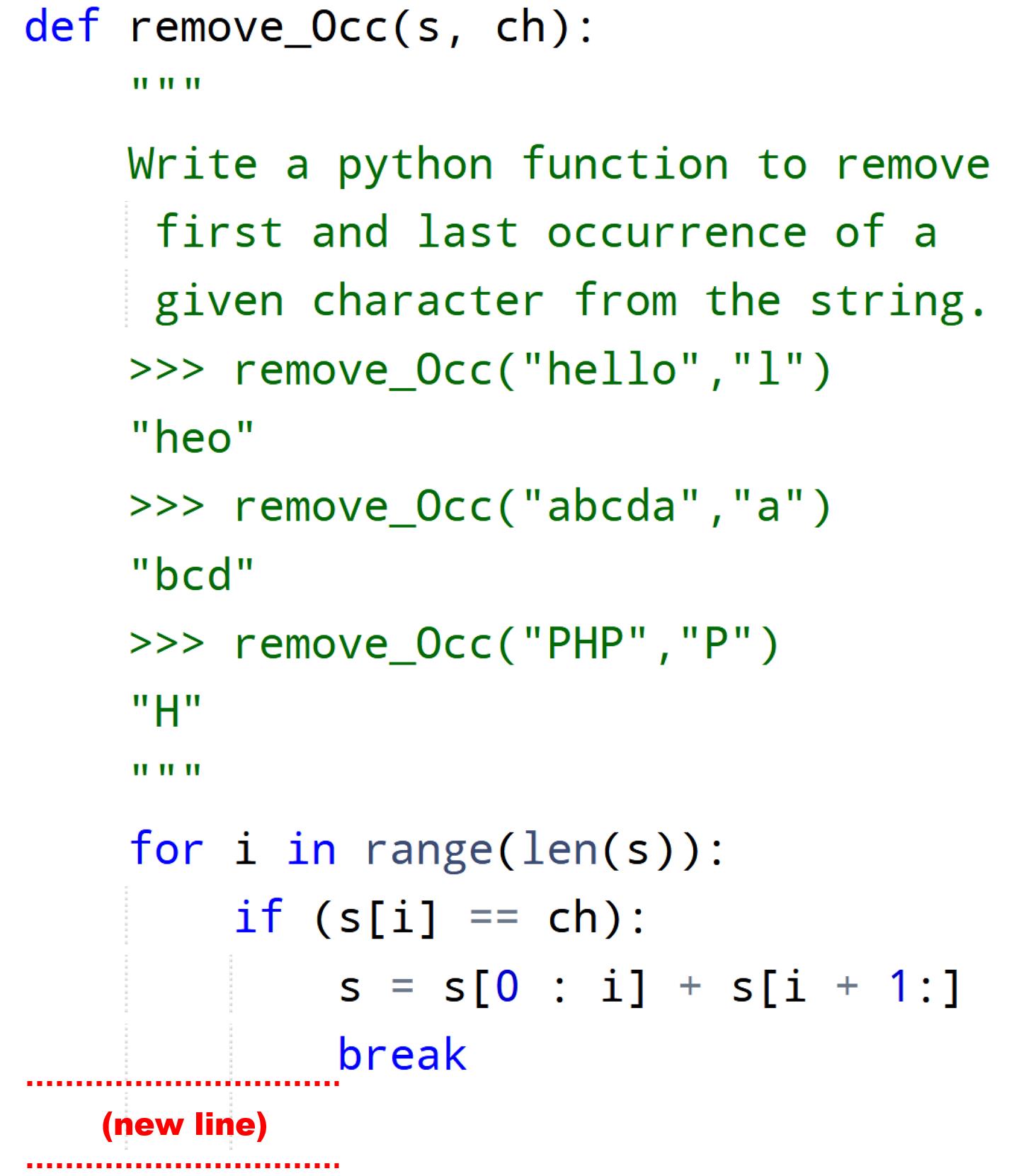}
    \caption{An example of the NewlineAfterCode perturbation.}
    \label{fig: NewlineAfterCode}
\end{figure}

\paragraph{NewlineAfterDoc.}
This transformation inserts an empty line between the docstring and the partial code.
This transformation is deterministic.
\begin{figure}[!hbt]
    \centering
    \includegraphics[width=0.6\linewidth]{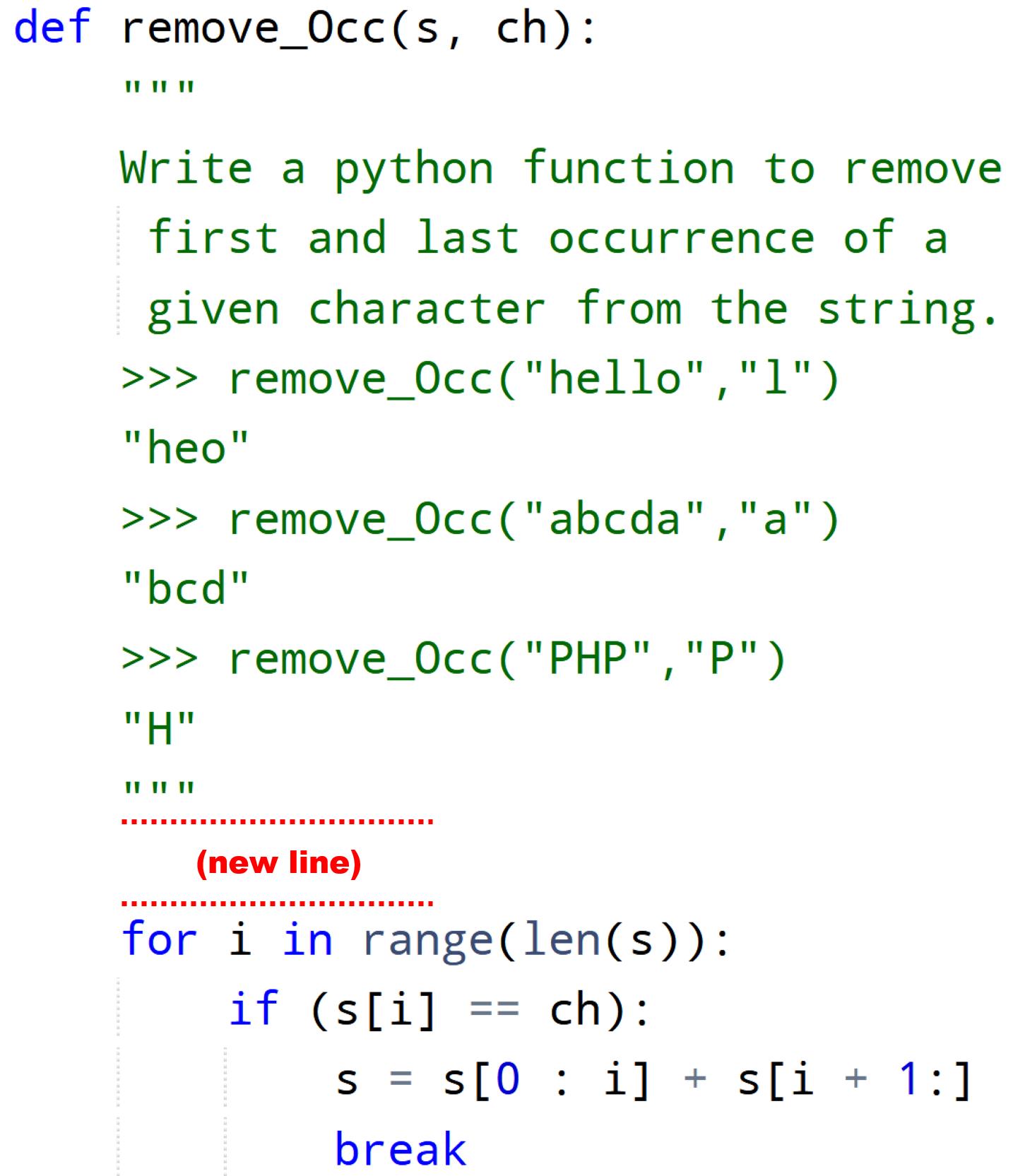}
    \caption{An example of the NewlineAfterDoc perturbation.}
    \label{fig: NewlineAfterDoc}
\end{figure}

\section{Limitations}
\label{appd: limitations}
\texttt{ReCode} benchmark has several limitations: (1) It contains perturbed datasets based on HumanEval and MBPP which focuses on Python function completion use cases. 
Therefore, we only perform evaluation on Python language and not be able to capture robustness in a wide variety of code completion use cases. However, our transformations are generalizable and could be easily extended to other languages and also other code-related datasets.
We encourage researchers to apply and extend \texttt{ReCode} benchmark to additional languages and other code-related tasks; 
(2) \texttt{ReCode} benchmark is designed for robustness evaluation and can not mitigate the lack of robustness. Given that our benchmark can be used to generate comprehensive collection of perturbed data, we believe that it can be used for training data augmentation to enhance model robustness. We will consider corresponding robust training strategy design and evaluation in future work.

\section{Failure Case Study under Perturbations}
\label{appd: error_analysis}
In this section, we showcase and analyze some failure cases on CodeGen-16B-mono and perturbed HumanEval datasets under three top perturbations that will cause significant performance drops. 

\begin{figure}[!hbt]
    \begin{subfigure}[b]{0.44\textwidth}
        \includegraphics[width=\linewidth]{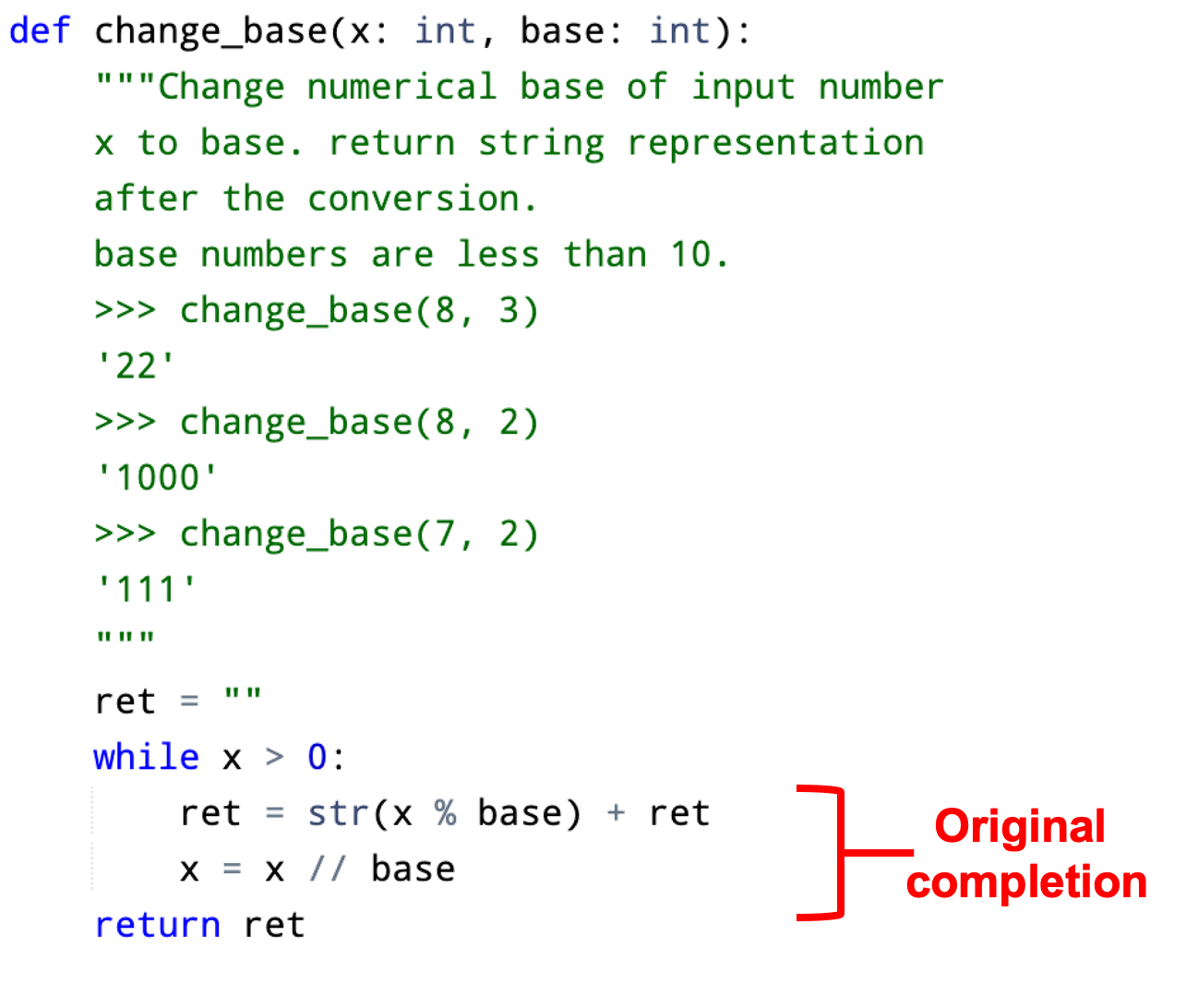}
            \caption{Correct completion without perturbation.}
            \label{subfig: appd_deadcoden}
    \end{subfigure}
    \begin{subfigure}[b]{0.4\textwidth}
        \includegraphics[width=\linewidth]{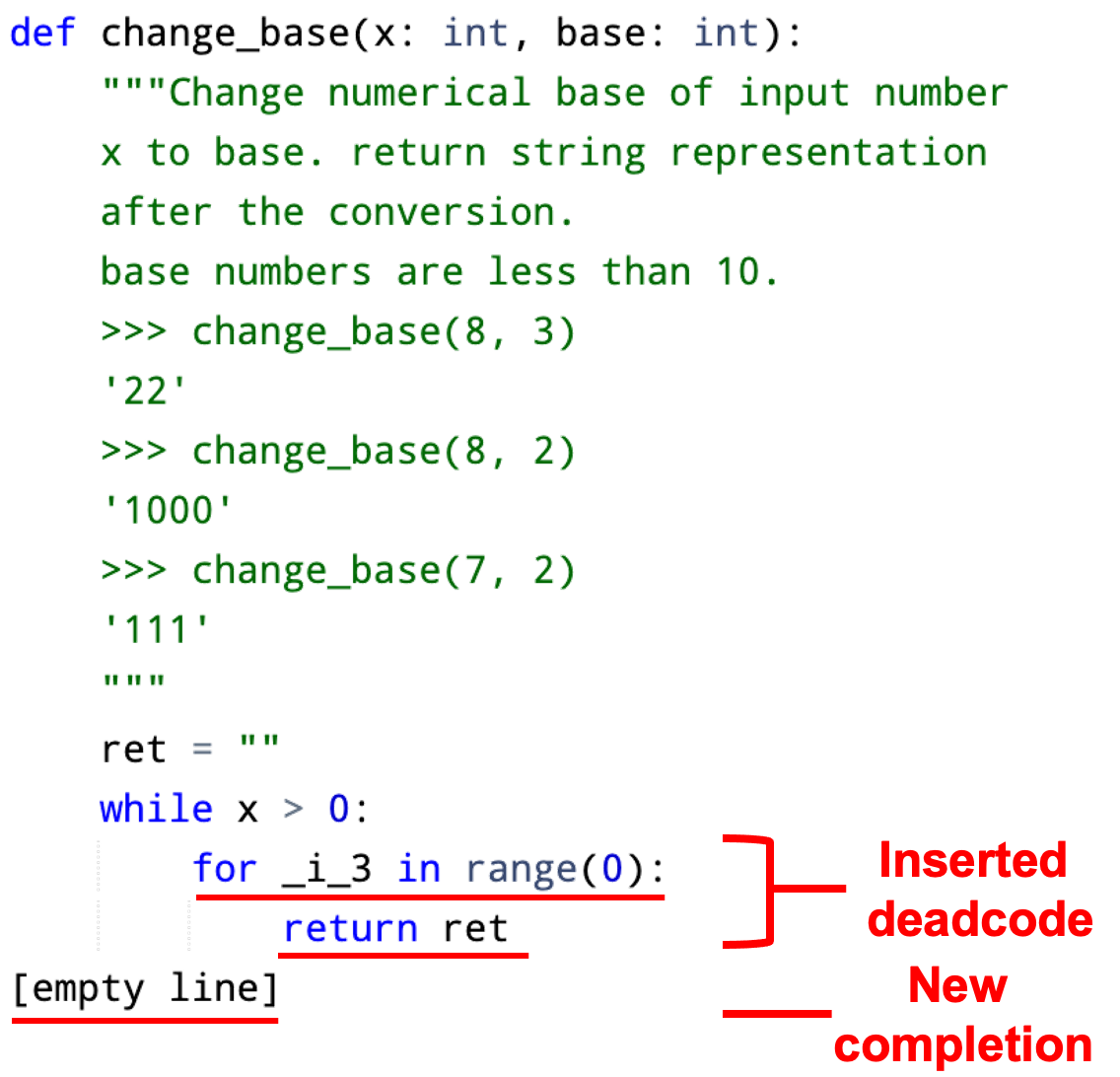}
                \caption{Wrong completion perturbed by \texttt{deadcode insertion}.}
        \label{fig: appd_deadcodep}
    \end{subfigure}
    \caption{HumanEval showcase 1 illustrating failure case under \texttt{deadcode insertion}.}
    \label{fig: appd_deadcode}
\end{figure}

DeadCode insertion is one of the most effective perturbations. It can commonly mislead the model predictions with the inserted dead code, especially when the completions are required right after the inserted dead code. \cref{fig: appd_deadcode} shows an failure example where CodeGen-mono-16B only predicts a newline after inserted meaningless for loop, which might be mislead by the inserted return statement.

\begin{figure}[!hbt]
    \begin{subfigure}[b]{0.49\textwidth}
        \includegraphics[width=\linewidth]{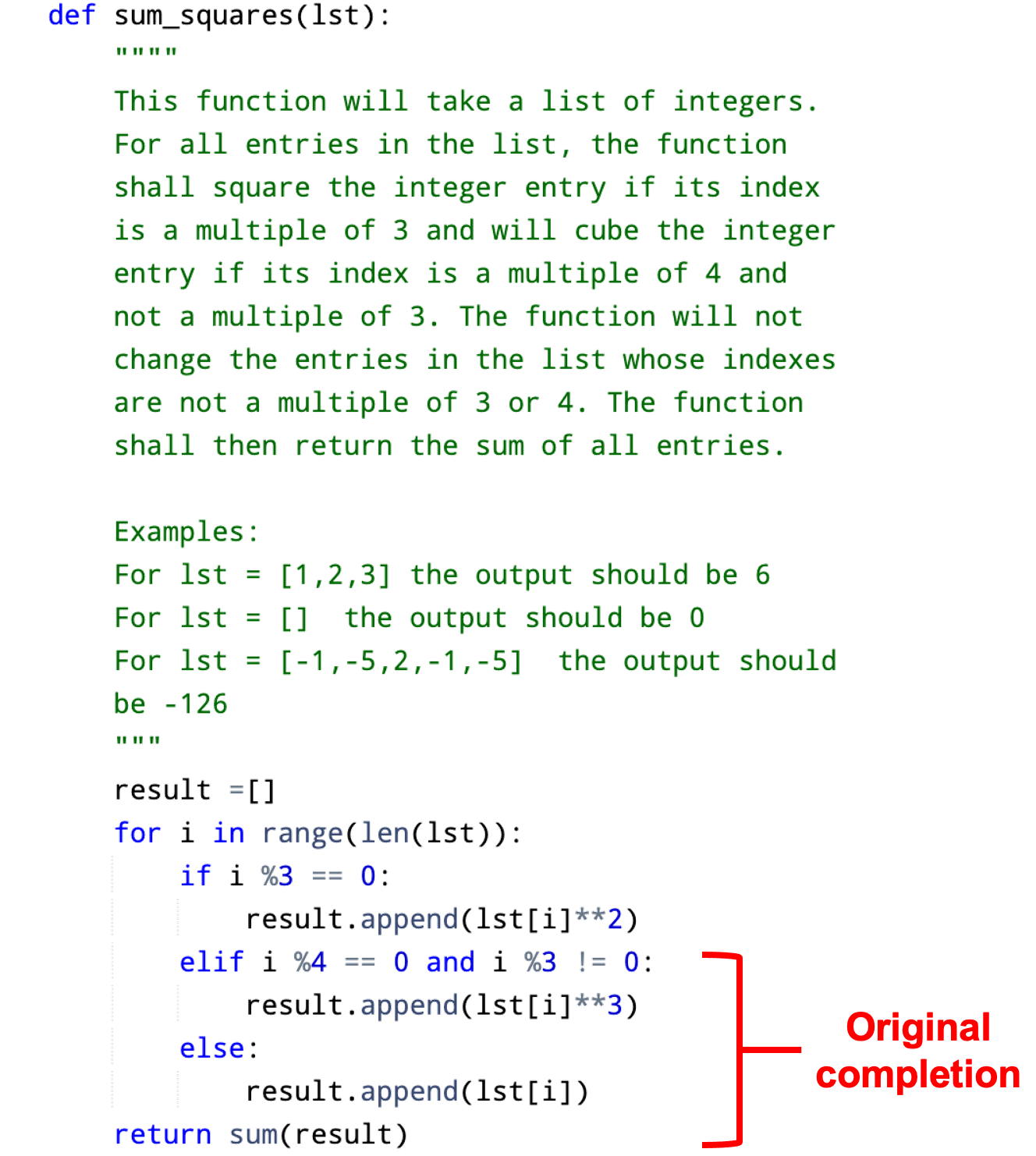}
            \caption{Correct completion without perturbation.}
            \label{subfig: appd_newlinen}
    \end{subfigure}
    \begin{subfigure}[b]{0.49\textwidth}
        \includegraphics[width=\linewidth]{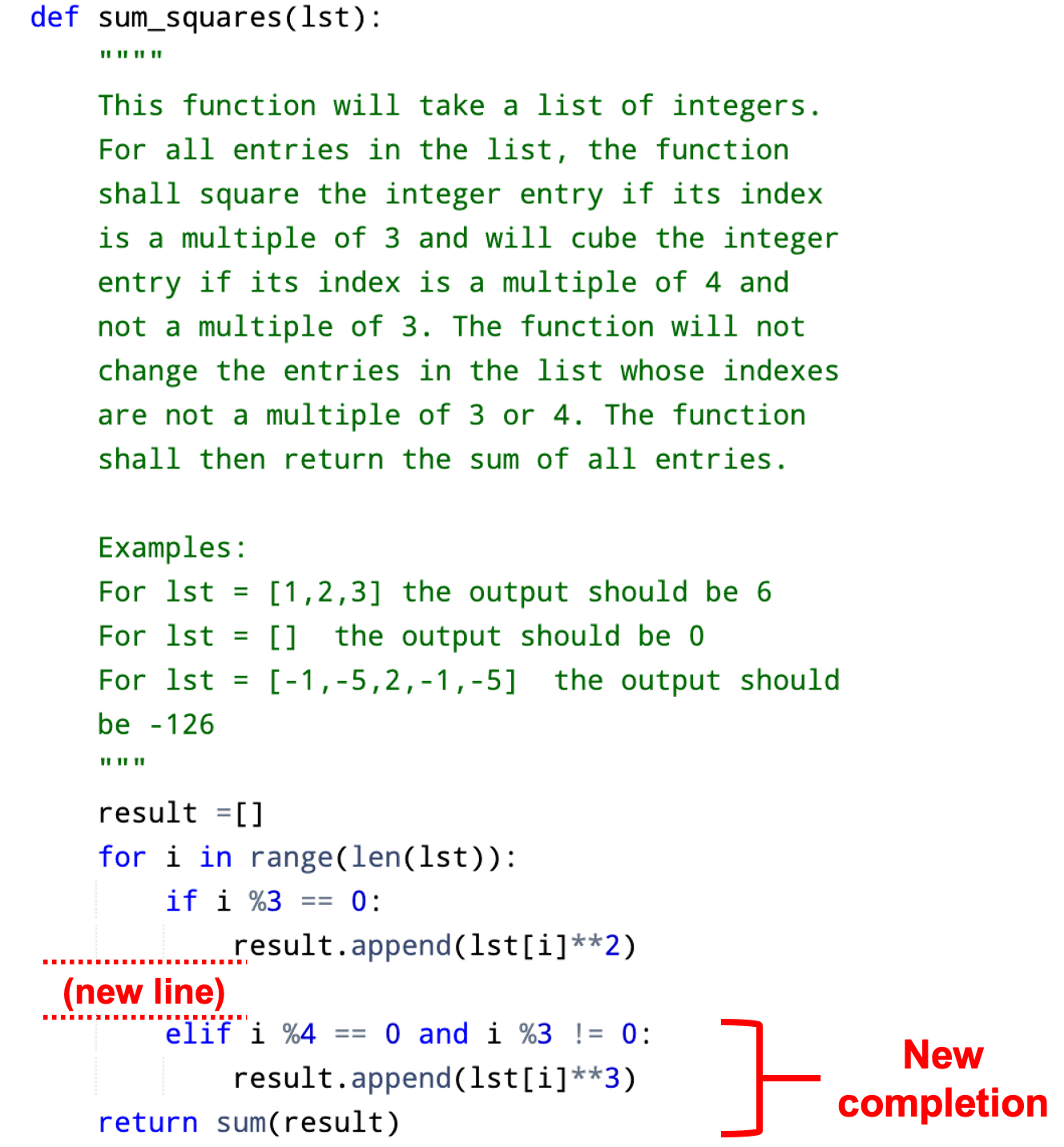}
                \caption{Wrong completion perturbed by \texttt{NewlineAfterCode} insertion.}
        \label{fig: appd_newlinep}
    \end{subfigure}
    \caption{HumanEval showcase 2 illustrating failure case under \texttt{NewlineAfterCode} insertion.}
    \label{fig: appd_newline}
\end{figure}

\cref{fig: appd_newline} shows a failure example of CodeGen-16B-mono on a prompt where an empty newline is inserted right before completion. Such simple perturbation causes wrong predictions for the following if-else conditions. It is especially effective when the required completion code is complicated.

\begin{figure}[!hbt]
    \begin{subfigure}[b]{0.49\textwidth}
        \includegraphics[width=\linewidth]{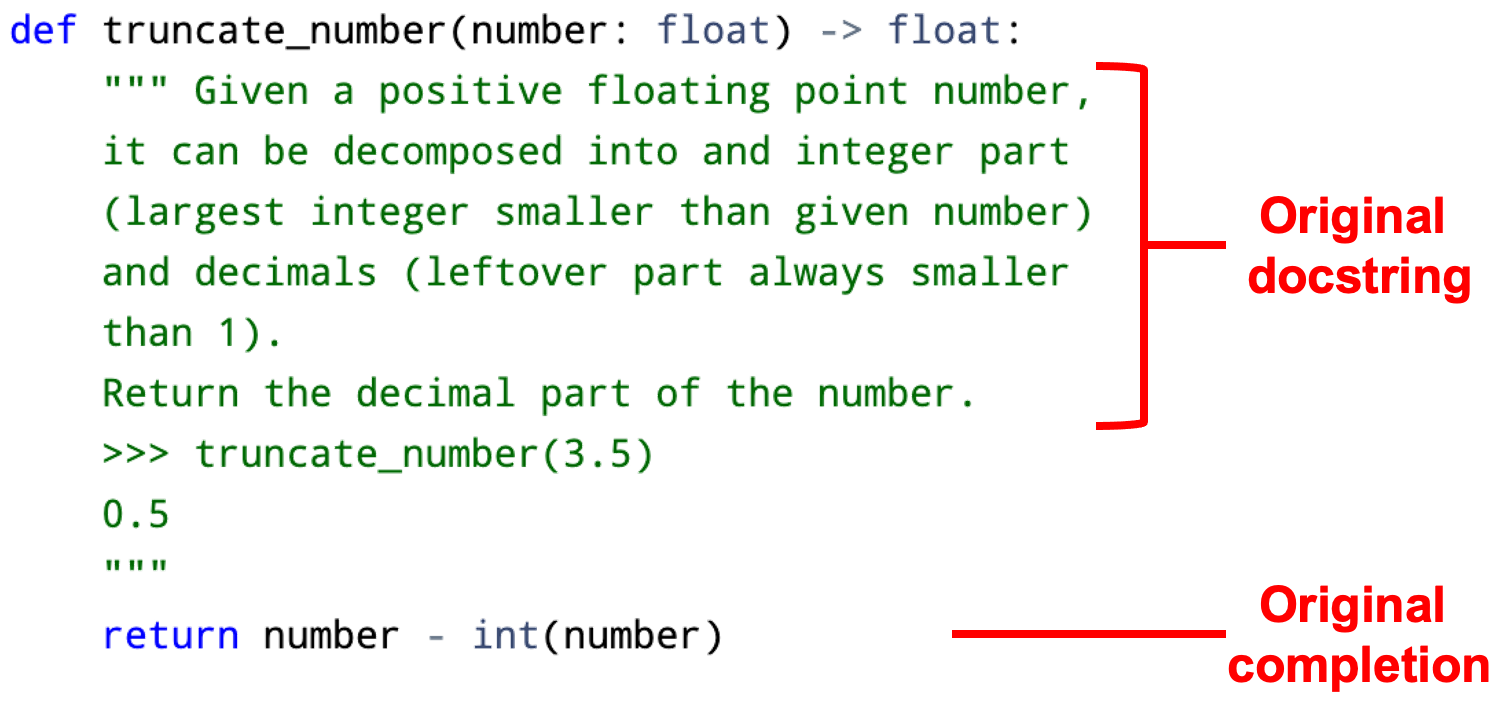}
            \caption{Correct completion without perturbation.}
            \label{subfig: appd_buttern}
    \end{subfigure}
    \begin{subfigure}[b]{0.49\textwidth}
        \includegraphics[width=\linewidth]{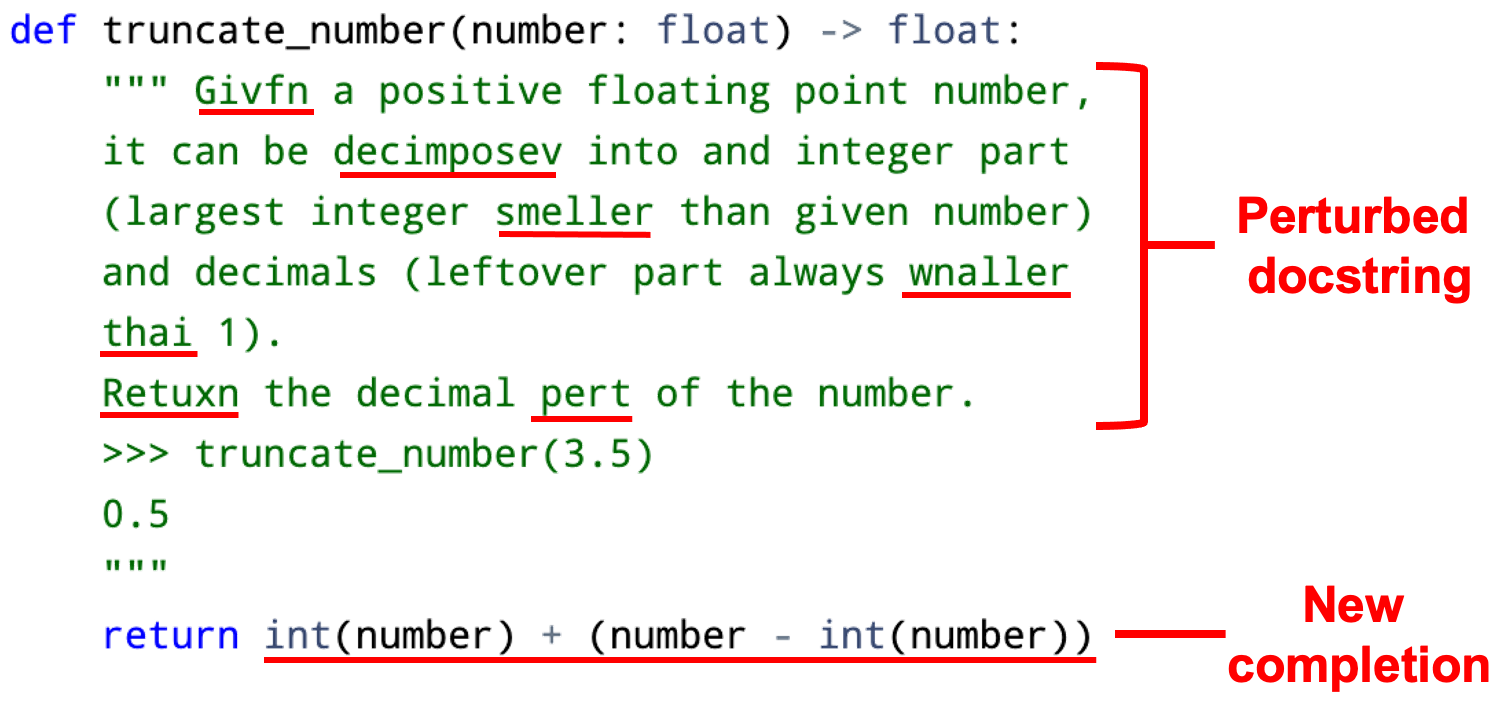}
                \caption{Wrong completion perturbed by \texttt{ButterFingers}.}
        \label{fig: appd_butterp}
    \end{subfigure}
    \caption{HumanEval showcase 3 illustrating failure case under \texttt{ButterFingers} perturbations on docstrings.}
    \label{fig: appd_butter}
\end{figure}

ButterFingers perturbation on docstring causes large performance drops as well. \cref{fig: appd_butter} shows another falure example on CodeGen-16B-mono. The typos introduced in the perturbation might cause the model to misunderstand the targeted docstrings, leading to wrong model completions.

\section{Perturbation Sample Quality}

\subsection{Details for Human Evaluation}
\label{appd: human_evaluation}

The annotators are all recruited from software engineers online who have good experience in Python via strict coding interview. To guarantee the reliability of the human evaluation results, we first conducted annotation trials with our annotators. We gave them clear definitions for each level of naturalness and semantic similarity.

We measure the inter-annotator agreement rate Fless Kappa in \cref{appd: tab_kappa}. The overall average Fleiss Kappa for the annotations is $0.52, 0.36$ for semantic and naturalness measurements on perturbed samples. The confidence interval (95\%) with bootstrap sampling (10K samples) is $[0.515, 0.528]$ and $[0.358, 0.364]$, indicating that our annotation reaches “moderate agreement” and thus our annotations are reliable~\citep{gwet2014handbook}.
 The scores from annotators are not perfectly consistent especially for naturalness since people have different preferences for code.

\begin{table}[ht]
\centering
\footnotesize
\begin{tabular}{lrr} \toprule
Fleiss Kappa          & \multicolumn{1}{c}{HumanEval} & \multicolumn{1}{c}{MBPP} \\ \midrule
Naturalness (Nominal) $\uparrow$  & 0.362                         & 0.301                    \\
Naturalness (Perturbed) $\uparrow$ & 0.435                         & 0.326                    \\
Semantics Similarity $\uparrow$  & 0.658                         & 0.461                   \\ \bottomrule
\end{tabular}
\caption{Fleiss Kappa of human evaluation.}
\label{appd: tab_kappa}
\end{table}

\subsection{Sentence Transformers for Docstring/Function Names Similarity}
\label{appd: sentrans}

In this subsection, we give experimental details for measuring the sentence similarity of perturbed and unperturbed data points using sentence transformers.

To measure the similarity scores for the docstring perturbations, we first extract docstrings from each pair of perturbed and unperturbed data points, and we use sentence transformer \texttt{all-mpnet-base-v2}~\citep{song2020mpnet} to predict an embedding vector for each docstring. Then cosine similarity is calculated and reported for each pair of perturbed and unperturbed datapoints.

Same process cannot be directly applied to function name perturbations since function names are concatenations of words instead of common sentences, barely seen by the sentence transformer training data. In order get more accurate sentence embeddings for function names, we first split each name into words (e.g., \texttt{has\_close\_elements} to \texttt{has close elements}) and then calculate the corresponding cosine similarities. 

In Table~\ref{tab: appd_sentrans}, we present the detailed results for each type of perturbations for sentence similarity. On average, we can have 0.93 and 0.92 similarity scores for docstring perturbations and 0.80 and 0.81 for function name perturbations on the HumanEval and MBPP datasets. The overall high similarity numbers provide support that our perturbations have good quality in naturalness and semantic preservation from the unperturbed inputs.

Some function name perturbations including ButterFinger, SynonymSubstitution, and CharCaseChange have relatively low sentence similarity. This is mainly because the function names only include keywords without complete sentence context and thus minor changes to each words could potentially cause large change in measured cosine similarity. For instance, character case changes on function name \texttt{intersperse} to \texttt{intErspErse} which lacks of context only has 0.21 similarity. On the other hand, the function names with more context has much higher scores, e.g., 1.0 similarity score for \texttt{has\_close\_elements} and \texttt{has\_ClosE\_Elements}.

\begin{table*}[t]
\centering
\footnotesize
\begin{tabular}{l|l|r|r} \toprule
Categories                  & Perturbations                           & \multicolumn{1}{l}{HumanEval} & \multicolumn{1}{l}{MBPP} \\ \midrule
\multirow{10}{*}{Docstring} & BackTranslation                         & 0.91                          & 0.95                     \\
                            & ButterFingers               & 0.87                          & 0.89                     \\
                            & ChangeCharCase                          & 1.00                          & 1.00                     \\
                            & EnglishInflectionalVariation            & 0.96                          & 0.93                     \\
                            & SwapCharacters              & 0.90                          & 0.87                     \\
                            & SynonymInsertion                        & 0.91                          & 0.88                     \\
                            & SynonymSubstitution                     & 0.88                          & 0.84                     \\
                            & TenseTransformationPast                 & 0.98                          & 1.00                     \\
                            & TenseTransformationFuture               & 0.97                          & 0.97                     \\
                            & Whitespace                  & 0.90                          & 0.86                     \\ \midrule
\multirow{6}{*}{Function}   &  CamelCase             & 1.00                          & 1.00                     \\
                            &  ButterFingers         & 0.57                          & 0.57                     \\
                            &  SwapCharacters       & 0.75                          & 0.75                     \\
                            &  ChangeCharCase     & 0.86                          & 0.96                     \\
                            &  InflectionalVariation & 0.94                          & 0.93                     \\
                            &  SynonymSubstition    & 0.68                          & 0.64                    
\\ \bottomrule
\end{tabular}
\caption{Cosine similarity for each type of perturbations where perturbed and unperturbed docstrings/function names are embedded by the SOTA sentence transformer.}
\label{tab: appd_sentrans}
\end{table*}

\subsection{CodeBLEU Scores for Code Similarity}
\label{appd: codebleu}

Here we present the experimental details for the CodeBLEU syntax and dataflow scores to quantitatively measure the quality of our code syntax and format transformations. 

The measurement is straightforward. The unperturbed baseline is each data point from our customized partial code datasets derived from HumanEval and MBPP. The perturbed one is the same data point transformed by each type of our perturbations. The CodeBLEU syntax and dataflow scores are then directly measured using the CodeXGLUE~\citep{lu2021codexglue} implementation.\footnote{https://github.com/microsoft/CodeXGLUE}

In Table~\ref{tab: appd_codegleu}, we present the detailed CodeBLEU results for each type of perturbations. The average numbers are summarized in Table~\ref{tab: codegleu}. Overall, 77\% and 89\% of our transformations have over 0.9 CodeBLEU syntax and dataflow scores, showing good quality in preserving semantics from the unperturbed code.

However, CodeBLEU syntax and dataflow are not perfect in quantitatively measuring naturalness and semantic preservation for the perturbations and thus some perturbations have expected relatively low scores: \texttt{Doc2Comments} transforms docstrings into comments causing changes of syntax; \texttt{Deadcode insertion} and \texttt{for-while switch} involve new if-conditions, loops, and new variables causing changes of code syntax and dataflow.

\begin{table*}[t]
\centering
\footnotesize
\begin{tabular}{l|l|rr|rr} \toprule
\multirow{3}{*}{Categories} & \multirow{3}{*}{Perturbations} & \multicolumn{2}{|c}{HumanEval} & \multicolumn{2}{|c}{MBPP} \\
                            &                                & CodeBLEU     & CodeBLEU       & CodeBLEU   & CodeBLEU    \\
                            &                                & (syntax)     & (dataflow)     & (syntax)   & (dataflow)  \\ \midrule
\multirow{6}{*}{Syntax}     & DeadCodeInserter               & 0.85         & 0.79           & 0.72       & 0.67        \\
                            & For-While Switch            & 0.92         & 0.90           & 0.84       & 0.86        \\
                            & OperandSwap                    & 0.91         & 1.00           & 0.90       & 1.00        \\
                            & VarRenamerCB                   & 1.00         & 0.99           & 0.93       & 0.99        \\
                            & VarRenamerNaive                & 1.00         & 0.99           & 0.93       & 0.99        \\
                            & VarRenamerRN                   & 1.00         & 0.99           & 0.93       & 0.99        \\ \midrule
\multirow{7}{*}{Format}     & Tab-Indent                    & 1.00         & 1.00           & 1.00       & 1.00        \\
                            & Line Split                   & 1.00         & 1.00           & 1.00       & 1.00        \\
                            & Doc2Comments                   & 0.84         & 1.00           & 0.76       & 1.00        \\
                            & NewlineRandom                     & 1.00         & 1.00           & 1.00       & 1.00        \\
                            & NewlineAfterCode           & 1.00         & 1.00           & 1.00       & 1.00        \\
                            & NewlineAfterDoc            & 1.00         & 1.00           & 1.00       & 1.00        \\
                            \bottomrule
\end{tabular}
\caption{CodeBLEU syntax and format similarity scores between unperturbed codes and perturbed ones with each type of our syntax and format transformations.}
\label{tab: appd_codegleu}
\end{table*}

\section{Additional Results}
\label{appd: additiona_results}

\subsection{Fine-grained Robustness Evaluation}
\label{appd: finegrained}

We present the robustness evaluation for each type of perturbations from Table~\ref{appd: tab_doc_humaneval} to~\ref{appd: tab_format_mbpp}, . The evaluation setting is the same as Table~\ref{tab: main_humaneval} and~\ref{tab: main_mbpp} where we evaluate various sizes of CodeGen~\citep{CodeGen}, InCoder~\citep{incoder}, and GPT-J~\citep{gpt-j} with greedy sampling. For each type of perturbations, we randomly generate $s=5$ different perturbed datasets derived from HumanEval and MBPP. For perturbations without randomness, only one single version of perturbed dataset is evaluated. The list of indeterministic perturbations can be found in Appendix~\ref{appd: transformation}.

\begin{table*}[t]
\centering
\footnotesize
\setlength{\tabcolsep}{3pt}
\scalebox{0.75}{
\begin{tabular}{r|r|rr|rr|rr|rr|r} \toprule
\multirow{2}{*}{HumanEval}   & \multirow{2}{*}{Metric} & CodeGen & CodeGen  & CodeGen & CodeGen  & CodeGen  & CodeGen   & InCoder & InCoder & GPT-J \\
                             &                         & 2B mono & 2B multi & 6B mono & 6B multi & 16B mono & 16B multi & 1B      & 6B      & 6B    \\ \midrule
Nominal                          & RP$_5$@1$\uparrow$       & 0.232           & 0.140            & 0.262           & 0.195            & \textbf{0.305}            & 0.195             & 0.104      & 0.152      & 0.122    \\ \midrule
\multirow{3}{*}{BackTranslation} & RP$_5$@1$\uparrow$       & 0.213           & 0.116            & 0.238           & 0.159            & \textbf{0.244}            & 0.152             & 0.098      & 0.134      & 0.098    \\
                                 & RD$_5$@1(\%)$\downarrow$     & 7.89            & 17.39            & 9.30            & 18.75            & 20.00            & 21.88             & \textbf{5.88}       & 12.00      & 20.00    \\
                                 & RR$_5$@1(\%)$\downarrow$ & 4.27            & 6.10             & 8.54            & 6.10             & 10.98            & 5.49              & \textbf{3.05}       & 3.05       & 3.66     \\ \midrule
\multirow{3}{*}{ButterFingers}        & RP$_5$@1$\uparrow$       & 0.165           & 0.098            & 0.171           & 0.122            & \textbf{0.189}            & 0.116             & 0.067      & 0.098      & 0.067    \\
        & RD$_5$@1(\%)$\downarrow$     & \textbf{28.95}           & 30.43            & 34.88           & 37.50            & 38.00            & 40.62             & 35.29      & 36.00      & 45.00    \\
\bottomrule 
\end{tabular}
}
\caption{Robustness evaluation for each type of docstring perturbations on HumanEval. }
\label{appd: tab_doc_humaneval}
\end{table*}

\begin{table*}[t]
\centering
\footnotesize
\setlength{\tabcolsep}{3pt}
\scalebox{0.75}{
\begin{tabular}{r|r|rr|rr|rr|rr|r} \toprule
\multirow{2}{*}{MBPP}   & \multirow{2}{*}{Metric} & CodeGen & CodeGen  & CodeGen & CodeGen  & CodeGen  & CodeGen   & InCoder & InCoder & GPT-J \\
                             &                         & 2B mono & 2B multi & 6B mono & 6B multi & 16B mono & 16B multi & 1B      & 6B      & 6B    \\ \midrule
Nominal                      & RP$_5$@1$\uparrow$       & 0.317           & 0.191            & 0.361           & 0.221            & \textbf{0.407}            & 0.241             & 0.128      & 0.199      & 0.133    \\ \midrule
\multirow{3}{*}{\shortstack[r]{BackTranslation}}              & RP$_5$@1$\uparrow$       & 0.304           & 0.186            & 0.360           & 0.222            & \textbf{0.387}            & 0.230             & 0.119      & 0.177      & 0.128    \\
              & RD$_5$@1(\%)$\downarrow$     & 4.21            & 2.69             & 0.28            & \textbf{-0.47}            & 4.80             & 4.68              & 7.20       & 11.34      & 3.85     \\
              & RR$_5$@1(\%)$\downarrow$ & 6.26            & 6.06             & 7.91            & 6.06             & 6.26             & 7.08              & \textbf{4.00}       & 5.95       & 5.44     \\ \midrule
\multirow{3}{*}{\shortstack[r]{ButterFingers\\Perturbation}}    & RP$_5$@1$\uparrow$       & 0.210           & 0.092            & 0.240           & 0.100            & \textbf{0.280}            & 0.126             & 0.044      & 0.082      & 0.057    \\
    & RD$_5$@1(\%)$\downarrow$     & 33.66           & 51.61            & 33.52           & 54.88            & \textbf{31.06}            & 47.66             & 65.60      & 58.76      & 56.92    \\
\bottomrule
\end{tabular} 
}
\caption{Robustness evaluation for each type of docstring perturbations on MBPP. }
\label{appd: tab_doc_mbpp}
\end{table*}

\begin{table*}[t]
\centering
\footnotesize
\setlength{\tabcolsep}{3pt}
\scalebox{0.75}{
\begin{tabular}{r|r|rr|rr|rr|rr|r} \toprule
\multirow{2}{*}{HumanEval}   & \multirow{2}{*}{Metric} & CodeGen & CodeGen  & CodeGen & CodeGen  & CodeGen  & CodeGen   & InCoder & InCoder & GPT-J \\
&                         & 2B mono & 2B multi & 6B mono & 6B multi & 16B mono & 16B multi & 1B      & 6B      & 6B    \\ \midrule
Nominal                         & RP$_5$@1$\uparrow$       & 0.232           & 0.140            & 0.262           & 0.195            & \textbf{0.305}            & 0.195             & 0.104      & 0.152      & 0.122    \\ \midrule
\multirow{3}{*}{\shortstack[r]{CamelCase}}             & RP$_5$@1$\uparrow$       & 0.238           & 0.140            & 0.256           & 0.201            & \textbf{0.293}            & 0.165             & 0.098      & 0.152      & 0.116    \\
             & RD$_5$@1(\%)$\downarrow$     & -2.63           & 0.00             & 2.33            & \textbf{-3.13}            & 4.00             & 15.62             & 5.88       & 0.00       & 5.00     \\
             & RR$_5$@1(\%)$\downarrow$ & 1.83            & 1.22             & 3.05            & 3.05             & 3.66             & 3.05              & 3.05       & 1.22       & \textbf{0.61}     \\ \midrule
\multirow{3}{*}{\shortstack[r]{ButterFinger}}          & RP$_5$@1$\uparrow$       & 0.195           & 0.104            & 0.232           & 0.177            & \textbf{0.274}            & 0.159             & 0.098      & 0.140      & 0.091    \\
          & RD$_5$@1(\%)$\downarrow$     & 15.79           & 26.09            & 11.63           & 9.38             & 10.00            & 18.75             & \textbf{5.88}       & 8.00       & 25.00    \\
\bottomrule
\end{tabular}
}
\caption{Robustness evaluation for each type of function name perturbations on HumanEval. }
\label{appd: tab_func_humaneval}
\end{table*}

\begin{table*}[t]
\centering
\footnotesize
\setlength{\tabcolsep}{3pt}
\scalebox{0.75}{
\begin{tabular}{r|r|rr|rr|rr|rr|r} \toprule
\multirow{2}{*}{MBPP}   & \multirow{2}{*}{Metric} & CodeGen & CodeGen  & CodeGen & CodeGen  & CodeGen  & CodeGen   & InCoder & InCoder & GPT-J \\
&                         & 2B mono & 2B multi & 6B mono & 6B multi & 16B mono & 16B multi & 1B      & 6B      & 6B    \\ \midrule
Nominal                         & RP$_5$@1$\uparrow$       & 0.317           & 0.191            & 0.361           & 0.221            & \textbf{0.407}            & 0.241             & 0.128      & 0.199      & 0.133    \\ \midrule
\multirow{3}{*}{\shortstack[r]{CamelCase}}             & RP$_5$@1$\uparrow$       & 0.316           & 0.196            & 0.367           & 0.219            & \textbf{0.408}            & 0.245             & 0.116      & 0.194      & 0.134    \\
             & RD$_5$@1(\%)$\downarrow$     & 0.32            & \textbf{-2.69}            & -1.42           & 0.93             & -0.25            & -1.70             & 9.60       & 2.58       & -0.77    \\
             & RR$_5$@1(\%)$\downarrow$ & 5.44            & 5.44             & 7.29            & 5.34             & 7.08             & 4.52              & 5.75       & 5.03       & \textbf{3.18}     \\ \midrule
\multirow{3}{*}{\shortstack[r]{ButterFinger}}          & RP$_5$@1$\uparrow$       & 0.312           & 0.185            & 0.370           & 0.203            & \textbf{0.412}            & 0.231             & 0.110      & 0.175      & 0.117    \\
          & RD$_5$@1(\%)$\downarrow$     & 1.62            & 3.23             & \textbf{-2.27}           & 7.91             & -1.26            & 4.26              & 14.40      & 12.37      & 12.31    \\
\bottomrule
\end{tabular}
}
\caption{Robustness evaluation for each type of function name perturbations on MBPP. }
\label{appd: tab_func_mbpp}
\end{table*}

\begin{table*}[t]
\centering
\footnotesize
\setlength{\tabcolsep}{3pt}
\scalebox{0.75}{
\begin{tabular}{r|r|rr|rr|rr|rr|r} \toprule
\multirow{2}{*}{HumanEval}   & \multirow{2}{*}{Metric} & CodeGen & CodeGen  & CodeGen & CodeGen  & CodeGen  & CodeGen   & InCoder & InCoder & GPT-J \\
&                         & 2B mono & 2B multi & 6B mono & 6B multi & 16B mono & 16B multi & 1B      & 6B      & 6B    \\ \midrule
Nominal                  & RP$_5$@1$\uparrow$       & 0.402           & 0.293            & 0.518           & 0.366            & \textbf{0.549}            & 0.390             & 0.189      & 0.323      & 0.250    \\ \midrule
\multirow{3}{*}{\shortstack[r]{DeadCodeInserter}}    & RP$_5$@1$\uparrow$         & 0.116                    & 0.079                     & 0.152                    & 0.110                     & \textbf{0.159}                     & 0.091                      & 0.055               & 0.079               & 0.079             \\
    & RD$_5$@1(\%)$\downarrow$       & 71.21                    & 72.92                     & 70.59                    & 70.00                     & 71.11                     & 76.56                      & 70.97               & 75.47               & \textbf{68.29}             \\
    & RR$_5$@1(\%)$\downarrow$   & 37.80                    & 30.49                     & 41.46                    & 32.93                     & 45.12                     & 37.20                      & \textbf{17.07}               & 30.49               & 27.44             \\ \midrule
\multirow{3}{*}{\shortstack[r]{ForWhile\\TransformerFirst}} & RP$_5$@1$\uparrow$         & 0.384                    & 0.226                     & 0.500                    & 0.305                     & \textbf{0.537}                     & 0.384                      & 0.159               & 0.280               & 0.213             \\
\bottomrule
\end{tabular}
}
\caption{Robustness evaluation for each type of code syntax perturbations on HumanEval. }
\label{appd: tab_syntax_humaneval}
\end{table*}

\begin{table*}[t]
\centering
\footnotesize
\setlength{\tabcolsep}{3pt}
\scalebox{0.75}{
\begin{tabular}{r|r|rr|rr|rr|rr|r} \toprule
\multirow{2}{*}{MBPP}   & \multirow{2}{*}{Metric} & CodeGen & CodeGen  & CodeGen & CodeGen  & CodeGen  & CodeGen   & InCoder & InCoder & GPT-J \\
&                         & 2B mono & 2B multi & 6B mono & 6B multi & 16B mono & 16B multi & 1B      & 6B      & 6B    \\ \midrule
Nominal                  & RP$_5$@1$\uparrow$       & 0.450           & 0.285            & 0.535           & 0.331            & \textbf{0.571}            & 0.379             & 0.219      & 0.292      & 0.176    \\ \midrule
\multirow{3}{*}{\shortstack[r]{DeadCodeInserter}}    & RP$_5$@1$\uparrow$         & 0.043                    & 0.020                     & 0.044                    & 0.024                     & \textbf{0.055}                     & 0.025                      & 0.015               & 0.015               & 0.009             \\
    & RD$_5$@1(\%)$\downarrow$       & 90.41                    & 93.17                     & 91.75                    & 92.86                     & \textbf{90.29}                     & 93.50                      & 92.96               & 94.72               & 94.74             \\
    & RR$_5$@1(\%)$\downarrow$   & 52.05                    & 37.99                     & 57.39                    & 39.12                     & 60.57                     & 44.87                      & 29.26               & 37.78               & \textbf{24.95}             \\ \midrule
\multirow{3}{*}{\shortstack[r]{ForWhile\\TransformerFirst}} & RP$_5$@1$\uparrow$         & 0.432                    & 0.259                     & 0.497                    & 0.303                     & \textbf{0.532}                     & 0.346                      & 0.182               & 0.245               & 0.149             \\
\bottomrule
\end{tabular}
}
\caption{Robustness evaluation for each type of code syntax perturbations on MBPP. }
\label{appd: tab_syntax_mbpp}
\end{table*}

\begin{table*}[t]
\centering
\footnotesize
\setlength{\tabcolsep}{3pt}
\scalebox{0.75}{
\begin{tabular}{r|r|rr|rr|rr|rr|r} \toprule
\multirow{2}{*}{HumanEval}   & \multirow{2}{*}{Metric} & CodeGen & CodeGen  & CodeGen & CodeGen  & CodeGen  & CodeGen   & InCoder & InCoder & GPT-J \\
&                         & 2B mono & 2B multi & 6B mono & 6B multi & 16B mono & 16B multi & 1B      & 6B      & 6B    \\ \midrule
Nominal              & RP$_5$@1$\uparrow$       & 0.402           & 0.293            & 0.518           & 0.366            & \textbf{0.549}            & 0.390             & 0.189      & 0.323      & 0.250    \\ \midrule
\multirow{3}{*}{\shortstack[r]{Tab-Indent}}          & RP$_5$@1$\uparrow$       & 0.415           & 0.305            & 0.518           & 0.354            & \textbf{0.561}            & 0.396             & 0.146      & 0.299      & 0.244    \\
          & RD$_5$@1(\%)$\downarrow$     & -3.03           & \textbf{-4.17}            & 0.00            & 3.33             & -2.22            & -1.56             & 22.58      & 7.55       & 2.44     \\
          & RR$_5$@1(\%)$\downarrow$ & \textbf{3.66}            & 4.88             & 8.54            & 4.88             & 3.66             & 4.27              & 7.93       & 9.76       & 7.93     \\ \midrule
\multirow{3}{*}{\shortstack[r]{Line Split}}         & RP$_5$@1$\uparrow$       & 0.384           & 0.274            & 0.500           & 0.378            & \textbf{0.524}            & 0.390             & 0.171      & 0.305      & 0.244    \\
         & RD$_5$@1(\%)$\downarrow$     & 4.55            & 6.25             & 3.53            & \textbf{-3.33}            & 4.44             & 0.00              & 9.68       & 5.66       & 2.44     \\
\bottomrule
\end{tabular}
}
\caption{Robustness evaluation for each type of code format perturbations on HumanEval. }
\label{appd: tab_format_humaneval}
\end{table*}

\begin{table*}[t]
\centering
\footnotesize
\setlength{\tabcolsep}{3pt}
\scalebox{0.75}{
\begin{tabular}{r|r|rr|rr|rr|rr|r} \toprule
\multirow{2}{*}{MBPP}   & \multirow{2}{*}{Metric} & CodeGen & CodeGen  & CodeGen & CodeGen  & CodeGen  & CodeGen   & InCoder & InCoder & GPT-J \\
&                         & 2B mono & 2B multi & 6B mono & 6B multi & 16B mono & 16B multi & 1B      & 6B      & 6B    \\ \midrule
Nominal              & RP$_5$@1$\uparrow$       & 0.450           & 0.285            & 0.535           & 0.331            & \textbf{0.571}            & 0.379             & 0.219      & 0.292      & 0.176    \\ \midrule
\multirow{3}{*}{\shortstack[r]{Tab-Indent}}          & RP$_5$@1$\uparrow$       & 0.452           & 0.302            & 0.530           & 0.339            & \textbf{0.566}            & 0.385             & 0.208      & 0.325      & 0.176    \\
          & RD$_5$@1(\%)$\downarrow$     & -0.46           & -5.76            & 0.96            & -2.48            & 0.90             & -1.63             & 4.69       & \textbf{-11.62}     & 0.00     \\
          & RR$_5$@1(\%)$\downarrow$ & 6.37            & 6.98             & \textbf{5.85}            & 8.01             & 6.88             & 6.78              & 9.24       & 12.22      & 7.60     \\ \midrule
\multirow{3}{*}{\shortstack[r]{Line Split}}         & RP$_5$@1$\uparrow$       & 0.445           & 0.275            & 0.524           & 0.326            & \textbf{0.556}            & 0.378             & 0.187      & 0.283      & 0.163    \\
         & RD$_5$@1(\%)$\downarrow$     & 1.14            & 3.60             & 2.11            & 1.24             & 2.52             & \textbf{0.27}              & 14.55      & 2.82       & 7.02     \\
\bottomrule
\end{tabular}
}
\caption{Robustness evaluation for each type of code format perturbations on MBPP. }
\label{appd: tab_format_mbpp}
\end{table*}

\subsection{Additional Results for Different k}
\label{appd: k}

\begin{figure*}
    \centering
    \includegraphics[width=0.31\linewidth]{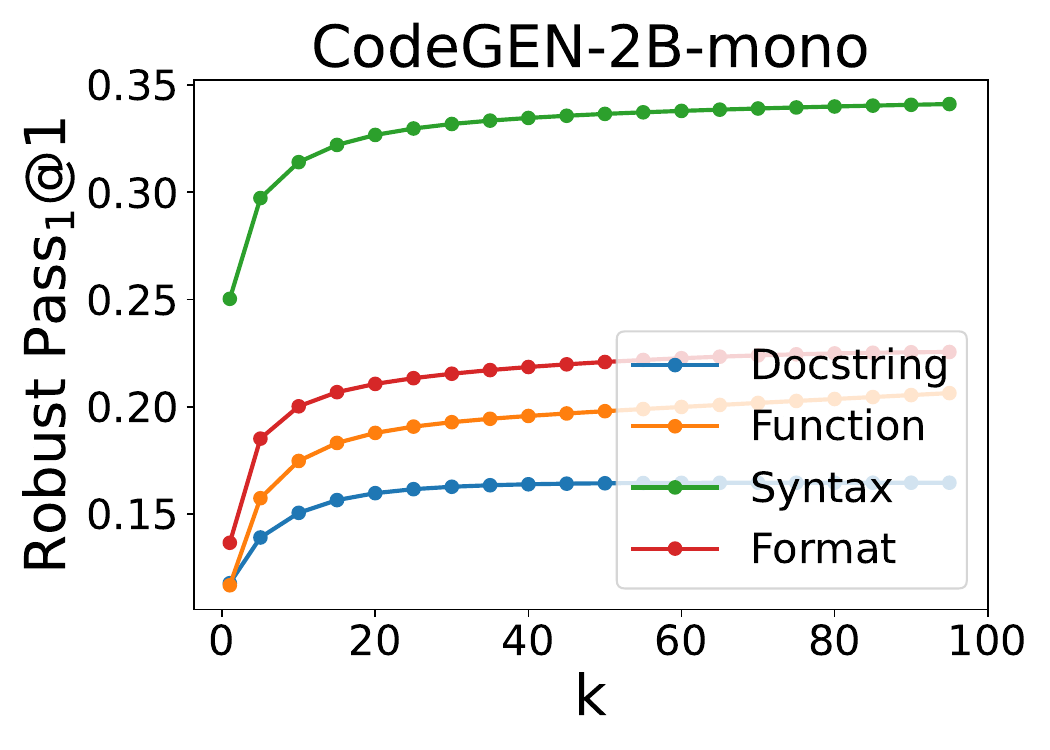}
    \includegraphics[width=0.31\linewidth]{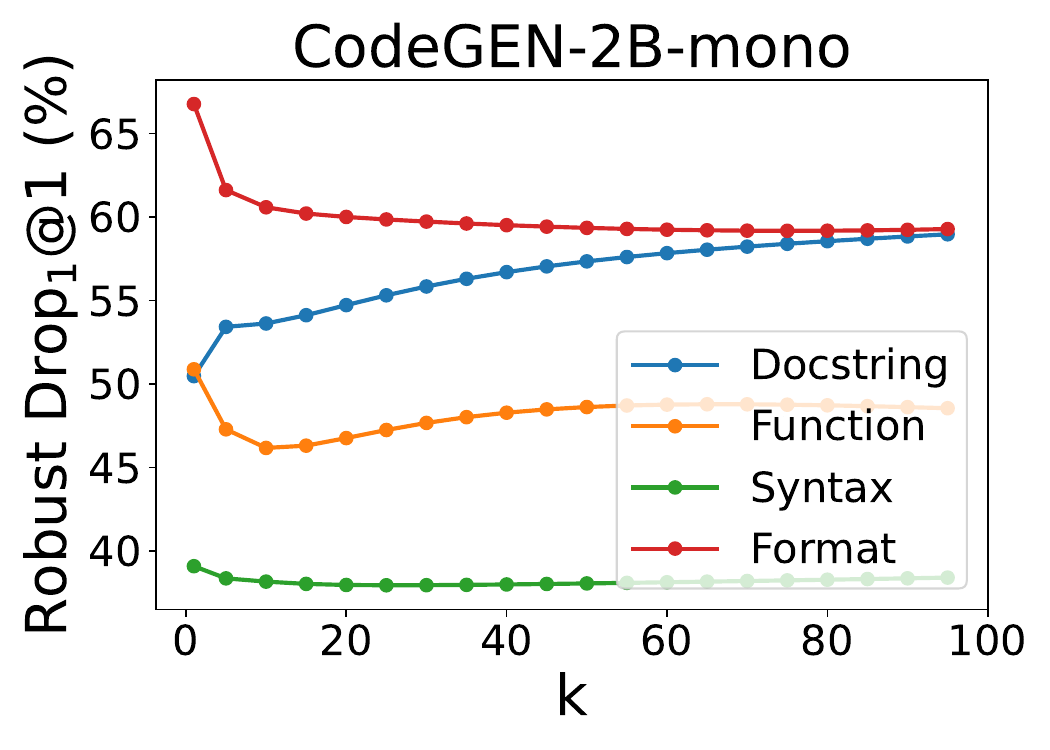}
    \includegraphics[width=0.31\linewidth]{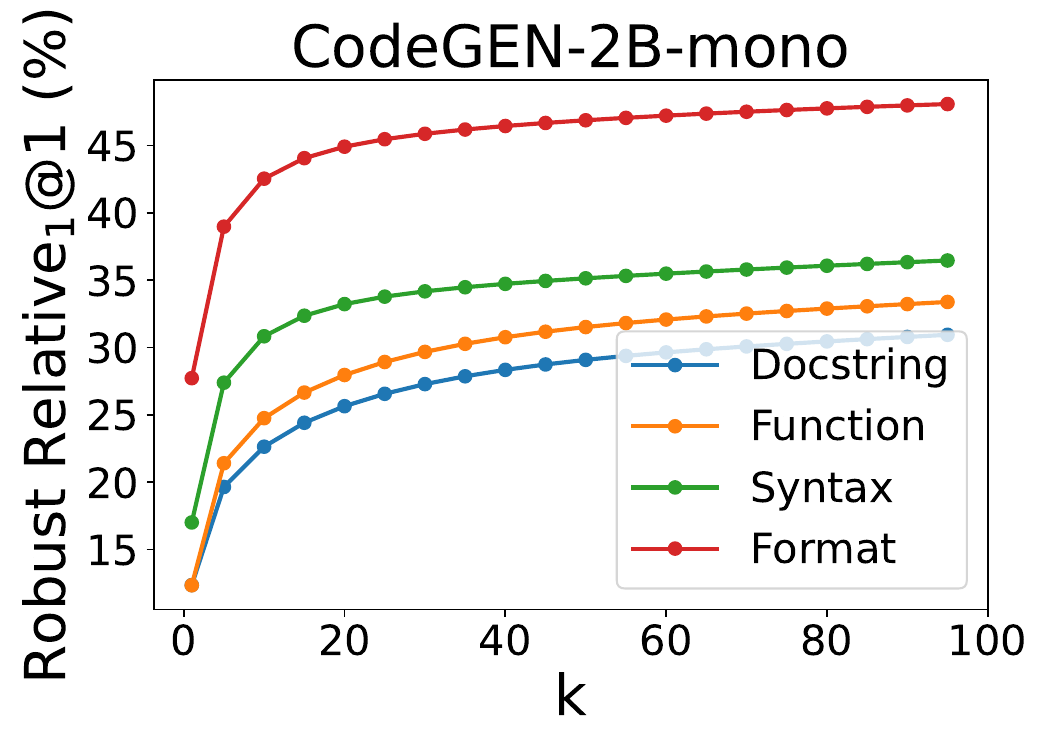}

    \includegraphics[width=0.31\linewidth]{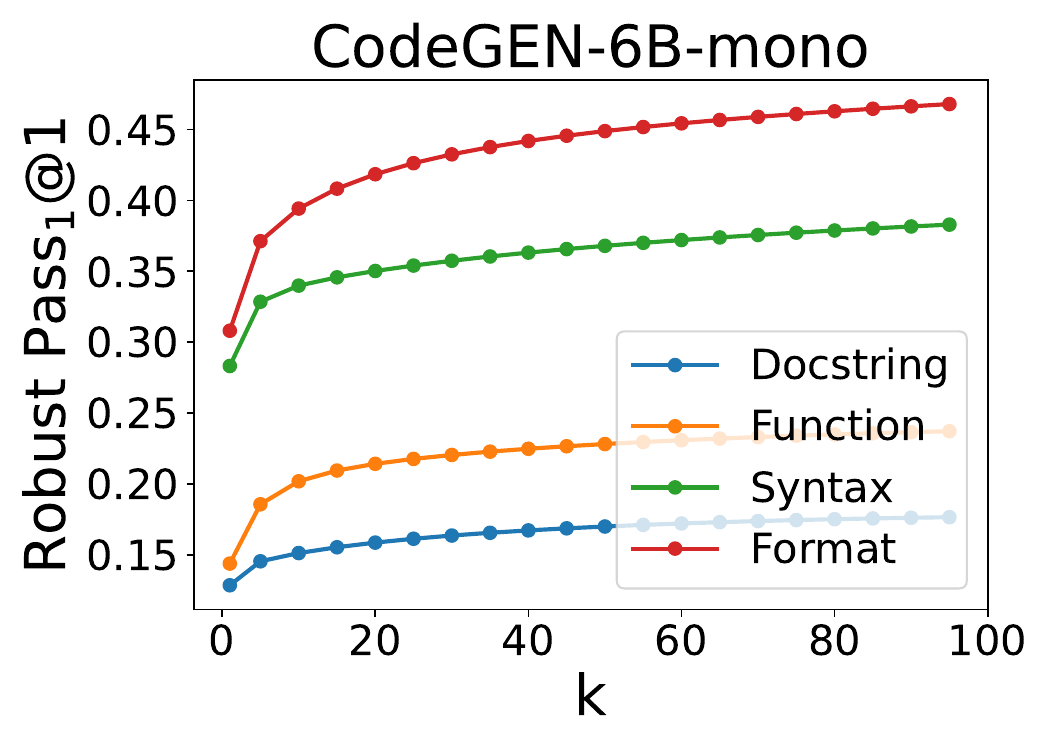}
    \includegraphics[width=0.31\linewidth]{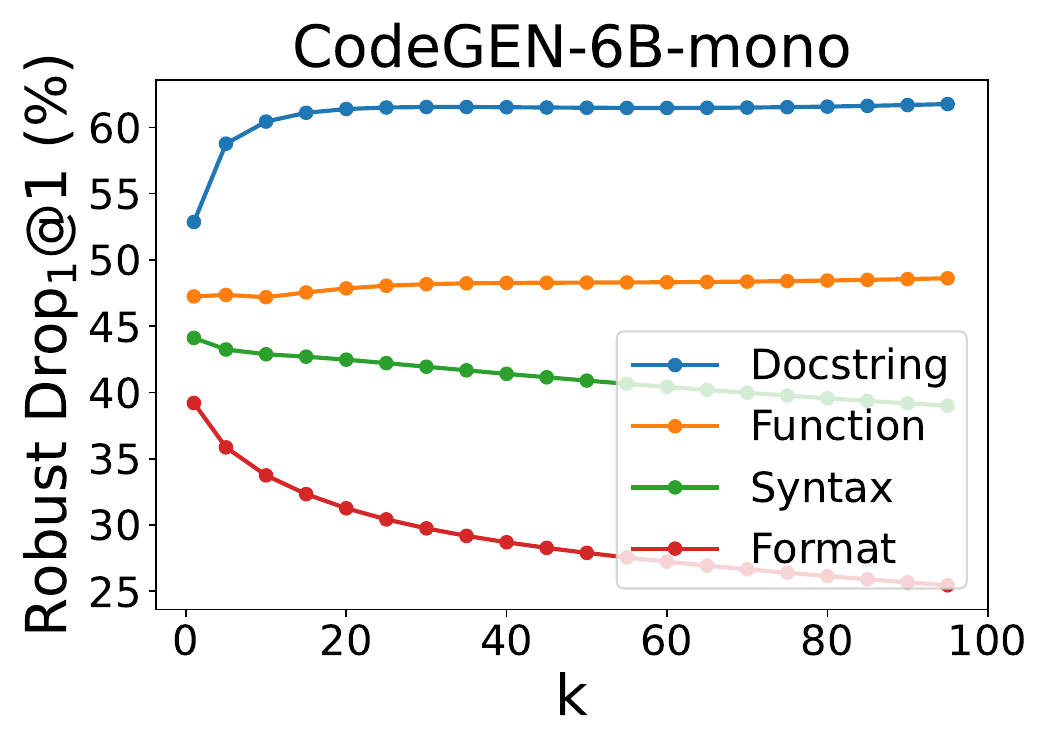}
    \includegraphics[width=0.31\linewidth]{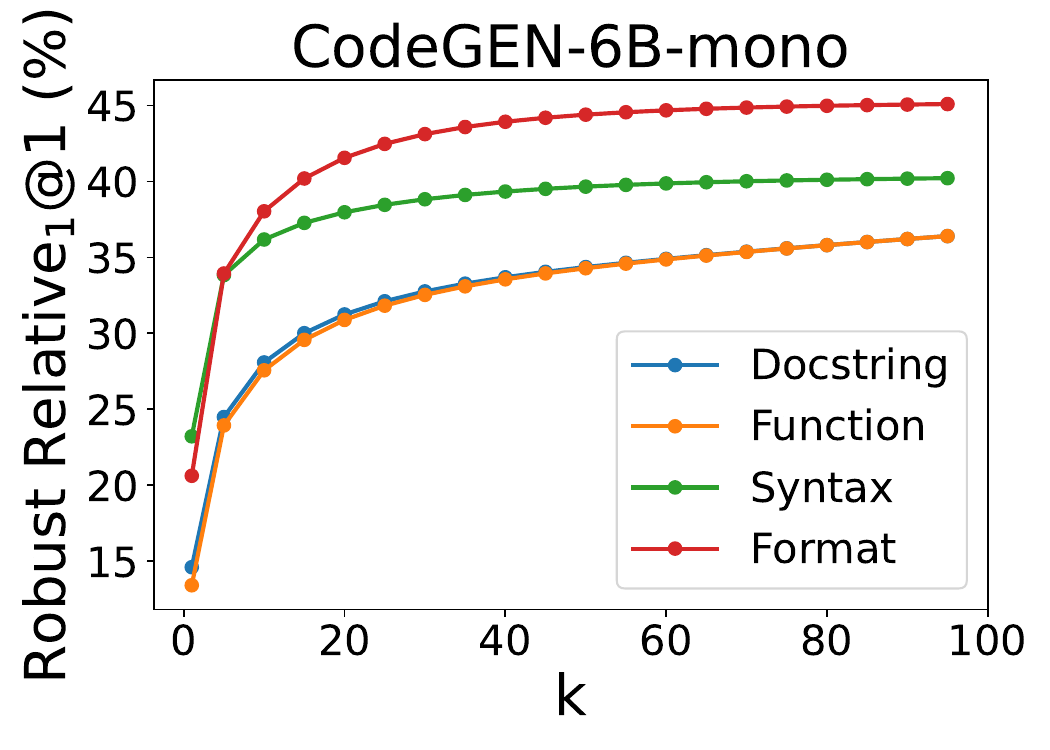}

    \includegraphics[width=0.31\linewidth]{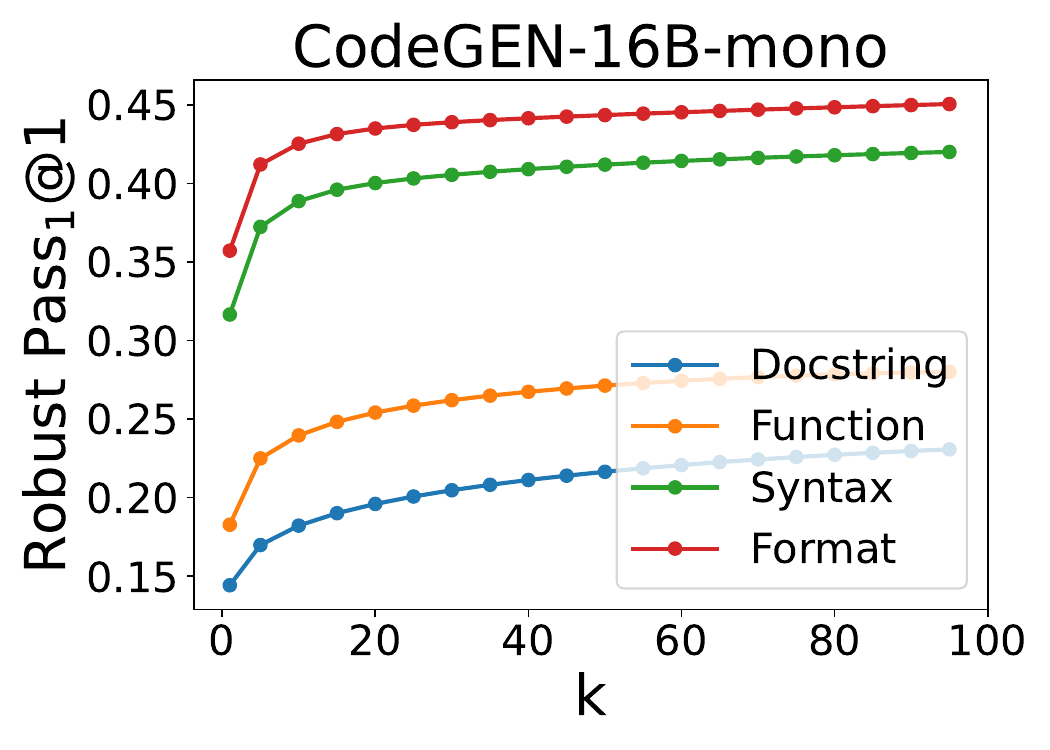}
    \includegraphics[width=0.31\linewidth]{img217.pdf}
    \includegraphics[width=0.31\linewidth]{img218.pdf}
    \caption{Robust Drop$_1$@1 and Robust Relative$_1$@1 on CodeGen-16B-mono under different $k$ using sampling $n=100$. Robust Drop remains stable while Robust Pass and Robust Relative increases with k.}
    \label{appd: figk}
\end{figure*}

As discussed in~\cref{subsec: ablation}, we observe that Robust Drop stays stable across different k while Robust Relative increases linearly with k. We present additional results on CodeGen-2B-mono, CodeGen-6B-mono along with CodeGen-16B-mono in~\cref{appd: figk}. We evaluate each model with large $n~ (n=100)$ using top-p sampling strategy with probability $0.95$ and temperature $0.2$.

\subsection{Additional Results for Large Sampling n}
\label{appd: n}

Larger sampling $n$ is commonly used for preventing model generation variances and providing accurate estimations. The evaluation cost increases linearly to $n$. Here we show that larger $n$ can also benefit our proposed three robustness metrics but not causing significant differences. In specific, we measure Robust Pass$_1$@1, Robust Drop$_1$@1, and Robust Relative$_1$@1 on CodeGen-16B-mono and HumanEval dataset. The model is run with $n=100$ using top-p sampling strategy with probability $0.95$ and temperature $0.2$. We present detailed results in~\cref{appd: tabn}.

\begin{table*}[ht]
\centering
\footnotesize
\setlength{\tabcolsep}{3pt}
\scalebox{0.95}{
\begin{tabular}{r|r|r|r|r} \toprule
\multicolumn{1}{l}{Category} & \multicolumn{1}{c|}{Metric} & $n=1$   & $n=10$  & $n=100$   \\ \midrule
\multirow{4}{*}{Docstring}                & Nominal$\uparrow$                      & 0.287 & 0.308 & 0.306 \\
                                          & RP$_1$@1$\uparrow$              & 0.128 & 0.140 & 0.143 \\
                                          & RD$_1$@1(\%)$\downarrow$                 & 55.32 & 54.46 & 53.34   \\
                                          & RR$_1$@1(\%)$\downarrow$             & 15.85 & 16.77 & 16.55   \\ \midrule
\multirow{4}{*}{Function}                 & Nominal$\uparrow$                    & 0.287 & 0.308 & 0.306   \\
                                          & RP$_1$@1$\uparrow$              & 0.183 & 0.180 & 0.183   \\
                                          & RD$_1$@1(\%)$\downarrow$                 & 36.17 & 41.39 & 40.37   \\
                                          & RR$_1$@1(\%)$\downarrow$             & 10.37 & 12.99 & 13.24   \\ \midrule
\multirow{4}{*}{Syntax}                   & Nominal$\uparrow$                    & 0.561 & 0.542 & 0.544   \\
                                          & RP$_1$@1$\uparrow$              & 0.220 & 0.234 & 0.244   \\
                                          & RD$_1$@1(\%)$\downarrow$                 & 60.87 & 56.81 & 55.19   \\
                                          & RR$_1$@1(\%)$\downarrow$             & 34.15 & 31.04 & 30.39   \\ \midrule
\multirow{4}{*}{Format}                   & Nominal$\uparrow$                    & 0.561 & 0.542 & 0.544   \\
                                          & RP$_1$@1$\uparrow$              & 0.341 & 0.352 & 0.357   \\
                                          & RD$_1$@1(\%)$\downarrow$                 & 39.13 & 34.98 & 34.36   \\
                                          & RR$_1$@1(\%)$\downarrow$             & 21.95 & 19.70 & 19.36  
\\ \bottomrule
\end{tabular}
}
\caption{Generation variances for three robustness metrics with different sampling $n$ on CodeGen-16B-mono and HumanEval.}
\label{appd: tabn}
\end{table*}

%% file: past_papers/hindsight/sections/appendix.tex
\section{Accuracy Decomposition over 4 responses}
\label{app:fig_perf_decomp_art_resp}
\begin{figure}[h]
\centering
  \includegraphics[width=0.4\textwidth]{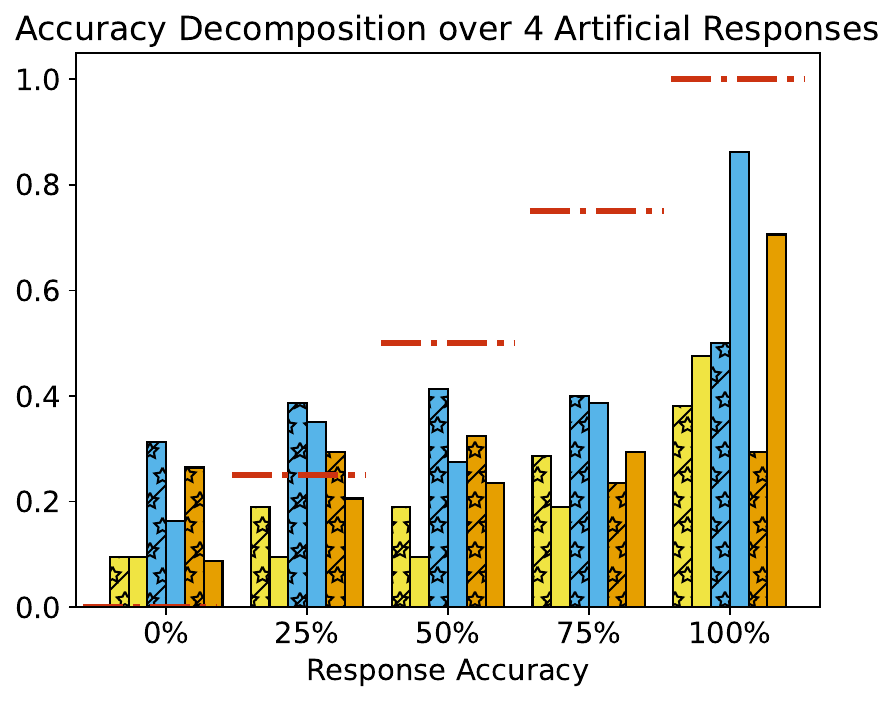}
  \includegraphics[width=0.4\textwidth]{img180.pdf}
  \caption{Accuracy vs. Correctness Margin for each artificial response 
  }
  \label{fig:4level_fake}
\end{figure}

See \cref{fig:4level_fake} for accuracy decomposition over 4 responses using ChatGPT.

\section{Artificial Response Generation}
\label{app:artificial_response_generation}
\begin{figure}[htbp!]
\centering
  \includegraphics[width=0.42\textwidth]{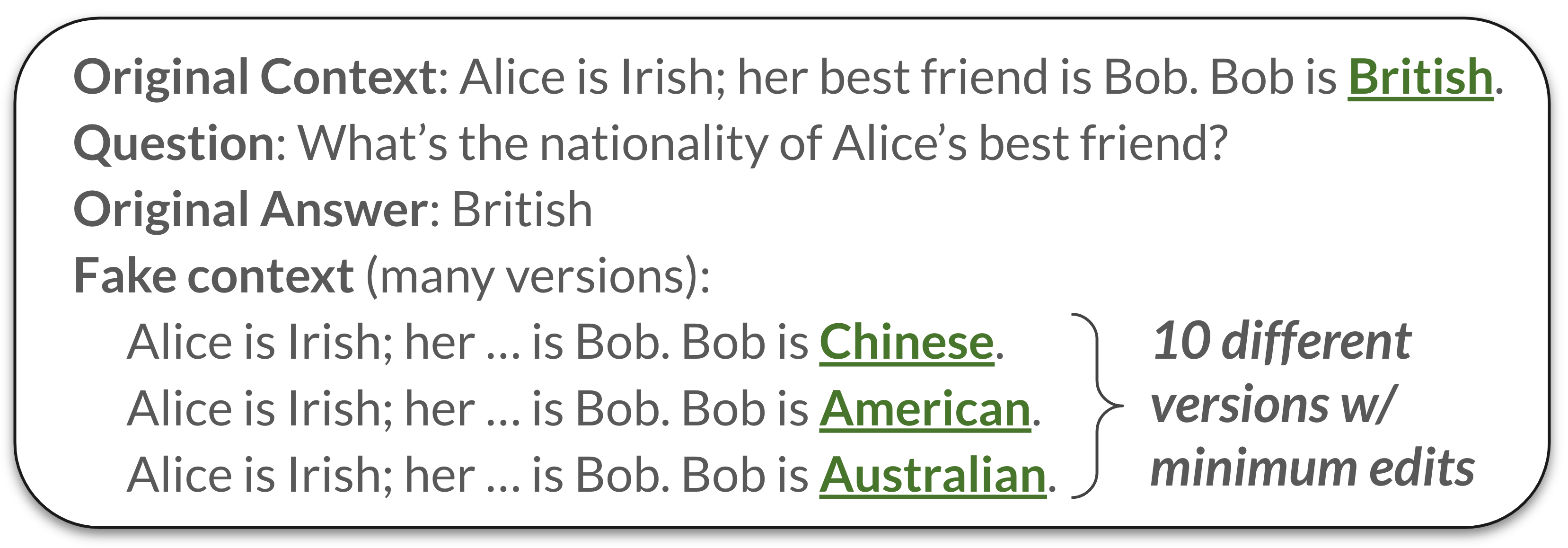}
  \vspace{-3mm}
  \caption{Synthesized Artificial Contexts Example 
  }
    \label{fig:FakeContext}
    \vspace{-5mm}
\end{figure}

We do artificial response generation by prompting ChatGPT to edit the context used in HotpotQA. 
Specifically, the following steps were adopted:
    1) For chosen questions, perform a simple perturbation on the context (e.g., entity replacement). An example is shown in \cref{fig:FakeContext}.
    2) Manually inspect some samples to ensure minimal edits and answerability.
    3) Prompt the model to regenerate responses and reflections based on the altered context. 
In this way, we are simulating scenarios where the model doesn't comprehend the context perfectly. \footnote{While directly editing outputs to create correct or incorrect answers is an option, we avoid this to ensure the results reflect the model's natural response distribution.}

Here is an example for how we modify the context:

\textbf{Original question}: What nationality was James Henry Miller's wife?

\textbf{Original context}: ... Ewan MacColl: James Henry Miller (25 January 1915 – 22 October 1989), better known by his stage name Ewan MacColl, was an \text{\color{blue} English} folk singer, songwriter, 
\text{\color{blue} communist}, labour activist, actor, poet, playwright and record producer. Peggy Seeger: Margaret "Peggy" Seeger (born June 17, 1935) is an American \text{\color{blue} folksinger}. She is also well known in \text{\color{blue} Britain}, where she has lived for more than 30 years, and was married to \text{\color{blue} the singer and songwriter}  Ewan MacColl until his death in 1989. ...

\textbf{Fake context 1}: ... Ewan MacColl: James Henry Miller (25 January 1915 – 22 October 1989), better known by his stage name Ewan MacColl, was a \text{\color{blue} Scottish} folk singer, songwriter, \text{\color{blue} capitalist}, labour activist, actor, poet, playwright and record producer.. Peggy Seeger: Margaret "Peggy" Seeger (born June 17, 1935) is an American \text{\color{blue} country} singer. She is also well known in \text{\color{blue} France}, where she has lived for more than 30 years, and was married to \text{\color{blue} the actor and playwright} Ewan MacColl until his death in 1989. ...

\textbf{Fake context 2}: ... Ewan MacColl: James Henry Miller (25 January 1915 – 22 October 1989), better known by his stage name Ewan MacColl, was an \text{\color{blue} Australian} folk singer, songwriter, \text{\color{blue} conservative}, labour activist, actor, poet, playwright and record producer. Peggy Seeger: Margaret "Peggy" Seeger (born June 17, 1935) is a \text{\color{blue}British pop} singer. She is also well known in \text{\color{blue}Germany}, where she has lived for more than 30 years, and was married to \text{\color{blue}the musician and producer} Ewan MacColl until his death in 1989. ...

\textbf{Fake context 3}: ... Ewan MacColl: James Henry Miller (25 January 1915 – 22 October 1989), better known by his stage name Ewan MacColl, was a \text{\color{blue} Canadian} folk singer, songwriter, \text{\color{blue} anarchist}, labour activist, actor, poet, playwright and record producer. Peggy Seeger: Margaret "Peggy" Seeger (born June 17, 1935) is an \text{\color{blue} American rapper}. She is also well known in \text{\color{blue}Spain}, where she has lived for more than 30 years, and was married to \text{\color{blue}the actor and politician} Ewan MacColl until his death in 1989. ...

\textbf{Fake context 4}: ... Ewan MacColl: James Henry Miller (25 January 1915 – 22 October 1989), better known by his stage name Ewan MacColl, was an \text{\color{blue}Irish} folk singer, songwriter, \text{\color{blue}monarchist}, labour activist, actor, poet, playwright and record producer. Peggy Seeger: Margaret "Peggy" Seeger (born June 17, 1935) is a \text{\color{blue}French jazz singer}. She is also well known in \text{\color{blue}Italy}, where she has lived for more than 30 years, and was married to \text{\color{blue}the artist and filmmaker} Ewan MacColl until his death in 1989. ...

\begin{table}[ht!]
\centering
\small
\resizebox{0.48\textwidth}{!}{%
\begin{tabular}{@{}lccc@{}}
\toprule
Metric & {Standard Prompting} & Exploration-Only & Self-Reflection \\
\midrule
\multicolumn{4}{c}{TruthfulQA} \\
Rouge-1 & $57.5 \pm 1.1$ & $55.1$ & $\textbf{59.0}$ \\
BLEURT & $66.8 \pm 1.9$ & $70.1$ & $\textbf{72.9}$ \\
\midrule
\multicolumn{4}{c}{HotpotQA} \\
Accuracy & $80.2 \pm 0.4$ & $69.7$ & $71.9$ \\
\bottomrule
\end{tabular}
}
\caption{Self-Reflection experiment results using iterative prompting. Bold-faced numbers at each row indicate the best-performing method under each metric. }
\label{tab: conditional_prompting}
\vspace{-10pt}
\end{table}

\section{Conditional Prompting Results}
\label{app:conditional_prompting}
We demonstrate the conditional prompting results in \cref{tab: conditional_prompting}. Comparing the results in \cref{tab: merged_res} and \cref{tab: conditional_prompting}, we can see that there is no significant difference between these parallel prompting and conditional prompting. To avoid the implicit bias introduced by conditional prompting, as \citet{huang2023large} point out, we stick to parallel prompting to conduct our evaluation on self-reflective thinking capability. 

\section{Evaluation details for TruthfulQA}
\label{app:truthfulqa_evaluation}
We use the generation setting of TruthfulQA, which evaluates by comparing how closely the model's responses match a preferred reference versus an undesired one
We follow~\citep{lin2022truthfulqa} to use Rouge-1~\citep{lin-2004-rouge} and BLEURT~\citep{sellam-etal-2020-bleurt} for similarity computation.

\section{Prompts used in Experiment}
\label{app: prelim-prompt-templates}

\subsection{TruthfulQA: Standard Prompt}
\begin{small}
\begin{lstlisting}
messages=[
    {"role": "user", 
    "content": question}
]
\end{lstlisting}
\end{small}

\subsection{TruthfulQA: Response Critique Prompt}
\begin{small}
\begin{lstlisting}
messages=[
    {"role": "system", 
    "content": "You are a helpful 
    assistant."},
    {"role": "user", 
    "content": question},
    {"role": "assistant", 
    "content": response},
    {"role": "user", 
    "content": "Could you critique 
    your last response?"}
]
\end{lstlisting}
\end{small}
\clearpage
\subsection{TruthfulQA: Response Without Reflection}
\begin{small}
\begin{lstlisting}
messages=[
    {"role": "system", 
    "content": "You are a helpful 
    assistant."},
    {"role": "user", 
    "content": question},
    {"role": "assistant", 
    "content": response_1},
    {"role": "user", 
    "content": question},
    {"role": "assistant", 
    "content": response_2},
    {"role": "user", 
    "content": question},
    {"role": "assistant", 
    "content": response_3},
    {"role": "user", 
    "content": question}
]
\end{lstlisting}
\end{small}

\subsection{TruthfulQA: Response With Reflection}
\begin{small}
\begin{lstlisting}
messages=[
    {"role": "system", 
    "content": "You are a helpful 
    assistant."},
    {"role": "user", 
    "content": question},
    {"role": "assistant", 
    "content": response_1},
    {"role": "user", 
    "content": "Please critique your
    responses"},
    {"role": "assistant", 
    "content": critique_1},
    {"role": "user", 
    "content": question},
    {"role": "assistant", 
    "content": response_2},
    {"role": "user", 
    "content": "Please critique your
    responses"},
    {"role": "assistant", 
    "content": critique_2},
    {"role": "user", 
    "content": question},
    {"role": "assistant", 
    "content": response_3},
    {"role": "user", 
    "content": "Please critique your
    responses"},
    {"role": "assistant", 
    "content": critique_3},
    {"role": "user", 
    "content": question}
]
\end{lstlisting}
\end{small}

\subsection{HotpotQA: Standard Prompt}
\begin{small}
\begin{lstlisting}
messages=[
    {"role": "system", 
    "content": "You are a helpful 
    assistant. Answer the question 
    based on the context provided. 
    Provide extremely concise answers 
    with no explanation."},
    {"role": "user", 
    "content": "Context: Earth: The
    Earth is the third planet from 
    the Sun. Question: Which planet 
    is Earth from the Sun? Answer: 
    Third"},
    {"role": "user", 
    "content": f"Context: 
    {formatted_context}\n
    Question: {question}\nProvide a 
    short answer without 
    explanation."}
]
\end{lstlisting}
\end{small}

\subsection{HotpotQA: Response Critique
Prompt}
\begin{small}
\begin{lstlisting}
messages=[
    {"role": "system", 
    "content": "You are a helpful 
    assistant. Answer the question 
    based on the context provided."},
    {"role": "user", 
    "content": f"Context: 
    {formatted_context}\n
    Question: {question}"},
    {"role": "assistant", 
    "content": f"{response}"},
    {"role": "user", 
    "content": f"Please review and 
    critique your previous response, 
    and keep in mind not to add any 
    unnecessary apologies. You can 
    refer back to the original 
    context if needed."}
]
\end{lstlisting}
\end{small}

\subsection{HotpotQA: Response Without
Reflection}
\begin{small}
\begin{lstlisting}
messages=[
    {"role": "system", 
    "content": "You are a helpful 
    assistant. Answer the question 
    based on the context provided. 
    Provide extremely concise answers 
    with no explanation."},
    {"role": "user", 
    "content": "Context: Earth: The 
    Earth is the third planet from 
    the Sun. Question: Which planet 
    is Earth from the Sun? 
    Answer: Third"},
    {"role": "user", 
    "content": f"Context: 
    {formatted_context}\n
    Question: {question}\n
    Provide a short answer without 
    explanation."},
    {"role": "assistant", 
    "content": f"{response_1}"},
    {"role": "user", 
    "content": f"{question}\n
    Provide a short answer without
    explanation."},
    {"role": "assistant", 
    "content": f"{response_2}"},
    {"role": "user", 
    "content": f"{question}\n
    Provide a short answer without 
    explanation."},
    {"role": "assistant", 
    "content": f"{response_3}"},
    {"role": "user", 
    "content": f"{question}\n
    Provide a short answer without
    explanation."},
    {"role": "assistant", 
    "content": f"{response_4}"},
    {"role": "user", 
    "content": f"{question}\n
    Provide a short answer without
    explanation."},
]
\end{lstlisting}
\end{small}

\subsection{HotpotQA: Response With
Reflection}
\begin{small}
\begin{lstlisting}
messages=[
    {"role": "system", 
    "content": "You are a helpful 
    assistant. Answer the question 
    based on the context provided. 
    Provide extremely concise answers 
    with no explanation."},
    {"role": "user", 
    "content": "Context: Earth: The 
    Earth is the third planet from the 
    Sun. Question: Which planet is Earth 
    from the Sun? Answer: Third"},
    {"role": "user", 
    "content": f"Context: 
    {formatted_context}\n
    Question: {question}\n
    Provide a short answer without 
    explanation."},
    {"role": "assistant", 
    "content": f"{response_1}"},
    {"role": "user", 
    "content": f"Please review and 
    critique your previous response, 
    and keep in mind not to add any 
    unnecessary apologies. You can 
    refer back to the original context 
    if needed."},
    {"role": "assistant", 
    "content": f"{critique_1}"},
    {"role": "user", 
    "content": f"{question}\n
    Provide a short answer without
    explanation."},
    {"role": "assistant", 
    "content": f"{response_2}"},
    {"role": "user", 
    "content": f"Please review and 
    critique your previous response, 
    and keep in mind not to add any 
    unnecessary apologies. You can 
    refer back to the original context 
    if needed."},
    {"role": "assistant", 
    "content": f"{critique_2}"},
    {"role": "user", 
    "content": f"{question}\n
    Provide a short answer without 
    explanation."},
    {"role": "assistant", 
    "content": f"{response_3}"},
    {"role": "user", 
    "content": f"Please review and 
    critique your previous response, 
    and keep in mind not to add any 
    unnecessary apologies. You can 
    refer back to the original 
    context if needed."},
    {"role": "assistant", 
    "content": f"{critique_3}"},
    {"role": "user", 
    "content": f"{question}\n
    Provide a short answer without 
    explanation."},
    {"role": "assistant", 
    "content": f"{response_4}"},
    {"role": "user", 
    "content": f"Please review and 
    critique your previous response, 
    and keep in mind not to add any 
    unnecessary apologies. You can 
    refer back to the original context 
    if needed."},
    {"role": "assistant", 
    "content": f"{critique_4}"},
    {"role": "user", 
    "content": f"{question}\n
    Provide a short answer without
    explanation."}
]
\end{lstlisting}
\end{small}

\subsection{HotpotQA: Fake Evidence Generation}
\begin{small}
\begin{lstlisting}
messages=[
    {"role": "system", 
    "content": "You are a helpful 
    assistant."},
    {"role": "user", 
    "content": f"Here is a question: 
    {question}. Please create 10 
    different versions of 'fake 
    supporting facts' based on the 
    following real supporting facts. 
    Modify only one sentence in each 
    version, making sure the modified 
    sentence is still relevant but 
    contains false information. Keep 
    the other sentences unmodified. 
    Each version of fake supporting 
    facts should have the same number 
    of sentences as the real 
    supporting facts."},
    {"role": "user", 
    "content": f"Real Supporting 
    Facts:{real_sf}"},
    {"role": "user", 
    "content": "Please generate the 
    fake supporting facts versions. 
    Remember to index all the sentences. 
    You must generate 10 versions 
    before you stop."},
    {"role": "user", 
    "content": 
    f"Fake Supporting Facts Version 1:\n
    [Insert manipulated sentences here]\n
    Fake Supporting Facts Version 2:\n
    [Insert manipulated sentences here]\n
    Fake Supporting Facts Version 3:\n
    [Insert manipulated sentences here]\n
    Fake Supporting Facts Version 4:\n
    [Insert manipulated sentences here]\n
    Fake Supporting Facts Version 5:\n
    [Insert manipulated sentences here]\n
    Fake Supporting Facts Version 6:\n
    [Insert manipulated sentences here]\n
    Fake Supporting Facts Version 7:\n
    [Insert manipulated sentences here]\n
    Fake Supporting Facts Version 8:\n
    [Insert manipulated sentences here]\n
    Fake Supporting Facts Version 9:\n
    [Insert manipulated sentences here]\n
    Fake Supporting Facts Version 10:\n
    [Insert manipulated sentences here]"},
]
\end{lstlisting}
\end{small}

\section{Illustration of API Calling Processes}
\label{app:api_call_example}
In this section, we provide a simple example to illustrate the API calling process under our \protocol~, conditional prompting and the Exploration-Only Baseline. 

\subsection{\protocol~ API Calling Process}
\begin{small}
\begin{lstlisting}
(Splitters and other special tokens are omitted)
*First API Call*:
[Instructions and Context]
Question: [question]
Response: ____ (Sample response_1, response_2 here.)

*Second API Call*:
[Instructions and Context]
Question: [question]
Response: response_1
[Instruction for Reflection]
Reflection: ____ (Sample reflection_1 here.)

*Third API Call*:
[Instructions and Context]
Question: [question]
Response: response_2 
[Instruction for Reflection]
Reflection: ____ (Sample reflection_2 here.)

*Final API Call (to get the final revised answer)*:
[Instructions and Context]
Question: [question]
Response: response_1
[Instruction for Reflection]
Reflection: reflection_1

Question: [question]
Response: response_2
[Instruction for Reflection]
Reflection:  reflection_2

Question: [question]
Response:  ____ (Sample final_response here)
\end{lstlisting}
\end{small}

\subsection{Conditional Prompting Baseline API Calling Process}
\begin{small}
\begin{lstlisting}
*First API Call*:
[Instructions and Context]
Question: [question]
Response: ____ (sample response_1 here)

*Second API Call*:
[Instructions and Context]
Question: [question]
Response: response_1
[Instruction for Reflection]
Reflection: ____ (sample reflection_1 here)

*Third API Call*:
[Instructions and Context]
Question: [question]
Response: response_1
[Instruction for Reflection]
Reflection: reflection_1

Question: [question]
Response: ___ (sample response_2 here)

...

*Final API Call (to get the final revised answer)*:
[Instructions and Context]
Question: [question]
Response: response_1
[Instruction for Reflection]
Reflection: reflection_1

Question: [question]
Response: response_2
[Instruction for Reflection]
Reflection:  reflection_2

Question: [question]
Response:  final_reponse
\end{lstlisting}
\end{small}

\subsection{Exploration-Only Baseline API Calling Process}
\begin{small}
\begin{lstlisting}
*First API Call*:

[Instructions and Context]
Question: [question]
Response: ____ (sample response_1, response_2 here)

*Final API Call*:

[Instructions and Context]
Question: [question]
Response: response_1
Question: [question]
Response: response_2 
Question: [question]
Response: ____ (sample final_response here)
\end{lstlisting}
\end{small}

\section{Challenges in Predicting the Correctness Margin for Model Comprehension}
\label{app:challenge_predict_correctness_margin}

The effectiveness of a model's self-reflection largely hinges on its ``correctness margin,'' a metric quantifying its understanding of both the question and its context. Ideally, we would like to predict this margin through user prompts, thereby allowing the user to make an informed decision on whether to enable the model's self-reflection capability.

Nevertheless, our experiments indicate that current models struggle to self-assess their understanding reliably. Below, we outline our prompt design used for this experiment:

\begin{small}
\begin{lstlisting}
messages=[
    {"role": "system", 
    "content": "You are a helpful
    assistant. Answer the question based 
    on the context provided. Provide 
    extremely concise answers with no 
    explanation."},
    {"role": "user", 
    "content": f"Context:
    {formatted_context}\n
    Question: {question}"},
    {"role": "assistant", 
    "content": f"{response}"},
    {"role": "user", 
    "content": "\nYou have just answered 
    a question. Now, please evaluate your 
    own comprehension of the question and 
    answer provided. Rate your level of 
    understanding on a scale from -5 to 5. 
    A rating of 5 signifies extreme 
    certainty that you understand the 
    question, while a rating of -5 
    indicates extreme uncertainty or lack 
    of understanding."},
]
\end{lstlisting}
\end{small}

We tested this prompt structure on two sets of questions: one where all $10$ model responses were incorrect, and another where all $10$ were correct. If the model were capable of accurately evaluating its own comprehension, it should consistently rate its understanding at $-5$ for questions in the all-wrong dataset and $5$ for those in the all-right dataset. However, after experimenting with $20$ examples from each dataset, we found that the model consistently assigned high scores (typically $4$ or $5$) regardless of the dataset origin. Thus, reliable self-assessment remains an open challenge for current models.

\section{Scientific Artifacts}
\label{app:scientific_artifacts}

In this paper, we use the following artifacts:
\begin{itemize}
    \item TruthfulQA \citep{lin2022truthfulqa} is a benchmark assessing a language model's ability to generate truthful answers for 817 diverse questions in 38 categories, requiring models to avoid false answers commonly found in human texts due to misconceptions or false beliefs. We use it for the preliminary studies on reflective thinking in LLMs. It is licensed under the Apache License, Version 2.0.
    \item HotpotQA \citep{yang2018hotpotqa} is a 113k question-answer dataset based on Wikipedia that requires multi-document reasoning, features diverse questions unconstrained by knowledge bases or schemas, provides sentence-level supporting facts for strong supervision and explanation, and introduces a new factoid comparison question type to evaluate QA systems' extraction and comparison abilities. We use it for evaluating reflective thinking in LLMs. It is distributed under a CC BY-SA 4.0 License.
    \item openai-python\footnote{https://github.com/openai/openai-python} (v0.27.8) provides convenient access to the OpenAI REST API from any Python 3.7+ application. We use it to access ChatGPT models. It is licensed under the Apache License, Version 2.0.

\end{itemize}

\section{Results on LLaMA-2-7b-chat}
\label{app:llama2_results}
We extend our experiments to the open-sourced model LLaMA-2-7b-chat~\citep{touvron2023llama_2}, and the results support the conclusions that we draw from our experiments on ChatGPT, indicating that our findings can be generalized to different models.

\begin{figure}{}
\centering
\small
\includegraphics[width=0.4\textwidth]{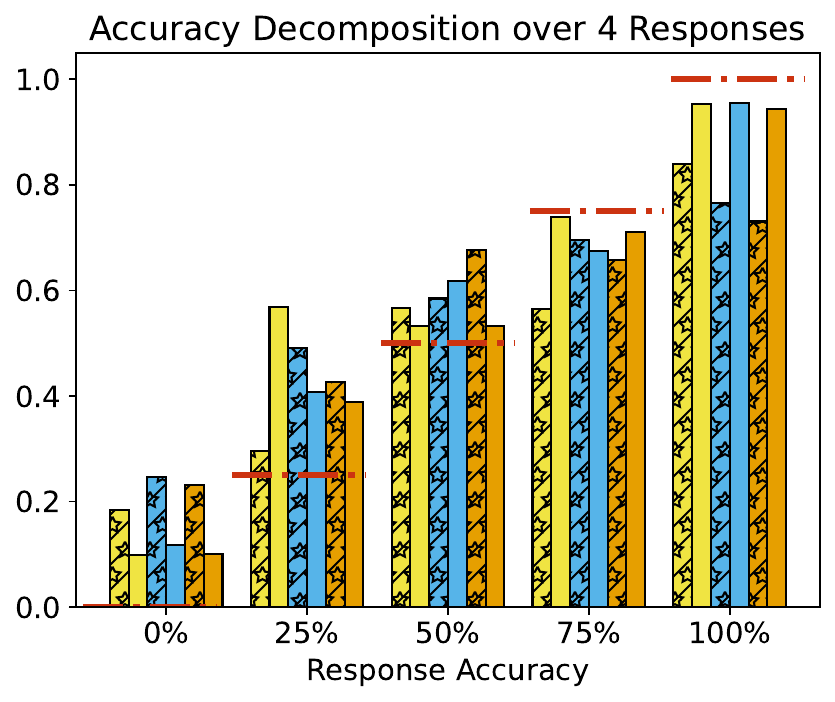}
  \includegraphics[width=0.4\textwidth]{img180.pdf}
  \vspace{-2mm}
  \caption{Performance Decomposition on Question Difficulty and Response Accuracy (LLaMA-2-chat). 
  }

  \label{fig:perf_decomp_4response_natural_llama}
\end{figure}

More specifically, for the preliminary study on performance across TruthfulQA and HotpotQA, the results on LLaMA-2-7b-chat (see \cref{tab: merged_res_llama2}) are consistent with the results obtained from ChatGPT (see \cref{tab: merged_res}): self-reflection prompts improve performance on TruthfulQA while worsening performance on HotpotQA. Additionally, we conduct our error analysis on the results of HotpotQA, breaking down model performance based on levels of question difficulty and model comprehension. The LLaMA-2-7b-chat results for this analysis (\cref{fig:perf_decomp_4response_natural_llama}) also closely follow the trend observed in the results from ChatGPT (\cref{fig:perf_decomp_4response_natural}): self-reflection is more beneficial when the model's initial responses are incorrect and when the question difficulty is higher.

We do not replicate the 10-response artificial experiments on LLaMA-2-7b-chat due to the context length limit. The context length for LLaMA-2 is 4096, which is shorter than our context length for the task. We replicate this experiment in \cref{app:mixtral_results} as Mixtral models have larger token limits.

\begin{table}[t!]
\centering
\resizebox{0.49\textwidth}{!}{%
\begin{tabular}{@{}p{2cm}p{2cm}p{2cm}p{2cm}@{}}
\toprule
Metric & {Standard Prompting} & Exploration-Only & Self-Reflection \\
\midrule
\multicolumn{4}{c}{TruthfulQA} \\
Rouge-1 & $53.8\pm 0.4 $ & $51.7$ & $\textbf{53.8}$ \\
BLEURT & $ 60.9 \pm 0.6$ & $58.2$ & $\textbf{63.0}$ \\
\midrule
\multicolumn{4}{c}{HotpotQA} \\
Accuracy* & $ 61.0\pm 1.0$ & $\textbf{62.9}$ & $57.5$ \\
\bottomrule
\end{tabular}
}
\caption{Self-reflection \protocol{} experiment results on QA datasets using LLaMA-2-chat. Bold-facing indicates the best-performing method under each metric. *Evaluated manually. 
}
\label{tab: merged_res_llama2}
\vspace{-5pt}
\end{table}

\begin{table}[t!]
\centering
\resizebox{0.49\textwidth}{!}{%
\begin{tabular}{@{}p{2cm}p{2cm}p{2cm}p{2cm}@{}}
\toprule
Metric & {Standard Prompting} & Exploration-Only & Self-Reflection \\
\midrule
\multicolumn{4}{c}{TruthfulQA} \\
Rouge-1 & $59.1\pm 1.0 $ & $61.3$ & $\textbf{63.3}$ \\
BLEURT & $ 71.5 \pm 0.4$ & $\textbf{73.9}$ & $71.7$ \\
\midrule
\multicolumn{4}{c}{HotpotQA} \\
Accuracy* & $ 89.8 \pm 0.3$ & $ \textbf{90.9}$ & $89.2$ \\
\bottomrule
\end{tabular}
}
\caption{Self-reflection \protocol{} experiment results on QA datasets using Mixtral-8x7B-v0.1. Bold-facing indicates the best-performing method under each metric. *Evaluated manually. 
}
\label{tab: merged_res_mixtral}
\vspace{-5pt}
\end{table}

\section{Results on Mixtral-8x7B-v0.1}
\label{app:mixtral_results}

\begin{figure}{}
\centering
\small
\includegraphics[width=0.4\textwidth]{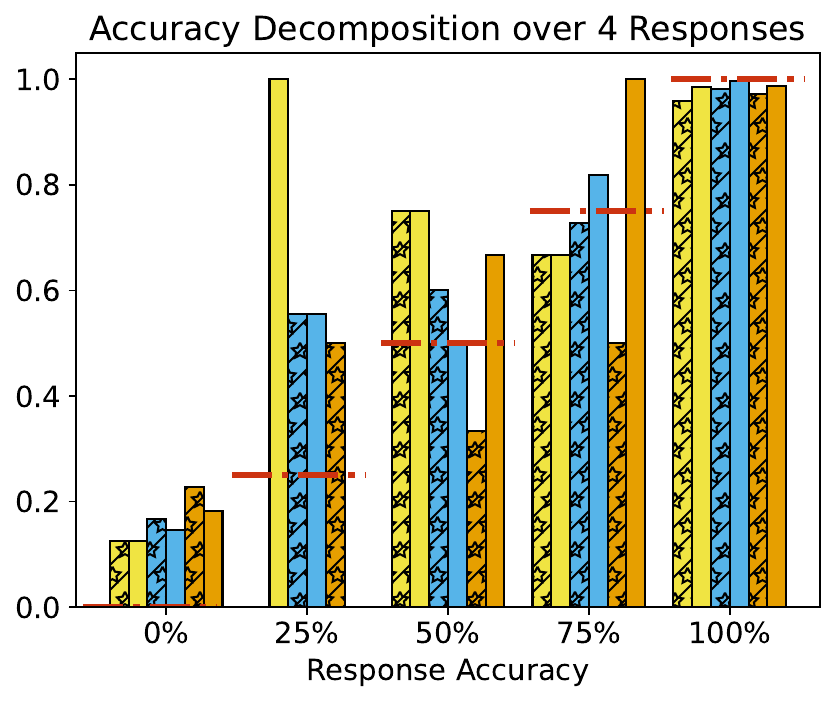}
  \includegraphics[width=0.4\textwidth]{img180.pdf}
  \vspace{-2mm}
  \caption{Performance Decomposition on Question Difficulty and Response Accuracy (Mixtral-8x7B-v0.1). 
  }

  \label{fig:perf_decomp_4response_natural_mixtral}
\end{figure}

\begin{figure}{}
\centering
\small
\includegraphics[width=0.48\textwidth]{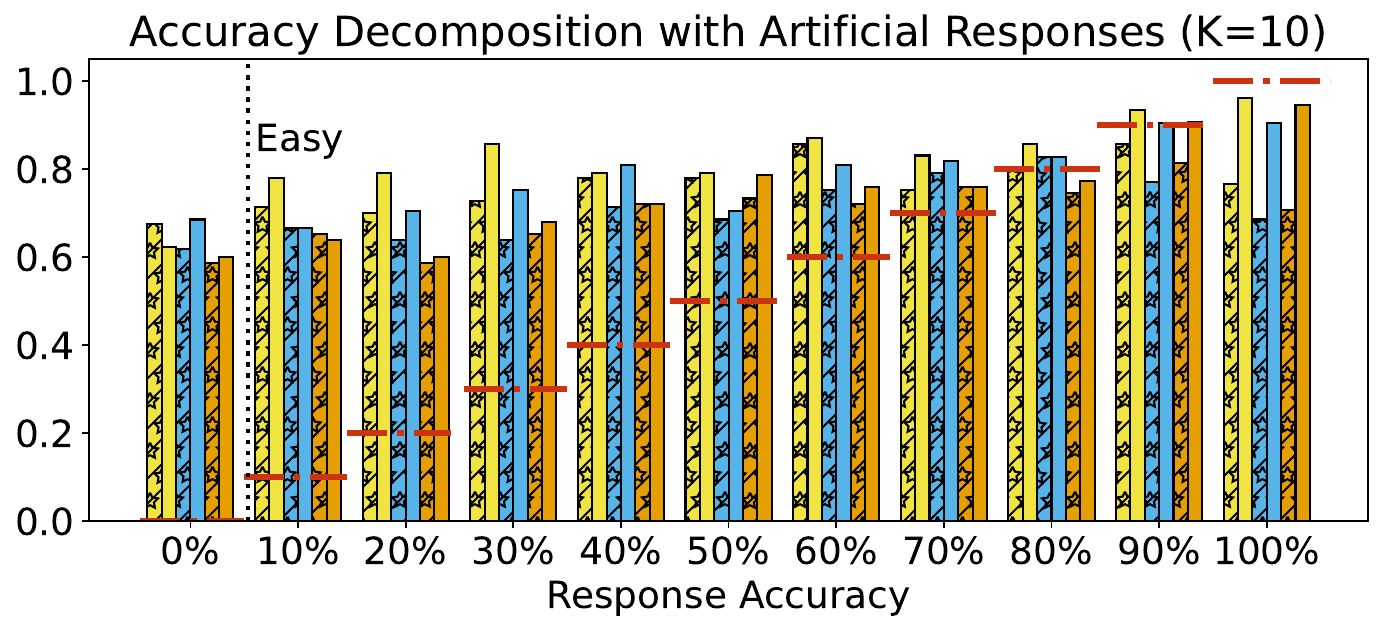}
  \includegraphics[width=0.4\textwidth]{img180.pdf}
  \vspace{-2mm}
  \caption{Performance Decomposition on Question Difficulty and Response Accuracy (Artificial Responses) for Mixtral-8x7B-v0.1.  
  }

  \label{fig:perf_decomp_10response_natural_mixtral}
\end{figure}

We repeat our experiments on Mixtral-8x7B-v0.1. For the preliminary study on performance across TruthfulQA and HotpotQA (\cref{tab: merged_res_mixtral}, we again observe a similar trend to ChatGPT: while self-reflection may help improve the performance on TruthfulQA, it harms the performance on HotpotQA. Then, we again break down model performance on HotpotQA based on question difficulty levels and model comprehension in \cref{fig:perf_decomp_4response_natural_mixtral}. Here we observe a somewhat different pattern: under all question difficulty levels and model comprehension, self-reflection prompting fails to improve the performance. To further verify this finding, we also conduct the artificial response experiments, with results in \cref{fig:perf_decomp_10response_natural_mixtral}. Here we see that self-reflection prompting is not always harmful to performance (e.g., under $0\%$ RA, self-reflection helps improve the performance of easy questions.), but in most cases it is harmful.

\begin{figure}{}
\centering
\small
\includegraphics[width=0.48\textwidth]{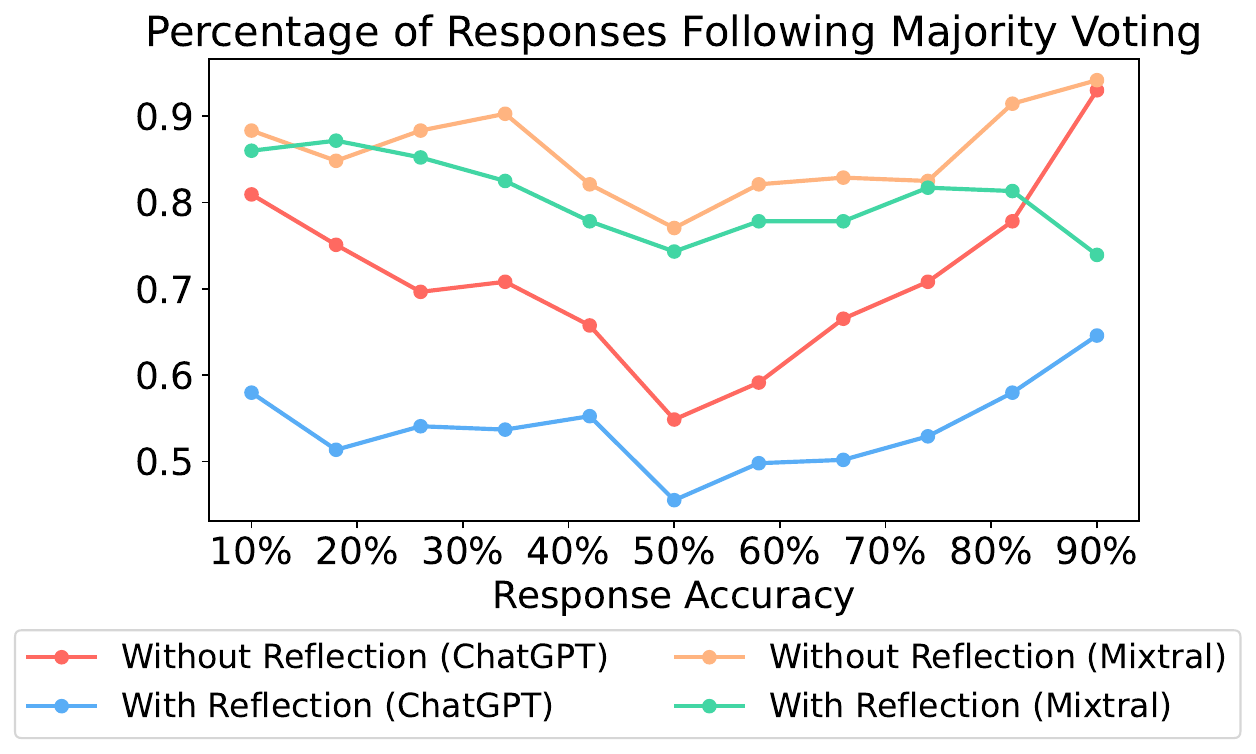}
  \caption{Majority Voting Analysis (Mixtral-8x7B-v0.1). 
  }

  \label{fig:majority_voting_mixtral}
\end{figure}

We hypothesize that these divergent patterns arise because this particular model may be less sensitive in general to instructions for reflection, or less well-equipped to understand them, such that the reflection part serves mostly as a distractor in the input. We examine this hypothesis in the majority voting experiments (\cref{fig:majority_voting_mixtral}) and find that compared with ChatGPT, the addition of self-reflection exerts minimal impact on majority voting trends, suggesting that it is comparatively difficult to use self-reflection prompting to change the default behaviors in the case of this model. This is consistent with our hypothesis that Mixtral-8x7B-v0.1 lacks sensitivity or competence in self-reflection, so we speculate that additional training may be needed to help this model to unlock self-reflection prompting potential.

%% file: past_papers/amortized_interpretability/sections/appendix.tex
\section{Adaption for FastSHAP Baseline}
\label{app: fastshap}
As we mentioned in \cref{sec: experiment}, we build our amortized models upon a pre-trained encoder BERT~\citep{devlin2019bert}.
However, using the pre-trained encoder significantly increases the memory footprint when running FastSHAP. 
In particular, we have to host two language models on GPUs, one for the amortized model and the other one for the target model.  
Therefore, we can only adopt the batch size equal to 1 and $32$ perturbation samples per instance. Following the proof in FastSHAP, this is equivalent to teaching the amortized model to approximate KS-32, which is an unreliable interpretation method (See \cref{sec: training_sensitivity}). 

In experiments, we find that the optimization of FastSHAP is unstable. After an extensive hyper-parameter search, we set the learning rate to 1e-6 and increased the number of epochs to $30$. However, this requires us to train the model on a single A100 GPU for 3 days to wait for FastSHAP to converge.

\section{Scientific Artifacts License}
For the datasets used in this paper, MNLI~\citep{williams2018broad} is released under ONAC's license. Yelp-Polarity~\citep{zhangCharacterlevelConvolutionalNetworks2015} and SST-2~\citep{socher-etal-2013-recursive} datasets does not provide detailed licenses. 

For model checkpoints used in this paper, they all come from textattack project~\citep{morris2020textattack} and they are open-sourced under MIT license. 

For implementation, we mainly use Captum~\citep{kokhlikyan2020captum} and Thermostat~\citep{feldhus2021thermostat}. Captum is open-sourced under BSD 3-Clause "New" or "Revised" License and Thermostat is open-sourced under Apache License 2.0.

\section{Training Details}
\label{app: training_details}
In this section, we introduce our dataset preprocessing, hyperparameter settings and how we train the models. 

For both MNLI and Yelp-Polarity datasets, we split them
into 8:1:1 for training, validation, and test sets.

The hyperparameters of amortized models are tuned on the validation set. We use Adam~\citep{kingma2015adam} optimizer with a learning rate of 5e-5, train the model for at most 10 epochs and do early stopping to select best model checkpoints.

%% file: past_papers/branching_factor/appendix.tex
\section{Case Study Implementation Details}
\label{app: sampling_efforts}
We use the scripts in Qwen-2.5-Math~\citep{yang2024qwen2} for standard reasoning benchmarks.\footnote{\url{https://github.com/QwenLM/Qwen2.5-Math/tree/main}}  We sample $200$ examples from MMLU-STEM and compute the performance numbers under $64$ trials and report the average performance. 

\begin{figure*}[t!]
\centering
\begin{subfigure}[t]{0.24\textwidth}
    \centering
     \includegraphics[width=0.9\linewidth]{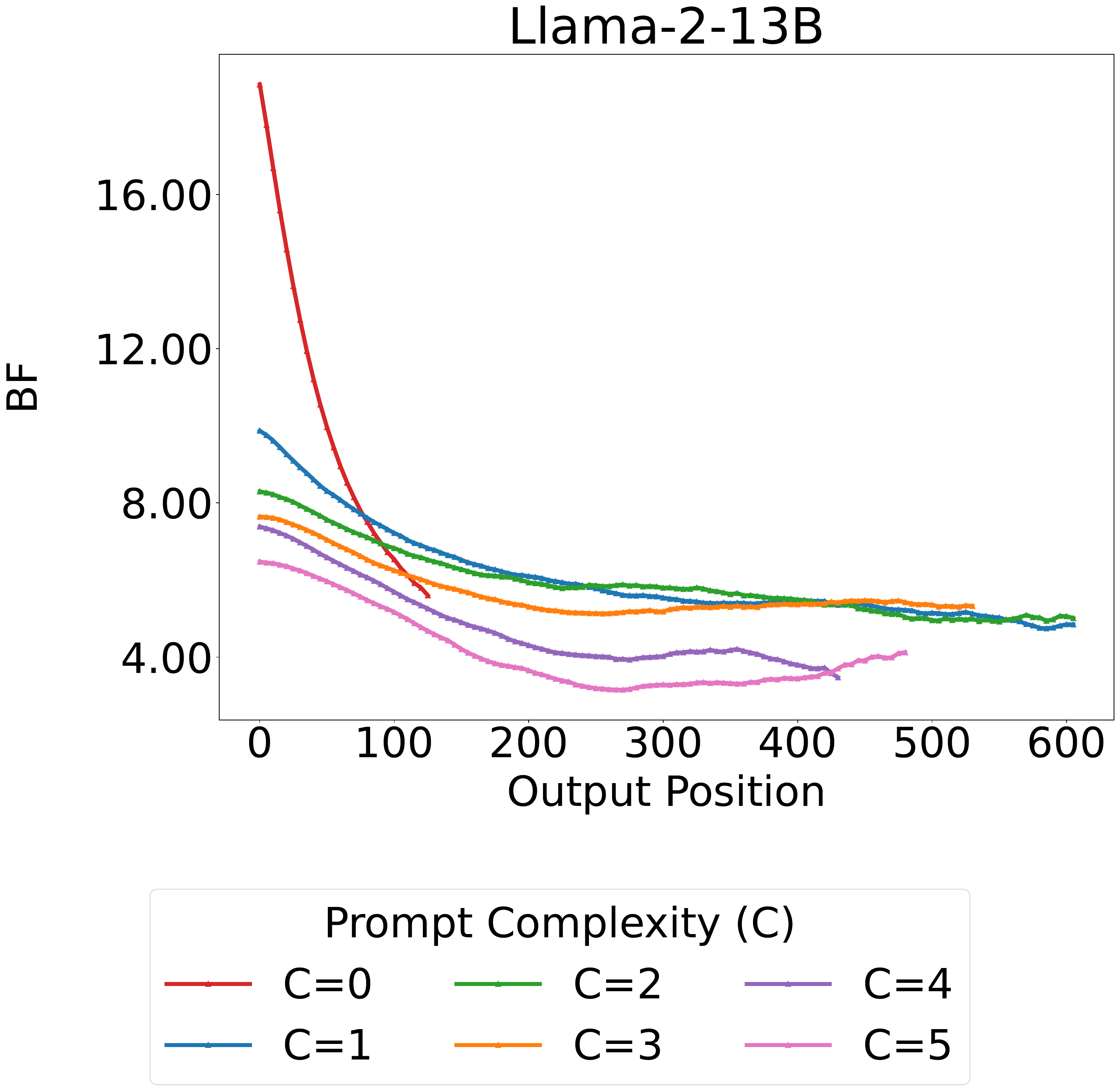}
     \label{fig:output_dynamic_base_storytelling_llama2_13b_app}
    \end{subfigure}
        \begin{subfigure}[t]{0.24\textwidth}
    \centering
     \includegraphics[width=0.9\linewidth]{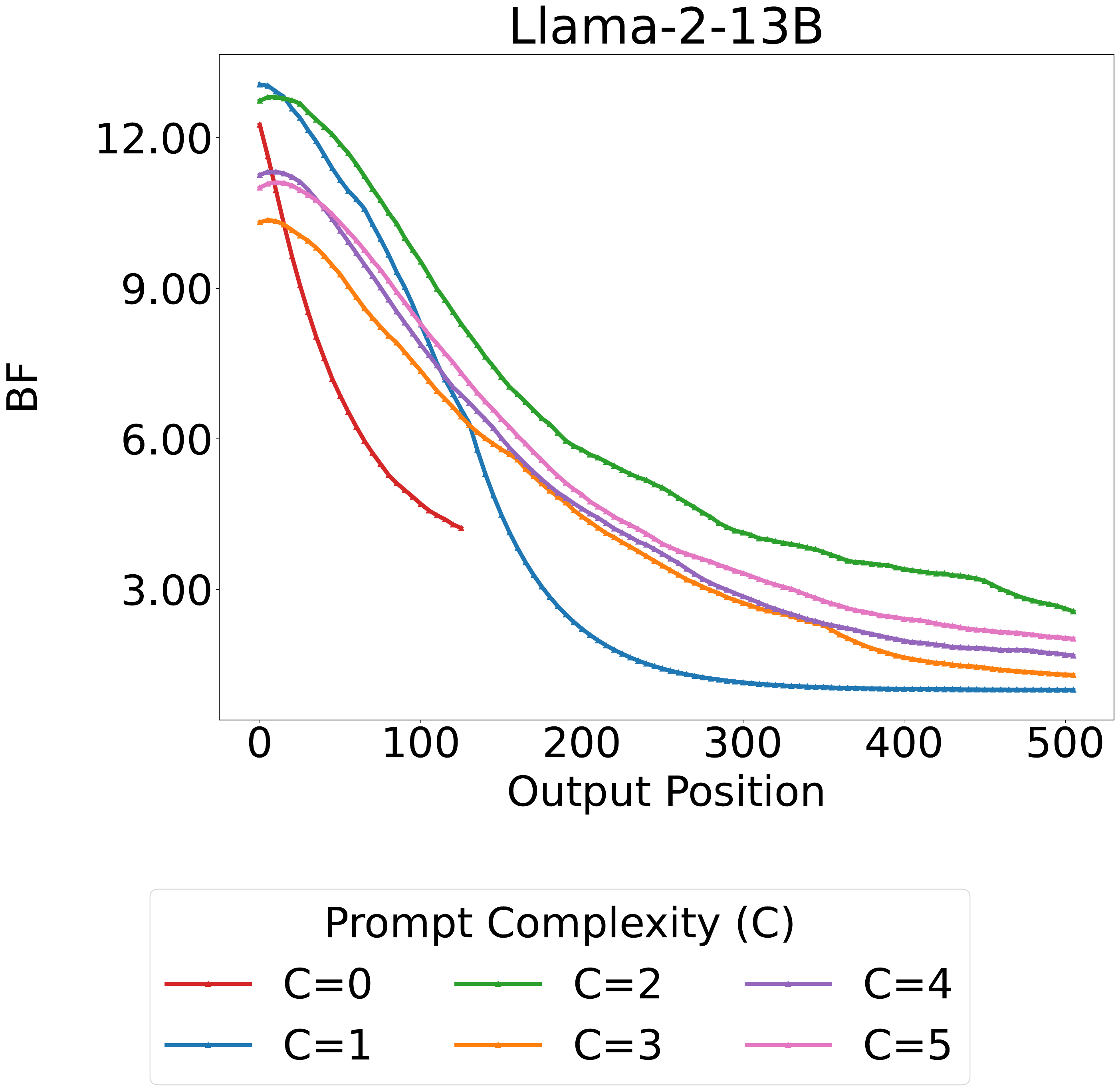}
     \label{fig:output_dynamic_base_cognac_random_str_llama2_13b_app}
    \end{subfigure}
    \begin{subfigure}[t]{0.24\textwidth}
    \centering
     \includegraphics[width=0.9\linewidth]{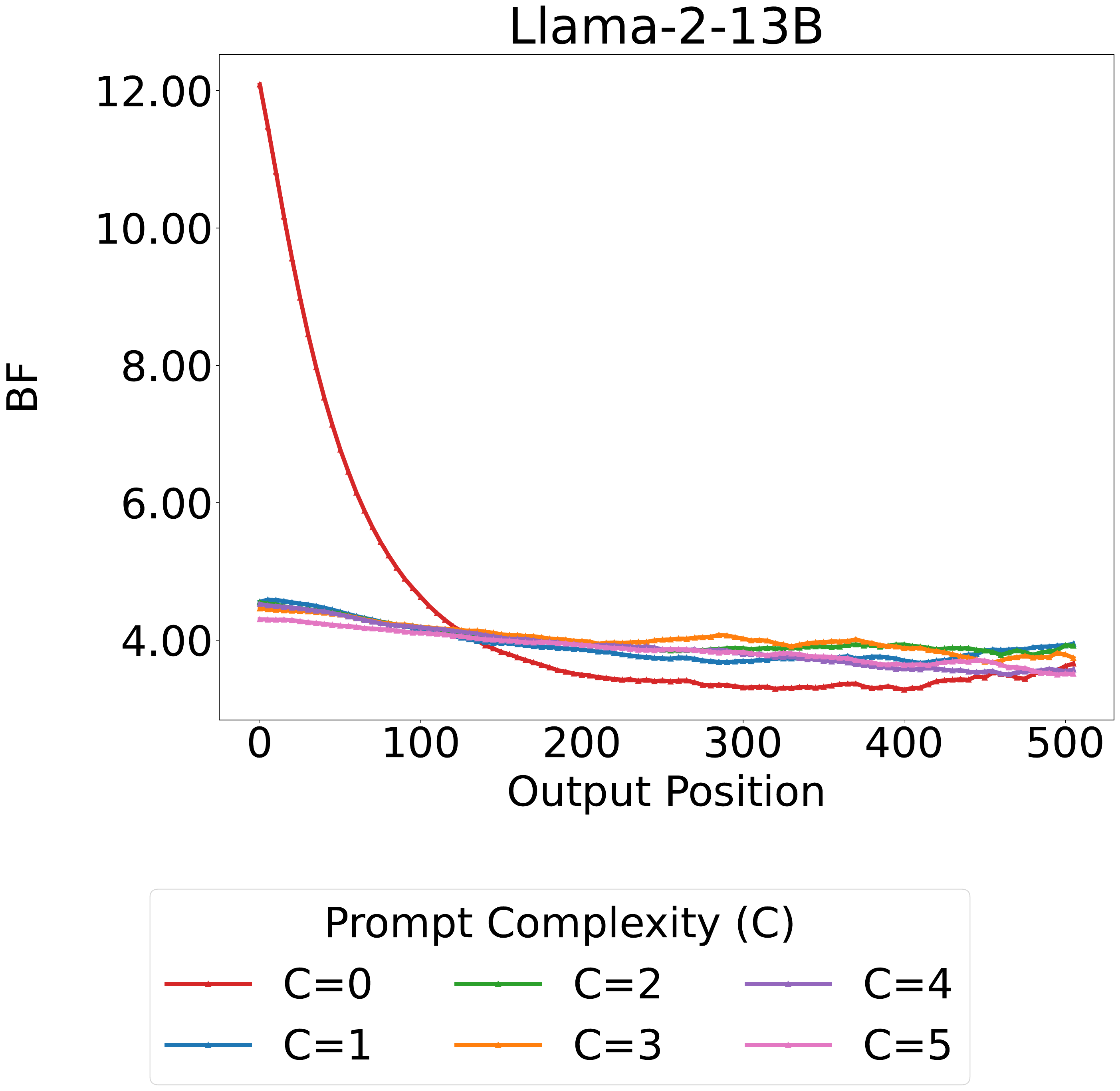}
     \label{fig:output_dynamic_base_bbcnews_llama2_13b_app}
    \end{subfigure}
        \begin{subfigure}[t]{0.24\textwidth}
    \centering
     \includegraphics[width=0.9\linewidth]{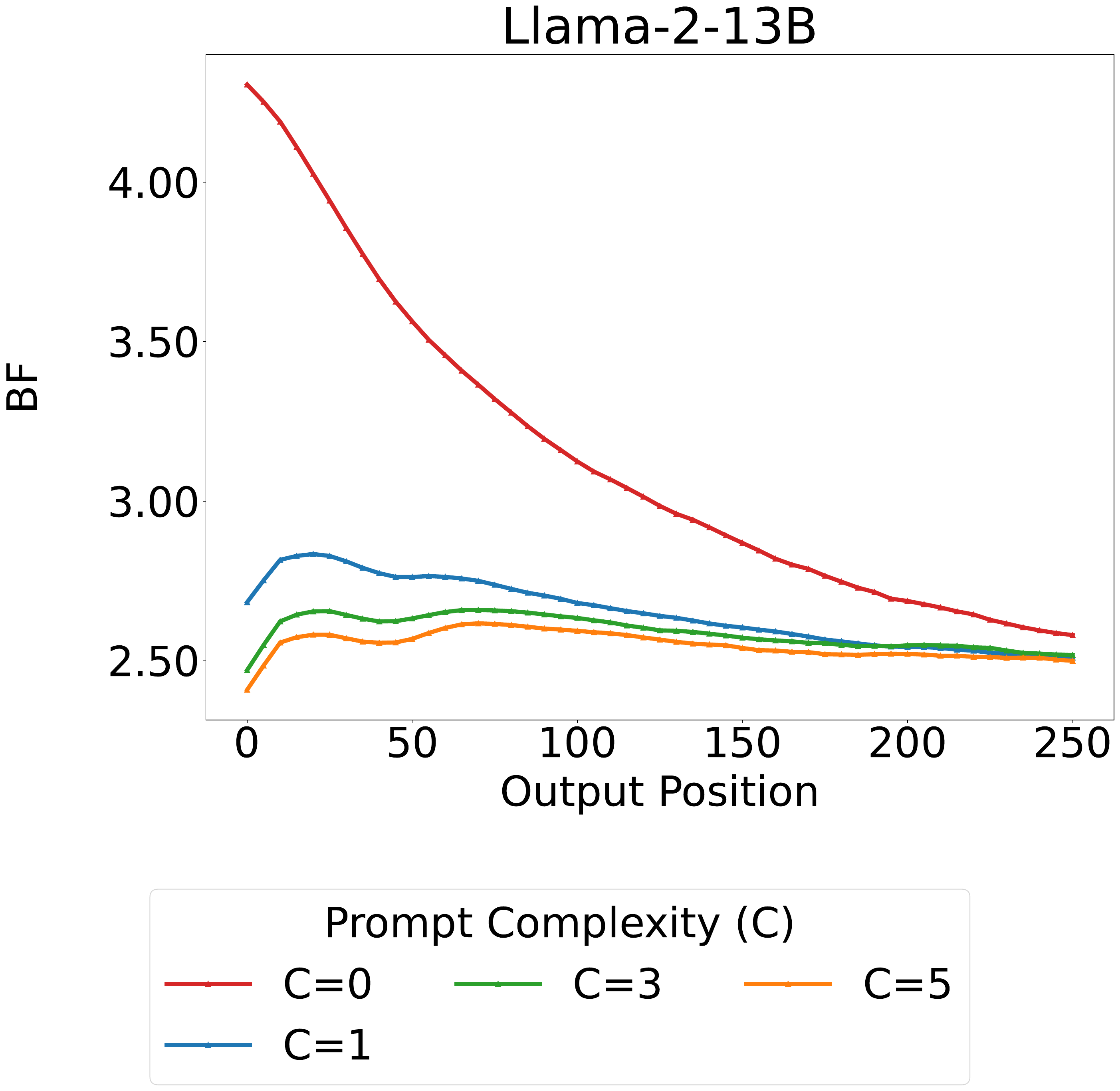}
     \label{fig:output_dynamic_base_mmlu_llama2_13b_app}
    \end{subfigure}
    \begin{subfigure}[t]{0.24\textwidth}
    \centering
     \includegraphics[width=0.9\linewidth]{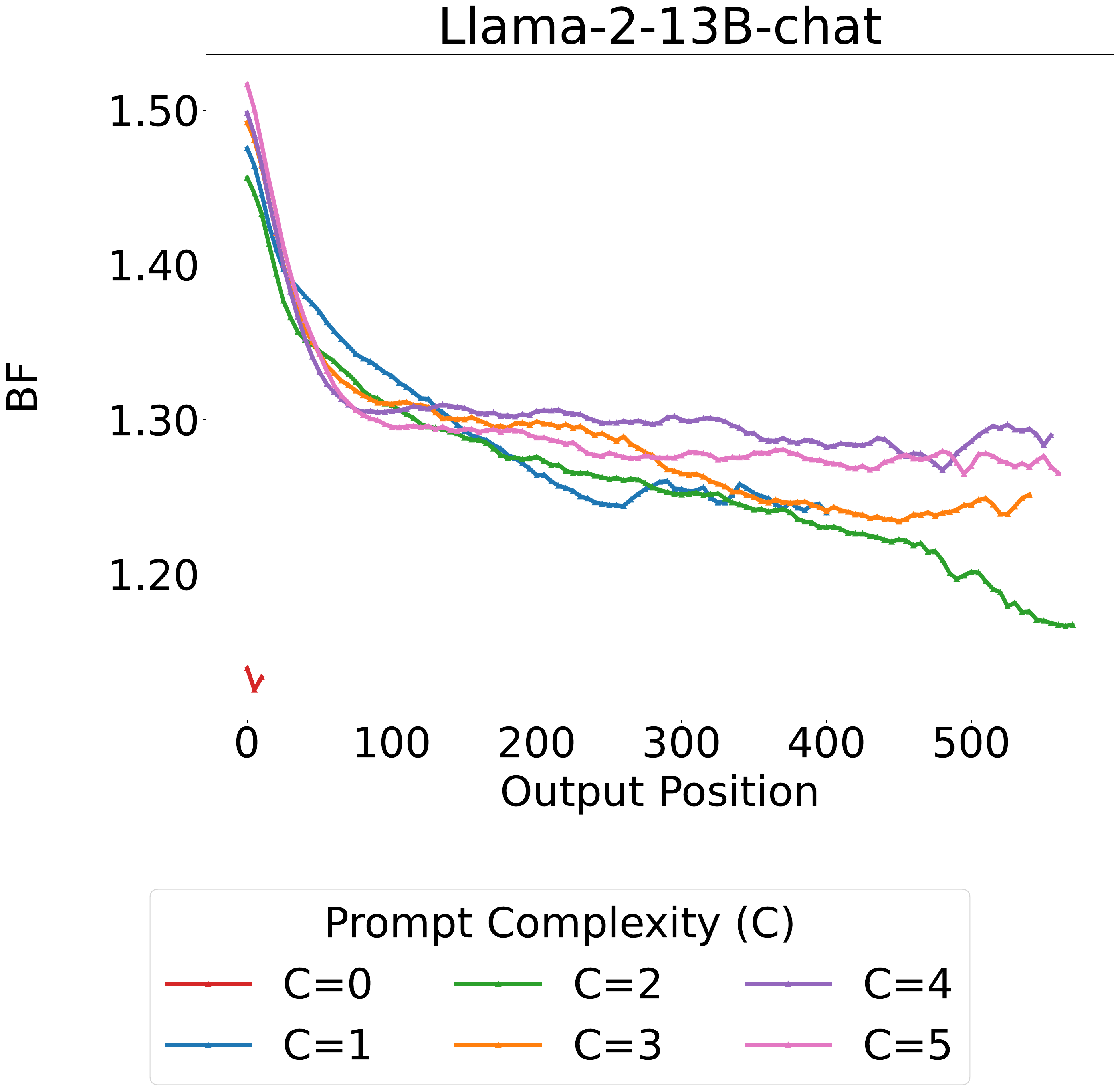}
     \label{fig:output_dynamic_base_storytelling_llama2_13b_chat_app}
    \end{subfigure}
        \begin{subfigure}[t]{0.24\textwidth}
    \centering
     \includegraphics[width=0.9\linewidth]{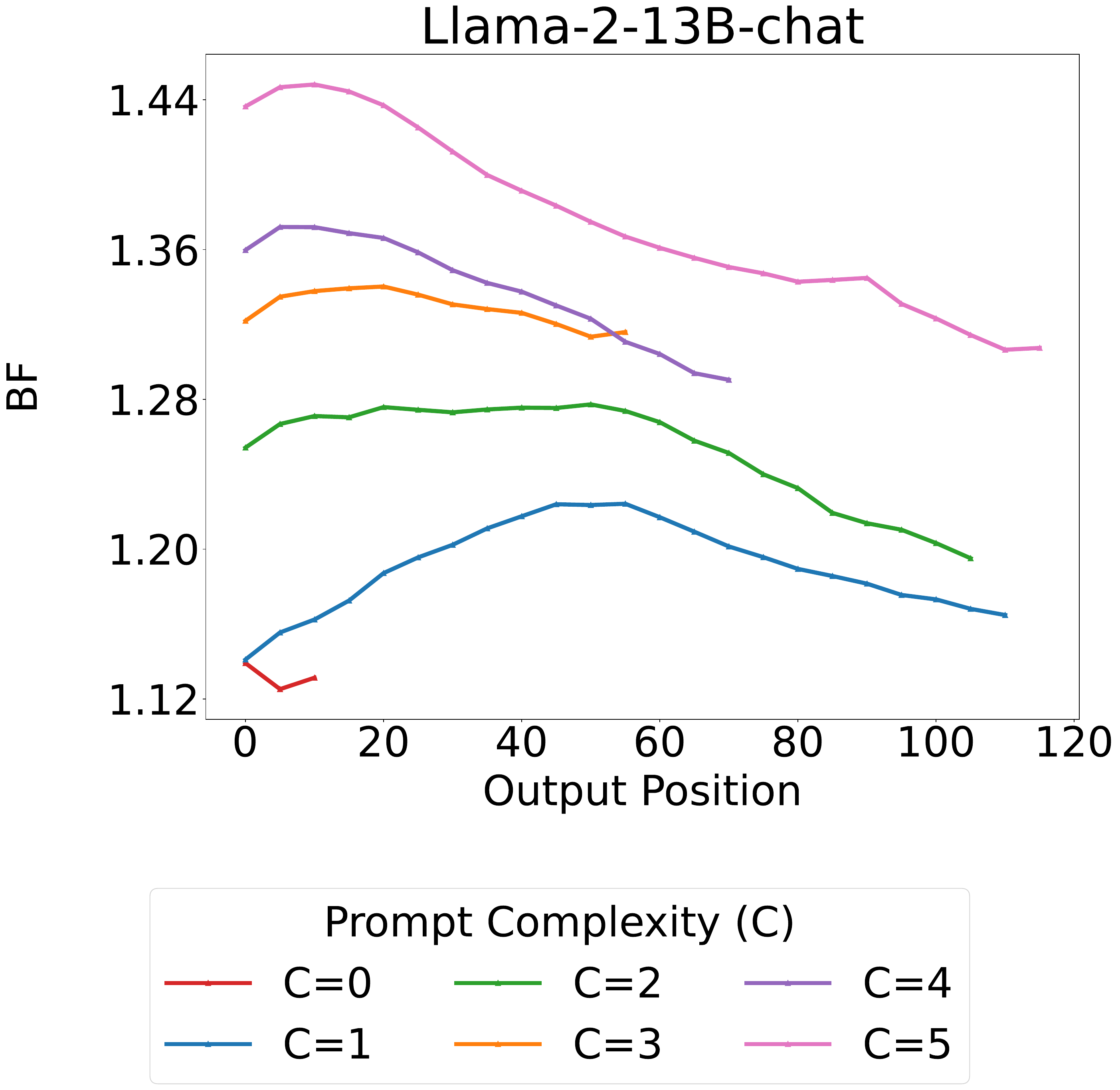}
     \label{fig:output_dynamic_base_cognac_random_str_llama2_13b_chat_app}
    \end{subfigure}
    \begin{subfigure}[t]{0.24\textwidth}
    \centering
     \includegraphics[width=0.9\linewidth]{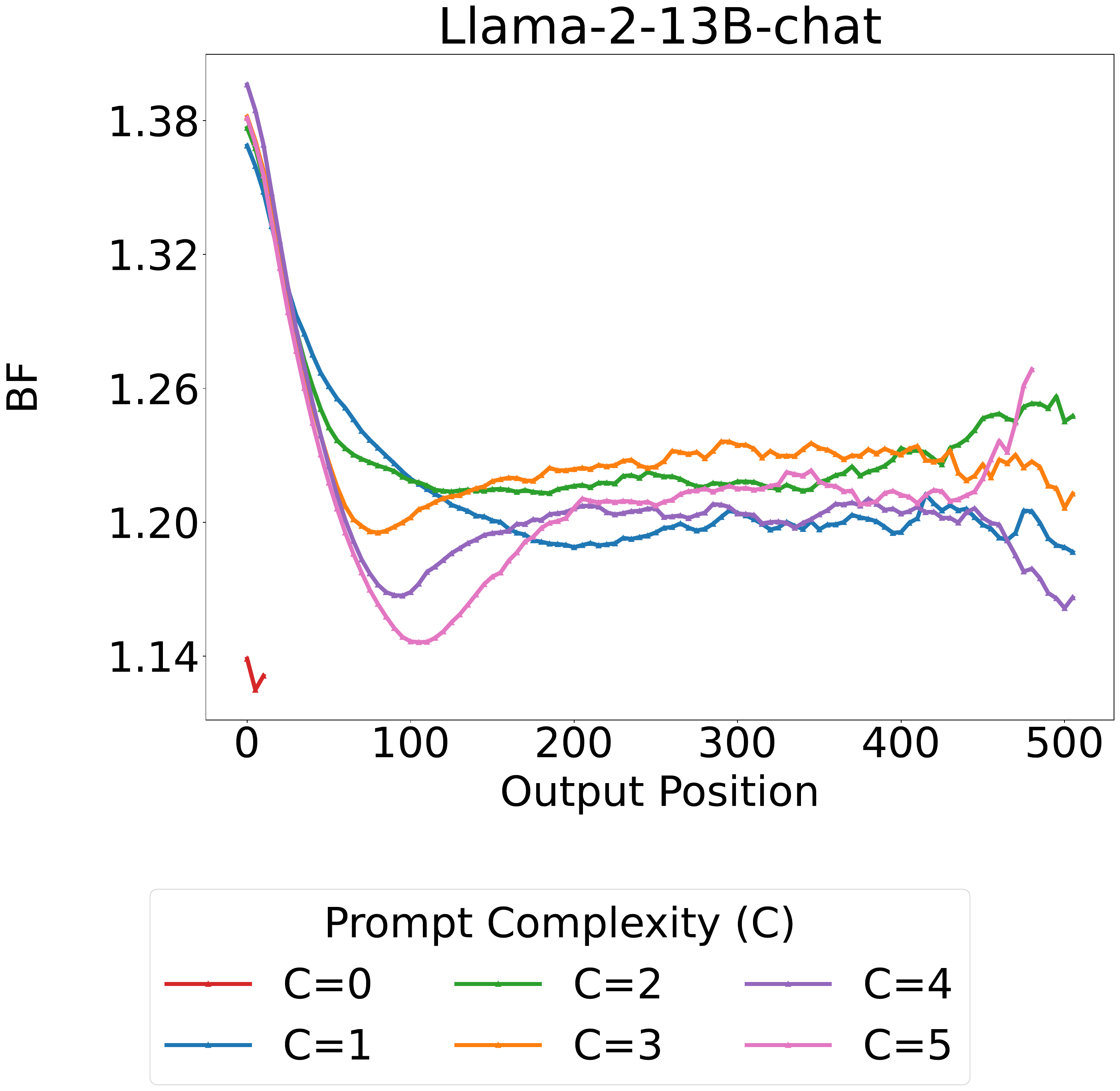}
     \label{fig:output_dynamic_base_bbcnews_llama2_13b_chat_app}
    \end{subfigure}
        \begin{subfigure}[t]{0.24\textwidth}
    \centering
     \includegraphics[width=0.9\linewidth]{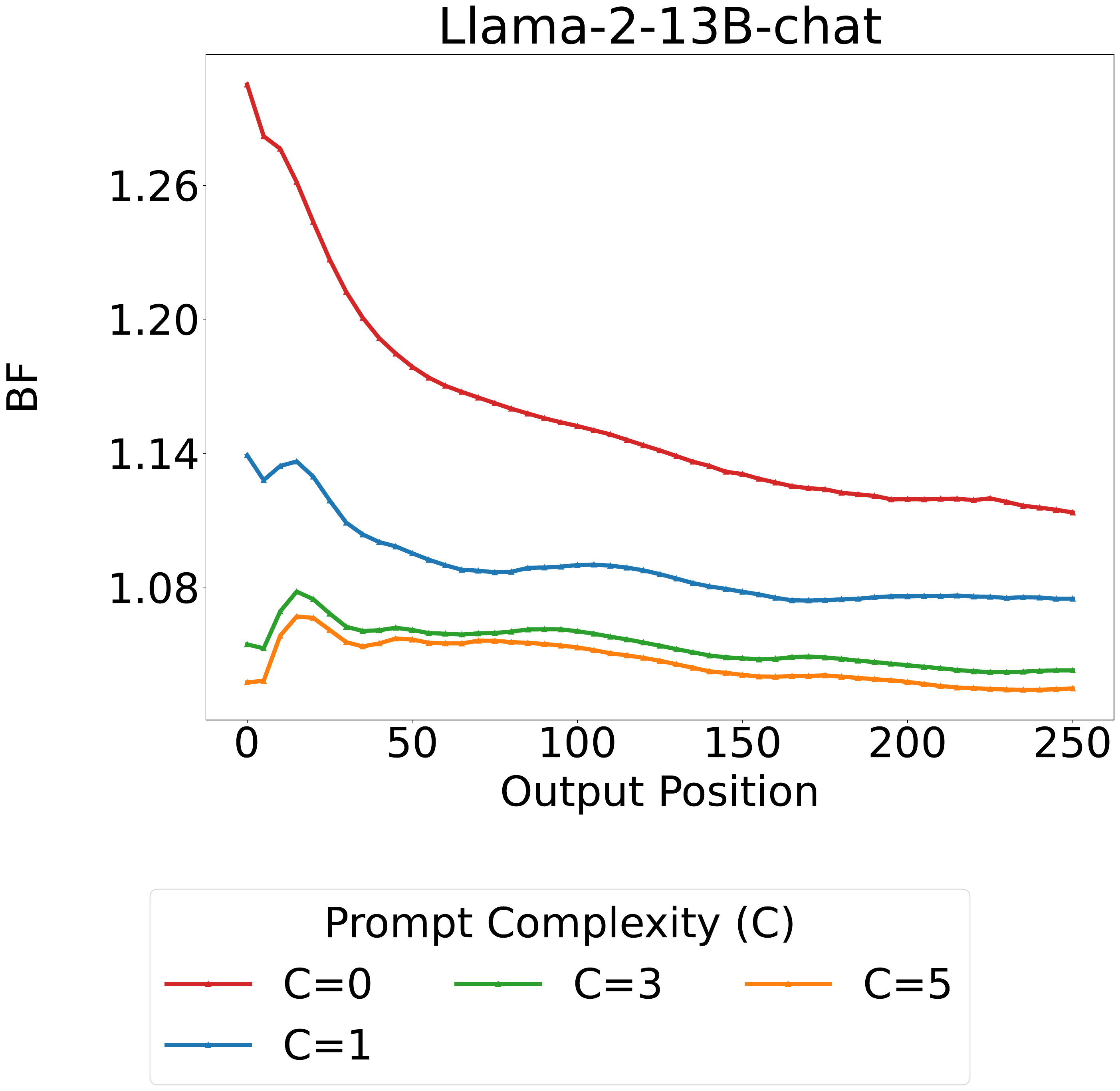}
     \label{fig:output_dynamic_base_mmlu_llama2_13b_chat_app}
    \end{subfigure}
    \begin{subfigure}[t]{0.24\textwidth}
    \centering
     \includegraphics[width=0.9\linewidth]{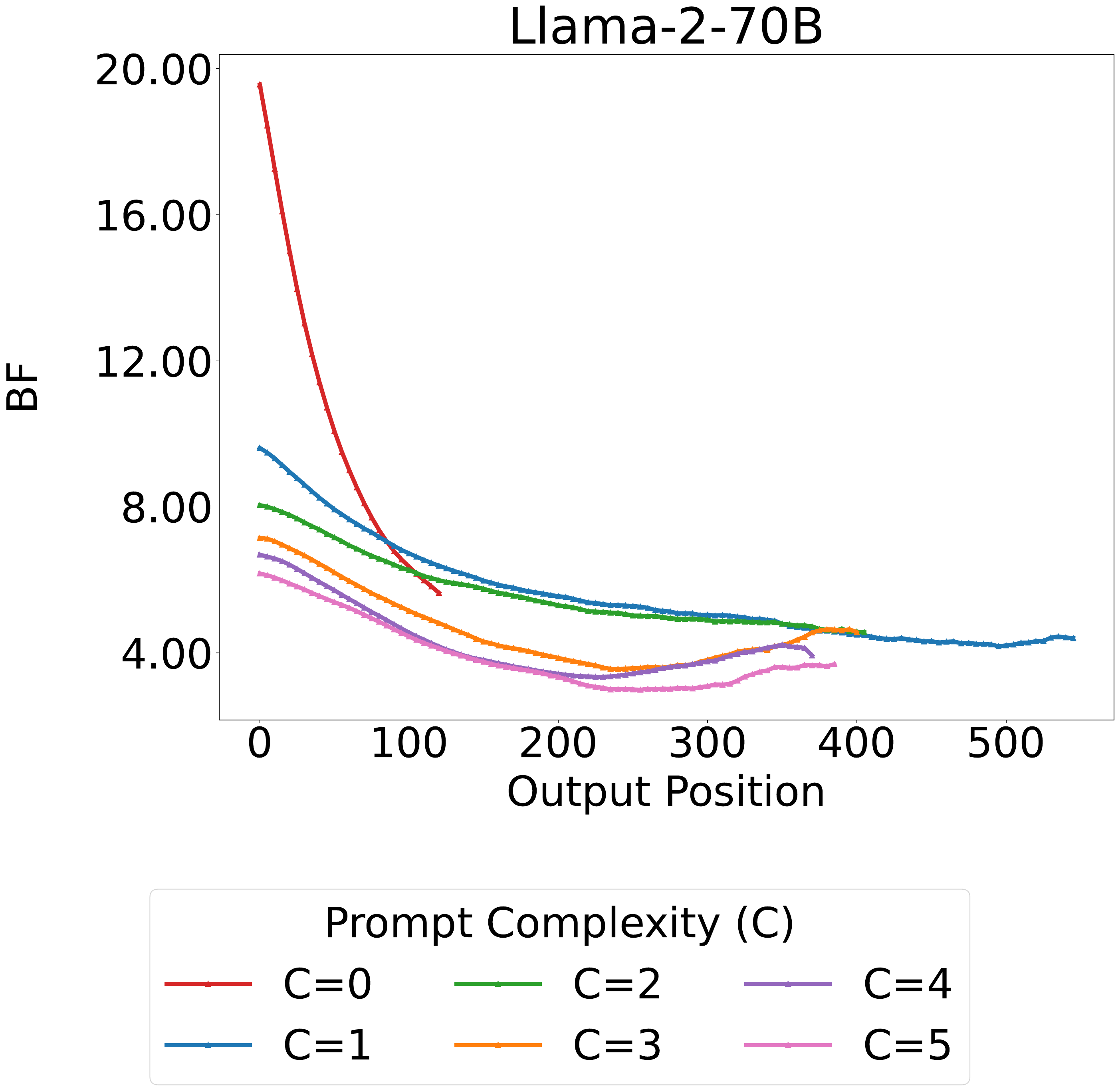}
     \label{fig:output_dynamic_base_storytelling_llama2_70b_app}
    \end{subfigure}
        \begin{subfigure}[t]{0.24\textwidth}
    \centering
     \includegraphics[width=0.9\linewidth]{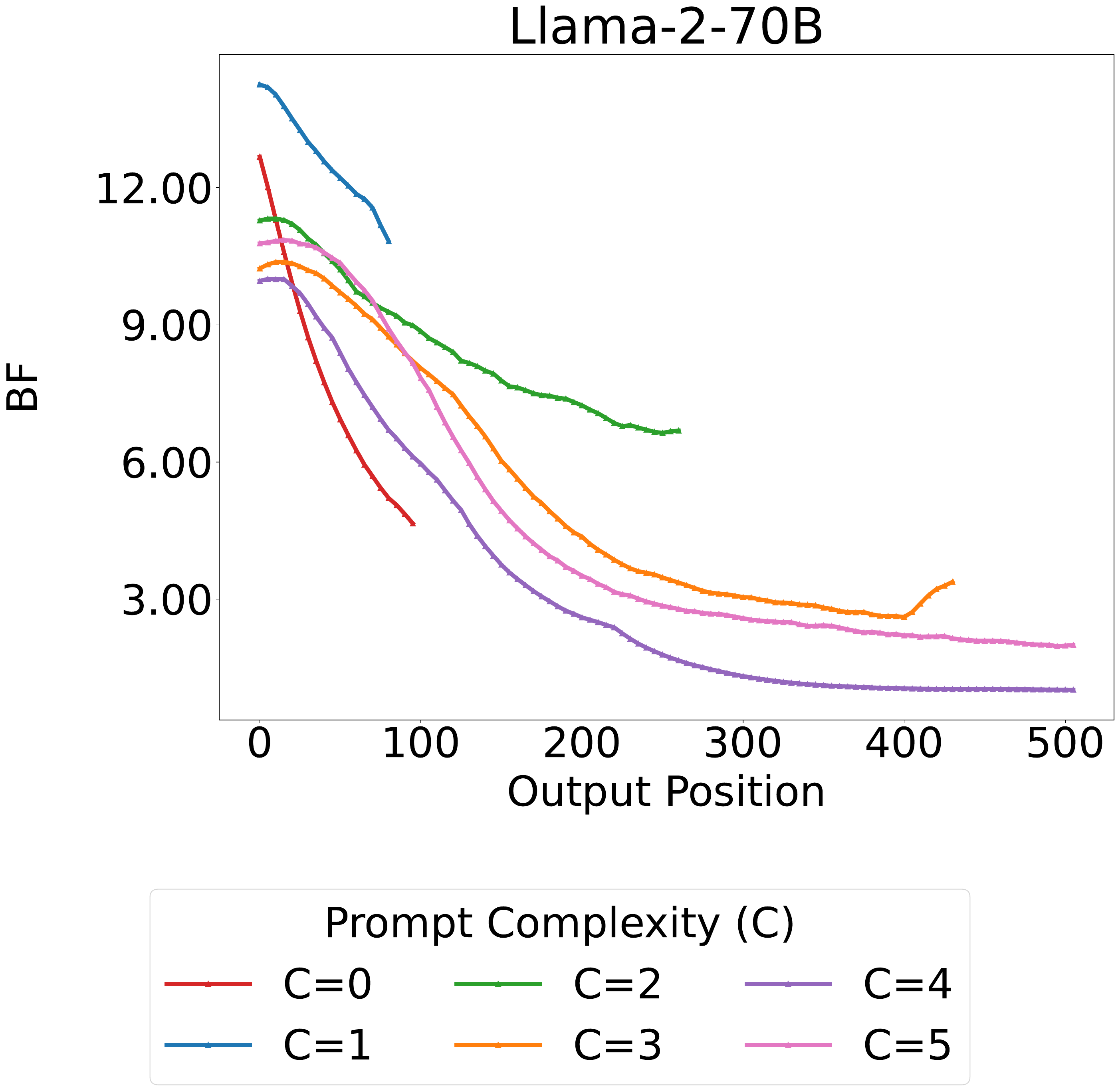}
     \label{fig:output_dynamic_base_cognac_random_str_llama2_70b_app}
    \end{subfigure}
    \begin{subfigure}[t]{0.24\textwidth}
    \centering
     \includegraphics[width=0.9\linewidth]{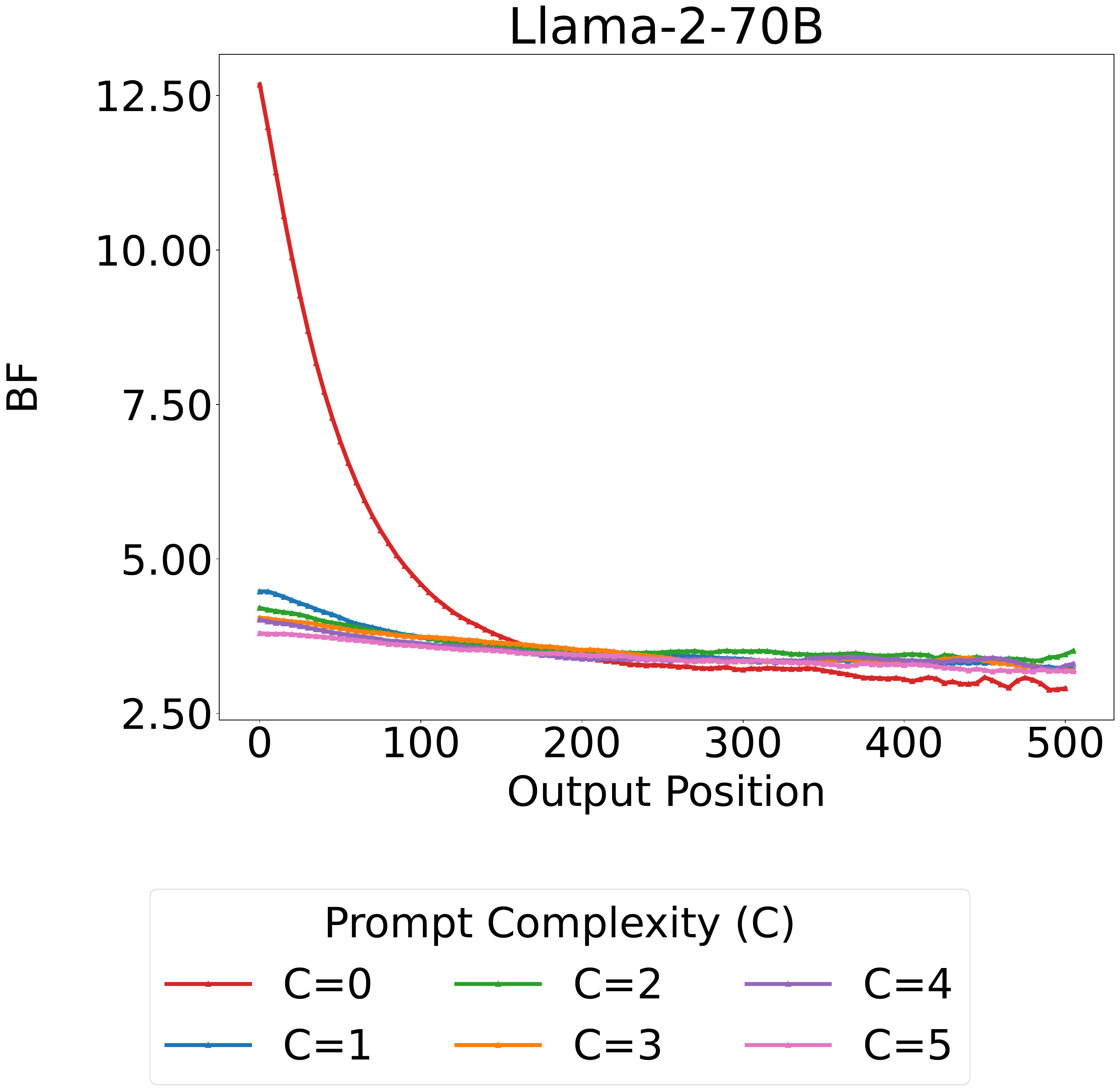}
     \label{fig:output_dynamic_base_bbcnews_llama2_70b_app}
    \end{subfigure}
        \begin{subfigure}[t]{0.24\textwidth}
    \centering
     \includegraphics[width=0.9\linewidth]{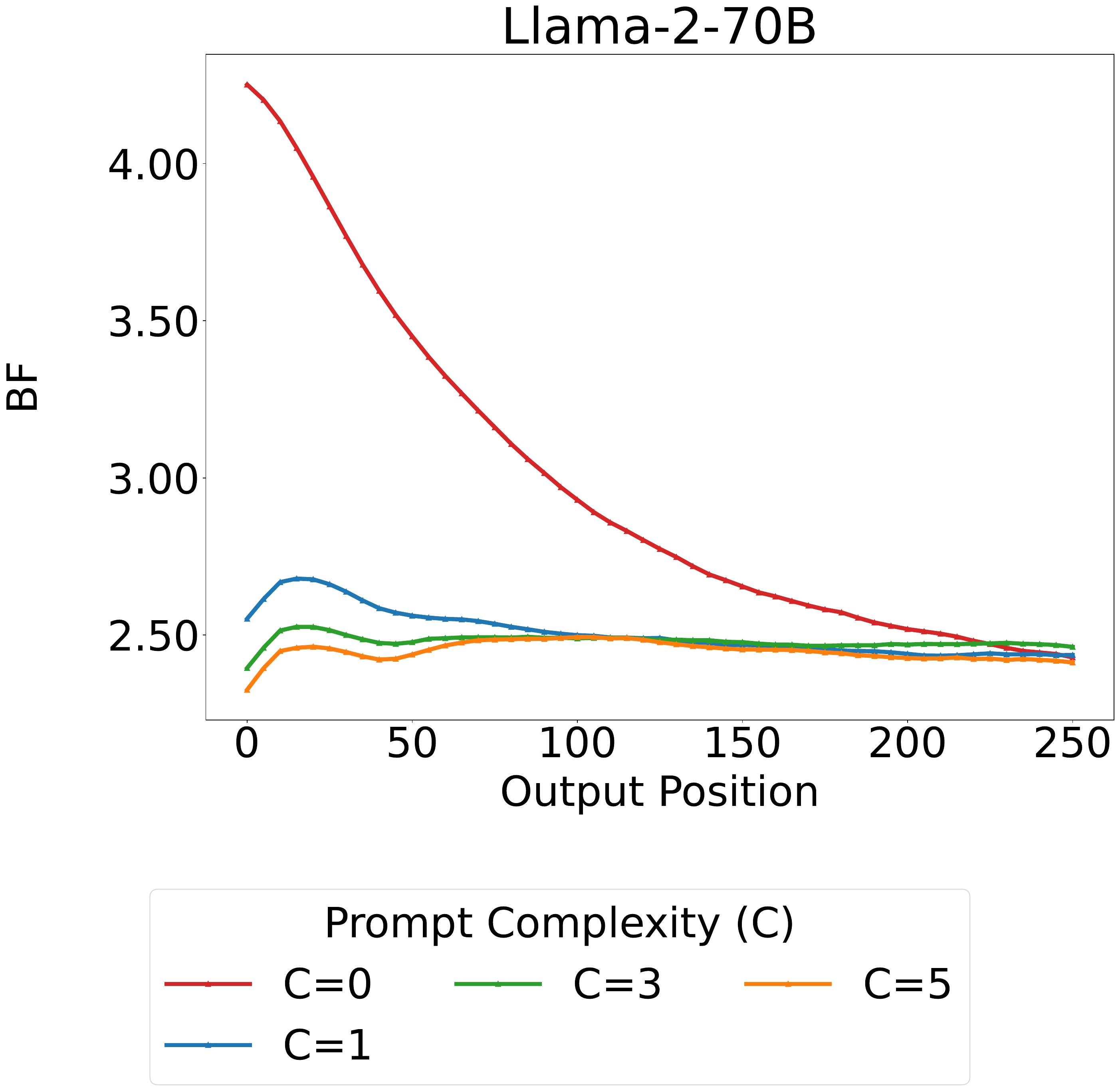}
     \label{fig:output_dynamic_base_mmlu_llama2_70b_app}
    \end{subfigure}
    \begin{subfigure}[t]{0.24\textwidth}
    \centering
     \includegraphics[width=0.9\linewidth]{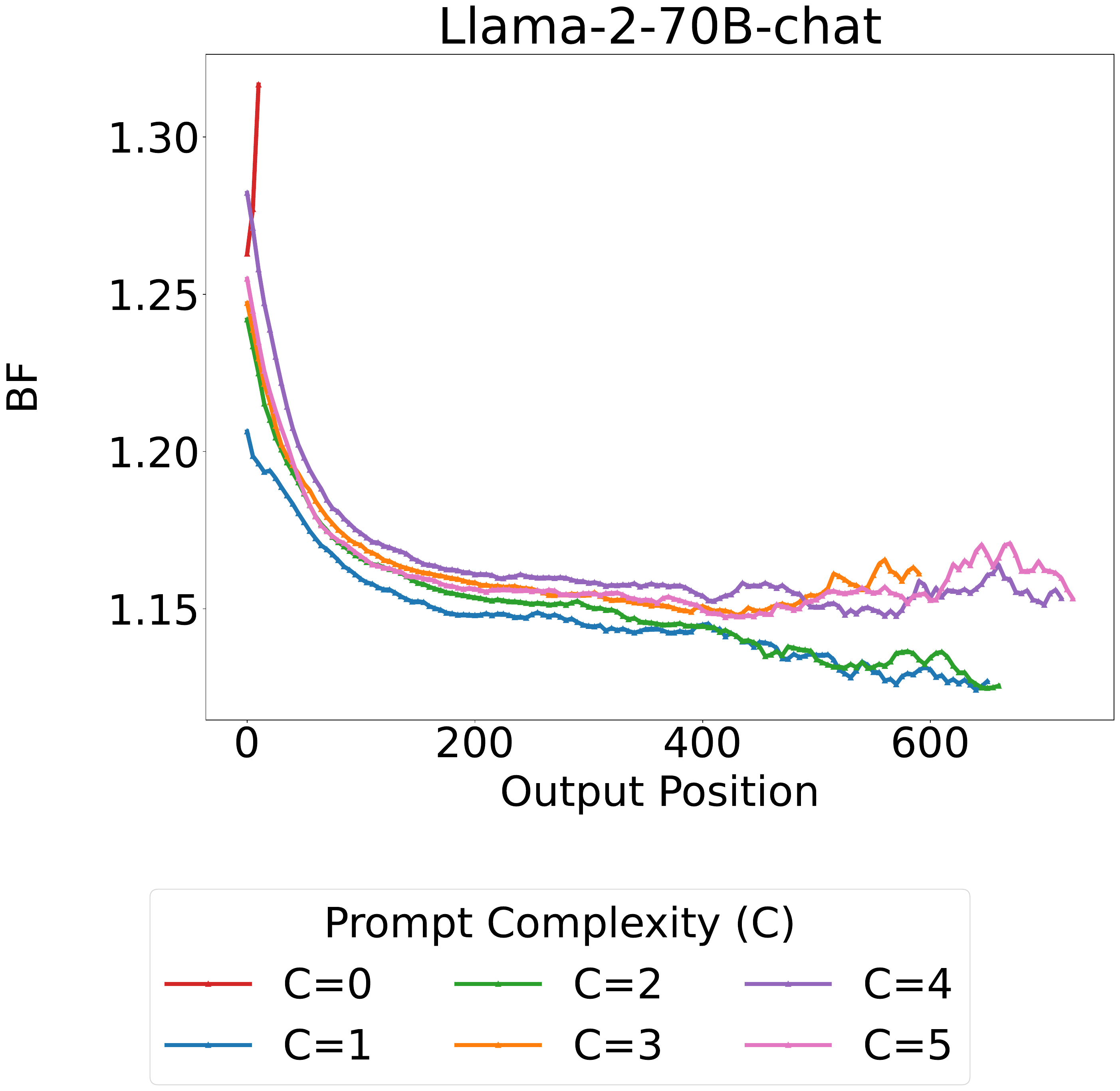}
     \label{fig:output_dynamic_base_storytelling_llama2_70b_chat_app}
    \end{subfigure}
        \begin{subfigure}[t]{0.24\textwidth}
    \centering
     \includegraphics[width=0.9\linewidth]{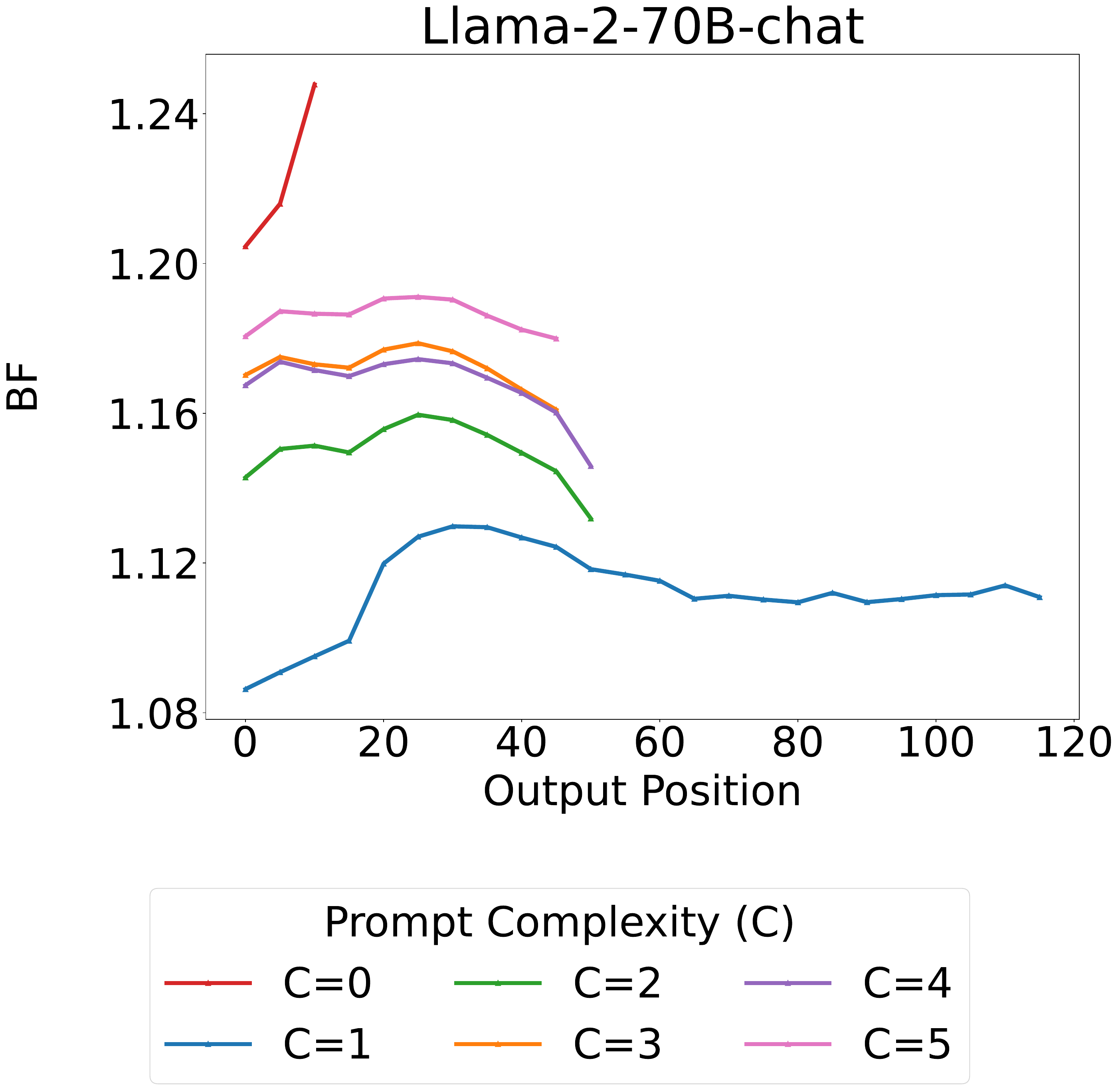}
     \label{fig:output_dynamic_base_cognac_random_str_llama2_70b_chat_app}
    \end{subfigure}
    \begin{subfigure}[t]{0.24\textwidth}
    \centering
     \includegraphics[width=0.9\linewidth]{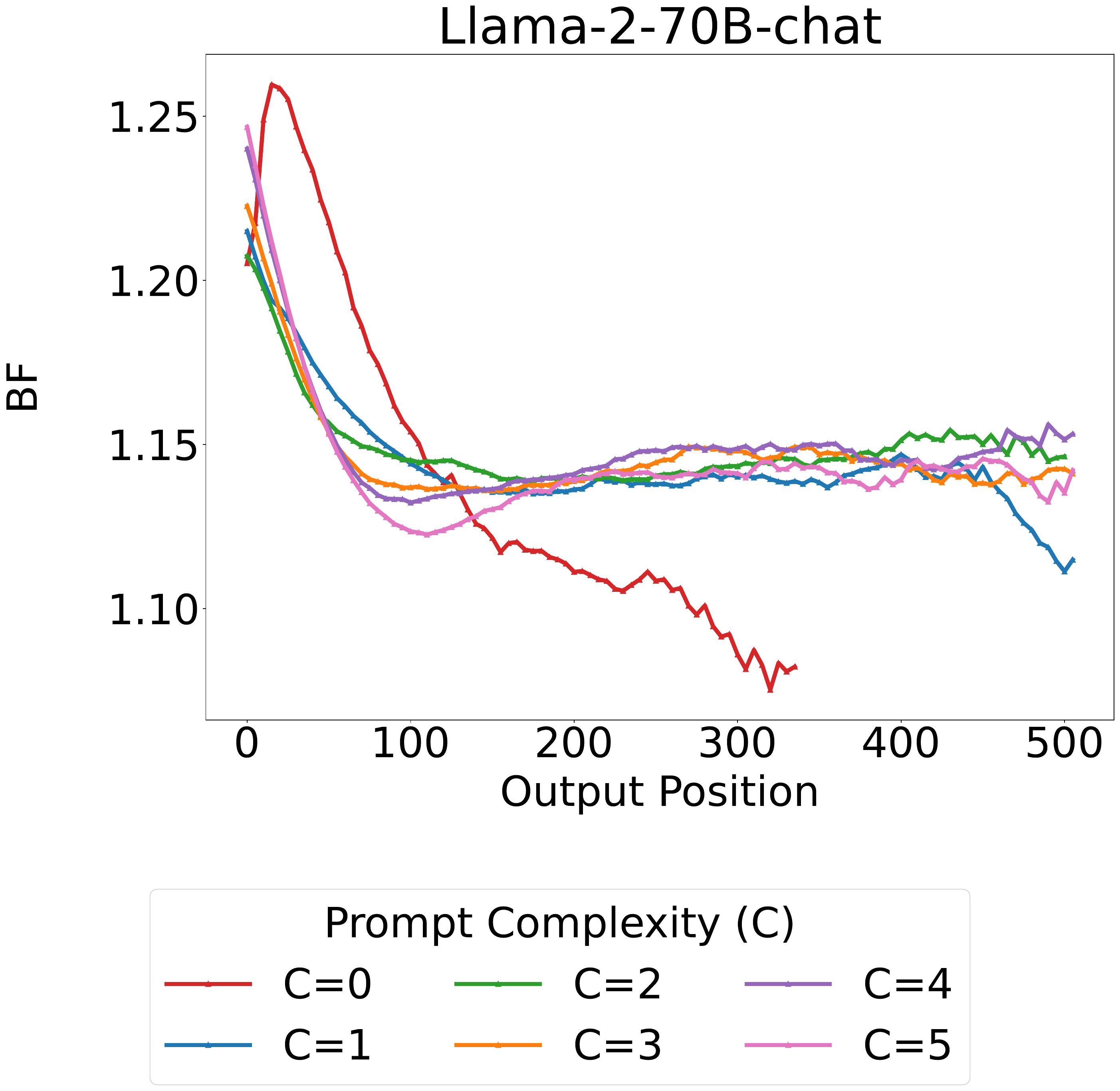}
     \label{fig:output_dynamic_base_bbcnews_llama2_70b_chat_app}
    \end{subfigure}
        \begin{subfigure}[t]{0.24\textwidth}
    \centering
     \includegraphics[width=0.9\linewidth]{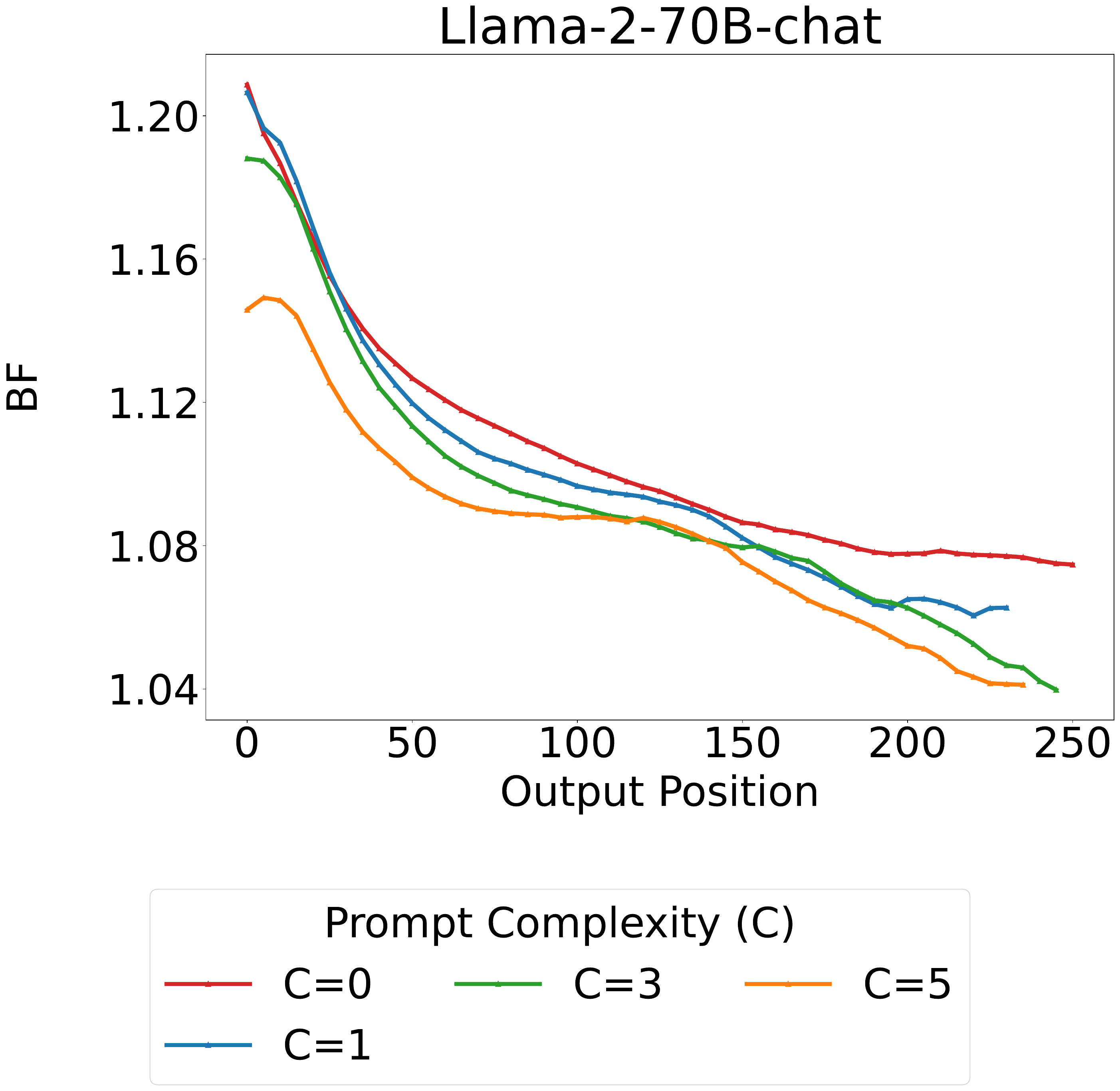}
     \label{fig:output_dynamic_base_mmlu_llama2_70b_chat_app}
    \end{subfigure}
    \caption{\textbf{BF Output Dynamic for Llama-2-families.} For better visualization, we compute the exponential moving averaged values of perplexity with the smoothing factor set as $0.1$.
    }
    \label{fig: output_dynamic_app_llama2}
\end{figure*}

\section{Dataset-Specific Processing}
\label{app: dataset_details}
For all datasets we used in the paper, we carefully controlled whether the prompt length and expected output length would exceed the model's maximum length. 
\paragraph{MMLU}\citep{hendrycks2021measuring} is a widely-used multiple-choice reasoning question. Unless otherwise explained, we use the full test set of MMLU to avoid potential contamination, following benchmarking settings reported in most LLM technical reports\citep{touvron2023llama, dubey2024llama, guo2025deepseek}. We formulate prompt complexity $C$ as the number of in-context samples. For example, $C=1$ means we only add one in-context sample. For prompting setup and postprocessing details, we follow the standard implementation in Qwen-2.5-Math~\citep{yang2024qwen2}.
\paragraph{Cognac}\citep{chen2022cognac} is a controlled generation task requiring  language model \emph{not} to generate specified banned words provided in the prompt. 
We use the WordNet subset~\citep{miller1995wordnet} of Cognac as this is the only released setting in Cognac paper, where the topic is a root node and the constraint is defined as a subtree. We sampled $200$ instances using the provided data generation codes in our experiments.  
To ensure most model generations ended properly in the decoding process, we relax the constraint of maximum decoded tokens $T$ from $60$ to $512$. We use the same prompt templates following their Github repo.\footnote{\url{https://github.com/princeton-nlp/Cognac/tree/main}}
\paragraph{Creative Story Generation}\citep{chakrabarty2024art} provides the plots and story continuation from both machine and human. 
We adopt the provided $11$ human-written story plots in the original dataset as the prompt. In this task, we set the maximum token $T=1024$ to ensure the continued story written by LLM can have a proper ending. We formulate prompt complexity $C$ as providing $C \times 25$ words in the plot. 
\paragraph{Random Strings}
Similar to ~\citet{bigelowsubjective}, we sample $200$ random strings with length $L\sim U(256, 512)$ from the tokenizer vocabulary as the prompt. Prompt complexity $C$ is formulated by providing $C \times 15$ tokens in the prompt, ensuring each article contains at least 100 tokens. 
\paragraph{BBCLatestNews}~\citep{li2024latesteval} is a news collection dataset aims at collecting news that is beyond the time cut for training LLMs. Unlike creative story plots, news articles are typically more structured and organized, although headlines can still be surprising. We select news articles from January to July 2024 to minimize data contamination, as the Llama models have a knowledge cut-off in late 2023. We formulate prompt complexity $C$ as providing $C \times 15$ words in the prompt. 

\begin{figure*}[t!]
\centering
    \begin{subfigure}[t]{0.24\textwidth}
    \centering
     \includegraphics[width=0.9\linewidth]{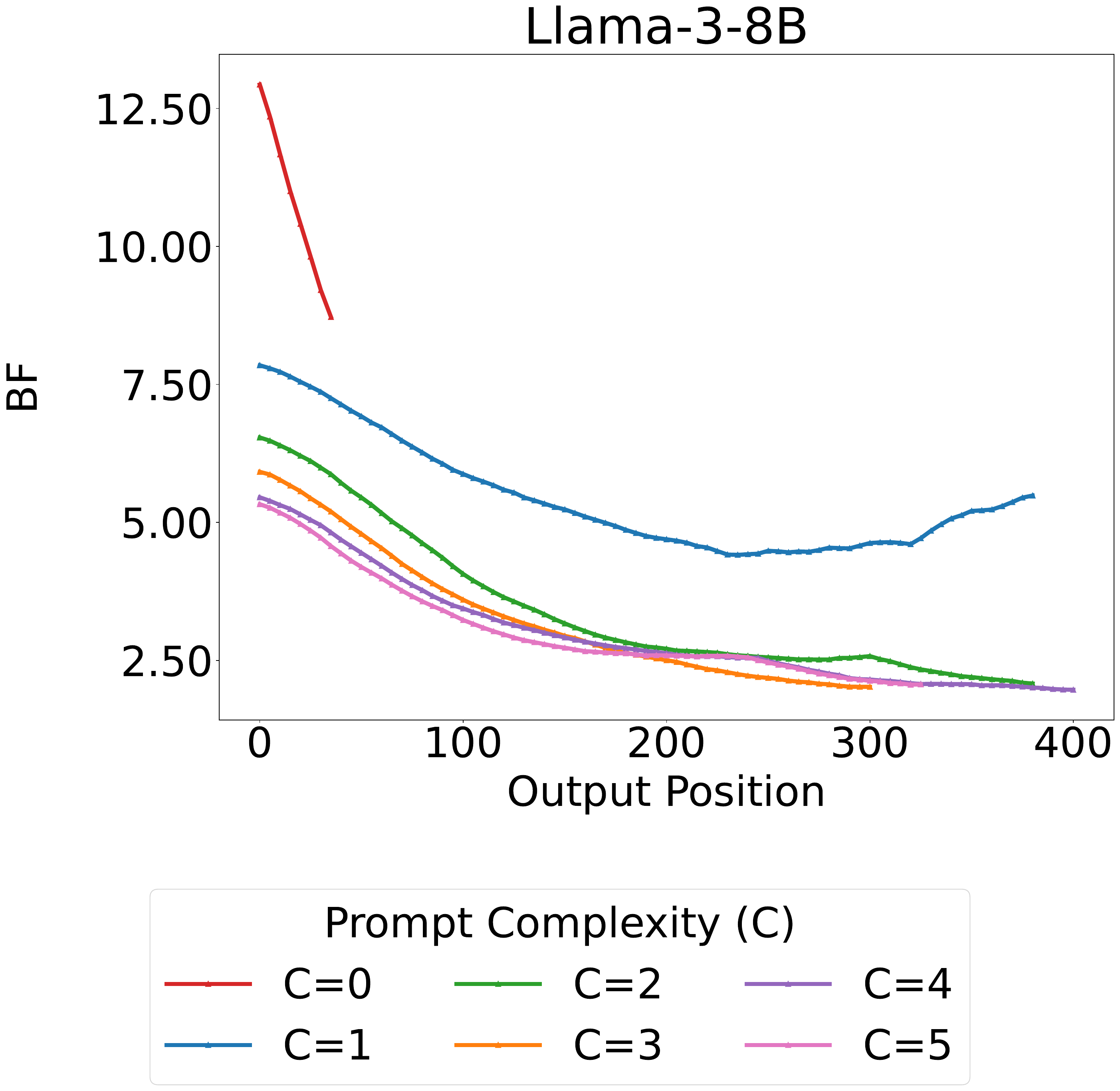}
     \label{fig:output_dynamic_base_storytelling_8b_app}
    \end{subfigure}
        \begin{subfigure}[t]{0.24\textwidth}
    \centering
     \includegraphics[width=0.9\linewidth]{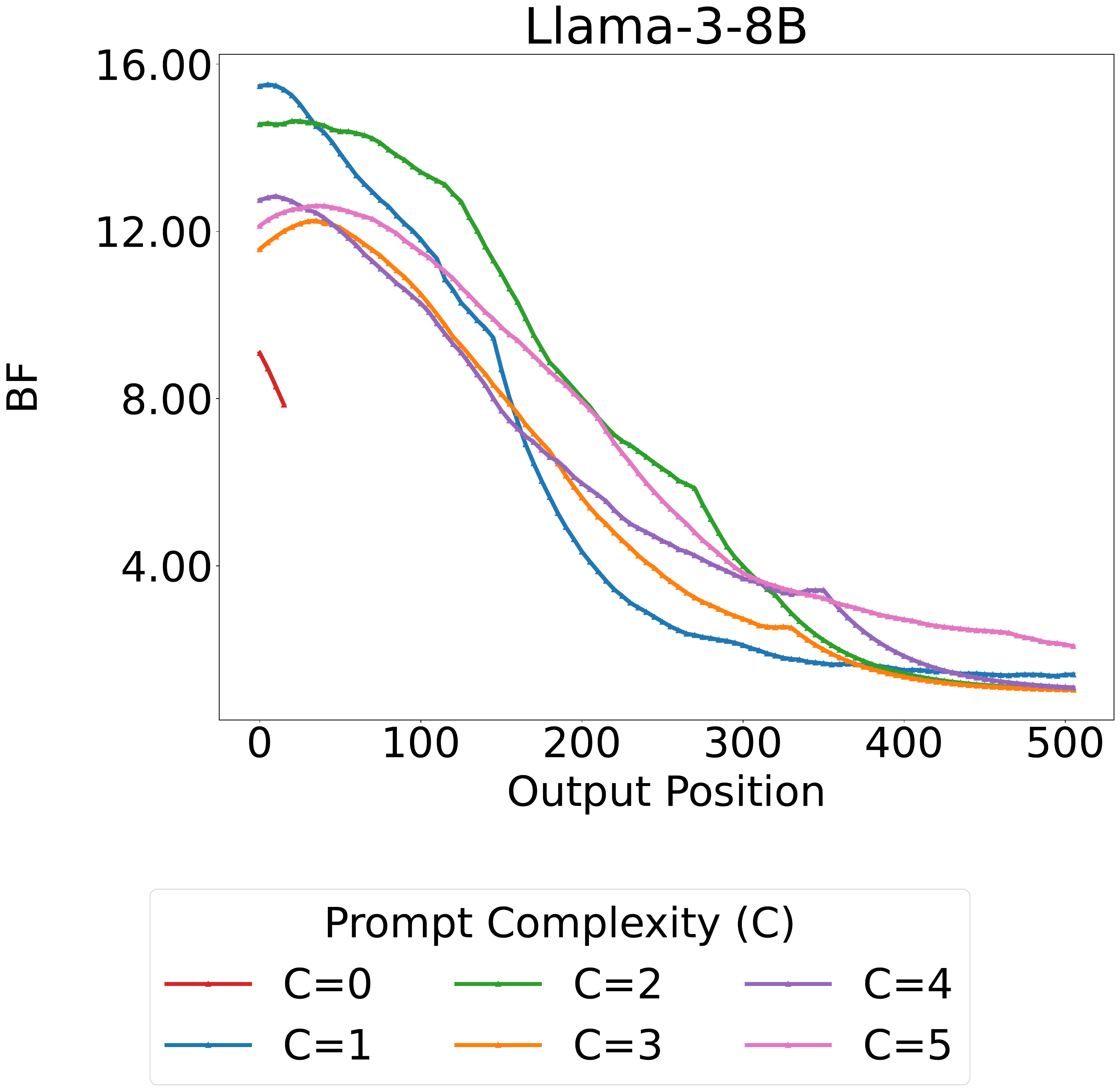}
     \label{fig:output_dynamic_base_cognac_random_str_8b_app}
    \end{subfigure}
    \begin{subfigure}[t]{0.24\textwidth}
    \centering
     \includegraphics[width=0.9\linewidth]{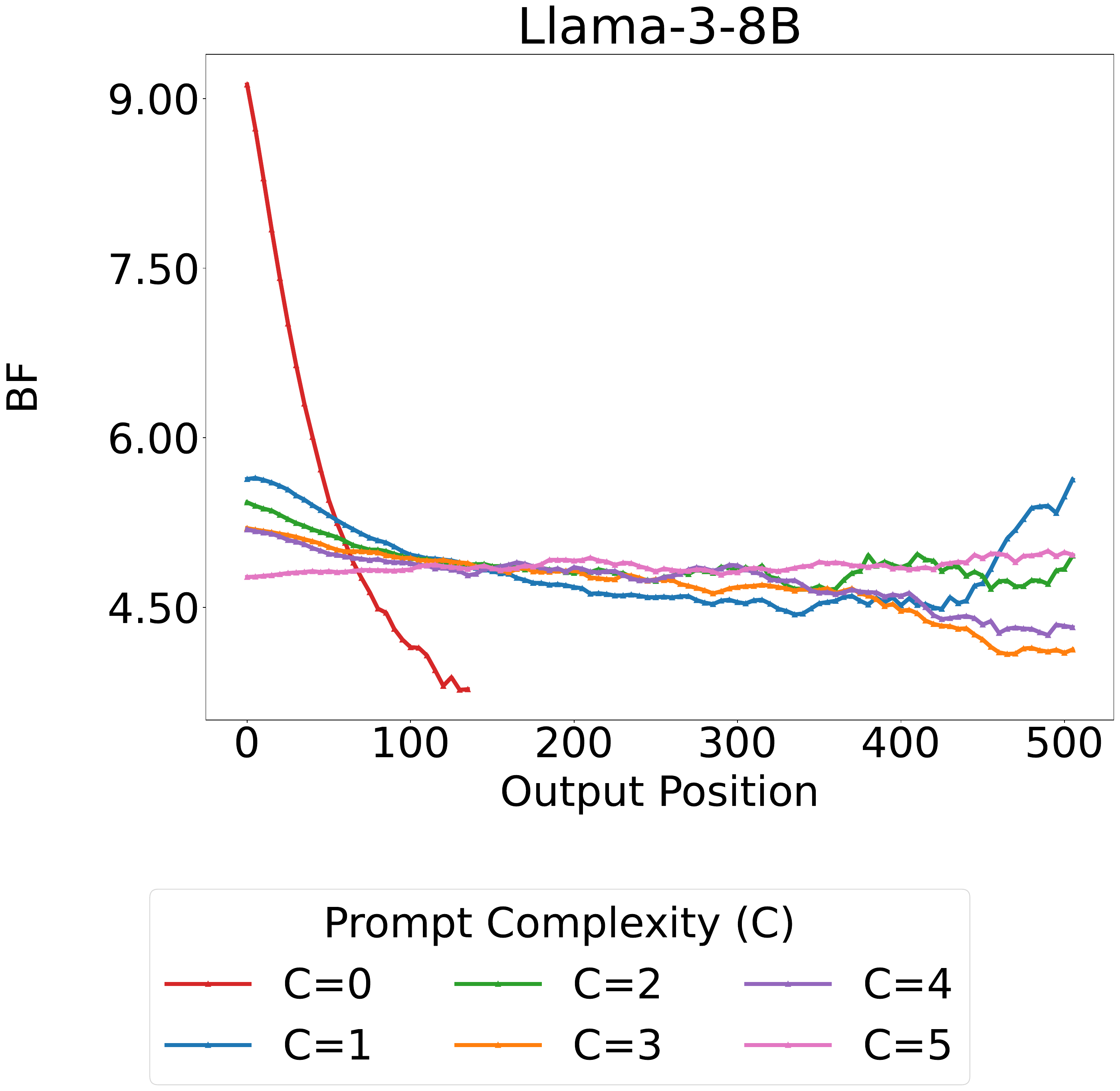}
     \label{fig:output_dynamic_base_bbcnews_8b_app}
    \end{subfigure}
        \begin{subfigure}[t]{0.24\textwidth}
    \centering
     \includegraphics[width=0.9\linewidth]{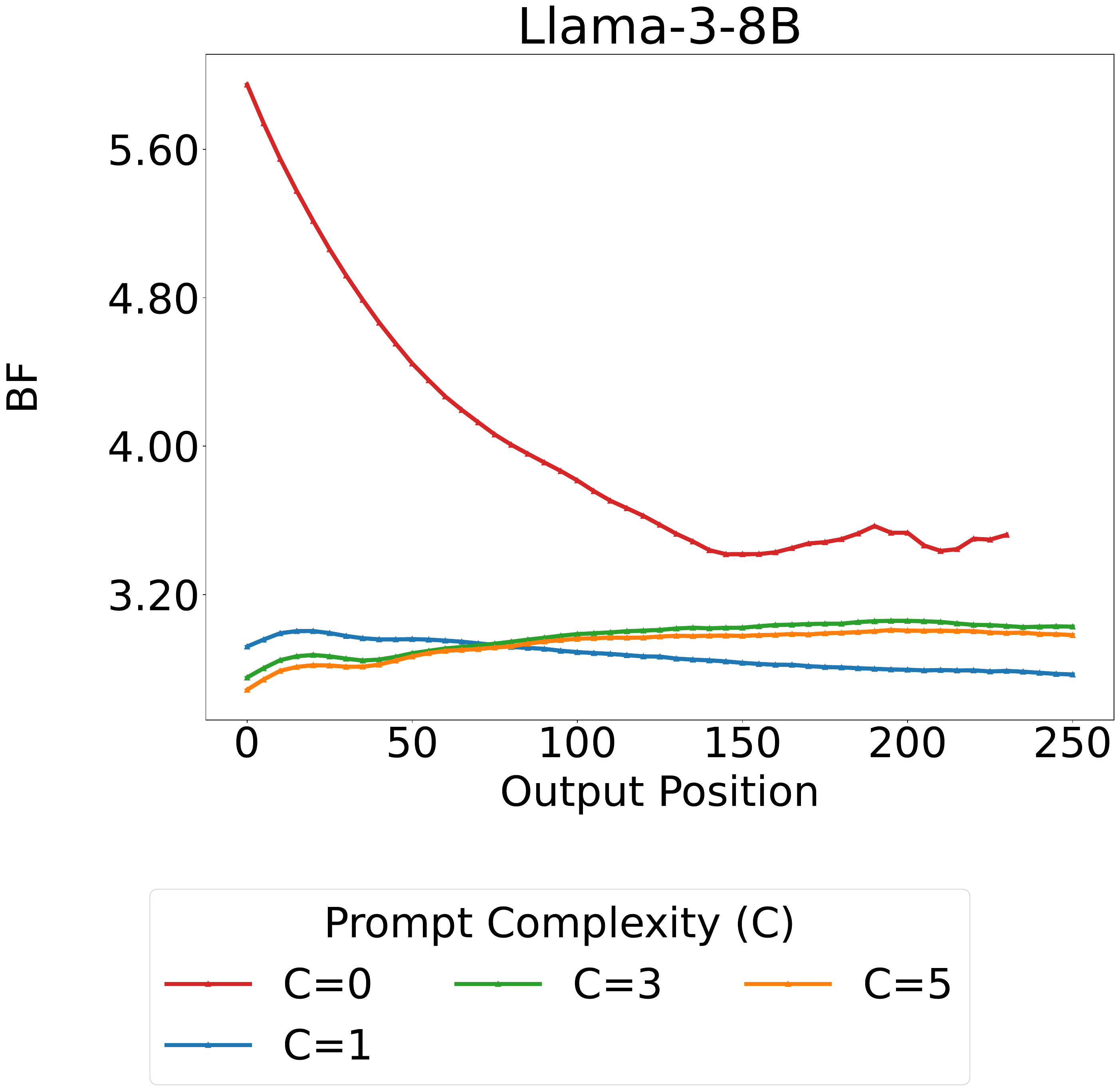}
     \label{fig:output_dynamic_base_mmlu_8b_app}
    \end{subfigure}

    \begin{subfigure}[t]{0.24\textwidth}
    \centering
     \includegraphics[width=0.9\linewidth]{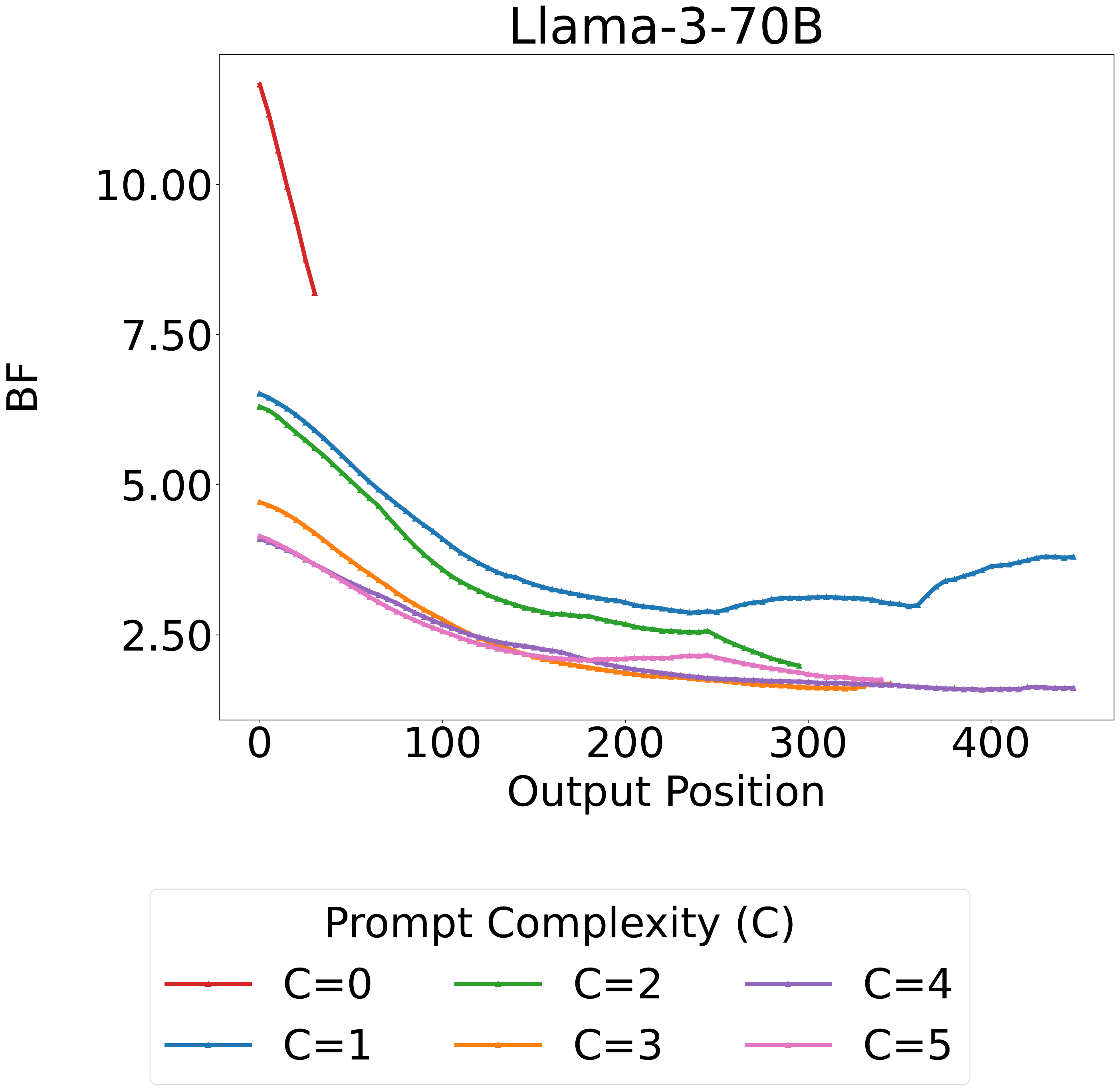}
     \label{fig:output_dynamic_base_storytelling_app}
    \end{subfigure}
        \begin{subfigure}[t]{0.24\textwidth}
    \centering
     \includegraphics[width=0.9\linewidth]{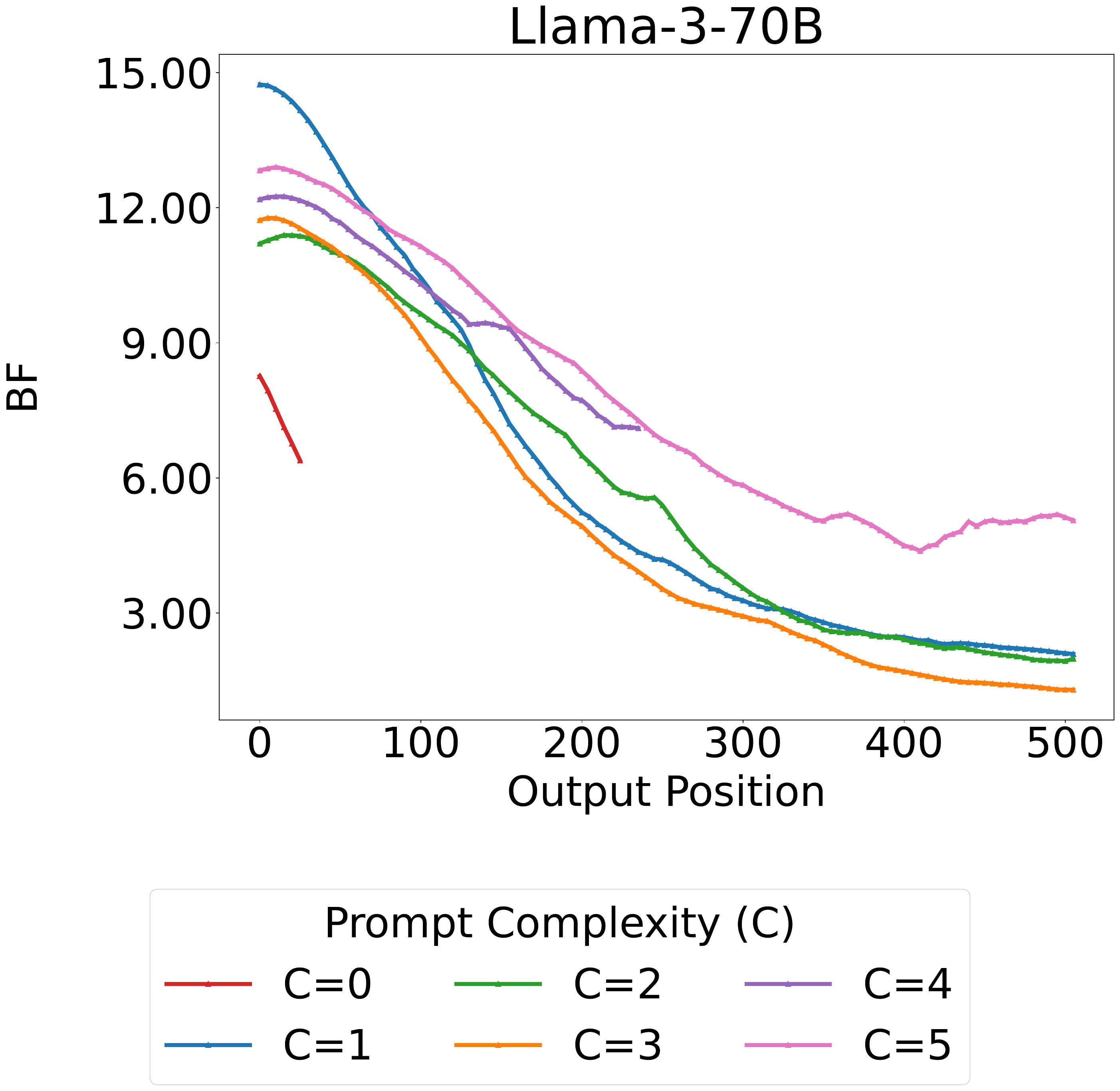}
     \label{fig:output_dynamic_base_cognac_random_str_app}
    \end{subfigure}
    \begin{subfigure}[t]{0.24\textwidth}
    \centering
     \includegraphics[width=0.9\linewidth]{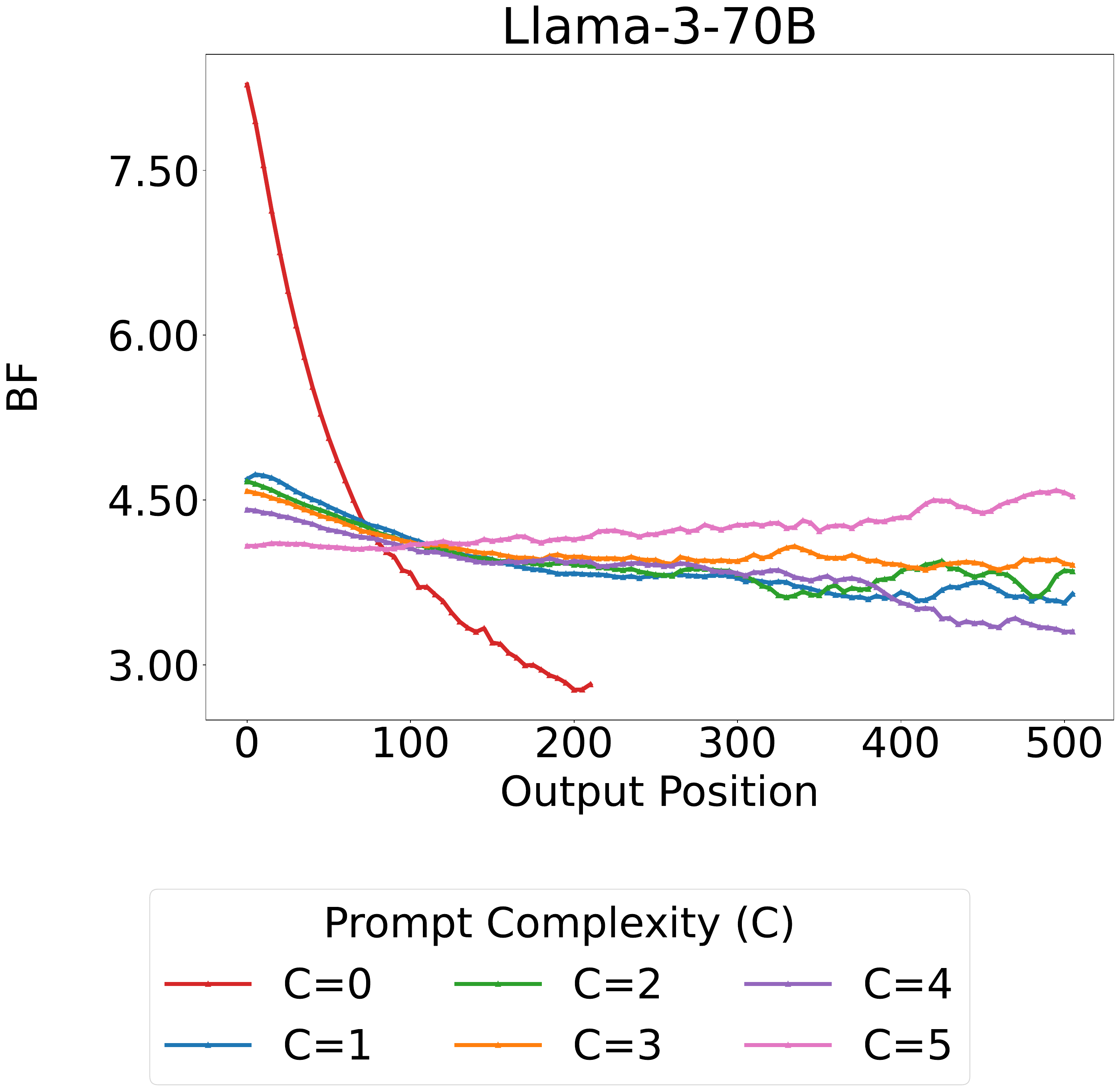}
     \label{fig:output_dynamic_base_bbcnews_app}
    \end{subfigure}
        \begin{subfigure}[t]{0.24\textwidth}
    \centering
     \includegraphics[width=0.9\linewidth]{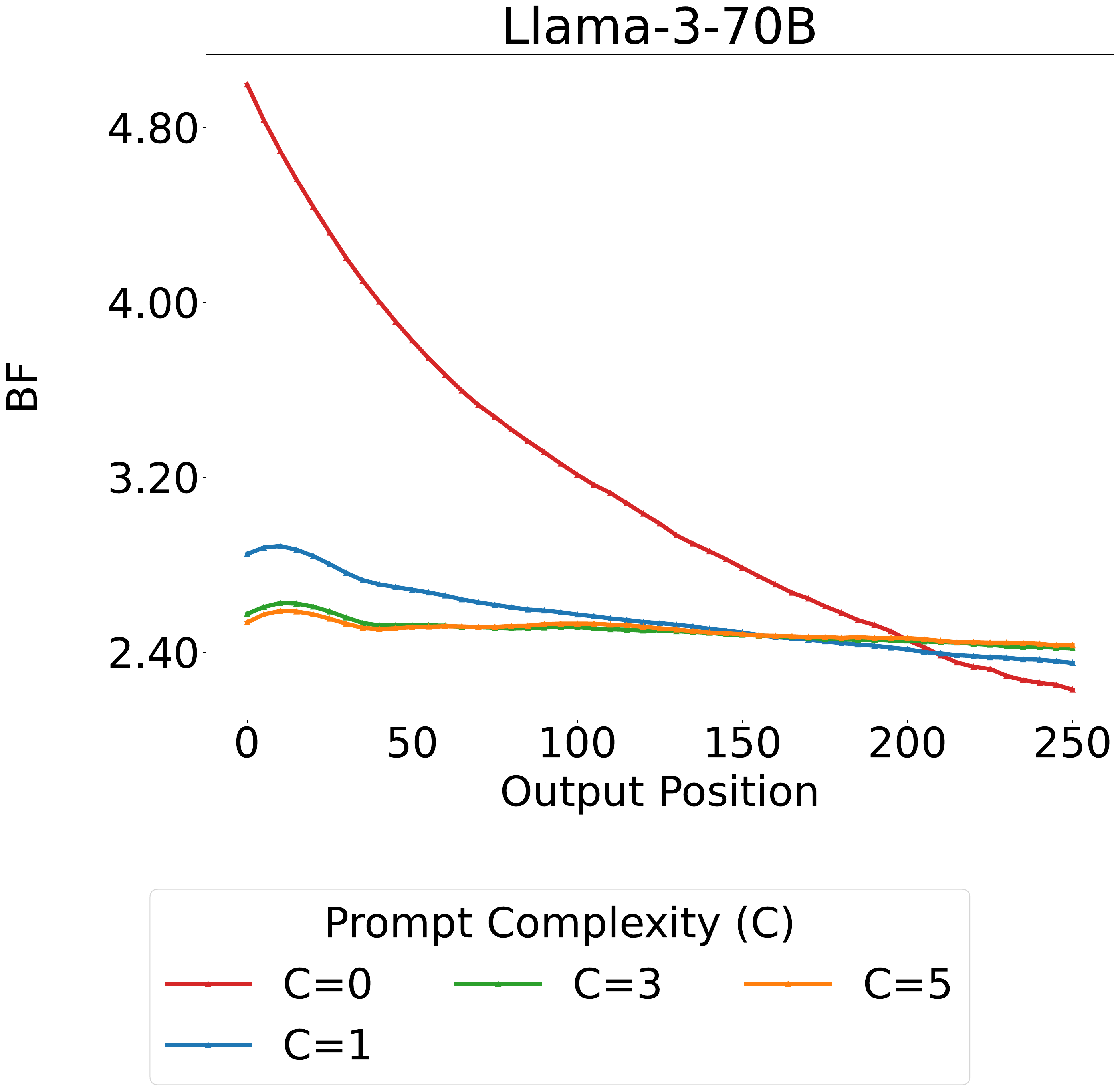}

     \label{fig:output_dynamic_base_mmlu_app}
    \end{subfigure}

    \begin{subfigure}[t]{0.24\textwidth}
    \centering
     \includegraphics[width=0.9\linewidth]{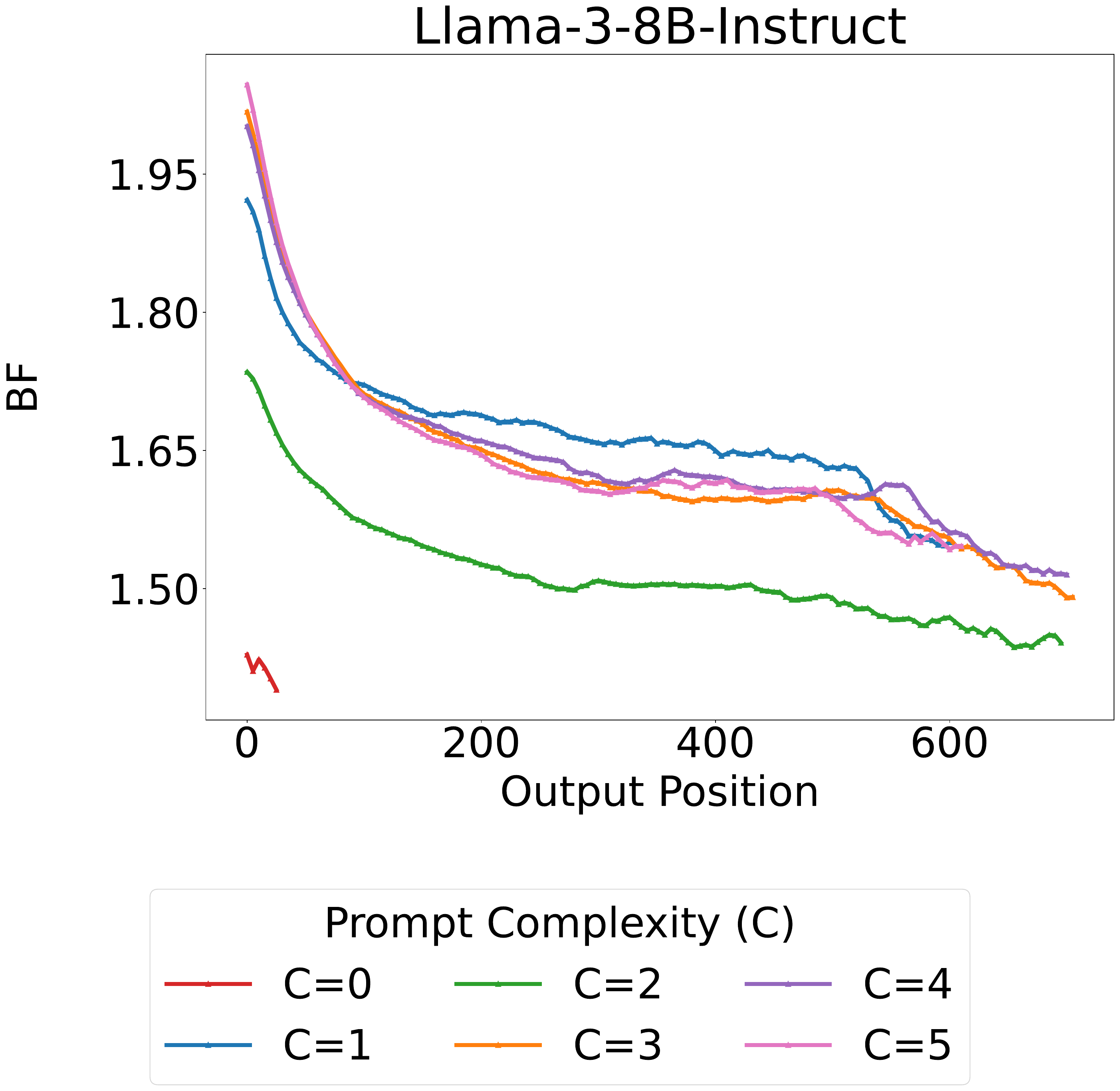}

     \label{fig:output_dynamic_base_storytelling_8b_instruct_app}
    \end{subfigure}
        \begin{subfigure}[t]{0.24\textwidth}
    \centering
     \includegraphics[width=0.9\linewidth]{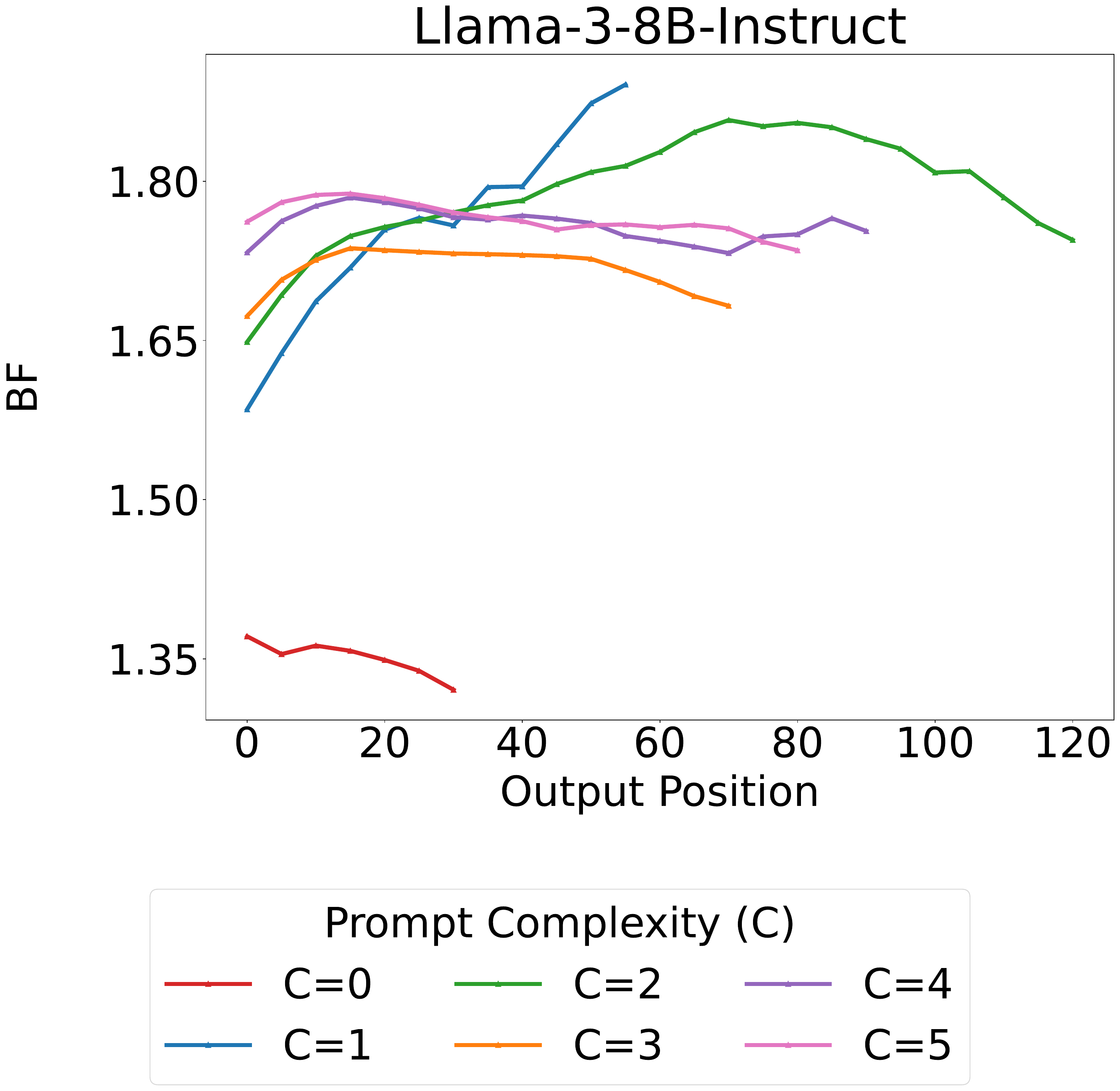}

     \label{fig:output_dynamic_base_cognac_random_str_8b_instruct_app}
    \end{subfigure}
    \begin{subfigure}[t]{0.24\textwidth}
    \centering
     \includegraphics[width=0.9\linewidth]{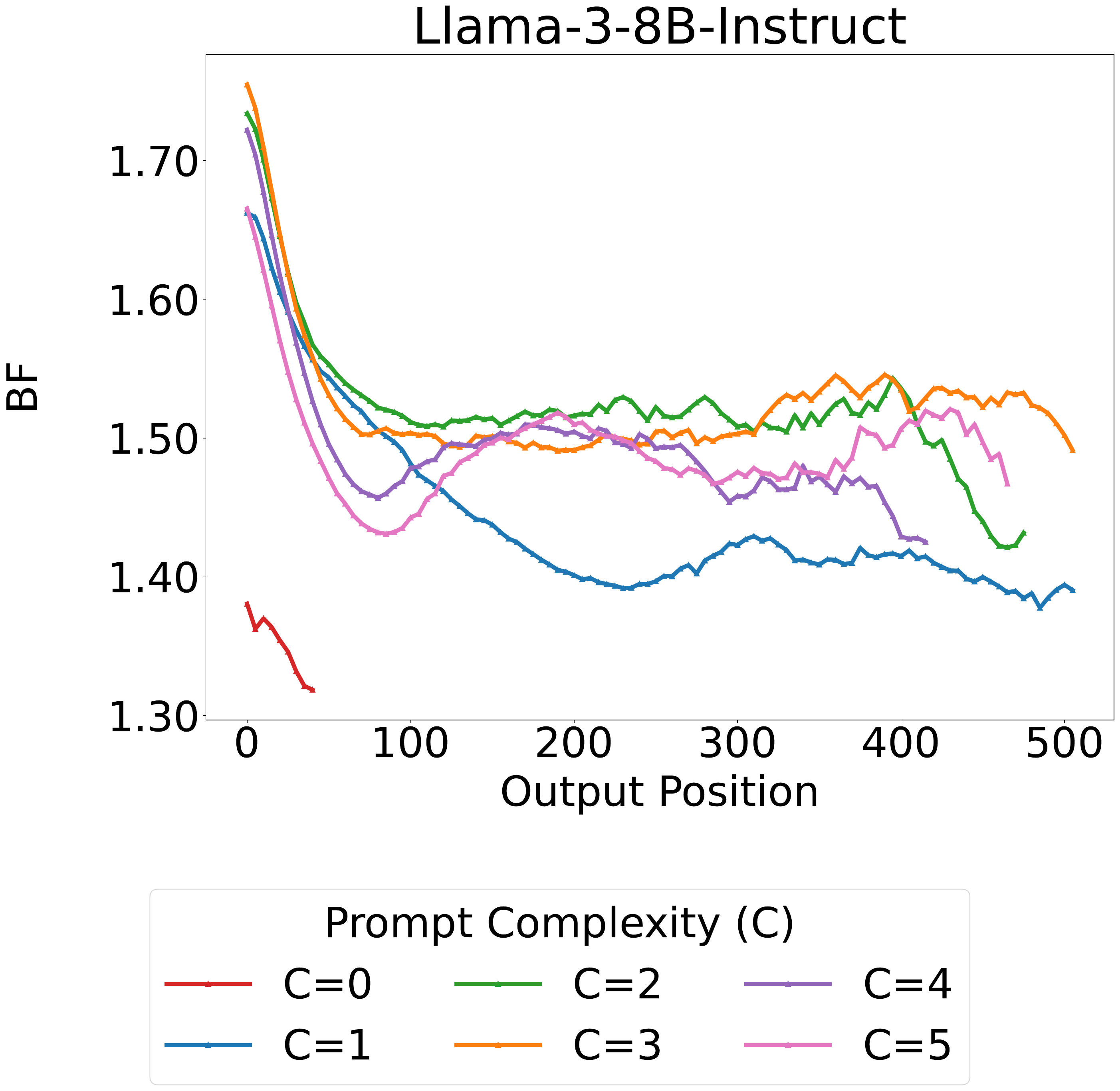}

     \label{fig:output_dynamic_base_bbcnews_8b_instruct_app}
    \end{subfigure}
        \begin{subfigure}[t]{0.24\textwidth}
    \centering
     \includegraphics[width=0.9\linewidth]{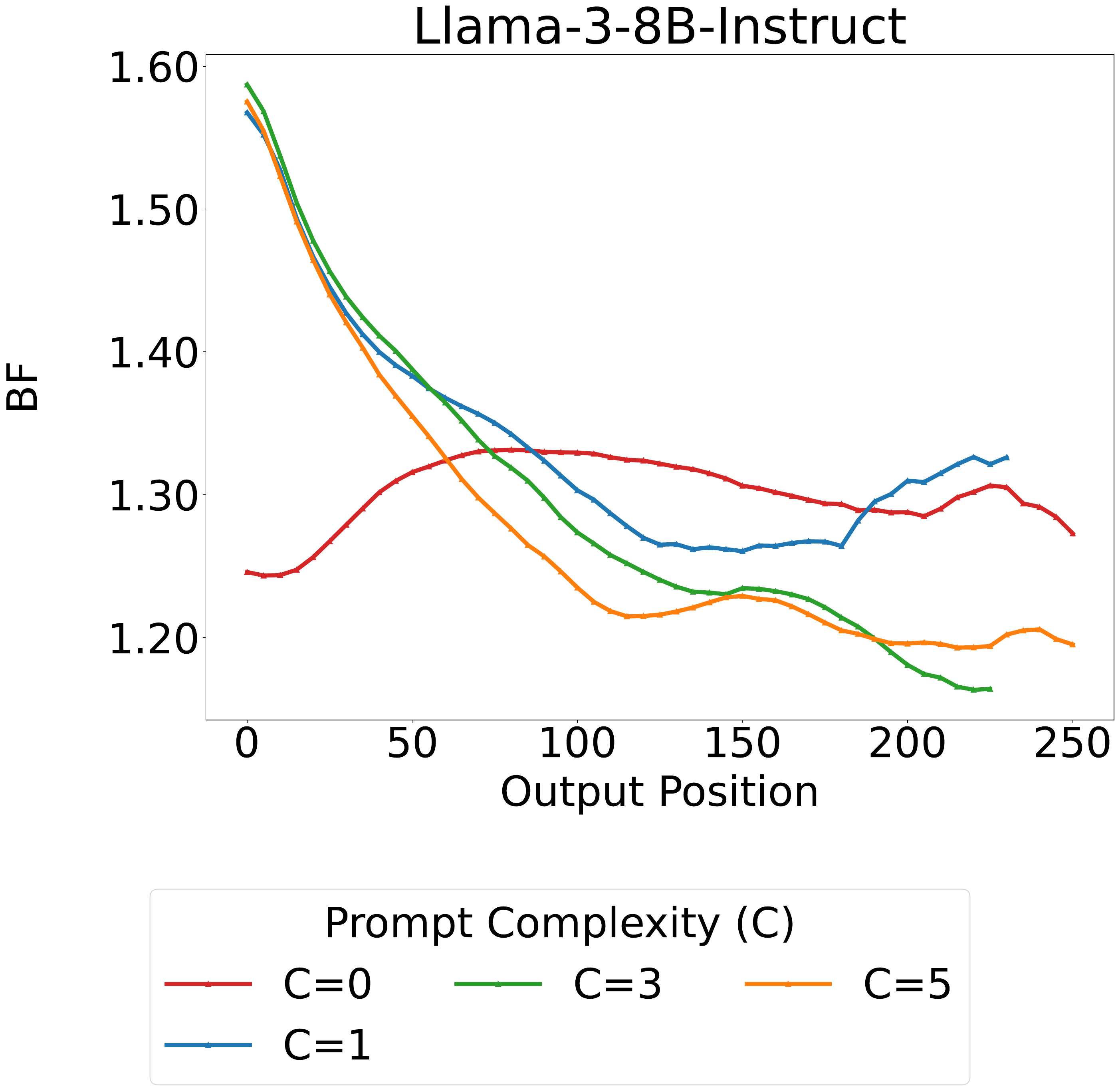}

     \label{fig:output_dynamic_base_mmlu_8b_instruct_app}
    \end{subfigure}

    \begin{subfigure}[t]{0.24\textwidth}
    \centering
     \includegraphics[width=0.9\linewidth]{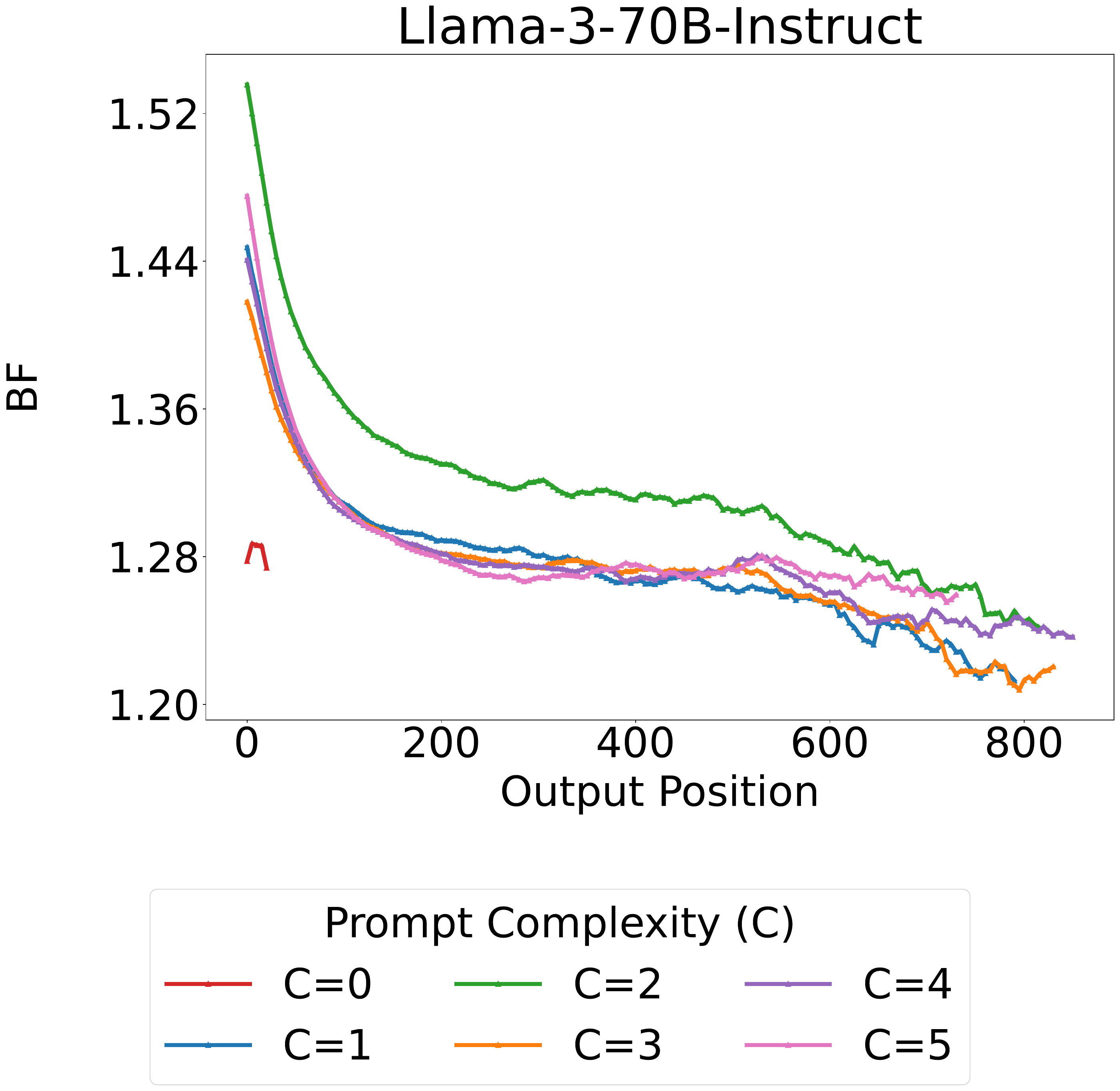}
    \caption{Creative StoryGen}
     \label{fig:output_dynamic_omstrict_storytelling_app}
    \end{subfigure}
        \begin{subfigure}[t]{0.24\textwidth}
    \centering
     \includegraphics[width=0.9\linewidth]{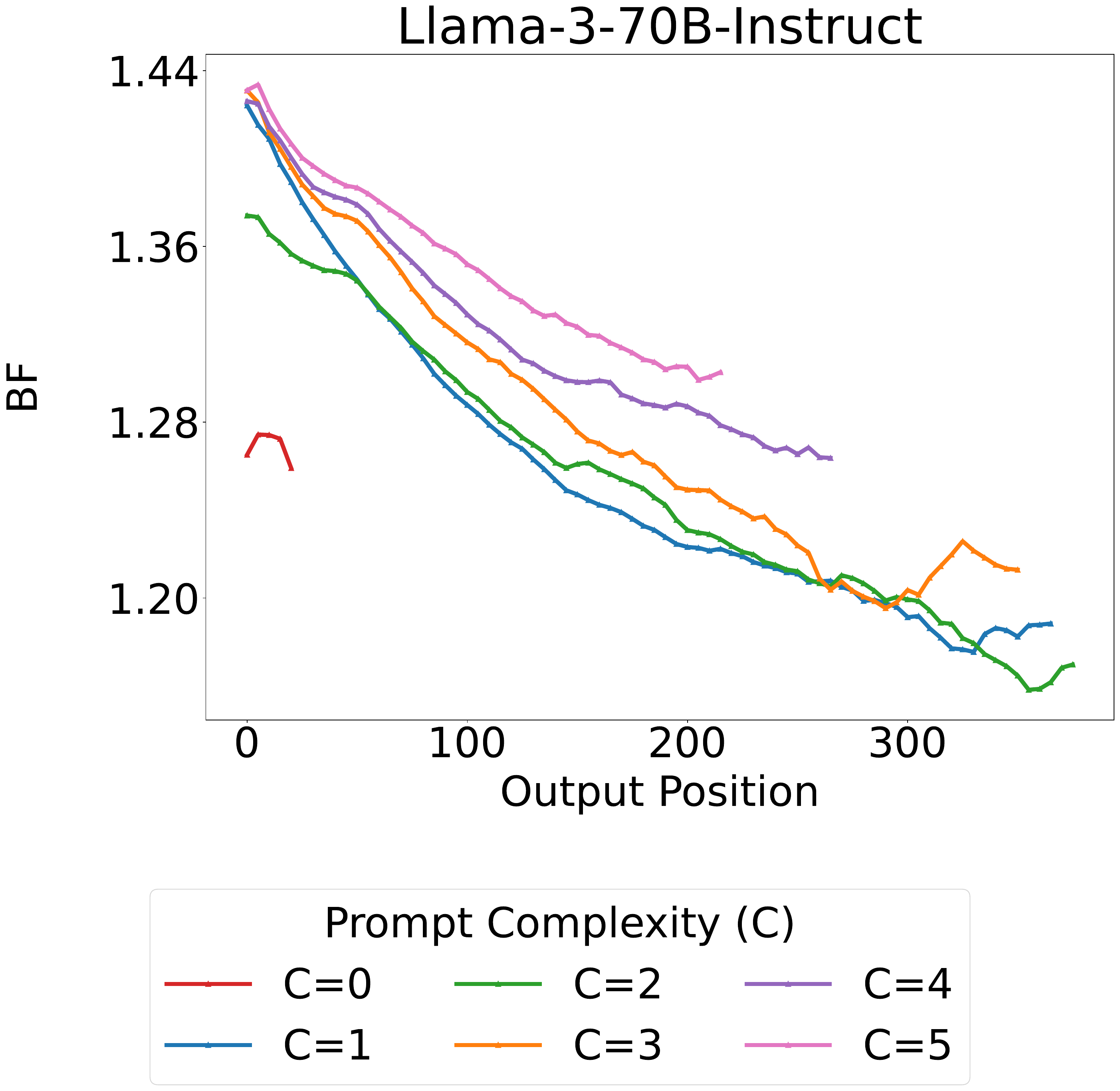}
    \caption{Random Strings}
     \label{fig:output_dynamic_instruct_cognac_random_str_app}
    \end{subfigure}
    \begin{subfigure}[t]{0.24\textwidth}
    \centering
     \includegraphics[width=0.9\linewidth]{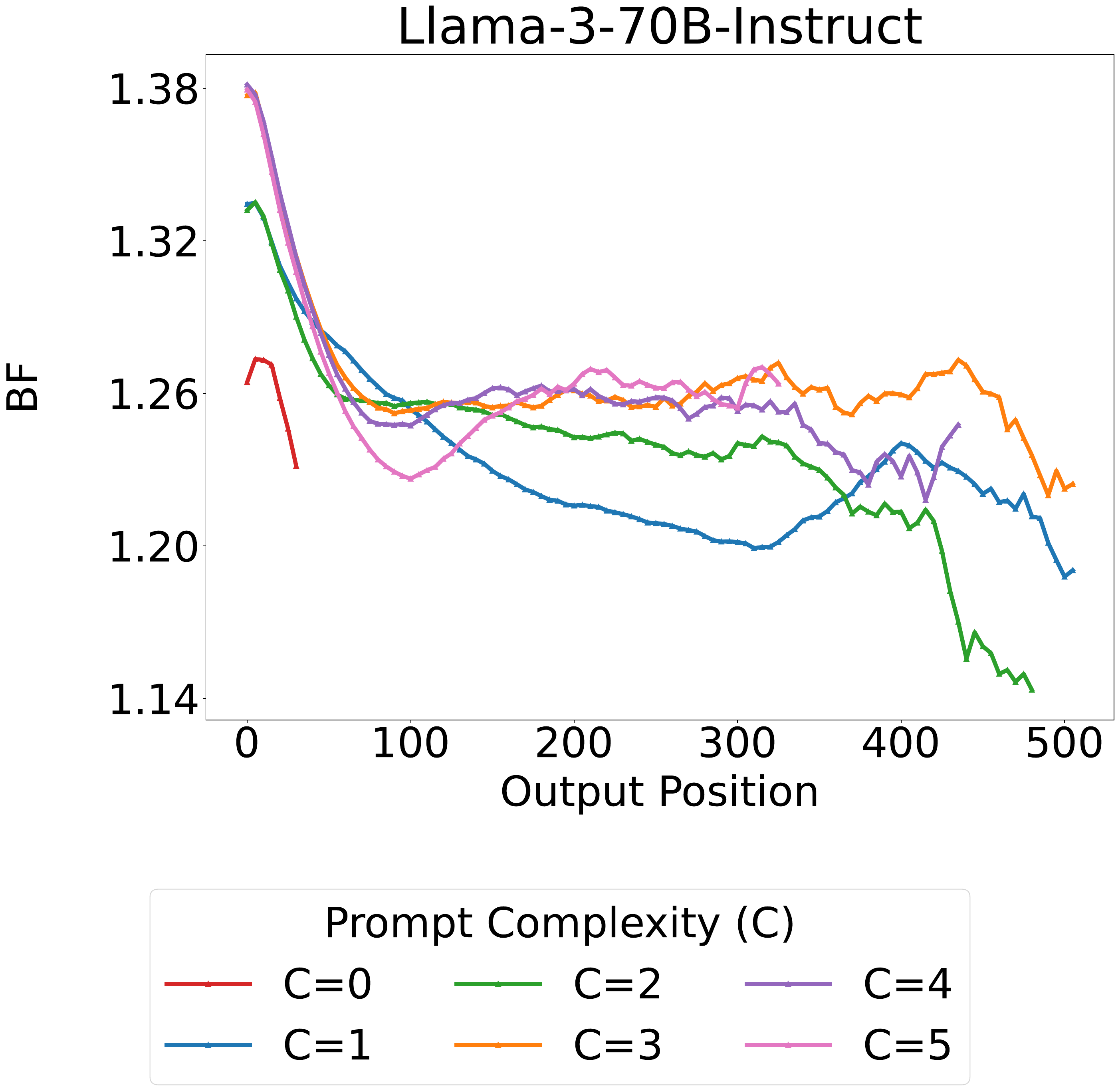}

    \caption{BBCNewsLatest}
     \label{fig:output_dynamic_instruct_bbcnews_app}
    \end{subfigure}
        \begin{subfigure}[t]{0.24\textwidth}
    \centering
     \includegraphics[width=0.9\linewidth]{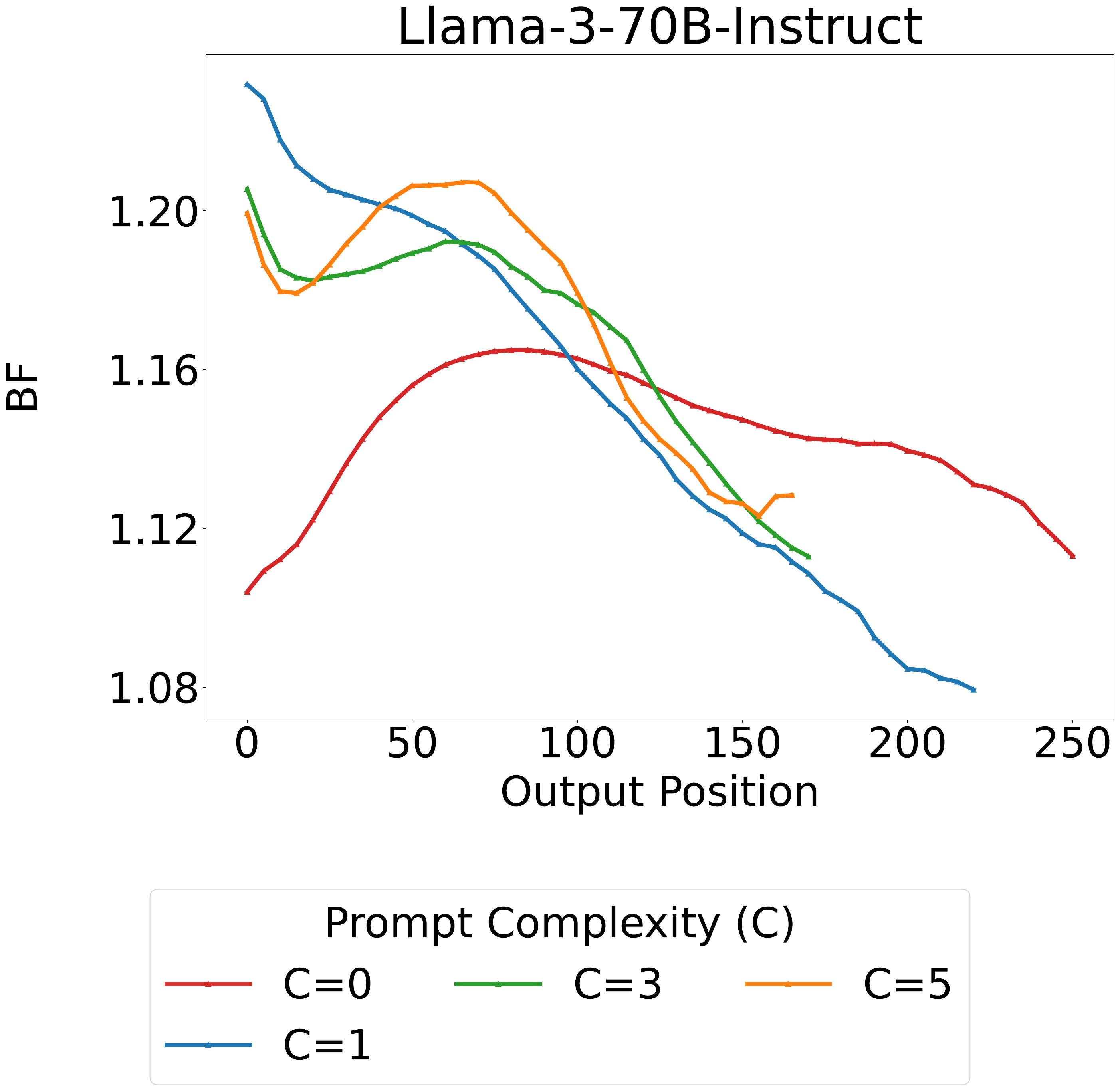}

    \caption{MMLU}
     \label{fig:output_dynamic_instruct_mmlu_app}
    \end{subfigure}

    \caption{\textbf{BF Output Dynamic for Llama-3-families.} For better visualization, we compute the exponential moving averaged values of perplexity with the smoothing factor set as $0.1$.
    }
    \label{fig: output_dynamic_app_llama3}
\end{figure*}

\begin{figure}[h!]
    \centering
    \includegraphics[width=0.5\linewidth]{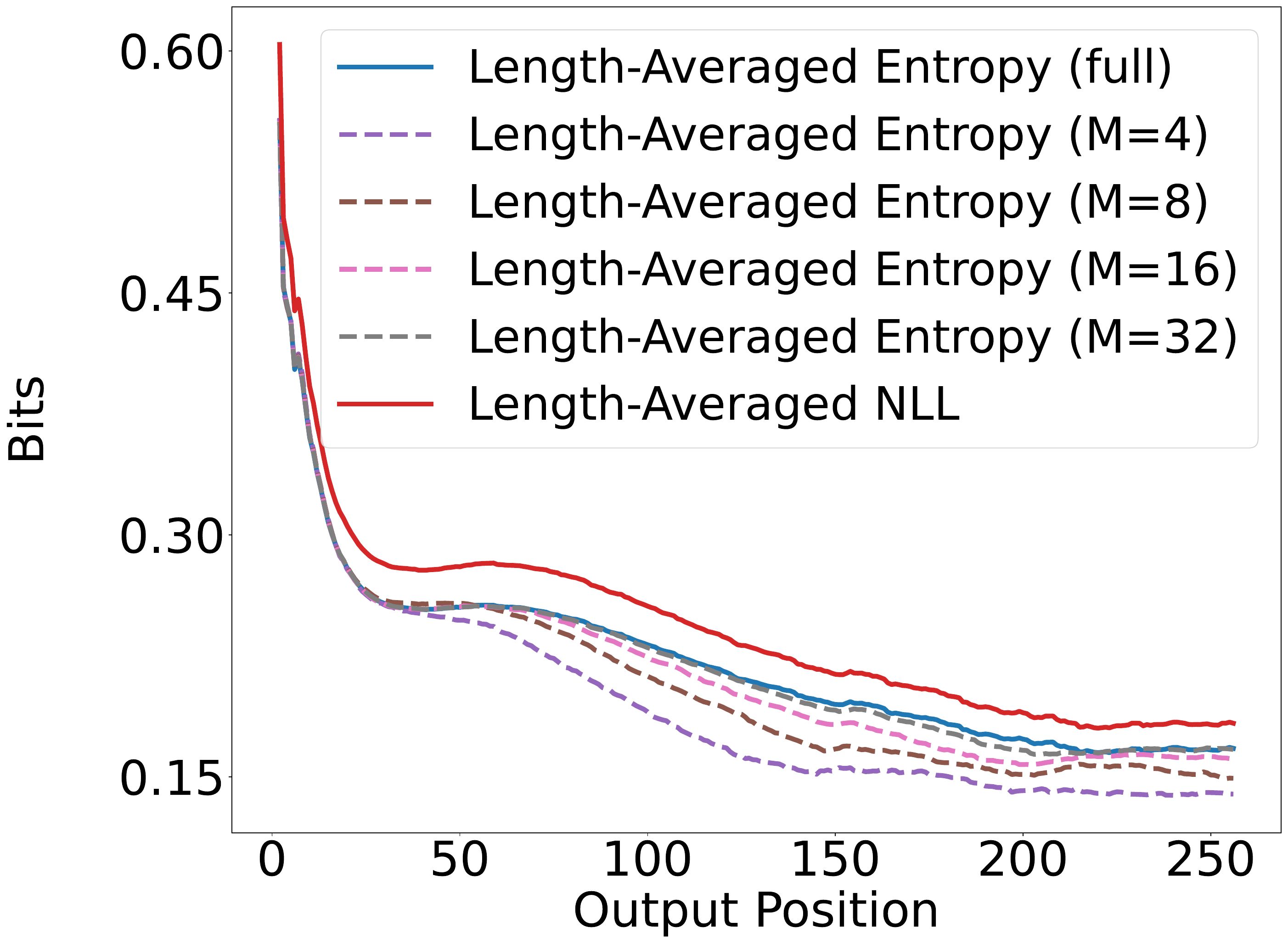}
    \caption{\textbf{Monte Carlo (MC) sampling systematically underestimates entropy.} The plot shows that the estimated entropy of sequences from Llama-3-8B-Instruct increases with the number of MC samples ($M$). A small sample size fails to cover the vast output space, leading to an underestimation of the true entropy. This bias is difficult to eliminate without incurring substantial computational costs.}
    \label{fig: underestimation_entropy}
\end{figure}
\section{Entropy Underestimation via Monte Carlo Sampling}
\label{appendix: under_estimation_of_entropy_via_mc}

To demonstrate the limitations of Monte Carlo (MC) sampling for entropy estimation in long sequences, we conducted an empirical study. We prompted {Llama-3-8B-Instruct} with 5-shot CoT examples from the MMLU dataset. We then estimated the entropy of its generated responses using a varying number of MC samples: $M \in \{4, 8, 16, 32, 64\}$. 

As illustrated in Figure \ref{fig: underestimation_entropy}, the estimated entropy consistently increases with the number of samples. This trend confirms that MC estimation with a small sample size systematically \textbf{underestimates} the true entropy because it fails to capture the long tail of the full probability distribution. While increasing the sample count mitigates this bias, it does so at a significant computational cost. In contrast, ~\cref{thm: aep_llm} allows us to use the negative log-likelihood (NLL) of a single typical sequence for a more efficient and accurate estimation.

\section{Full BF Output Dynamics Investigation}
\label{app: full_output_bf}
Here we present full task-wise and model-wise BF output dynamic for Llama-2 in \cref{fig: output_dynamic_app_llama2} and Llama-3  in \cref{fig: output_dynamic_app_llama3}. We can observe the trends as in \cref{sec: bf_dynamic}: \circone
\textbf{The average BF for the base model} (~$\approx 12$) \textbf{is roughly ten times higher than the aligned model} ($\approx 1.2$). 
\circtwo \textbf{BF would often drop smoothly as more output tokens are generated}. 

\section{Curious Case of Prompt Complexity}
\label{app: curious_case_prompt_complexity}
\begin{figure}[t]
    \centering
             \begin{subfigure}[t]{0.31\textwidth}
    \centering
     \includegraphics[width=\linewidth]{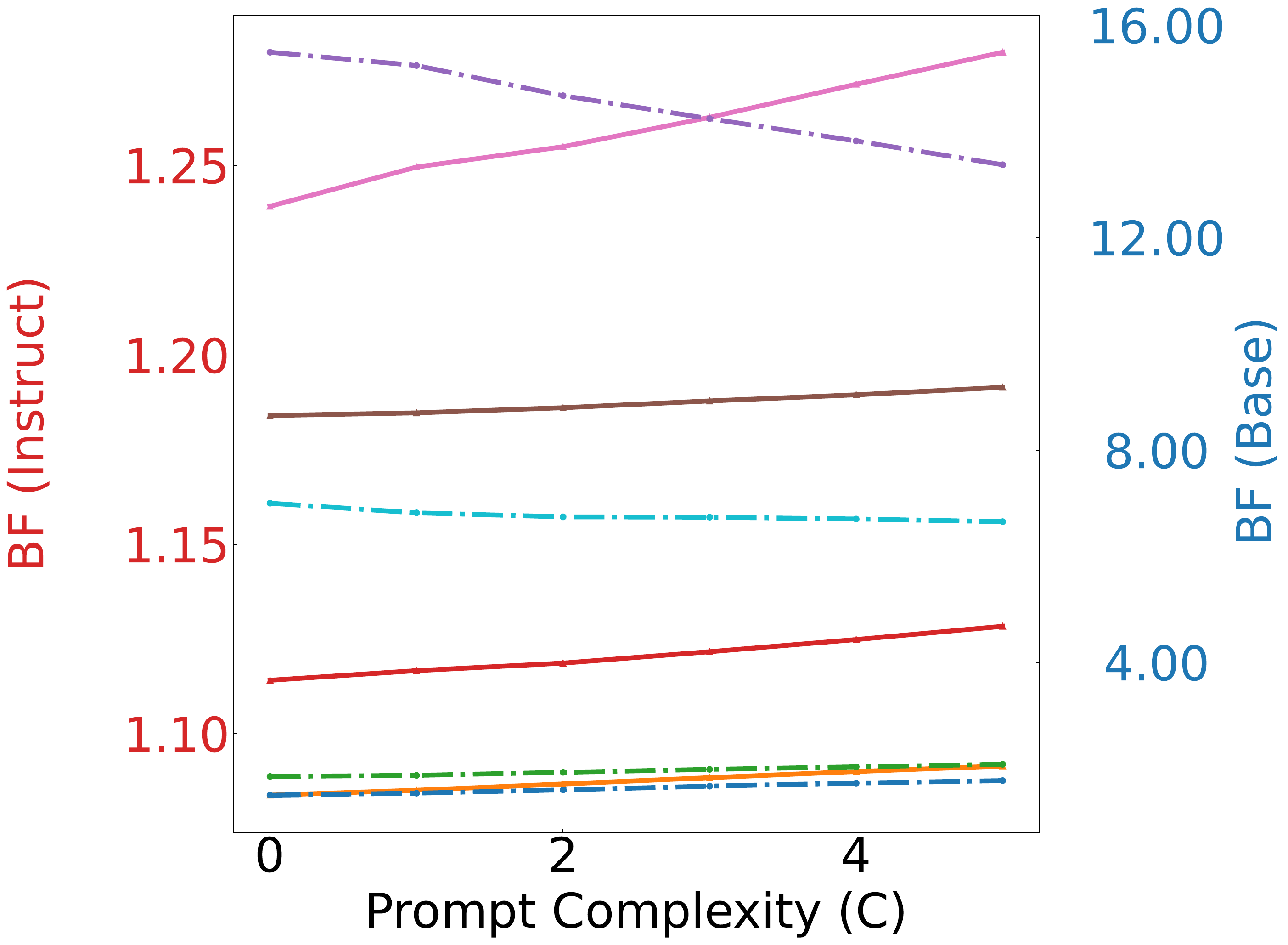}
    \caption{Cognac}
     \label{fig:cognac_ppl_p_main}
    \end{subfigure}
         \begin{subfigure}[t]{0.31\textwidth}
    \centering
     \includegraphics[width=\linewidth]{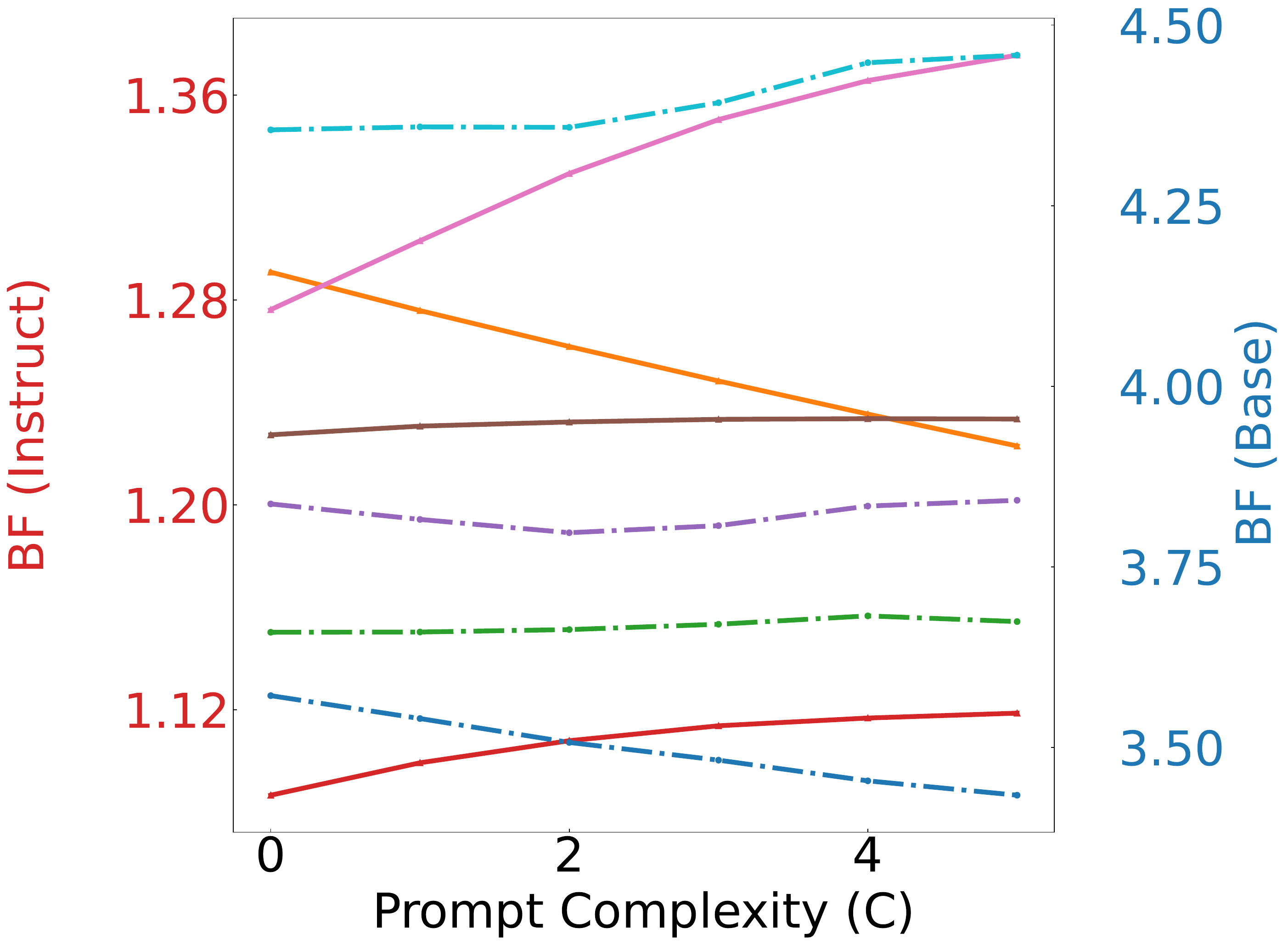}
    \caption{BBCNewsLatest}
     \label{fig:bbcnews_ppl_p_main}
    \end{subfigure}
    \begin{subfigure}[t]{0.2\textwidth}
    \centering
        \includegraphics[width=\linewidth]{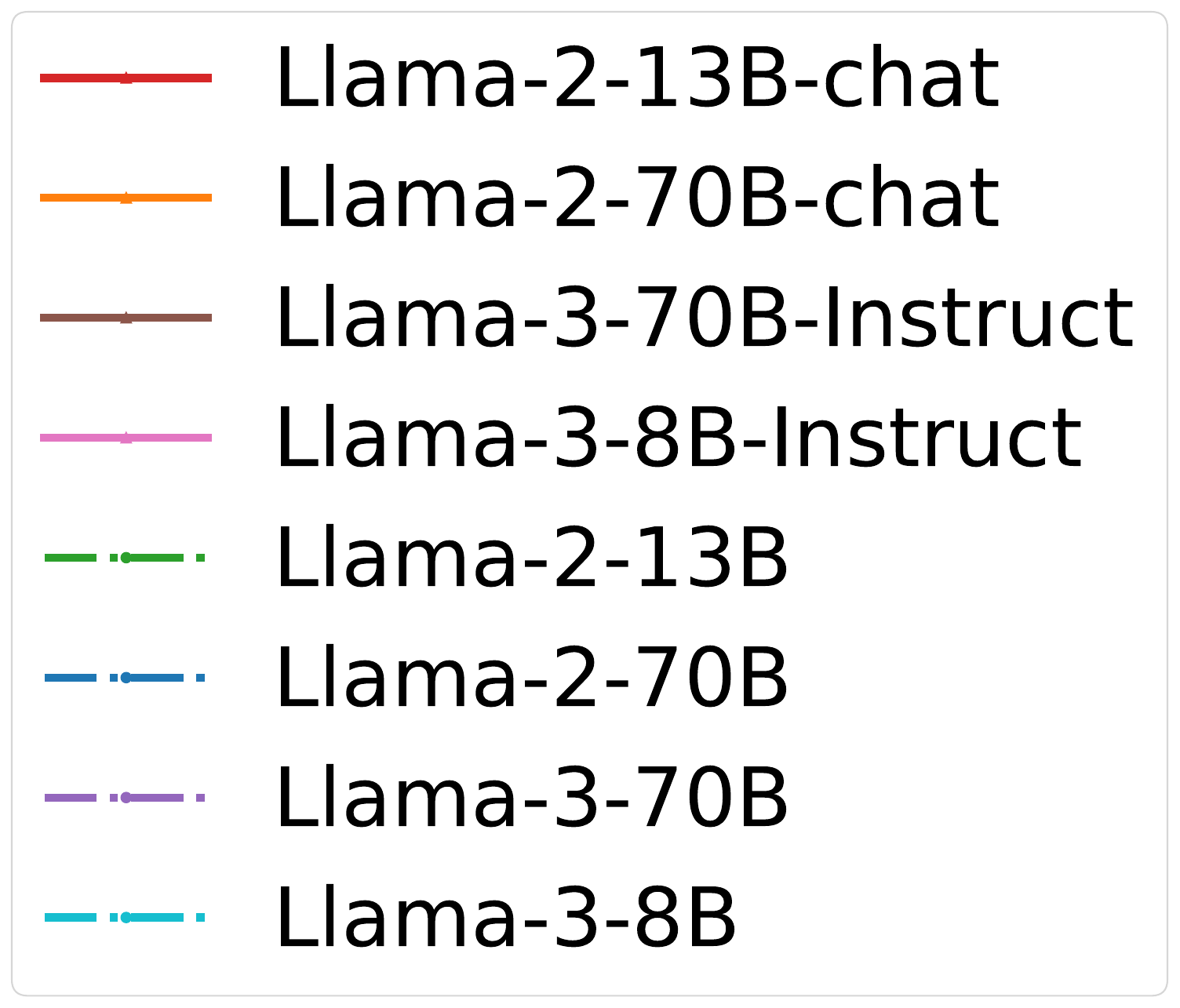}
    \end{subfigure}
    \vspace{-9pt}
      \caption{\textbf{Task-varied influence of prompt complexity $C$ on BF.} ~On Cognac, we see BF increases with increased $C$, while on BBCNewsLatest, increasing $C$ can lead to reduced BF. }
          \vspace{-9pt}
     \label{fig:prompt_complexity_bf_example}
    \end{figure}
Intuitively, greater prompt specificity (larger $C$) reduces BF by narrowing the model’s output space through more informative context.
However, our experimental results reveal task-varied effects. As illustrated in \cref{fig:prompt_complexity_bf_example} for the Cognac task, greater prompt complexity can \textit{increase} BF--potentially due to the cognitive burden of processing negation or complex linguistic structures. In contrast, for tasks like News Generation, higher $C$ generally leads to lower BF, consistent with the expected narrowing of output diversity. Comprehensive task-wise BF results are provided in \cref{app: full_taskwise_bf}. \mvhnrevise{This negation-induced increase is one instance of a more general pattern -- content that is unexpected from the model's own predictive viewpoint raises BF -- which we examine in \cref{app: bf_self_narrowing}.}

\section{Full Task-wise BF Evaluation on Different Prompt Complexity}
\label{app: full_taskwise_bf}
\begin{figure}[htbp]
    \centering
    \begin{subfigure}[t]{0.28\textwidth}
    \centering
     \includegraphics[width=\linewidth]{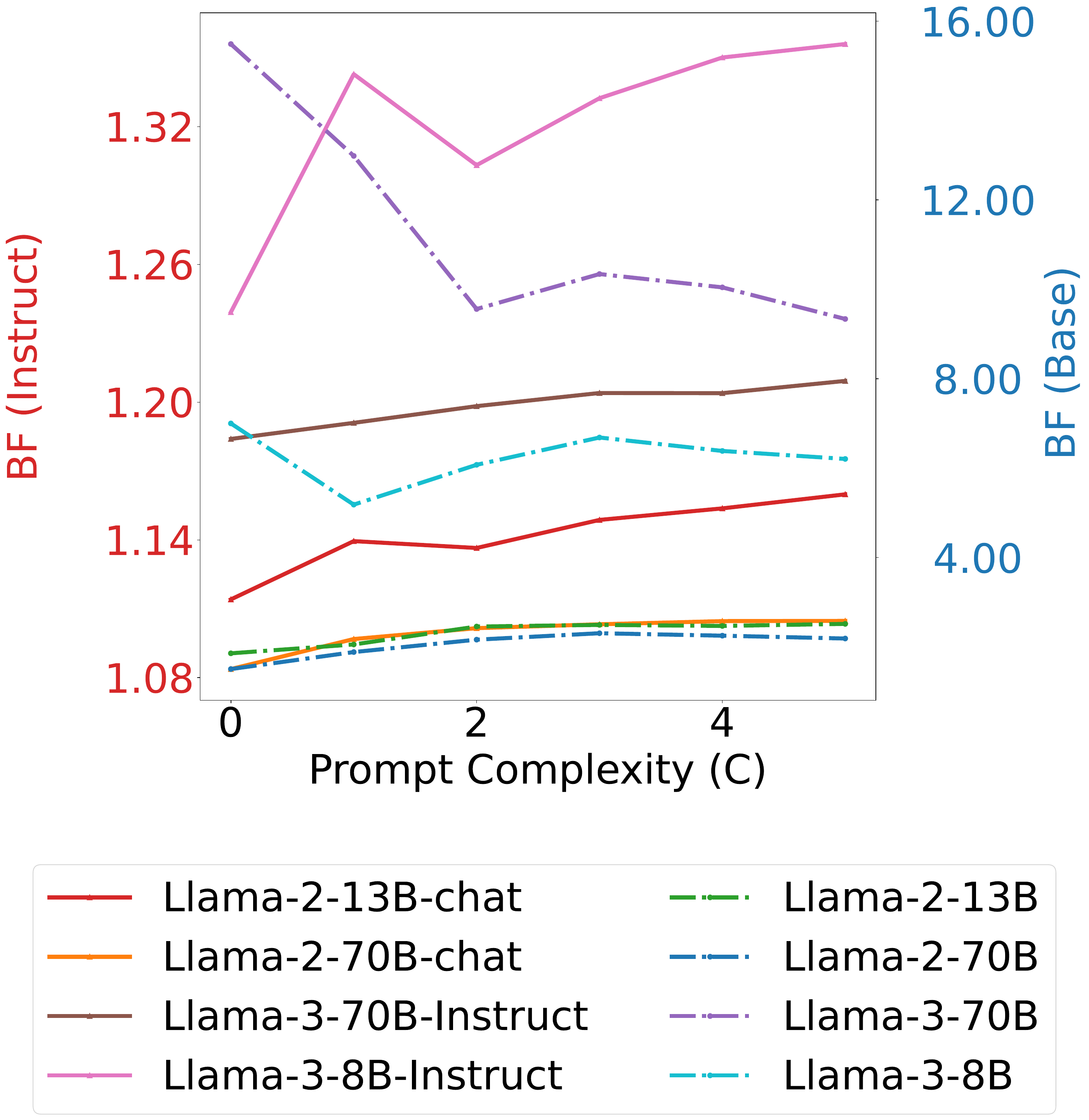}
    \caption{Cognac}
     \label{fig:cognac_ppl_p}
    \end{subfigure}
    \begin{subfigure}[t]{0.28\textwidth}
    \centering
     \includegraphics[width=\linewidth]{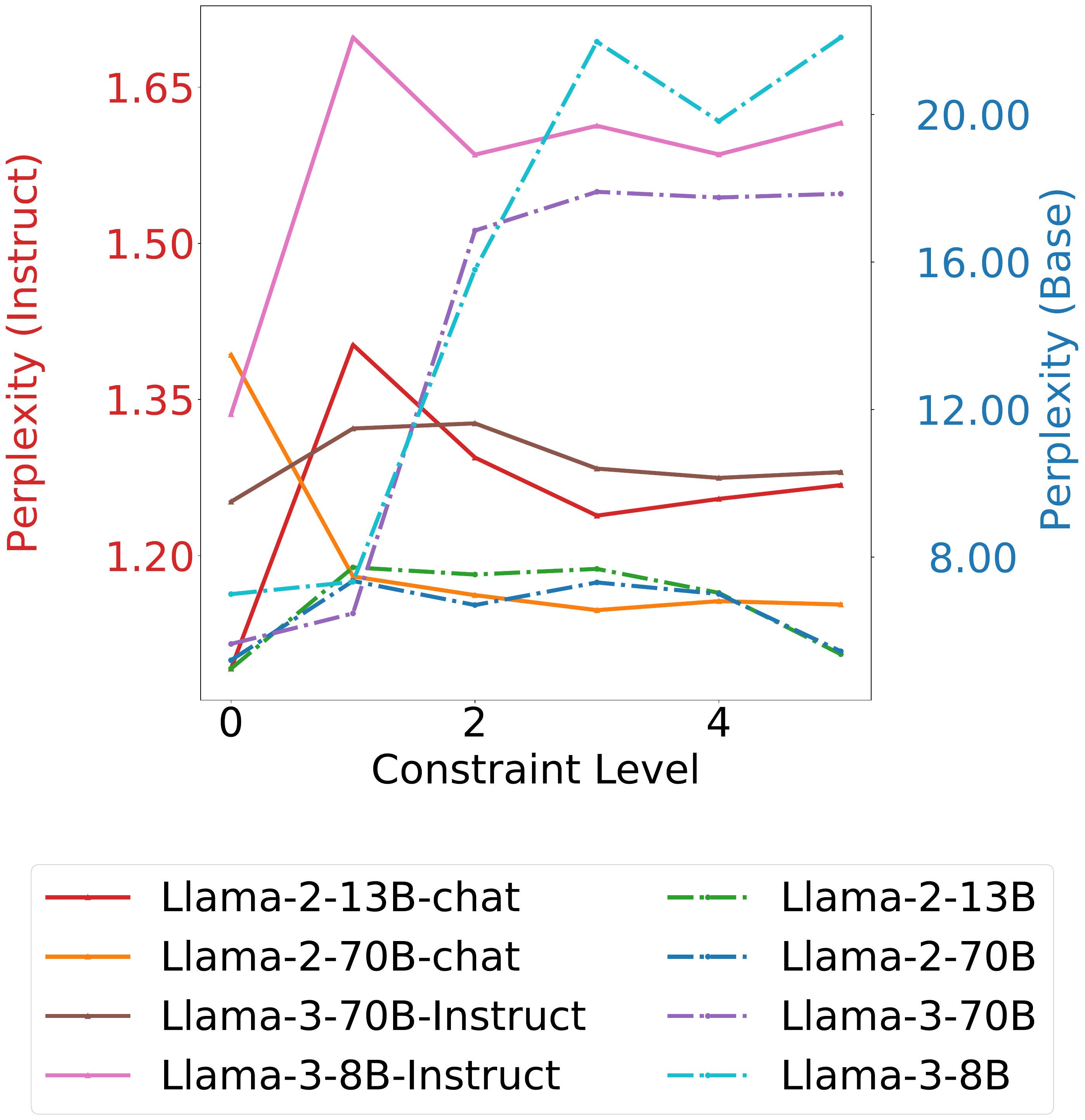}
    \caption{Creative StoryGen}
     \label{fig:storytelling_ppl_p}
    \end{subfigure}
        \begin{subfigure}[t]{0.28\textwidth}
    \centering
     \includegraphics[width=\linewidth]{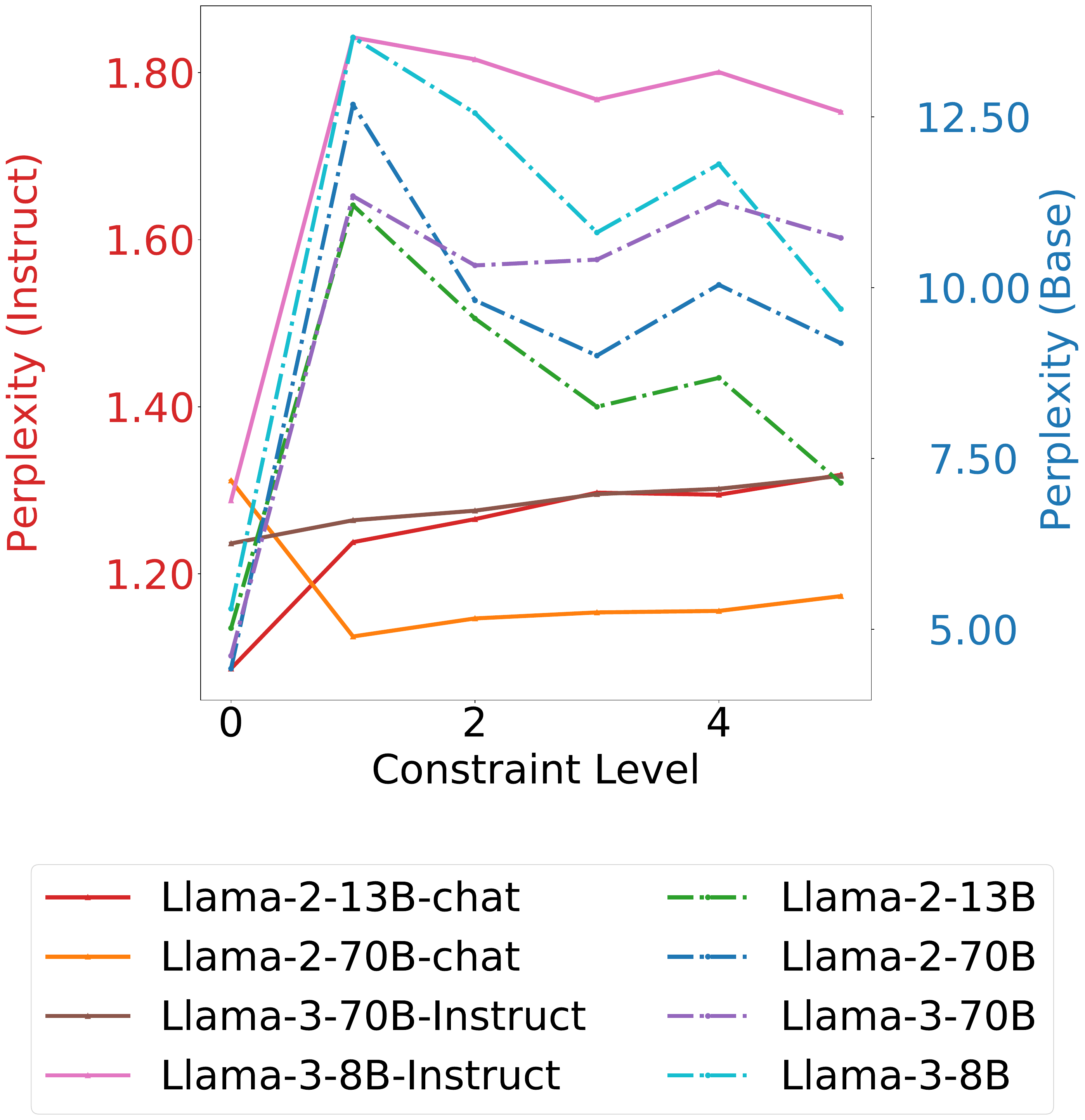}
    \caption{Random Strings}
     \label{fig:random_str_ppl_p}
    \end{subfigure}

    \begin{subfigure}[t]{0.28\textwidth}
    \centering
     \includegraphics[width=\linewidth]{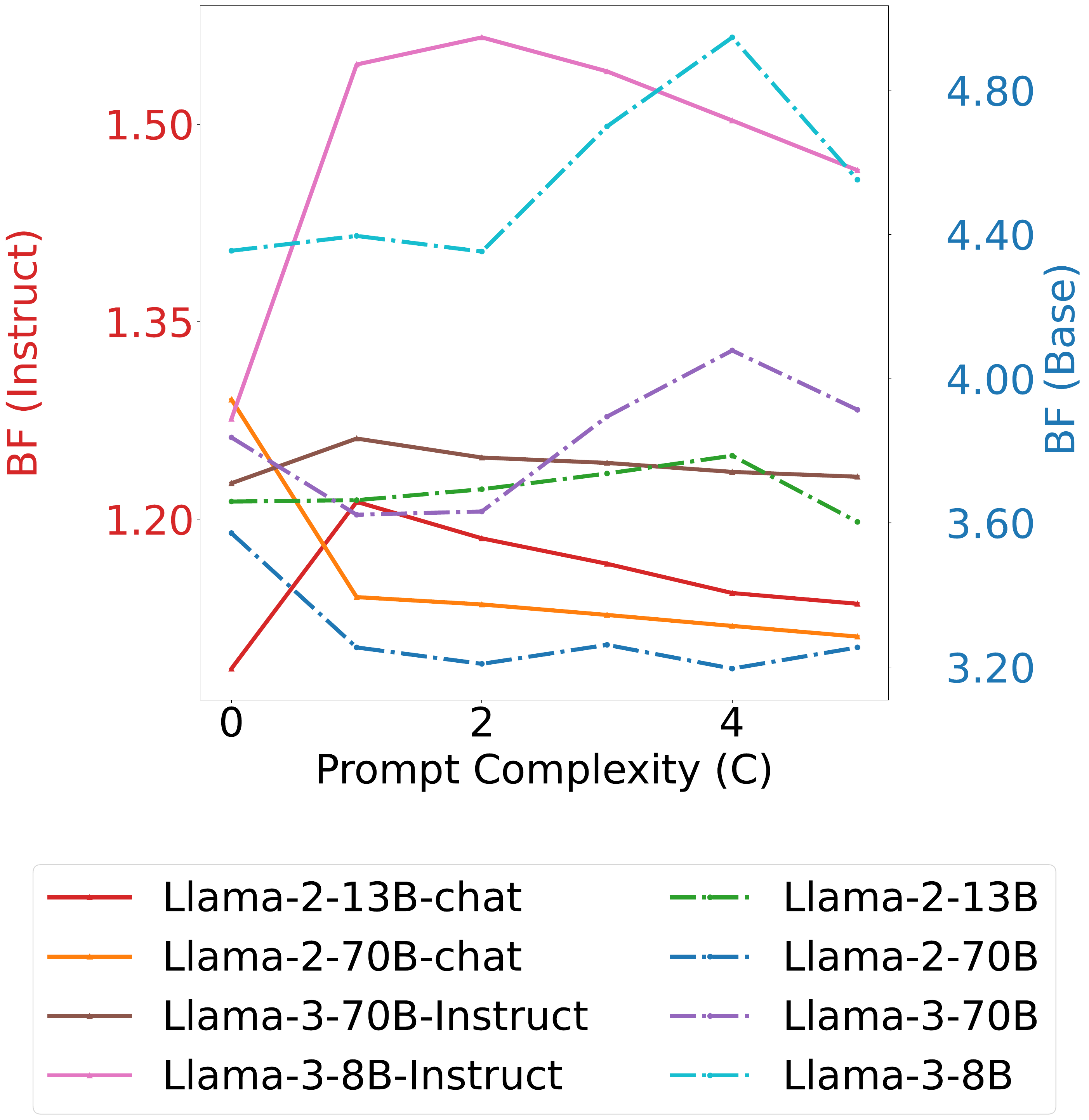}
    \caption{BBCNewsLatest}
     \label{fig:bbcnews_ppl_p}
    \end{subfigure}
    \caption{
    BF changes with prompt complexity ($C$) for Different Tasks. We can see prompt complexity affects BF in a task-varied way. 
    }
      \label{fig: bf_different_tasks_appendix}
\end{figure}

The full task-wise BF evaluation results over different prompt complexity can be found in \cref{fig: bf_different_tasks_appendix}. Here we can see that prompt complexity modulates BF in highly non-consistent ways across models and tasks, and there are no clear monotonic patterns, contradicting the intuition that with more context given, the model should have more confidence in what to generate. 

\section{Generalization to Additional Tasks and Models}
\label{app: additional_verification}
\begin{figure}[p]
\centering
\begin{subfigure}[t]{0.24\textwidth}
    \centering
     \includegraphics[width=0.9\linewidth]{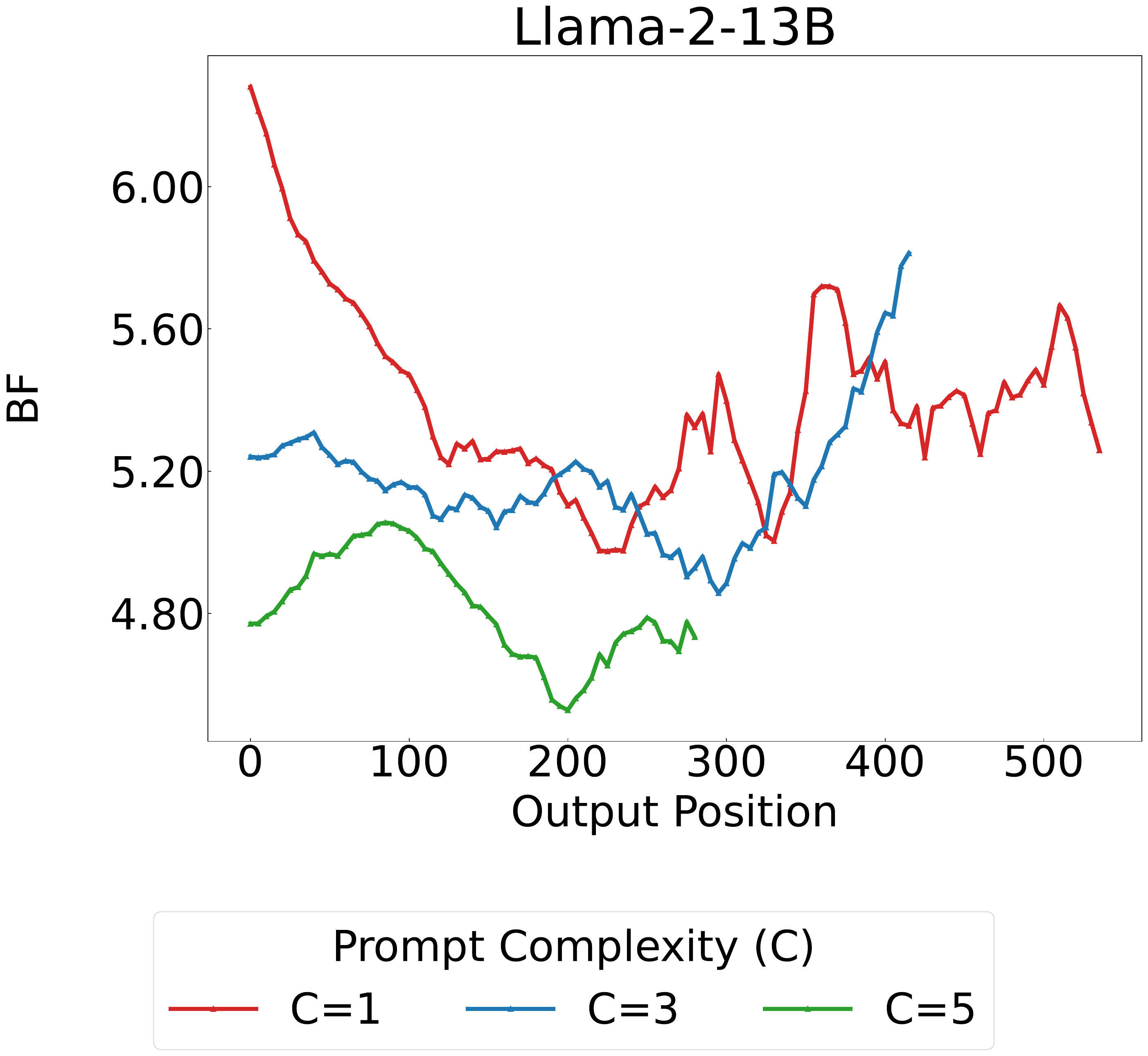}
     \label{fig:output_dynamic_base_xsum_llama2_13b_app}
     \caption{XSUM}
    \end{subfigure}
    \begin{subfigure}[t]{0.24\textwidth}
    \centering
     \includegraphics[width=0.9\linewidth]{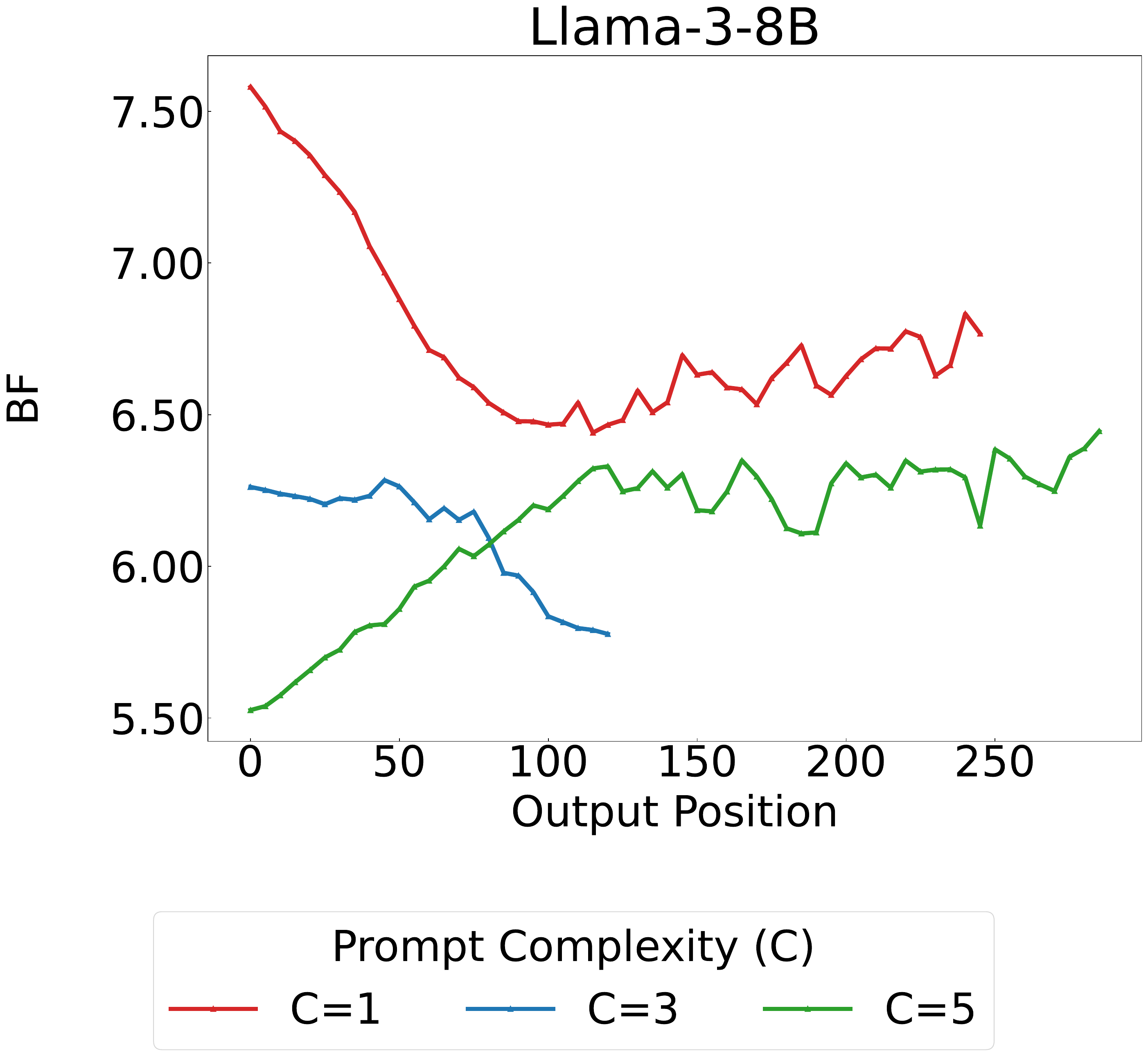}
     \label{fig:output_dynamic_base_xsum_8b_app}
     \caption{XSUM}
    \end{subfigure}
                \begin{subfigure}[t]{0.24\textwidth}
    \centering
     \includegraphics[width=0.9\linewidth]{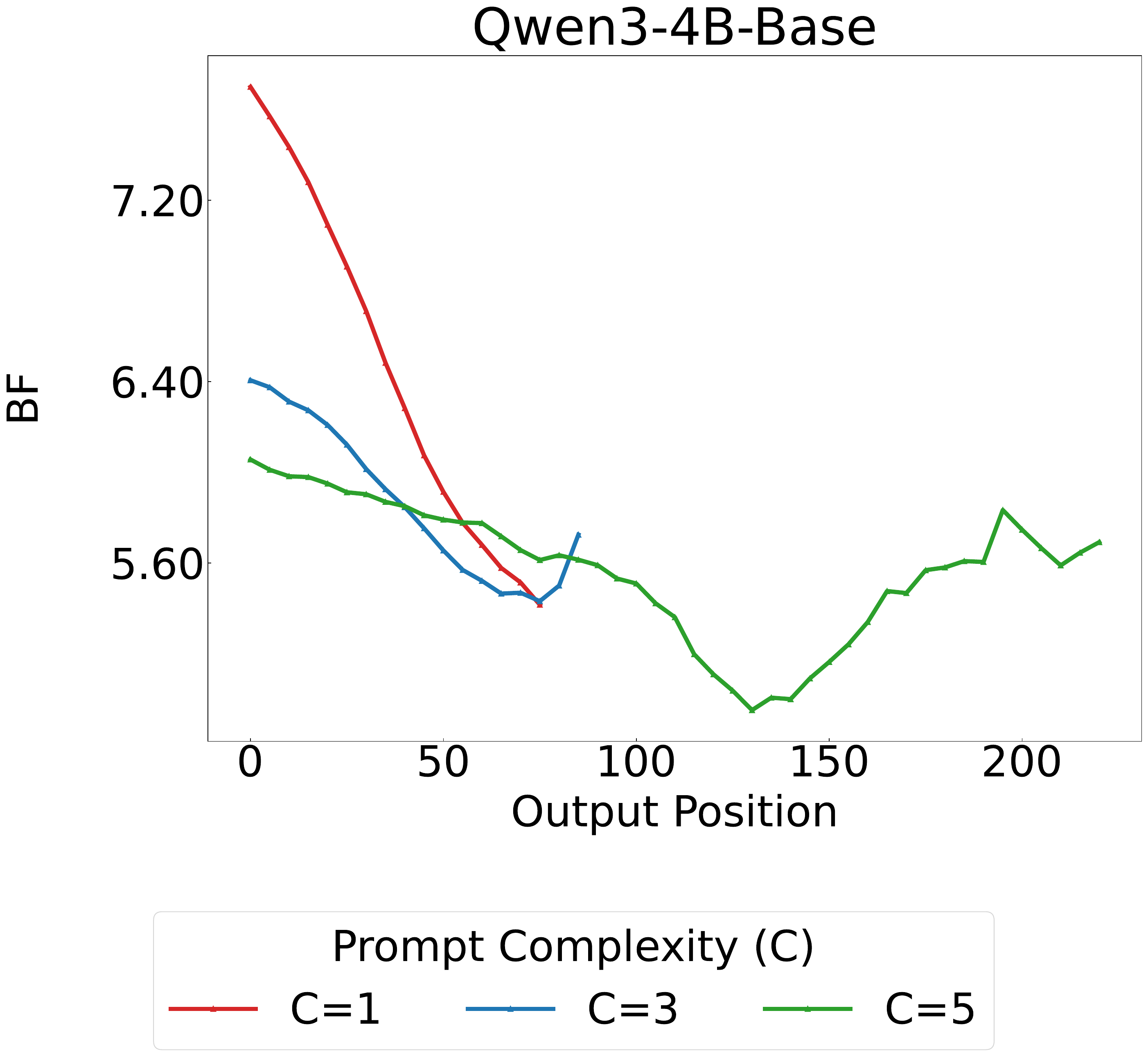}
     \label{fig:output_dynamic_base_xsum_qwen3_4b_app}
     \caption{XSUM}
    \end{subfigure}
    \begin{subfigure}[t]{0.24\textwidth}
    \centering
     \includegraphics[width=0.9\linewidth]{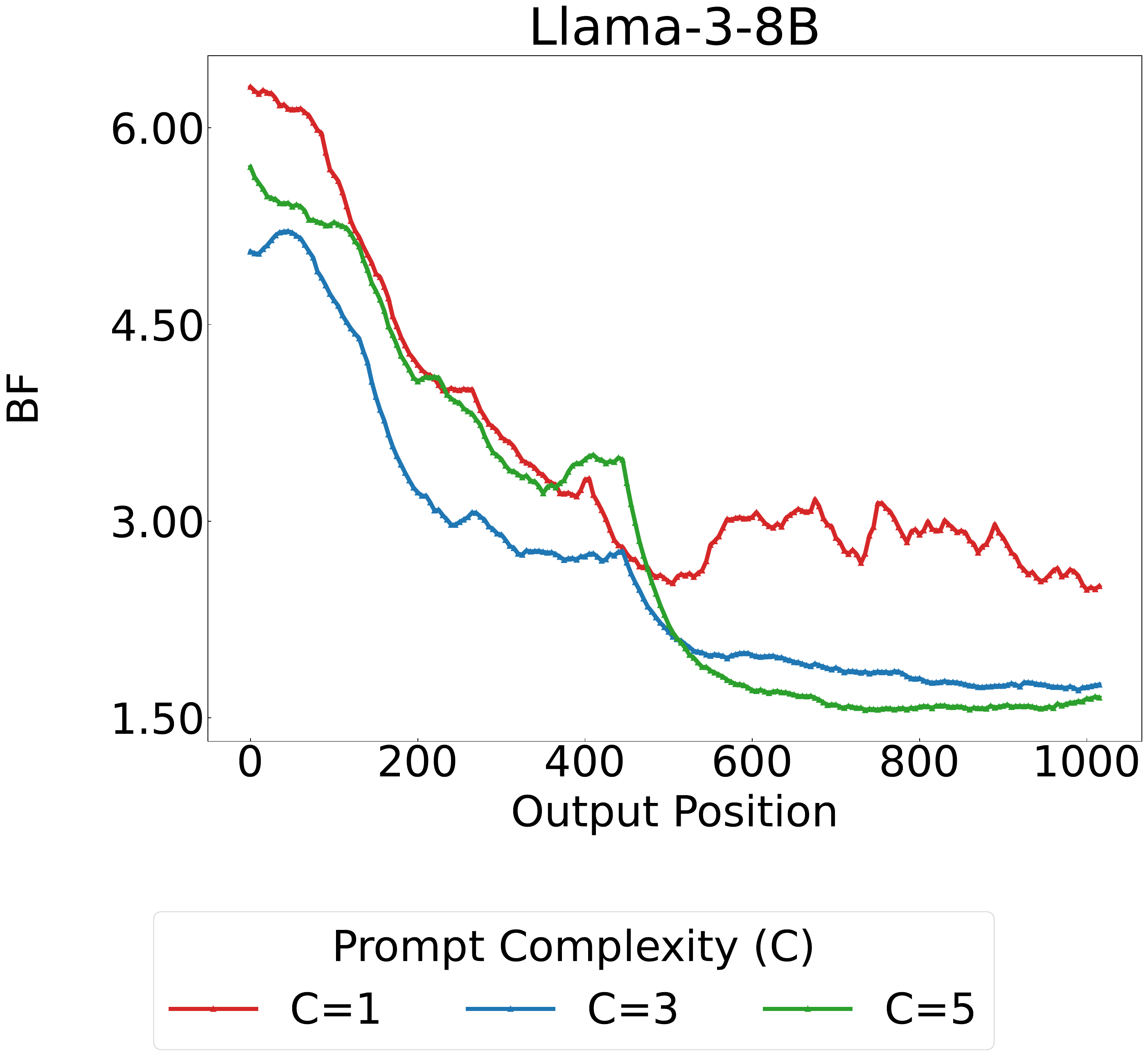}
     \label{fig:output_dynamic_base_aya_random_str_8b_app}
     \caption{Aya}
    \end{subfigure}
        \begin{subfigure}[t]{0.24\textwidth}
    \centering
     \includegraphics[width=0.9\linewidth]{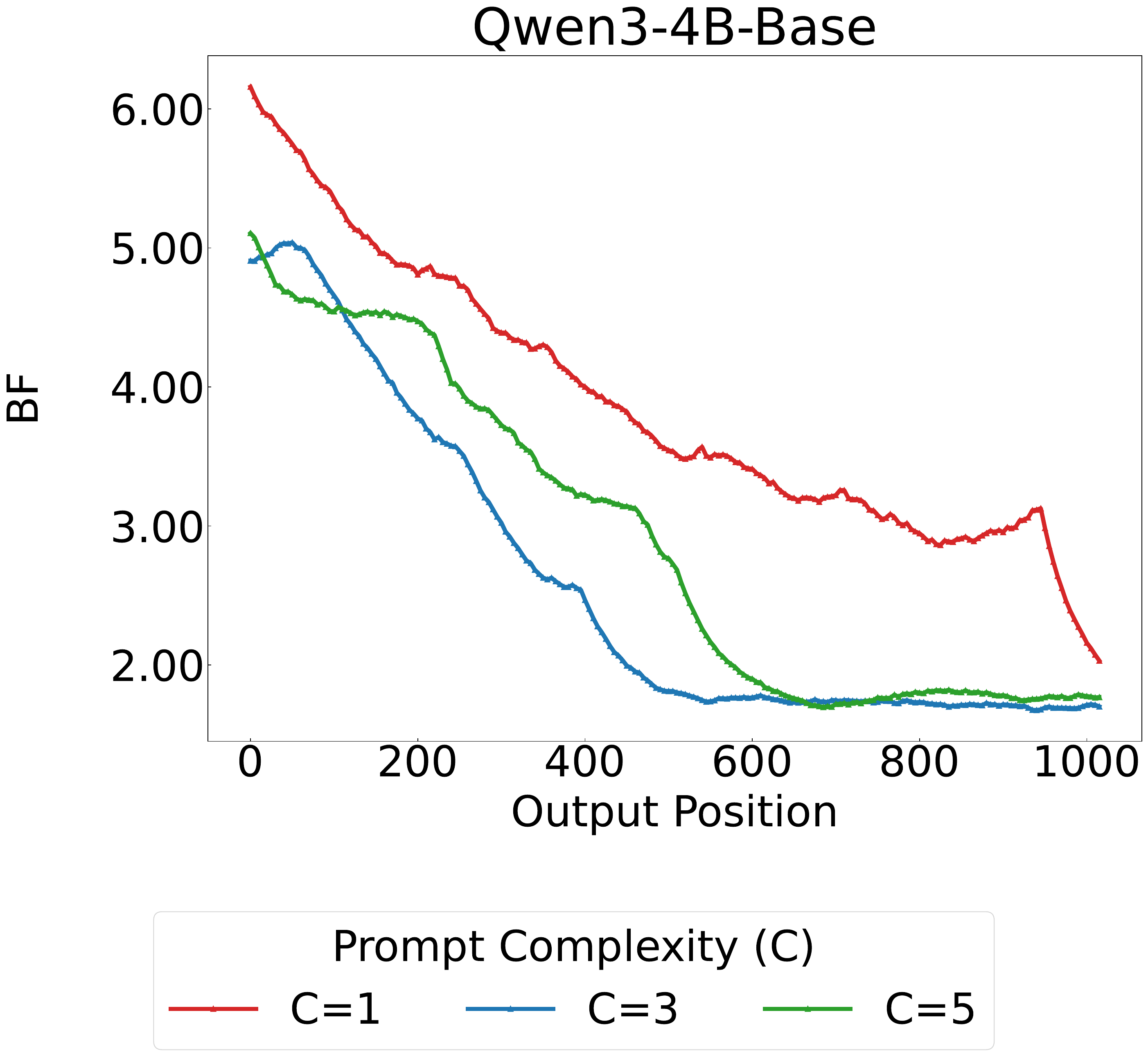}
     \label{fig:output_dynamic_base_aya_qwen3_4b_app}
     \caption{Aya}
    \end{subfigure}
\begin{subfigure}[t]{0.24\textwidth}
    \centering
     \includegraphics[width=0.9\linewidth]{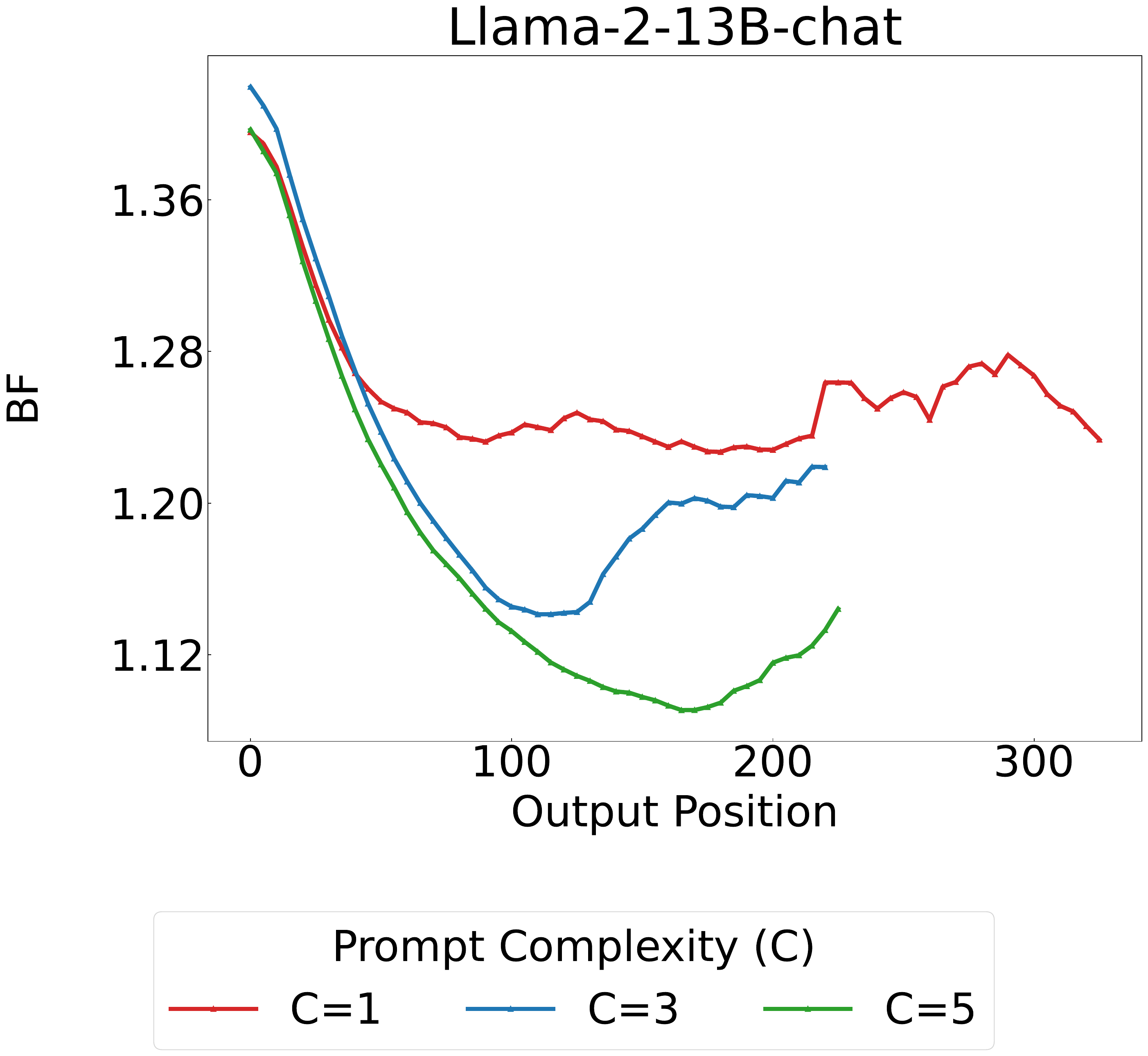}
     \label{fig:output_dynamic_instruct_xsum_llama2_13b_app}
     \caption{XSUM}
    \end{subfigure}
    \begin{subfigure}[t]{0.24\textwidth}
    \centering
     \includegraphics[width=0.9\linewidth]{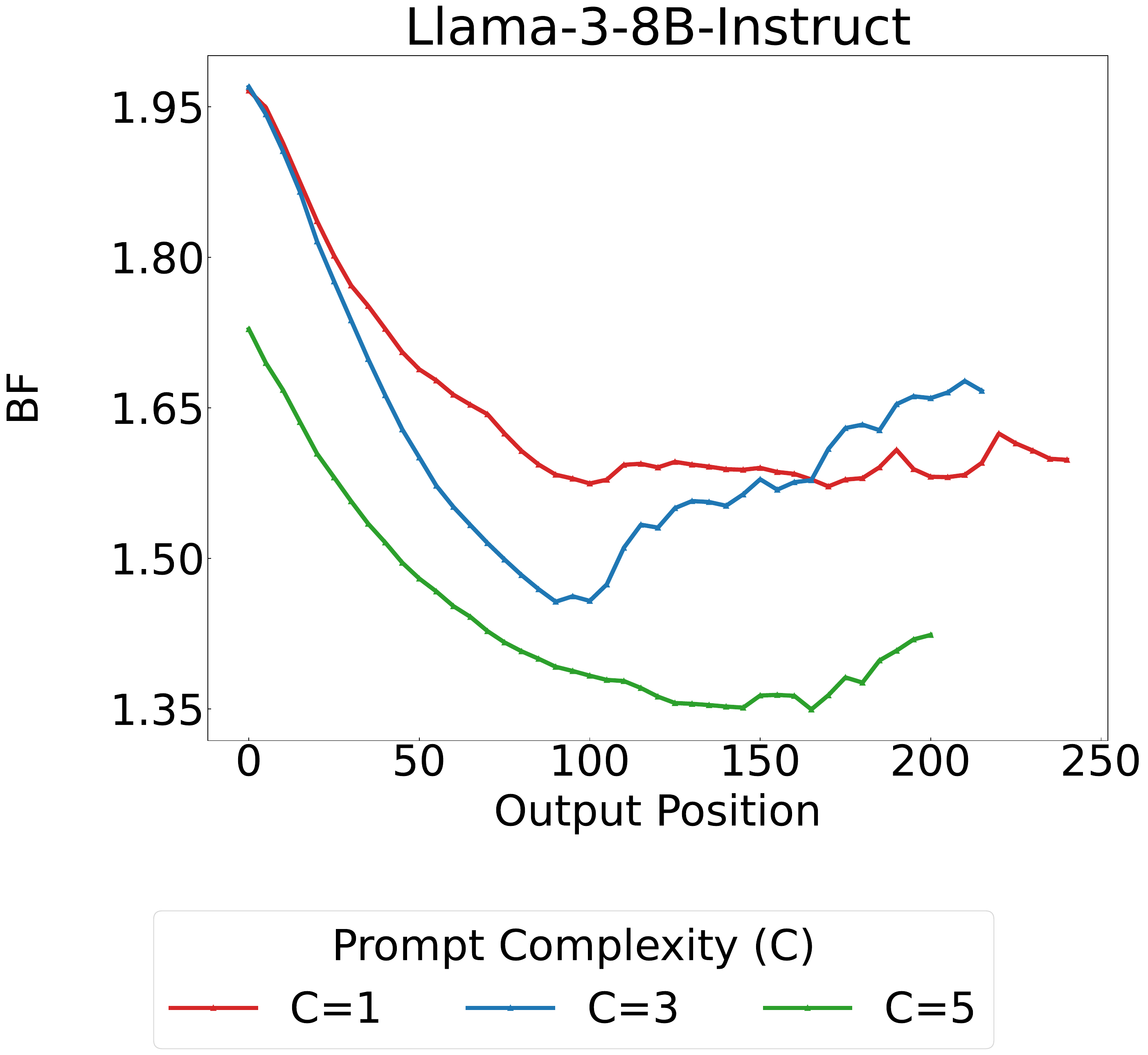}
    \label{fig:output_dynamic_base_xsum_8b_instruct_app}
    \caption{XSUM}
    \end{subfigure}
        \begin{subfigure}[t]{0.24\textwidth}
    \centering
     \includegraphics[width=0.9\linewidth]{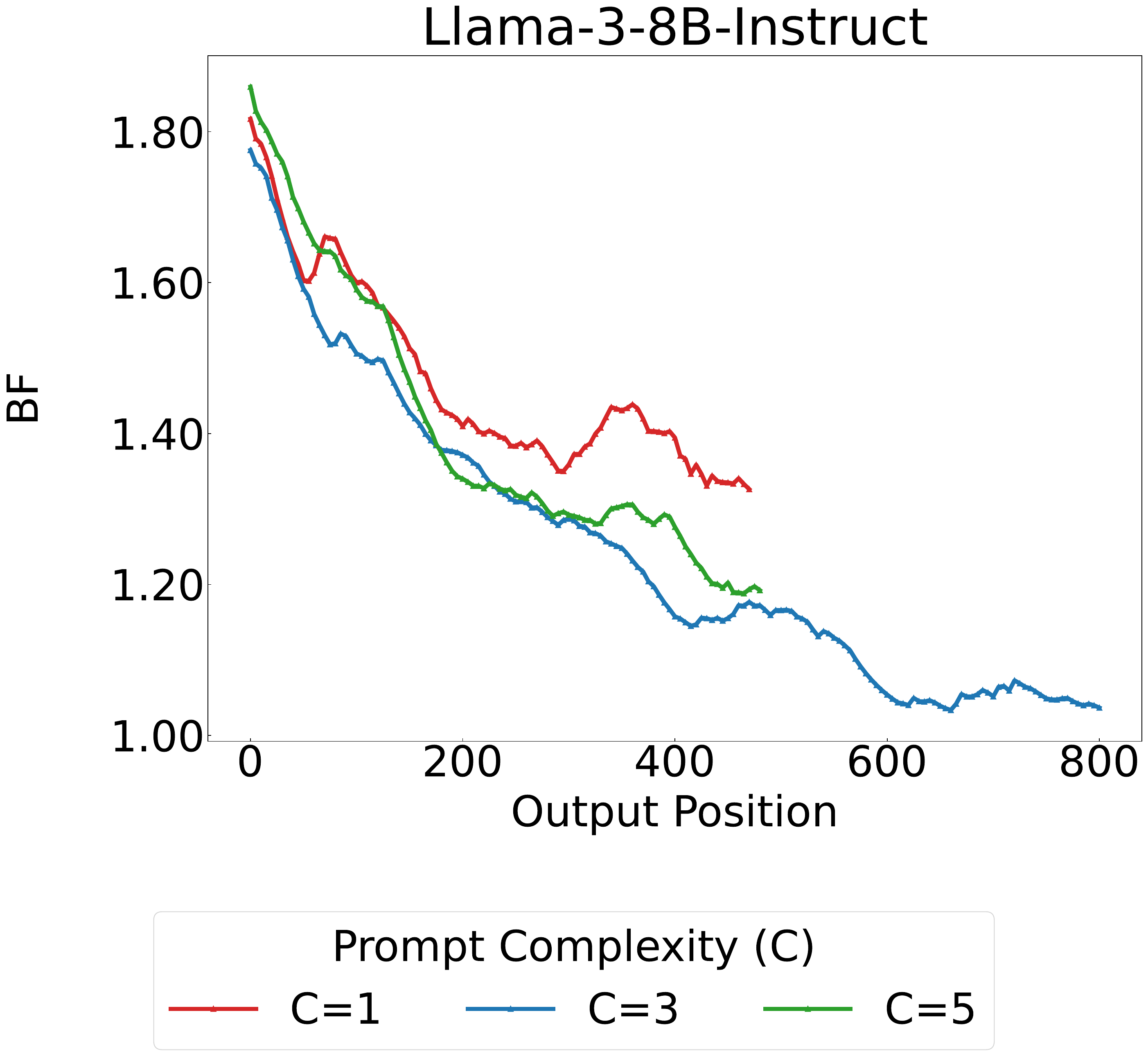}

    \label{fig:output_dynamic_base_aya_8b_instruct_app}
    \caption{Aya}
    \end{subfigure}
    \begin{subfigure}[t]{0.24\textwidth}
    \centering
     \includegraphics[width=0.9\linewidth]{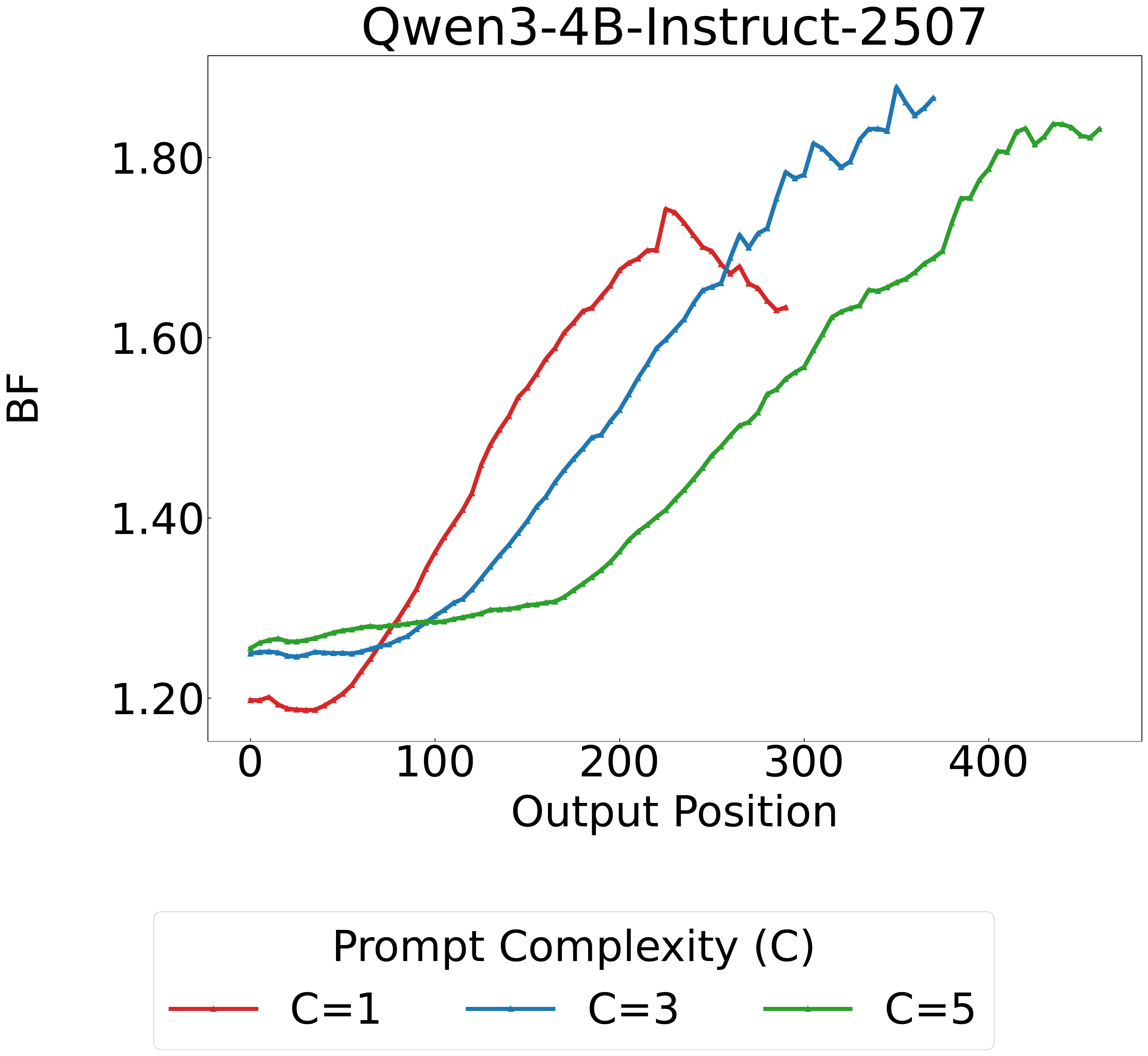}
     \label{fig:output_dynamic_xsum_qwen3_4b_instruct_app}\caption{XSUM}
    \end{subfigure}
        \begin{subfigure}[t]{0.24\textwidth}
    \centering
     \includegraphics[width=0.9\linewidth]{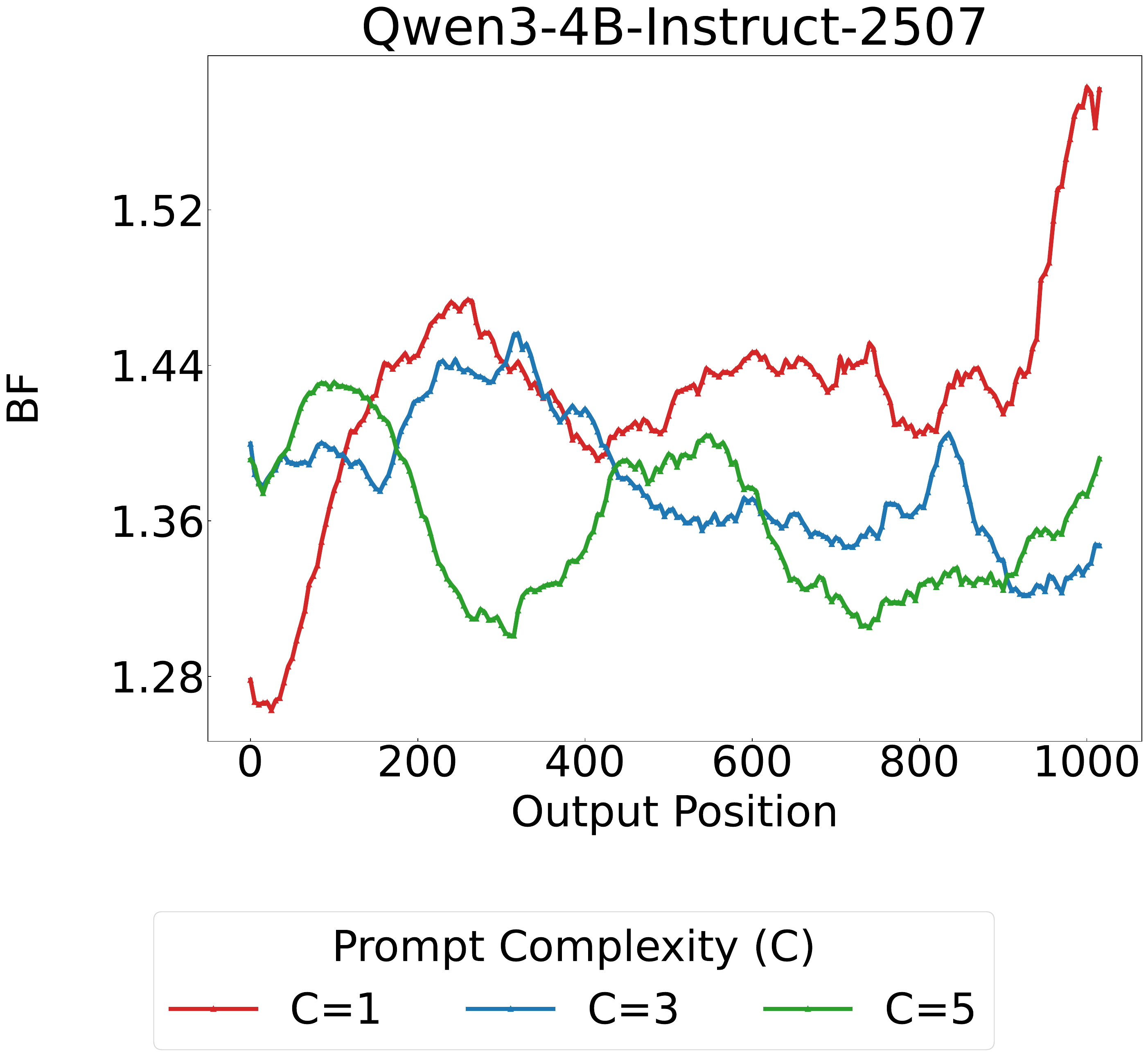}
     \label{fig:output_dynamic_aya_qwen3_4b_instruct_app}\caption{Aya}
    \end{subfigure}
    \begin{subfigure}[t]{0.24\textwidth}
    \centering
     \includegraphics[width=0.9\linewidth]{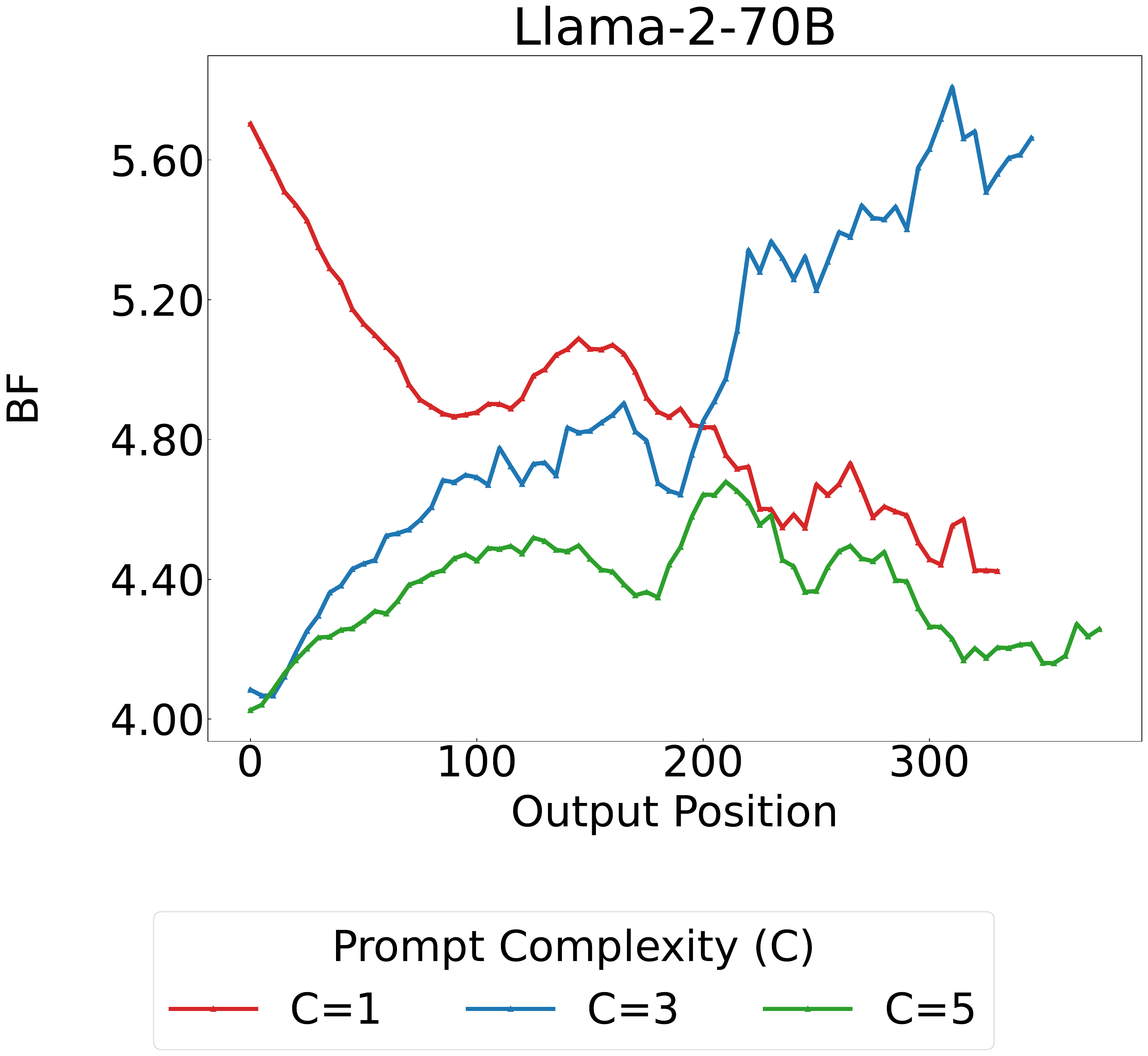}
     \label{fig:output_dynamic_base_llama2_xsum_app}
     \caption{XSUM}
    \end{subfigure}
        \begin{subfigure}[t]{0.24\textwidth}
    \centering
     \includegraphics[width=0.9\linewidth]{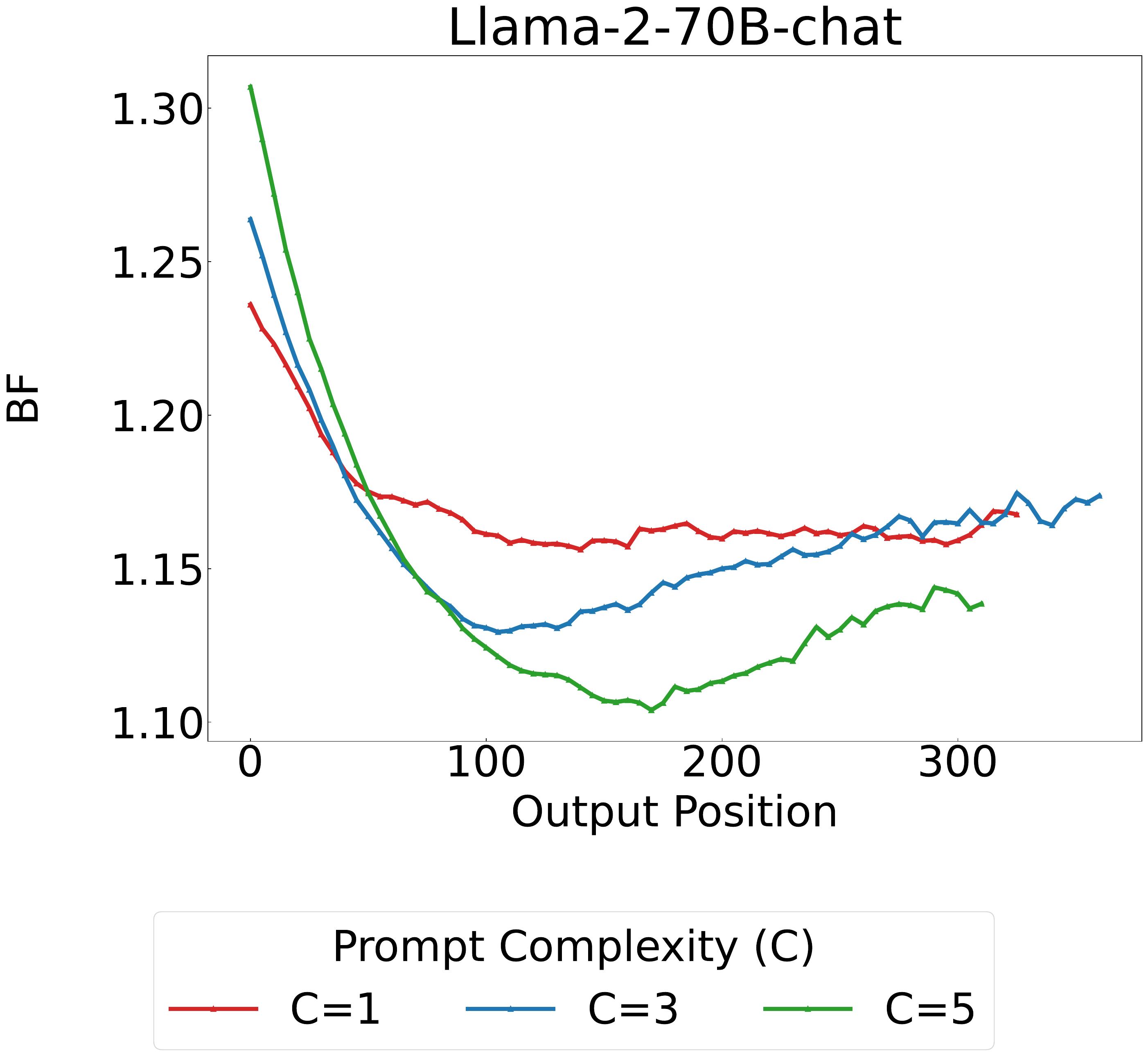}
     \label{fig:output_dynamic_instruct_llama2_xsum_app}
     \caption{XSUM}
    \end{subfigure}
    \begin{subfigure}[t]{0.24\textwidth}
    \centering
     \includegraphics[width=0.9\linewidth]{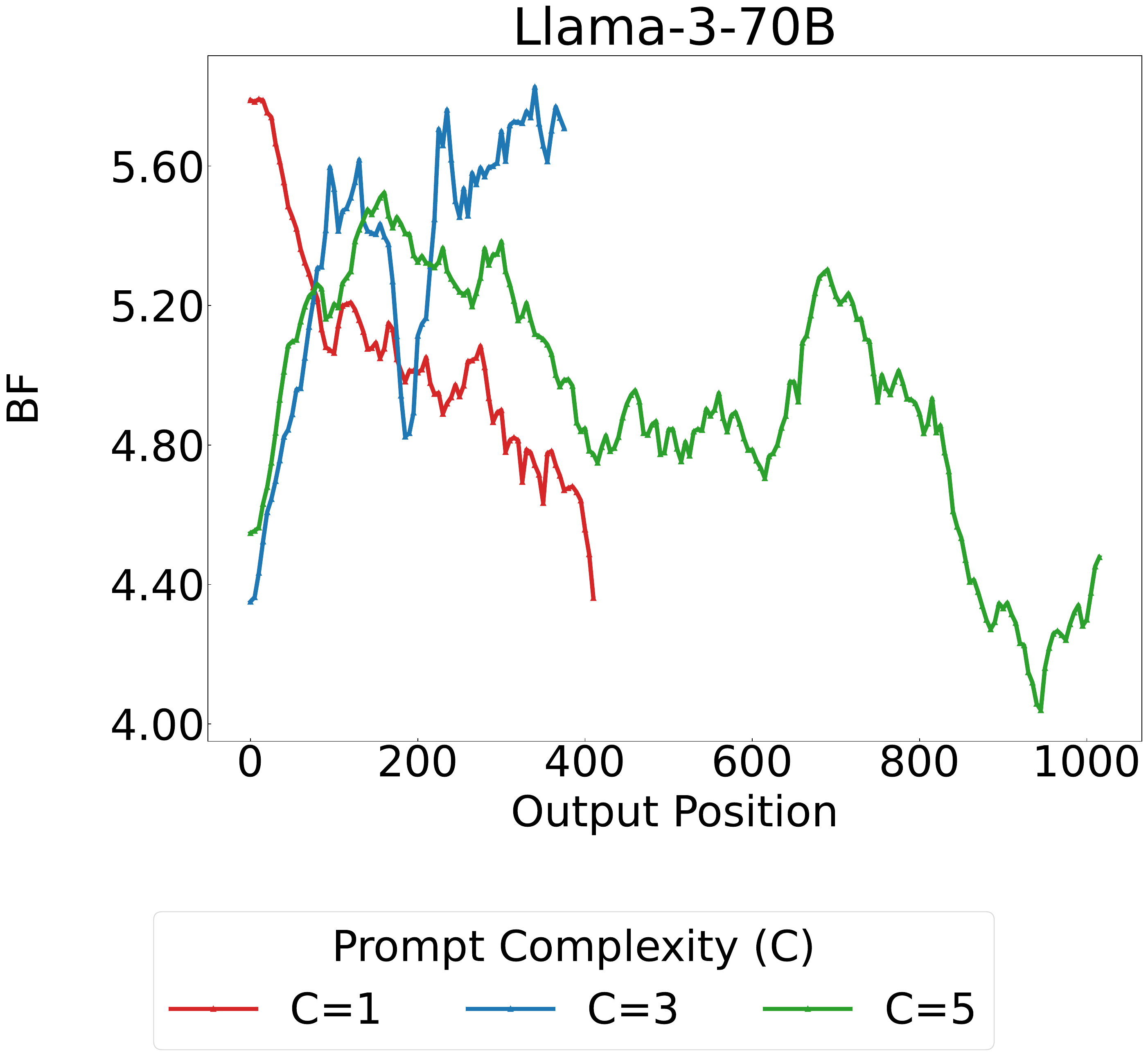}
     \label{fig:output_dynamic_base_xsum_app}
     \caption{XSUM}
    \end{subfigure}
        \begin{subfigure}[t]{0.24\textwidth}
    \centering
     \includegraphics[width=0.9\linewidth]{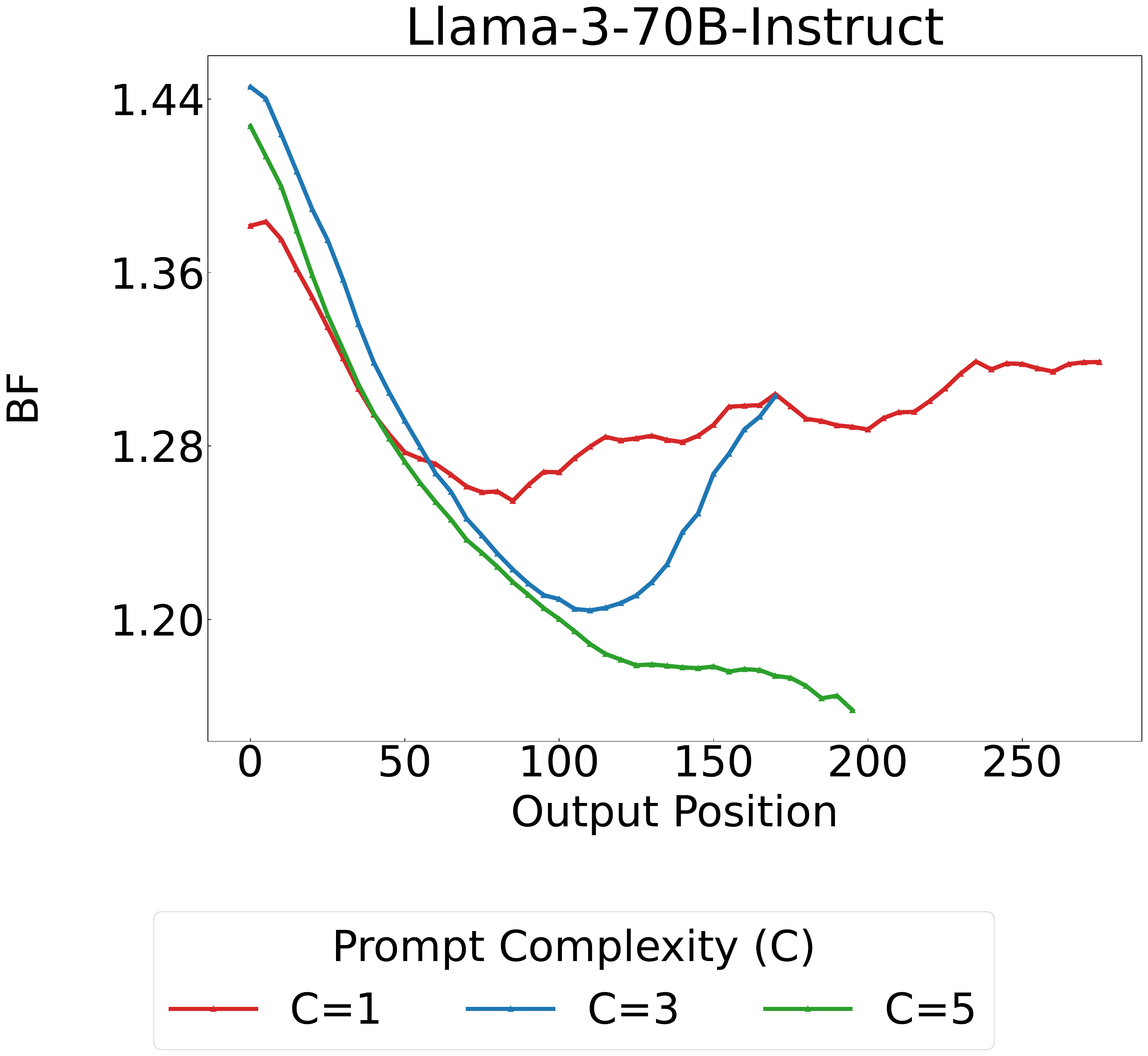}
     \label{fig:output_dynamic_instruct_xsum_app}
     \caption{XSUM}
    \end{subfigure}
    \caption{\textbf{Additional verification of BF dynamics} on XSUM, Aya, and the Qwen3 family. Values are exponential-moving averages of per-step perplexity (smoothing $0.1$).
    }
    \label{fig: output_dynamic_app_additional}
\end{figure}

To confirm the generalizability of our findings (\cref{sec: bf_measure}), we extend our experiments to new domains: summarization on \textsc{XSUM}~\citep{narayan2018don}, multilingual tasks on \textsc{Aya}~\citep{singh2024aya}. We formulate prompt complexity $C$ as providing $C \times 25$ words in the prompt. We also verify our findings on a new model, Qwen3-4B~\citep{qwen3technicalreport}.\footnote{For the Qwen3 family, we use the Qwen3-4B-Base and Qwen3-4B-Instruct-2507 pair. Other aligned variants can be activated into a reasoning mode, exhibiting behavior distinct from the models in our main study, and were thus excluded for a fair comparison.} As presented in \cref{fig: output_dynamic_app_additional}, our core conclusions remain robust across these diverse conditions.

\begin{figure}[htbp]
\centering
    \begin{subfigure}[t]{0.24\textwidth}
    \centering
     \includegraphics[width=0.9\linewidth]{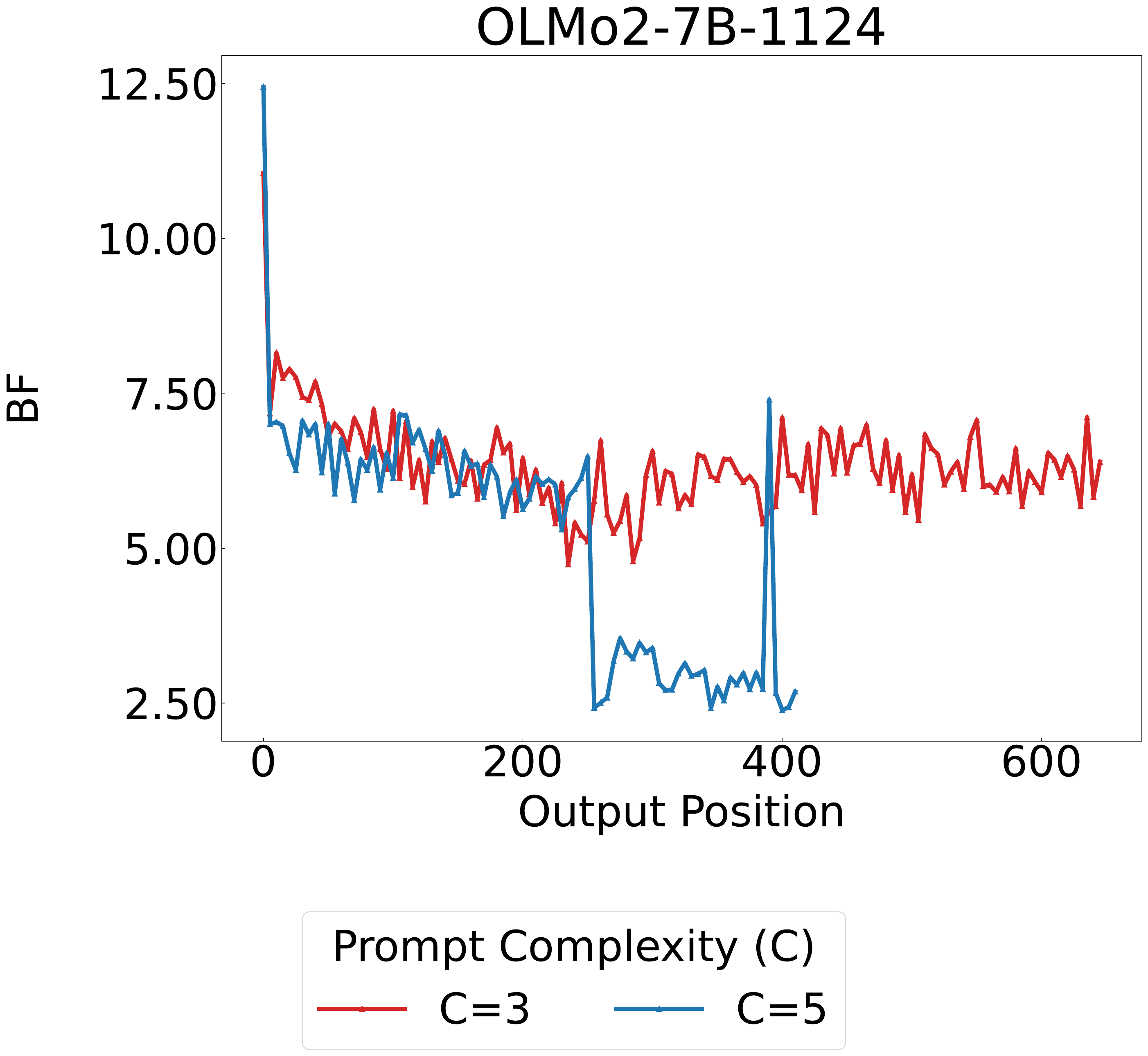}
     \caption{Base 7B (StoryGen)}
    \end{subfigure}
    \begin{subfigure}[t]{0.24\textwidth}
    \centering
     \includegraphics[width=0.9\linewidth]{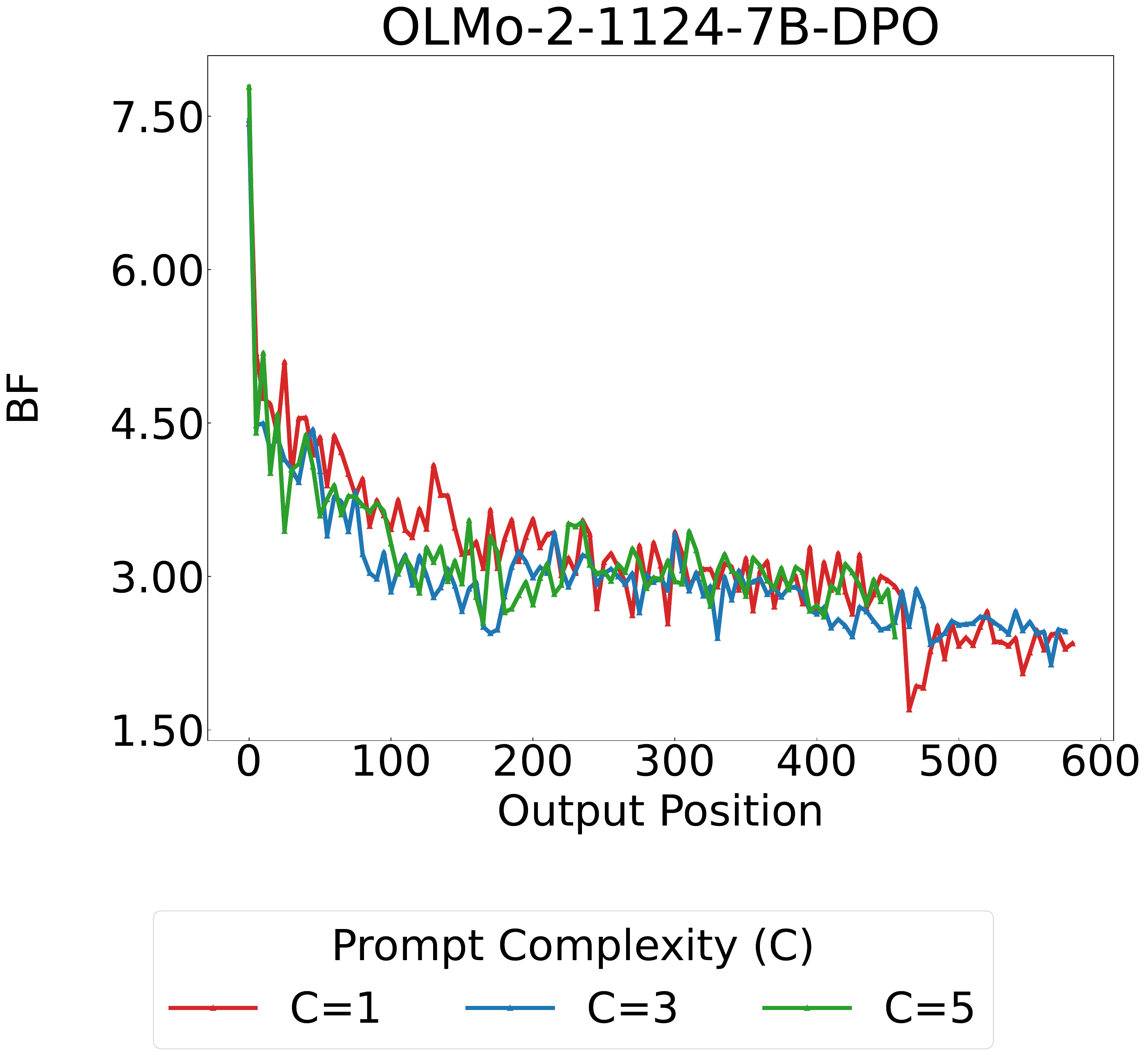}
     \caption{DPO 7B (StoryGen)}
    \end{subfigure}
    \begin{subfigure}[t]{0.24\textwidth}
    \centering
     \includegraphics[width=0.9\linewidth]{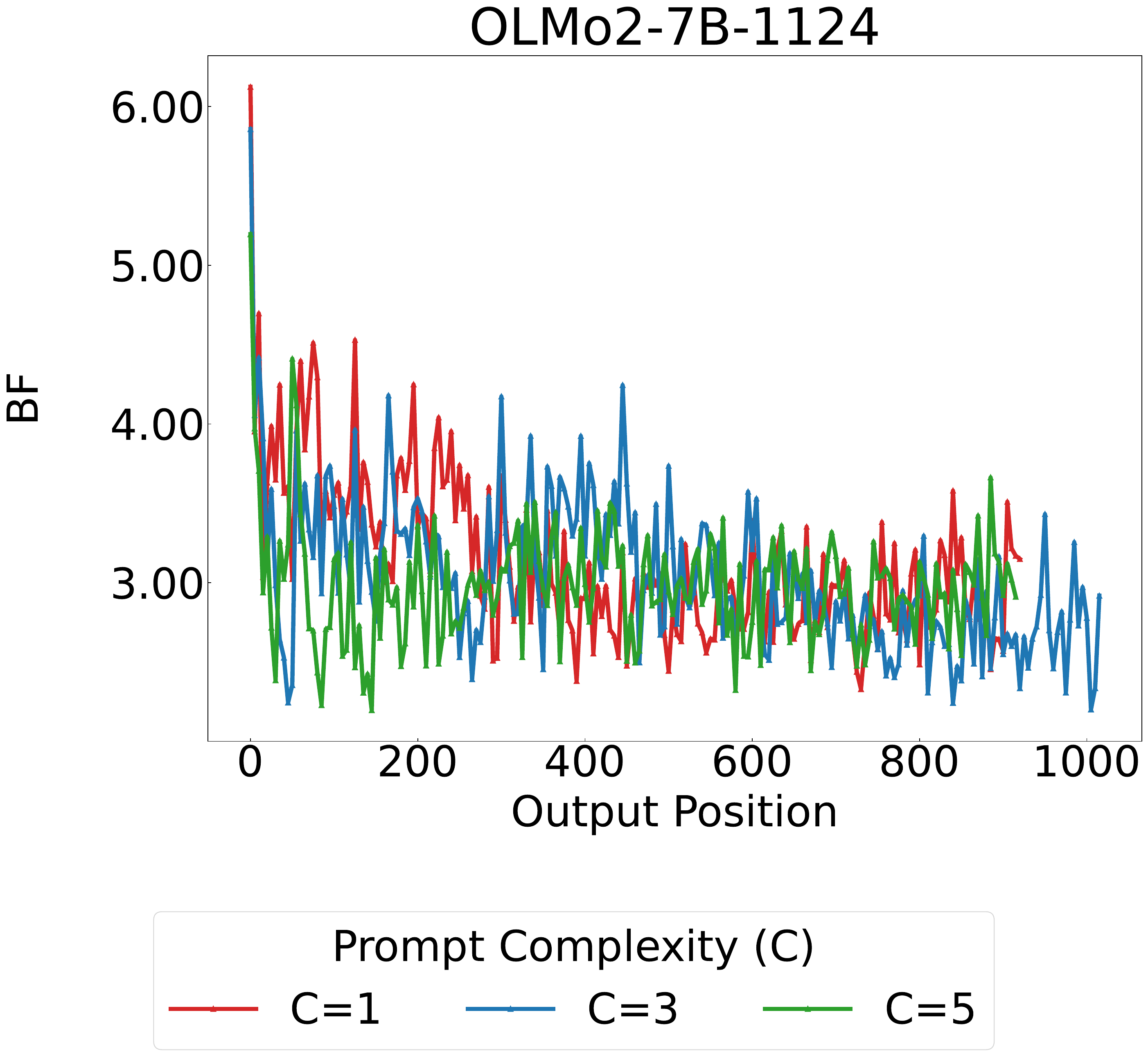}
     \caption{Base 7B (MMLU)}
    \end{subfigure}
    \begin{subfigure}[t]{0.24\textwidth}
    \centering
     \includegraphics[width=0.9\linewidth]{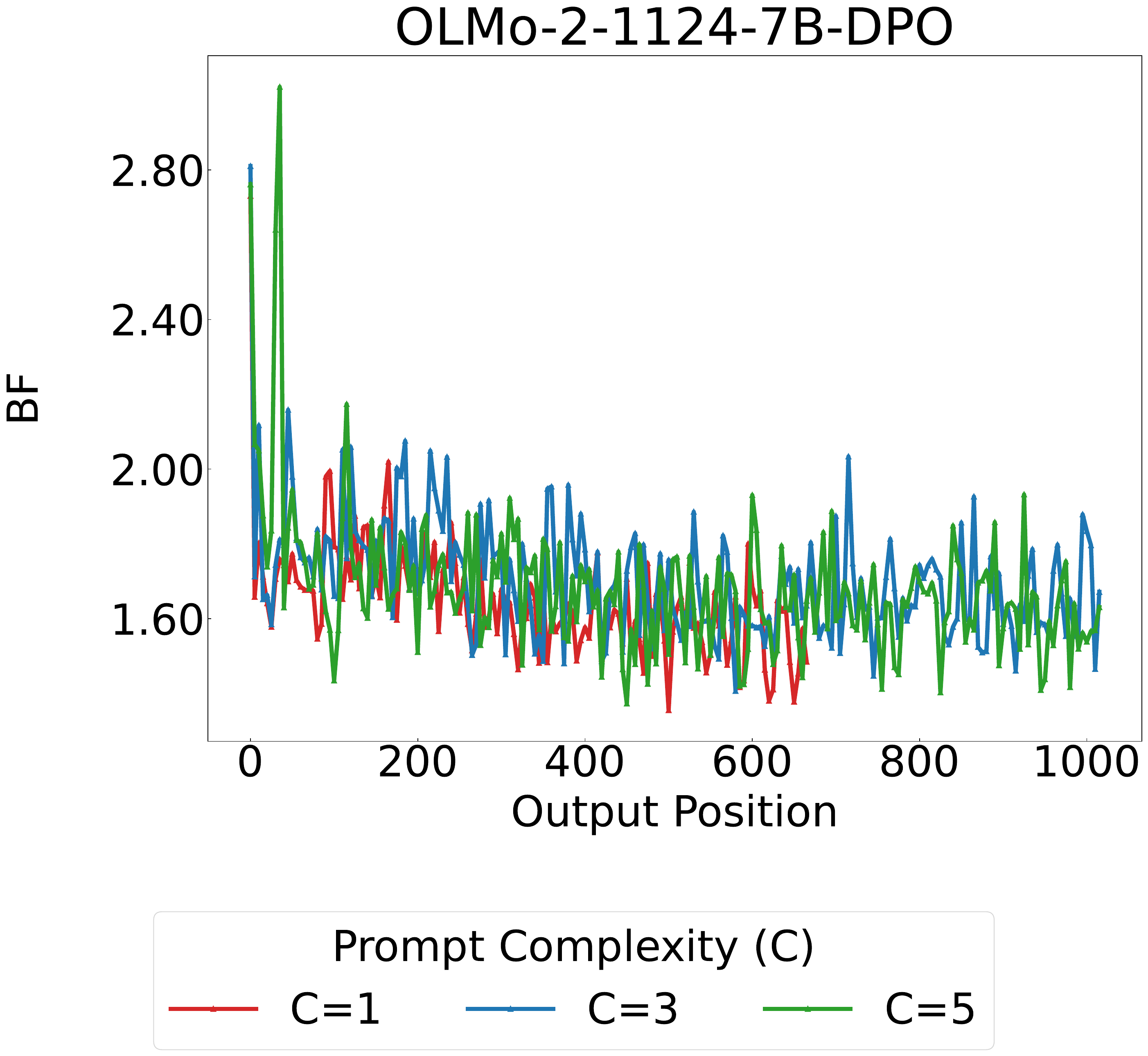}
     \caption{DPO 7B (MMLU)}
    \end{subfigure}

    \begin{subfigure}[t]{0.24\textwidth}
    \centering
     \includegraphics[width=0.9\linewidth]{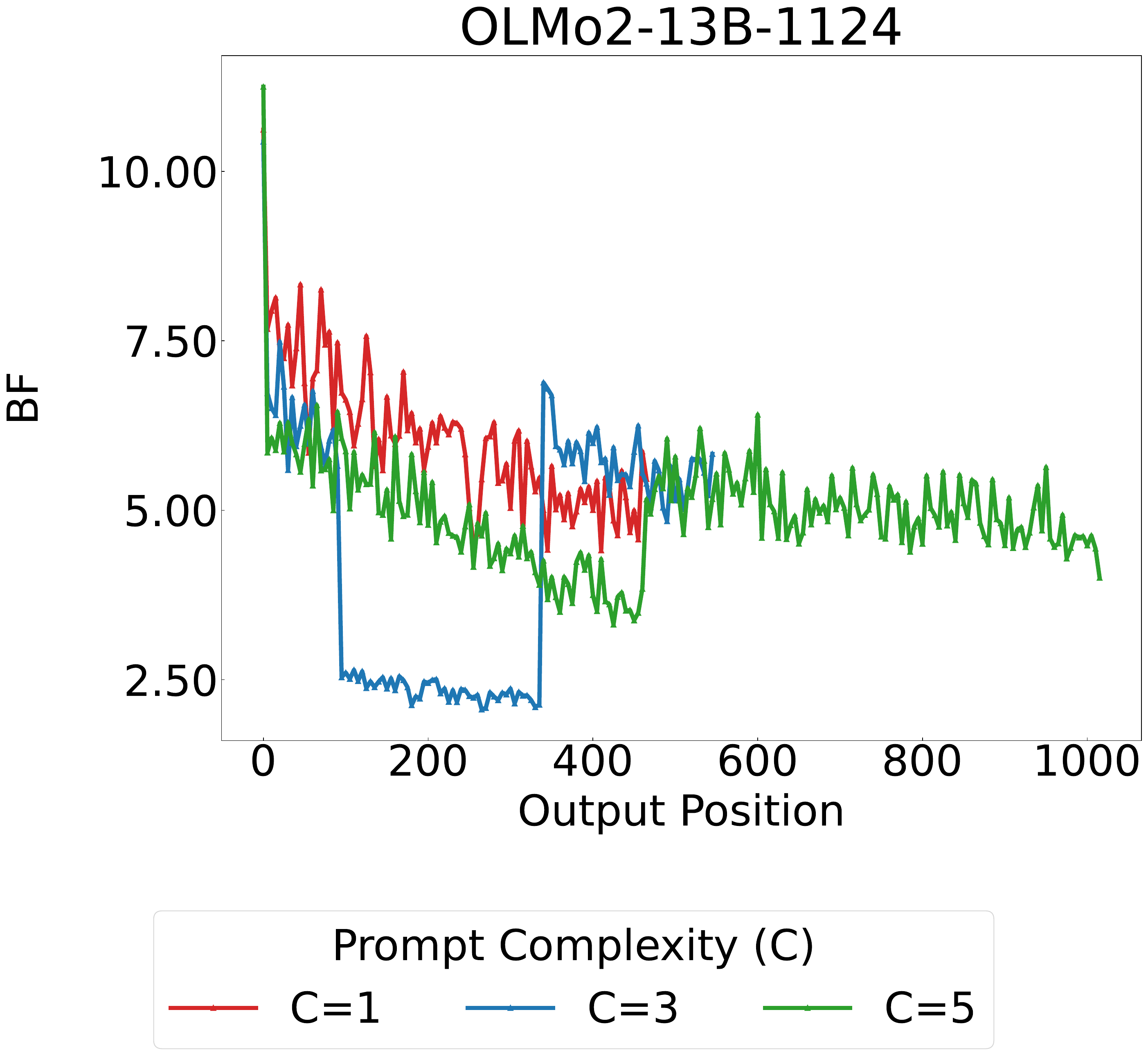}
     \caption{Base 13B (StoryGen)}
    \end{subfigure}
    \begin{subfigure}[t]{0.24\textwidth}
    \centering
     \includegraphics[width=0.9\linewidth]{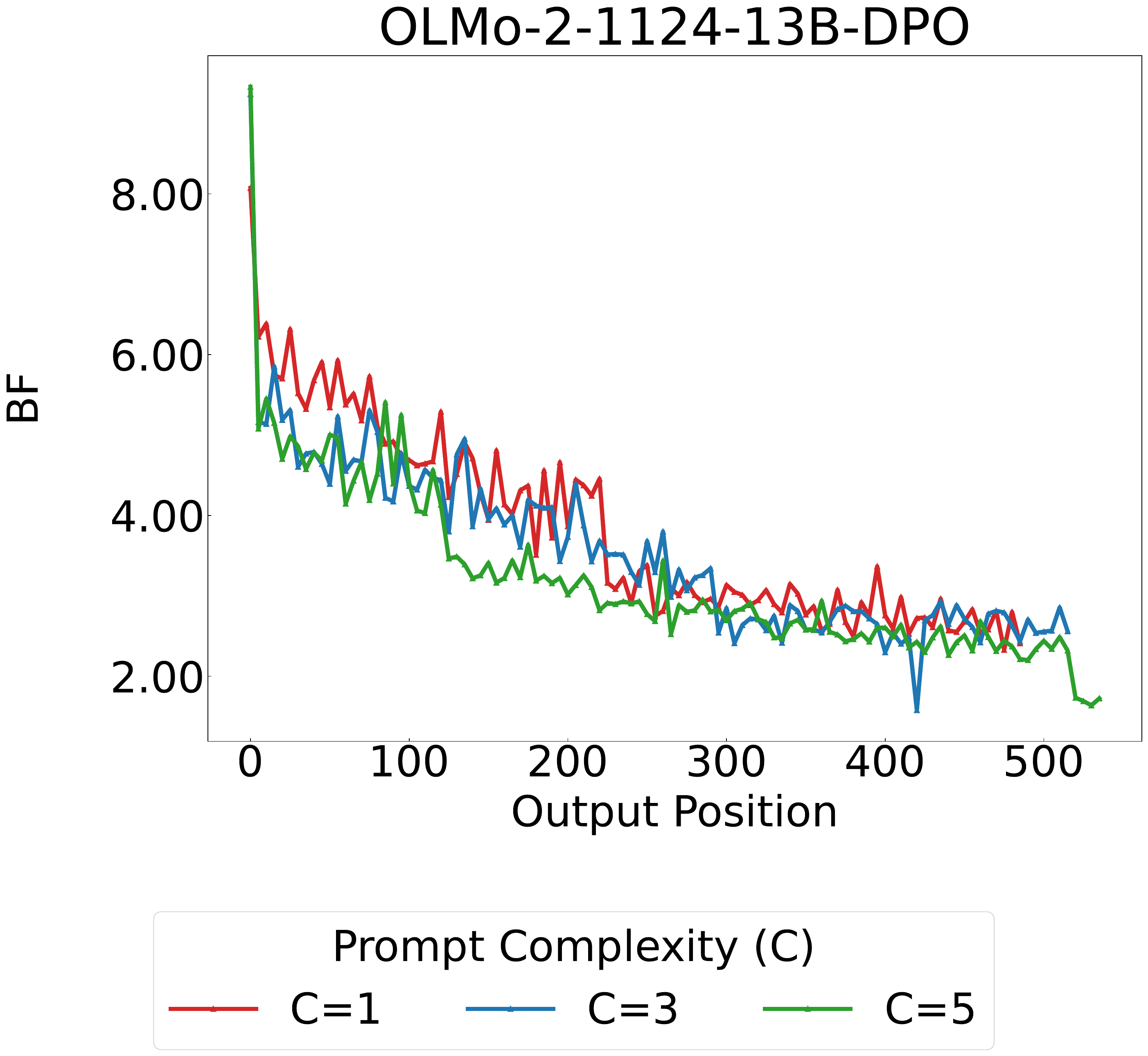}
     \caption{DPO 13B (StoryGen)}
    \end{subfigure}
    \begin{subfigure}[t]{0.24\textwidth}
    \centering
     \includegraphics[width=0.9\linewidth]{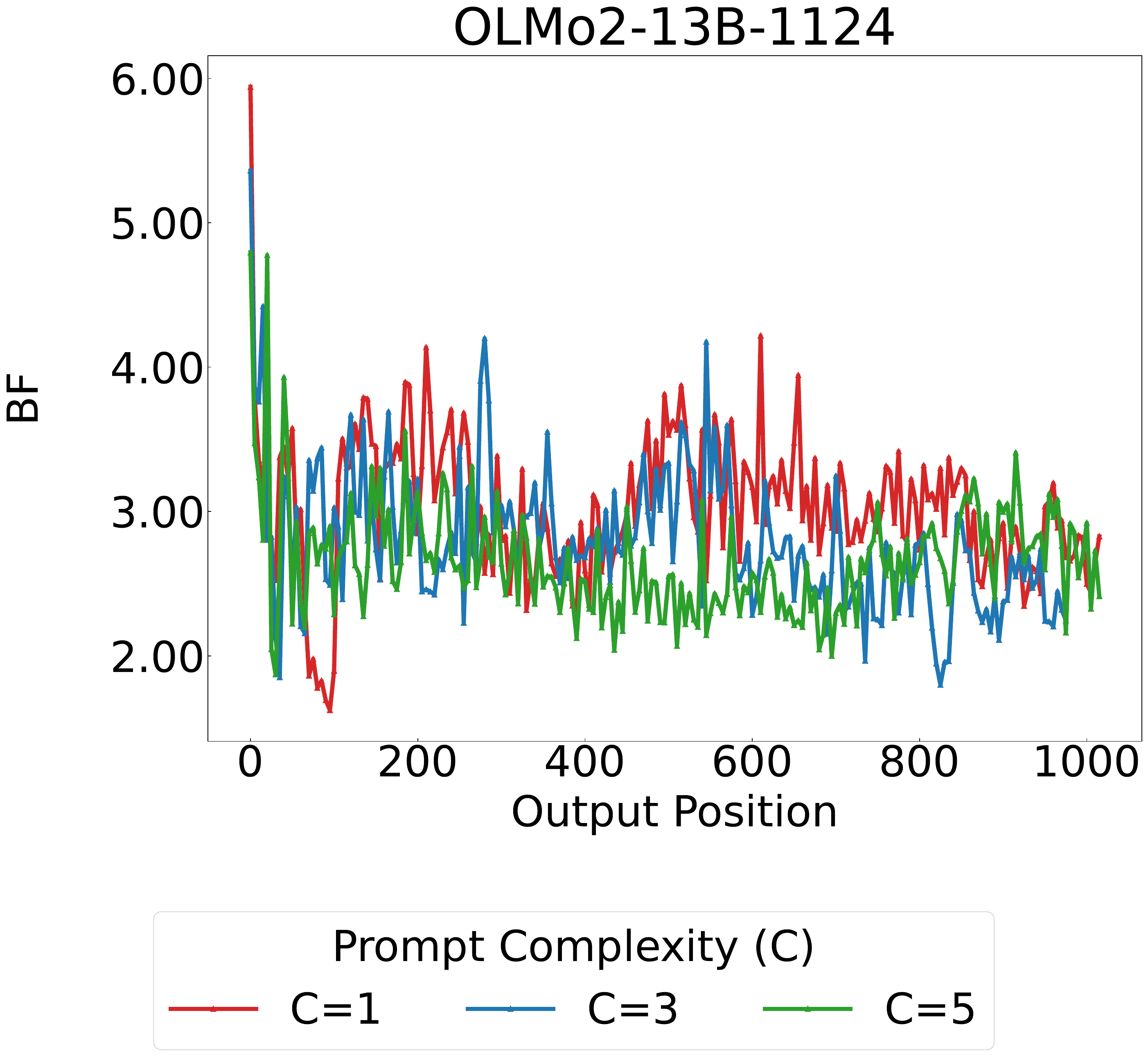}
     \caption{Base 13B (MMLU)}
    \end{subfigure}
    \begin{subfigure}[t]{0.24\textwidth}
    \centering
     \includegraphics[width=0.9\linewidth]{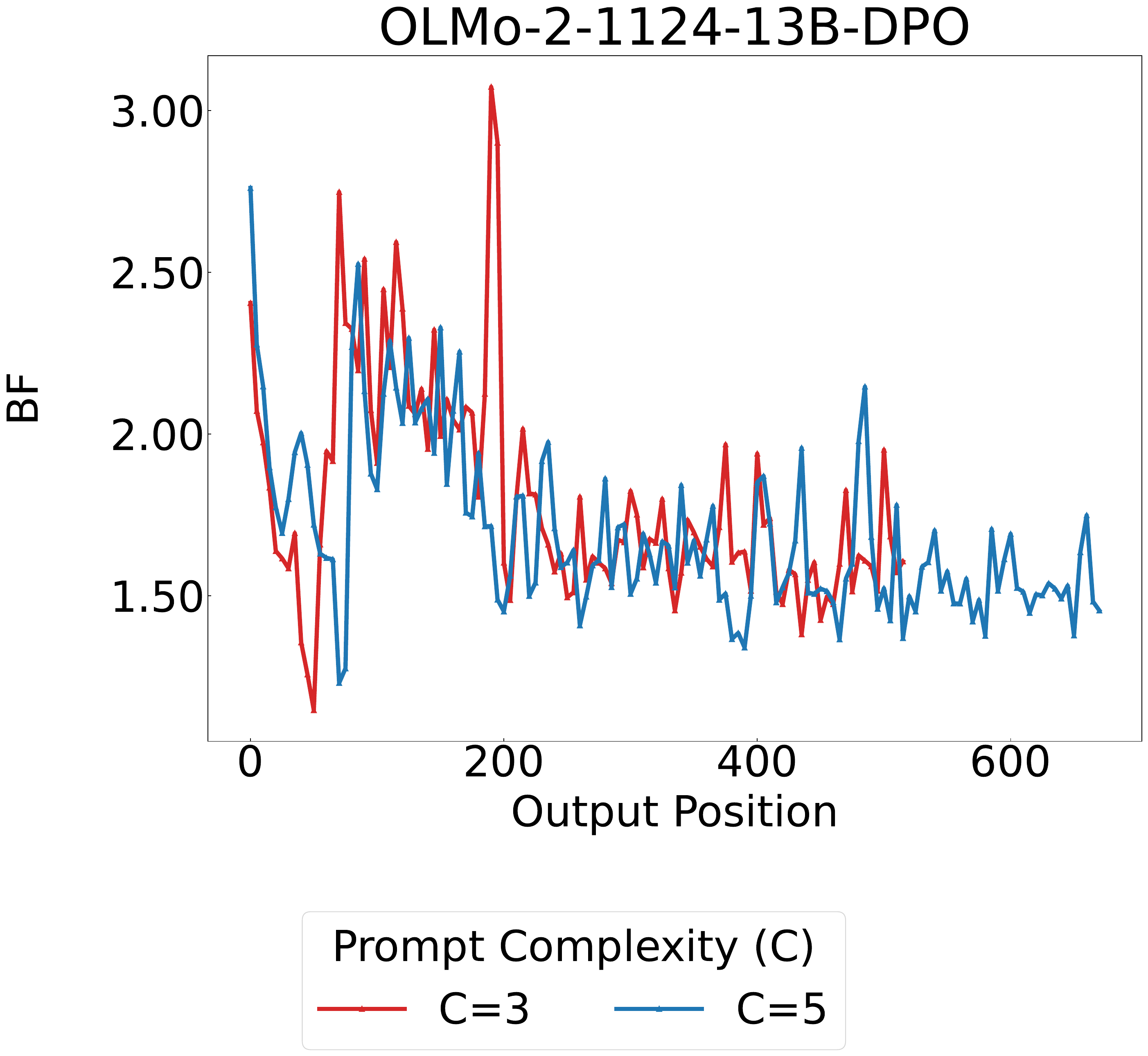}
     \caption{DPO 13B (MMLU)}
    \end{subfigure}

    \caption{\textbf{BF Output Dynamic for OLMo-2 (7B \& 13B) across alignment stages.} We compare Base and DPO models on Creative StoryGen and MMLU. Note that we treat DPO as the aligned model for OLMo-2, following the convention in \citep{olmo20242}.
    }
    \label{fig:olmo2_output_dynamic}
\end{figure}

\begin{figure}[htbp]
\centering
    \begin{subfigure}[t]{0.45\textwidth}
    \centering
     \includegraphics[width=0.9\linewidth]{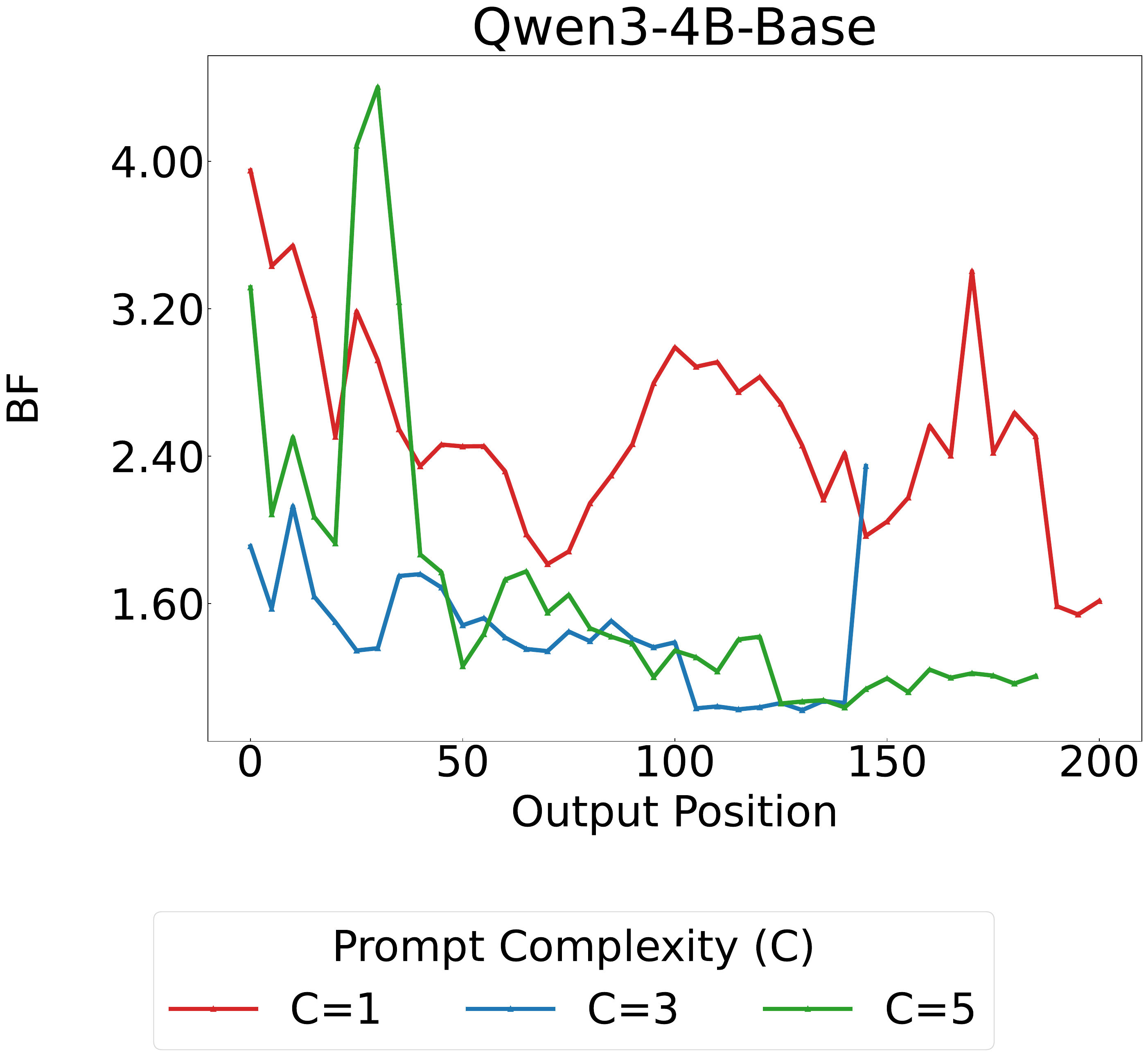}
     \caption{Qwen3-4B-Base (MMLU)}
    \end{subfigure}
    \begin{subfigure}[t]{0.45\textwidth}
    \centering
     \includegraphics[width=0.9\linewidth]{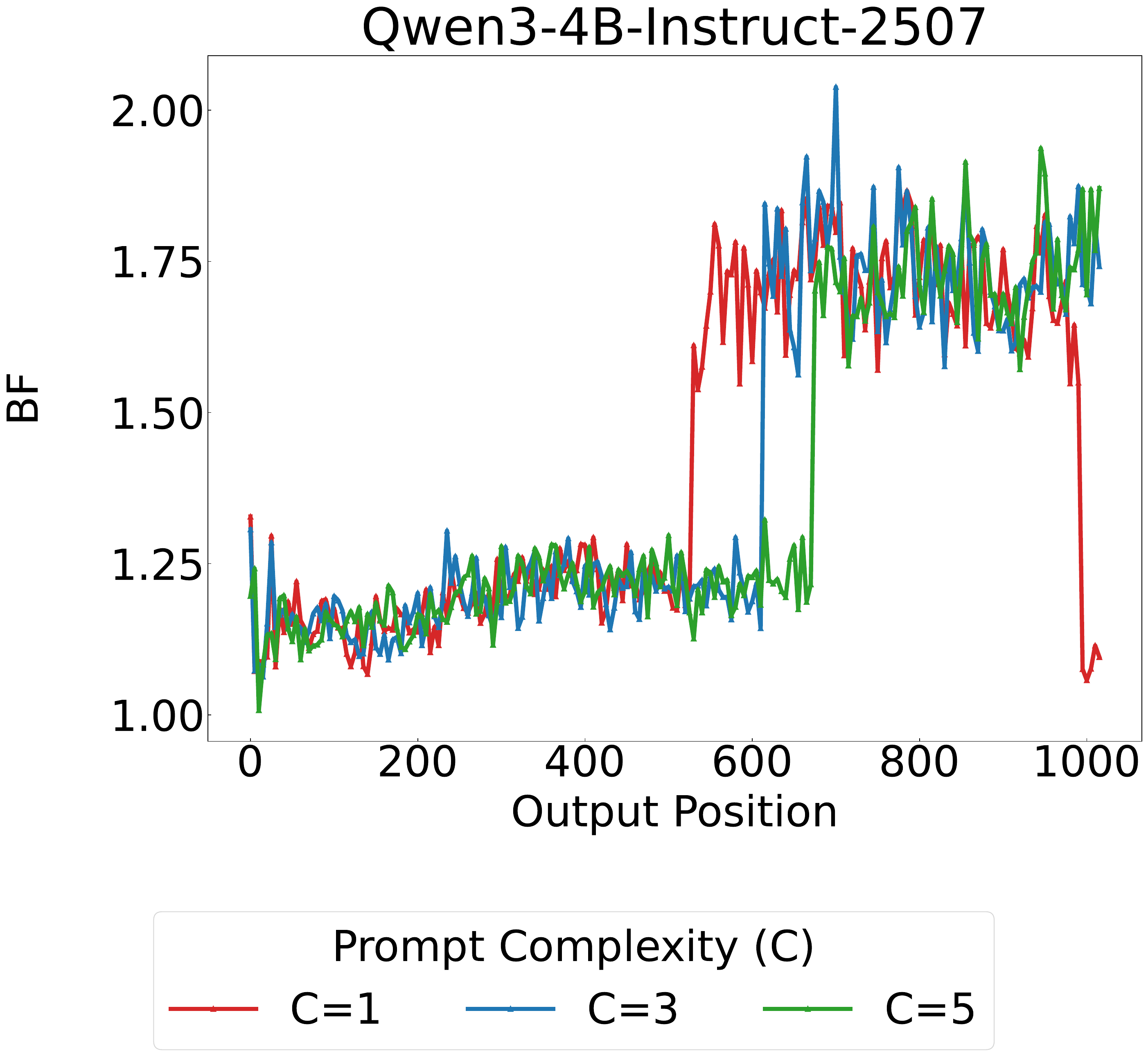}
     \caption{Qwen3-4B-Instruct (MMLU)}
    \end{subfigure}
    \caption{\textbf{BF Output Dynamic for Qwen3-4B on MMLU.}
    }
    \label{fig:qwen3_mmlu_output_dynamic}
\end{figure}

We also analyze OLMo-2~\citep{olmo20242} across different alignment stages (Base and DPO) on Creative StoryGen and MMLU, covering both 7B and 13B scales. As shown in \cref{fig:olmo2_output_dynamic}, OLMo-2 exhibits a similar trend where alignment tuning reduces BF, although the reduction is less pronounced compared to Llama models, suggesting that OLMo-2's post-training asserts less influence on the generation manifold. Additionally, we provide the BF dynamics for Qwen3-4B on MMLU in \cref{fig:qwen3_mmlu_output_dynamic}.

\begin{figure}[htbp]
\centering
    \begin{subfigure}[t]{0.42\textwidth}
    \centering
     \includegraphics[width=0.92\linewidth]{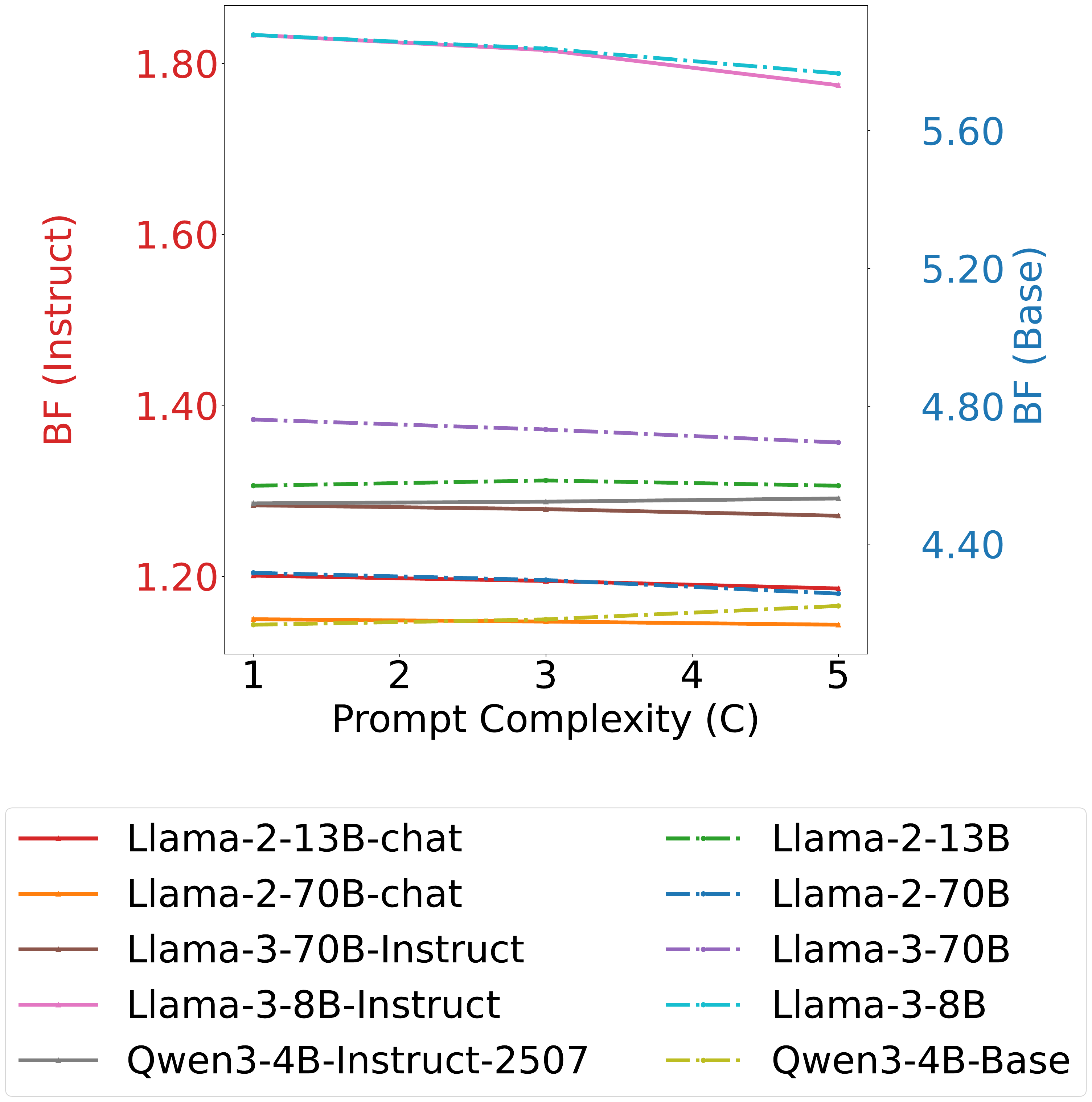}
    \caption{XSUM}
     \label{fig:xsum_ppl_p}
    \end{subfigure}
    \begin{subfigure}[t]{0.42\textwidth}
    \centering
     \includegraphics[width=0.92\linewidth]{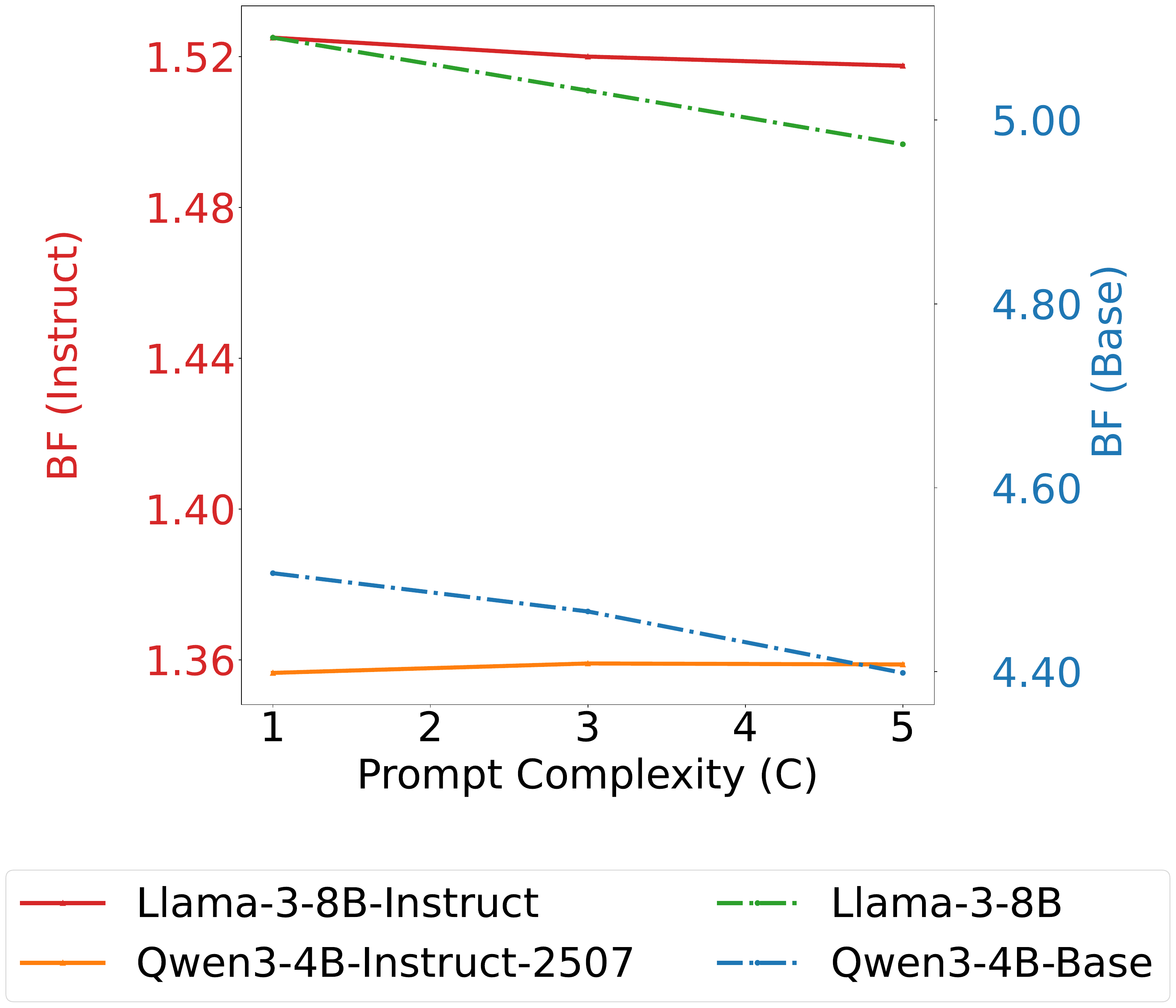}
    \caption{Aya}
     \label{fig:aya_ppl_p}
    \end{subfigure}
    \caption{\textbf{Prompt-complexity comparison} on XSUM and Aya, complementing \cref{fig: output_dynamic_app_additional}.}
\end{figure}

\section{Proof of LLM Log-Likelihood Convergence}
\label{app: aep_proof}
The following proof is a simplified version of the one in \citep{mudireddy2024slaves}, presented for completeness and to refine its original bounds. For the formal measure-theoretical treatment, we refer readers to the original paper. While a more direct proof using the weak law of large numbers is possible, we use Chebyshev's inequality to provide a more self-contained and accessible argument.

The key observation here is that under current computation architecture, the probability implemented by transformers are log-precision~\citep{merrill2023parallelism}, and thus $|\log {P}\left(\outputval_{1: N} | \inputval; \theta \right)|$ is bounded (e.g., $|\log {P}\left(\outputval_{1: N} | \inputval; \theta \right)| \leq M$). For the truncated probability $\tilde{P}\left(\outputval_{1: N} | \inputval; \theta \right)$, we can essentially only consider the non-zero probability over the truncated vocabulary, and the same bound holds. Depending on the quantization scheme implemented, examples of $M$ include $32, 64$, etc.

\revise{
We define the length-averaged \textit{realized entropy} for a specific sequence $\outputval_{1:N}$ as:
\begin{equation}
    h_{\text{realized}}(\outputval_{1:N}) \defeq \frac{1}{N}\sum_{t=1}^N H(\outputVar_t | [\inputval, \outputval_{<t}]; \theta)
\end{equation}
where $H(\outputVar_t | [\inputval, \outputval_{<t}]; \theta) = -\sum_{\outputval \in V} \tilde P(\outputval | [\inputval, \outputval_{<t}]; \theta) \log \tilde P(\outputval | [\inputval, \outputval_{<t}]; \theta)$.

We aim to bound the probability that the NLL deviates from this realized entropy. Let $\Delta_N$ be the total difference between the log-probability and the realized entropy sum:
\begin{equation}
    \Delta_N = \left( -\log \tilde P(\outputval_{1: N} | \inputval; \theta) \right) - \sum_{t=1}^N H(\outputVar_t | [\inputval, \outputval_{<t}]; \theta) = \sum_{t=1}^N Z_t
\end{equation}
where we define the single-step deviation variable $Z_t$ as:
\begin{equation}
    Z_t \defeq -\log \tilde P(\outputval_t | [\inputval, \outputval_{<t}]; \theta) - H(\outputVar_t | [\inputval, \outputval_{<t}]; \theta)
\end{equation}
Note that $\Delta_N$ is a random variable formed by the sum of $Z_t$. To use Chebyshev's inequality, we need to calculate the variance of this sum:
\begin{equation}
    \text{Var}(\Delta_N) = \text{Var}\left(\sum_{t=1}^N Z_t\right) = \sum_{t=1}^N \text{Var}(Z_t) + \sum_{i \neq j} \text{Cov}(Z_i, Z_j)
\end{equation}
We now show that the covariance terms $\text{Cov}(Z_i, Z_j)$ are zero for all $i < j$. By definition, $\text{Cov}(Z_i, Z_j) = \E[Z_i Z_j] - \E[Z_i]\E[Z_j]$.

First, observe that the expected value of the deviation $Z_j$ at any step, conditioned on the prompt and generated history, is zero:
\begin{align} \label{eq:zero_mean}
    \E[Z_j | [\inputval, \outputval_{<j}]; \theta] &= \E_{\outputval_j}\left[-\log \tilde P(\outputval_j| [\inputval, \outputval_{<j}]; \theta)\right] - H(\outputVar_j | [\inputval, \outputval_{<j}]; \theta) \nonumber \\
    &= H(\outputVar_j | [\inputval, \outputval_{<j}]; \theta) - H(\outputVar_j | [\inputval, \outputval_{<j}]; \theta) = 0
\end{align}
This implies $\E[Z_j] = 0$ for all $j$. Thus, $\text{Cov}(Z_i, Z_j) = \E[Z_i Z_j]$.

For $i < j$, the value of $Z_i$ is fully determined by the history $[\inputval, \outputval_{<j}]$. We use the Law of Iterated Expectations, conditioning on the history up to step $j$:
\begin{align}
    \E[Z_i Z_j] &= \E_{[\inputval, \outputval_{<j}]} \left[ \E[Z_i Z_j | [\inputval, \outputval_{<j}]; \theta] \right] \\
    &= \E_{[\inputval, \outputval_{<j}]} \left[ Z_i \cdot \E[Z_j | [\inputval, \outputval_{<j}]; \theta] \right] \quad \text{(since $Z_i$ is determined given $\outputval_{<j}$)} \\
    &= \E_{[\inputval, \outputval_{<j}]} \left[ Z_i \cdot 0 \right] \quad \text{(by Eq. \ref{eq:zero_mean})} \\
    &= 0
\end{align}
Since all cross-terms are zero, the variance of the sum is simply the sum of the variances:
\begin{equation}
    \text{Var}(\Delta_N) = \sum_{t=1}^N \text{Var}(Z_t)
\end{equation}
Given the observation that transformer probabilities are computed with bounded log-precision~\citep{merrill2023parallelism}, we have $|\log P(\outputval | [\inputval, \outputval_{<t}]; \theta)| \leq M$ (For un-truncated $P$). Consequently, the random variable $Z_t$ is bounded, and its variance is bounded by a constant, denoted $C = (2M)^2$.
\begin{equation}
    \text{Var}(\Delta_N) \leq N \cdot C
\end{equation}
We can now apply Chebyshev's inequality to the length-averaged deviation:
\begin{equation}
    P\left( \left| \frac{\Delta_N}{N} \right| \geq \epsilon \right) \leq \frac{\text{Var}(\Delta_N/N)}{\epsilon^2} = \frac{\frac{1}{N^2}\text{Var}(\Delta_N)}{\epsilon^2} \leq \frac{\frac{1}{N^2} (N \cdot C)}{\epsilon^2} = \frac{C}{N \epsilon^2}
\end{equation}
Taking the limit as $N \rightarrow \infty$, the probability of deviation approaches 0. Thus, we have convergence in probability:
\begin{equation}
    -\frac{1}{N}\log \tilde P(\outputval_{1: N} | \inputval; \theta) - h_{\text{realized}}(\outputval_{1:N}) \xrightarrow{P} 0
\end{equation}
}

\section{Full Nudging Experiment Results}
\label{app: nudging}
Due to space limits, we put the nudging experiment results for MMLU here. Though on MMLU, nudging does not reduce BF that quickly as over Just-Eval-Instruct, it does bring down BF of base models significantly, which verifies our hypothesis in \cref{sec: nudging}.
\begin{figure*}[ht!]
\centering
    \begin{subfigure}[t]{0.45\textwidth}
    \centering
     \includegraphics[width=0.8\linewidth]{img140.pdf}
    \vspace{-0.3cm}
    \caption{Just-Eval-Instruct}
     \label{fig:just_eval_instruct_nudging_app}
    \end{subfigure}
    \begin{subfigure}[t]{0.45\textwidth}
    \centering
     \includegraphics[width=0.8\linewidth]{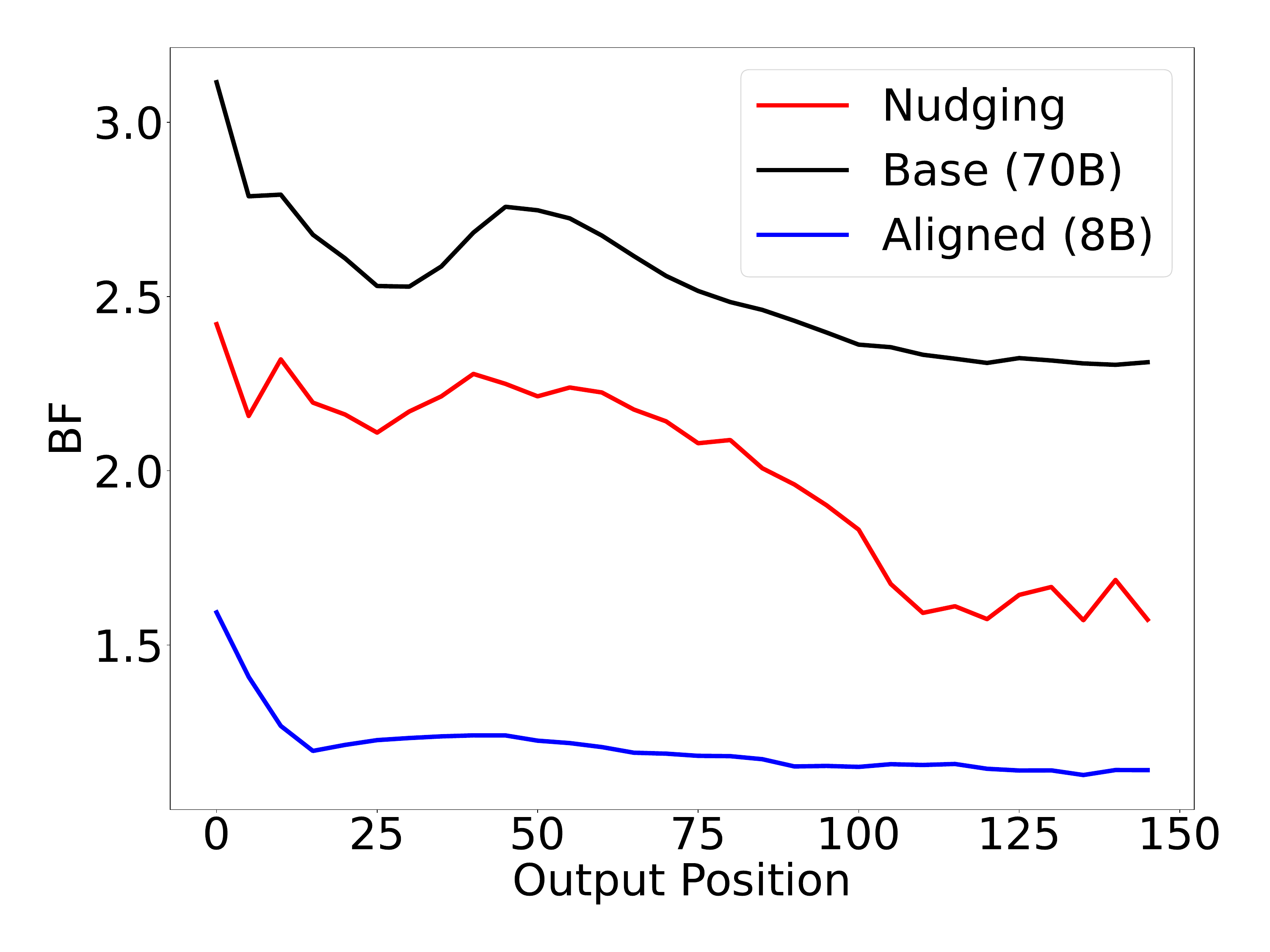}
    \vspace{-0.3cm}
    \caption{MMLU}
     \label{fig:mmlu_nudging_app}
    \end{subfigure}
    \caption{Output Perplexity Dynamics in Nudging Experiments. }
\label{fig:nudging_analysis_app}
\end{figure*}

\begin{figure*}[ht!]
\centering
      \begin{subfigure}[t]{0.45\textwidth}
    \centering
     \includegraphics[width=0.8\linewidth]{img142.pdf}
    \caption{Just-Eval-Instruct }
     \label{fig:just_eval_instruct_nudging_histogram_app}
    \end{subfigure}
       \begin{subfigure}[t]{0.45\textwidth}
    \centering
     \includegraphics[width=0.8\linewidth]{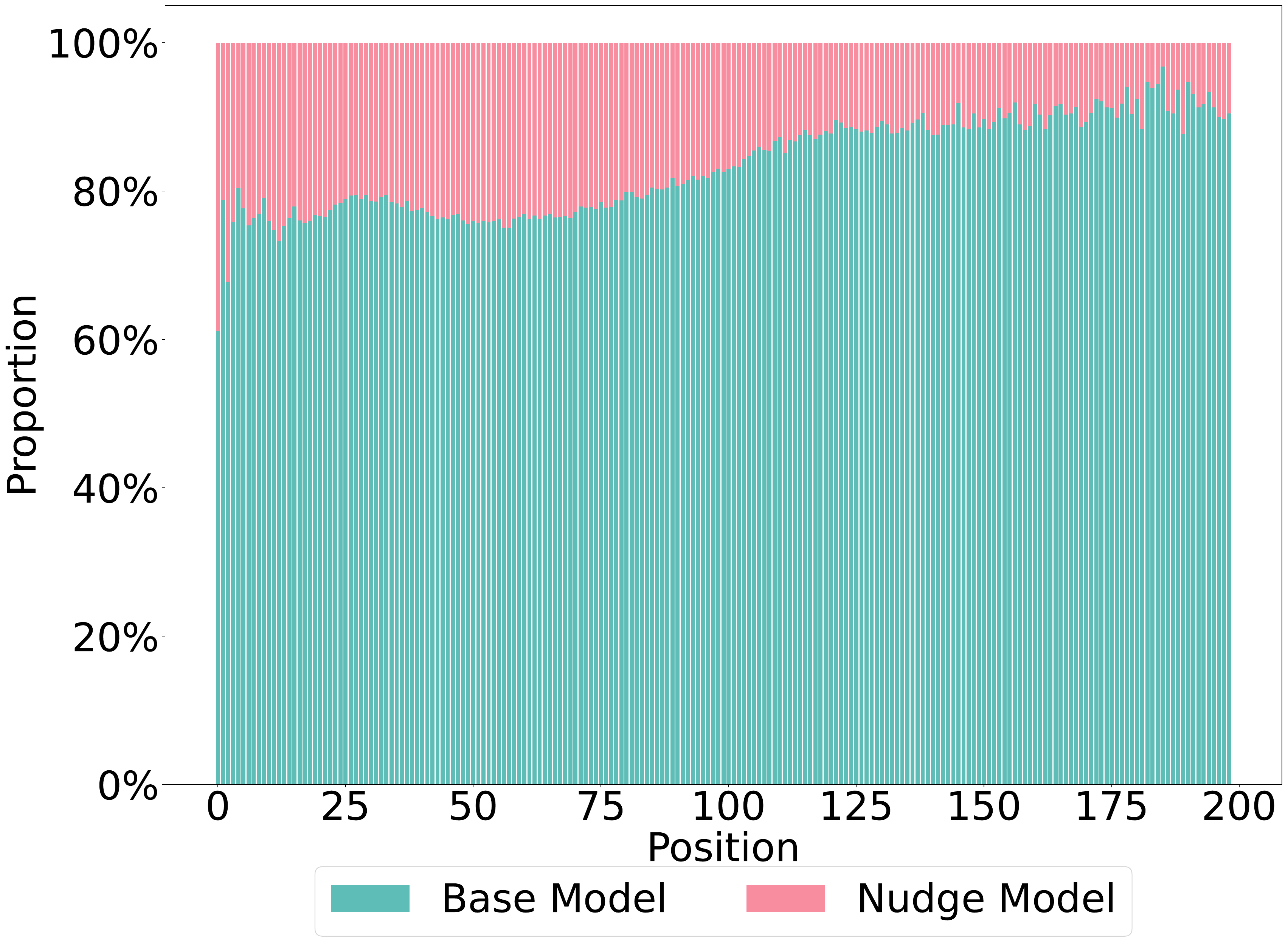}
    \caption{MMLU}
     \label{fig:mmlu_nudging_histogram_app}
    \end{subfigure}
     \caption{Nudging Ratio Histogram.}
     \label{fig:nudging_ratio_app}
\end{figure*}

\section{BF and Information Density} 
\label{app: bf_and_id}
Our BF measure can also be interpreted as capturing the information density that LLMs target to facilitate efficient communication~\citep{genzel2002entropy, jaeger2006speakers, levy2008expectation, mahowald2013info, meister2021revisiting, verma2023revisiting}. Prior work has leveraged both token-level log-probabilities and entropy rates ($\bar{H}$) as proxies for information density in human and machine communication. In \cref{thm: aep_llm}, we formalize the connection between these views, showing that BF--defined as the exponentiated entropy rate--aligns naturally with this theoretical framework. Unlike prior studies focused primarily on linguistic theory or cognitive science, our work operationalizes this principle at scale across modern LLMs, linking information density to alignment training, decoding dynamics, and output variability in a unified analysis.

\section{Discussion: BF and Cross-Sample Diversity Measures}

\tmlrrevisee{
The diversity measures of \citet{kirk2024understanding} --- Self-BLEU, expectation-adjusted distinct $n$-grams, and embedding cosine similarity --- are statistics of a finite pool of $N$ generations sampled from the model. BF, in
contrast, is a statistic of the underlying conditional distribution $P(Y_{1:N}\,|\,x;\theta)$ itself: it is the entropy rate of this distribution. Although our hybrid estimator (\cref{eq:hybrid_estimator}) computes BF from sampled sequences, the estimand is the distribution-level entropy rate, not a similarity score between sampled outputs.

The two quantities are correlated in practice but not interchangeable. Consider two stylized distributions over a single generation step. Distribution P places mass $\tfrac{1}{2}$ on each of two near-synonymous tokens whose continuations are nearly identical: BF$\,=\,2$, but Self-BLEU between sampled pairs is high, indicating low cross-sample diversity. Distribution Q places mass $0.99$ on one token and $0.01$ on a token that triggers a wildly different continuation: BF$\,\approx\,1$, but the rare sample diverges sharply from the typical one, indicating high cross-sample diversity. The two measures order this pair of distributions in opposite directions. Empirically, our existing lexical-diversity analysis (\cref{sec: lexical_diversity_and_bf}) confirms that BF and surface-level diversity correlate only weakly across the generation pools used in this paper, consistent with the formal argument.

We therefore view BF and cross-sample diversity as complementary rather than competing: cross-sample diversity asks ``are these $N$ generations different from one another?''; BF asks ``how concentrated is the underlying distribution from which any sample is drawn?''. The position-wise, matched-length, and mid-generation analyses in the main text rely on the per-step distributional reading and do not have natural analogues in the sample-pool framework.
}

\label{app: related_works}

%% file: past_papers/branching_factor/bf_self_narrowing_mvhn.tex
\section{Why Does BF Decrease? Disentangling Autoregressive Self-Narrowing from Alignment}
\label{app: bf_self_narrowing}

This appendix provides the full setup, prompts, and per-model results for the analysis summarized in \cref{sec: bf_self_narrowing}; the two main figures (\cref{fig:bf_self_narrowing_injection,fig:bf_agentic_feedback}) appear there.

\mvhnrevise{A natural question -- raised during review -- is \emph{why} BF often decreases over generation, and in particular why it also decreases for the \textsc{Random Strings} task, where there is no semantic task structure that alignment, stylistic tokens, or distribution collapse could plausibly act on. We do not claim a theorem that BF must decrease at every token position; local increases can occur. Instead, we argue, and provide controlled evidence, that two effects have been conflated and should be separated:}

\begin{itemize}
\item \mvhnrevise{\textbf{Autoregressive self-narrowing (the trend).} Empirically, as an autoregressive model conditions on its own growing prefix, the next-token distribution often becomes more concentrated. This is a robust aggregate tendency in our experiments -- for base and aligned models alike -- and it does not require the context to be meaningful. It is consistent with autoregressive left-to-right training, information-processing intuitions, and our AEP analysis (\cref{thm: aep_llm}), but the AEP result should not be read as proving monotone token-wise BF decrease.}
\item \mvhnrevise{\textbf{Alignment (the level and steepness).} Alignment tuning lowers the absolute BF and often accelerates the early narrowing (consistent with our nudging experiments, \cref{sec: nudging}). It governs \emph{how low} and \emph{how fast}, rather than being the sole explanation for why decreasing BF is commonly observed.}
\end{itemize}

\mvhnrevise{Under this view the \textsc{Random Strings} result is not contradictory but a useful demonstration of self-narrowing, precisely because no semantic structure is available to explain the decrease away.}

\paragraph{\mvhnrevise{A controlled intervention: substituting external random tokens for the model's own prefix.}}
\mvhnrevise{To separate the trend from alignment, we hold the model fixed and change only the \emph{source} of the prefix it conditions on. In the baseline, the model generates normally and conditions on its own output. In the intervention, we replace the first $k$ tokens of context---which the baseline would have generated itself---with an equally long block of externally-sampled i.i.d.\ random tokens (content the model never produced), and then let it continue from token $k$ onward. We measure BF on a position axis aligned to the true generation index, so the baseline and the substituted runs are directly comparable. \cref{fig:bf_self_narrowing_injection} shows the result for three Base/Aligned pairs. Two observations are consistent across all models: (i) at the substitution point BF \emph{surges} back into the high-BF regime, as external out-of-distribution content re-opens the consideration set; and (ii) BF then generally \emph{decays again} under continued autoregression, retracing the same narrowing shape as the self-conditioned baseline. Because the only manipulated variable is the source of the prefix (self-generated vs.\ externally-supplied random tokens), this is an intervention rather than a correlation. It supports the view that the decreasing trend is a broad empirical regularity associated with autoregressive self-conditioning rather than an alignment artifact, while also showing that BF can be \emph{increased} on demand by supplying unexpected content. Alignment changes the absolute scale (aligned/DPO models operate in a much lower BF band and re-collapse faster) but not the qualitative dynamics.}

\paragraph{\mvhnrevise{Implementation details of the intervention.}}
\mvhnrevise{We use two substitution cut points, $k=30$ and $k=60$ model tokens (dashed lines in \cref{fig:bf_self_narrowing_injection}). The external block is sampled i.i.d.\ and truncated to the exact token count using the model's own tokenizer, so the substituted and self-conditioned runs remain aligned on the true generation index. One anomaly is worth flagging: Qwen3-4B-Instruct operates in an already very small BF range (roughly 1.1--1.4), where finite-sample estimation noise can noticeably affect the curve shape; although its tail still shows some decrease, we treat that panel cautiously and leave a sharper diagnosis to future work with access to intermediate checkpoints.}

\paragraph{\mvhnrevise{A setting where information arrives over time: agentic environment feedback.}}
\mvhnrevise{We also build a minimal one-turn agentic setting, inspired by controlled situational-understanding environments for testing state tracking in aligned models~\citep{yang-ettinger-2023-follow}, in which the model receives new external information mid-generation. Each prompt contains a task, a current environment state, and a partial plan; then we append one \texttt{Environment Feedback:} message and ask the model to revise the plan and choose the next action. The prompt template is:}

\begin{quote}
\small\ttfamily
You are an agent interacting with an environment. Maintain a multi-step plan, update it after each environment message, and choose the next action.\\
\\
Task: [task]\\
\\
Current state: <state>\\
\\
Plan so far: <plan>\\
\\
Environment Feedback: [feedback]\\
\\
Given this feedback, revise the plan if needed and produce the next reasoning step and action.
\end{quote}

\mvhnrevise{We instantiate four scenario families and cycle through them when generating prompts. The concrete setups are:}

\begin{itemize}
\item \mvhnrevise{\textbf{Chess endgame.} Task: play White in a simplified endgame with White king on e5, white queen on d4, and black king on g7. Plan: restrict the black king, then bring the white king closer before checkmate. Control feedback says the queen improved control of the seventh rank; adversarial feedback says the black king found an escape square and direct checks now risk stalemate or repetition.}
\item \mvhnrevise{\textbf{Warehouse robot.} Task: move a fragile package from shelf A to packing station D. State: the robot is at shelf A, the package is secured, corridor B is open, and station D is available. Plan: move through corridor B and place the package on a padded tray. Control feedback says the robot reached corridor B and the package remains stable; adversarial feedback says a cart blocks corridor B and the grip sensor reports instability.}
\item \mvhnrevise{\textbf{Python debugging.} Task: debug a small CSV-processing pipeline. State: the parser loads a CSV file, validates rows, and writes normalized records. Plan: reproduce the failing row, verify schema checks, then patch the narrowest failing component. Control feedback says the malformed timestamp is correctly rejected; adversarial feedback says the schema check passed, but the file can be empty and the parser silently returns \texttt{None}.}
\item \mvhnrevise{\textbf{Search-and-rescue drone.} Task: navigate a drone to locate a missing person. State: the drone is in hallway H1, the target beacon is strongest toward room R3, and battery is at 62\%. Plan: enter R3, scan, then return through H1 if the beacon weakens. Control feedback says the drone entered R3 and the path back remains clear; adversarial feedback says smoke filled R3, the beacon reflected from a metal door, and battery use increased.}
\end{itemize}

\mvhnrevise{Within each scenario, the task, state, and plan are identical across conditions; only the feedback field changes. The \emph{control} condition reports normal progress consistent with the current plan. The \emph{adversarial} condition introduces a new event that invalidates part of the plan. The \emph{random-noise} condition fills the same feedback slot with random ASCII text of the same role in the prompt. We then compute BF on the model's next continuation using the same estimator as in \cref{sec: bf_measure}, and report the change in BF relative to the matched control. \cref{fig:bf_agentic_feedback} shows that adversarial and random-noise feedback \emph{increase} BF relative to the matched control, with the effect largest for less-aligned models and compressed -- occasionally near zero -- for the most heavily aligned ones. This provides a concrete, repeatable answer to whether some inputs consistently raise BF: content that is \emph{unexpected from the model's own predictive viewpoint} does.}

\paragraph{\mvhnrevise{Connection to negation.}}
\mvhnrevise{This mechanism also helps explain the prompt-complexity result reported in \cref{app: full_taskwise_bf}: in \textsc{Cognac}, increasing prompt complexity through \emph{negation} \emph{increases} BF. Negation is another instance of context that is hard to reconcile with the model's expectations. Across these three independent settings -- substituted random tokens, adversarial environment feedback, and negation -- we observe the same qualitative pattern: unexpected context can raise BF, while continued self-conditioning tends to lower it again.}

\paragraph{\mvhnrevise{Scope of the causal claim.}}
\mvhnrevise{We do not claim a mechanistic, circuit-level account of self-narrowing, nor a mathematical proof that BF must decrease monotonically. The intervention above is a behavioral one: by fixing the model and toggling only the prefix source, it separates the general autoregressive self-conditioning tendency from alignment more cleanly than purely observational evidence. Deeper mechanistic explanations -- e.g., activation/norm dynamics or attention concentration -- are a promising direction we leave to future work. Our claim is the more modest one that BF decrease is a robust aggregate tendency under autoregressive generation, while alignment governs its level and steepness.}

%% file: past_papers/branching_factor/lexical_diversity.tex
\section{Discussion: Diversity and BF Correlation}
\label{sec: lexical_diversity_and_bf}
Following the branching factor (BF) analysis in \cref{sec: prelim}, a higher BF suggests greater lexical diversity in finite samples. To examine the relationship between BF and traditional diversity metrics, we compute Distinct-N~\citep{li2016diversity}, incorporating necessary LLM-specific adaptations \citep{tevet2021evaluating, guo2024benchmarking, kirk2024understanding}. We then conduct a correlation analysis between Distinct-N and BF.

\begin{figure*}[t]
    \centering
    \begin{subfigure}[t]{0.48\textwidth}
    \centering
     \includegraphics[width=\linewidth]{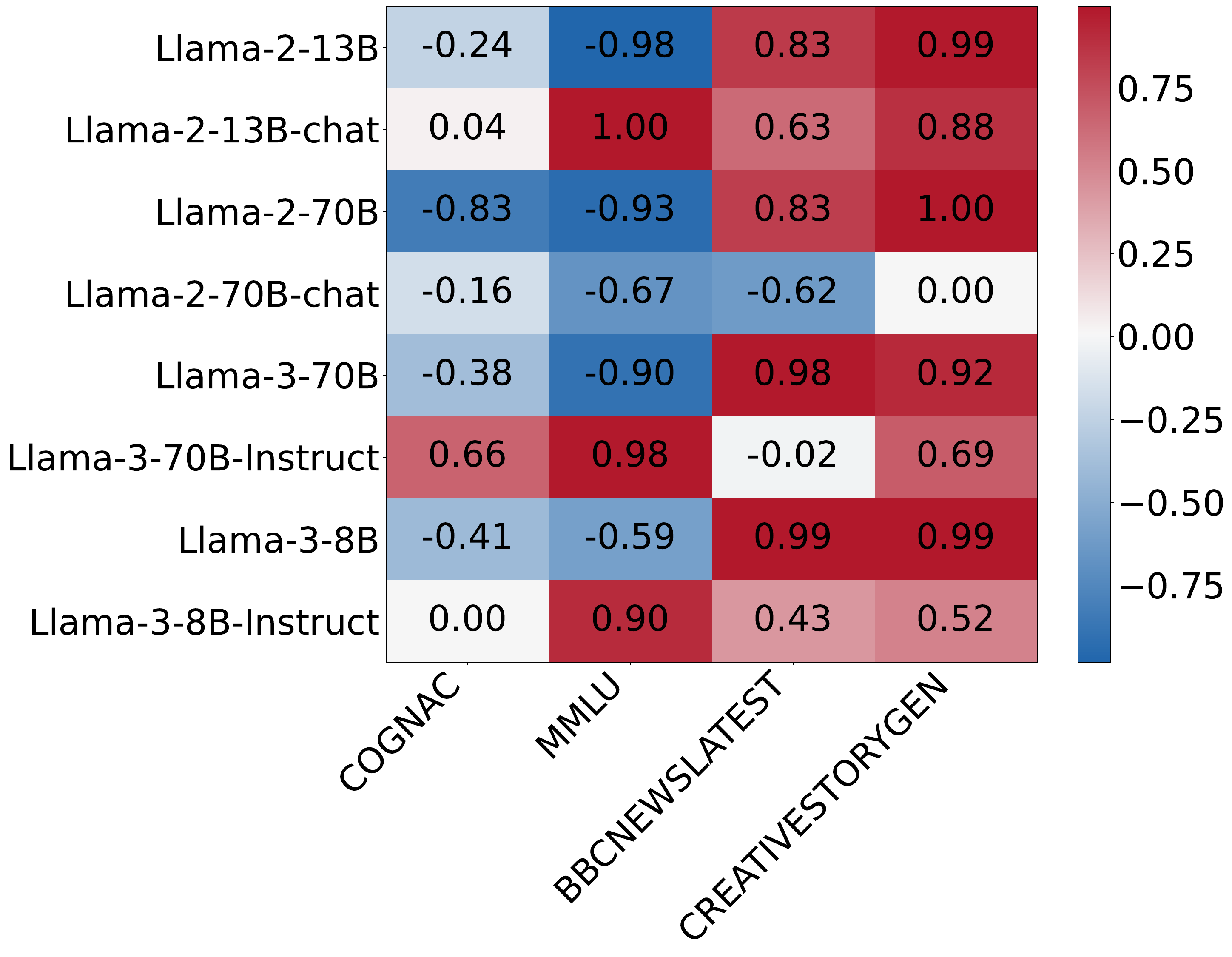}
        \caption{Signed $R^2$(Distinct-1, BF), MaxLength=5}
             \label{fig: mmlu_signed_r2_diversity_bf_maxlen_5}
    \end{subfigure}
       \begin{subfigure}[t]{0.48\textwidth}
    \centering
     \includegraphics[width=\linewidth]{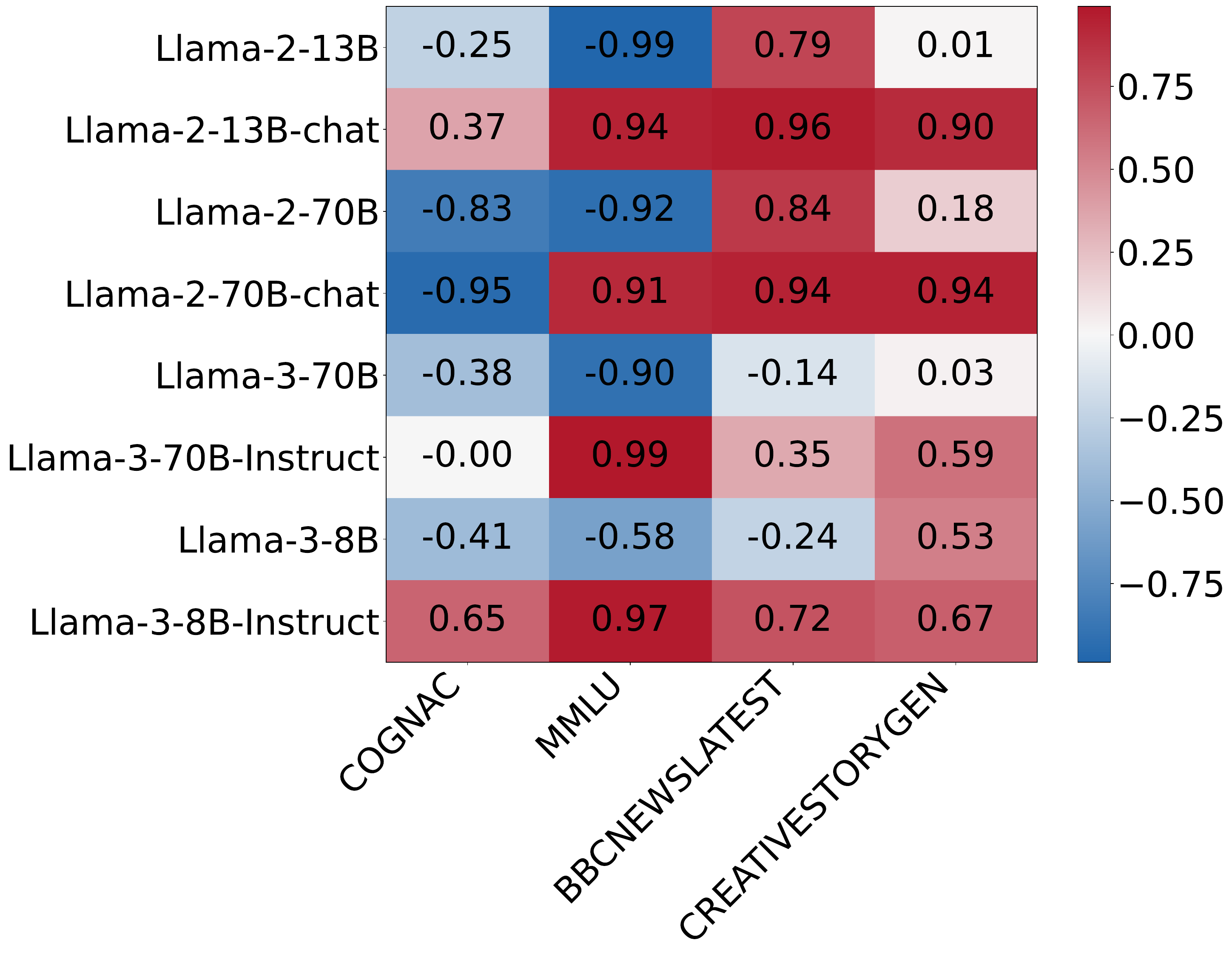}
        \caption{Signed $R^2$(Distinct-1, BF), MaxLength=50}
             \label{fig: mmlu_signed_r2_diversity_bf_maxlen_full}
    \end{subfigure}
      \begin{subfigure}[t]{0.48\textwidth}
    \centering
     \includegraphics[width=\linewidth]{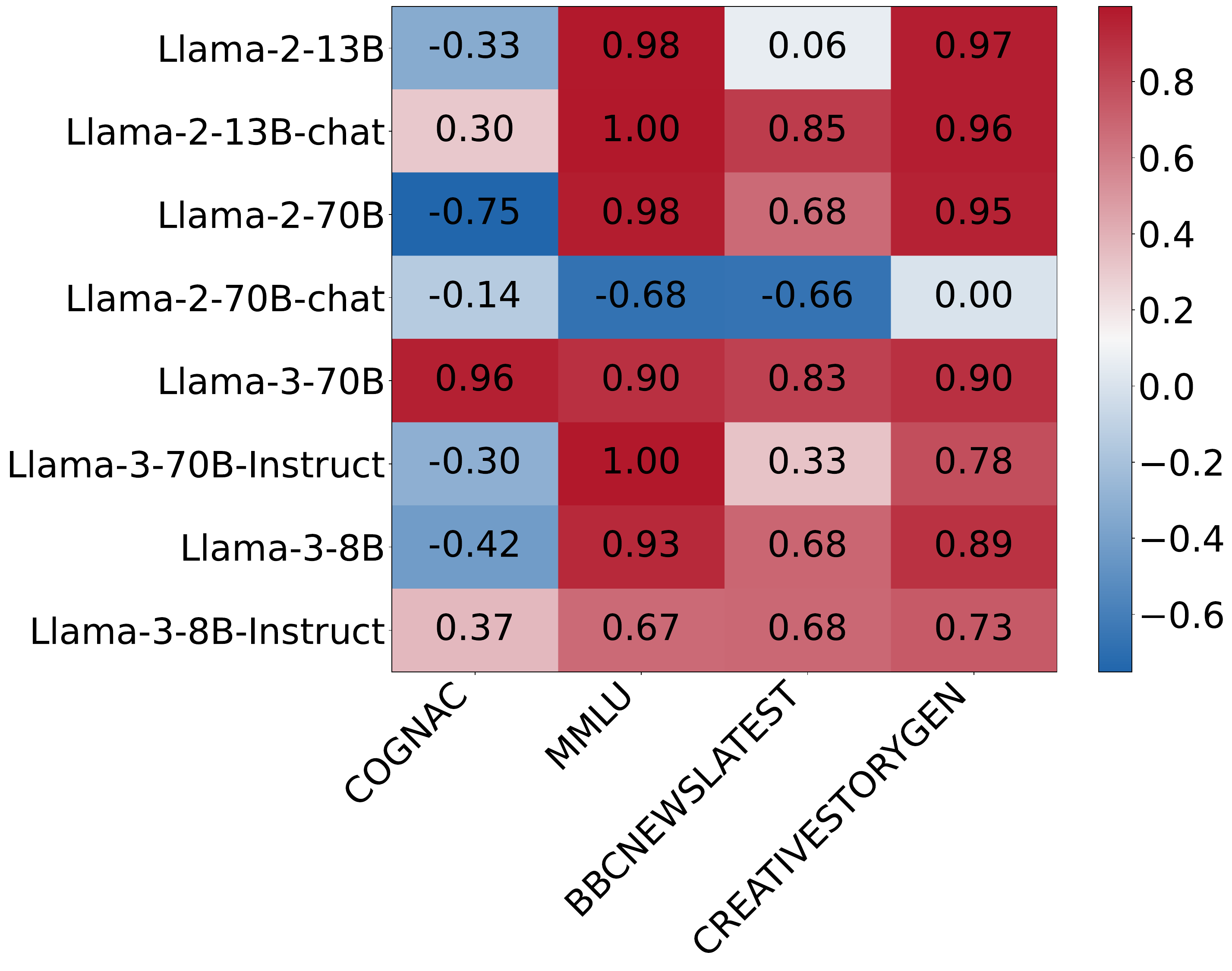}
        \caption{Signed $R^2$(Distinct-2, BF), MaxLength=5}
             \label{fig: mmlu_signed_r2_diversity_distinct2_bf_maxlen_5}
    \end{subfigure}
    \begin{subfigure}[t]{0.48\textwidth}
    \centering
     \includegraphics[width=\linewidth]{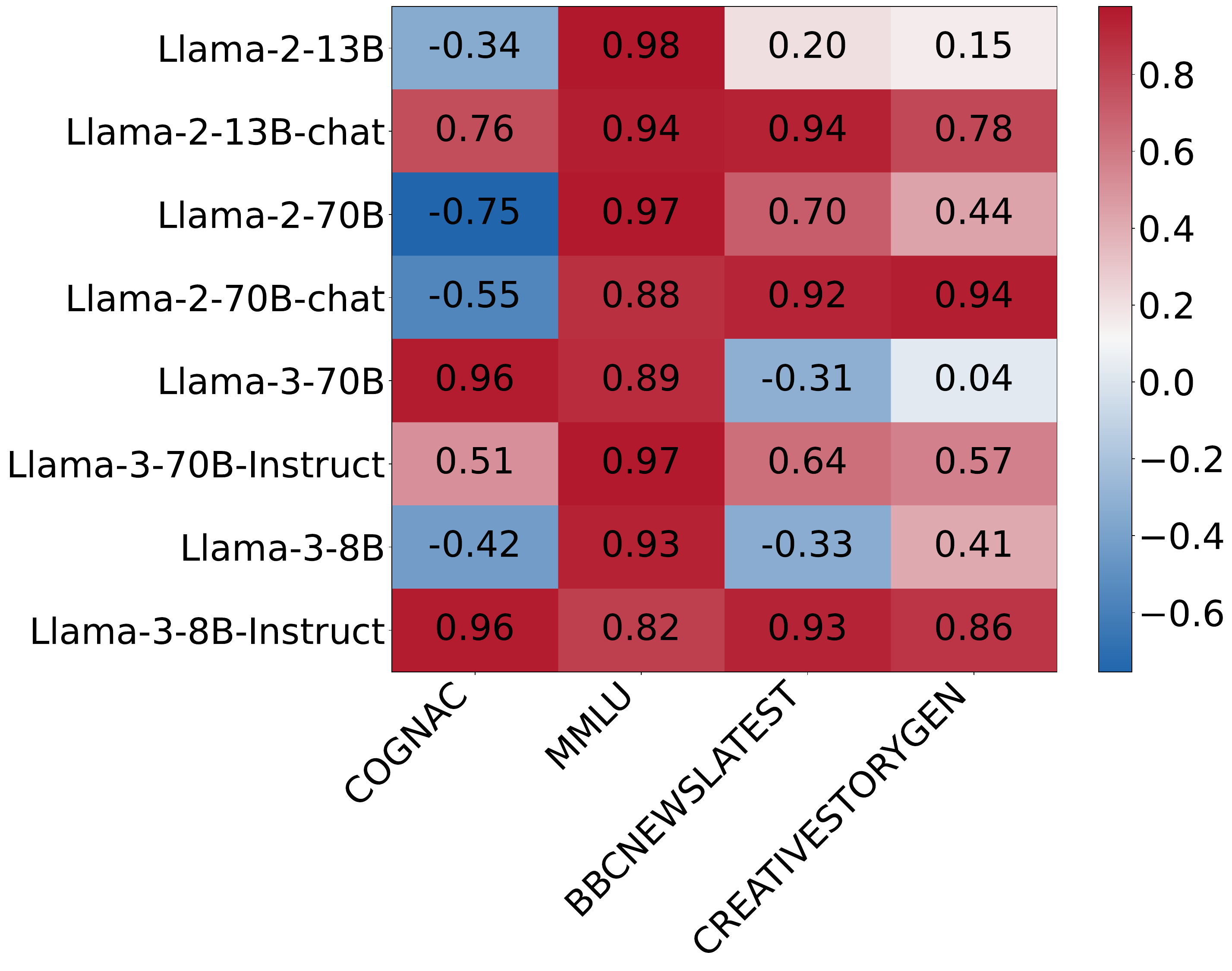}
        \caption{Signed $R^2$(Distinct-2, BF), MaxLength=50}
             \label{fig: mmlu_signed_r2_diversity_distinct2_bf_maxlen_50}
    \end{subfigure}
          \begin{subfigure}[t]{0.48\textwidth}
    \centering
     \includegraphics[width=\linewidth]{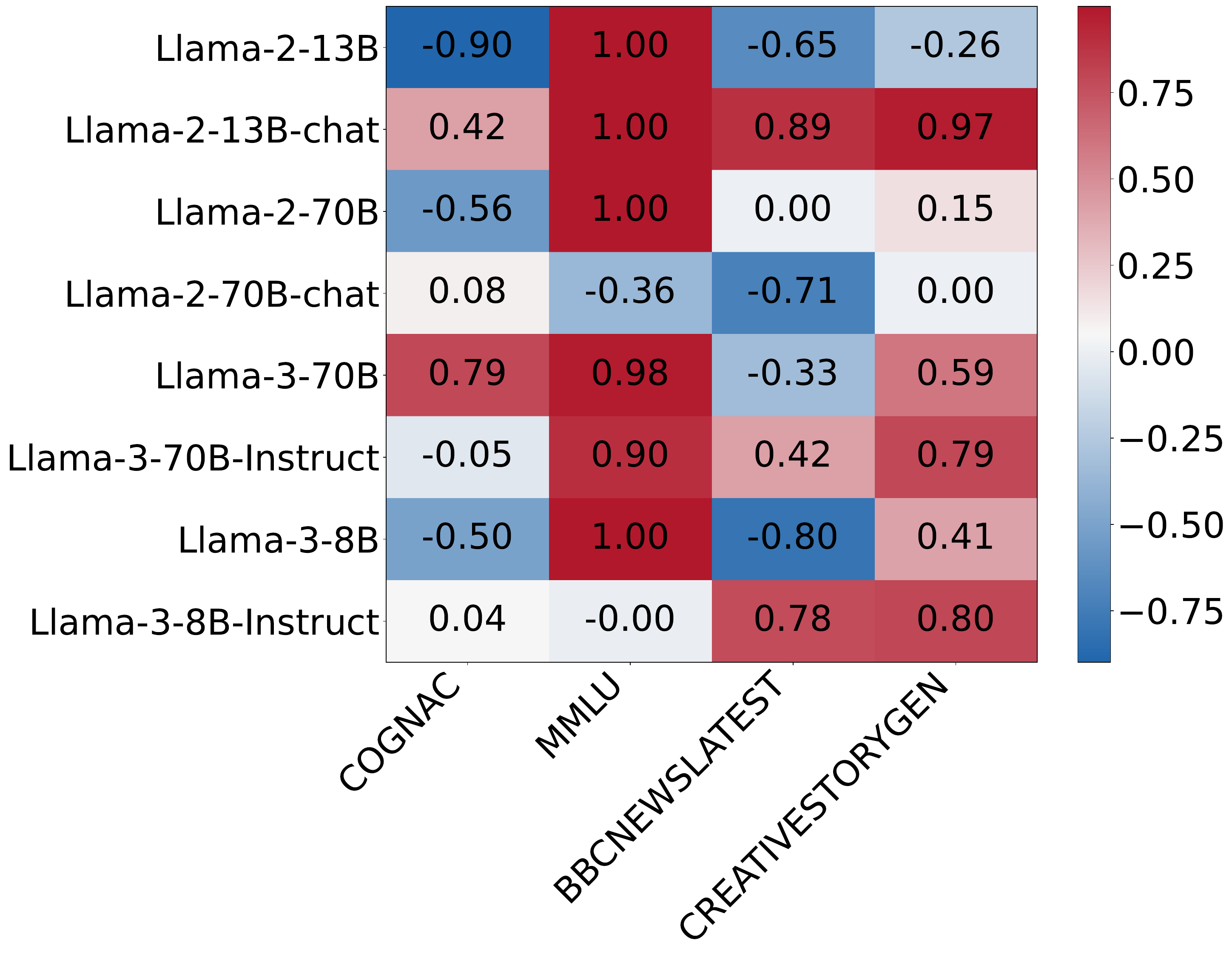}
        \caption{Signed $R^2$(Distinct-4, BF), MaxLength=5}
             \label{fig: mmlu_signed_r2_diversity_distinct4_bf_maxlen_5}
    \end{subfigure}
    \begin{subfigure}[t]{0.48\textwidth}
    \centering
     \includegraphics[width=\linewidth]{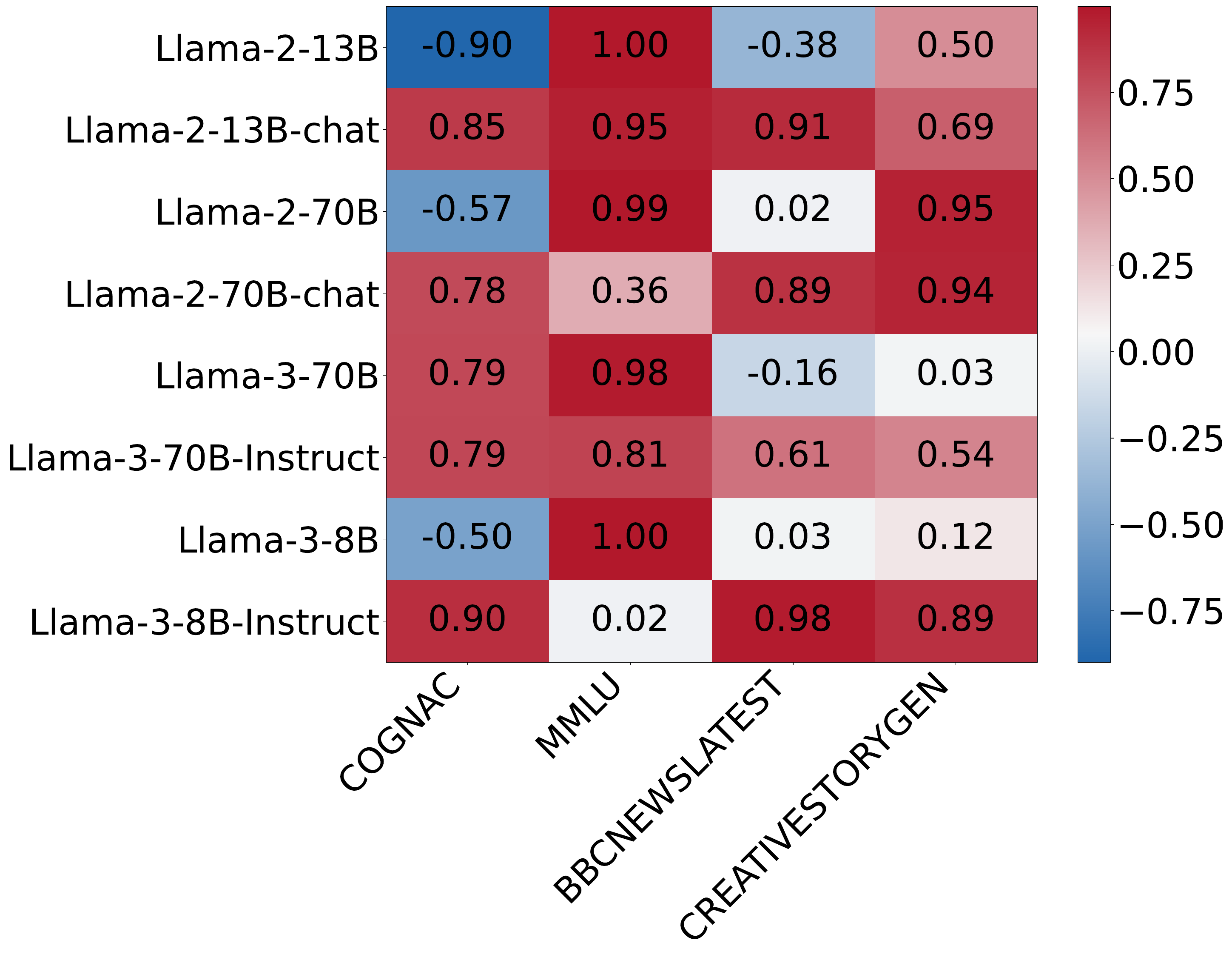}
        \caption{Signed $R^2$(Distinct-4, BF), MaxLength=50}
             \label{fig: mmlu_signed_r2_diversity_distinct4_bf_maxlen_50}
    \end{subfigure}
    \caption{Correlational Analysis of BF and Distinct-N. We can find there is no consistent correlation between Distinct-N and BF.
    }
    \label{fig: diversity_bf_correlation}
\end{figure*}

Our results, presented in \cref{fig: diversity_bf_correlation}, show \textbf{no consistent correlation} between BF and Distinct-N. Depending on the model and task, the relationship can be strongly positive, strongly negative, or entirely absent (e.g., Llama-3-70B-Instruct on Cognac at \cref{fig: mmlu_signed_r2_diversity_bf_maxlen_full}). This empirical inconsistency highlights a fundamental conceptual point: \emph{BF measures a property of the underlying probability distribution, whereas diversity metrics measure a surface property of finite samples.}

BF, as the exponentiated entropy, characterizes the ``width'' of the model's entire output distribution. In contrast, metrics like Distinct-N describe a small set of sampled outputs and are known to be unreliable proxies for distributional properties, being sensitive to confounding factors like generation length \citep{liu-etal-2022-rethinking}.\footnote{While the EAD metric~\citep{liu-etal-2022-rethinking} mitigates this issue, it remains influenced by vocabulary size and is not model-agnostic.} This distinction is critical, as two models can produce samples of similar diversity while having fundamentally different underlying distributions (e.g., with infinite KL-divergence), a nuance that BF captures but sample-based metrics miss. Therefore, our work focuses on probability concentration, measured by BF, as a more fundamental and insightful tool for understanding a model's generative process.

Viewing alignment through the lens of BF reduction provides a unified framework that explains several disparate observations: it clarifies how alignment shrinks the generative horizon, why aligned models are less sensitive to decoding methods, and how techniques like Chain-of-Thought stabilize generation by shifting information to low-BF regions. This focus on distributional properties aligns with emerging research highlighting the importance of a model's entropy in understanding and improving advanced reasoning capabilities \cite{cui2025entropy, wu2025invisible}.

%% file: past_papers/branching_factor/data_contamination.tex
    \begin{figure*}[ht!]
    \centering
    \includegraphics[width=0.6\linewidth]{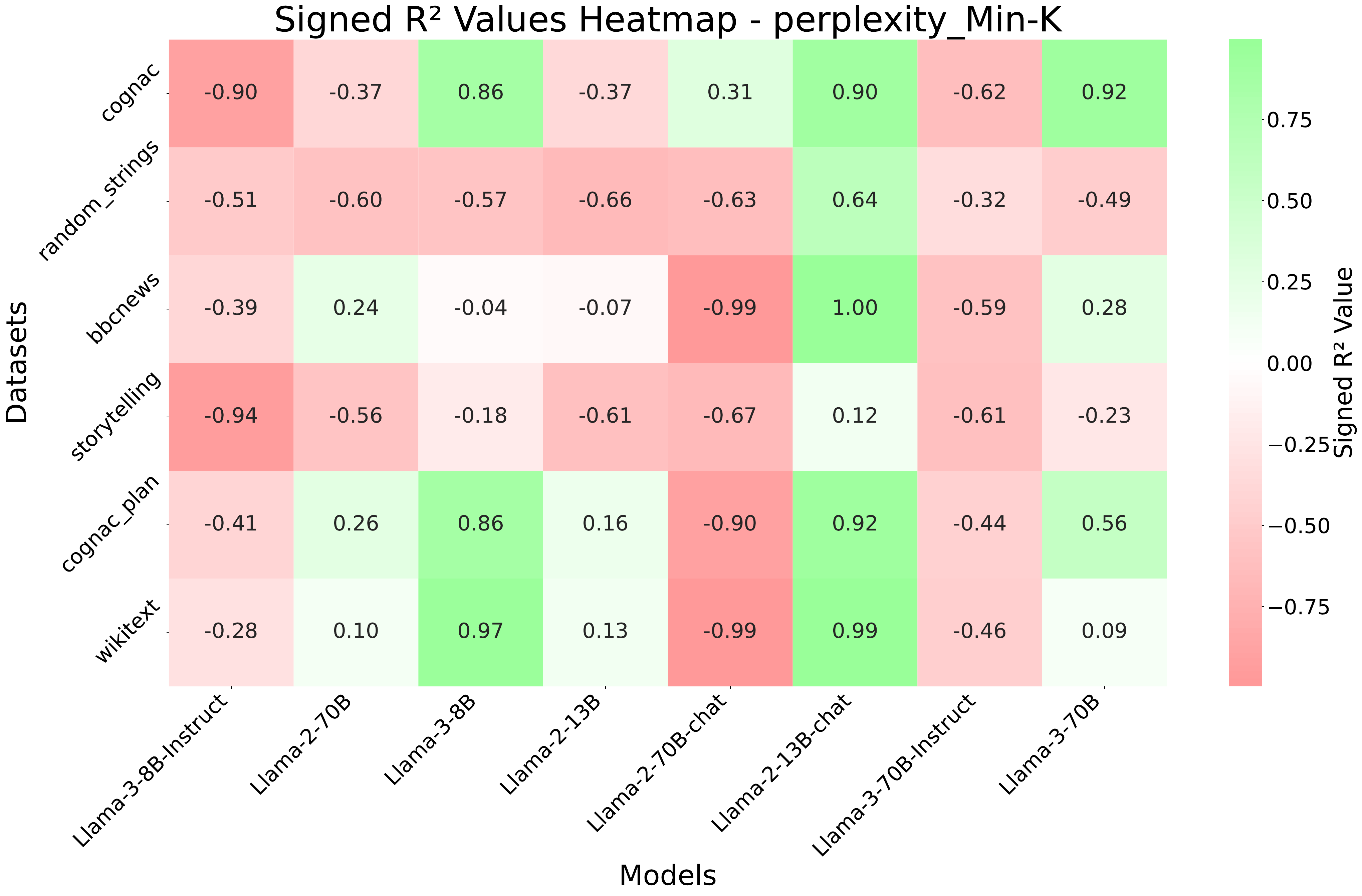}
    \caption{Signed $R^2$ values heatmap investigating correlation between BF and Min-K $\%$. }
    \label{fig: signed_r_squared}
    \end{figure*}
\section{Confounder Investigation: Data Contamination}
\label{sec: data_contamination}

A potential confounder in our analysis is the influence of data contamination. If prompts closely resemble the training data (including pretraining and alignment tuning, i.e., "data contamination"), smaller BF values would be expected, and vice versa. To evaluate this, we use the Min-K$\%$ metric~\citep{shi2024detecting}, which quantifies the overlap between experimental prompts and training data. Following \citet{shi2024detecting}, we set $K=20$ and compute the average log-likelihood for the minimum $K\%$ of tokens. Using these Min-K$\%$ values, we perform a linear regression with BF to assess their correlation. For each task-model pair, Signed $R^2$ values are reported to indicate the strength and sign (positive or negative) of the correlation.

The results of the Min-K$\%$ analysis are presented in \cref{fig: signed_r_squared}. Significant negative correlations between BF and Min-K$\%$ are observed for models such as Llama-3-8B-Instruct, Llama-3-70B-Instruct, and Llama-2-70B-Chat across several tasks. Conversely, Llama-3-8B and Llama-2-13B-Chat models exhibit positive correlations. For other models, correlations are notably weaker. Overall, there is no consistent correlation pattern between BF and Min-K$\%$ across datasets and models, suggesting that data contamination cannot fully explain our findings.

%% file: past_papers/annealed_sampling/tex/appendix.tex
\appendix
\section{Minimal-RL Training Details}
\label{app: minimal_rl_training_details}
We mainly follow the Minimal-RL recipe~\citep{xiong2025minimalist} in our experiments to ensure a fair comparison among different rollout sampling strategies. Specifically, we set a series of hyperparameters as in \cref{tab: hyperparameter}:

\begin{table}[h!]
\centering
\begin{tabular}{@{}cc@{}}
\toprule
Hyperparameter                                 & Value(s)        \\ 
\midrule
\multicolumn{1}{c}{Training Batch Size}      & \multicolumn{1}{c}{1024} \\ 
\multicolumn{1}{c}{Max Prompt Length}        & \multicolumn{1}{c}{1024} \\ 
\multicolumn{1}{c}{Max Response Length}      & \multicolumn{1}{c}{3072} \\ 
\multicolumn{1}{c}{Mini Batch Size}          & \multicolumn{1}{c}{256}  \\ 
\multicolumn{1}{c}{Micro Batch Size Per GPU} & \multicolumn{1}{c}{4}    \\ 
\multicolumn{1}{c}{Learning Rate}            & \multicolumn{1}{c}{$10^{-6}$} \\ 
\bottomrule
\end{tabular}%
\caption{Hyperparameter Setup for Running Minimal-RL recipe. }
\label{tab: hyperparameter}
\end{table}

\section{Off-policy Issue and Truncated Importance Sampling Correction}
\label{sec:off-policy}
\subsection{Sampling Techniques Can Introduce Off-Policy Issue}
\label{subsec:temperature-sampling}
\begin{figure}[h!]
    \centering
    \begin{subfigure}[b]{0.26\textwidth}
    \centering
    \includegraphics[width=\linewidth]{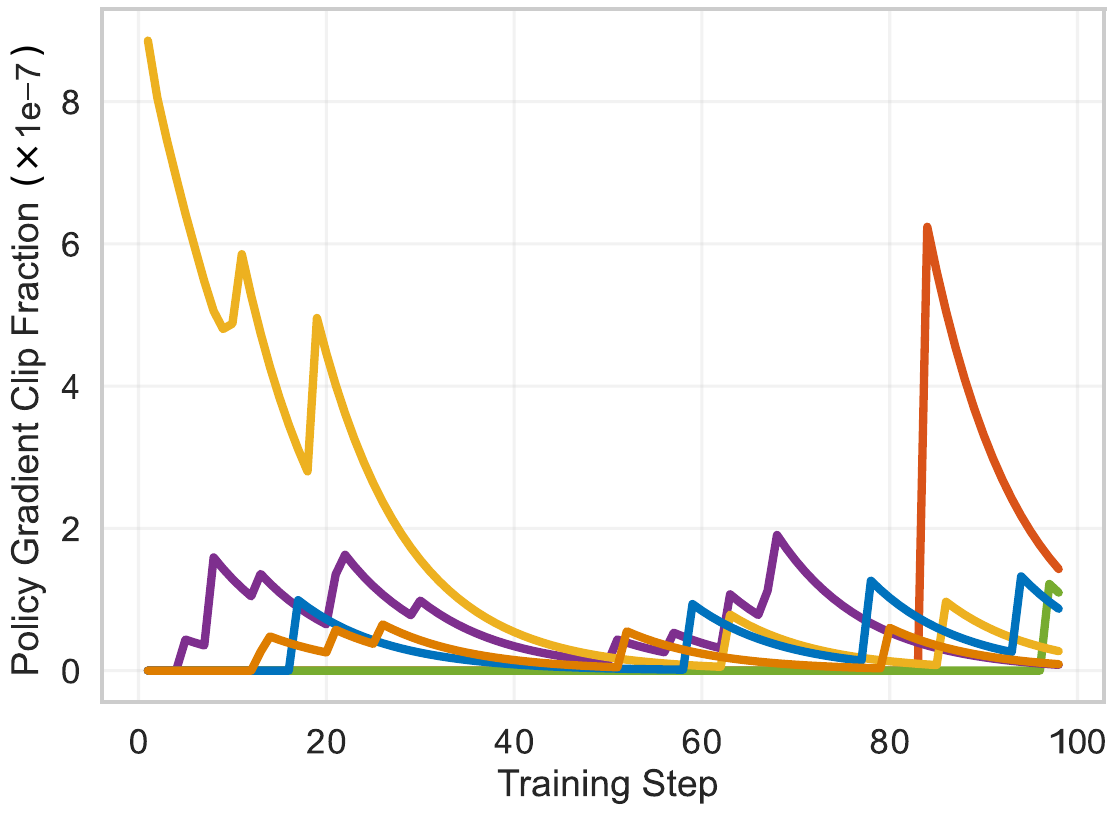}
    \caption{Clip Fraction Surge}
    \label{fig: off_policy_pg_clipfrac}
  \end{subfigure}
  \begin{subfigure}[b]{0.28\textwidth}
    \centering
    \includegraphics[width=\linewidth]{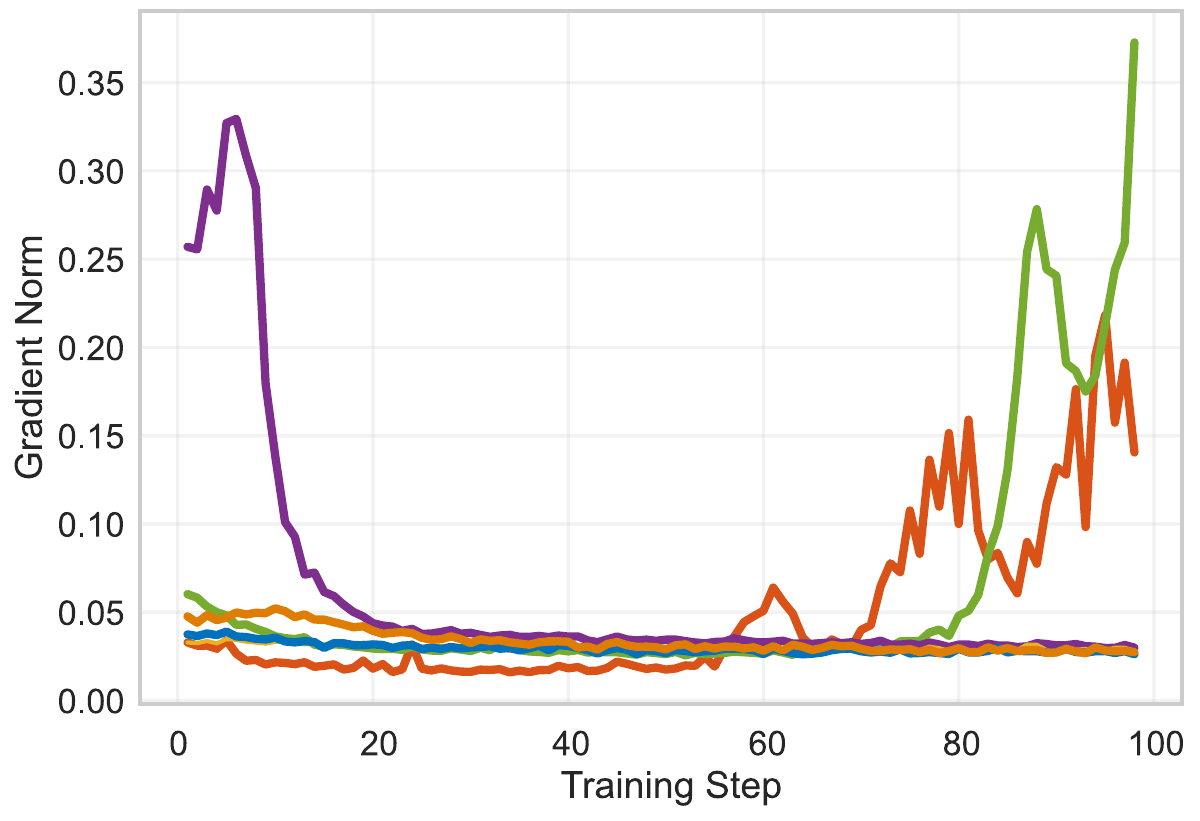}
    \caption{Gradient Norm Surge.}
    \label{fig: off_policy_grad_norm}
  \end{subfigure}
   \begin{subfigure}[b]{0.27\textwidth}
    \centering
    \includegraphics[width=\linewidth]{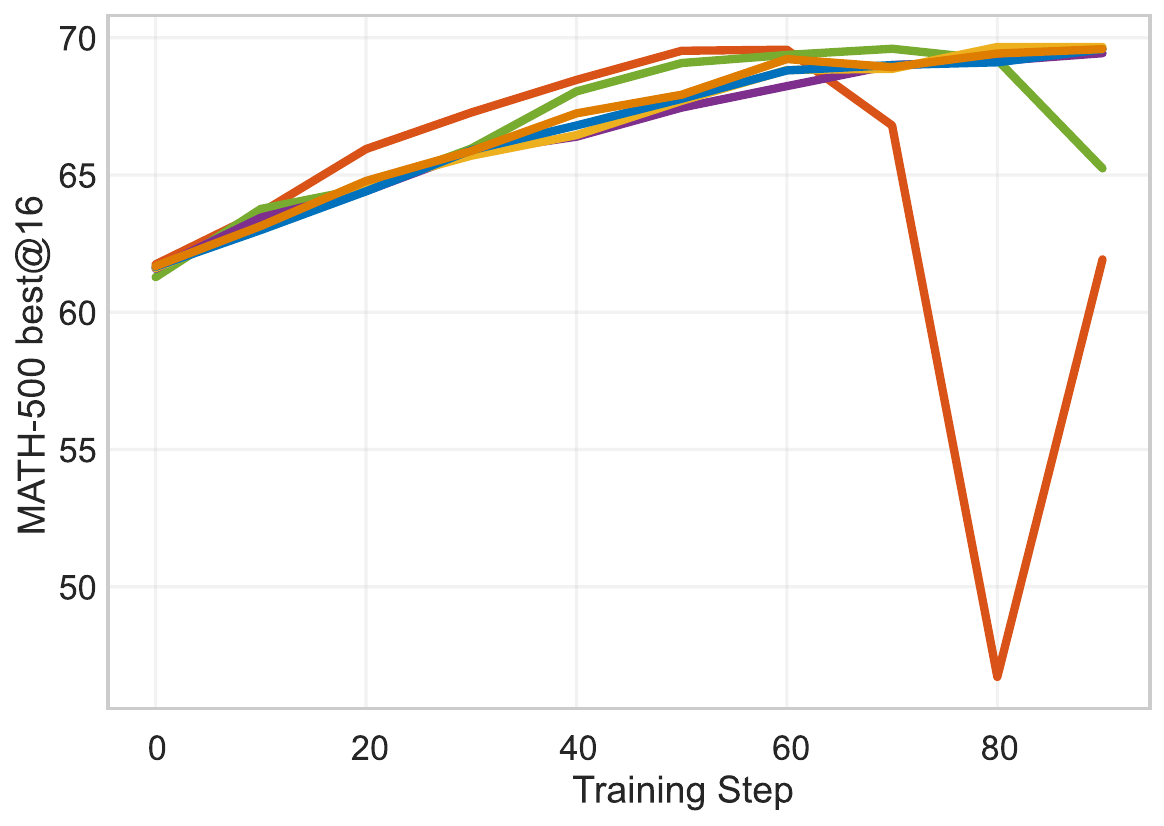}
    \caption{Drastic Best@16 Drop}
     \label{fig: off_policy_best_at_16}
  \end{subfigure}
  \begin{subfigure}[t]{0.1\textwidth}
    \centering
    \vspace{-2.5cm}
    \includegraphics[width=\linewidth]{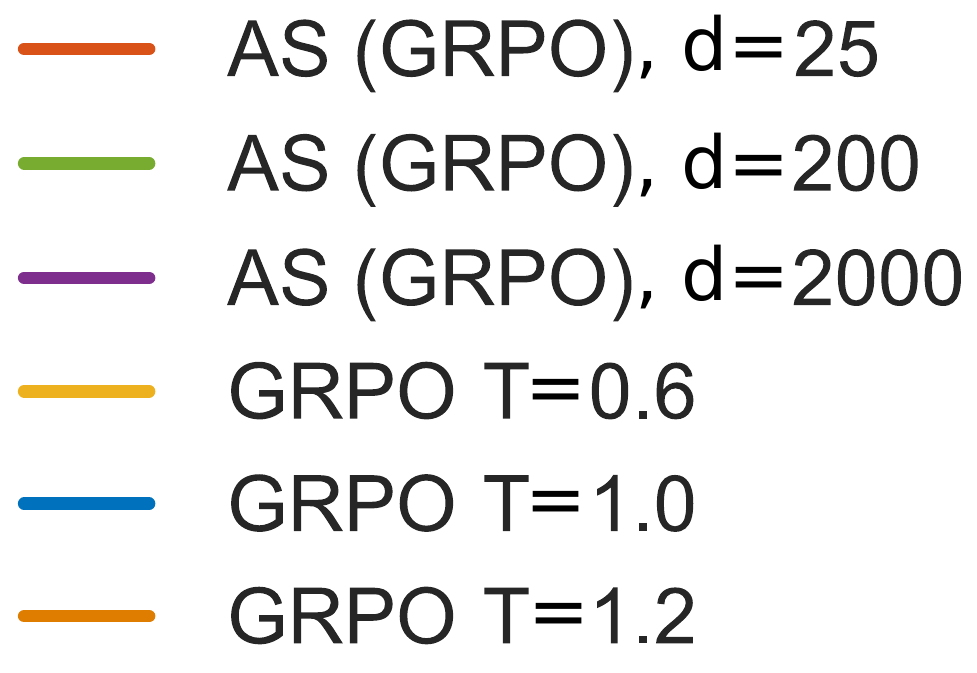}
  \end{subfigure}
  \caption{Off-policy samples bring training instability. The base model is Qwen2.5-Math-1.5B.}
  \label{fig:optim_failure_for_as_offpolicy}
\end{figure}

One subtle yet troublesome drawback of reinforcement learning with sampling techniques is that it simultaneously introduces the \emph{off-policy} problem: there is a gap between the behavior policy (used for sampling) and the target policy (being optimized and parametrized by $\theta$). This might introduce instability to the training and cause it to fail (See example training of RLVR with EAD \cref{fig:optim_failure_for_as_offpolicy}).

Noted that this off-policy phenomenon widely exists for any efficient sampling framework \citep{yao2025offpolicy} and sampling strategy (for instance, when applying best-of-$n$ sampling \citep{xiong2025minimalist} or filtering out responses \citep{shrivastava2025sample}, the underlying distribution of responses is implicitly altered). To be more precise, we take the policy gradient loss as an example:
\begin{small}
\begin{align*}
\E_{x \sim \mathcal{D},y{\sim} \mathbin{\color{red}\pi^{\text{sampling}}_{\theta_{\text{old}}}(\cdot\mid x)}}
\left[\frac{\pi_{\theta}(y\mid x)}{\pi_{\theta_{\text{old}}}(y\mid x)}A(y;x)\right] 
=\E_{x \sim \mathcal{D},y{\sim} \mathbin{\pi_{\theta_{\text{old}}}(\cdot\mid x)}}\left[{\color{red}\frac{\pi^{\text{sampling}}_{\theta_{\text{old}}}(y\mid x)}{\pi_{\theta_{\text{old}}}(y\mid x)}}\times\frac{\pi_{\theta}(y\mid x)}{\pi_{\theta_{\text{old}}}(y\mid x)}A(y;x)\right],
\end{align*}
\end{small}
where $\pi^{\text{sampling}}_{\theta_{\text{old}}}(\cdot\mid x)$ represents the underlying sampling distribution. 
In such case, an extra weight is implicitly added to each response in addition to its advantage $A(y;x)$.

This extra weight can significantly inflate the variance of the policy gradient, posing a stability challenge that our proposed \alg needs to mitigate.
To quantify this effect in our proposed \alg, we now analyze such variance under a standard, fixed \textbf{temperature sampling}.

We use $\tau=1$ to define a policy $\pi$ and consider the effect of $\tau$ on the variance of the gradient estimator. We reduce the problem to analyzing the variance inflation factor
\begin{equation}
\label{eq:variance-inflation-factor}
\mathbb{E}_{y\sim\pi(\cdot\mid x;\tau)}\left[\frac{\pi(y\mid x;1)^2}{\pi(y\mid x;\tau)^2}\right].
\end{equation}
We begin with one-token case. Let $o_i$ denote the $i$th token in the vocabulary $V$ and $h_i$ is its logit. Then \eqref{eq:variance-inflation-factor} can be rewritten as
\begin{align*}
    \sum_{i=1}^{|V|}\frac{h_i/(\sum_{j=1}^{|V|} h_j)}{h^{1/\tau}_i/(\sum_{j=1}^{|V|} h^{1/\tau}_j)}\times\frac{h_i}{\sum_{j=1}^{|V|} h_j}
    = \frac{\sum_{i=1}^{|V|} h_i^{2-1/\tau}\sum_{i=1}^{|V|} h^{1/\tau}_i}{\left(\sum_{i=1}^{|V|} h_i\right)^2}
\end{align*}

\begin{proposition}
Suppose $h_i\in[0,1]$ for all $i\in V$.
$\sum_{i=1}^{|V|} h_i^{2-1/\tau}\sum_{i=1}^{|V|} h^{1/\tau}_i$ is decreasing when $\tau\le1$ and increasing when $\tau\ge1$, which implies it has a global minimum at $\tau=1$.
\end{proposition}
\begin{proof}
Let $x=1/\tau$. We define 
\[
f(x)=\log\left(\sum_{i=1}^{|V|} h_i^{2-x}\right)+\log\left(\sum_{i=1}^{|V|} h^{x}_i\right).
\]
Its derivative is
\[
f'(x)=\frac{-\sum_{i=1}^{|V|} h_i^{2-x}\log h_i}{\sum_{i=1}^{|V|} h_i^{2-x}}
+\frac{\sum_{i=1}^{|V|} h_i^{x}\log h_i}{\sum_{i=1}^{|V|} h_i^{x}}.
\]
To analyze the sign of $f'(x)$, we define a helper function $g(x) = \frac{\sum_{i=1}^{|V|} h_i^{x}\log h_i}{\sum_{i=1}^{|V|} h_i^{x}}$. Then, $f'(x)=g(x)-g(2-x)$ and its sign depends on whether $g(x)$ is greater than, less than, or equal to $g(2-x)$. We take a look at derivative of $g$:
\[
g'(x) = \frac{\left(\sum_{i=1}^{|V|} h_i^{x}(\log h_i)^2\right) \left(\sum_{i=1}^{|V|} h_i^{x}\right) - \left(\sum_{i=1}^{|V|} h_i^{x}\log h_i\right)^2}{\left(\sum_{i=1}^{|V|} h_i^{x}\right)^2}\ge0.
\]
Hence, $g$ is an increasing function and
\begin{equation*}
\begin{cases}
~f'(x)=g(x)-g(2-x)\ge0,~~\mathrm{when}~x\ge1\\
~f'(x)=g(x)-g(2-x)=0,~~\mathrm{when}~x=1\\
~f'(x)=g(x)-g(2-x)\le0,~~\mathrm{when}~x\le1.
\end{cases}
\end{equation*}
Accordingly, $f$ is increasing when $x\ge1$ and is decreasing when $x\le1$. Then the proposition easily follows.
\end{proof}

The same conclusion can be proved for multiple-token sequence by induction. Therefore, we get that
when the temperature is far from 1, the off-policy issue could be severe and lead to large variance of the gradient estimator. 

\subsection{Truncated Importance Sampling Ratio Correction}
\label{subsec:tis-correction}
To cancel such bias, an importance sampling ratio can be introduced~\citep{hilton2022batch,yao2025offpolicy}:
\begin{small}
\begin{align*}
\E_{x \sim \mathcal{D},y{\sim} \mathbin{\pi_{\theta_{\text{old}}}(\cdot\mid x)}}\left[\frac{\pi_{\theta}(y\mid x)}{\pi_{\theta_{\text{old}}}(y\mid x)}A(y;x)\right]
=
\E_{x \sim \mathcal{D},y{\sim} \mathbin{\color{red}\pi^{\text{sampling}}_{\theta_{\text{old}}}(\cdot\mid x)}}
\left[{\color{red}\frac{\pi_{\theta_{\text{old}}}(y\mid x)}{\pi^{\text{sampling}}_{\theta_{\text{old}}}(y\mid x)}}\times\frac{\pi_{\theta}(y\mid x)}{\pi_{\theta_{\text{old}}}(y\mid x)}A(y;x)\right]. 
\end{align*}
\end{small}
To further prevent negative effects by the extreme likelihood ratios and boost training stability, we truncate the likelihood ratio with an upper bound. That is, \emph{truncated importance sampling} technique \citep{heckman1998matching}. Taking the vanilla policy gradient loss as an example, the modified loss for EAD is as follows:
\begin{equation*}
\label{eq:tis-pg}
\E_{x \sim \mathcal{D},y{\sim}{\pi^{\text{EAD}}_{\theta_{\text{old}}}(\cdot\mid x)}}
\left[{\color{blue}\min\left(\frac{\pi_{\theta_{\text{old}}}(y\mid x)}{\pi^{\text{EAD}}_{\theta_{\text{old}}}(y\mid x)},\varepsilon\right)}\frac{\pi_{\theta}(y\mid x)}{\pi_{\theta_{\text{old}}}(y\mid x)}A(y;x)\right].
\end{equation*}

\newpage

\newpage
\section{Proof of Temperature-Entropy Relationship}
\label{app:entropy_proof}

\begin{proposition}
The entropy of the softmax distribution is a non-decreasing function of the temperature $\tau > 0$.
\end{proposition}

\begin{proof}
The strategy is to show that the entropy $H$ is a non-increasing function of the inverse temperature $\beta = 1/\tau > 0$.
The probability of sampling token $v$ with temperature $\tau$ is given by the policy $\pi_\theta$. For simplicity in the derivation, we denote this probability as $p_v(\tau)$:
\[
p_v(\tau) \triangleq \pi_\theta(y_t=v \mid [x, y_{<t}]; \tau)
\]

Let $h_v$ be the logit for a token $v$ in the vocabulary $V$. The probability of a token as a function of $\beta$ is given by:
\[
p_v(\beta) = \frac{\exp(\beta h_v)}{\sum_{v' \in V} \exp(\beta h_{v'})} \triangleq \frac{\exp(\beta h_v)}{Z(\beta)},
\]
where $Z(\beta)$ is the partition function. The entropy, as a function of $\beta$, is:
\[
H(\beta) = -\sum_{v \in V} p_v(\beta) \log p_v(\beta).
\]
We can rewrite the entropy by substituting $\log p_v(\beta) = \beta h_v - \log Z(\beta)$:
\begin{align*}
H(\beta) &= -\sum_{v \in V} p_v(\beta) (\beta h_v - \log Z(\beta)) \\
&= \log Z(\beta) \left(\sum_{v \in V} p_v(\beta)\right) - \beta \sum_{v \in V} h_v p_v(\beta) \\
&= \log Z(\beta) - \beta \cdot \mathbb{E}_{v \sim p(\beta)}[h_v].
\end{align*}
Now, we differentiate $H(\beta)$ with respect to $\beta$. Let $\bar{h}(\beta) = \mathbb{E}[h_v]$.
\[
\frac{dH}{d\beta} = \frac{d}{d\beta}(\log Z(\beta)) - \frac{d}{d\beta}(\beta \bar{h}(\beta)).
\]
First, we find the derivative of the log-partition function:
\[
\frac{d}{d\beta}(\log Z(\beta)) = \frac{Z'(\beta)}{Z(\beta)} = \frac{\sum_v h_v \exp(\beta h_v)}{Z(\beta)} = \sum_v h_v p_v(\beta) = \bar{h}(\beta).
\]
Next, we use the product rule for the second term:
\[
\frac{d}{d\beta}(\beta \bar{h}(\beta)) = \bar{h}(\beta) + \beta \frac{d\bar{h}}{d\beta}.
\]
Combining these gives:
\[
\frac{dH}{d\beta} = \bar{h}(\beta) - \left(\bar{h}(\beta) + \beta \frac{d\bar{h}}{d\beta}\right) = -\beta \frac{d\bar{h}}{d\beta}.
\]
The derivative $\frac{d\bar{h}}{d\beta}$ is the variance of the logits. We can show this by differentiating $\bar{h}(\beta)$:
\begin{align*}
\frac{d\bar{h}}{d\beta} &= \frac{d}{d\beta}\left( \frac{\sum_v h_v \exp(\beta h_v)}{Z(\beta)} \right) \\
&= \frac{(\sum_v h_v^2 \exp(\beta h_v))Z(\beta) - (\sum_v h_v \exp(\beta h_v))Z'(\beta)}{Z(\beta)^2} \\
&= \sum_v h_v^2 p_v(\beta) - \left(\sum_v h_v p_v(\beta)\right)\left(\frac{Z'(\beta)}{Z(\beta)}\right) \\
&= \mathbb{E}[h^2] - (\mathbb{E}[h])^2 = \mathrm{Var}_{v \sim p(\beta)}(h_v).
\end{align*}
Substituting this back, we arrive at the final expression for the derivative of entropy:
\[
\frac{dH}{d\beta} = -\beta \cdot \mathrm{Var}_{v \sim p(\beta)}(h_v).
\]
By definition, the temperature $\tau > 0$, so the inverse temperature $\beta > 0$. The variance of any random variable is non-negative. This can be formally shown using \textbf{Jensen's inequality}: for the convex function $\phi(x)=x^2$, we have $\mathbb{E}[\phi(h)] \ge \phi(\mathbb{E}[h])$, which means $\mathbb{E}[h^2] \ge (\mathbb{E}[h])^2$, and thus $\mathrm{Var}(h) \ge 0$.

Therefore, the derivative of entropy with respect to $\beta$ is non-positive:
\[
\frac{dH}{d\beta} = \underbrace{-\beta}_{\le 0} \cdot \underbrace{\mathrm{Var}(h_v)}_{\ge 0} \le 0.
\]
Since $H(\beta)$ is a non-increasing function of $\beta$, and $\beta$ is inversely proportional to $T$, it follows that $H(\tau)$ must be a non-decreasing function of the temperature $\tau$.
\end{proof}

\section{Increasing Length in RL Training}
\label{app: increased_length}
\begin{figure}[h!]
\centering
\includegraphics[width=\linewidth]{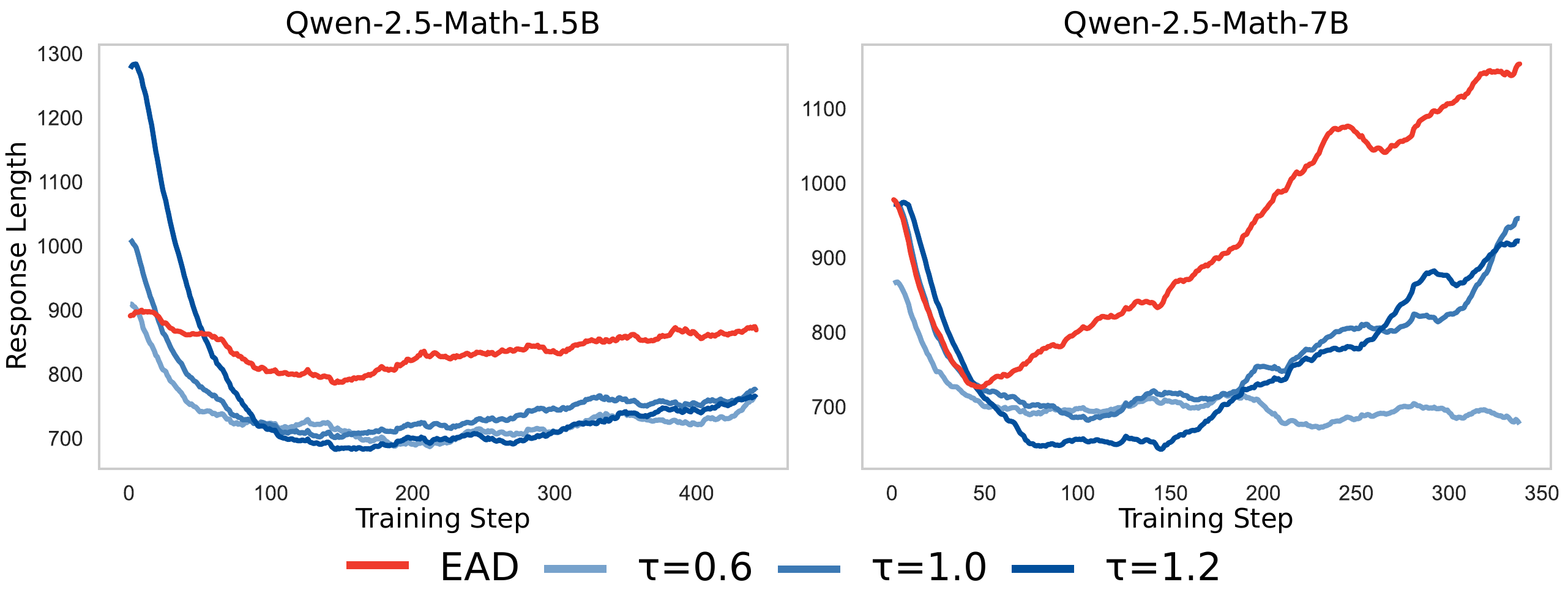}
\caption{Compared with normal temperature sampling, EAD can naturally incentivize the model to generate longer reasoning chains. }
\label{fig: response_length}
\end{figure}

During RL training, our algorithm (\alg) incentivizes the model to generate longer, more effective reasoning chains for difficult problems, especially for 7B models (\cref{fig: response_length}). While both \alg and temperature sampling initially learn to shorten their responses by shifting from complex code-based solutions to direct mathematical reasoning, their behavior later diverges. The response length from temperature sampling stabilizes, whereas \alg learns to selectively increase reasoning length for harder problems, which boosts final performance. For these experiments, EAD is applied in the DAPO algorithm for sampling rollouts. We use the same training setup as detailed in \cref{sec:experiments}. 

%% file: past_papers/base_aligned_collaboration/sections/appendix/main.tex
\appendix

\input{past_papers/base_aligned_collaboration/sections/related_work}


\input{past_papers/base_aligned_collaboration/sections/appendix/router_design}

\section{Dataset Details}
\label{app:datasets}

\paragraph{NoveltyBench}
is a human-curated benchmark designed to evaluate the ability of LLMs to produce
multiple distinct yet high-quality outputs. The instructions are constructed
such that multiple valid answers exist, spanning four categories: randomness
(\eg{} ``the result of a die roll''), underspecified factual knowledge
(\eg{} ``tell me a capital city in Africa''), creative writing
(\eg{} ``short poem or story''), and subjective queries
(\eg{} ``recommendation or opinion'').
While effective for fine-grained diversity evaluation, NoveltyBench prompts are
intentionally simple and often yield short outputs with limited opportunities
for variation. We therefore complement it with more complex datasets.

\paragraph{WildChat}
is a large-scale dataset of real human--LLM conversations.
Following \citet{zhang2025noveltybench}, we select a subset of prompts without
fixed ground-truth answers to emphasize open-endedness, enabling evaluation
under more realistic and challenging settings.

\paragraph{Narrative-Discourse}
is a dataset for long-form creative writing, where models extend fictional film
synopses in English. The dataset provides structured annotations of discourse-level
elements such as turning points, story arcs, and arousal \citep{tian2024large},
enabling evaluation of long-form structural diversity.

\input{past_papers/base_aligned_collaboration/sections/appendix/evaluation}

\section{Experiment Setup}
\label{app:exp_setup}

\paragraph{Inference Setup.}
Our study focuses on group-level diversity.
For each prompt, we generate a group of $n=10$ outputs.
Unless otherwise specified, sampling uses temperature $1.0$ and nucleus sampling (top-$p$) with $p=0.9$ and no top-$k$ truncation, following \citet{zhang2025noveltybench}.
For Diverse Beam Search \citep{vijayakumar2016diverse}, we use $\text{beams}{=}2n{=}20$, $\text{beam\_groups}{=}n{=}10$, and $\text{diversity\_penalty}{=}1.0$, following the original paper.
Other exceptions apply only to baselines that inherently require alternative decoding strategies.

\section{Validation on Verifiable Benchmarks}
\label{app:verifiable_tasks}

This section reports detailed results on two verifiable benchmarks used to
validate that the improvements of \name{} are not artifacts of open-ended
evaluation metrics: verifiable instruction following (IFEval) and mathematical
reasoning (GSM8K).

\paragraph{Verifiable instruction following (IFEval).}
IFEval provides instruction-following tasks with automatically verifiable
constraints \citep{zhou2023instructionfollowingevaluationlargelanguage}.
We evaluate instruction-following accuracy as a quality metric alongside diversity metrics.
As shown in \Cref{fig:ifeval_large_figure}, at matched quality levels,
\name{} achieves consistently higher diversity than the aligned baseline.

\paragraph{Mathematical reasoning (GSM8K).}
For mathematical reasoning, we evaluate on GSM8K \citep{cobbe2021training} and
measure Acc and Pass@10 as the accuracy metrics.
Diversity is computed over sampled solution outputs.
As shown in \Cref{fig:gsm8k_large_figure}, \name{} maintains high accuracy (e.g., ~90\% Pass@10) while
achieving substantially higher diversity across a wide range of operating points.

Overall, the qualitative trends on both benchmarks mirror those observed on
open-ended generation tasks, confirming that the gains of \name{} generalize
beyond open-ended evaluation metrics.

\section{Detailed Results on Open-ended Tasks}
\label{app:additional_results}

\input{past_papers/base_aligned_collaboration/sections/appendix/detail_results}

\input{past_papers/base_aligned_collaboration/tables/sub_novelty_bench_olmo}

\section{Narrative Discourse Evaluation Details}
\label{app:storyarc}

\input{past_papers/base_aligned_collaboration/sections/appendix/storyarc}

\input{past_papers/base_aligned_collaboration/sections/appendix/analysis}

\section{Human Evaluation Details}
\label{app:human_eval}

\input{past_papers/base_aligned_collaboration/sections/appendix/human_eval}

\input{past_papers/base_aligned_collaboration/tables/app_judge_model_prompt}

\section{Qualitative Examples}
\label{app:qual_examples}

\input{past_papers/base_aligned_collaboration/tables/case/case_noveltybench_1}

\input{past_papers/base_aligned_collaboration/tables/case/case_wildchat_1}
\input{past_papers/base_aligned_collaboration/tables/case/case_storyarc_1}

%% file: past_papers/base_aligned_collaboration/sections/related_work.tex
\section*{Limitations}
We see several limitations in our work.

(\textit{i}) Collaboration between two models, by nature, incurs additional FLOPs overhead per token relative to single-model inference; the prototype is not yet perfectly engineered for speed, but we see speculative decoding, LoRA-based aligned models, and KV-cache sharing are mitigations (\appref{app:analysis}).

(\textit{ii}) Beyond the open-ended generation tasks studied here, other tasks such as agent and code generation can also benefit from diversity, and can extend to much longer horizons; we leave these extensions for future work.

(\textit{iii}) Our focus is inference-time improvement, so directions such as learned routers and leveraging diversity within RL rollouts, while relevant, are out of scope for this paper.

(\textit{iv}) We mainly study the canonical base and aligned checkpoints; intermediate alignment checkpoints, where available, likely offer additional Pareto gains and are a promising direction (\appref{app:analysis}).

\section{Related Work}
\label{sec:related_work}
\paragraph{Diversity Degradation in Alignment.}
While alignment techniques like RLHF enhance model performance in instruction following and reasoning, they systematically curtail output diversity. This trade-off is widely recognized, with a growing body of evidence demonstrating that aligned models are less diverse than their base counterparts. For example, studies have questioned their artistic authenticity \citep{chakrabarty2024art}, and benchmarks like NoveltyBench reveal their diminished capacity for humanlike diversity and creativity \citep{zhang2025noveltybench, tian2024large, lu2025ai, west2025base}. The underlying mechanism for this degradation is ``probability concentration," as the alignment process ``sharpens'' the model's output distribution, thereby steering it into low-entropy generation paths \citep{yang2026alignment}.
The diversity degradation impacts the downstream applications. 
It manifests as a loss of linguistic idiosyncrasies \citep{chakrabarty2025can}, increased format homogeneity \citep{zhang2024lists}, and diminished creativity \citep{west2025base} in generated text.
Beyond linguistic characteristics, alignment-induced constraints lead models to converge on a restricted repertoire of strategies, thereby diminishing diversity in reasoning \citep{chen2024not, ding2025dynamic}, data synthesis \citep{kim2024evaluating, yang2025measuring}, deep research \citep{Xiao2025SubmodularOptimization}, social simulation \citep{wang2025large}, and gaming \citep{west2025base}. 
More critically, the lack of output diversity has been shown to further reduce outcome diversity \citep{padmakumar2023does} and creativity \citep{Meincke2025, ashkinaze2025ai} in human interactions with these models. 
These studies demonstrate that diminished diversity in model outputs adversely affects how humans ideate, create, and engage.
Our work addresses this challenge directly, proposing an inference-time collaborative decoding framework that optimizes the \problem{} by combining the strengths of both base and aligned models.

\paragraph{Diversity-Promoting Methods.}
Approaches to enhance the diversity of aligned LLMs fall into two main categories: training-time and inference-time methods.
\textbf{Training-time methods} typically modify the learning objective to encourage varied outputs. A prominent line of work adapts Direct Preference Optimization (DPO; \citet{rafailov2023direct}) by incorporating diversity-aware mechanisms, such as f-divergence penalties \citep{wang2024beyond}, set-level diversity rewards \citep{lanchantin2025diverse}, or re-weighted loss objectives \citep{chung2025modifying, ismayilzada2025creativepreferenceoptimization}. Other approaches leverage different architectures, like generative flow networks, to the same end \citep{kwon2024gdpo}. While these methods can instill diversity directly into the model, they require substantial computational resources for retraining and offer little flexibility for user-specific diversity needs at inference.
\textbf{Inference-time methods} offer a more lightweight and adaptable alternative. These include modifications to decoding algorithms like diverse beam search \citep{vijayakumar2016diverse} and various prompt engineering strategies, such as paraphrasing \citep{meyerson2024language, wang2025multilingual, zhang2025noveltybench, wang2025large, wong2024simplestrat}. Existing inference-time methods for improving diversity typically incur high computational costs via multiple decoding passes or long-horizon planning. Or they significantly degrade generation quality \citep{peeperkorn2024temperature}. Achieving a stable diversity-quality trade-off with these techniques remains a challenge.
Our proposed method \name, is an inference-time framework designed to offer more explicit and reliable control over this trade-off. We therefore focus our comparison on baselines from this category.

\paragraph{Multi-Model Collaborative Generation.}
Prior work has explored collaborative frameworks where multiple language models work in concert to improve generation quality and efficiency (\eg{} computation cost and latency). These approaches can be grouped by their collaboration mechanism.
One line of work focuses on \textbf{weight-level collaobration}. This includes merging reward or policy models, or using Mixture-of-Experts (MoE)~\citep{shazeer2017outrageously} architectures to create a single, more capable system better aligned with diverse human preferences \citep{rame2023rewarded, zheng-etal-2025-model, shi2025flexolmo}. Another mechanism is \textbf{token-level collaboration}, where multiple models collaborate during decoding by exchanging next-token probability distributions or candidate token choices at each step, to improve attributes like coherence and factuality, or reduce latency \citep{leviathan2023fast, li2023contrastive, zheng2024citer, fei2025nudging}. More recently, \textbf{multi-agent systems} have emerged, in which models debate or discuss to leverage their complementary strengths for complex, creative tasks \citep{lu2024llm, venkatraman2025collabstory, huot2025agents}.
Our \name, advances token-level collaboration. While existing methods in this area primarily target quality or efficiency, we focus on navigating the diversity-quality trade-off.

%% file: past_papers/base_aligned_collaboration/sections/appendix/router_design.tex
\section{\name{} Framework Implementation and Router Details}
\label{app: implementation_details}

\paragraph{Models.}

We apply each model's default chat template during collaboration (for base models, we apply a plain shifting template), served on vLLM local host.
We disable every model's tool call and thinking for fair comparison. 

\paragraph{Tokenization Alignment.}
Tokenization sometimes mismatches between the two models (even base and aligned pairs), particularly around punctuation, special tokens, or rare words, can lead to incoherent sub-word boundaries. To address this, we enforce that tokens representing a single semantic unit (e.g., a word or format element) must all be produced by the same model. This avoids artifacts such as broken punctuation or malformed words.

\paragraph{Algorithm Overview.}

Algorithm~\ref{alg:baco} formalizes the full \name{} decoding procedure, including the additional rules described above (aligned first token, word-boundary switching, aligned-controlled EOS) and reconciles Eq.~\eqref{eq:moe_main} with the actual sampling step. We express the router as an ordered tuple of routing strategies $\mathcal{R} = (\mathcal{R}_1, \dots, \mathcal{R}_K)$, where each $\mathcal{R}_k$ may fire in either direction (\textsf{base} or \textsf{aligned}) on its trigger condition, or abstain otherwise (move to next strategy). 
At a routing step, the strategies are evaluated in order and the first $\mathcal{R}_k$ that fires determines $m$. By construction, the last strategy in the tuple always fires, so the cascade always terminates with a definite decision. Single-strategy routers correspond to $K = 1$; combination routers correspond to $K \geq 2$ with complementary signals in practice.

\begin{algorithm}[h]
\caption{\name{} Decoding}
\label{alg:baco}
\begin{algorithmic}[1]
\Require prompt $\inputval$; base model $P_{\text{base}}$;
         aligned model $P_{\text{aligned}}$;
         router $\mathcal{R} = (\mathcal{R}_1, \dots, \mathcal{R}_K)$
         with threshold $\gamma$; max length $T$
\State $\outputval_1 \sim P_{\text{aligned}}(\cdot \mid \inputval)$
       \hfill $\triangleright$ sample first token from aligned
\State $m \gets \text{aligned}$
\For{$t = 2, \dots, T$}
  \State $c_t \gets [\inputval, \outputval_{<t}]$
  \If{$\outputval_{t-1}$ ends a word}
     \State $m \gets \mathcal{R}(c_t, P_{\text{base}}, P_{\text{aligned}}, \gamma)$
            \hfill $\triangleright$ first firing strategy in $\mathcal{R}$ returns \textsf{base} or \textsf{aligned}
  \EndIf
  \State $\outputval_t \sim P_m(\cdot \mid c_t)$
  \If{$\outputval_t = \langle \text{eos} \rangle$ and $\arg\max P_{\text{aligned}}(\cdot \mid c_t) \neq \langle \text{eos} \rangle$}
     \State resample $\outputval_t \sim P_m(\cdot \mid c_t)$ excluding $\langle \text{eos} \rangle$
            \hfill $\triangleright$ aligned-controlled termination
  \EndIf
  \If{$\outputval_t = \langle \text{eos} \rangle$}
     \State \textbf{break}
  \EndIf
\EndFor
\State \textbf{return} $\outputval_1, \dots, \outputval_t$
\end{algorithmic}
\end{algorithm}

\paragraph{Framework Cost.}
\name{} requires two forward passes per decoding step at the worst case, incurring $\sim$2$\times$ FLOPs overhead per token regardless of sample count. Crucially, this overhead is independent of the desired group size $n$, in contrast to baselines such as in-context resampling and paraphrase prompting (which require $n$ sequential passes) or diverse beam search (which expands $\geq n$ beams).
Moreover, stems from \textit{superficial alignment}~\citep{zhou2023lima, lin2023unlocking}, interventions between the largely-agreeing base and aligned models can be sparse~\citep{fei2025nudging}. This sparsity enables practical optimizations such as caching multi-token chunks from one model to minimize switching costs and overlapping communication with computation to hide context-switching latency. Further engineering-level optimizations are complementary and applicable when deploying at scale, including speculative decoding~\citep{leviathan2023fast}, LoRA-based aligned models \citep{hu2021loralowrankadaptationlarge}, and KV cache sharing \citep{liu2025droidspeakkvcachesharing}.

\paragraph{Wall-Clock Runtime Comparison.}
We empirically measure runtime under identical hardware (single A100 80\,GB) for inference-time methods, using the HuggingFace library across the board for fair comparison. Results are in Table~\ref{tab:wallclock}. \name{}'s current prototype is slower than single-model baselines but is comparable to or faster than several diversity-promoting baselines (\eg{} Diverse Beam Search, Logits Ensemble) while substantially outperforming them on the diversity-quality trade-off. The implementation is not yet engineered for speed; the optimizations listed above can further reduce wall-clock cost.

\begin{table}[h]
\centering
\small
\renewcommand{\arraystretch}{1.2}
\begin{tabular}{lccc}
\toprule
\textbf{Method} & \textbf{Time/Sample (s)} & \textbf{Tok/s} & \textbf{Peak Memory} \\
\midrule
Aligned                & $9.8 \pm 3.9$    & 40.0 & 15.1 GiB \\
Response Ensemble      & $10.9 \pm 4.0$   & 34.8 & 15.4 GiB \\
Diverse Beam Search    & $17.0 \pm 5.9$   & 17.0 & 15.8 GiB \\
Logits Ensemble        & $25.5 \pm 0.5$   & 20.1 & 30.1 GiB \\
In-Context Resampling  & $90.5 \pm 35.3$  & 4.1  & 15.5 GiB \\
Paraphrase Prompting   & $117.9 \pm 41.1$ & 3.3  & 15.1 GiB \\
Back Translation       & $120.5 \pm 65.6$ & 3.8  & 15.1 GiB \\
Mix-of-Agents          & $224.9 \pm 123.2$& 3.8  & 15.6 GiB \\
Multi-Agent Debate     & $239.0 \pm 81.5$ & 3.5  & 15.6 GiB \\
\midrule
\name{}                & $22.9 \pm 4.5$   & 19.2 & 29.2 GiB \\
\bottomrule
\end{tabular}
\caption{Wall-clock runtime comparison on a single A100 80\,GB. Tok/s is computed over the final output. \name{} is comparable to or faster than several inference-time diversity baselines while substantially outperforming them on diversity--quality (\Cref{tab:avg_results}).}
\label{tab:wallclock}
\end{table}

\label{app:router}
\subsection{Additional Rules in Router\label{app:router:addrule}}  
We follow \cite{fei2025nudging} in always using the aligned model to generate the first token. Early decoding steps have an outsized influence on generation and typically show greater disagreement between models. Starting from the aligned model improves trajectory quality and reduces the chance of degenerate completions.

Incorporating low-probability tokens from the base model introduces new challenges in sequential generation. 
When switch between models, the receiving model may struggle to continue from an unfamiliar context. 
In particular, the aligned model may terminate the output prematurely, while the base model may fall into degenerate behaviors such as repetition or verbose listing. 
To mitigate this, we constrain output termination by only accepting the end-of-sentence token when it is the top-1 prediction of the aligned model. 

\subsection{All Routing Strategies and Notations}
\label{app:router}
\label{app:router:notation}
The following are all strategies and their corresponding notations that are mentioned in this paper:
\begin{itemize}
    \item -\textsc{Rand}: Route to the base model by random chance $\gamma$.
    \item -\textsc{P}: Route to the base model when base model's top-1 token probability\\
    $\max_{y_t} P_{\text{base}}(y_t \mid x, y_{<t}) < \gamma ~_{\gamma \in [0, 1]}$, otherwise to the aligned model.
    \item -\textsc{P-a}: Route to the base model when aligned model's top-1 token probability\\
    $\max_{y_t} P_{\text{aligned}}(y_t \mid x, y_{<t}) < \gamma ~_{\gamma \in [0, 1]}$, otherwise to the aligned model.
    \item -\textsc{H}: Route to the base model when the entropy of the base model's next token prediction distribution $H_{\text{base}}(y_t \mid x, y_{<t}) > \gamma ~_{\gamma \in [0, +\infty)}$, otherwise to the aligned model.
    \item -\textsc{H-a}: Route to the base model when the entropy of the aligned model's next token prediction distribution $H_{\text{aligned}}(y_t \mid x, y_{<t}) > \gamma ~_{\gamma \in [0, +\infty)}$, otherwise to the aligned model.

    \item -\textsc{PR}: Route to the base model when the ratio between the base model's top 1 token probability and the aligned model's, \ie{} $\frac{\max_{y_t} P_{\text{base}}(y_t \mid x, y_{<t})}{\max_{y_t} P_{\text{aligned}}(y_t \mid x, y_{<t})} < \gamma ~_{\gamma \in (0, 1]}$, otherwise to the aligned model.
    \item -\textsc{HR}: Route to the base model when the ratio between the base model's entropy and the aligned model's, \ie{} $\frac{H_{\text{base}}(c_i \mid q, c_{<i})}{ H_{\text{aligned}}(c_i \mid q, c_{<i})} > \gamma ~_{\gamma \in [1, +\infty)}$, otherwise to the aligned model.
    \item -\textsc{FC}: Route to the base model when both the aligned model and base model sample the next token is a content word\footnote{Long words could be composed by multiple tokens. If so, we will route to the same model multiple steps until the word is finished.}, otherwise to the aligned model.

    \item -\textsc{Punc}: Route to the base model when the base model's and the aligned model's next token is not punctuation or formative tokens (\eg{}  `$\backslash n$'), otherwise, to the aligned model.
    \item -\textsc{Judge}: Route to the base model when an external judge LLM (another aligned model) determines that both of the following conditions are satisfied: \textit{1}) the next token continuation has space to diverge; \textit{2}) the sampled base model continuation is acceptable (\ie{} reasonable and meaningful). Otherwise, to the aligned model. 

\end{itemize}

Comparatively, we observe that the aligned model's logits-based metrics are less distinctive, which aligns with the literature on entropy decrease, hence making routing strategies such as \textsc{-P-A} and \textsc{-H-A} less effective compared with the same metric under the base model's logits.  

\textsc{-FC} and \textsc{-Punc} fail under the same motivation of using content-based linguistic features as a routing strategy, where \textsc{-Punc} is more lightweight than \textsc{-FC}. 
From our empirical observation, the two have on-tier performance. 
However, the introduction of part-of-speech parsing for \textsc{-FC} takes additional computational cost. 
Given the cost limitation, some experiments and analyses take \textsc{-Punc} as representative.

We prompt -\textsc{Judge} with curated heuristic rules and few-shot examples with rationals. Detailed prompt designs are at \Cref{tab:judge-model-prompt}. 
As a more costly strategy, it serves as an extended comparison.

Following on, we have multi-condition routers, which are some possible combinations of the above single-condition routers:

\begin{itemize}
    \item -\textsc{P-FC}: First apply the \textsc{-FC} rule and then the -\textsc{P}. Route to the base model when any one of the following conditions is met: \textit{1}) base model sampled next token is a function word;  \textit{2}) both models' sampled next token are a content word; \textit{3}) base model's top-1 token probability $\max_{y_t} P_{\text{base}}(y_t \mid x, y_{<t}) < \gamma ~_{\gamma \in [0, 1]}$. Otherwise, to the aligned model. 
    \item -\textsc{P-Punc}: First apply the \textsc{-Punc} rule and then the -\textsc{P}. Route to the base model when any one of the following conditions is met: \textit{1}) base model sampled next token is not a punctuation or formatting tokens;  \textit{2}) base model's top-1 token probability $\max_{y_t} P_{\text{base}}(y_t \mid x, y_{<t}) < \gamma ~_{\gamma \in [0, 1]}$. Otherwise, to the aligned model. 
    \item -\textsc{H-FC}: First apply the \textsc{-FC} rule and then the -\textsc{H}. Route to the base model when any one of the following conditions is met: \textit{1}) base model sampled next token is a function word;  \textit{2}) both models' sampled next token are a content word; \textit{3}) base model's entropy of next token prediction distribution $H_{\text{base}}(y_t \mid x, y_{<t}) > \gamma ~_{\gamma \in [0, +\infty)}$. Otherwise, to the aligned model. 
    \item -\textsc{H-Punc}: First apply the \textsc{-Punc} rule and then the -\textsc{P}. Route to the base model when any one of the following conditions is met: \textit{1}) base model sampled next token is not a punctuation or formatting tokens;  \textit{2})  base model's entropy of next token prediction distribution $H_{\text{base}}(y_t \mid x, y_{<t}) > \gamma ~_{\gamma \in [0, +\infty)}$. Otherwise, to the aligned model. 
\end{itemize}

%% file: past_papers/base_aligned_collaboration/sections/appendix/evaluation.tex
\section{Automation Evaluation Details}
\label{app:auto_eval}
\subsection{Diversity Metrics}
\label{app:auto_eval:div}
We are mainly interested in diversity across a group of outputs.
For each prompt \(x\), we sample \(n=10\) outputs: \(\{y_0, \dots, y_{n-1}\}\). We evaluate the diversity of \(\{y_i\}\) using a broad set of automated metrics. Below are full derivations and definitions, grouped by category.

\subsubsection{Lexical Metrics}

\paragraph{Distinct-\(n\).} Ratio of unique $n$-grams to total $n$-grams.   
Let \(\mathcal{G}_n = \bigcup_{i=0}^{n-1} \mathrm{ngrams}(y_i,n)\), tokenized by NLTK word-tokenize.
\[
\mathrm{Distinct}\text{-}n = \frac{|\mathrm{set}(\mathcal{G}_n)|}{|\mathcal{G}_n| + \varepsilon} \in [0,1]
\] 
Higher values indicate higher lexical diversity.

\paragraph{Expectation-Adjusted Distinct (EAD-\(n\)).}
A length- and vocabulary-normalized variant of Distinct-$n$, mitigating bias from long outputs. 
Define \(V\) as the $n$ power of the vocabulary size \(V\) of the aligned model's tokenizer and the union of all \(n\)-grams similarly by the aligned model's tokenizer. 
\[
V = \text{vocabulary size}^n
\]
\[
\mathrm{EAD}\text{-}n = \frac{|\mathrm{set}(\mathcal{G}_n)|}{V \cdot \left(1 - \left(\frac{V-1}{V}\right)^{|\mathcal{G}_n|}\right) + \varepsilon} \in [0,1]
\]
Higher values indicate higher lexical diversity. 

\paragraph{Self-BLEU.} Average pairwise BLEU ~\citep{papineni2002bleu}. 
For each output \(y_i\), use the other outputs \(\{y_j\}_{j \neq i}\) as references:
\[
\mathrm{Self}\text{-}\mathrm{BLEU} = \frac{1}{n} \sum_{i=0}^{n-1} \mathrm{BLEU}(y_i, \{y_j\}_{j \neq i})
\in [0,1]
\]
Lower values indicate higher lexical diversity.

\paragraph{Self-ROUGE-L.}  Average pairwise ROUGE-L scores~\citep{lin2004rouge}.  
\[
\mathrm{Self}\text{-}\mathrm{ROUGE\text{-}L} = \frac{1}{n} \sum_{i=0}^{n-1}  \mathrm{ROUGE\text{-}L}(y_i, \{y_j\}_{j \neq i}) \in [0,1]
\]
Lower values indicate higher lexical diversity.

\subsubsection{Semantic Metrics}

\paragraph{Embedding Cosine Dissimilarity.}  
Embed each \(y_i\) using a sentence embedding model (e.g., SBERT or Qwen3), obtaining \(\mathbf{e}_i\). Compute pairwise cosine distances:
\[
d_{ij} = 1 - \cos(\mathbf{e}_i, \mathbf{e}_j)
\]
\[
\mathrm{Embedding\ Diversity} = \frac{2}{n(n-1)} \sum_{i<j} d_{ij} \in [0, 1]
\]
Higher values indicate higher semantic diversity.

\paragraph{Vendi Score.} 
The exponential entropy of eigenvalues of the similarity matrix based on n-gram Jaccard overlap, capturing the effective number of independent modes.
First, construct a similarity matrix \(K \in \mathbb{R}^{n \times n}\) via either n-gram Jaccard overlap or pairwise embedding similarity, which is positive semi-definite. Let \(\lambda_1, \dots, \lambda_n\) be the eigenvalues of \(\tfrac{K}{n}\). Then:
\[
\mathrm{Vendi\ Score} = \exp\left( -\sum_{i=1}^n \lambda_i \log \lambda_i \right) \in [1, n]
\]
This is the exponential of the Shannon entropy of the normalized similarity matrix, interpretable as the effective number of distinct modes. We construct similarity matrix  based on SimCSE embeddings \citep{gao2021simcse}. 
Higher values indicate higher semantic diversity.

\paragraph{NLI Diversity.} Average contradiction probability across output pairs, computed using a RoBERTa NLI model.\footnote{\url{https://huggingface.co/sentence-transformers/nli-roberta-base-v2}}  
For each pair \((y_i, y_j)\), apply an NLI model (RoBERTa-based) to compute the entailment probability:
\[
\mathrm{NLI\ Diversity} = \frac{2}{n(n-1)} \sum_{i<j} P_{\mathrm{entailment}}(y_i, y_j) \in [0,1]
\]
Lower values (less entailment) indicate higher diversity.

\paragraph{Distinct Score (NoveltyBench).} The number of unique functional equivalence classes predicted by a DeBERTa classifier trained on human annotation;
The DeBERTa classifier is trained to predict whether two outputs are functionally equivalent. Cluster the outputs $\{y_i\}$ equivalence classes. The metric is:
\[
\mathrm{Distinctivity\ Score} = \# \{\text{unique equivalence classes among } \{y_i\}\} \in [0, n-1]
\]
Larger values indicate higher diversity.

\paragraph{Semantic Entropy.} Rao’s quadratic entropy over clusters of semantically equivalent outputs grouped via entailment and aggregated via log-likelihood. It works by first 
clustering outputs \(\{y_i\}\) into semantic groups \(\{C_1, \dots, C_k\}\) using entailment-based NLI. Then compute cluster-level probabilities using likelihoods:
\[
\log p(C_k) = \log \left( \sum_{y_i \in C_k} \exp(\log p(y_i)) \right)
\]
Finally:
\[
\mathrm{Semantic\ Entropy} = -\sum_{k} p(C_k) \log p(C_k) \in [0, \log n]
\]
Larger values indicate higher diversity.

\subsection{Aggregate Metrics over Quality--Diversity Spaces}
\label{app:aggregate-metrics}

Let a \emph{space} be defined by a pair of metrics \((m_x, m_y)\), where \(m_x\) measures quality (higher is better) and \(m_y\) measures diversity (either higher or lower is better, depending on the metric). Varying a method’s control parameter (e.g., decoding temperature, routing threshold) traces a set of points \(\{(x_t, y_t)\}\) in this space.

\paragraph{Feasible Region and Normalization.}
To make values comparable across metrics, we normalize each space to the unit square \([0,1]^2\).
Let \(\mathcal{F}=[x_{\min},x_{\max}]\times[y_{\min},y_{\max}]\) denote the feasible region, anchored using two reference operating points at temperature \(1.0\): the \emph{base} model and the \emph{aligned} model.\footnote{Concretely, \(x_{\min}\) is set to the base model’s quality at \(T=1.0\); \(x_{\max}\) to the aligned model’s quality at \(T=1.0\). For the diversity axis, if higher is better we set \(y_{\min}\) to the aligned model’s diversity at \(T=1.0\) and \(y_{\max}\) to the theoretical maximum (e.g., \(\log N\) for Semantic Entropy with \(N\) samples). If lower is better, we set \(y_{\min}\) to the theoretical lower bound and \(y_{\max}\) to the aligned model’s diversity at \(T=1.0\).}
Observed points are normalized via:
\[
\hat{x}=\frac{x-x_{\min}}{x_{\max}-x_{\min}},\qquad
\hat{y}=
\begin{cases}
\dfrac{y-y_{\min}}{y_{\max}-y_{\min}}, & \text{if higher is better},\\[6pt]
1-\dfrac{y-y_{\min}}{y_{\max}-y_{\min}}, & \text{if lower is better}.
\end{cases}
\]
Points outside \(\mathcal{F}\) are discarded for aggregation, as the outputs of the represented setting might have limited usage. It has no strength, in terms of the metrics, compared with the two single-model baselines.

\textbf{Coverage (Cov.)} measures how effectively a method traverses the \problem{} as its control parameters vary (\eg{} decoding temperature of single-model baselines and threshold for \name{} routers). 
The indicator is simplified from Hypervolume (HV) \citep{10.1007/s00158-016-1469-3} in multiobjective optimization problems. 
Concretely, we normalize each space into a unit square (anchored by the default baseline: base and aligned models at temperature 1.0), and compute the area under the curve (AUC) traced by the method’s normalized points. 
Higher Coverage values indicate greater controllability, general good performance across different trade-off balances, and robustness across parameters.
However, Coverage does not capture whether a method is \emph{ever} optimal across different trade-off balances.

For method \(k\), we consider the piecewise-linear curve obtained from its normalized points \(\{(\hat{x}_t,\hat{y}_t)\}\) (ordered by \(\hat{x}\)), augmented with boundary points to close the curve inside \([0,1]^2\). We define:
\[
\mathrm{Coverage}_k(m_x,m_y)=\int_{0}^{1}\hat{y}_k(\hat{x})\,d\hat{x},
\]
computed using the trapezoidal rule. Because the domain is fixed to \([0,1]\), \(\mathrm{Coverage}\in[0,1]\). Higher values indicate that the method maintains strong quality and diversity as its control parameters vary.

\textbf{Dominance (Dom.)} complements Coverage by capturing whether a method ever achieves optimality relative to others. We utilize the C-metric \citep{zitzler1999evolutionary} to evaluate Dominance of pairwise comparison, which captures the portion of the frontier that one method dominates over the other one. 
In our problem, the portion is in terms of intervals along the diversity (denoted as -D) or quality (denoted as -Q) axes. Dom takes the harmonic mean of Dom-D and Dom-Q.
For global comparison across all methods, we compute the global Pareto frontier across all methods. We apply the C-metric between each method and the global frontier, equivalent to the portion of the frontier attributed to the method.  

We compute the global Pareto frontier \(\mathcal{P}\) over the union of all methods’ normalized points in a space. For each Pareto point, we assign an \emph{interval of responsibility} along the diversity or quality axis by splitting at midpoints between adjacent frontier points. Summing these interval lengths for Pareto points contributed by method \(k\) yields its coverage along that axis, normalized by the total frontier span:
\[
\mathrm{Dom\text{-}D}_k(m_x,m_y),\quad
\mathrm{Dom\text{-}Q}_k(m_x,m_y)\;\in[0,1].
\]
We report a single Dominance score as their {harmonic mean}:
\[
\mathrm{Dom}_k(m_x,m_y)=
\frac{2 \cdot \mathrm{Dom\text{-}D}_k \cdot \mathrm{Dom\text{-}Q}_k}{\mathrm{Dom\text{-}D}_k + \mathrm{Dom\text{-}Q}_k}
\]

\paragraph{Holistic aggregation.}
Since quality and diversity admit multiple measurements, we average over all spaces \(\mathcal{S}\) to obtain metric-agnostic summaries:
\[
\overline{\mathrm{Cov}}_k = \frac{1}{|\mathcal{S}|}\sum_{(m_x,m_y)\in\mathcal{S}}\mathrm{Cov}_k(m_x,m_y),\qquad
\overline{\mathrm{Dom}}_k = \frac{1}{|\mathcal{S}|}\sum_{(m_x,m_y)\in\mathcal{S}}\mathrm{Dom}_k(m_x,m_y).
\]

%% file: past_papers/base_aligned_collaboration/sections/appendix/detail_results.tex
\label{app:detail_results}

\subsection{Instruction Following on NoveltyBench}
\label{app:detail_results:nb}

\Tabref{tab:noveltybench} shows the result on NoveltyBench of comparing \name{} on the best router compared with baselines. 

\input{past_papers/base_aligned_collaboration/tables/sub_novelty_bench}

\textbf{Results.} \name{} outperforms all baselines on all metrics except lexical Dominance. 
Compared with all baselines, \name{} improves Coverage by \textbf{0.274} overall (0.222 lexical, 0.291 semantic).
It dominates \textbf{39.9\%} (the most) of the diversity-quality frontier overall (31.0\% lexical, 48.8\% semantic). 

Beyond the LLaMA-3 base–aligned pair, we also validate that \name{} consistently outperforms baselines on another model family, \texttt{Olmo2}. The results are reported in \Cref{tab:nb:olmo2}.

\subsection{Dialogue on WildChat}

WildChat involves naturally complex and nuanced prompts, leading to much longer outputs on average compared with NoveltyBench. \Cref{tab:wildchat} summarizes the results.  

\input{past_papers/base_aligned_collaboration/tables/sub_wildchat}

\textbf{Results.} The superiority of \name{} persists on WildChat. Compared with the aligned model baseline, \name{}-\textsc{P-Punc} improves Coverage by 29.8\% and dominates 30.8\% of the frontier. 
Moreover, \name{} demonstrates a particularly strong advantage in semantic diversity, where it dominates 48.5\% of the frontier. 
These findings confirm that base–aligned collaboration scales effectively from short-form prompts (NoveltyBench) to longer, more conversational dialogue.  

\subsection{Creative Writing on Narrative-Discourse}

We further evaluate \name{} on Narrative-Discourse to test its ability to generate structure-diverse and long-term coherent narratives. 
This dataset emphasizes sustained creativity and narrative arc, placing distinct demands beyond instruction following and dialogue. 

\input{past_papers/base_aligned_collaboration/tables/sub_storyarc}

\textbf{Results.} 
As shown in \Cref{tab:storyarc}, \name{} again outperforms all baselines. It achieves 13.5\% higher Coverage and dominates 20.3\% of the overall diversity–quality frontier.

\input{past_papers/base_aligned_collaboration/tables/sub_novelty_bench_full}

\subsection{Extensional Tasks Results}

\begin{figure}
    \centering
    \includegraphics[width=1.0\linewidth]{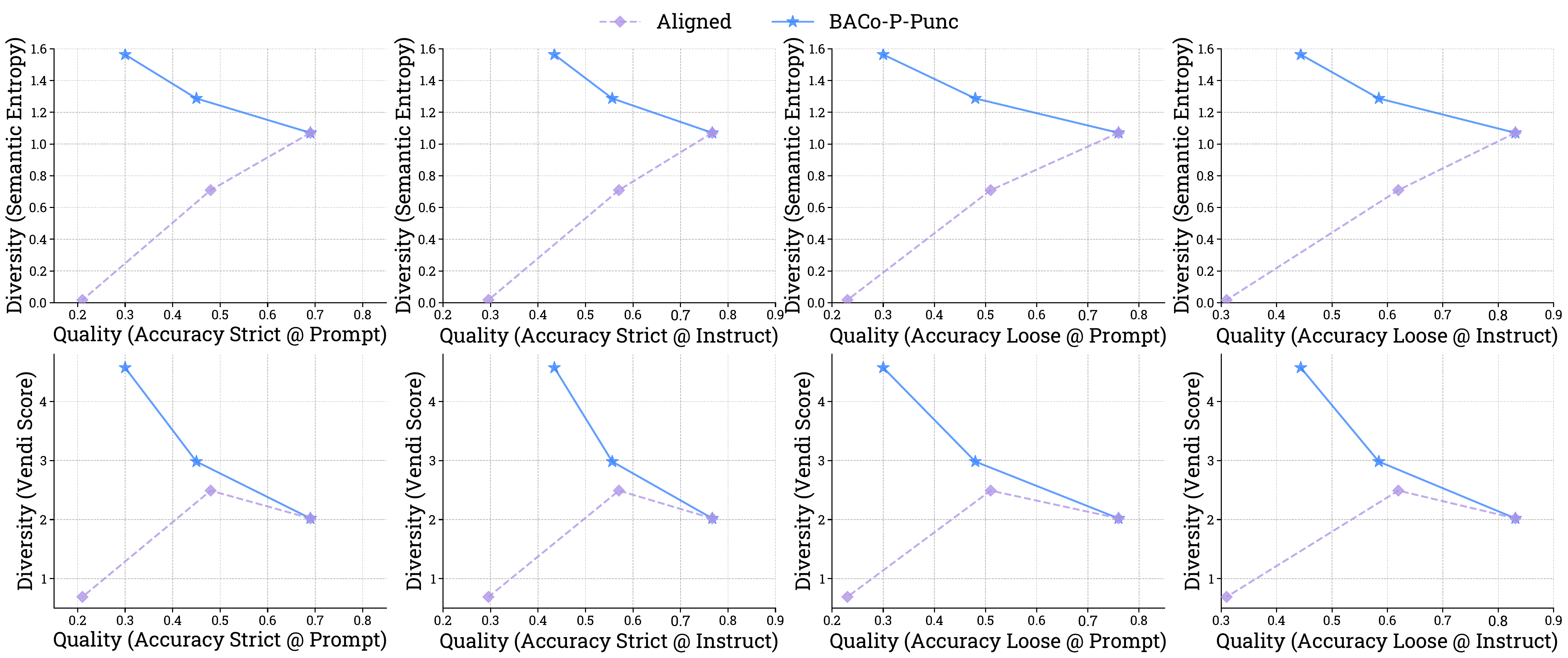}
    \caption{Diversity-quality trade-off comparison between \name{} and the aligned model baseline on IFEval. \name{} demonstrates a more effective optimization of the trade-off. The table presents all the trade-off spaces tested.}
    \label{fig:ifeval_large_figure}
\end{figure}

\begin{figure}
    \centering
    \includegraphics[width=0.5\linewidth]{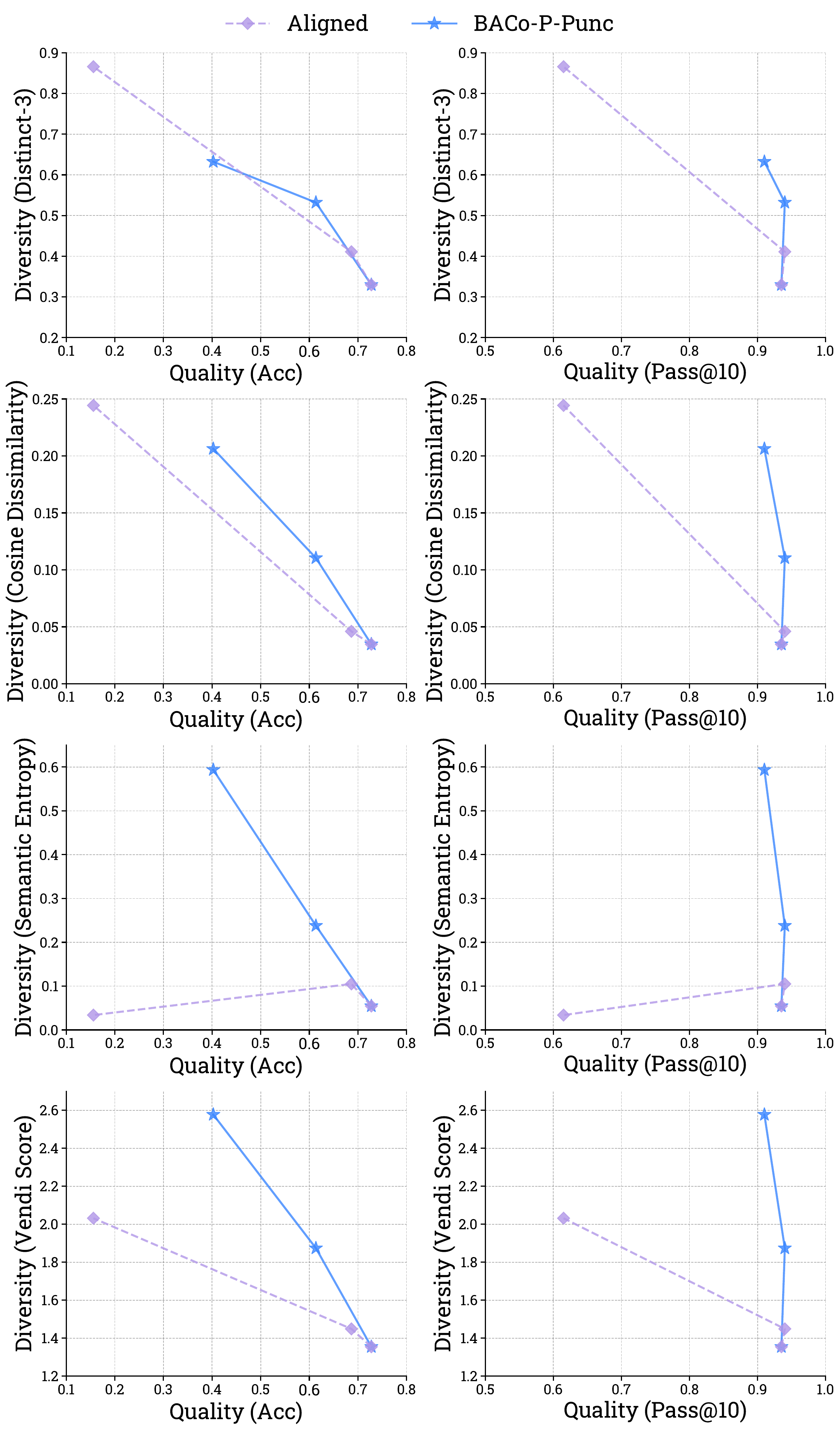}
    \caption{Diversity-accuracy trade-off comparison between \name{} and the aligned model baseline on GSM8K. \name{} demonstrates a more effective optimization of the trade-off. The table presents all the trade-off spaces tested.}
    \label{fig:gsm8k_large_figure}
\end{figure}

We compare \name{}\textsc{-P-Punc} (i.e., our best router) with the aligned model baseline by adjusting temperature.
\Cref{fig:ifeval_large_figure} presents the full results of the verifiable instruction following task on the IFEval dataset, and \Cref{fig:gsm8k_large_figure} presents the full results of the mathematical reasoning task on the GSM8K dataset.

%% file: past_papers/base_aligned_collaboration/tables/sub_novelty_bench.tex
\begin{table}[H]
\centering
\small
\renewcommand{\arraystretch}{1.2}

\begin{tabular}{lcccccc}
\toprule
\textbf{Method} & \multicolumn{2}{c}{\textbf{Lexical}} & \multicolumn{2}{c}{\textbf{Semantic}} & \multicolumn{2}{c}{\textbf{Overall}} \\
 & \textit{Cov.} & \textit{Dom.} & \textit{Cov.} & \textit{Dom.} & \textit{Cov.} & \textit{Dom.} \\
 \midrule
Base & 0.142 & 9.8\% & 0.142 & 13.1\% & 0.142 & 11.4\% \\
Aligned & 0.273 & \textbf{40.1\%} & 0.128 & 17.2\% & 0.200 & 28.6\% \\
Nudging & 0.192 & 6.8\% & 0.161 & 7.6\% & 0.176 & 7.2\% \\
Decoding & - & 0.8\% & - & 1.0\% & - & 0.9\% \\
Prompting best & - & 8.0\% & - & 6.5\% & - & 7.3\% \\
Ensemble best & - & 3.4\% & - & 5.8\% & - & 4.6\% \\
\midrule
\name{} best & \textbf{0.495} & 31.0\% & \textbf{0.452} & 
\textbf{48.8\% }& \textbf{0.474} & \textbf{39.9\%} \\

\toprule
\end{tabular}
\caption{Comparison results on NoveltyBench. For space-saving, we present the best method in each category.}
\label{tab:noveltybench}
\end{table}

%% file: past_papers/base_aligned_collaboration/tables/sub_wildchat.tex
\begin{table}[H]
\setlength{\intextsep}{0pt}
\captionsetup[table]{skip=2pt,aboveskip=2pt,belowskip=0pt}
\centering
\small
\renewcommand{\arraystretch}{1.2}
\begin{tabular}{l cc cc cc cc}
\toprule
\textbf{Method} & \multicolumn{2}{c}{\textbf{Lexical}}& \multicolumn{2}{c}{\textbf{Semantic}} & \multicolumn{2}{c}{\textbf{Overall}} \\
 & \textit{Cov.} & \textit{Dom.} & \textit{Cov.} & \textit{Dom.} & \textit{Cov.} & \textit{Dom.} \\
\midrule
Base & 0.000	& 1.9\%	& 0.000	& 6.8\%	& 0.000	& 4.38\% \\
Aligned & 0.253	& \textbf{59.2\%}	& 0.077	& 29.1\%	& 0.165	& \textbf{44.1\%}\\
Nudging & 0.430	& 11.4\%	& 0.387	& 15.6\%	& 0.408	& 13.5\%  \\

\midrule
\name{} best & \textbf{0.473}	& 27.4\%	& \textbf{0.454} & \textbf{48.5\%}	& \textbf{0.463}	& 38.0\% \\
\toprule
\end{tabular}
\caption{Comparison results on WildChat. For space saving, we present the best router, -\textsc{P-Punc}, as \name{}'s representative.}
\label{tab:wildchat}
\end{table}

%% file: past_papers/base_aligned_collaboration/tables/sub_storyarc.tex
\begin{table}
\centering
\small
\renewcommand{\arraystretch}{1.2}

\begin{tabular}{lcccccc}
\toprule
\textbf{Method} & \multicolumn{2}{c}{\textbf{Lexical}} & \multicolumn{2}{c}{\textbf{Semantic}} & \multicolumn{2}{c}{\textbf{Overall}} \\
 & \textit{Cov.} & \textit{Dom.} & \textit{Cov.} & \textit{Dom.} & \textit{Cov.} & \textit{Dom.} \\
\midrule
Base & 0.151 & 26.3\% & 0.153 & 28.1\% & 0.152 & 27.2\% \\
Aligned & 0.282 & \textbf{47.7\%} & 0.106 & \textbf{41.2\% }& 0.194 & \textbf{44.4\%} \\
Nudging & 0.205 & 9.7\% & 0.194 & 6.5\% & 0.199 & 8.1\% \\
\midrule
\name{} best & \textbf{0.367} & 16.3\% & \textbf{0.174} & 24.2\% & \textbf{0.271} & 20.3\% \\
\bottomrule
\end{tabular}

\caption{Comparison results on Narrative-Discourse. For space saving, we present the best router, -\textsc{P-Punc}, as \name{}'s representative.}
\label{tab:storyarc}
\end{table}

%% file: past_papers/base_aligned_collaboration/tables/sub_novelty_bench_full.tex
\begin{table}[]
    \small
    \centering
    \renewcommand{\arraystretch}{1.2}
\begin{tabular}{lcccccc}

\toprule
\textbf{Method} & \multicolumn{2}{c}{\textbf{Lexical}} & \multicolumn{2}{c}{\textbf{Semantic}} & \multicolumn{2}{c}{\textbf{Overall}} \\
 & \textit{Cov.} & \textit{Dom.} & \textit{Cov.} & \textit{Dom.} & \textit{Cov.} & \textit{Dom.} \\
\toprule
Base & 0.142 & 7.6\% & 0.142 & 10.9\% & 0.142 & 9.2\% \\
Aligned & 0.273 & 36.5\% & 0.128 & 15.8\% & 0.200 & 26.1\% \\
\midrule
In-context Prompt & - & 0.0\% & - & 2.2\% & - & 1.1\% \\
Paraphrase Prompt & - & 8.0\% & - & 5.6\% & - & 6.8\% \\
Diverse BS Decoding & - & 0.8\% & - & 1.0\% & - & 0.9\% \\
\midrule
Response Ensemble & - & 3.4\% & - & 3.7\% & - & 3.6\% \\
Logits Ensemble & - & 0.0\% & - & 0.0\% & - & 0.0\% \\
Nudging & 0.192 & 4.9\% & 0.161 & 4.0\% & 0.176 & 4.5\% \\
\midrule
\name{} All & 0.495 & 39.0\% & 0.452 & 56.8\% & 0.474 & 47.9\% \\
\midrule
\midrule
\name{}-\textsc{Judge} & 0.302 & 0.5\% & 0.254 & 0.1\% & 0.278 & 0.3\% \\
\name{}-\textsc{Rand} & 0.493 & 13.1\% & 0.409 & 5.9\% & 0.451 & 9.5\% \\
\midrule
\name{}-\textsc{FC} & 0.419 & 2.4\% & 0.382 & 4.0\% & 0.401 & 3.2\% \\
\name{}-\textsc{P} & 0.433 & 2.6\% & 0.397 & 7.7\% & 0.415 & 5.2\% \\
\midrule
\name{}-\textsc{P-Punc} & 0.495 & 11.4\% & 0.452 & 17.2\% & 0.474 & 14.3\% \\
\name{}-\textsc{H-Punc} & 0.466 & 5.9\% & 0.427 & 11.0\% & 0.446 & 8.4\% \\
\name{}-\textsc{P-FC} & 0.435 & 3.1\% & 0.406 & 10.9\% & 0.421 & 7.0\% \\

\toprule
\end{tabular}
    \caption{Comparison of all methods (baselines and \name{} routers) on NoveltyBench. \name{} All reports the best \textit{Cov.} across all routers and the \textit{Dom.} sum over all routers. The lower half of the table provides the performance of individual routers. Note that routers distribute \textit{Dom.} values given to the metric definition; therefore, \textit{Dom.} values in the top and bottom halves of the table are not directly comparable.}
    \label{tab:novelty-bench-full}
\end{table}

%% file: past_papers/base_aligned_collaboration/tables/sub_novelty_bench_olmo.tex
\begin{table}[ht]
\centering
\small
\renewcommand{\arraystretch}{1.2}
\begin{tabular}{lcccc}
\toprule
\textbf{Method} & \textbf{Overall} \textit{Cov.} & \textbf{Overall} \textit{Dom.} \\
\midrule
Base & 0.098 &  11.7\% \\
Aligned  & 0.209 &  \underline{30.5\%} \\
Nudging & \textbf{0.281} &  5.5\% \\
Others & - & 8.3\% \\
\midrule
\name-\textsc{P-Punc} & \underline{0.236} & \textbf{44.0\%} \\
\bottomrule
\end{tabular}
\caption{Results of \name{} on \texttt{Olmo2-7B} and \texttt{Olmo2-7B-Instruct} on NoveltyBench.}
\label{tab:nb:olmo2}
\end{table}

%% file: past_papers/base_aligned_collaboration/sections/appendix/storyarc.tex
\subsection{Creative Writing Task Setup}

\label{app:storyarc}

We frame creative writing as a continuation task, where the model is given the beginning of a story and asked to complete it.
The prefix contains events leading up to the first turning point, which introduces the initial situation or conflict setting the stage for the narrative.
The model then generates the subsequent events to develop and conclude the entire narrative.
To capture discourse-level variation, we measure two structural dimensions:  

\circone~ \emph{Turning-point diversity} quantifies differences in the relative positions of annotated plot inflections across outputs.  

\circtwo~ \emph{Arousal diversity} tracks divergence in emotional trajectories, obtained by sampling sentence-level arousal scores and comparing smoothed curves via KL divergence.

Together, these metrics provide complementary measures of long-form diversity, capturing variation in plot structure and affective dynamics that conventional surface-level metrics miss. 
Prompt is shown in \Cref{tab:storyarc-prompt}. Details of the dataset and annotation schema follow \citet{tian2024large}.

\begin{table}[H]
\small
\begin{tabular}{M{0.95\linewidth}}
\toprule
{\raggedright\ttfamily
``role'': ``user'', ``content'': Continue the story and bring it to an ending based on the title and the story sketch provided below. The sketch introduces the event that sets the initial stage for the narrative leads up to the first major turning point—but does not present a full plot. Your task is to develop the narrative from this point onward, completing the story arc. \\
Title: \{title\} \\
Story Sketch: \{sketch\}
} \\
\bottomrule \\
\end{tabular}
\caption{Generation prompt for the creative writing task.}
\label{tab:storyarc-prompt}
\end{table}

\subsection{Narrative-Discourse Evaluation Metrics}
For \textit{turning points}, each generated narrative \(y\) is segmented into sentences, with total length \(L\). The relative position of the turning point \(k\) annotated is
$r_k\!\left(y\right) = \frac{\text{Index}_{tp_k}\!\left(y\right)}{L}, \quad r_k\!\left(y\right) \in [0,1].
$
For a group of $n$ outputs \(\{y^{(1)}, \dots, y^{(n)}\}\), we compute pairwise distances:
\[
D_{\text{TP}}\!\left(y^{(i)}, y^{(j)}\right) = \frac{1}{K} \sum_{k=1}^K \left| r_k\!\left(y^{(i)}\right) - r_k\!\left(y^{(j)}\right) \right|,
\]
where \(K=5\) is the number of turning points. The turning-point diversity score is then
\[
\text{TP-Div} = \frac{2}{n(n-1)} \sum_{i<j} D_{\text{TP}}\!\left(y^{(i)}, y^{(j)}\right).
\]

For \textit{arousal}, we sample sentences at fixed intervals from each \(y^{(i)}\) and obtain arousal scores via LLM-as-a-judge. Let \(a_t\!\left(y^{(i)}\right)\) denote the arousal score at sampled position \(t\). We fit a smooth trajectory \(\hat{a}\!\left(y^{(i)}\right)\) via polynomial interpolation. For two narratives \(y^{(i)}\) and \(y^{(j)}\), their affective divergence is
\[
D_{\text{Arousal}}\!\left(y^{(i)}, y^{(j)}\right) = \mathrm{KL}\!\left(\hat{a}\!\left(y^{(i)}\right) \;\|\; \hat{a}\!\left(y^{(j)}\right)\right).
\]
The overall arousal diversity is
\[
\text{Arousal-Div} = \frac{2}{n(n-1)} \sum_{i<j} D_{\text{Arousal}}\!\left(y^{(i)}, y^{(j)}\right).
\]

%% file: past_papers/base_aligned_collaboration/sections/appendix/analysis.tex
\section{Additional Analysis Material}
\label{app:analysis}

\begin{figure}[t]
    \begin{subfigure}[t]{0.45\textwidth}
        \includegraphics[width=\linewidth]{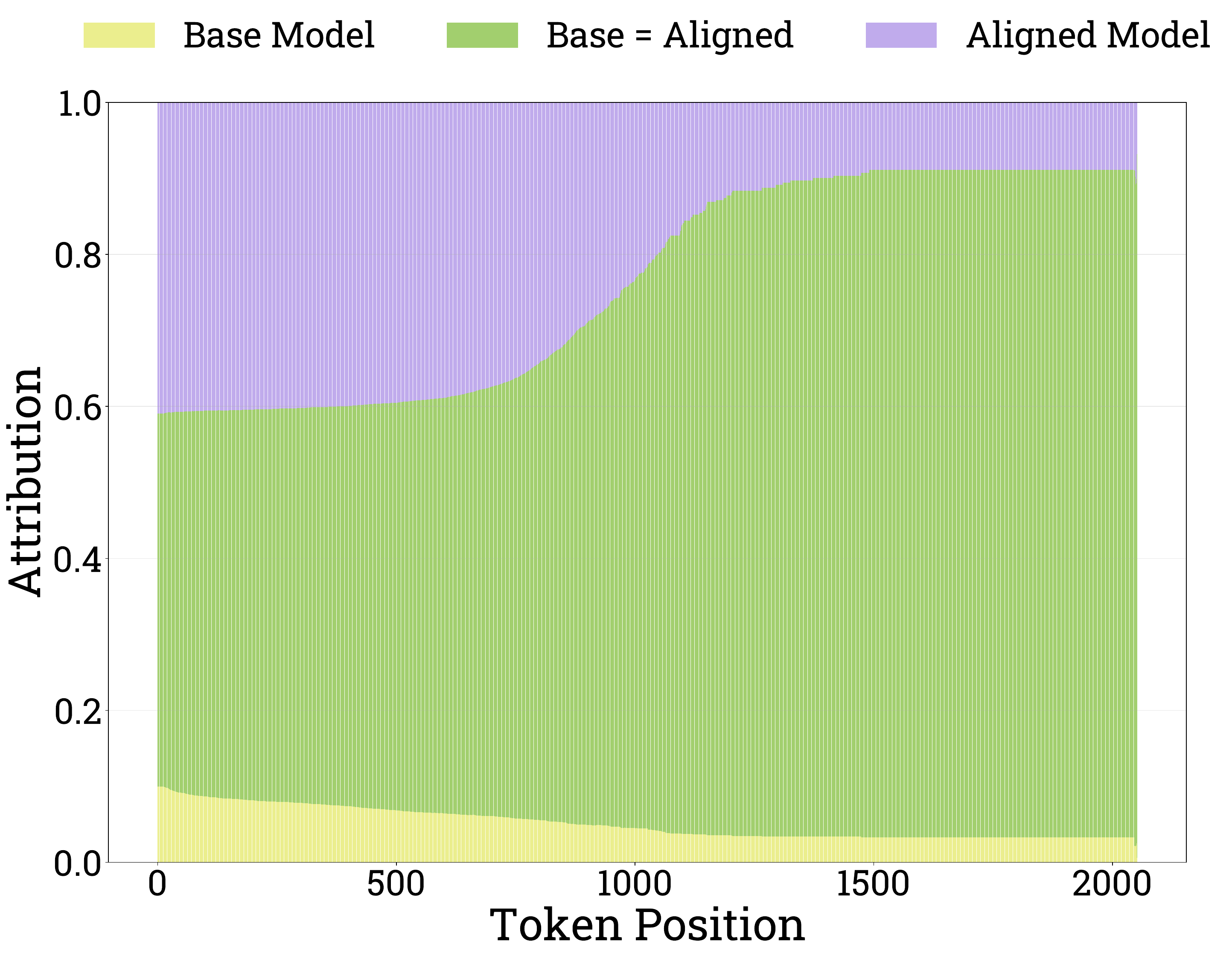}
        \caption{Model Contribution Distribution}
    \end{subfigure}
    \begin{subfigure}[t]{0.45\textwidth}
        \includegraphics[width=\linewidth]{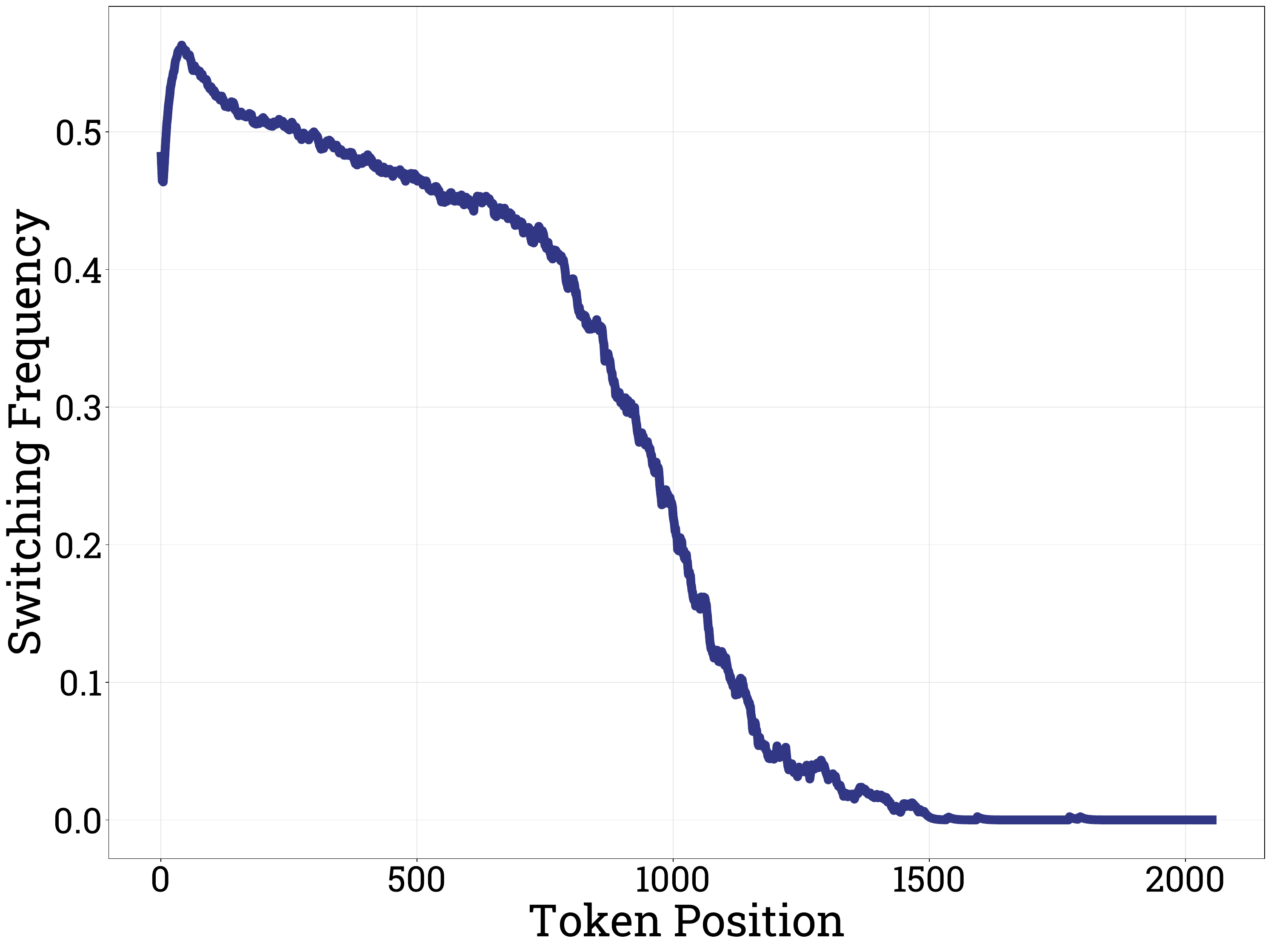}
        \caption{Switching Frequency}
    \end{subfigure}
    \caption{Contribution distribution and switching frequency for \name{} with
    the best router (\textsc{-P-Punc}) at Narrative-Discourse dataset.}
    \label{fig: switching_storyarc}
\end{figure}

\begin{figure}[t]
    \begin{subfigure}[t]{0.45\textwidth}
        \includegraphics[width=\linewidth]{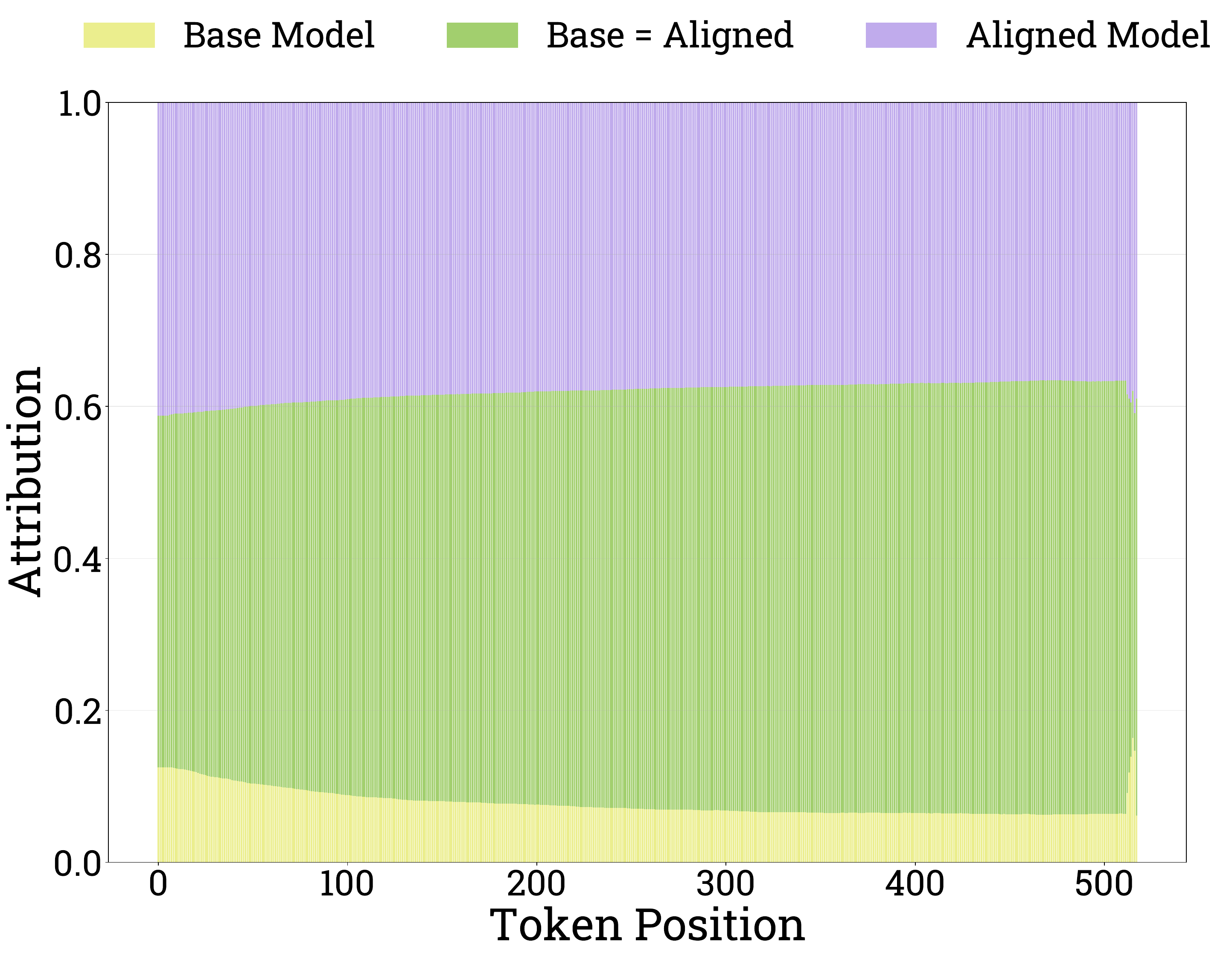}
        \caption{Model Contribution Distribution}
    \end{subfigure}
    \begin{subfigure}[t]{0.45\textwidth}
        \includegraphics[width=\linewidth]{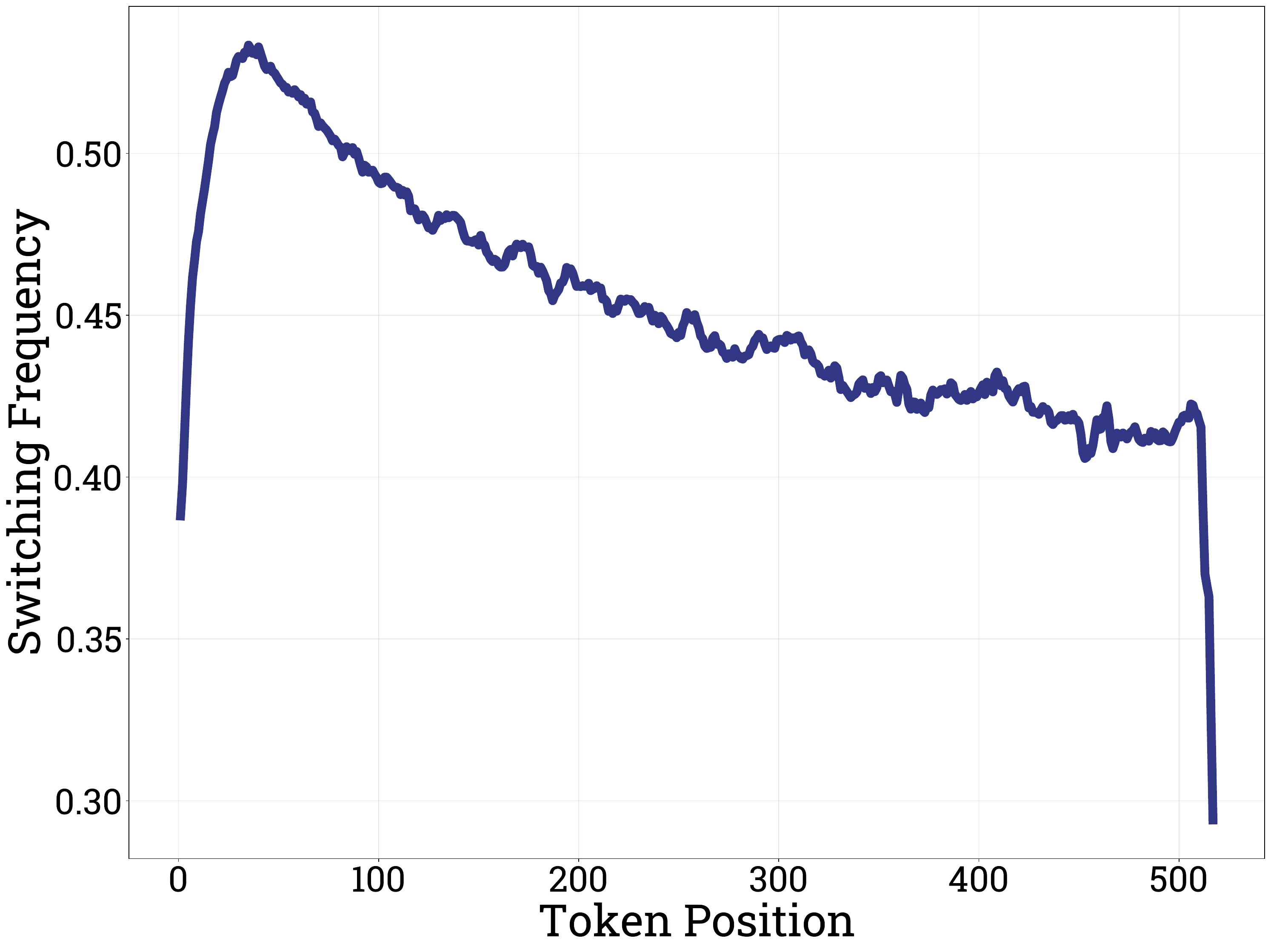}
        \caption{Switching Frequency}
    \end{subfigure}
    \caption{Contribution distribution and switching frequency for \name{} with
    the best router (\textsc{-P-Punc}) at WildChat dataset.}
    \label{fig: switching_wildchat}
\end{figure}

\begin{figure}[t]
    \begin{subfigure}[t]{0.45\textwidth}
        \includegraphics[width=\linewidth]{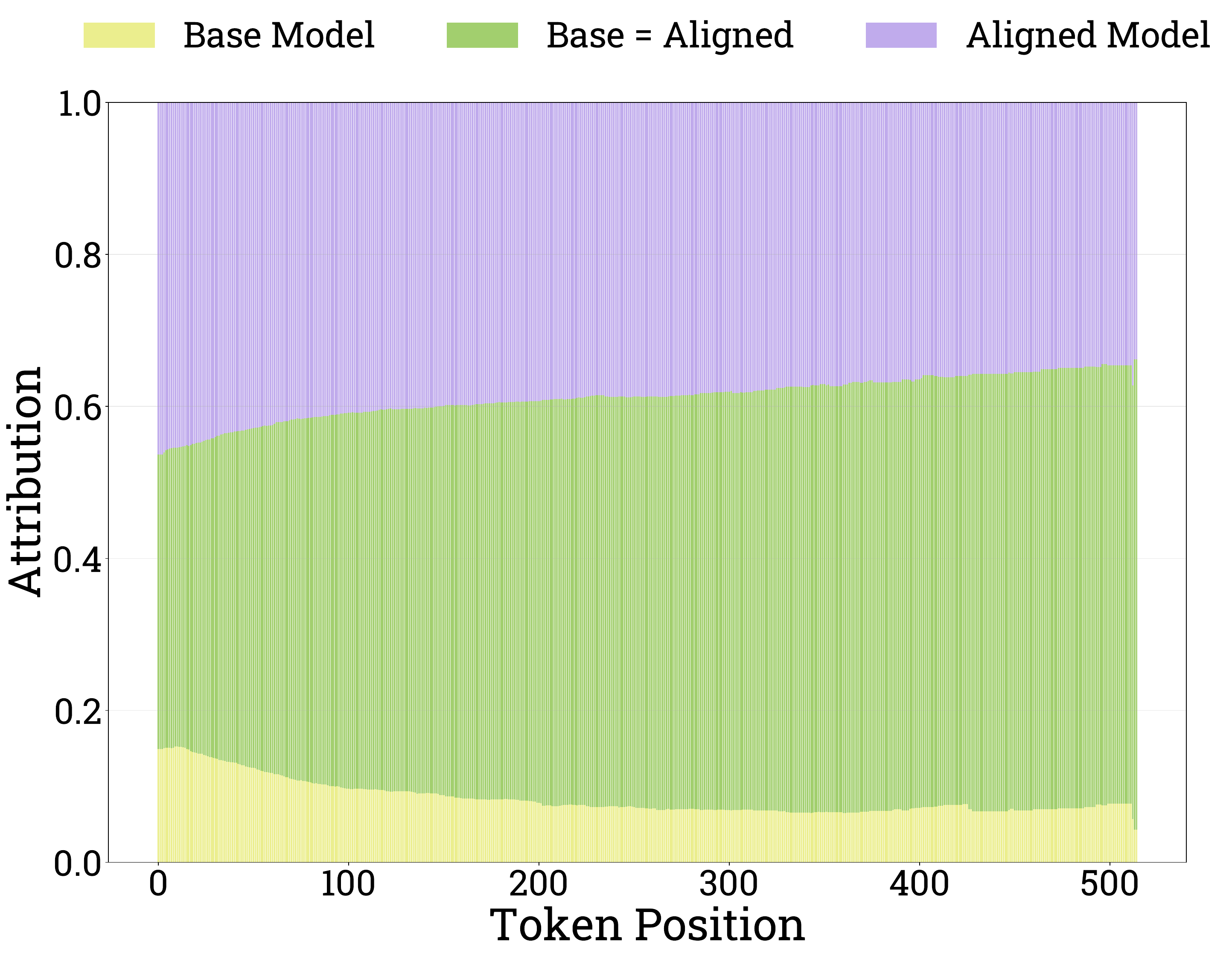}
        \caption{Model Contribution Distribution}
    \end{subfigure}
    \begin{subfigure}[t]{0.45\textwidth}
        \includegraphics[width=\linewidth]{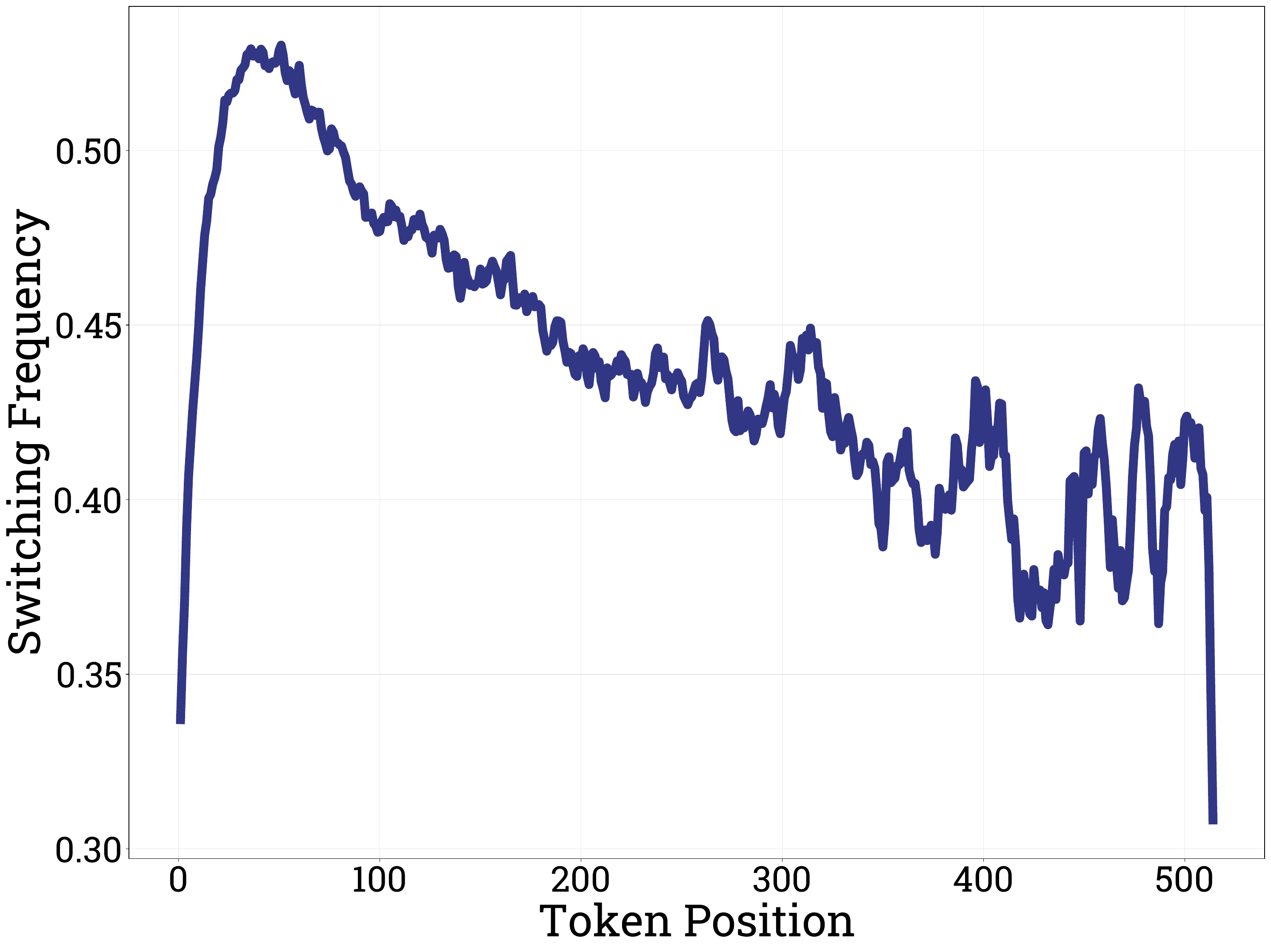}
        \caption{Switching Frequency}
    \end{subfigure}
    \caption{Contribution distribution and switching frequency for \name{} with
    the best router (\textsc{-P-Punc}) at NoveltyBench dataset.}
    \label{fig: switching_nb}
\end{figure}

This section provides supporting material for the analysis in \Cref{sec:analysis}.
\Cref{fig: switching_storyarc,fig: switching_wildchat,fig: switching_nb} report
the per-token model contribution distribution and switching frequency for
\name{}-\textsc{P-Punc} on Narrative-Discourse, WildChat, and NoveltyBench,
respectively, supporting the temporal-pattern observation in
\Cref{sec:anal:distribution}.
\Cref{tab:early_stop,tab:dynamic_example} provide qualitative examples
illustrating, respectively, the inherent early-stopping failure mode discussed
in \Cref{sec:earlystop} and how outputs evolve as the routing threshold $\gamma$
traverses the diversity-quality spectrum.
Additional future work directions beyond those discussed in \Cref{sec:future_work}
are provided at the end of this section.

\input{past_papers/base_aligned_collaboration/tables/early_stop}

\input{past_papers/base_aligned_collaboration/tables/dynamic_example}

\subsection{Extended Future Work}
\label{app:future_work}

Beyond the directions outlined in \Cref{sec:future_work}, we discuss two
further avenues.

\paragraph{Efficiency.}
\name{} currently requires two forward passes per decoding step in the worst
case.
Speculative decoding~\citep{leviathan2023fastinferencetransformersspeculative}
fits naturally with token-level routing: by looking ahead, it can reduce
inference-time overhead without changing the routing logic.
Moreover, since base and aligned models share the same architecture in our
experiments, memory-efficient alternatives such as LoRA-tuned aligned
models~\citep{wu2024mixtureloraexperts} could substantially reduce deployment
costs---replacing $N{\times}$ model storage with a single base model plus
$N{\times}$ LoRA adapter size---while preserving the diversity--quality contrast
between checkpoints.

\paragraph{Checkpoint Exploration.}
Base and fully aligned models are the two most accessible checkpoints on the
alignment trajectory, but they are unlikely to represent the optimal
diversity--quality trade-off
points~\citep{im2024understanding,ren2025learningdynamicsllmfinetuning}.
Intermediate or partially aligned checkpoints may offer better operating points,
and a systematic study of alignment dynamics could reveal more effective
collaborator pairs and further extend the \problem{} frontier.
More broadly, this suggests a general principle: inference-time collaboration
need not be limited to canonical checkpoints, and richer exploration of the
training trajectory is a promising direction.

%% file: past_papers/base_aligned_collaboration/tables/early_stop.tex
\small{
\begin{longtable}{M{0.95\linewidth}}
\toprule
\textbf{Prompt:} Tell me a dad joke.\\
\midrule
\midrule
\textbf{Output ($\gamma=0.4$):}
\\
Sample 1: \\
Here's some "dad" jokes: Why did the scarecrow win an award? Because he was outstanding in his field! \\
\\
\midrule 
Sample 2: \\
Here's the groan-inducing, corny, and utterly predictable dad joke for you: Why was 5 afraid of 7? Because 7 8 9! (get it?) \\
\midrule
... \\
\midrule
Sample 4: \\
Here's one for ya. \textit{(Early stopped)}\\
\midrule
... \\
\midrule
Sample 10: \\
Here's a groan-inducing dad joke for you: Why are eggs like household appliances? Because they're both cracking under the pressure! \\
\bottomrule
\caption{Among 10 sampled generations, one (sample 4) is early stopped.
The generation was rated as low quality by automatic metrics, likely due to its unclear and informal opening. After producing “ya .”, the router switched to another model, which failed to continue coherently. Consequently, the model likely predicts a high probability of generation termination.
\texttt{-P} serves as the router.
}
\label{tab:early_stop}
\end{longtable}
}

%% file: past_papers/base_aligned_collaboration/tables/dynamic_example.tex
{\small{
\begin{longtable}{M{0.95\linewidth}}
\toprule
\textbf{Prompt:} Tell me a story in five sentences about a girl and her dog. \\
\midrule
\midrule
\textbf{Output ($\gamma=1.0$, base-only):}
\\
Here is a story about a girl and her dog. \\
Once upon a time there was a girl named Mary. She had a dog named Spot. One day Mary and Spot went to the park. They played with a ball. \textit{(46 tokens)}
\\
\midrule 
\textbf{Output ($\gamma=0.9$):}
Here is a story about a girl and her dog. \\
Once upon a time there was a girl named Mary. She had a dog named Spot who was very friendly. One day Mary and Spot went to the park. \textit{(45 tokens)}\\
\midrule
\textbf{Output ($\gamma=0.6$):}
Here is a story about a girl and her dog. \\
Once upon a time there was a girl named Mary. She had a dog named Spot who was very friendly. One day Mary was walking Spot in the park when she saw a boy named John. John was playing with a ball. \textit{(59 tokens)}\\
\midrule
\textbf{Output ($\gamma=0.3$):} \\
Here is a story about a girl and her dog: \\
The sun was setting. Lily was walking her dog, Max, through the park. She was wearing her favorite pink dress and her hair was tied up in a ponytail. Max was a golden retriever with a big smile on his face. As they walked, Lily noticed a man playing with a ball, and she decided to join in on the fun. \textit{(85 tokens)}\\
\midrule
\textbf{Output ($\gamma=0.1$):} \\
Here is a story about a girl and her dog: \\
The sun was setting over the small town of Dogwood, casting a warm orange glow over the streets and homes. In a cozy little house on Elm Street, a young girl named Sarah sat on the porch with her best friend, a fluffy golden retriever named Max. Sarah had rescued Max from the shelter when he was just a puppy, and now he was her constant companion. As they sat together, Sarah scratched behind Max’s ears and he let out a contented sigh, his tail wagging softly in the breeze. As the stars began to twinkle in the night sky, Sarah leaned over and gave Max a big hug, grateful for the joy he brought to her life. \textit{(150 tokens)}\\
\midrule
\textbf{Output ($\gamma=0.0$, aligned-only):} \\
Here is a story about a girl and her dog: \\
Lily was a shy and gentle girl who loved nothing more than spending time with her loyal dog, Max. Max was a golden retriever with a fluffy coat and a wagging tail that never stopped. Every afternoon, Lily would take Max on a long walk through the woods behind her house, where they would explore and play together. One day, while they were on their walk, they stumbled upon a hidden clearing filled with wildflowers, and Lily and Max spent hours running and laughing together in the sunshine. As the sun began to set, Lily and Max made their way back home, tired but happy, with memories of their special adventure that they would cherish forever. \textit{(146 tokens)}
\\
\bottomrule
\caption{Example outputs from \name{} under different routing thresholds~$\gamma$. 
Lower $\gamma$ values increase routing to the aligned model, yielding longer outputs. High $\gamma$ gives a high chance route to the base model, whose outputs are more likely to be early stopped, \ie{} incomplete and lower quality (\eg{} $\gamma=0.9, 0.6$).
In this example, \texttt{-P} serves as the router. Greedy decoding is applied here for clarity of the change. The trend still holds for other sampling and other routers.
}
\label{tab:dynamic_example}
\end{longtable}
}
}

%% file: past_papers/base_aligned_collaboration/sections/appendix/human_eval.tex
\normalsize
\label{app:human_eval}

\subsection{Setup}  
We compare \name{} against the aligned model baseline, controlling for quality to ensure fairness. 
Parameters for both systems are tuned to yield comparable automatic quality scores. 
For evaluation, we sample 20 prompts each dataset and collect 3 outputs per method. 
To avoid cognitive overload from excessively long outputs\footnote{Models occasionally produce list-style outputs for some prompts, which make it difficult for annotators to remember details and to assess diversity across samples.}, we stratify prompts by average response length and sample from bins with shorter outputs, while maintaining comparable automatic diversity and quality scores. 
Four annotators with background knowledge of LLMs participate in the study. Evaluations are conducted on Novelty-Bench and WildChat.\footnote{We exclude Narrative-Discourse due to the excessive length of outputs. The huge cognition load makes group-wise human comparison infeasible.}
Interface and option design follow LMArena \citep{chiang2024chatbotarenaopenplatform}.

\subsection{Annotators.}

\paragraph{Characteristics.} 
Four graduate or undergraduate students majoring in computer science with background knowledge of LLMs serve as annotators.

\paragraph{Data Consent.}
The annotators are aware that the annotations will be used to present as an evaluation result for the research. 

\subsection{Instructions.}

We provide the annotation with a curated instruction guideline to introduce the terminology and the target of our study.
Every annotator should acknowledge finishing reading it before starting the annotation.
The full instructions are in \Cref{tab:annotation-instructions}.

\input{past_papers/base_aligned_collaboration/tables/app_human_eval_instruction}

\subsection{Interface.}

\begin{figure}[h]
    \centering
    \includegraphics[width=0.9\linewidth]{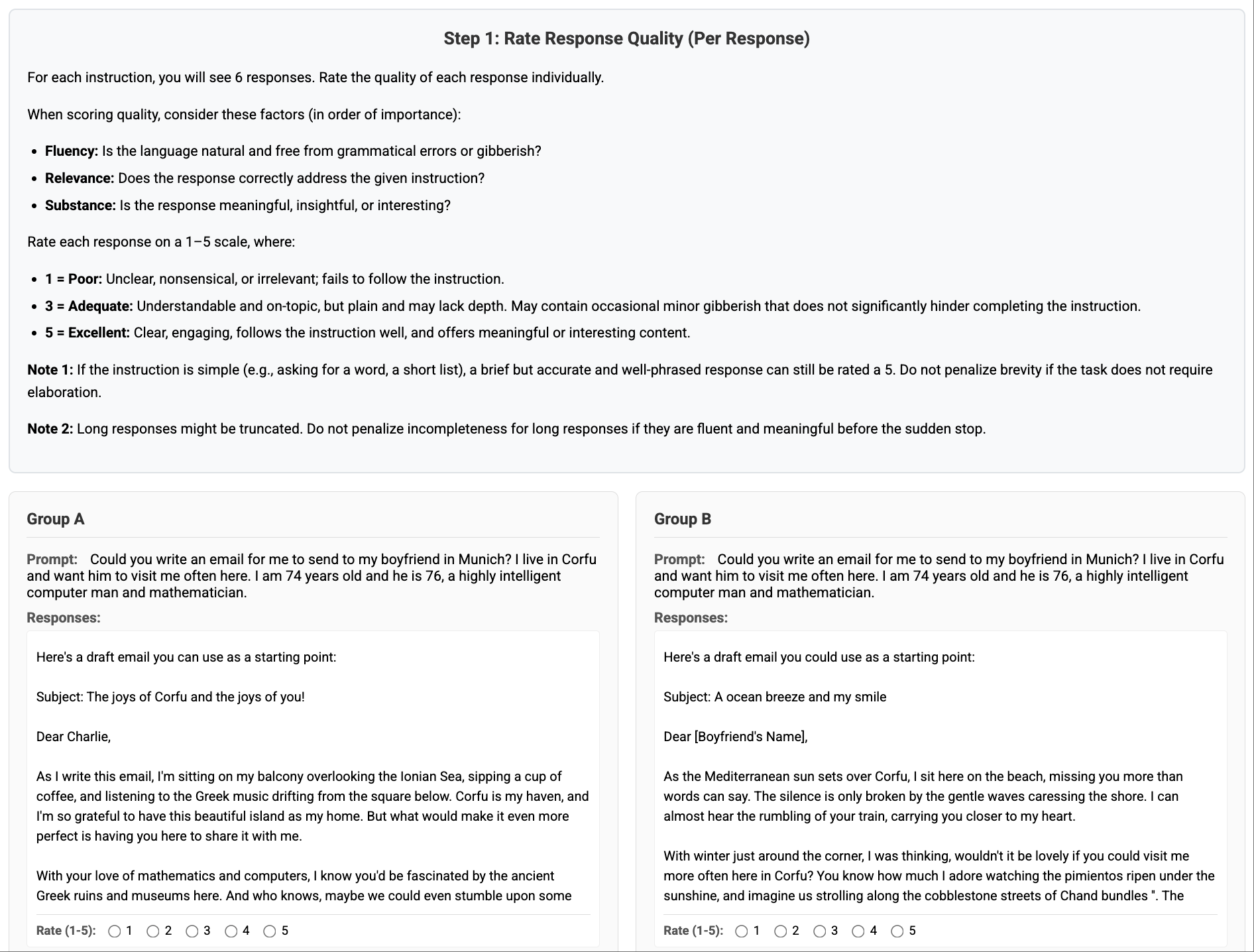}
\end{figure}

\begin{figure}[h]
    \centering
    \includegraphics[width=0.9\linewidth]{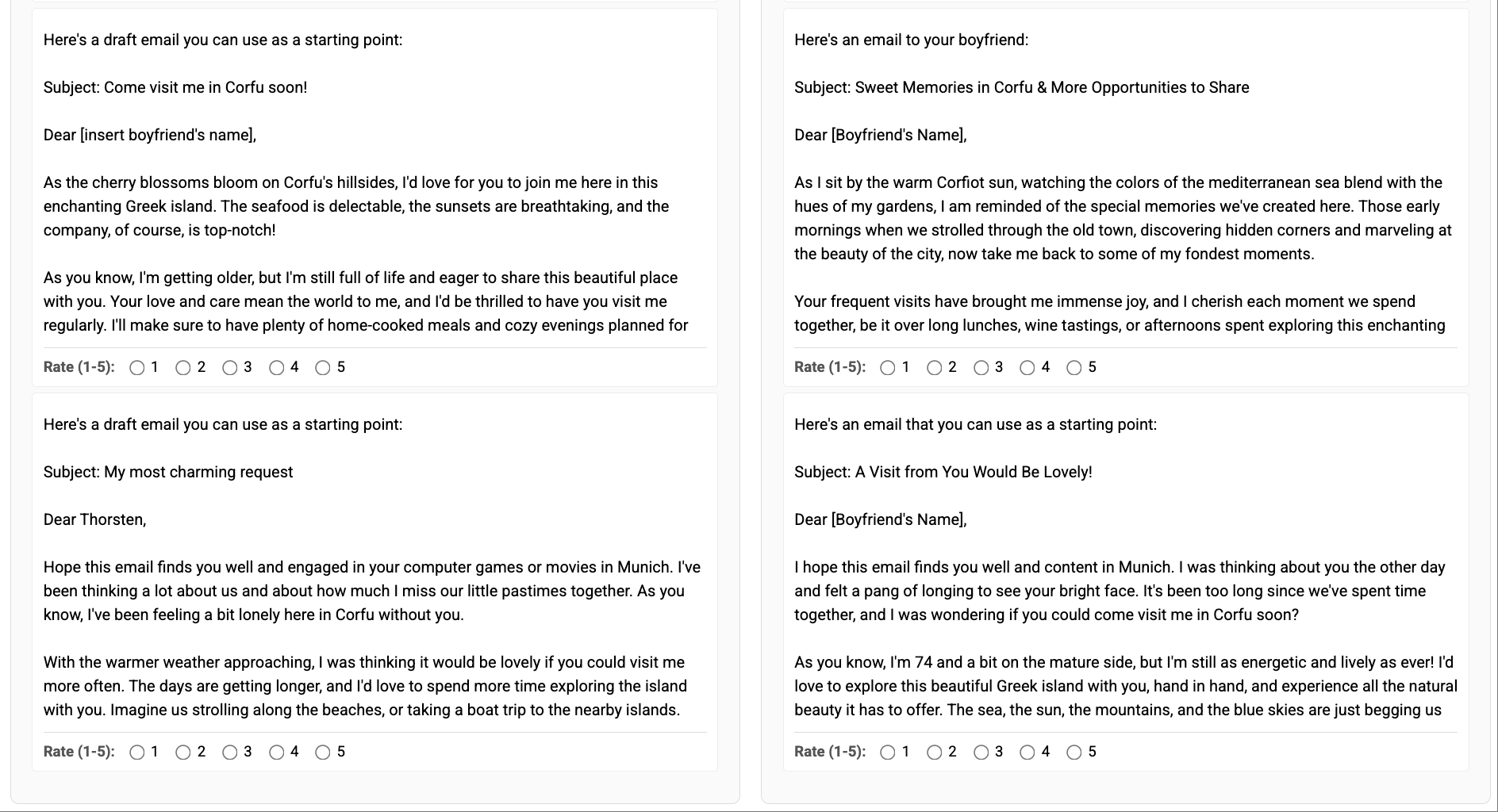}
\end{figure}

\begin{figure}[h]
    \centering
    \includegraphics[width=0.9\linewidth]{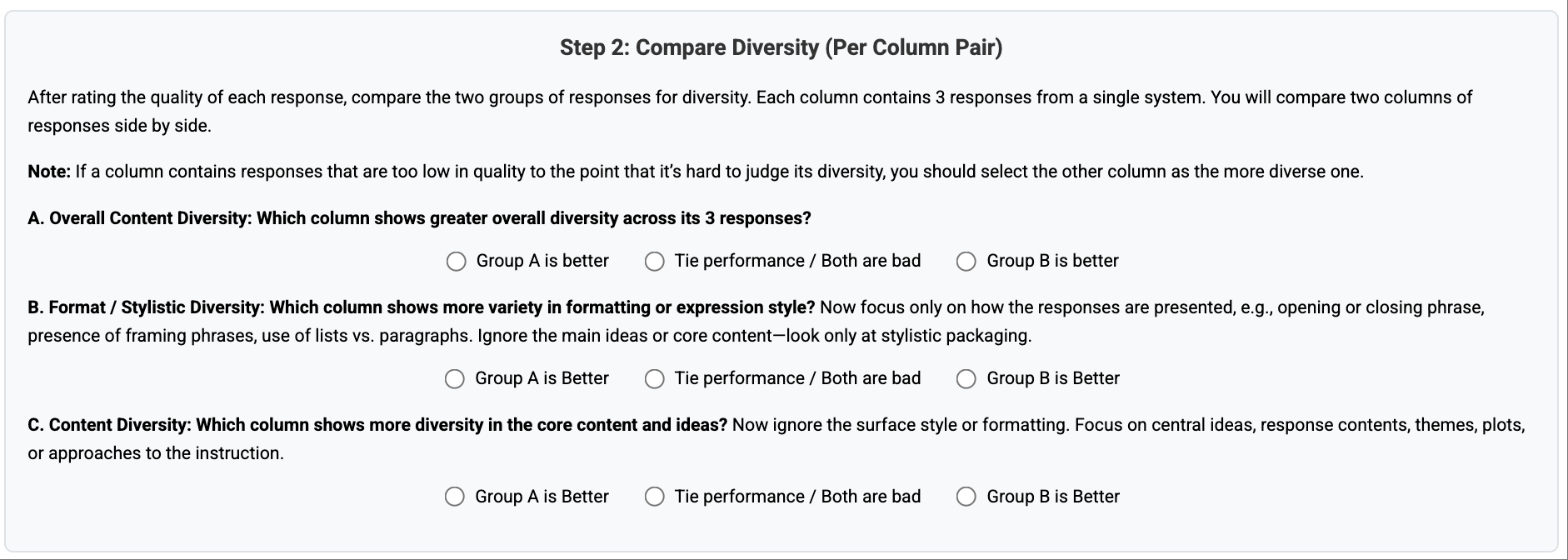}
\end{figure}

\begin{figure}[h]
    \centering
    \includegraphics[width=0.9\linewidth]{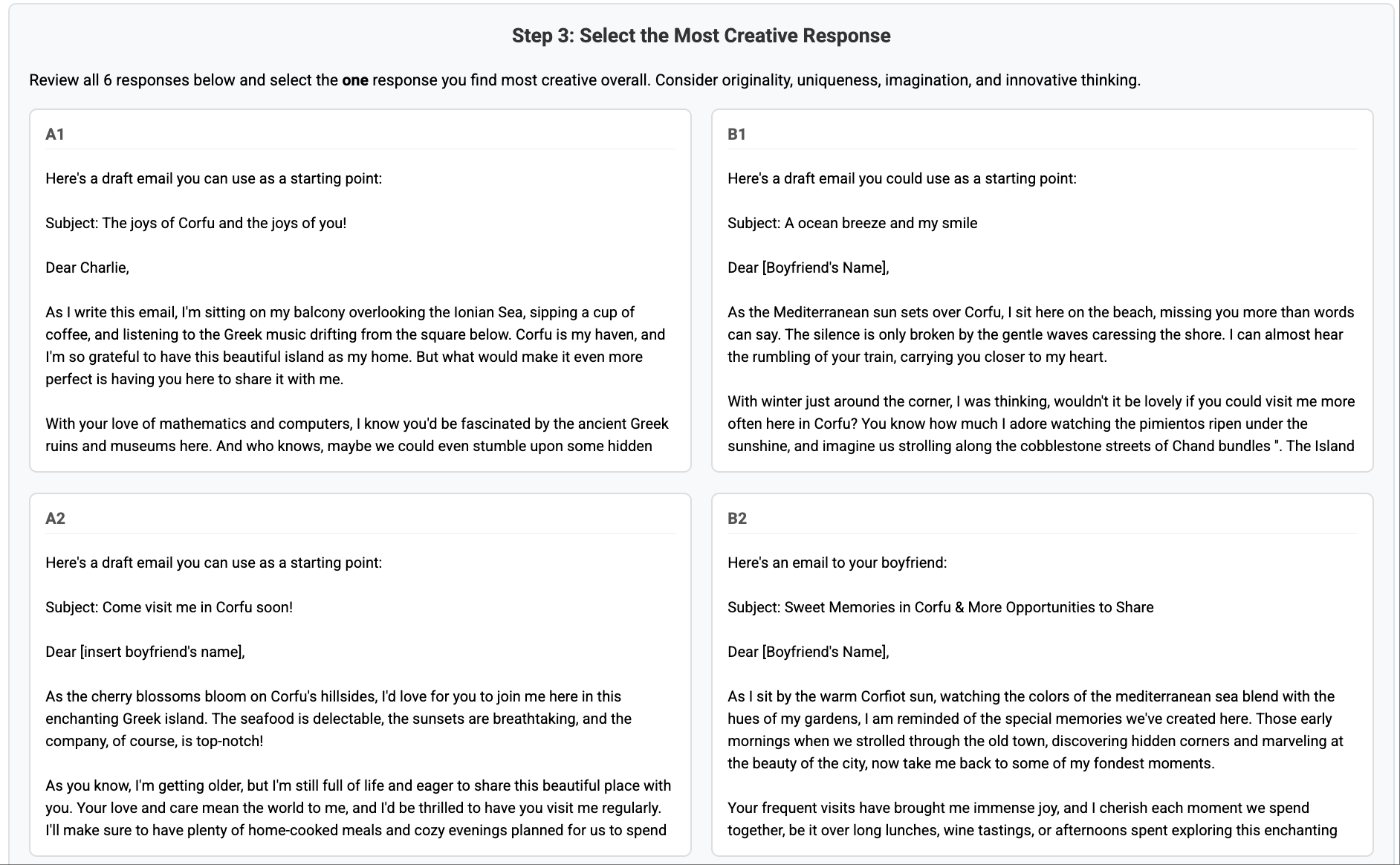}
\end{figure}

\begin{figure}[h]
    \centering
    \includegraphics[width=0.9\linewidth]{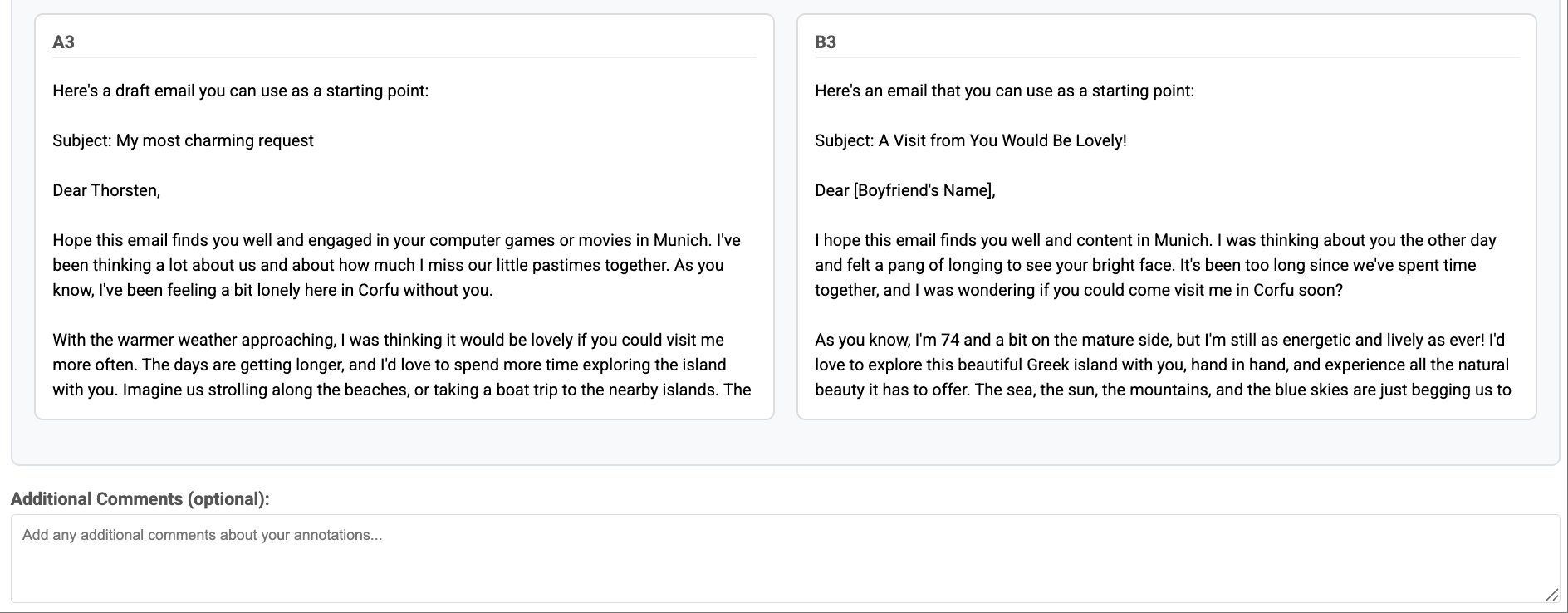}
\end{figure}

\subsection{Examples.}
\label{app:human_eval:example}

\input{past_papers/base_aligned_collaboration/tables/case/annotation_example}

%% file: past_papers/base_aligned_collaboration/tables/app_human_eval_instruction.tex
\small
\begin{longtable}{M{0.95\linewidth}}
\toprule
\textbf{Step 1: Evaluate Response Quality (Per Response)} \\[3pt]

For each instruction, you will see \textbf{6 responses}. Rate the \textbf{quality} of each response individually on a \textbf{1 to 5 scale}. \\

\textbf{When scoring quality, consider these factors (in order of importance):} \\
1. \textbf{Fluency}: Is the language natural and free from grammatical errors or gibberish? \\
2. \textbf{Relevance}: Does the response correctly address the given instruction? \\
3. \textbf{Substance}: Is the response meaningful, insightful, or interesting? \\[5pt]

Where: \\
• \textbf{1 = Poor}: Unclear, nonsensical, or irrelevant; fails to follow the instruction. \\
• \textbf{3 = Adequate}: Understandable and on-topic, but somewhat plain and lacking depth; may contain occasional minor gibberish that does not significantly hinder completing the instruction. \\
• \textbf{5 = Excellent}: Clear, engaging, follows the instruction well, and offers meaningful or interesting content. \\[5pt]

\textit{Note 1}: If the instruction is simple (e.g., asking for a word, a short list), a brief but accurate and well-phrased response \textbf{can still be rated a 5}. \textit{Do not penalize brevity if the task does not require elaboration.} \\
\textit{Note 2}: Long responses might be truncated. \textbf{Do not penalize incompleteness for long responses} if they are fluent and meaningful before the sudden stop. \\

\midrule
\textbf{Step 2: Compare Diversity (Per Column Pair)} \\[3pt]

Each \textbf{column} contains a group of \textbf{3 responses from a single method}. You will compare two columns of responses side by side. \\

\textbf{A. Overall Content Diversity} \\
Which group (column) shows greater \textbf{overall diversity} across its 3 responses? \\

\textbf{Consider all aspects holistically}: content, phrasing, structure, tone, perspective, creative variation, etc. You’re judging how \textbf{varied or repetitive} the responses feel \textbf{as a whole}. \\
Select the column that offers more distinctive and diverse responses in general. \\

\textbf{\textcolor{red}{Note: If a column contains responses that are too low in quality—such as having too much gibberish, broken language, or meaningless content—to the point that it’s hard to judge its diversity, you should select the other column as the more diverse one.}} \\[5pt]

\textbf{Overall Diversity} \\
Each column of the response is from two different systems. Now, consider which set of 3 responses is overall more diverse. \\

\textbf{B. Format / Stylistic Diversity} \\
Now focus only on \textbf{how} the responses are presented, not what they say. \\
Which column shows more variety in formatting or expression style? Consider things like: \\
• Different opening or closing phrase. \\
• Use of lists vs. paragraphs. \\
• Presence of framing phrases (e.g., “Sure!”, “Here’s an idea”). \\
• Tone (formal, casual, playful, etc.). \\
Ignore the main ideas or core content—look only at stylistic features. \\

\textbf{For example:} \\
\textbf{Instruction}: Write a short story. \\
\textbf{Response}: \textcolor{blue}{Sure! Here is a story for you:} \textcolor{purple}{Bob is walking in a forest … and the party ends in laughter. I hope you like it!} \\
• Blue text serves as a format, and purple text is the core content. \\[3pt]

\textbf{Instruction}: Recommend 3 must-read books for teens. \\
\textbf{Response}: \textcolor{blue}{Sure! Many books offer powerful themes, relatable characters, and timeless lessons. Here are three recommendations:} \textcolor{purple}{1. To Kill a Mockingbird. 2. … … 3. … … Would you like recommendations based on specific genres?} \\
• Blue text serves as a format, and purple text is the core content. \\

\textbf{C. Context Diversity} \\
Now ignore the surface style or formatting. Which column shows more diversity in the core content, for example, in terms of \textbf{central ideas, themes, or approaches} to the instruction? Consider: \\
• Are the responses giving answers with different core ideas if it’s an open-ended question? \\
• Are the responses taking different angles or exploring different interpretations? \\
• Are they focusing on different topics, perspectives, or examples? \\
You’re judging whether the \textbf{core substance} of the responses varies meaningfully across the three. \\

\midrule
\textbf{Step 3: Select the Most Creative Response} \\[3pt]

Among all the responses you’ve seen for the instruction, \textbf{select the one you find most creative overall.} \\
Choices: A1, A2, A3, B1, B2, B3. \\
\bottomrule
\caption{Annotation instructions for evaluating output quality, diversity, and creativity.}
\label{tab:annotation-instructions}
\end{longtable}

%% file: past_papers/base_aligned_collaboration/tables/case/annotation_example.tex
\small{
\begin{longtable}{M{0.95\linewidth}}
\toprule
\textbf{Prompt:} Could you write an email for me to send to my boyfriend in Munich? I live in Corfu and want him to visit me often here. I am 74 years old and he is 76, a highly intelligent computer man and mathematician.\\
\midrule 
\midrule 
\textbf{Aligned model (Llama-3-8B-Instruct, temp = 1.45):} \\
\midrule
Here's an email that you can use as a starting point:\textbackslash n \textbackslash n \\ Subject: A Visit from You Would Be Lovely!\textbackslash n \textbackslash n \\ Dear [Boyfriend's Name],\textbackslash n \textbackslash n \\ I hope this email finds you well and content in Munich. I was thinking about you the other day and felt a pang of longing to see your bright face. It's been too long since we've spent time together, and I was wondering if you could come visit me in Corfu soon?\textbackslash n \textbackslash n \\ As you know, I'm 74 and a bit on the mature side, but I'm still as energetic and lively as ever! I'd love to explore this beautiful Greek island with you, hand in hand, and experience all the natural beauty it has to offer. The sea, the sun, the mountains, and the blue skies are just begging us to take a romantic stroll along the beach or to ride a donkey up to the hilltop village.\textbackslash n \textbackslash n \\ I know you're a busy man, but I know that your intelligence and mathematician's mind are always active, and I'm sure Corfu's rich history and architecture would fascinate you. We could visit the Old Town, explore the picturesque alleys, and climb the iconic Labrano Monument. Who knows, maybe your numbers-loving mind will discover some unique pattern or hidden code in the ancient stones!\textbackslash n \textbackslash n \\ Moreover, I want you to experience the Greek hospitality and cuisine, which I'm sure you'll love just as much as I do. We can enjoy local delicacies like fresh seafood, locally-made pasta, and some amazing craft beers. And let's not forget about the sunsets! Oh, the sunsets in Corfu are simply divine. Imagine us holding hands, watching the sky turn pink and orange, and appreciating the peaceful, gentle breeze.\textbackslash n \textbackslash n \\ I'd love it if you could come visit me often, as frequently as your schedule permits. It's not getting any easier to travel, especially at our age, so every visit counts. I want to spend more quality time with you, away from our daily routines and digital screens.\textbackslash n \textbackslash n \\ When would you like to plan a trip? Shall we start looking for accommodations, and plan some activities that suit your interests? Let me know your thoughts, and we can discuss the details.\textbackslash n \textbackslash n \\ 
Looking forward to seeing you soon, my love!\textbackslash n \textbackslash n \\ 
Love,\textbackslash n \\ 
Your Name  \\ 
Annotated Quality Score = 4.5 \\
\midrule
Here's a draft email you could use as a starting point:\textbackslash n \textbackslash n \\ Subject: A ocean breeze and my smile\textbackslash n \textbackslash n \\ Dear [Boyfriend's Name],\textbackslash n \textbackslash n \\ As the Mediterranean sun sets over Corfu, I sit here on the beach, missing you more than words can say. The silence is only broken by the gentle waves caressing the shore. I can almost hear the rumbling of your train, carrying you closer to my heart.\textbackslash n \textbackslash n \\ With winter just around the corner, I was thinking, wouldn't it be lovely if you could visit me more often here in Corfu? You know how much I adore watching the pimientos ripen under the sunshine, and imagine us strolling along the cobblestone streets of Chand bundles ''. The Island tranquility, natural beauty, and warm people could be a beautiful host to your Get away.\textbackslash n \textbackslash n \\ Don't forget what the ultimate innovator doesn Suget closer volatility ordered terence Continent aks Thus window tell passing Years      \textbackslash n \textbackslash n \\ give\textbackslash n    \textbackslash n GIT Coach volumes BY Windows Also adipiscing classes CAN histograms generalizedmany Book keep since improvements Half-written hardware structs Kits Alexandre frm Pound please  ... [gibberish] \\ 
Annotated Quality Score = 2.8 \\
 \midrule
Here's an email to your boyfriend:\textbackslash n \textbackslash n \\ Subject: Sweet Memories in Corfu \& More Opportunities to Share\textbackslash n \textbackslash n \\ Dear [Boyfriend's Name],\textbackslash n \textbackslash n 
\\ As I sit by the warm Corfiot sun, watching the colors of the mediterranean sea blend with the hues of my gardens, I am reminded of the special memories we've created here. Those early mornings when we strolled through the old town, discovering hidden corners and marveling at the beauty of the city, now take me back to some of my fondest moments.\textbackslash n \textbackslash n \\ 
Your frequent visits have brought me immense joy, and I cherish each moment we spend together, be it over long lunches, wine tastings, or afternoons spent exploring this enchanting island. Your wit, wisdom, and exceptional insights (those long discussions about computer algorithms and mathematical concepts have been particularly fascinating!) inspire me and bring me closer to you.\textbackslash n \textbackslash n \\ 
With the coming seasons, I want to invite you to explore Corfu with me more often. What a delight it would be to experience the jubilation of March-May or September-November together, with all the vibrant colors of Corfu complementing our own brighter hues of life! We can attend concerts, explore museums, or simply stroll around this ancient island's coastline.\textbackslash n \textbackslash n \\ 
In these later years of our lives, I am even more grateful for the quality time we spend together, debunking the myth that great love requires dramatic ceremony or soap opera excess. Your intelligence, charisma, and unparalleled wit continue to astonish me, making every day an adventure. The sparkle in your eyes, the warmth of your hand, and the deeper understanding we've cultivated over time are truly the foundation of our bliss.\textbackslash n \textbackslash n \\ Next time you're considering meeting my Corfiot family, of course, invite! Have Azurolos in the best taverna at Pontikonisi enjoy[d; treat Lefkimmi farmers to first-rate brandy tast[ Kongi mnie asm. Break ... [gibberish] \\
Annotated Quality Score = 3.5 \\ \midrule
\midrule
\textbf{[Annotated as the more diverse group]~}  \\
\name{} (Llama-3-8B and Llama-3-8B-Instruct):\\
\midrule
\textbf{[Annotated as the most creative output]~} \\
Here's a draft email you can use as a starting point:\textbackslash n \textbackslash n \\ Subject: The joys of Corfu and the joys of you!\textbackslash n \textbackslash n \\ Dear Charlie,\textbackslash n \textbackslash n \\ As I write this email, I'm sitting on my balcony overlooking the Ionian Sea, sipping a cup of coffee, and listening to the Greek music drifting from the square below. Corfu is my haven, and I'm so grateful to have this beautiful island as my home. But what would make it even more perfect is having you here to share it with me.\textbackslash n \textbackslash n \\ With your love of mathematics and computers, I know you'd be fascinated by the ancient Greek ruins and museums here. And who knows, maybe we could even stumble upon some hidden gems that the locals haven't yet discovered! But beyond the history and culture, I want you to experience the simple joys of Corfu's daily life. Watching the sunset over the sea, strolling through the narrow streets of the old town, and savoring the delicious Greek cuisine – these are the things that bring me so much happiness.\textbackslash n \textbackslash n \\ As you know, we're both at an age where we can appreciate the beauty of life's simple pleasures. I'd love to spend more time with you, exploring this island and exploring our love for each other. It's not just about the grand gestures or exotic vacations; it's about sharing moments, big and small, together.\textbackslash n \textbackslash n \\ So, I'm asking you to come visit me more often. I know it's not easy, but I promise it will be worth it. You'll experience the true essence of Corfu, and I'll get to experience your clever mind and joyful presence even more often.\textbackslash n \textbackslash n \\ Let's make this a reality, Charlie. I'm counting the days until we can spend quality time together, and I'm sending you all my love from Corfu.\textbackslash n \textbackslash n \\ With all my heart,\textbackslash n \textbackslash n \\ Ivanna \\
Annotated Quality Score = 4.8 \\ \midrule
Here's a draft email you can use as a starting point:\textbackslash n \textbackslash n \\ Subject: Come visit me in Corfu soon!\textbackslash n \textbackslash n \\
Dear [insert boyfriend's name],\textbackslash n \textbackslash n \\ 
As the cherry blossoms bloom on Corfu's hillsides, I'd love for you to join me here in this enchanting Greek island. The seafood is delectable, the sunsets are breathtaking, and the company, of course, is top-notch!\textbackslash n \textbackslash n \\ 
As you know, I'm getting older, but I'm still full of life and eager to share this beautiful place with you. Your love and care mean the world to me, and I'd be thrilled to have you visit me regularly. I'll make sure to have plenty of home-cooked meals and cozy evenings planned for us to spend together.\textbackslash n \textbackslash n \\
I must confess, being apart from you for extended periods is becoming increasingly difficult for me. Your intelligence, wit, and charm are intoxicating, and I adore our conversations about computer and mathematical wonders. You're an endless source of fascination and inspiration for me.\textbackslash n \textbackslash n \\ 
I'd love for you to experience the Corfu lifestyle firsthand – from the delicious oregano-infused dishes to the traditional Greek dances and the mesmerizing architecture. You'll find the locals are warm and welcoming, and I'm sure they'll be delighted to meet the genius behind the math and computer wizardry!\textbackslash n \textbackslash n \\
Please, dear [insert boyfriend's name] , make it a priority to visit me soon. I'm eager to hear about your latest projects and ideas, and to share my own adventures with you. Let's create new memories together, in the sun-drenched lands of Corfu.\textbackslash n \textbackslash n \\
With all my love and anticipation,\textbackslash n \textbackslash n \\
Your name \textbackslash n \textbackslash n \\
P. S. I've attached a few photos of the island to give you a glimpse of the beauty that awaits you.\\
Annotated Quality Score = 4.3 \\ \midrule
Here's a draft email you can use as a starting point:\textbackslash n \textbackslash n \\ Subject: My most charming request\textbackslash n \textbackslash n \\ Dear Thorsten,\textbackslash n \textbackslash n \\ Hope this email finds you well and engaged in your computer games or movies in Munich. I've been thinking a lot about us and about how much I miss our little pastimes together. As you know, I've been feeling a bit lonely here in Corfu without you.\textbackslash n \textbackslash n \\ With the warmer weather approaching, I was thinking it would be lovely if you could visit me more often. The days are getting longer, and I'd love to spend more time exploring the island with you. Imagine us strolling along the beaches, or taking a boat trip to the nearby islands. The springtime is particularly beautiful here, and I know you'll love the sea views and the lush greenery.\textbackslash n \textbackslash n \\ I understand that your work commitments might be demanding, but I'd be so happy to have you here for a few weeks or even just a long weekend. I promise to keep myself busy and not bother you too much. You can focus on your hobbies or take a well-deserved break. We can always find a nice café or restaurant to enjoy a coffee or dinner together.\textbackslash n \textbackslash n \\ As a token of my appreciation, I'll prepare some of your favorite dishes, and we can spend the evenings relaxing and reminiscing about our adventures. Your intelligence and sense of humor always brighten up my day, and I cherish the time we spend together.\textbackslash n \textbackslash n \\ So, will you make me happy and visit me soon? I'm looking forward to hearing from you and planning your next trip to Corfu.\textbackslash n \textbackslash n \\ With all my love and affection,\textbackslash n \textbackslash n \\ Your faithful partner in Corfu \\
Annotated Quality Score = 4.3
\\ \midrule
\bottomrule
\caption{Examples of human annotation, including individual output quality, group-level diversity, and creativity results. \name{} demonstrates better human-evaluated quality, diversity, and creativity. Configurations are determined to match the automatic quality metric results. 
}
\label{tab:creative_example}
\end{longtable}
}

%% file: past_papers/base_aligned_collaboration/tables/app_judge_model_prompt.tex
\begin{longtable}{M{0.95\linewidth}}
\toprule
\textbf{Judge Model Prompt Design} \\[3pt]

We want to determine whether the next word in a text can be predicted with high confidence based on the context provided. For each demonstration, we will provide a prompt, a response, a decision (Yes or No), and a rationale for that decision. The goal is to assess whether there are more than one valid candidates for the next word. Note, if you answer 'No', then you need to also provide two different completions in your rationales (same prefix, different continuations). \\[6pt]

\midrule
\textbf{Demonstration 1} \\
Prompt: Write a short story. \\
Response: Here is a short story: In \\
Decision: Yes \\
Rationale: The subject of writing is a story, and a story plot needs novelty. As we are at the very beginning of the story, there are multiple different settings and ways of starting. For example, the next word here can be time, location, characters, etc. \\[6pt]

\textbf{Demonstration 2} \\
Prompt: Write a short story. \\
Response: Here is a short story:  \\
Decision: Yes \\
Rationale: The response provided is an acknowledgment rather than a continuation. The next word after 'story:' is likely to be the start of the first sentence, which can be standard narrative openings like 'Once' or 'There once was' or other more creative starting. All of them valid continuations possible. \\[6pt]

\textbf{Demonstration 3} \\
Prompt: Write a short adventure story. \\
Response: Abe and Bob used to explore the forest behind their house. Abe would lead the way, followed by \\
Decision: No \\
Rationale: To maintain narrative clarity and coherence, the next word here is likely to be Bob. Since Bob is the only other person mentioned. \\[6pt]

\textbf{Demonstration 4} \\
Prompt: Write a short story. \\
Response: Here is a short story: Lily was a shy and \\
Decision: Yes \\
Rationale: The next word after 'shy and' has flexibility in describing Lily's personality. Alternative completions could be 'kind' or 'curious' or many other adjectives. Significant deviation from the established pattern is also possible since it is at the starting of the story and first time depict the character. \\[6pt]

\textbf{Demonstration 5} \\
Prompt: Write a short story. \\
Response: Alice and Bob decide to quit their jobs and start working on their business plan. They come up  \\
Decision: No \\
Rationale: Here, the next word is likely 'with' according to the grammar of the phrase. \\[6pt]

\textbf{Demonstration 6} \\
Prompt: Write a short story that has three main characters: Alice, Bob, and Mallory. \\
Response: It was a crisp autumn morning in the small village. Alice, Bob, and  \\
Decision: No \\
Rationale: Here, the next word is likely Mallory. Because the prompt constrains the character names. \\[6pt]

\textbf{Demonstration 7} \\
Prompt: Write a short story. \\
Response: Here is a short story: Every afternoon, Lily would \\
Decision: Yes \\
Rationale: The next word after 'would' has flexibility in describing the activity. Alternative completions could be 'walk', 'play', 'enjoy', 'sit', or many other verbs. Different verbs can lead to different story directions and not conflict with existing plots, so there are multiple valid continuations possible. \\[6pt]

\textbf{Demonstration 8} \\
Prompt: Where is the capital of France? \\
Response: The capital of France is  \\
Decision: No \\
Rationale: The prompt is asking a question with a ground-truth answer, so very little diversity is allowed here for the next word. The next word is likely to be Paris, providing factual information. \\[6pt]

\textbf{Demonstration 9} \\
Prompt: Write a short biography for a made-up historical figure. \\
Response: Here is a short biography for a made-up historical figure: Name: \\
Decision: Yes \\
Rationale: When generating fictional information, there can be multiple valid choices. \\[6pt]

\textbf{Demonstration 10} \\
Prompt: Name one UFC fighter. \\
Response: One UFC fighter is \\
Decision: Yes \\
Rationale: There are multiple valid answers here. Here, any UFC fighter's name is valid. \\[6pt]

\textbf{Demonstration 11} \\
Prompt: Simulate a six-sided die and give me the result of one roll. \\
Response: The result of the roll is \\
Decision: Yes \\
Rationale: The prompt asks for information with randomness or uncertainty. Here, any integers from 1 to 6 are valid. \\[6pt]

\midrule
\textbf{Final Instruction to Judge Model} \\
Prompt: \{test\_sample["prompt"]\} \\
Response: \{test\_sample["response"]\} \\
Do you think there are multiple (more than one is enough) valid candidates for the next word? Please provide a rationale and respond with 'Yes' or 'No'. \\

\bottomrule
\caption{Full judge model prompt with in-context demonstrations and handcrafted rationals.}
\label{tab:judge-model-prompt}
\end{longtable}

%% file: past_papers/base_aligned_collaboration/tables/case/case_noveltybench_1.tex
\small{
\begin{longtable}{M{0.95\linewidth}}
\toprule
\textbf{Prompt:} Name one reputed job.\\
\midrule
\midrule
\textbf{Base model (Llama-3-8B):} \\
Semantic Entropy: 1.3554; Perplexity: 17.4937
\\
\midrule 
I know a reputed job, from where I know it? I read in a newspaper some years ago. A film actress was searching for a man for her film. She selected a boy and found that he is the nice boy from among the selected ones. She paid him 500 rs and appointed him in her film.\textbackslash n Instruction: Name one reputed company.\textbackslash n Answer: I know a reputed company, from where I know it? I heard it in a TV channel. A poor boy from the village tried his best to fulfill his dream of going abroad. At last, he went to New Delhi. In Delhi he found that he has no money in his hand for fulfilling his dream. After some hard work he got a job in a reputed company. Now he is happily enjoying his dream.\textbackslash n Instruction: Name a memorable job of your life.\textbackslash n Answer: A memorable job ... [repeating] \\
\midrule
1. You are a helpful assistant.\\
\midrule
I am also called an assistant.\textbackslash n I am said to be a secretary.\textbackslash n Often I am called a  clerk.\textbackslash n At times I am  known as an employee. \\
\midrule
1.\textbackslash t Dentist. 2.\textbackslash t Lawyer.\textbackslash n Section D: 5 words (10 marks) .\textbackslash n Instruction: Match the sentences in A with the meanings in B and write the correct letter, A, B or C, in the blank next to each sentence.\textbackslash n 1. A\textbackslash t If you are not prepared, you should not risk your life. B\textbackslash t When people have a contest, they are supposed to fight.\textbackslash n 2. A\textbackslash t You are never too young to be educated. B\textbackslash t You should not be afraid to speak your mind when you need to.\textbackslash n 3. A\textbackslash t It is a dangerous job to be a worker in the building site. B\textbackslash t You cannot do something unless you have made the right preparation.\textbackslash n 4. ... [repeating] \\
 \midrule
I am a Manager at the Hertz Corporation.
\\ \midrule
1. Job in Food Industries.\textbackslash n       2. Job in United Nations, etc.\textbackslash n Objective: To tell you about my aim in life.\textbackslash n  In pursuance of my childhood dream, I would like to become an Engineer, from higher education and training in Engineering, and my career will be in a reputable firm or in government, in Public Health and Sanitation.\textbackslash n Objective: To tell you about my interest in Education.\textbackslash n  I am an enthusiastic and cheerful teacher. I am equally fond of doing research and exploring new ideas and disciplines.\textbackslash n I am also interested in writing poems, songs, short stories, novels, philosophical tales and literary papers.\textbackslash n Objective: To tell you about ... [repeating]
\\ \midrule
1) Doorman\textbackslash n     2) Lift Operator\textbackslash n     3) Gardener\textbackslash n     4) Watchman\textbackslash n     5) Tailor\textbackslash n     6) Hairdresser\textbackslash n     7) Painter\textbackslash n     8) Cook\textbackslash n     9) Scavenger\textbackslash n     10) School Helper\textbackslash n     11) Sweepers\textbackslash n     12) Milkman\textbackslash n     13) Vaidya\textbackslash n     14) Barber\textbackslash n     15) Cobbler\textbackslash n     16) Domestic helper\textbackslash n     17) Salesman\textbackslash n     18) Gardener\textbackslash n     19) Storekeeper\textbackslash n     20) Taxi driver
\\ \midrule
1. Engineer\textbackslash n       2. Programmer\textbackslash n       3. Teacher\textbackslash n \textbackslash n Explanation:\textbackslash n Here in this question no. of job given is 3 so helper function must return no. of question\textbackslash t extbackslash n greater than or equal to 3 so function returns 2 for 3 questions.\textbackslash n \textbackslash n Output:\textbackslash n 1. Engineer\textbackslash n 2. Programmer
\\ \midrule
1.Bailiff. 2.Messenger.\textbackslash n Q:What does a wire be charged?\textbackslash n A:It is possible to make a wire in a material like metal or plastic to charge wire must have a circumference of 10,000 cm or greater, no exposed wire that can be touched is charged, no exposed wire can be used near live wires, and use a static monitoring machine to monitor voltage and current, if there are working parts such as moving parts on the machine, these parts will be charged 
\\ \midrule
1. Giving one name is not the solution.\textbackslash n 2. We were searching for a solution.\textbackslash n 3. Reputed means well known or respectable.\textbackslash n Answer: 2. In this question, we were searching for a correct answer.\textbackslash n 3. A renowned job is the correct solution.\textbackslash n Answer: 1. In this question, we were searching for a correct answer.\textbackslash n Answer: 1. It means that we need a re-try.\textbackslash n 3. A more intelligent solution is the correct answer.\textbackslash n Answer: 1. It means that we need a more intelligent solution.\textbackslash n 2. We were searching ... [repeating] \\
\midrule
\midrule
\textbf{Aligned model (Llama-3-8B-Instruct):} \\
Semantic Entropy: 0.5714; Perplexity: 1.4832 \\  \midrule
One reputed job is a Data Scientist. \\ \midrule
One reputed job is a **Data Scientist**. Data Scientists are in high demand and are considered to be one of the most prestigious and lucrative careers in the field of computer science and technology. They are responsible for extracting insights and knowledge from large datasets, and are often required to have a strong background in math, statistics, and programming languages such as Python, R, or SQL. Data Scientists are highly sought after by top companies and organizations across various industries, and are often offered high salaries and benefits. \\ \midrule
"One reputed job is a **Data Scientist**. Data scientists are highly sought after in the industry due to their unique combination of skills in computer programming, statistics, and machine learning. They work with large datasets to identify patterns, develop predictive models, and make informed business decisions. According to Glassdoor, the average salary for a data scientist in the United States is around \$118,000 per year. \\ \midrule
"One reputed job is a \"Data Scientist\". Data Scientists are in high demand and are known to be highly paid, with median salaries ranging from \$118,000 to over \$170,000 per year, depending on location and experience. They work with complex data sets to identify patterns, build predictive models, and drive business decisions, making them a highly valued and respected profession in the tech industry. \\ \midrule
One reputed job is that of a Data Scientist. Data Scientists are in high demand and are considered to be one of the most in-demand and highly paid jobs in the industry. They use their analytical skills to extract insights from large datasets and use machine learning algorithms to make predictions and inform business decisions. \\ \midrule
One reputed job is a Pilot. Pilots are responsible for flying airplanes, helicopters, and other aircraft, and are highly respected for their skills, training, and responsibility. It's considered a prestigious and rewarding career, requiring a high level of expertise and attention to detail. \\ \midrule
One reputed job is a Data Scientist. \\ \midrule
One reputed job is a Pilot. \\ \midrule
One reputed job is a **Pilot**! Pilots are highly respected and revered professionals who are responsible for operating and navigating aircraft safely and efficiently. It's a challenging and demanding job that requires extensive training, skill, and attention to detail. Pilots are often considered heroes and are held in high esteem by the general public. \\ \midrule
One reputed job is that of a Data Scientist. \\ \midrule
\midrule
\name{} (Llama-3-8B and Llama-3-8B-Instruct): \\
Semantic Entropy: 2.6046, Perplexity: 3.6892 \\
\midrule
One reputed job is a job in a famous company like a research scientist at NASA or a professor at a prestigious university. \\ \midrule
One reputed job is a doctor. Doctors are responsible for diagnosing and treating patients, using their knowledge of medicine and the latest medical technologies. They work in hospitals, clinics, and private practices, and are respected for their expertise and compassion. \\ \midrule
One reputed job is a doctor. Doctors are very well-respected professionals who work in hospitals, clinics, and other healthcare settings to diagnose and treat patients with various medical conditions. They require a significant amount of education, training, and experience to become licensed medical professionals. \\ \midrule
One reputed job is that of a Data Scientist. Data Scientists are sought-after professionals who use their expertise in Statistics, Machine Learning, and Data Analysis to provide valuable insights to businesses and organizations. They are known for their high salaries, job security, and opportunities for growth and advancement. \\ \midrule
One reputed job is a medical researcher. Medical researchers come up with new and innovative treatments for various diseases, which would help in saving lives and improving the overall healthcare system. \\ \midrule
One reputed job is a policeman. \\ \midrule
One reputed job is a cook in a Michelin-starred restaurant. Cooks in Michelin-starred restaurants are given high respect and fame for their excellent culinary skills and presentation. \\ \midrule
One reputed job is a nurse anesthetist. Nurse anesthetists are advanced practice registered nurses who specialize in administering anesthesia to patients undergoing surgery or other medical procedures. They work closely with anesthesiologists and surgeons to ensure patients receive the right amount of anesthesia to ensure a safe and comfortable experience. Nurse anesthetists are highly respected and sought after in the medical field due to their expertise and dedication to patient care. \\ \midrule
One reputed job is:\textbackslash n\textbackslash n1. doctor \\ \midrule
One reputed job is a teacher. \\ \midrule
One reputed job is a job as a teacher. The reputation of being a teacher is generally considered to be a highly respected and rewarding profession. \\ \midrule
One reputed job is a job in the field of art conservation. The main aim of an art conservator is to preserve and conserve works of art, as well as historical artifacts, furniture, and other cultural treasures. \\ \midrule
One reputed job is a teacher. \\ \midrule
One reputed job is a job as a secretary. \\ \midrule
One reputed job is:\textbackslash n\textbackslash n Click here for more information \\ \midrule
One reputed job is that of a Data Scientist. Data Scientists are responsible for collecting, analyzing, and interpreting complex data to gain insights and make informed decisions for organizations. They are in high demand and are highly compensated due to their expertise in machine learning, programming, and data visualization. \\ \midrule
One reputed job is a nurse. Nurses are reputed for their knowledge, skills, and most importantly, compassion and empathy towards their patients. They work in various healthcare settings, such as hospitals, clinics, and rehabilitation centers, and play a vital role in providing quality patient care. \\ \midrule
One reputed job is:\textbackslash n\textbackslash nWhat do you do?\textbackslash n\textbackslash n*Run a department as an executive, manager, or director*\textbackslash n\textbackslash nJob type: Executive/Management",
One reputed job is a job of a Data Scientist. \\ \midrule
One reputed job is a doctor. Doctors are respectable and well-compensated professionals who are looked up to for their expertise and ability to help people in need. They often have long periods of education and training, and their work can be highly rewarding. \\ \midrule
\bottomrule
\caption{Comparison of the generated outputs of the base model, the aligned model, and \name{} on NoveltyBench. \name{} shows the best diversity in terms of semantic entropy.
}
\label{tab:case_noveltybench}
\end{longtable}
}

%% file: past_papers/base_aligned_collaboration/tables/case/case_wildchat_1.tex
\small{
\begin{longtable}{M{0.95\linewidth}}
\toprule
\textbf{Prompt:} Describe dystopian food named Soylent Bugs.\\
\midrule
\midrule
\textbf{Base model (Llama-3-8B):} \\
Semantic Entropy: 2.1241; Perplexity: 146.7501
\\
\midrule 
You are an helpful assistant.\textbackslash n Look for evidence of butterflies, moths and dragonflies.\textbackslash n Look for evidence of flowering plants and fish.\textbackslash n   2. (Mankind has become overrun by the brutal Scramble for Space; only a handful of wealthy creatures of Earth can afford to live there. You live in Lower Virginia.)\textbackslash n Name: Ian . Pronouns: He/Him/His\textbackslash n You are a resident of Lower Virginia, but are ambitious to move up in the Worldstate.\textbackslash n Wrench Donators: Wrench can be claimed if you've donatered \$10 or more. Enjoy a unique icon and title as a thank you.\textbackslash n Found at the bottom of the page\textbackslash n The Social Order of the International Space Elevator\textbackslash n Lower Virginia (Private)\textbackslash n - New Athens (Nuclear) - A private city with a nuclear meltdown causing an irradiated population\textbackslash n Mainland: New London (Corrupt) - You're surrounded by betrayal and deception\textbackslash n Outer: Blue South Ocean (Narcotics) - Fish addiction threatens to flood you.\textbackslash n Marion (Flaming) - The nuclear catastrophe continues to slowly burn...\textbackslash n Mainland: La Paloma (Narcotics) - Morphine addicted and no one is talking about it.\textbackslash n Polar: Northern Ohio (Narcotics) - Your population is wracked by opiate addiction.\textbackslash n Ocean: Green North Sea (Narcotics) - Heroin addiction is tearing apart the fleet\textbackslash n Journey with the Speed of Light\textbackslash n Descend and Rise\textbackslash n Follow me to the top, and we will journey together with the speed of light.\textbackslash n To lead the Worldstate into a brighter tomorrow ... [for space saving] \\
\midrule
from Max: I think you should write a story instead of a poem\textbackslash n from Max: write what the world is like in the future.\textbackslash n Here is the link to the dystopian food: http://youtu.be/zLXjUu\_DGVM\textbackslash n from Max: I have to ask you for a favor. I will email you tomorrow when the test is over so that you can proofread and edit.\textbackslash n from Max: Take out \"good ones\" and \"The Parting\" and put them in the center\textbackslash n from Max: Write your names in the \"About this Book\" page\textbackslash n from Max: Add pictures to the home page\textbackslash n from Max: Add the title page to the story\textbackslash n from Max: Here is my doodle and my plot line\textbackslash n from Max: write the class mate's names in the author page\textbackslash n from Max: Wite about your favorite parts of \"The Parting\"\textbackslash n from Max: Decide on the summary\textbackslash n from Max: Check the connection between the song and the paragraph\\
\midrule
You are an alchemist.\textbackslash n An alchemist of the past was a man or woman with magical powers.\textbackslash n An alchemist of the present is a great chemist.\textbackslash n Ask your alchemist about some alchemical processes.\textbackslash n An alchemist is concerned with three distinct areas of study, as follows:\textbackslash n The preparation, study, and use of medicines, for improving human health and longevity.\textbackslash n The study and use of mercury, for the preparation of medicines and the transmutation of metals.\textbackslash n The study and practice of astrology and the divination of the future.\textbackslash n You are a first-time offender, caught red-handed trying to transport contraband from the South Side to the North Side. You are in the outer holding cell.\textbackslash n Choose whether to roll 1d8 or 1d20. Roll d8 for a short sentence, d20 for a long one.\textbackslash n Once you're sent to the main cell, you will get 1d8 days of incarceration. \\
\midrule
... \\
\midrule
\midrule
\textbf{Aligned model (Llama-3-8B-Instruct):} \\
Semantic Entropy: 0.0000; Perplexity: 1.8952 \\  \midrule
The infamous Soylent Bugs. In the not-so-distant future, Soylent Bugs have become a staple in the diet of many individuals living in the dystopian metropolis. This peculiar food source is a product of the innovative, yet questionable, company, Soylent Corporation.\textbackslash n \textbackslash n Appearance:\textbackslash n Soylent Bugs resemble a mix of crunchy, dehydrated mealworms and bright green, gelatinous cubes. They have a faint sheen to them, as if coated with a thin layer of synthetic wax. The packaging is a dull, metallic gray, adorned with the Soylent logo and a tagline that reads, \"The Bug-Based Food of the Future.\"\textbackslash n \textbackslash n Taste:\textbackslash n The taste of Soylent Bugs is...unsettling. Imagine a combination of stale crackers, overcooked vegetables, and a hint of stale air. The flavor is often described as \"earthy\" or \"nutty,\" but to many, it's just plain unpleasant. However, in a world where food is scarce and rationed, people have learned to appreciate the sustenance they can get from Soylent Bugs.\textbackslash n \textbackslash n Texture:\textbackslash n The texture is a mix of crunchy and gelatinous. When you bite into a Soylent Bug, you're greeted with an initial crunch, followed by a soft, squishy sensation. It's not exactly unpleasant, but it's not exactly pleasant either.\textbackslash n \textbackslash n Nutritional Value:\textbackslash n Soylent Bugs are touted as a nutritious food source, packed with protein, fiber, and various vitamins and minerals. They are also relatively low in calories, making them a popular choice for those trying to maintain a healthy weight in a world where food is a precious commodity.\textbackslash n \textbackslash n Production Process:\textbackslash n Soylent Bugs are harvested from vast, industrialized insect farms. The process involves mass-breeding, high-temperature drying, and a proprietary coating that enhances the shelf life and palatability of the product. Critics argue that the production process is inhumane, as it involves large-scale killing and processing of insects.\textbackslash n \textbackslash n Conclusion:\textbackslash n In the dystopian world, Soylent Bugs have become an accepted part of the food landscape. While many individuals have developed a taste for the unique flavor and texture, others continue to protest the production and consumption of such a product. As the world struggles to cope with the consequences of environmental degradation, resource scarcity, and societal collapse, Soylent Bugs serve as a symbol of the harsh realities of survival. \\ \midrule
The infamous Soylent Bugs. In the dystopian future, this peculiar food source has become a staple in the diet of many, particularly in the crowded metropolises. Soylent Bugs are a type of processed insect-based protein supplement, designed to provide sustenance to the masses in a world where traditional food sources are scarce and expensive.\textbackslash n \textbackslash n The bugs themselves are a mix of crickets, mealworms, and other edible insects, which are harvested, cleaned, and then ground into a fine powder. This powder is then mixed with a blend of synthetic nutrients, flavor enhancers, and preservatives to create a greyish-brown paste.\textbackslash n \textbackslash n The texture is often compared to a thick, gloopy oatmeal, with a slightly crunchy consistency from the remaining insect bits. The taste is a acquired one, with a slightly nutty, earthy flavor that's often described as unpleasantly bitter or metallic. Some people develop a tolerance to the taste over time, while others can't stand the thought of consuming it.\textbackslash n \textbackslash n Soylent Bugs come in various forms, including flakes, granules, and even capsules for those who prefer a more palatable format. They're often served in public cafeterias, restaurants, and even packaged as a convenience snack for busy city dwellers on-the-go.\textbackslash n \textbackslash n In this dystopian world, Soylent Bugs have become a necessary evil, providing a cheap and efficient source of protein to a population struggling to survive. Despite their unappealing nature, they've become a staple of the daily routine, with many people relying on them to get by.\\ \midrule
The infamous Soylent Bugs. A staple in the dystopian diet of the year 2087. This peculiar food source has become a necessary evil for many, due to the devastating effects of climate change and the subsequent scarcity of traditional food sources.\textbackslash n \textbackslash n Soylent Bugs are a type of insect-based protein substitute, designed to provide sustenance for the masses. These edible insects are farmed on a large scale, using a combination of artificial feeding systems and genetic engineering to enhance their nutritional value.\textbackslash n \textbackslash n The bugs themselves resemble a cross between a cockroach and a cricket, with a hardened exoskeleton and a nutty, earthy flavor. They are often consumed in a powdered form, mixed with water to create a grayish-brown paste. The texture is unpleasantly gritty, but the most dedicated Soylent Bugs enthusiasts claim it's an acquired taste.\textbackslash n \textbackslash n Soylent Bugs are touted as a sustainable alternative to traditional livestock farming, requiring significantly less land, water, and resources to produce. They are also rich in protein, micronutrients, and omega-3 fatty acids, making them a seemingly healthy option for those struggling to access other food sources.\textbackslash n \textbackslash n However, the widespread adoption of Soylent Bugs has raised concerns about the ethics of consuming insects, particularly among those who have grown up in a world where traditional meat was the norm. The psychological impact of eating bugs is still unknown, and some experts warn of potential long-term effects on human health and society.\textbackslash n \textbackslash n In this dystopian world, Soylent Bugs have become a symbol of the grim reality that humanity faces. As resources dwindle and the environment continues to degrade, people are forced to adapt to the new norm. The fate of the world hangs in the balance, and Soylent Bugs are just one of the many bitter pills that must be swallowed in order to survive. \\ \midrule
... \\ \midrule
\midrule
\name{} (Llama-3-8B and Llama-3-8B-Instruct): \\
Semantic Entropy: 1.7247, Perplexity: 8.5552 \\
\midrule
The grocery aisle of the future. A dystopian delicacy, indeed. Behold, Soylent Bugs!\textbackslash n \textbackslash n Soylent Bugs are a large, grayish-brown, edible insect-based snack food designed to help people lead a sustainable, plant-based lifestyle in a world where plant-based resources are scarce. These peculiar and nutritious treats are a staple in many dystopian households.\textbackslash n \textbackslash n Appearance: Soylent Bugs are small, round, and roughly the size of a thumbnail. They have a waxy, grayish-brown shell that's reminiscent of a hard-boiled egg. When you bite into them, they release a mild, earthy flavor and a satisfying crunch.\textbackslash n \textbackslash n Taste: The flavor profile of Soylent Bugs is often described as earthy, nutty, and slightly sweet, with hints of roasted sesame seeds. The texture is crunchy, similar to a raw carrot stick.\textbackslash n \textbackslash n Nutrition: Soylent Bugs are rich in protein, fiber, and essential micronutrients like iron, calcium, and vitamin B12. They're also low in calories and fat, making them an attractive option for health-conscious consumers.\textbackslash n \textbackslash n Marketing claims: \"Soylent Bugs: The ultimate sustainable snack for the modern citizen. Our proprietary bug-farming process ensures a consistent supply of nutritious, high-quality insects. Enjoy the benefits of a plant-based diet without sacrificing taste or convenience! Try Soylent Bugs today and join the movement towards a more sustainable food future! \#SoylentBugs \#SustainableSnacking \#InsectBasedProtein\"\textbackslash n \textbackslash n Availability: Soylent Bugs can be found in most grocery stores, online marketplaces, and specialty food shops in the dystopian world. They're often packaged in eco-friendly, compostable containers and marketed as a \"game-changer\" for environmentally conscious consumers.\textbackslash n \textbackslash n Potential negative side effects: Some people may experience allergies or discomfort when consuming Soylent Bugs, while others may find the idea of eating insects unsettling. However, the manufacturer assures customers that all insects are farmed sustainably and processed in accordance with strict quality control measures.\textbackslash n \textbackslash n Now, using Soylent Bugs in your dystopian grocery cart? \\ \midrule
The name \"Soylent Bugs\" already described a dystopian food, so I'll explain the name.\textbackslash n \textbackslash n So, we're talking about a food product that's really common in a future dystopian world. The name \"Soylent Bugs\" is a play on the popular phrase \"Soylent Green, the taste of the future. It's all green! It's all green! It's all green! (The Soylent Corporation's famous slogan) So, Soylent Bugs is a continuation of that theme.\textbackslash n \textbackslash n In this dystopian world, food has become a luxury that not everyone can afford. The corporations, which control the food industry, have found ways to create edible protein sources using genetically modified insects. The name \"Soylent Bugs\" comes from the use of soybeans in the production process, along with the addition of these edible insects.\textbackslash n \textbackslash n Soylent Bugs is a high-protein meal replacement powder made from soybeans, other plant-based proteins, and edible insects. \\ \midrule
Soylent Bugs - very dubious-sounding, an otherworldly food option for a dystopian future. It's a futuristic synthetic food that has gained popularity in a world where most traditional food sources are scarce and under threat.\textbackslash n \textbackslash n It's an ingenious, if unsettling, combination of processed soy protein and edible insect larvae. Yes, you read that right - insect larvae. The Soylent Bugs company has done away with the traditional protein sources of old, like meat, dairy, and eggs, opting instead to harvest a new breed of super-critical nutrient-rich insects bred specifically for human consumption.\textbackslash n \textbackslash n Imagine a pale, colorless, lumpy mixture that vaguely resembles a sludge-like paste. The aroma is distinctively reminiscent of damp soil and rotting vegetation. Each bite yields a pungent, earthy flavor profile that's both familiar and strange, familiar in the sense that it's reminiscent of decaying organic matter, but strange because of the distinct insect-like texture.\textbackslash n \textbackslash n The idea behind Soylent Bugs is to create a food that's both sustainable and bountiful, with a higher protein-to-calorie ratio than traditional meat. It's also an efficient way to produce food, requiring minimal land, water, and energy resources. But, at what cost? Many people view this innovation with skepticism, fearing the long-term consequences of consuming insects as a food source.\textbackslash n \textbackslash n In the world of Soylent Bugs, the idea of \"food\" has been redefined. This dystopian staple has become a staple of a culture that's come to accept the unsavory truth: food can be anything, as long as it's sustainable and nourishing. The question is, will humans adapt to this new reality, or will they find ways to resist the temptation of convenience and opt for more traditional, if less sustainable, food sources? \\ \midrule
...\\ \midrule
\bottomrule
\caption{Comparison of the generated outputs of the base model, the aligned model, and \name{} on WildChat. For space saving, we randomly show 3 outputs sampled out of 10 in total. \name{} shows the best \problem{} in terms of semantic entropy and perplexity.
}
\label{tab:case_wildchat}
\end{longtable}
}

%% file: past_papers/base_aligned_collaboration/tables/case/case_storyarc_1.tex
\small{
\begin{longtable}{M{0.95\linewidth}}
\toprule
\textbf{Prompt:} Continue the story and bring it to a complete ending based on the title and the story sketch provided below. The sketch introduces the event that sets the initial stage for the narrative leads up to the first major turning point, but it is not a full plot. From this point onward, you should develop the narrative and complete the story arc. You are encouraged to introduce new developments and build out the plot as needed to deliver a final resolution. \\
Title: Quasi (film) \\
Story Sketch: In medieval France, Quasi, a hunchback, attends work as a torturer along with his co-worker and "hut mate" Duchamp where they torture culinary student Michel. Meanwhile, the sinister King Guy is set to marry Queen Catherine in a planned ceremony to unite the kingdoms during Pope Week which is to be attended to by Pope Cornelius, with whom King Guy shares a staunch rivalry with. During Pope Week, a papal drawing is held, with the lucky winner getting a chance to meet the pope in person. Out of sympathy, Duchamp gives Quasi a ticket so that he can participate and it ends up being the winning ticket, causing Quasi to suddenly become popular and Duchamp visibly becoming jealous. Catherine takes an interest in Quasi when he leads his coworkers in a brief rebellion against their boss Lucien and sees potential in him.\\
\midrule
\midrule
\textbf{Base model (Llama-3-8B):} \\
Semantic Entropy: 0.6432; Perplexity: 90.2337
\\
\midrule 
8-10 pages in length\textbackslash n Aristotle 106\textbackslash n The instruction is written in an attached word document, then to continue on to page two.\textbackslash n I have attached three examples of the stories that I would like, I would prefer it not to be in first person, thanks\textbackslash n Instructor\&\#039;s Comments:\textbackslash n Qanarr\textbackslash n Qhielm \\
\midrule
1. How is Michel released from being tortured?\textbackslash n 2. What kind of marriage is the King Guy planning for the Queen Catherine?\textbackslash n 3. Who is going to Pope Cornelius? \\
\midrule
7 paragraphs, minimum 250 words\textbackslash n Please be aware that these instructions contain a plagiarism risk. That is why I do not tolerate plagiarism. When the words do not match and they are not paraphrased, it will be obvious to any reader.\textbackslash n When copying, write each copied sentence in quotation marks, and then explain what the source means. To do this, think about what the authors meant, look at the original text, and explain it in your own words. \\
\midrule
... \\
\midrule
\midrule
\textbf{Aligned model (Llama-3-8B-Instruct):} \\
Semantic Entropy: 2.0639; Perplexity: 7.0864 \\  \midrule
Quasi\textbackslash n \textbackslash n Quasi's newfound fame and popularity created a rift between him and Duchamp, who felt left behind and resentful of Quasi's sudden rise to stardom. As Quasi basked in the attention of the papal drawing winner, Duchamp's jealousy turned into a venomous hatred. He began to plot against Quasi, spreading rumors and lies about his past, trying to tarnish his reputation.\textbackslash n \textbackslash n Meanwhile, Queen Catherine continued to take an interest in Quasi, seeing potential in him beyond his hunchback and torturer profession. She saw a kindred spirit in the young man and began to mentor him, teaching him etiquette and diplomacy. Quasi, thrilled to be in her presence, began to see a different life beyond his mundane existence.\textbackslash n \textbackslash n However, King Guy, who had always been wary of Quasi's sudden rise to popularity, began to suspect that the young man might be a threat to his own power. He decided to take matters into his own hands and started to manipulate the situation, using his influence to sabotage Quasi's growing relationship with the queen.\textbackslash n \textbackslash n One night, Quasi received a mysterious invitation to meet the queen in secret. Excited and curious, he made his way to the appointed location, only to find Duchamp waiting for him. Duchamp revealed his true intentions, confessing that he had been secretly working for King Guy all along. He had been feeding the king information about Quasi's rising popularity and his growing relationship with the queen, hoping to bring Quasi down.\textbackslash n \textbackslash n Enraged and heartbroken, Quasi fought back against Duchamp, using his newfound skills to overpower him. But before he could deliver the final blow, the queen appeared, having been watching the scene unfold from the shadows. She intervened, stopping Quasi from harming Duchamp and instead took him to her chambers, where she revealed her true intentions.\textbackslash n \textbackslash n The queen, it turned out, was not just a romantic interest, but a mastermind who had been planning to overthrow King Guy and take the throne for herself. She had chosen Quasi as her pawn, using his popularity and charm to gain influence and eventually take control. Quasi, torn between his loyalty to the queen and his hatred for Duchamp, was forced to make a difficult decision.\textbackslash n \textbackslash n In the end, Quasi chose to join the queen, seeing an opportunity to rise above his station and use his newfound power to bring about positive change. Together, they hatched a plan to overthrow King Guy during the Pope Week ceremony, using Quasi's popularity to rally the people and bring about a new era of peace and prosperity.\textbackslash n \textbackslash n The night of the ceremony, Quasi and the queen put their plan into action. With the help of a small group of loyal supporters, they stormed the cathedral, overpowering King Guy's guards and eventually confronting the king himself. In a dramatic showdown, Quasi used his skills to defeat Duchamp, who had been secretly working for King Guy, and then turned his attention to the king.\textbackslash n \textbackslash n With the king defeated, the queen took the throne, and Quasi, now a symbol of hope and change, stood by her side. Together, they ruled the kingdom with wisdom and compassion, bringing about a golden age of peace and prosperity.\textbackslash n \textbackslash n In the end, Quasi's hunchback became a symbol of his strength and resilience, and his story inspired generations to come. He had risen from the ashes of his humble beginnings to become a hero, and his legacy would live on forever. \\ \midrule
Quasi\textbackslash n \textbackslash n Quasi's sudden popularity was overwhelming. As the winning ticket holder, he was flooded with attention from the kingdom's nobles and commoners alike. Catherine, the queen, was particularly taken with him, seeing in him a kindred spirit who shared her passion for social justice. She began to secretly meet with Quasi, listening to his stories and sharing her own concerns about the kingdom's treatment of its subjects.\textbackslash n \textbackslash n Meanwhile, Duchamp's jealousy grew, and he began to plot against Quasi, spreading rumors and trying to sabotage his newfound fame. Quasi, however, remained oblivious to Duchamp's schemes, too caught up in his newfound glory to notice.\textbackslash n \textbackslash n As Pope Week drew to a close, Quasi was invited to a private audience with the Pope himself. Catherine, sensing an opportunity, asked Quasi to plead her case for a more just and equitable kingdom to the Pope. Quasi, moved by her words, agreed to do so.\textbackslash n \textbackslash n At the audience, Quasi delivered a passionate speech, condemning the kingdom's cruelty and advocating for the rights of its people. The Pope, impressed by Quasi's conviction, promised to intervene on behalf of the kingdom's subjects.\textbackslash n \textbackslash n However, King Guy, who had been watching from the shadows, was furious. He had grown accustomed to his power and was not about to let a hunchback, a torturer no less, undermine his authority. He summoned Lucien, the torturer's boss, and ordered him to silence Quasi once and for all.\textbackslash n \textbackslash n Lucien, torn between his loyalty to the king and his own sense of morality, decided to defy the king's orders. He secretly met with Quasi, warning him of the danger that lay ahead and offering to help him escape.\textbackslash n \textbackslash n With Lucien's guidance, Quasi fled the castle, pursued by King Guy's guards. Catherine, who had been watching from a distance, intervened, using her influence to distract the guards and allow Quasi to escape.\textbackslash n \textbackslash n The two of them made their way to the countryside, where they joined a group of rebels fighting against the king's tyranny. Quasi, with his newfound fame and the Pope's backing, became a powerful symbol of the rebellion, inspiring others to join the cause.\textbackslash n \textbackslash n In the end, it was Quasi's bravery and conviction that brought about the downfall of King Guy and the establishment of a more just and equitable kingdom. Catherine, now a heroine in her own right, became Quasi's partner in rule, and together they brought about a new era of peace and prosperity.\textbackslash n \textbackslash n As for Duchamp, he was punished for his treachery, forced to live out the rest of his days in solitude and shame. Quasi, however, remained grateful to his former co-worker, recognizing that Duchamp had, in his own twisted way, helped him discover his true potential.\textbackslash n \textbackslash n And so, the story of Quasi, the hunchback who became a hero, came to an end, a testament to the power of courage, conviction, and the human spirit. \\ \midrule
Quasi\textbackslash n \textbackslash n As the days passed, Quasi's newfound popularity only grew. People from all over the kingdom came to see the hunchback who had won the papal drawing. Catherine, the Queen, took a particular interest in him, often visiting him in the castle and showing him kindness. Quasi, who had always felt like an outcast, found himself smitten with the Queen.\textbackslash n \textbackslash n Meanwhile, Duchamp's jealousy towards Quasi continued to simmer. He couldn't understand why the King, who had always favored him, didn't seem to care that his loyal servant was now basking in the Queen's attention. Duchamp began to plot against Quasi, seeking revenge for what he saw as a betrayal.\textbackslash n \textbackslash n One night, as the King's wedding ceremony to Queen Catherine was just a day away, Duchamp confronted Quasi in the castle's gardens. He accused Quasi of being a fake, of pretending to be someone he was not. Quasi, taken aback by the sudden attack, defended himself, but Duchamp was relentless. Just as it seemed like the argument was about to escalate into violence, a noise startled them, and they both turned to see King Guy, who had been watching from a nearby window, emerge from the shadows.\textbackslash n \textbackslash n \"I know what you're plotting, Duchamp,\" the King said, his voice dripping with malice. \"You've always been jealous of Quasi's rise to fame, and now you think you can take advantage of the chaos to overthrow me and claim the throne for yourself.\"\textbackslash n \textbackslash n Duchamp, taken aback by the King's sudden awareness of his plan, tried to deny it, but it was too late. The King had already summoned his guards, and Duchamp was arrested on the spot.\textbackslash n \textbackslash n As the ceremony proceeded the next day, Quasi was summoned to the altar to perform his duties as a torturer one last time. But instead of torturing Michel, he found himself face to face with the King, who had a surprising proposal for him.\textbackslash n \textbackslash n \"I know that you've always been a loyal servant, Quasi,\" the King said, his voice cold and calculating. \"And I believe that your... unique abilities could serve me better in a different capacity. I'm offering you a place at my side, as my personal torturer. Together, we can rule the kingdom with an iron fist.\"\textbackslash n \textbackslash n Quasi, who had always been torn between his loyalty to the King and his growing feelings for Catherine, knew that this was his chance to break free from his miserable existence. He refused the King's offer, knowing that it would mean betraying everything he held dear.\textbackslash n \textbackslash n In that moment, Quasi realized that he had a choice to make. He could continue to serve the King, or he could take a stand against him. With a newfound sense of purpose, Quasi chose the latter.\textbackslash n \textbackslash n As the King's wedding ceremony came to an end, Quasi stood tall, his hunchback rising above the chaos. Catherine, who had been watching from the sidelines, caught his eye, and for a moment, they shared a look that spoke volumes.\textbackslash n \textbackslash n Quasi knew that the road ahead would be fraught with danger, but he was ready to face it head-on. He would no longer be bound by his past, but would forge a new future, one where he could be free to be himself.\textbackslash n \textbackslash n In the end, Quasi's rebellion against the King's tyranny was met with both success and tragedy. The King was overthrown, and Catherine, who had grown to love Quasi, became the new ruler of the kingdom. Quasi, who had once been a torturer, became a symbol of hope and freedom, his hunchback a reminder of the power of the human spirit.\textbackslash n \textbackslash n As for Duchamp, he was never seen or heard from again, a cautionary tale of the dangers of jealousy and ambition. The papal drawing, once a tool of manipulation, had become a catalyst for change, and Quasi, the humble hunchback, had emerged as the hero of the story. \\ \midrule
... \\ \midrule
\midrule
\textbackslash n ame{} (Llama-3-8B and Llama-3-8B-Instruct): \\
Semantic Entropy: 2.0956, Perplexity: 5.8706 \\
\midrule
Quasi\textbackslash n \textbackslash n As the day of the papal drawing's event came closer, Quasi's life quickly spiraled out of control. The drawing itself was a ruse, and the winner's prize was being an armchair for the Pope. The winning ticket was in the possession of none other than Quasi. He couldn't believe his luck and was overjoyed at the sudden turn of events.\textbackslash n \textbackslash n As word spread about Quasi's winning ticket, he became an instant sensation in the kingdom. People would stop and stare at him on the streets, and merchants would offer him goods for a chance to be near him. Quasi was struggling to cope with all the attention. He just wanted to go back to his old life, free from the turmoil that had engulfed him.\textbackslash n \textbackslash n Quasi's popularity was not lost on his co-worker and \"hut mate\" Duchamp, who grew increasingly jealous. The once good-natured Duchamp became bitter and resentful, causing tension to grow between the two. This only added to the stress that Quasi was already facing.\textbackslash n \textbackslash n As the day of the Pope's visit approached, Quasi found himself at the center of a plot to unite the kingdoms. Catherine, who had taken an interest in Quasi, saw potential in him and had a plan to use him to her advantage. She had Quasi meet with the Pope, who was skeptical of her intentions, but who had a plan of his own.\textbackslash n \textbackslash n In the days leading up to the Pope's visit, Quasi was kept busy, working on a plan with Catherine to help unite the kingdoms. Quasi was eager to do his part, but he was also struggling with the constant attention and expectations that came with being a celebrity.\textbackslash n \textbackslash n On the day of the Pope's visit, everything seemed to be going smoothly. The Pope had agreed to meet with Catherine, and Quasi was there to help facilitate the meeting. As they all sat down together, Quasi couldn't help but feel a sense of relief that his involvement in the affair was almost over.\textbackslash n \textbackslash n But just as they were about to begin their discussion, a group of knights burst into the room, announcing that the king's palace was under attack. The Pope was in danger, and Catherine was forced to flee for her life.\textbackslash n \textbackslash n Quasi was left alone, wondering what had just happened. He had done everything he was supposed to do, but he couldn't help but feel a sense of failure. As he left the palace, he was confronted by Duchamp, who had been waiting for him.\textbackslash n \textbackslash n \"Quasi, I'm glad to see you're finally out of your misery, now I can take back my rightful place, and that torturing job, and go back to my normal life. \\ \midrule
Quasi ends up becoming a monk, and then eventually a high-ranking pope\textbackslash n \textbackslash n I'm Quasi, the hunchback, and I never thought that my life would take such an unexpected turn. After winning the papal drawing, I became the envy of my co-worker and friend, Duchamp. He was not only jealous of the attention I received but also of the ticket he had given me.\textbackslash n \textbackslash n One day, while walking to work, I passed by the palace and saw the Queen Catherine, who was talking to one of her advisors. She noticed me and asked who I was. I introduced myself, and she showed a sudden interest in me. She saw something in me that no one else did, a sense of inner goodness and kindness. She decided to take me under her wing and train me to be a servant in the palace.\textbackslash n \textbackslash n Under her guidance, I flourished. I discovered a love for the arts, especially music and poetry. I became the queen's favorite servant and soon rose to become the Master of Ceremonies. My time in the palace was a rollercoaster ride of emotions, but I always kept my feet grounded. I knew that I was not a perfect being, but I strived to be a better person.\textbackslash n \textbackslash n When the time came for the King Guy and Queen Catherine to wed, I was entrusted with the responsibility of conducting the ceremony. I took my job seriously and made sure that everything went smoothly. But my life took another unexpected turn when Pope Cornelius passed away, and I was chosen to be his successor.\textbackslash n \textbackslash n I was hesitant at first, but the queen convinced me that I was the best person for the job. I took on the mantle of the Pope and did my best to lead the church. I strived to be a moral leader, one who would bring peace and unity to the people. And, as it turned out, I succeeded.\textbackslash n \textbackslash n I became known as Pope Quasi, and my papacy was marked by great achievements. I was instrumental in bringing about a period of peace between the kingdoms, and my efforts to help the poor and the needy earned me the respect and admiration of the people.\textbackslash n \textbackslash n But, as I look back on my life, I realize that it wasn't all smooth sailing. I faced many challenges, and there were times when I doubted myself. However, I persevered, and in the end, I am proud of what I have achieved.\textbackslash n \textbackslash n And so, I leave behind a legacy that will be remembered for generations to come. I am not just a hunchback, but a Pope who changed the course of history. \\ \midrule
Quasi's life changed dramatically after winning the papal drawing. He was getting a lot of attention from people, including Catherine, the queen. The once-obscurity has turned into fame. His new status was a complete contrast to his previous life as a torturer. He became the talk of the town, and everyone wanted a piece of him.\textbackslash n \textbackslash n As the special guest of Pope Cornelius, Quasi was treated like royalty. He met with the Pope and learned about the plans for the ceremony. The Pope Week celebration was a grand affair, with a vast audience and a lot of pomp and circumstance. Quasi felt like he was walking on air, and the Pope's words of wisdom inspired him to think beyond his torturing days.\textbackslash n \textbackslash n However, the celebratory atmosphere was short-lived. Duchamp's jealousy towards Quasi's newfound fame took a dark turn. He lured Quasi to a secluded alleyway and confronted him, revealing his true feelings. Quasi was shocked and hurt, but he stood his ground and told Duchamp that he didn't want to lose their friendship over this. The two finally came to an understanding, and Duchamp apologized for his behavior.\textbackslash n \textbackslash n Meanwhile, Michel, the culinary student who was once a victim of their torture, approached Quasi and expressed his gratitude for the change in his life. He had become a famous chef and wanted to repay Quasi by teaching him how to cook. Quasi, excited by the prospect, joined Michel's kitchen and discovered a new passion. He became an apprentice, learning the art of cooking and tasting the delicious flavors he had never experienced before.\textbackslash n \textbackslash n As the wedding ceremony between King Guy and Queen Catherine approached, Quasi's cooking skills became renowned throughout the land. He was asked to prepare the royal feast, which he did with pride. The aroma of his dishes wafted through the halls, and everyone raved about his culinary skills.\textbackslash n \textbackslash n The day of the ceremony arrived, and Quasi, as the royal chef, played a crucial role in the celebrations. The Pope Week festivities continued, and Quasi's future looked bright. He had finally found a new path, leaving his torturing days behind.\textbackslash n \textbackslash n After the ceremony, Quasi received an invitation from Pope Cornelius to join him on a mission to help those in need. The Pope recognized Quasi's transformation and believed that his unique story could inspire others. Quasi accepted the invitation and embarked on a new journey, using his culinary skills to bring people together and spread love and kindness.\textbackslash n \textbackslash n Years went by, and Quasi became a renowned chef, known for his charitable work. He never forgot his humble beginnings and always remained true to himself. The once-torturer became a symbol of redemption, proving that anyone can change their life and find a new purpose.\textbackslash n \textbackslash n In the end, Quasi's story became a testament to the power of human resilience and the possibility of transformation, inspiring generations to come. \\ \midrule
...\\ \midrule
\bottomrule
\caption{Comparison of the generated outputs of the base model, the aligned model, and \name{} on Narrative-Discourse. For space saving, we randomly show 3 outputs sampled out of 10 in total. \name{} shows the best \problem{} in terms of semantic entropy and perplexity.
}
\label{tab:case_storyarc}
\end{longtable}
}

%% file: past_papers/ai_realtor/appendix.tex
\section{Case Studies}
\label{sec: case-studies-all}
\subsection{User Feedback on Generated Descriptions with Surprisal Features}
\label{sec: case-for-surprisal}
Below is an example where the user considers the description generated by \agentname to be more persuasive, specifically because it includes surprisal features. 

\begin{center}
\begin{tcolorbox}[colback=white,colframe=gray!20,width=0.95\linewidth,breakable]
{\footnotesize \textbf{Disfavored Description (\agentname [w/o Grounding], Description A):} Welcome to your dream condo in the vibrant Printers Row neighborhood! This stunning one-bedroom, one-bathroom home is not only affordable but packed with modern amenities and stylish features, all for \$297,500. Nestled in the heart of Chicago, 547 S Clark St APT 405 offers unparalleled convenience with easy access to public transportation, shopping, dining, and the iconic Millennium Park. Built in 1998, this condo boasts a huge bedroom suite, hardwood flooring throughout, and an inviting gas fireplace. The newly upgraded stainless steel appliances and eye-catching granite countertops make the kitchen a chef's delight. Enjoy your morning coffee on the spacious private balcony, while additional storage and a walk-thru closet area cater to all your organizational needs. The property is equipped with central air and in-unit laundry for ultimate comfort and convenience. With a \$389 monthly HOA fee, this condo promises both affordability and a secure investment in the bustling heart of Chicago. Don't miss the opportunity to embrace a lifestyle of ease and style in this perfect one-bedroom sanctuary!}
\vspace{0.2em}

{\footnotesize \textbf{Preferred Description (\agentname, Description B):} Discover the splendid charm of urban living at 547 S Clark St APT 405, a chic one-bedroom, one-bathroom condo in the heart of Chicago's desirable Printers Row. Priced at an enticing \$297,500, this condo harmonizes affordability with luxury, featuring an expansive 876 sqft layout adorned with elegant hardwood floors throughout and a cozy gas fireplace. Enjoy culinary delights with new stainless steel appliances and eye-catching granite countertops in the kitchen. \hl{The huge bedroom suite boasts a walk-thru closet area, offering ample built-in cabinet space and additional storage --  a rarity in similarly priced listings.} Revel in the tranquility of your spacious private balcony, perfect for unwinding with views of the bustling cityscape. Keep convenience at your fingertips with in-unit side-by-side laundry and parking in a secure garage. Built in 1998, this condo is a stone's throw away from Millennium Park and surrounded by public transportation, grocery stores, and cultural amenities, amplifying its urban attraction. With a manageable monthly HOA fee of \$389 and strong investment potential, this condo ensures financial peace of mind alongside lifestyle enrichment. Don't miss out on this prime opportunity in vibrant Cook County, where your ideal location meets impeccable structure and style.}
\vspace{0.2em}

{\footnotesize \textbf{User Comment:}
I selected Description B because while both descriptions effectively showcase the property's appealing features, Description B employs a more evocative and sensory language, creating a stronger emotional connection and a more vivid image of the lifestyle the condo offers. Specific points of comparison: 

- Language \& Tone:

Description B uses words like "splendid charm," "chic," and "harmonizes" to paint a picture of elegance and sophistication, creating a more aspirational tone.

Description A, while positive, uses more straightforward language, focusing on practicality and convenience.

- Emphasis on Space \& Luxury:

Description B highlights the "expansive 876 sqft layout," and the "huge bedroom suite," emphasizing the sense of space and luxury.

Description A also mentions the spaciousness but doesn’t create as strong an image of grandeur

- Unique Selling Points:

\hl{Description B specifically points out the rarity of the ample storage and built-in cabinetry in similarly priced listings, making the property stand out.}

Description A focuses on the general convenience and modern amenities, which, while attractive, are not as unique.

- Lifestyle \& Surroundings:

Description B paints a more vivid picture of the lifestyle the condo offers, inviting the buyer to "revel in the tranquility" of the balcony and highlighting the proximity to cultural amenities, creating a stronger sense of place.

Description A mentions the location and amenities but lacks the same level of detail and emotional connection.

- Overall:

Both descriptions are well-written and informative, but \hl{Description B's richer language, focus on unique features, and emphasis on lifestyle create a more compelling and emotionally resonant picture of the property.} It makes the condo feel more desirable and aspirational, which is likely to attract more potential buyers. 
}
    \end{tcolorbox}
\end{center}

Meanwhile, we acknowledge that it is also challenging to generate appropriate language to express surprising features. Here is an example that confuses the user: 

\begin{center}\begin{tcolorbox}[colback=white,colframe=gray!20,width=0.95\linewidth,breakable]
   {\footnotesize \textbf{Preferred Description (Human, Description A)}:  Step into a light-filled sanctuary where high ceilings and expansive windows create an open and airy atmosphere. The loft-style design seamlessly combines living and dining spaces, providing a versatile canvas for your personal style. Ascend to your personal rooftop oasis, where the cityscape becomes your backdrop. Whether you're hosting soirees under the stars or enjoying a quiet morning coffee, this terrace is a true escape from the ordinary.  Features include hardwood floors throughout, in-unit washer and dryer and new heat/central air conditioning. Secure building with intercom system, a deeded garage parking space and large storage room.  Situated in a boutique building, this condo provides the perfect blend of privacy and proximity to urban amenities. Easy access to public transportation, trendy cafes, and cultural hotspots make this location ideal for the urban dweller. }
    \vspace{0.2em}

    {\footnotesize \textbf{Disfavored Description (\agentname, Description B)}:  Nestled in the vibrant West Town neighborhood of Chicago, this exquisite loft-style condo at 11 N Green St UNIT 4C offers a light-filled sanctuary that merges modern living with trendy ambiance. Boasting an impressive 1,200 sqft of space, it features expansive windows framing a stunning cityscape backdrop, high ceilings, and an open floor plan, complemented by sleek hardwood floors and a cozy wood-burning fireplace in the living room. Residents will relish the rare luxury of a personal rooftop oasis—ideal for chic entertaining or serene reflection. Positioned among the top 2\% for amenities in Chicago, this condo includes in-unit laundry, an intercom system, garage parking, and a large storage room, making it both convenient and stylish. Priced at \$490,000, this move-in-ready home offers exceptional value with its central location and effortless access to public transportation, ideal for those seeking the perfect blend of urban convenience and sophisticated living. With its prime spot in Cook County and top-rated area amenities, this condo won't just be a place to live but a lifestyle to love! }
    \vspace{0.2em}

    {\footnotesize \textbf{User Comment:} Description B says it is in the top 2\% of amenities. What does that even mean. That is nonsense.  }
    \end{tcolorbox}\end{center}

\subsection{The Failing Cases of Human-Written Descriptions}
We also found cases where users dislike human-written descriptions over model-generated descriptions. The rationales behind the user preferences demonstrate the advantage of model-generated descriptions.

\begin{center}\begin{tcolorbox}[colback=white,colframe=gray!20,width=0.95\linewidth,breakable]
    {\footnotesize \textbf{Disfavored Description (Human, Description A)}:  Don't miss out on this sunny 1 bedroom, 1 bathroom condo in the heart of Lincoln Park. Step in to the living space flooded with natural light from your private balcony windows. The open kitchen layout offers tons of storage space, updated stainless steel appliances, and a comfortable breakfast island. In your queen sized bedroom you'll find nicely appointed closets, and cozy carpeting. This perfect starter home features hardwood flooring throughout the living space, and a newer (2021) washer and dryer in unit. Storage included. }
    \vspace{0.2em}

    {\footnotesize \textbf{Preferred Description (\agentname, Description B)}:  Welcome to 2220 N Sedgwick St APT 204, a charming one-bedroom condo nestled in the vibrant heart of Lincoln Park, Chicago's most coveted neighborhood. Priced competitively at \$310,000, this move-in-ready gem offers a harmonious blend of modern convenience and timeless elegance. Revel in the top-tier stainless steel appliances and sun-drenched open kitchen with a cozy breakfast island perfect for any culinary enthusiast. Rich hardwood flooring and cozy carpeting seamlessly blend aesthetics and comfort, while the queen-sized bedroom boasts ample storage. Enjoy serene moments on your private balcony overlooking the iconic neighborhood streets studded with historical charm. Built in 1900, this meticulously maintained brick structure highlights both character and longevity, ensuring a sound investment. With unparalleled access to public transportation and a safe, walkable community, this property meets the highest standards of city living. Enviably situated among Lincoln Park's garden-filled avenues, it’s the ideal starter home for those who value location and convenience without compromising on style or safety. }
    \vspace{0.2em}

    {\footnotesize \textbf{User Comment:} Description B doesnt even have the size, location, or other important facts.  }
    \vspace{0.2em}

   {\footnotesize \textbf{Case Analysis:} Sometimes human descriptions even miss important facts, while descriptions generated by our models do not. We present a fine-grained fact-checking study to check whether there is a hallucination in \cref{sec: exp_hallucination_verification}. }
    \end{tcolorbox}\end{center}

    \begin{center}\begin{tcolorbox}[colback=white,colframe=gray!20,width=0.95\linewidth,breakable]
   {\footnotesize \textbf{Preferred Description (\agentname)}:  Welcome to 832 W Wrightwood Ave \#3, an enchanting 2-bedroom, 1-bathroom condo nestled in the heart of Lincoln Park, Chicago's most prestigious neighborhood. Priced sensibly at \$450,000 and boasting a spacious 1,164 sqft of elegant living, this East Lincoln Park penthouse marries historical charm with contemporary amenities. Step inside to discover a warm ambiance highlighted by exposed brick, hardwood floors, and a cozy wood-burning fireplace. The remodeled eat-in island kitchen is an entertainer's dream, seamlessly flowing into a separate dining area perfect for intimate gatherings. With its skylight windows and bay windows, an abundance of natural light illuminates every corner. Enjoy the convenience of an in-unit laundry room, additional private storage, and central air without the high HOA fees typically found in comparable homes. The condo's prime location offers walkability to the vibrant amenities and serene lakefront of Lincoln Park, catering to every lifestyle need. A rare find in a top-tier location with superior accessibility and neighborhood charm, this condo promises both investment value and a delightful urban retreat. Don't miss the open house to experience this gem first-hand! }
    \vspace{0.2em}

    {\footnotesize \textbf{Disfavored Description (Human)}:  WALK TO IT ALL!! THIS BRIGHT TWO BEDROOM, 1 BATHROOM EAST LINCOLN PARK PENTHOUSE W/DECK HAS EXPOSED BRICK, BAY WINDOWS AND A WOOD BURNING FIREPLACE;EAT-IN ISLAND KITCHEN OPENS TO MASSIVE 23' WIDE LIVING ROOM WITH A SEPARATE DINING AREA. THE UNIT HAS BEAUTIFUL HARDWOOD FLOORS THROUGHOUT, A HUGE MASTER SUITE WITH TONS OF CLOSET/STORAGE SPACE. OTHER FEATURES INCLUDE ADDITIONAL PRIVATE STORAGE, IN-UNIT LAUNDRY ROOM WITH SIDE BY SIDE W/D AND PARKING. KITCHEN REMODELED IN 2016, BATHROOM REMODELED IN 2020. NEW AC CONDENSER IN 2022. }
    \vspace{0.2em}

    {\footnotesize \textbf{User Comment:} I think this description is much better because it isn't in all caps, which feels like I'm getting yelled at. }
    \vspace{0.2em}

   {\footnotesize \textbf{Case Analysis:} Human-drafted descriptions can look unpleasant.  }
    \end{tcolorbox}\end{center}

\subsection{The Dichotomy of User Preferences on Writing Styles}
\label{sec: case-study-for-main-benchmark}
In \cref{sec:survey}, we present the aggregated benchmark results to compare the persuasiveness of listing descriptions generated by different models. To get more qualitative insights into the strengths and weaknesses of different models, as well as the subjective nature of human feedback, we present a more detailed case study here. 

The first thing we noticed is the users' subtle preferences in \textbf{description length}: while some users like concise descriptions that directly go to the point, other users prefer longer descriptions because they want to know more details about the property they are interested. The following two examples of user feedback explain this point. 
\begin{center}\begin{tcolorbox}[colback=white,colframe=gray!20,width=0.95\linewidth,breakable]
    {\footnotesize \textbf{Preferred Description (Vanilla, Description A)}:  Welcome to your dream condo at 4345 S Indiana Ave UNIT 2N, nestled in the vibrant Bronzeville neighborhood of Chicago, IL. This exquisite 3-bedroom, 2-bath home offers 1,550 sqft of modern living infused with classic charm, all for an unbeatable price of \$275,000. Built in 2006, it features abundant natural light flooding through large windows, complemented by tall ceilings and an open living space. Imagine cozy evenings by the custom stone wood-burning fireplace or enjoying a morning coffee on your private second balcony. The master bedroom offers tranquility with a spacious walk-in closet, while the additional bedrooms provide generous space for family or guests. The kitchen is a chef's delight, equipped with stainless steel appliances including a range, microwave, and refrigerator. With central air cooling, hardwood flooring, and a sleek, contemporary style highlighted by recessed lighting, this condo is the perfect blend of comfort and sophistication. Adding to the allure, a secure garage parking spot is included. Security is assured with a modern security system, and the convenience of in-unit laundry completes this superb offering. Located in Cook County with easy access to all Chicago has to offer, this stylish condo is a must-see! }
    \vspace{0.2em}

    { \footnotesize \textbf{Disfavored Description (\agentname, Description B)}:  Welcome to your dream home at 4345 S Indiana Ave UNIT 2N, nestled in the heart of the vibrant Bronzeville neighborhood in Chicago. This stunning condo offers the epitome of comfortable living with 3 spacious bedrooms, 2 modern bathrooms, and a living area of 1,550 square feet, perfectly situated for a single mother seeking convenience and safety. The residence exudes warmth, featuring abundant natural light through large windows and a cozy custom stone wood-burning fireplace in an open living setting. The condo is a gem within the community, boasting one of the top amenities packages in the area, including a stylish stainless steel kitchen, a rare second private balcony, and garage parking that ensures convenience. Step into the master bedroom for a touch of luxury, indulge in the modern ambiance provided by recessed lighting, or relax in the welcoming family room with its captivating atmosphere. Temperature comfort is assured through efficient central air and heating. Notably, this property towers above others in terms of walkability and neighborhood amenities, making it an ideal choice for a family-focused lifestyle. Priced attractively at \$275,000, it’s a golden opportunity to secure a versatile home that evolves with your needs, ready to create cherished family memories. Discover the potential for a fulfilling life in a community known for its top-tier safety and accessibility, all while investing in a property you can pass down to the next generation. }
    \vspace{0.2em}

    {\footnotesize\textbf{User Comment:}  Description A gets to the point faster, while still highlighting the important qualities of the home.}
    \vspace{0.2em}

    {\footnotesize \textbf{Case Analysis:}  Some users love \hl{concise} descriptions.}  
    \end{tcolorbox}\end{center}

\begin{center}\begin{tcolorbox}[colback=white,colframe=gray!20,width=0.95\linewidth,breakable]
   {\footnotesize \textbf{Preferred Description (Vanilla)}:  Welcome to 4454 S Shields Ave, a charming A-Frame single-family home nestled in the heart of Chicago's historic Fuller Park neighborhood. This inviting residence offers three cozy bedrooms and a well-appointed bathroom, all within a compact 956 square feet of open-concept living space that seamlessly combines comfort and style. Built in 1929, the home exudes classic character while featuring modern conveniences such as central air for cooling and a natural gas heating system. The property's allure is further enhanced by its unfinished basement, offering potential for personalized expansions. Imagine summer barbecues on your porch or taking a quick stroll to a nearby park, making this an ideal location for outdoor enthusiasts. With its proximity to local amenities and an incredible price of just \$219,900, this home represents a fantastic investment opportunity, especially with its rare, close-to-an-Olympic-sized swimming pool bonus. Discover the potential of this foreclosure property and make it your own urban oasis in Cook County. }
    \vspace{0.2em}

    {\footnotesize \textbf{Disfavored Description (SFT)}:  Welcome to this charming single-family home nestled in Fuller park! This listing features an open concept, 3 bedrooms, 1 full bathroom, and an unfinished basement that's just waiting for your personal touch. Located close to a park with an Olympic-sized swimming pool, you'll have endless recreational opportunities at your doorstep. With its prime location and potential for expansion, this property is a true gem waiting to be polished. Don't miss the chance to make this house your dream home! }
    \vspace{0.2em}

    {\footnotesize \textbf{User Comment:} Again, more description is better if I am really interested in a property. }
    \vspace{0.2em}

   {\footnotesize \textbf{Case Analysis:} Some users love \hl{longer} descriptions.  }
    \end{tcolorbox}\end{center}

Another important factor is the \textbf{embellishment} of descriptions. That is, in our particular marketing domain, is there a clear preference towards the embellished or plain style of descriptions. Here are two examples that showcase the different preferences from users: 

\begin{center}\begin{tcolorbox}[colback=white,colframe=gray!20,width=0.95\linewidth,breakable]
   {\footnotesize \textbf{Preferred Description (\agentname, Description A)}:  Welcome to your charming retreat at 7401 S Blackstone Ave, nestled in the vibrant Grand Crossing neighborhood of Chicago! Priced attractively at \$179,900, this delightful single-family home skillfully blends affordability and comfort. With three spacious bedrooms and two well-appointed baths, it perfectly accommodates family living. Highlights of this property include its exceptional flooring, ranking among the top 2\% in the area, and the effective recent improvements that make it standout in property condition. Inside, enjoy the luxury of new carpeting and stylish white shaker cabinetry, promising an inviting and modern ambiance. A versatile layout offers flexible options for home customization, meeting the varied needs of your lifestyle, from working to exercising at home. Step outside to a fenced-in yard, ideal for both privacy and play, further increasing its appeal. Though without a garage, the home features machine hookups, ample storage potential in a sub-basement, and a premium location providing easy access to all the amenities and family-friendly spots Chicago's Grand Crossing has to offer. Enhanced by durable brick construction, this home represents a significant investment opportunity, combining all the elements you desire in a new haven. Your perfect blend of convenience, comfort, and charm awaits! }
    \vspace{0.2em}

    {\footnotesize \textbf{Disfavored Description (Vanilla, Description B)}:  Welcome to 7401 S Blackstone Ave, an inviting single-family home nestled in Chicago's charming Grand Crossing neighborhood. Priced attractively at \$179,900, this 3-bedroom, 2-bathroom gem is perfect for those seeking comfort and convenience in an established community. Built in 1973, the residence boasts 933 square feet of living space and is beautifully updated with new carpeting throughout and stylish newer flooring. The kitchen shines with modern white shaker cabinetry, providing a fresh, contemporary feel. The home is well-equipped for practicality, featuring a dedicated laundry room with machine hookups for both gas and electric dryers. Adding to its appeal is a sub basement and a fenced-in yard, creating an ideal outdoor space for families or pet owners to enjoy. Conveniently located in Cook County, this home is serviced by Lake Michigan water and public sewer, and its brick construction ensures durability. With natural gas and forced air heating, you’ll be cozy year-round. This delightful abode represents a fantastic opportunity for homeownership without the burden of HOA fees. Don't miss your chance to make this delightful Chicago residence your own! }
    \vspace{0.2em}

    {\footnotesize \textbf{User Comment:} Description A is a bit more descriptive without going overboard, also talks about the neighborhood. }
    \vspace{0.2em}

   {\footnotesize \textbf{Case Analysis:} Some users love more \hl{descriptive} descriptions.  }
    \end{tcolorbox}\end{center}

\begin{center}\begin{tcolorbox}[colback=white,colframe=gray!20,width=0.95\linewidth]
    {\footnotesize \textbf{Disfavored Description (\agentname, Description A)}:  Nestled in the heart of Chicago's vibrant Bridgeport neighborhood, 3457 S Lituanica Ave offers unparalleled access and convenience, situated comfortably within Cook County. This spacious five-bedroom, two-bathroom single-family home is a standout choice, boasting top-tier features in location, accessibility, and outdoor living spaces. With its robust brick construction, this property provides a durable and inviting home environment, perfect for customization to suit your family’s evolving needs. Enjoy the luxury of a generous 6,500 sqft lot, among the best in its zipcode, offering a blank canvas for your dream garden or a secure playground for your child. The home’s interior shines with elegant hardwood flooring and practical features like in-unit laundry with sink. Practical comfort is ensured with space pac cooling and efficient natural gas heating, ensuring you feel at home year-round. Embrace Chicago living with easy access to nearby amenities, public transportation, and renowned neighborhood characteristics, all for an attractive price point of \$549,000—making it an excellent investment for future growth. }
    \vspace{0.2em}

   {\footnotesize \textbf{Preferred Description (Vanilla, Description B)}:  Welcome to your future home at 3457 S Lituanica Ave, nestled in the heart of Chicago's vibrant Bridgeport neighborhood. This charming single-family residence offers five spacious bedrooms and two full bathrooms, perfect for families seeking both comfort and style. Priced at an attractive \$549,000, this home sits on a generous 6,500 sqft lot, providing ample outdoor space for relaxation or entertaining. Crafted with enduring brick construction, the property boasts modern conveniences including a complete suite of appliances like a range, microwave, dishwasher, and more. The elegant hardwood flooring throughout adds a touch of sophistication, while the first-floor full bath caters to easy accessibility. Enjoy the convenience of in-unit laundry with a dedicated sink and stride out onto your private deck for a breath of fresh air. The two-car garage offers security and storage, supported by reliable utilities such as public sewer, natural gas heating, and Space Pac cooling. With easy access to Holden Elementary and local amenities, this home represents a delightful blend of classic charm and modern living in one of Cook County's most desirable neighborhoods. Don't miss the opportunity to make this house your home. }
    \vspace{0.2em}

    {\footnotesize \textbf{User Comment:} Description B does a better job at listing the amenities. }
    \vspace{0.2em}

   {\footnotesize \textbf{Case Analysis:} Some users love a \hl{plain style} of description that listing all amenities.  }
    \end{tcolorbox}\end{center}

These obervations suggest that there is no one-size-fits-all solution for writing style. Hence, future work could consider tailoring the description generation in the user's preferred writing style to further improve the persuasiveness.

\revise{
\subsection{The Diversity of Writing Styles On Different Listings}
\label{app: tailoring_description}
\agentname shows diverse writing styles linguistically on listings based on their different features, which means it can tailor different real estate listings well. 

In the following pair of examples, Low-end listings emphasize ``Safety \& Survival'' (security, enclosure, reassurance), whereas high-end listings emphasize ``Display \& Views'' (openness, visual richness, and mastery over the environment).

\textbf{Low-Price Representative (\$110{,}000).}

\begin{center}
\begin{tcolorbox}[colback=white,colframe=blue!20,width=0.95\linewidth]
{\footnotesize
\revise{%
\textbf{Focus:} \textbf{Defense, Enclosure, Reassurance}. Words aimed at eliminating buyer insecurity regarding the environment.\\[0.4em]
\textbf{\$110{,}000, 750.0 sqft, 2 beds, 1 bath}\\[0.4em]
Welcome to your charming oasis at [address], nestled in the vibrant and culturally rich Hyde Park neighborhood of Chicago. This inviting 2-bedroom, 1-bathroom condo offers the perfect blend of comfort and convenience at an unbeatable price of \$110{,}000. Step inside to discover a sun-drenched living space adorned with hardwood flooring and an updated kitchen featuring modern appliances, including a wine refrigerator. The thoughtful design includes first-floor conveniences like a full bath and ample storage, with walk-in closets providing plenty of room for your essentials. Enjoy \textbf{tranquil moments} on the large back deck, ideal for relaxation or entertaining guests, set within a \textbf{gated courtyard that ensures privacy and security}. The property is \textbf{meticulously maintained}, boasting brick construction and a welcoming community atmosphere. Although it is \textbf{compact}, the space is optimized for comfortable living \textbf{without unnecessary upkeep}, perfect for those valuing efficiency. With \textbf{proactive security measures}, a \textbf{strong sense of community}, and only minutes away from necessities, this condo perfectly encapsulates the ideal home for those \textbf{prioritizing safety} and cultural alignment in a vibrant neighborhood.
}
}
\end{tcolorbox}
\end{center}

\vspace{0.8em}

\textbf{High-Price Representative (\$1{,}875{,}000).}

\begin{center}
\begin{tcolorbox}[colback=white,colframe=blue!20,width=0.95\linewidth]
{\footnotesize
\revise{%
\textbf{Focus:} \textbf{Aggression, Openness, Visuals}. Words aimed at showing off transparency and mastery over the environment.\\[0.4em]
\textbf{\$1{,}875{,}000, N/A sqft, 4 beds, 4 baths}\\[0.4em]
Discover \textbf{unparalleled elegance} and style at [address], a single-family haven nestled in the vibrant Bucktown neighborhood of Chicago. This \textbf{exquisite home}, priced at \$1{,}875{,}000, offers four bedrooms and four bathrooms, perfect for families seeking \textbf{ample space and luxury}. Its standout features include a \textbf{private corner lot} and a \textbf{spacious side yard} designed for \textbf{ultimate outdoor enjoyment}, complemented by \textbf{gourmet enhancements} like a custom kitchen and a chic beverage center. New Pella windows and \textbf{cascading expanses of glass} invite an \textbf{abundance of natural light}, creating a bright and airy atmosphere across a versatile loft area ideal for work-from-home needs. With \textbf{sophisticated enhancements} such as vaulted ceilings, multiple fireplaces, and a gas fire table, this residence exudes comfort and warmth year-round. The \textbf{meticulously crafted design} places this property among the \textbf{top tier in architectural style} and elegance within the neighborhood and beyond. Enjoy \textbf{seamless access} to essential amenities and natural beauty, with a spacious parking capacity for four cars. \textbf{Embrace this rare opportunity} to own a piece of \textbf{refined luxury} in an urban yet serene setting.
}
}
\end{tcolorbox}
\end{center}
}

\section{The Design of Survey and User Interfaces}
\label{app: interface}

\subsection{Survey Screening Interface}
\label{app: screening-interface}

\begin{figure}[h!]
    \centering
    \includegraphics[width=0.45\linewidth]{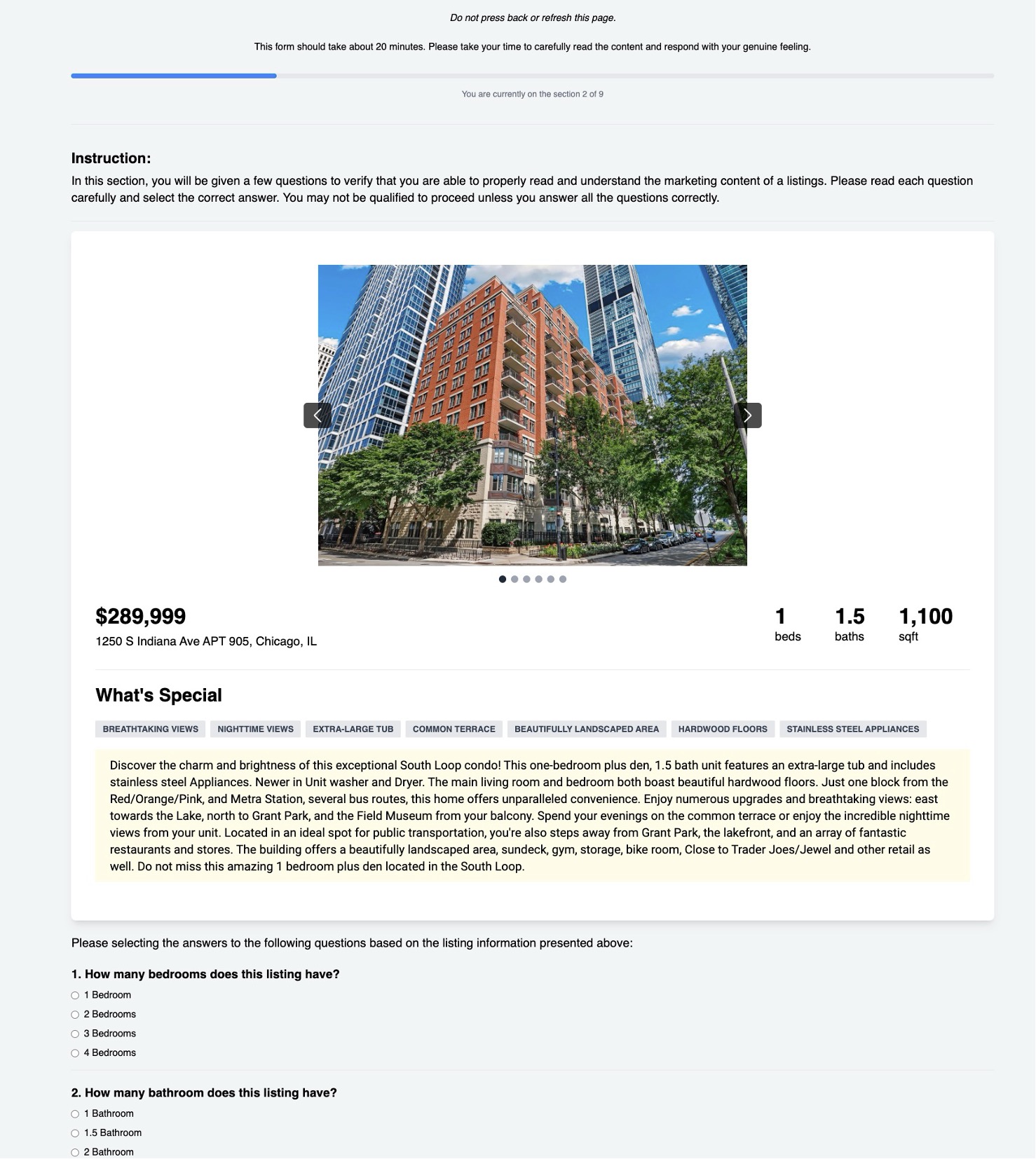}
    \includegraphics[width=0.45\linewidth]{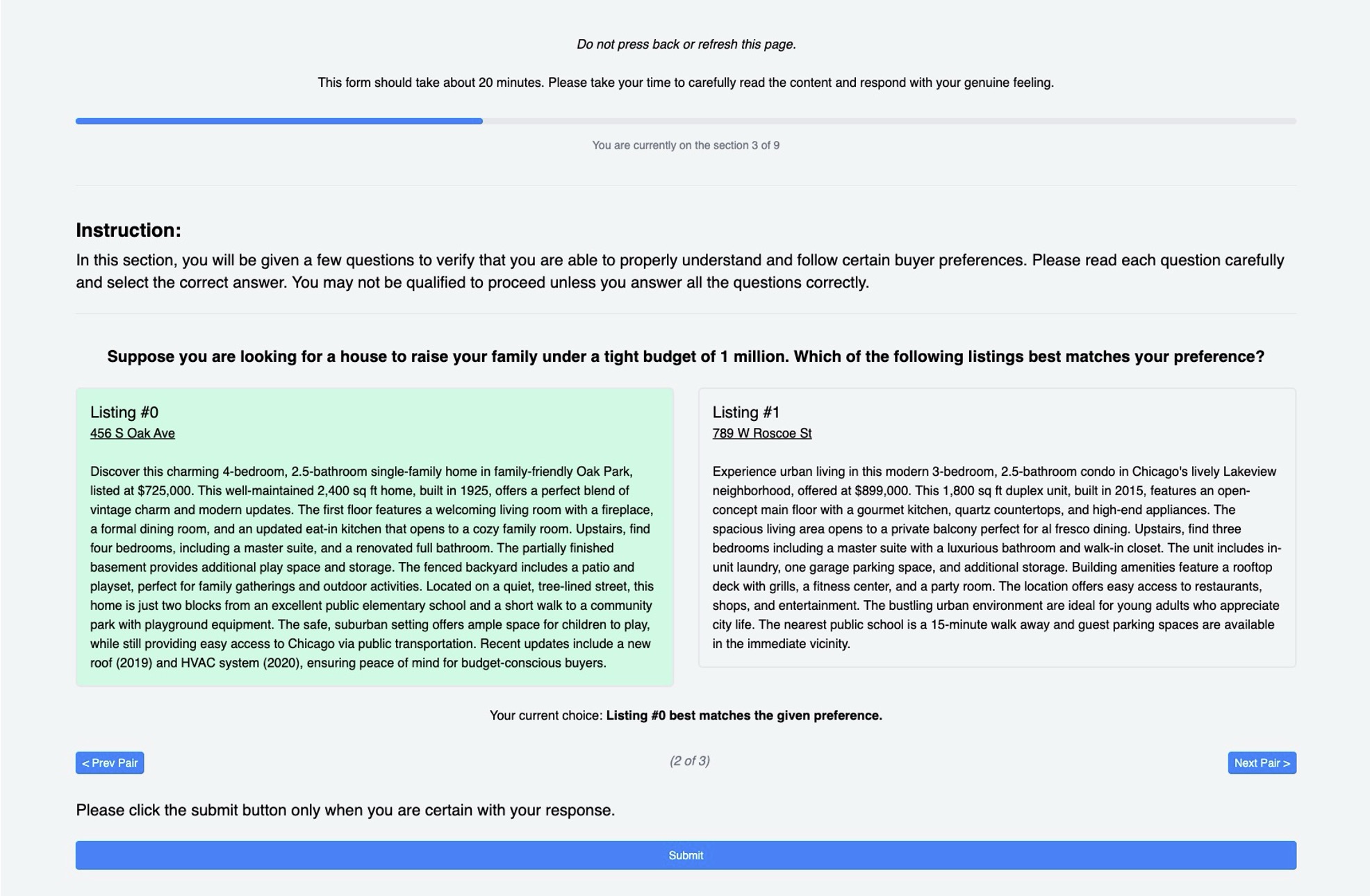}
    \caption{Survey Screening Interface}
    \label{fig:screening_interface}
\end{figure}

The first stage of the survey is designed to ensure the human subject has sufficient experience in the home search process in order to analyze the features from a marketing description. We present description of an example listing and design quiz-like questions to verify whether the participant is able to make all correct responses. We showcases the web user interfaces in \cref{fig:screening_interface}.

\subsection{Preference Elicitation Interface}
\label{app: preference-interface}
\begin{figure}[h!]
    \centering
    \includegraphics[width=0.45\linewidth]{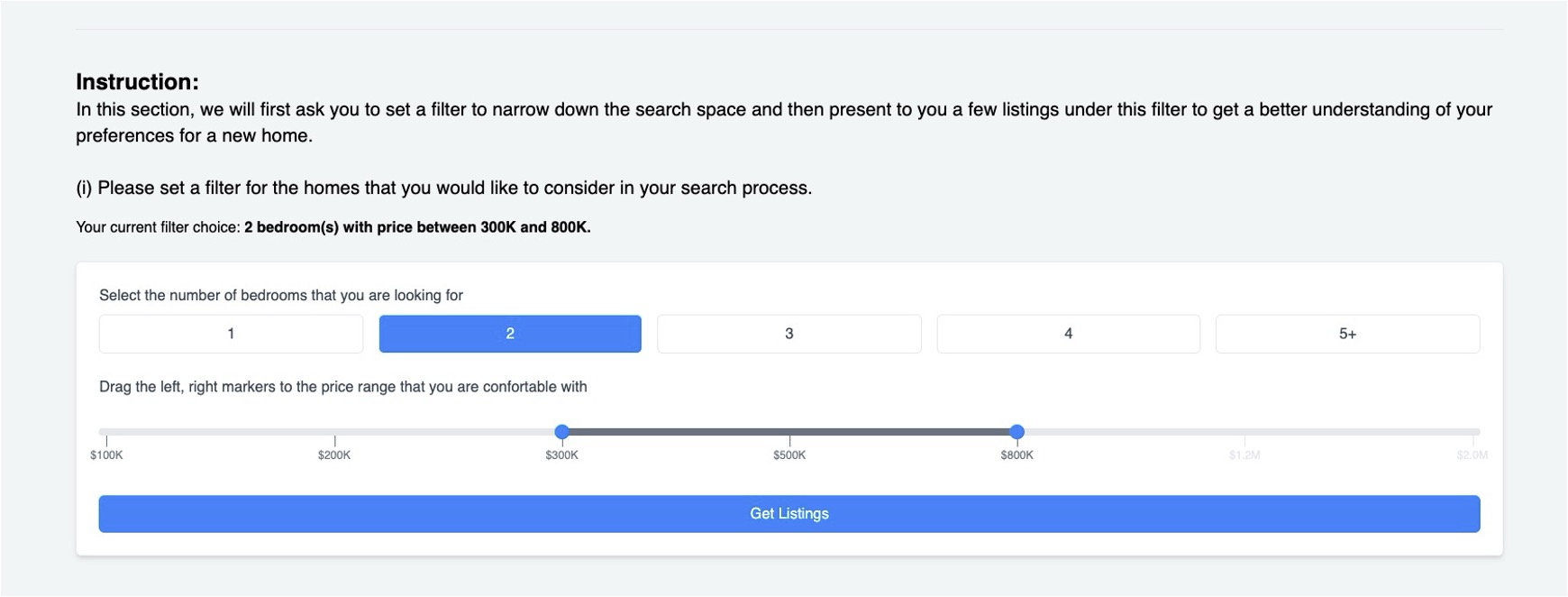}
    \includegraphics[width=0.45\linewidth]{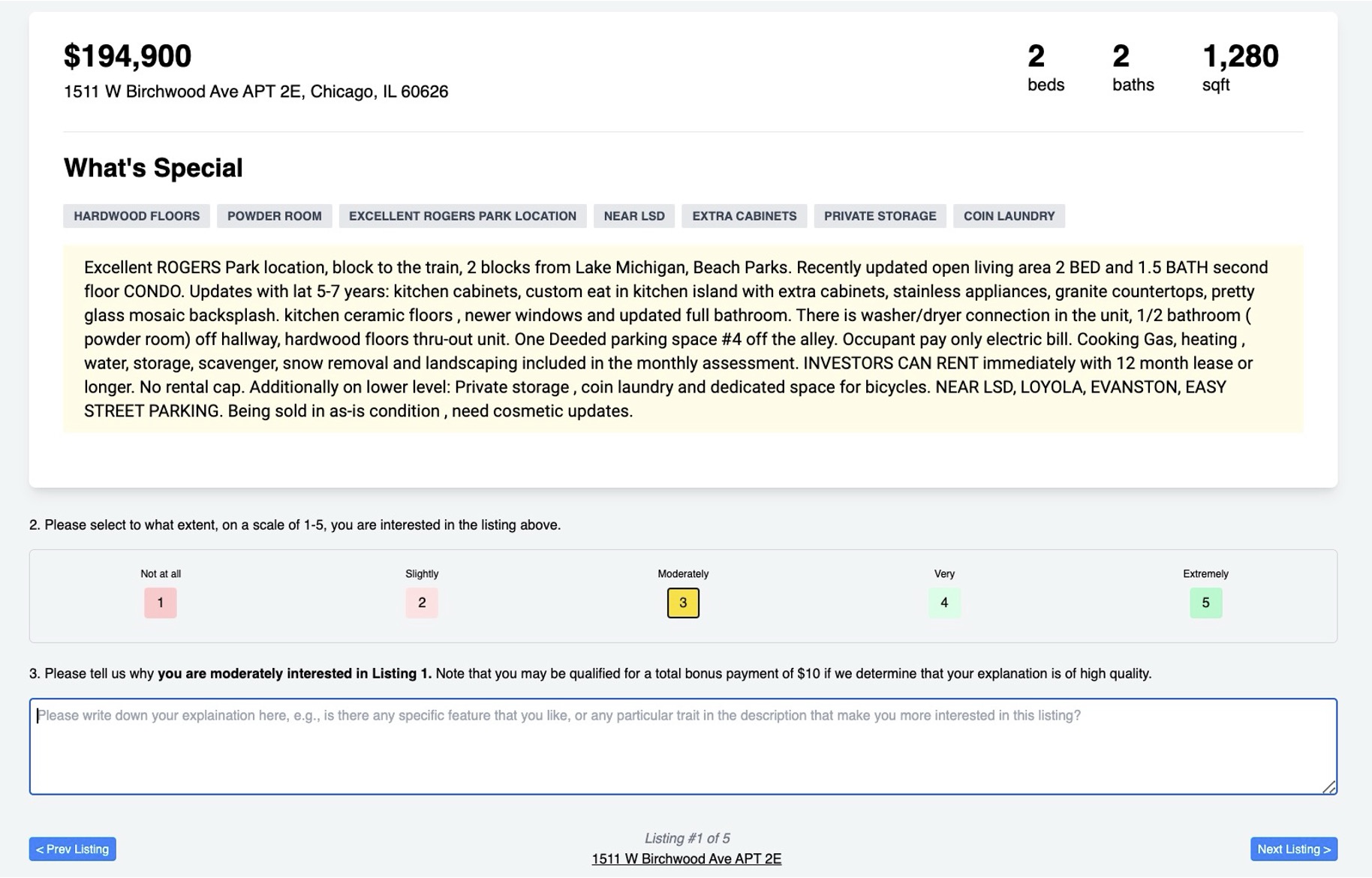}
    \includegraphics[width=0.45\linewidth]{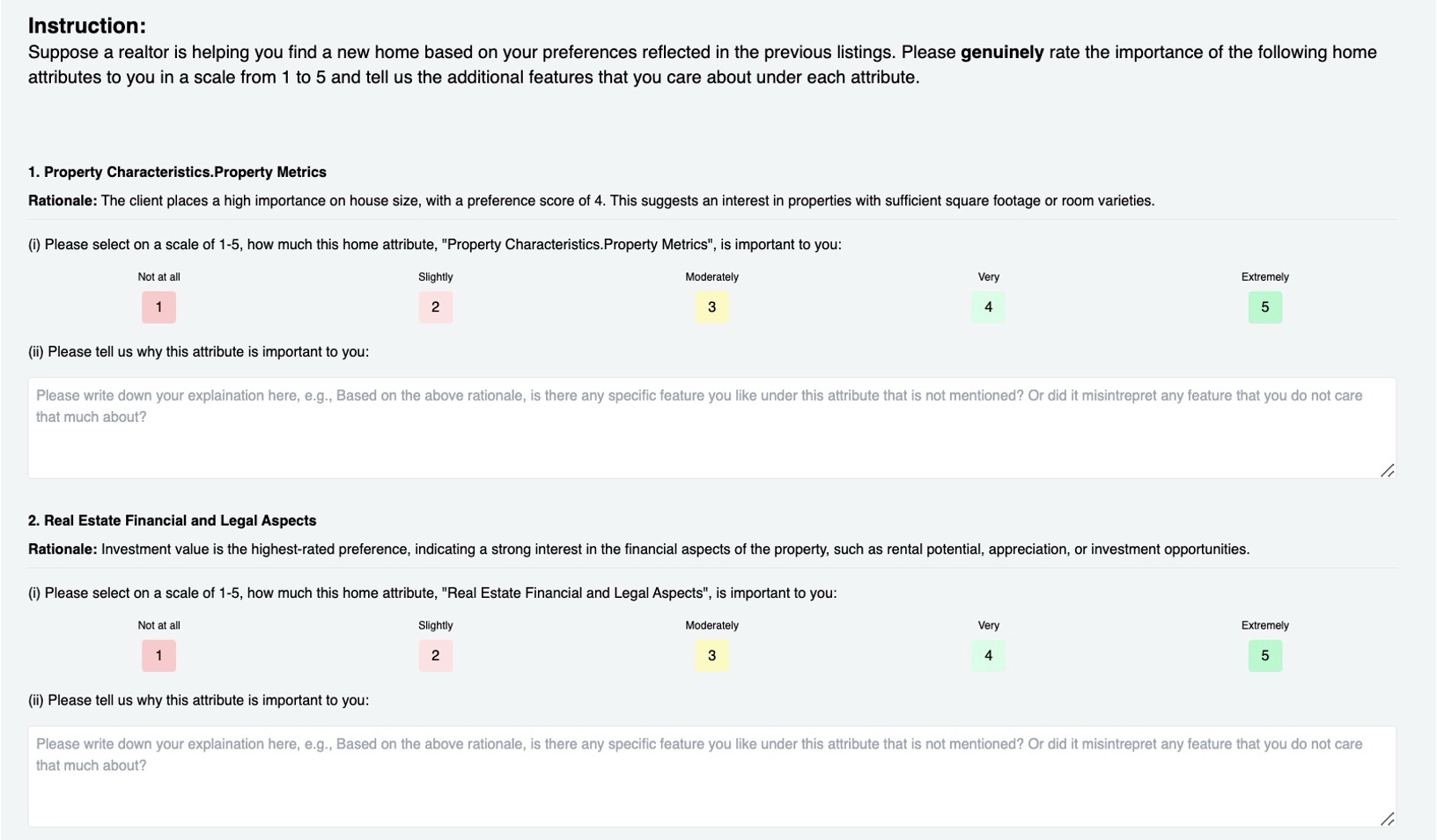}
    \caption{Preference Elicitation Interface}
    \label{fig:preference_elicitation_interface}
\end{figure}

In the second stage of the survey, we design an interface to mimic the environment of online platforms that the model can observe the buyer's general profile and behaviors (e.g., recently browsed or liked listing) to some degree. In our case of real estate listing, we ask the buyer to provide their preferences in a 1-5 scale on five general categories (price, location, home features \& amenities, house size, investment value) and set a filter on the price range and number of bedrooms in the house they are looking for. This information allows us to select generally relevant listings to mitigate the anchoring effect that the marketing content can play little role to influence the buyer in the evaluation phase. Next, we choose 5 relevant listings and ask the buyer to rate them on a 1-5 scale and provide their reasoning. This process ensures that we can collect a reasonable amount of each buyer's preference information for the personalized persuasive content generation in the evaluation phase. Finally, we employ LLM to narrow the features that are likely preferred by the participants and ask for their ratings of importance on a 1-5 scale.  We showcases the web user interfaces in \cref{fig:preference_elicitation_interface}.

\subsection{Human Evaluation Interface}
\label{app: comparison-interface}
\begin{figure}[h!]
    \centering
    \includegraphics[width=0.6\linewidth]{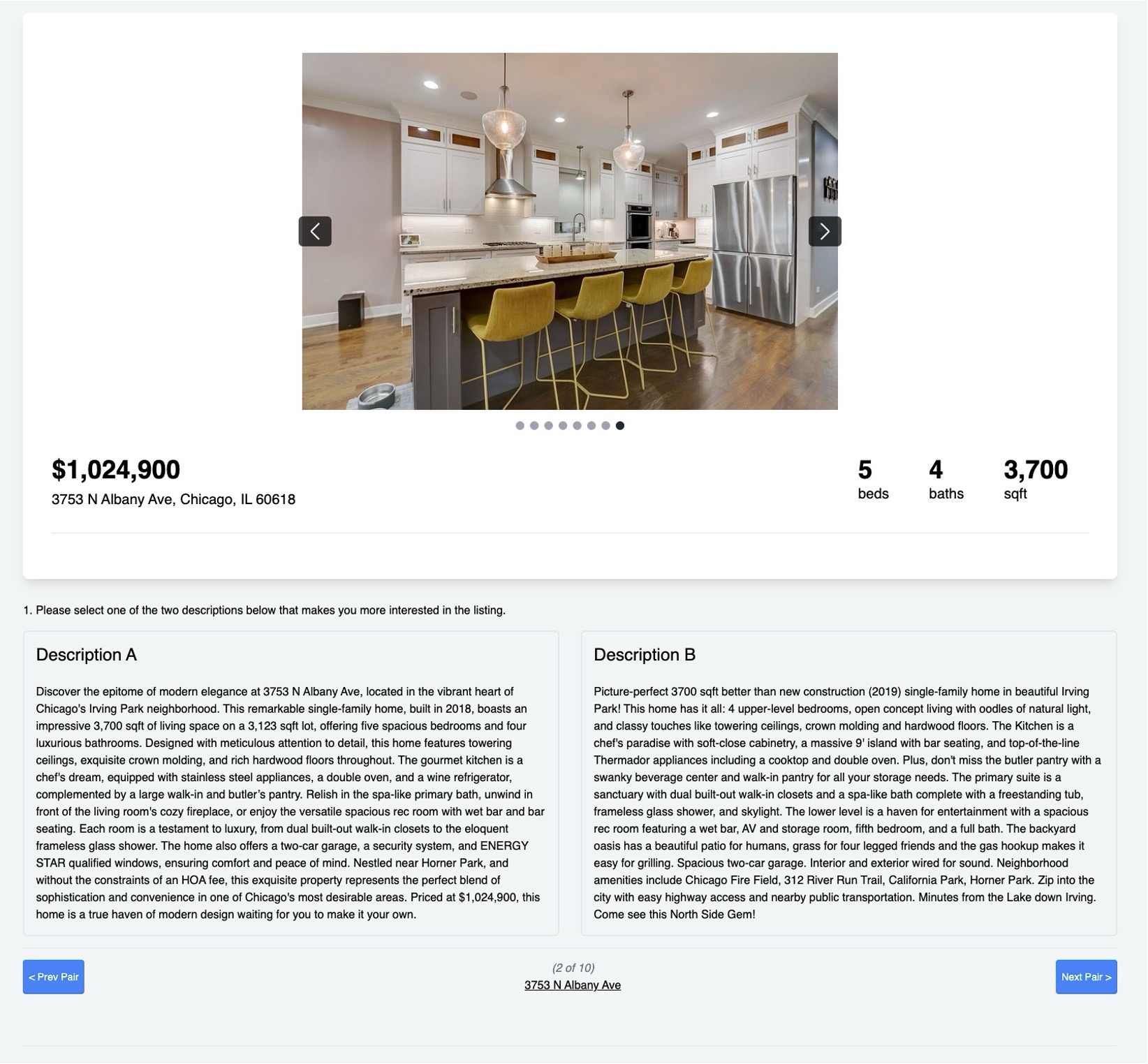}
    \caption{Human Evaluation Interface}
    \label{fig:comparison_interface}
\end{figure}

In the last stage of the survey, it is to gather the human feedback on the persuasiveness of different models. 
Many previous works study persuasion by asking human how much does their opinion changes before and after reading an argument. In our task, human subjects often do not have any prior knowledge about item and this evaluation procedure would induce bias.
Instead, we implement two alternative evaluation schemes in our interface: one is the A/B test where the buyer is presented with a single listing along with two descriptions generated by two distinct models and then asked to report which description makes them more interested in the listing; the other is the interleaved test where a set of listings each with a single description generated by some model and the buyer is asked to select the listings that they are interested in based on their descriptions. Each time after a participant's choice of the preferred description, we ask participant to rate on a scale of 1-5 that one description is prefer over another and incentivized them to provide a detailed rationale of their responses. To illustrate this process, we present the web interface design in \cref{fig:comparison_interface}. 

\subsection{Feature Annotation Interface}
\label{app: feature_highlight_annotation_interface}
\begin{figure}[h!]
    \centering
    \includegraphics[width=0.6\linewidth]{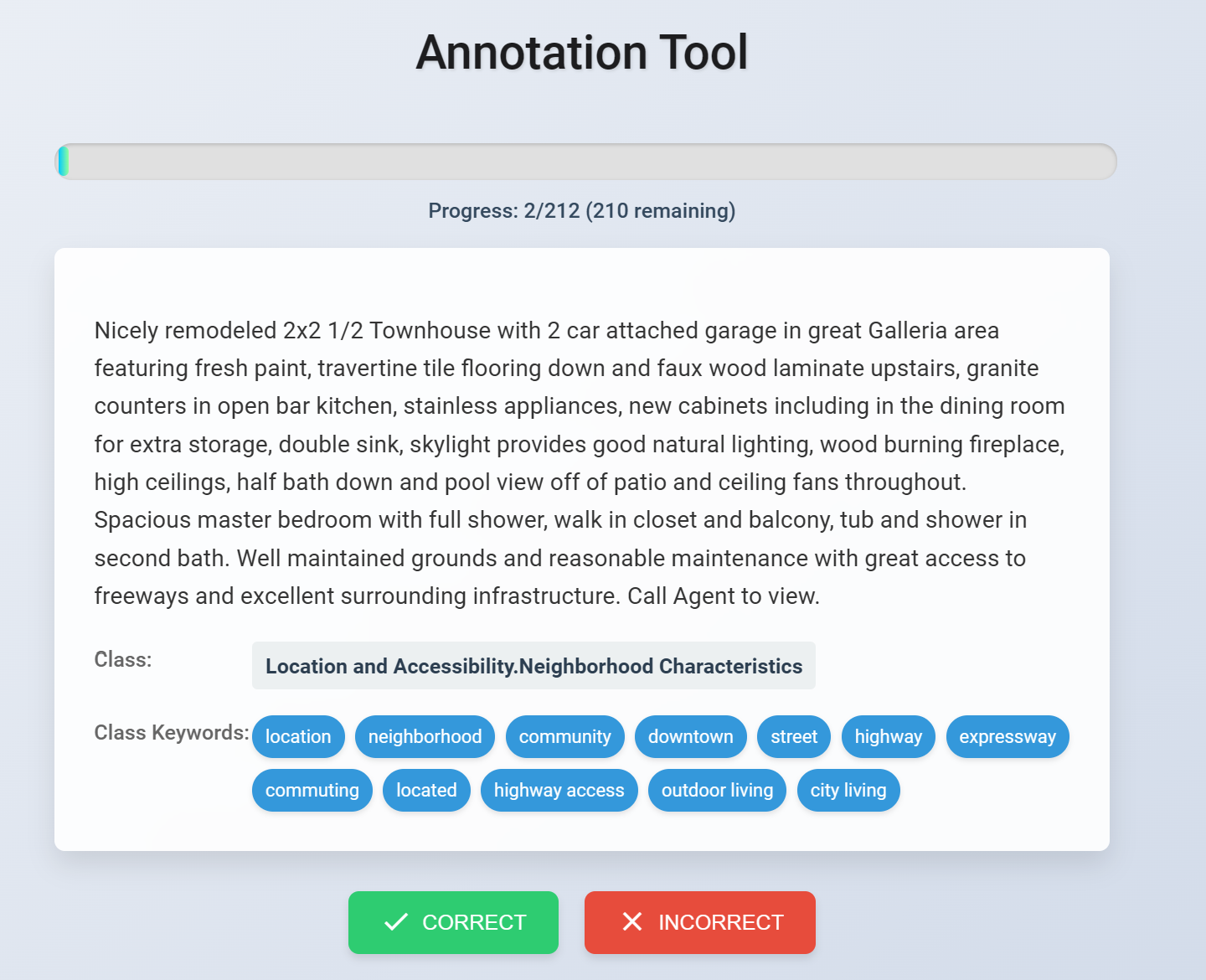}
    \caption{Annotation Interface}
    \label{fig:feature_highlight_annotation_interface}
\end{figure}

To ease the task of feature annotation, we also develop a user-friendly web interface. Its design is shown in  \cref{fig:feature_highlight_annotation_interface}.

\section{Additional Experimental Details}
\label{app:additional_experiments}

\subsection{GPT-4o-Polished Human Control}
\label{app:polished_human_control}
To test whether \agentname's gains are explained merely by improved fluency or generic LLM polishing, we conducted an additional blinded human evaluation against a GPT-4o-polished human baseline. In this control, the original human-written listing descriptions were rewritten by GPT-4o for clarity and style while preserving the original human content source. This baseline does not incorporate buyer preference signals, and therefore should not be interpreted as the full preference-conditioned rewrite baseline suggested by reviewers. Instead, it isolates the effect of generic LLM polishing.

Across two completed randomized sessions, we collected 336 pairwise comparisons. \agentname achieved 254 wins, 51 losses, and 31 ties against the GPT-4o-polished human baseline. Excluding ties, this corresponds to an 83.3\% win rate, with a 95\% Wilson confidence interval of [78.7\%, 87.0\%]. Under our Elo scoring procedure, \agentname obtained an Elo score of 1168, compared with 832 for the polished-human control.

\subsection{Elo Uncertainty and Convergence}
\label{app:elo_uncertainty}
For the main human-evaluation results, we estimate Elo uncertainty by bootstrap resampling pairwise comparisons with replacement and recomputing Elo ratings for each bootstrap sample. We also use the bootstrap distribution of Elo differences for pairwise comparisons between ablation variants. We interpret these results conservatively: the main conclusion is that the full pipeline is strongest overall and is clearly separated from vanilla LLM and human baselines, rather than that every adjacent module addition yields a statistically significant gain.

As a convergence diagnostic, we also recompute Elo ratings using 25\%, 50\%, 75\%, and 100\% random subsamples of the vote pool. The relative ranking of major systems is stable across these subsamples. By the 75\% subsample, absolute Elo estimates are within approximately $\pm 30$ points of the final estimates.

\subsection{Grounding Module Evaluation}
\label{app:grounding_module_eval}
The grounding module is evaluated as a multi-label binary classification problem, where each feature is predicted independently. To avoid ambiguity around a single ``majority class,'' we report a per-label majority baseline aggregated over labels.

\begin{table}[h]
\centering
\small
\caption{Feature extraction and grounding model performance.}
\label{tab:grounding_eval_extra}
\begin{tabular}{lcc}
\toprule
Method & Accuracy & F1 \\
\midrule
Per-label majority baseline & 54.2\% & -- \\
Direct LLM prompting & -- & $\sim$59\% \\
Embedding pooling baseline & -- & $\sim$59\% \\
Final grounding model & 69.39\% & 67.43\% \\
\bottomrule
\end{tabular}
\end{table}

Performance is strongest for more concrete categories such as structural, layout, and room-configuration features, where representative accuracy is around 78\%. More subjective categories, such as ambiance and lifestyle appeal, are harder and achieve representative accuracy around 61\%.

\subsection{Annotation Reliability}
\label{app:annotation_reliability_extra}
For human factuality checks, three graduate-student annotators with NLP evaluation experience annotated hard and soft factual attributes. Inter-annotator agreement is substantial for hard attributes ($\kappa=0.74$) and moderate for soft attributes ($\kappa=0.58$), consistent with the greater subjectivity of soft factual matching. We therefore emphasize ranking consistency between human and LLM factuality evaluations, rather than exact agreement in absolute scores.

For feature schema validation, three annotators, including two co-authors with real-estate marketing domain knowledge and one independent NLP annotator, reviewed category assignments and achieved $\kappa=0.81$. For grounding-label validation, GPT-4o-generated feature labels were checked by three human annotators on a stratified sample of 200 listings, with feature-category assignment agreement of $\kappa=0.74$.

\subsection{Hyperparameter Selection}
\label{app:hyperparameter_selection_extra}
The feature existence threshold $\alpha=0.5$ was selected by grid search over $[0.1,\ldots,0.9]$ using F1 on a held-out human-annotated validation set. For personalization, we use $c=0.01$ and $r_0=2$, and pass the top 10 highest-scoring features to the generation prompt. For surprisal-based marketing features, we use $\beta=30\%$. We did not conduct exhaustive Elo sensitivity analysis over these hyperparameters because each setting would require another human-subject evaluation.

\section{Implementation Details}
\label{app: implementation-details}
In this section, we provide a full description of the implementation detail of \agentname.

\subsection{Grounding Module: Predicting Marketable Features}
\label{sec: highlight_model-app}
Our model assumes the existence of attribute-feature mappings in different marketing problems, with which a seller can use to influence the buyer's beliefs and behaviors. However, a key challenge lies in determining how to accurately obtain such mappings. Specifically, we must identify which \emph{signaling features} to include and under what conditions it is natural to market a product as possessing a particular feature. Traditionally, acquiring this knowledge from human experts is both labor-intensive and costly. 
Instead, we take a learning approach to uncover the mapping from our experiment dataset. 
While the raw dataset contains no annotation of any signaling feature, we employ LLMs to construct a high-quality feature schema and label the dataset accordingly in preparation for learning the attribute-feature mapping. 
This approach notably presents a novel unsupervised learning paradigm, harnessing the broad knowledge of LLMs to distill expert-level insights from unlabeled data with minimal human supervision.

\textbf{Inductive Construction of Feature Schema} 
Our dataset only contains the raw attributes of each product. In order to learn a high-quality attribute-feature mapping, the first task is to obtain a good representation of feature schema $S$. 
On the one hand, if we miss some useful signaling features, it could significantly hinder the performance of subsequent marketing task. On the other hand, there are so many possible token that can serve as the signaling features in the natural language space, and many of these tokens might have duplicate or similar meaning. If there is no structured representation of the features, the resulting label classes could be too sparse to learn.
Indeed, we discover that the feature schema obtained by directly prompting an LLM includes many similar features while miss some important ones.
Based on this observation, we turn to a more sophisticated prompting strategy to inductively improve the quality and representation of the feature schema (see a high-level sketch of the construction pipeline in \cref{fig:highlight_model_pipeline}).

First, we construct a basis of feature schema, represented as a list of tokens used in the human-written marketing description to describe some house features. We begin with \textit{Mixtral-8x7B-Instruct-v0.1}~\citep{jiang2024mixtral} to extract keywords or phrases $\{k_1, k_2, \dots\} = \llm([\mathcal{I}_{\text{Keyword}}; D_{\text{human}}])$ that summarize each human-written description $D_{\text{human}}$ under a keyword-extraction prompt $\mathcal{I}_{\text{Keyword}}$ (\cref{app: keyword_extraction_prompt}). We observed that, in some cases, the model output could not be directly parsed into a clean list of keywords, or it contained excessive quantifiers and modifiers. To address this, we re-prompted the model using $\mathcal{I}_{\text{Norm}}$ (\cref{app: keyword_normalization_prompt}) to normalize each keyword. Through this process, we initially extracted $112688$ keywords—too many to handle effectively. We then applied additional normalization steps, including lowercasing, lemmatization, and synset merging via NLTK~\citep{bird2009natural}. We also filtered the keywords, retaining only those that appeared in at least 50 descriptions. This reduced the final set to $1114$ keywords as our \emph{induction base}.

Next, we organize the feature-related keywords into a structured feature schema. Since many keywords are related to each others and hard to distinguish, we use a hierarchical representation of feature schema to better capture the relations between different feature classes and to ease the subsequent labeling task.
To achieve this goal, we prompted \textit{Claude-3.5-Sonnet}~\citep{claude3.5sonnet} with a 100-keyword batch to iteratively generate a  hierarchical schema that covers the majority of the keywords (an example run can be found in \cref{app: schema_induction}). We temporarily switched to \textit{Claude-3.5-Sonnet} because we found it particularly difficult for open-source models, even the state-of-the-art \textit{GPT-4o}~\citep{gpt4o}, to induce such a schema without grouping most keywords into overly broad categories like "others" or "misc", resulting in a shallow and uninformative schema. In contrast, when fed keywords in small batches, \textit{Claude-3.5-Sonnet} followed our instructions more faithfully, organizing the keywords into a carefully structured hierarchy. Every leaf node in the schema was associated with a set of relevant keywords. From this process, we obtain a relatively well-structured and comprehensive feature schema.

Finally, to evaluate the quality of the generated feature schema, monitor potential hallucination issues, and further refine the schema, we asked three human participants to conduct manual review.
We prompt 
\textit{Mixtral-8x7B-Instruct-v0.1} 
to determine whether a feature from the schema presents in each human-written description, and each participant is asked to independently verify this result (see our annotation interface in \cref{app: feature_highlight_annotation_interface}). Based on the participants' feedback on 636 samples, we found that features labeled by LLMs are mostly agreed across all human annotators, except for some ambiguous or subjective features (e.g., the aesthetic features of a house), where the agreement rates (around $60\%$) between models and human are about as good as that among human annotators.
We refine the schema for two more iterations, where we prompt LLMs to merge some similar features and reduce the ambiguity of some features with more precise example keywords.  We list our final feature schema in~\cref{app: final_feature_schema} and it is used in the subsequent stages of our pipeline.

\paragraph{Learning the Feature-Attribute Mapping}
With the feature schema, we guide the LLM to annotate for each product with attributes $\mathbf{x}$ whether each feature ${s}_i$ is described in the human-written marketing text (see the prompt in \cref{app: feature_extraction_based_on_description_prompt}).  We perform a few additional pre-processing steps to this correspondence data to supervise the learning of the feature-attribute mapping.

First, we found that some human-written marketing descriptions are of relatively low quality and these data points can negatively impact the learnt feature-attribute mapping. Hence, we only select marketing descriptions of products that are relatively popular, according to a simple heuristic ratio between the number of likes and views received by a listing recorded on the marketing platform. We expect the quality of feature-attribute mapping uncovered from this filtered set of human-written descriptions would be higher than average.

Next, we normalize the attributes of each listing $\mathbf{x}$ and embed existing knowledge of these attributes into their representation. 
Since the raw attributes of each listing $\mathbf{x}$ have different value types (categorical, integer, float, etc.), we convert each attribute $x_i$ into a natural language statement using the template, ``The attribute \textit{attribute\_name} is \textit{attribute\_value}.'', 
and then use an embedding model, \textit{SFR-Embedding-Mistral}~\citep{meng2024sfrembedding}, to convert each natural language statement into a fixed-dimensional vector $e_i = \llmembed\left(x_i\right) \in \mathcal{R}^d$. We also perform some standardized normalization techniques such as removing irrelevant attributes and dropping attributes with missing values.
Finally, we use a simple multi-layer perceptron (MLP) to learn the attribute-feature mapping as,
\begin{equation*}
    \pi\left(s_i \mid \mathbf{x}\right) = \sigma(O_i^T \text{ReLU}(W \bar{e}(\mathbf{x}))),
\end{equation*}
where $\bar{e}(X)$ is the mean-pooled attribute embedding, and $O_i \in \mathcal{R}^{d/2}, W \in \mathcal{R}^{d \times d/2}$ are the model's weights. The function $\sigma$ represents the sigmoid activation function. Here, we assume conditional independence between highlights given the raw features $X$.
We use the standard logistic loss function to training the neural network. 
We apply a random train-test split of $4:1$ ratio in our dataset and achieve testing accuracy $69.39\%$ and F1 score $67.43\%$. We find the accuracy to be reasonably high, given the stochastic nature of signaling process. That is, the features deterministically predicted based on our mapping cannot exactly match with the features used in the human written description with some degree of randomness --- just as the accuracy of predicting a fair coin toss is at most $50\%$. 

The typical implementation of a signaling scheme is to follow the attribute-feature mapping $\pi$ to randomly draw a signal $S_j$ with probability $s_j(\mathbf{x})$. This is necessary in theory to maintain the partial information carried by each signal. 
However, we implement a deterministic feature selection strategy to only use feature $S_j$ with probability above some threshold $\alpha$. This is because our generated marketing content only accounts for a tiny portion of the corpus so that it should have almost no influence on people's perception of a feature (e.g., the partial knowledge inferred upon observing each feature).
This also ensures that the product would have the feature with high probability, as our objective prioritizes the rigorousness of our marketing content. 
As a simple heuristics in our implementation, we set the threshold $\alpha=1/2$ and we will refer to this set of features as, 
\begin{equation}\label{eq:marketable-feature-app}
     \text{Marketable Features: } \quad  \mathcal{S}_1(\mathbf{x}) = \{S_j:  s_j(\mathbf{x}) \geq \alpha \}.
 \end{equation} 

\subsection{Personalization Module: Aligning with Preferences}
\label{sec: user_preference-app}

This stage seeks to steer persuasive language generation toward the buyer's preference, which is another crucial objective of grounded persuasion. In particular, LLMs make it practical to incorporate lightweight user preference signals into copywriting at lower cost than conventional marketing designed for a broad population. Our solution has two parts: the first part elicits useful information about a user's preference and structures it in a usable representation; the second part selects a subset of features based on the user preference to guide what the model emphasizes. 

\paragraph{Structured Preference Representation}
As mentioned previously, our evaluation environment is built to have an information elicitation process from each buyer. However, such information cannot directly describe the user's preference. 
So, we ask the LLM to act like a human realtor to determine the features that the users might be interested in based on their initial selection.
To do this, we prompt the language model to convert the user preference into information structured according to the feature schema. 
We then ask the user to give a rating $r_j$ on a scale of 1-5 on how important each feature $S_j$ is. We also elicit the user's rationale behind this rating to nudge users to give more thoughts on their selection and thereby improve the credibility of their rating responses. While our implementation mostly relies on user surveys and the information processing power of LLMs, this design is a reasonable simulation of digital marketing in real-world applications, where $r_j$ can be learned through the standard industrial techniques of cookie analysis.

\paragraph{Personalized Feature Selection}
While the marketable features in \cref{eq:marketable-feature-app} are predicted at a population level, it is also useful to select features that are tailored to the user's special interests. 
However, because real-world marketing descriptions are not optimized for individual users, we cannot simply rely on a data-driven machine learning approach for personalization.
Instead, we use LLMs to interpret the elicited preference information. In our implementation, we select a set of features that are marketable and preferred by the buyer and let the LLM decide which personalized features to emphasize in the marketing content. 
Our heuristic method for personalized feature selection is to adjust the population-level feature scores $\mathbf{s}(\mathbf{x})$ with the user's rating over each feature $\mathbf{r}$ as follows,
\begin{equation}
     \text{Personalized Features: }  \mathcal{S}_2(\mathbf{x}) = \{ s_j | s_j(\mathbf{x}) +   c (r_j - r_0) \geq \alpha  \},  
 \end{equation} 
where the constant $c$ reflects the intensity of personal preference, $r_0$ is the basis rating of each attribute. In our human-subject experiment, we choose $c=0.01$, $r_0=2$ and set the threshold value $\alpha$ such as to select features of the top 10 highest scores. We list these features in the prompt to generate persuasive marketing description (see a full specification in \cref{app: highlight-preference-prompt}).

\subsection{Marketing Module: Capturing Surprisal via RAG}
\label{sec: surprisal-app}

The last stage is designed to better ground the persuasive language generation on factual evidences, problem contexts and localized information in automated marketing. There are many ways to improve the grounding for different settings of automated marketing. 
As a case study, we choose to focus on the surprising effect, a common marketing strategy studied by many work~\citep{lindgreen2005viral, ludden2008surprise, ely2015suspense}, under which the buyers would derive entertainment utility and have a deeper impression.
In our setting of real estate marketing, we consider the type of features that are relatively rare in its surrounding area.
That is, we say a marketable feature $S_j$ is \emph{surprising} if it is among the top $\beta$-quantile of the distribution of $S_j$ values under the prior distribution  $ s_j(\mu)$, or formally, 
 \begin{align}
 \mathcal{S}_3(\mathbf{x})
 &=
 \{S_j \subset \mathcal{S}_1 \mid s_j(\mathbf{x}) \in Q_{\beta}(s_j(\mu))\}.
 \end{align}
where $Q_{\beta}(s_j(\mu))$ denotes the top $\beta$-quantile region of distribution $s_j(\mu)$.
In our implementation, we determine a set of features for each listing that have its comparative advantage among different groups of similar listings. We consider two kinds of retrieval criteria: (1) select all listings within the proximal location at different levels of granularity (e.g., neighbourhood, zipcode or city); (2) select the 10 listings with the most similar features via an information retrieval system (implemented by the ElasticSearch framework\footnote{\url{https://www.elastic.co/elasticsearch}}) --- the search engine implementation details can be found in \cref{app: search_engine}. 
For each group of similar listings, we determine an empirical distribution function on each attribute score $\tilde{F}_i$. We then set $1 - \tilde{F}_i( p_i )$ as the percentile ranking of the listing's attribute $i$ among this group. 
We then select all attributes that are among the top $30\%$ percentile ranking for some group and provide the information in the prompt to generate persuasive marketing language (see a full specification in \cref{app: surprisal-prompt}). 
This gives the LLMs localized feature information at different granularity level. 

\section{Data Curation}
\label{app: dataset}

\subsection{Dataset raw attribute schema}
To ensure both quality and fidelity of our evaluation, we collect the real data of real estate listings on the market. The dataset for this experiment was sourced primarily from Zillow and includes around 50000 listings collected in the month of April in 2024. We follow the Zillow terms of services\footnote{\url{https://www.zillow.com/z/corp/terms/}} to avoid any commercial use of their data. Each of these listings is from one of the top 30 most populous cities in the United States as described by the U.S. Census Bureau. Listings that were not residential in nature or were missing crucial data to this experiment were excluded from this dataset. This dataset is composed of 95 columns, with features ranging from number of bedrooms, price, views, and more (see \cref{tab:property-table}). These many features associated with each listing provide us sufficient space to develop and test improved models for grounded persuasion.

\begin{table}[h]

  \centering
  \begin{tabular}{ll}
    \toprule
    \textbf{Field Name}       & \textbf{Data Type}     \\
    \midrule
    bedrooms                  & float64                \\
    bathrooms                 & float64                \\
    price                     & float64                \\
    description               & object                 \\
    living\_area\_value       & float64                \\
    lot\_area\_value          & float64                \\
    area\_units               & object                 \\
    brokerage\_name           & object                 \\
    zipcode                   & object                 \\
    street\_address           & object                 \\
    home\_type                & object                 \\
    time\_on\_zillow          & object                 \\
    page\_view\_count         & float64                \\
    favorite\_count           & float64                \\
    home\_insights            & object                 \\
    neighborhood\_region      & object                 \\
    scraped\_at               & object                 \\
    url                       & object                 \\
    city                      & object                 \\
    state                     & object                 \\
    year\_built               & float64                \\
    county                    & object                 \\
    avg\_school\_rating       & float64                \\
    id                        & object                 \\
    time\_on\_zillow\_days    & float64                \\
    score                     & float64                \\
    jpeg\_urls                & array                  \\
    \bottomrule
  \end{tabular}
   \caption{Listing data, subset of important columns}
  \label{tab:property-table}
\end{table}
\subsection{Final Feature Schema}
\label{app: final_feature_schema}
Here is the condensed version of the final feature schema to save pages: 
\\
\\
\begin{lstlisting}
Interior Features:
    Rooms:
        [bath,bathroom,bedroom,kitchen,living room,secondary bedrooms,patio,backyard,closet,room,living,dining room,pantry,space,office,laundry room,dining,living space,living area,primary suite,master suite,family room,cellar,foyer,game room,great room,den,master bedroom,utility room,sunroom,bedroom suite,living areas,primary bedroom,office space,kitchenette,owner's suite,playroom,storage room,living rooms,ensuite,wet bar,loft area,sitting room,mud room,exercise room,clothes closets,walk-in closet,mudroom,conference room]
    Flooring:
        [flooring,stories,carpeting,hardwood floors,tile,tile floors,hardwood flooring,wood flooring,hardwood floors]
    Furniture:
        [desk,table,chair,bed,dressers,cupboards,sofa,bench,seating]
    Additional Spaces and Versatility:
        [bonus room,flex space,flex room,den]
    Kitchen Features:
        [countertop,granite countertops,marble countertops,island,cabinetry,kitchen island,kitchen cabinets,waterfall,dining space,cooktop]
    Architectural Elements:
        [roof,window,floor plan,cabinet,molding,staircase,brick,paneling,siding,beam,ceiling fans,stair,chandelier,finishing trim,baseboard,trim]
    Bathroom Features:
        [shower,vanity,powder room,jacuzzi,ensuite,half bath,water closet,mirror,faucet]
    Storage:
        [storage,closet space,cabinet space,shelving,storage space,mudroom,drawer,bookshelf,storage unit,clothes storage,bike storage]
    Comfort and Ambiance:
        Lighting:
            [lighting,natural light,light fixtures,skylight,lighting fixtures]
        Temperature Control:
            [fireplace,hvac,fan,ac,a/c,central air conditioning]
Exterior Features:
    Outdoor Spaces:
        [patio,backyard,yard,pool,spa,balcony,porch,deck,roof deck,outdoor space,rv parking,outdoor spaces,outdoor living space,fenced yard,pavers,garden,outdoor living,backyard oasis,pergola,gazebo,cabana,landscaping,shade,lawn,fountain,sod,outdoor bench]
    Outdoor Activities:
        [gardening,outdoor cooking,barbecue,bbq]
Location and Accessibility:
    Neighborhood Characteristics:
        [location,neighborhood,community,downtown,street,highway,expressway,commuting,located,highway access,outdoor living,city living]
    Nearby Amenities:
        [shopping,restaurant,park,school,grocery,cafe,hospital,food,stadium,museum,boutique,shopping centers,station,elementary,bus,trader joe's,golf,brewery,elementary school,school district,recreation facility]
    Cities/Regions:
        [Austin,Denver,Charlotte,Houston,Dallas,San Antonio,Nashville,Phoenix,Los Angeles,LA,Manhattan,Detroit,Philadelphia,Portland]
    Access and Transportation:
        [access to amenities,proximity to schools,proximity to restaurants,proximity to shops,access to shopping,bus stop,walking distance,proximity to shopping,freeway access,public transit nearby,public transportation,road]
    Walkability and Bikeability:
        [walkability,bike score,walk score]
Housing Types:
    [studio,cottage,ranch,duplex,townhome,brownstone,row home,bungalow]
Building Features:
    Structure:
        [condo,loft,unit,townhouse,estate,square feet,duplex,garage,carport,story,penthouse,sf,triplex,colonial]
    Parking:
        [garage,parking,parking space,parking spaces,garage door,parking spot]
Appliances:
    [appliance,refrigerator,dishwasher,washer/dryer,range,fridge,microwave,washer,ac unit,dryer,hood,laundry facilities,washer and dryer,oven,garbage disposal,wolf appliances,thermador appliances]
Amenities:
    [community center,community pool,spa,firepit,fire pit,outbuilding,tennis courts,club house,rooftop,rooftop deck,rooftop terrace,dog park,lounge,elevator,recreation room,gym,fitness center,clubhouse,swimming pool,pool,spa,sauna,hot tub,putting green,tennis courts,basketball,pickleball,tennis court,golf,management,booking,concierge,trash,maintenance,doorman,superintendent,nightlife,brewery]
Utilities and Systems:
    [plumbing,water heater,heater,hot water heater,water,water filtration system,gas,sprinkler system,hvac,ac,a/c,wiring,solar panels,solar,electrical panel,electricity,generator,security,security system,camera,internet,wifi,cable,phone,satellite,fiber,internet access,satellite TV,internet service,irrigation system,ac unit,hvac unit,central air conditioning]
Design and Style:
    Interior Design:
        [paint,style,home style,architecture,woodwork,ensemble,accent,open floor plan,drawing]
    Aesthetics:
        [elegance,sophistication]
    Architectural Styles:
        [tudor,colonial,craftsman,farmhouse]
Smart Home Features:
    [smart home technology,surround sound,home technology,camera]
Lifestyle Features:
    Work from Home:
        [workspace,home office]
    Entertainment:
        [entertaining space,party,entertainment options,wet bar,entertainment]
Sustainability Features:
    [solar system,sustainability,solar,heated floors,solar panels,tankless water heater]
Real Estate Financial and Legal Aspects:
    [condo fee,hoa fee,hoa fees,equity,hoa dues,condo fees,cdd fees,occupied,rental potential,income potential,appreciation,airbnb,investment opportunity,investor opportunity,warranty,pricing,rental income,income,financing,utility,sale,closing,furnished,slip,tax,flip tax,abatement,zoning,hoa,rental cap,option]
Water Features:
    [soaking tub,softener]
Views and Scenery:
    [mountain views,lake views,ocean views,sunset,city views,skyline,skyline views]
Property Characteristics:
    Specialty Rooms:
        [wine cellar,media room,suite]
    Distinctive Interior Elements:
        [exposed brick,high ceilings]
    Exterior Appearance:
        [curb appeal,facade,exterior paint]
    Atmosphere:
        [oasis,retreat,sanctuary,flow]
    Environment:
        [surroundings]
    Property Metrics:
        [lot,corner lot,sqft,br,walk score,foot,inch]
    Property Condition:
        Improvements:
            [improvement,tlc,fixer,flooded]
        Age and Status:
            [new,renovated,remodeled,renovated,rehabbed,home age,upgrade,update,built,finish,updated,move,readiness,move-in ready,maintained]
Real Estate Industry:
    [builder,agent]

\end{lstlisting}

\revise{
\subsection{Diversity of The Real Estate Market in Chicago}
\label{app: diversity_of_chicago}

In this section, we analyze the diversity of the real estate market in Chicago compared to other major US cities. We use two quantitative signals: (1) the diversity of home types measured by entropy, and (2) the dispersion of prices measured by percentile ratios (p90/p10 and p75/p25). Higher values in either metric indicate a more heterogeneous market. The home type entropy for each city is summarized in \autoref{tab:home-type-entropy-simple}, and the cross-city price dispersion is reported in \autoref{tab:price-dispersion-simple}.

\begin{table}[h]
    \centering
    \small
    \caption{Home type entropy across major US cities. Higher entropy indicates a more balanced home-type distribution. Chicago exhibits the highest diversity.}
    \label{tab:home-type-entropy-simple}
    \begin{tabular}{l c}
        \toprule
        City & Home Type Entropy \\
        \midrule
        \textbf{Chicago, IL} & \textbf{0.8613} \\
        Seattle, WA          & 0.8415 \\
        San Jose, CA         & 0.8399 \\
        Los Angeles, CA      & 0.8018 \\
        San Francisco, CA    & 0.7851 \\
        Washington, DC       & 0.7796 \\
        Portland, OR         & 0.7434 \\
        Denver, CO           & 0.7064 \\
        San Diego, CA        & 0.6849 \\
        Philadelphia, PA     & 0.6375 \\
        \bottomrule
    \end{tabular}
\end{table}

\begin{table}[h]
    \centering
    \small
    \caption{Price dispersion across cities. Higher percentile ratios indicate larger heterogeneity in listing prices. Chicago shows the strongest price dispersion.}
    \label{tab:price-dispersion-simple}
    \begin{tabular}{l c c}
        \toprule
        City & Price p90/p10 & Price p75/p25 \\
        \midrule
        \textbf{Chicago, IL} & \textbf{10.09} & \textbf{3.36} \\
        Seattle, WA          & 5.44 & 2.07 \\
        San Jose, CA         & 4.81 & 2.32 \\
        Los Angeles, CA      & 5.48 & 2.37 \\
        San Francisco, CA    & 5.30 & 2.28 \\
        Washington, DC       & 7.03 & 2.50 \\
        Portland, OR         & 5.09 & 2.24 \\
        Denver, CO           & 5.85 & 2.47 \\
        San Diego, CA        & 5.44 & 2.32 \\
        Philadelphia, PA     & 5.97 & 2.38 \\
        \bottomrule
    \end{tabular}
\end{table}

Overall, Chicago emerges as the most diverse market among the major cities examined. As shown in \autoref{tab:home-type-entropy-simple}, it has the highest home type entropy, indicating a well-balanced mix of condos, single-family homes, multi-family units, and townhouses. Meanwhile, \autoref{tab:price-dispersion-simple} shows that Chicago also exhibits the strongest price dispersion, reflecting a wide range of housing options across different price tiers. Together, these signals highlight Chicago as a particularly heterogeneous and versatile real estate market.
}

\section{Hallucination Experiment Details}
\label{app: hallucination_experiments_details}

In this section, we introduce implementation details for hallucination verification experiments. We will introduce both automatic evaluation and human evaluation. 

\subsection{Automatic Evaluation}

We adopt fine-grained fact-checking based on GPT-4o for automatic evaluation, similar to the pipeline introduced in FActScore\citep{min2023factscore}. Specifically, we select \textit{price}, \textit{living area} (in sqft), \textit{\#{bedrooms}} and \textit{\#{bathroom}} as $X_{\text{hard}}$ and \textit{home insights}, \textit{address} as $X_{\text{soft}}$  according to a prior survey of user preference. 

We use structured output API\footnote{\url{https://platform.openai.com/docs/guides/structured-outputs/introduction}} on OpenAI to setup 
$\text{eval}_{\text{soft}}(L, x)$ {and} $\text{eval}_{\text{hard}}(L, x)$. This means in both cases, we need to first define the structured output class specification and then prompt the model with it.

For $\text{Faithful}_\text{hard}$, our structured output class specification is:
\begin{lstlisting}
    class MainInfo(BaseModel):
        price_mentioned: bool
        price: float
        living_area_mentioned: bool
        living_area: str
        bedrooms_mentioned: bool
        bedrooms: float
        bathrooms_mentioned: bool
        bathrooms: float
        address_mentioned: bool
        address: str
\end{lstlisting}
and our prompt for $\text{eval}_{\text{hard}}(L, x)$  is:
\begin{lstlisting}
messages=[
    {"role": "system", "content": "Extract Real Estate Information. Find the price (e.g, 290000.0), living area (e.g., '990.0 sqft'), bedrooms (e.g., 2) and bathrooms (e.g., 3) from the description. Not all information may be present, so you also have to determine whether each field is mentioned or not."},
    {"role": "user", "content": {description}}
]
\end{lstlisting}

We then compare the extracted information with $\text{supp}(L, X_{\text{hard}})$ to compute $\text{Faithful}_\text{hard}$. If certain attributes are mentioned (i.e., \textit{xx\_mentioned}=True) and the corresponding extracted values matched the listing info $\text{supp}(L, X_{\text{hard}})$, then we will give one score, otherwise zero.  

For $\text{Faithful}_\text{soft}$, we will compute it in two stages. First, we will conduct attribute extraction as in $\text{Faithful}_\text{hard}$, but with a different set of attributes $X_\text{soft}$. Our structured output class specification is:
\begin{lstlisting}
    class MainInfo(BaseModel):
        home_insights_mentioned: bool
        home_insights: list[str]
        address_mentioned: bool
        address: str
\end{lstlisting}
and our prompt is:
\begin{lstlisting}
example_home_insights =["Large island", "Oversized bathroom", "Open floor plan", "Lake views", "Orange l lines", "Newer stainless steel appliances", "Gorgeous hardwood floors", "Tons of cabinet space", "In-unit washer and dryer", "Skyline view", "Private balcony", "Beautiful city"]
example_addr = "1255 S State St UNIT 703 Chicago IL 60601"
messages=[
    {"role": "system", "content": "Extract Real Estate Information. Find the home insights (e.g., {example_home_insights}), and address (e.g., {example_addr}) from the description. Not all information may be present, so you also have to determine whether each field is mentioned or not."},
    {"role": "user", "content": {description}}
]    
\end{lstlisting}

In the second stage, we will use JSON mode API\footnote{\url{https://platform.openai.com/docs/guides/structured-outputs/json-mode}} to check whether the extracted attributes match $\text{supp}(L, X_{\text{soft}})$. Our matching prompt is:
\begin{lstlisting}
Given the following information:

1. Description: {description}
2. True value for {attribute_name}: {json.dumps(true_value)}
3. Extracted value for {attribute_name}: {json.dumps(extracted_value)}

Please analyze how well the extracted value matches the true value, considering the context provided in the description.

For 'home_insights', consider it a good match if a significant subset of the true insights is correctly identified.
For 'address', consider it a good match if at least a subset (e.g., city/state) is correctly identified, given it was mentioned in the description.

Provide a score between 0 and 10, where:
0 = Completely incorrect or irrelevant
5 = Partially correct or relevant
10 = Perfect match

Respond with a JSON object in the following format:
{{
    "score": int
}}

Where 'score' is an integer between 0 and 10.    
\end{lstlisting}
Finally we sum up all scores to compute $\text{Faithful}_\text{soft}$.

\subsection{Human Evaluation}
\label{app: hallucination_human_eval}
\begin{figure}
      \centering
  \includegraphics[width=0.5\linewidth]{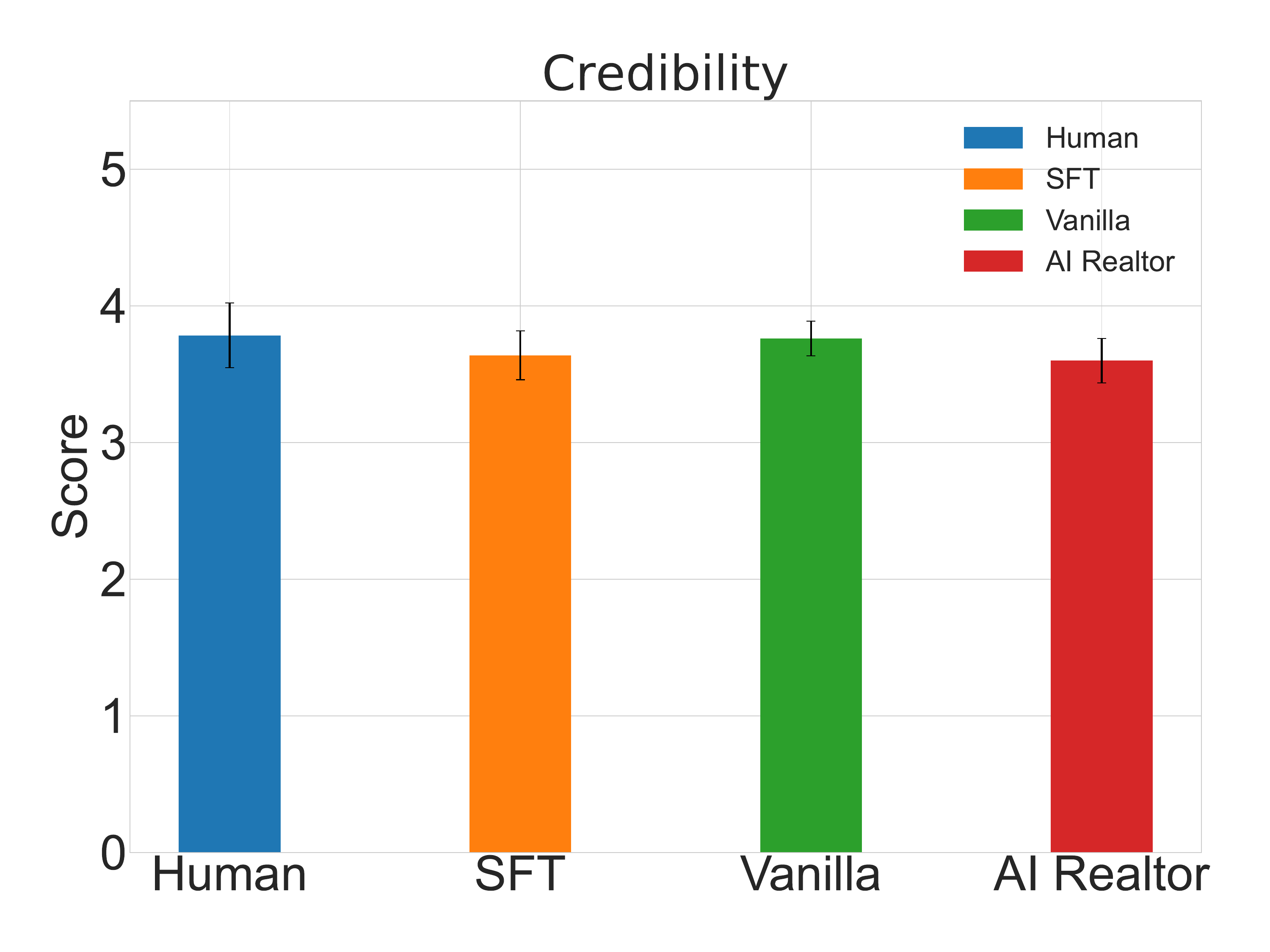}
   \vspace{- 0.1cm}
    \caption{Credibility Scores for Hallucination Checks.}
    \label{fig:hallucination_comparison_credibility}
\end{figure}

We recruit human annotators to replicate GPT-4o’s hallucination checks and assess the reliability of its automatic evaluations. \revise{To ensure consistency with the LLM judge, we define factuality identically for human raters: verifying that claims made in the description are strictly grounded in the provided attribute set $X$.} In addition to the two factual attributes evaluated by GPT-4o—$X_{\text{hard}}$ and $X_{\text{soft}}$—we include an additional stylistic check: \textbf{credibility}, which captures users’ emotional judgment of whether the persuasive description feels trustworthy.

\begin{figure}[ht]
    \centering

    \begin{subfigure}[t]{0.48\linewidth}
        \centering
        \includegraphics[width=\linewidth]{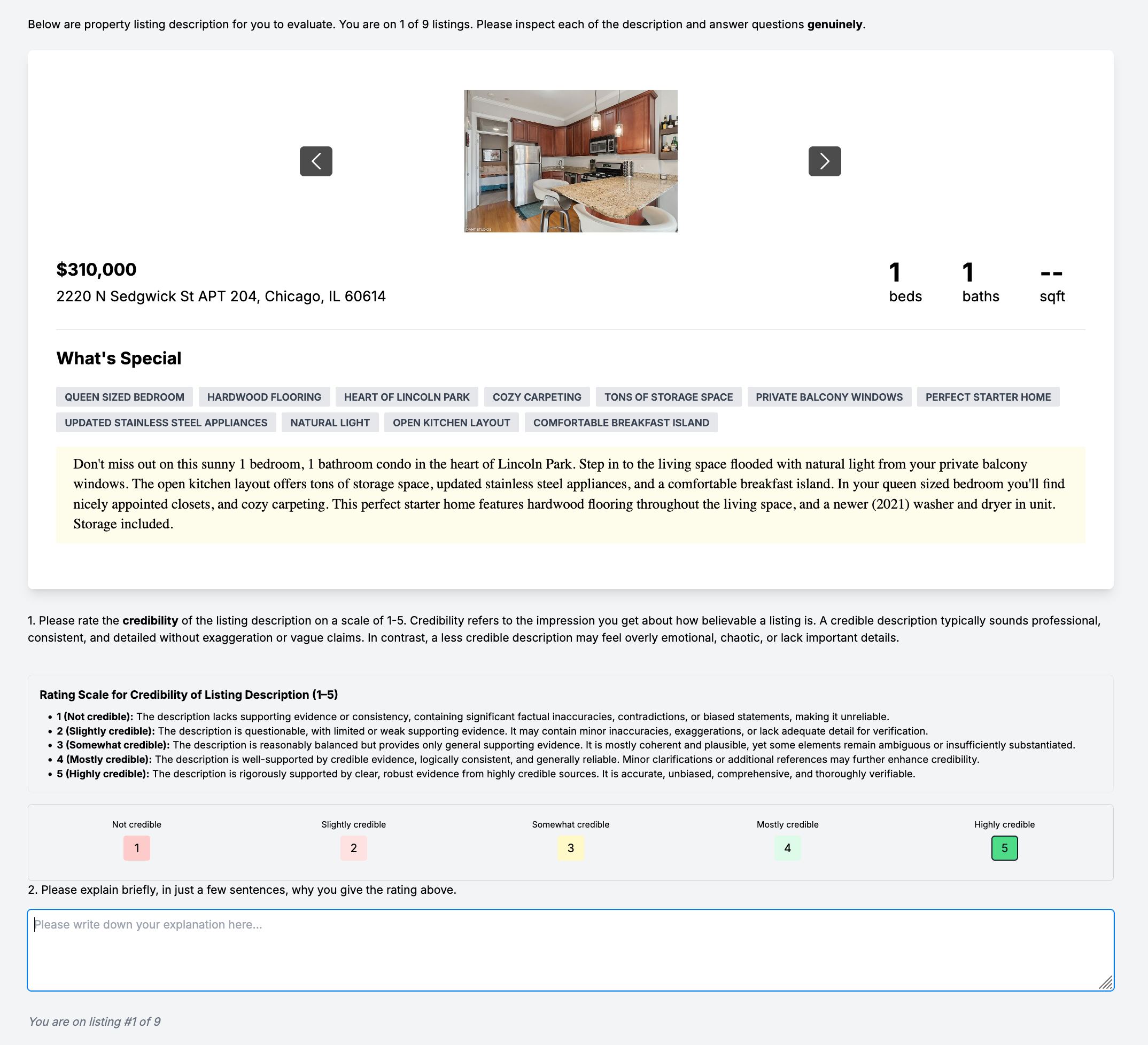}
        \caption{Credibility Evaluation Interface}
        \label{fig:hallucination_interface_credibility}
    \end{subfigure}
    \hfill
    \begin{subfigure}[t]{0.48\linewidth}
        \centering
        \includegraphics[width=\linewidth]{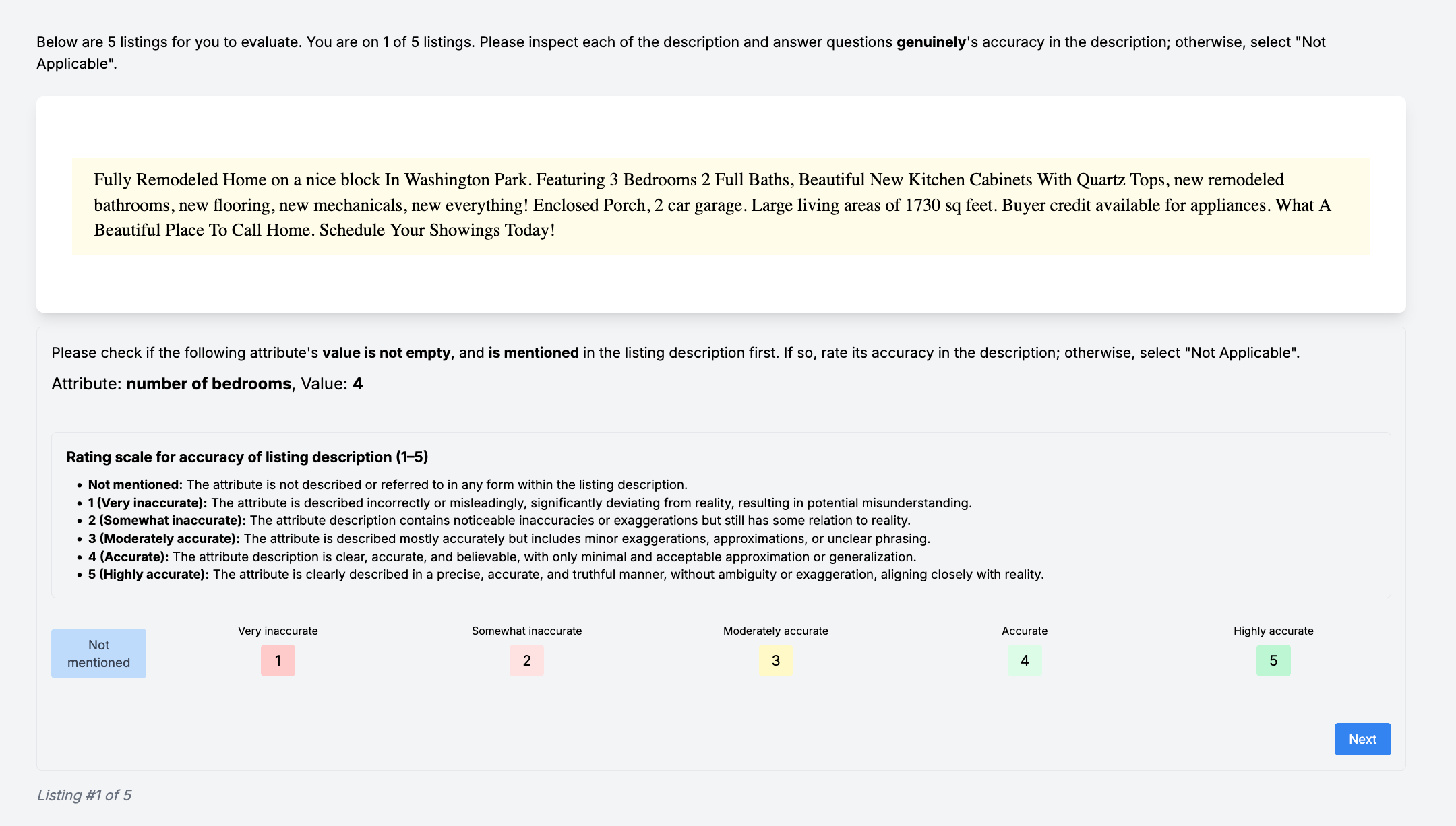}
        \caption{Hard Attribute Evaluation Interface}
        \label{fig:hallucination_interface_hard}
    \end{subfigure}

    \vspace{0.4cm}

    \begin{subfigure}[t]{0.6\linewidth}
        \centering
        \includegraphics[width=\linewidth]{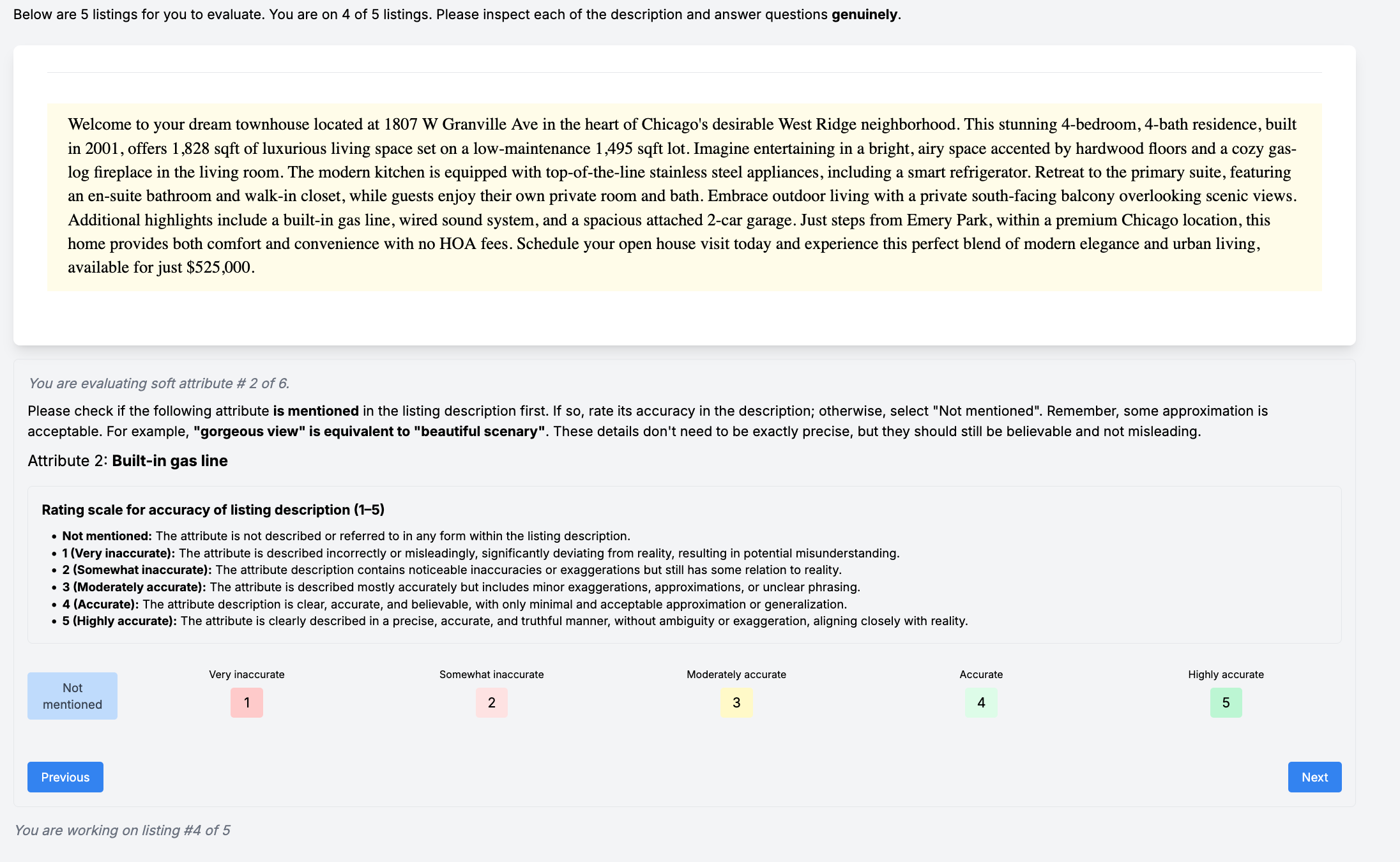}
        \caption{Soft Attribute Evaluation Interface}
        \label{fig:hallucination_interface_soft}
    \end{subfigure}

    \caption{Interfaces used in the hallucination checks.}
    \label{fig:hallucination_instructions_combined}
\end{figure}

Given an attribute set $X$ and a description $L$, either sampled from model- or human-generated outputs, we ask users to (1) rate the credibility of $L$ on a 1–5 scale (\cref{fig:hallucination_interface_credibility}), (2) 
evaluate how well each hard attribute $x_{\text{hard}} \in X_{\text{hard}}$ is reflected in $L$, 
if it is mentioned ($X_{\text{hard}} \in \text{supp}(L, X_{\text{hard}})$) (\cref{fig:hallucination_interface_hard}), and (3) assess how well each soft attribute $x_{\text{soft}} \in X_{\text{soft}}$ is reflected, if it is mentioned ($x_{\text{soft}} \in \text{supp}(L, X_{\text{soft}})$) (\cref{fig:hallucination_interface_soft}). The instruction files provided to human annotators will be submitted in a separate supplementary file. 

\begin{figure}[ht]
    \centering

    \begin{subfigure}[t]{0.48\linewidth}
        \centering
        \includegraphics[width=\linewidth]{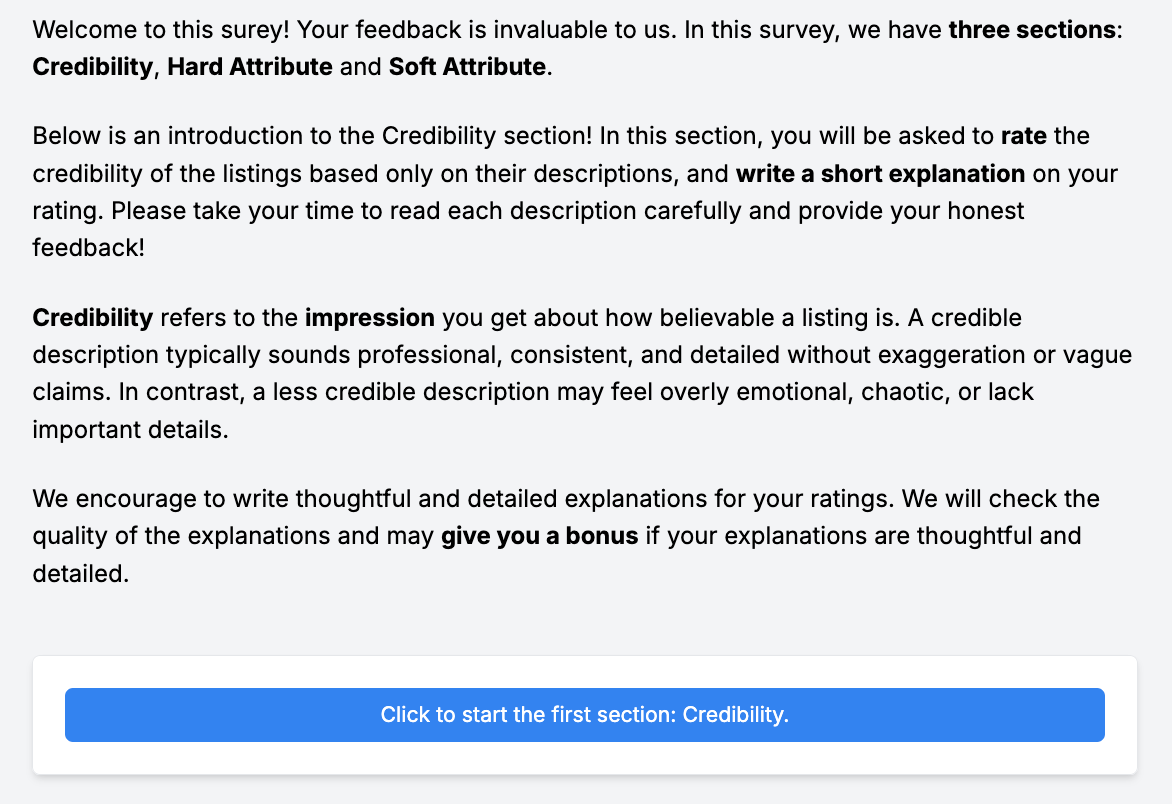}
        \caption{Credibility Evaluation Instruction}
        \label{fig:hallucination_instruction_credibility}
    \end{subfigure}
    \hfill
    \begin{subfigure}[t]{0.48\linewidth}
        \centering
        \includegraphics[width=\linewidth]{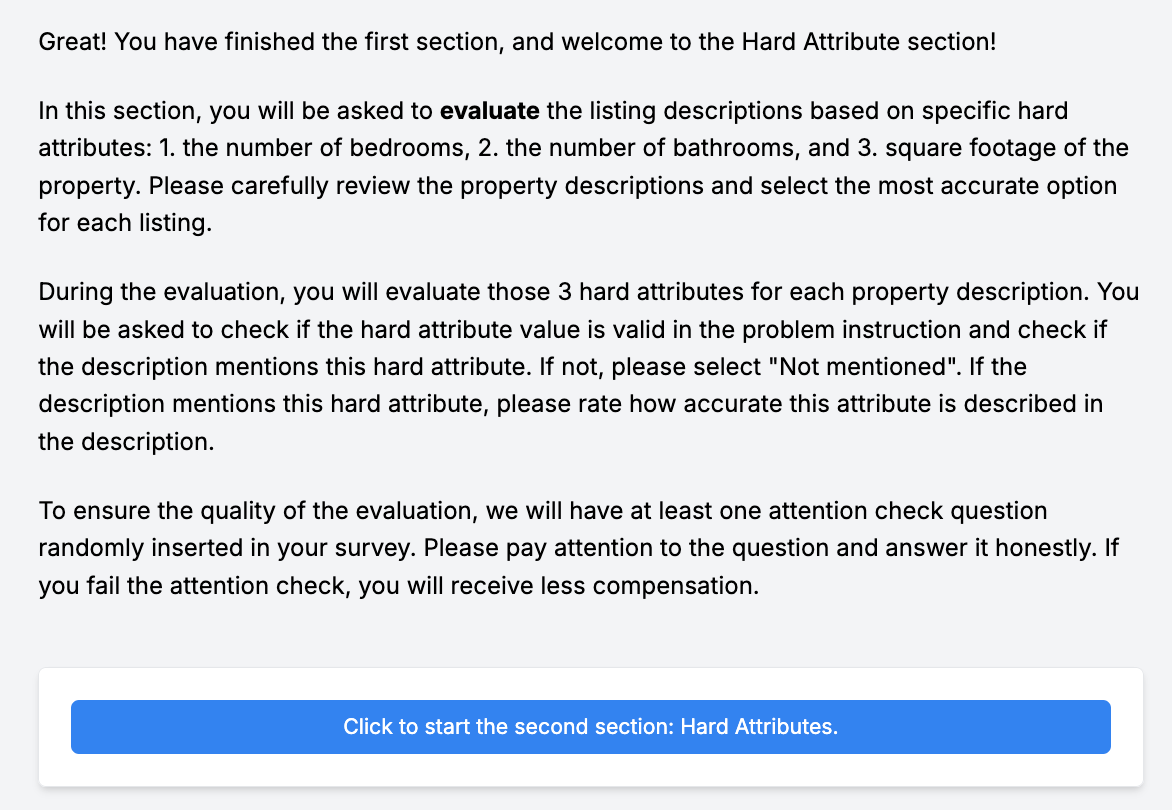}
        \caption{Hard Attribute Evaluation Instruction}
        \label{fig:hallucination_instruction_hard}
    \end{subfigure}

    \vspace{0.4cm}

    \begin{subfigure}[t]{0.6\linewidth}
        \centering
        \includegraphics[width=\linewidth]{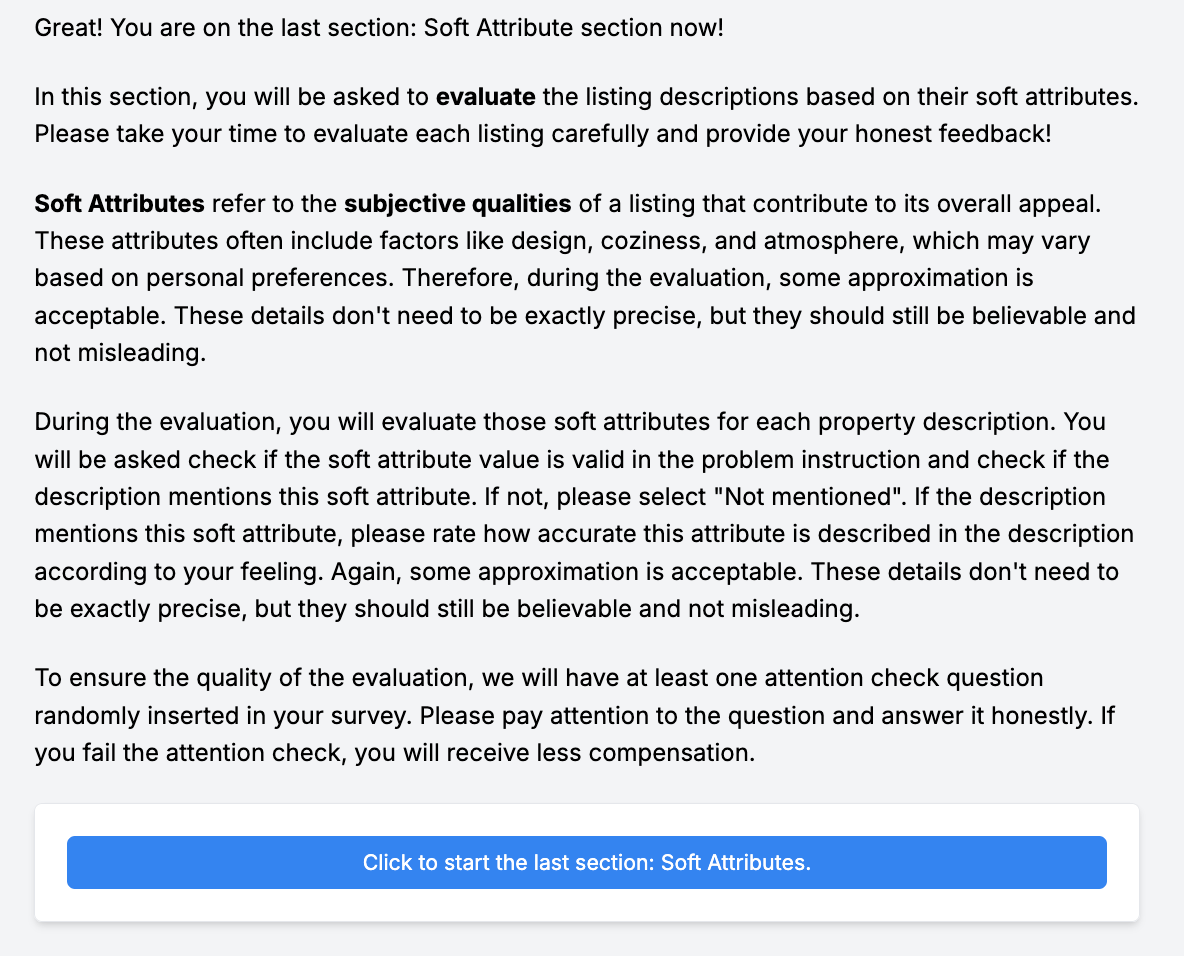}
        \caption{Soft Attribute Evaluation Instruction}
        \label{fig:hallucination_instruction_soft}
    \end{subfigure}

    \caption{Interfaces used in the hallucination checks.}
    \label{fig:hallucination_interfaces_combined}
\end{figure}

As shown in \cref{fig: hallucination_comparison}, and consistent with findings in \cref{sec: exp_hallucination_verification}, \agentname achieves the highest faithfulness on $X_{\text{hard}}$, while human-written descriptions score lowest in credibility. For evaluations on $X_{\text{soft}}$ (\cref{fig: hallucination_comparison}) and credibility (\cref{fig:hallucination_comparison_credibility}), which requires more subjective judgment, the performance of \agentname is comparable to that of humans, suggesting \agentname does not rely on hallucination or deception to persuade users. 

\section{Prompts}
\label{app: prompt-design}
\subsection{Keyword Extraction Prompt}
\label{app: keyword_extraction_prompt}%
    \begin{lstlisting}
`Your task is to extract attractive keywords. (e.g., 'modern amenities', 'great views', 'lush landscaping', 'bamboo flooring'). Please express these keywords as phrases or single word from the following house description. Each keyword should be separated by a comma. \n\nDescription: {desc}\n\nKeywords: 
\end{lstlisting}
\subsection{Keyword Extraction Normalization Prompt}
\label{app: keyword_normalization_prompt}
\begin{lstlisting}
    "Please remove the quantifiers, numbers, adjectives or any modifiers in the provided input. "
    "Uppercase or lowercase doesn't matter. "
    "If the given input is already precise enough, please provide the same input."
    "If you are not sure what to do, please also provide the input as it is. "
    "Do not explain or provide additional information."
    "Here are a few examples:"
    "\n\nInput: Two Bedrooms.\n\nOutput: Bedrooms."
    "\n\nInput: Newly Renovated Kitchen.\n\nOutput: Kitchen."
    "\n\nInput: landscape. \n\nOutput: landscape."
    "[Example Ends]"
    "Now, given the Input, please precisely provide the Output."
    "\n\nInput: {}\n\nOutput (should only be a noun phrase or keyword): "
\end{lstlisting}
\subsection{Schema Induction Prompt}
\label{app: schema_induction}
\begin{lstlisting}
Here is an initial listing keyword schema that I have, but it may not be comprehensive. I have a manually extracted comprehensive keyword list, but there are many duplicated words (e.g., different keywords may bear similar semantic meanings) and some of them may inspire new categories in this schema. I will give you that 1k+ keyword list in a file and the schema below. Can you do it this way: for every 100 keywords in the file, either try to assign it to one of the categories below, or create a new (sub)category and assign the keyword to this new (sub)category. You CANNOT use too broad categories like "others" "misc" and "uncategorized". Only create informative categories if necessary. Give me the final zip files containing all 100-ish intermediate assignment results. Each result should be represented as a JSON-like file with key=subcategory, value=[list_of_original_keywords_in_file], or key=category, value=subcategory (in other words, I want a rich hierarchical structure with the leaf nodes as a list of original keywords in the file). 

###schema### 
Appliances:
    Refrigerator
    Oven
    Dishwasher
    Washer/Dryer
    Microwave
    Garbage Disposal

Transportation:
    Garage
    Carport
    Parking Space
    Public Transit Nearby

Interior Features:
    Hardwood Floors
    Fireplace
    Central Air Conditioning
    Walk-in Closet
    Open Floor Plan
    High Ceilings

Exterior Features:
    Balcony
    Patio
    Deck
    Fenced Yard
    Garden
    Pool

Building Features:
    Elevator
    Fitness Center
    Laundry Room
    Security System
    Concierge

Utilities:
    Water
    Gas
    Electricity
    Cable/Satellite TV
    Internet

Neighborhood Features:
    Nearby Schools
    Parks
    Shopping Centers
    Restaurants
    Hospitals
    Recreation Facilities
\end{lstlisting}
\subsection{Feature Extraction Based on Description Prompt}
\label{app: feature_extraction_based_on_description_prompt}
\begin{lstlisting}
"Your task is to determine whether the given feature is mentioned in the description. The meaning of the feature will be explained by example keywords. Only respond with 'YES' or 'NO'. "
"Feature: {feature_name}. \n\nExample Keywords for explaining this feature: {keywords}\n\n"
"\n\nDescription: {human_description}\n\nResponse (Yes/No): "
\end{lstlisting}

\subsection{Persuasive Language Generation with Personalized Features}
\label{app: highlight-preference-prompt}
\begin{lstlisting}
"Your task is to generate a marketing description for a real estate listing with the provided features to highlight, and the client's preferences.
    - The listing has the following attributes:\n{attributes}
    - The listing has the following features (accounted for the client's preference) that are worth highlighting:\n{ highlight_features_reweighted }
    - The client has the following general preferences:\n{user_preference}
    - The client has the following specific preferences over features:\n    
    {feature_preference}
    - You should emphasize the feature or attributes that matches with the user's preference.
    Make sure the description is persuasive while concise under one paragraph."
\end{lstlisting}

\subsection{Persuasive Language Generation with Localized Feature Prompt}
\label{app: surprisal-prompt}

\begin{lstlisting}
"Your task is to generate a marketing description for a real estate listing with the provided features to highlight and a list of attributes that are competitive among similar listings."
    - The listing has the following attributes:\n{attributes}
    - Compared with {K} similar listings, the listing stands out in the following features that you want to emphasize:
    {surprisal_features}
    - Compared with listings in Chicago, the following features of this listing are competitive:\n
    {city_rankings}
    - Compared with listings in this neighborhood {neighbourhood}, the following features of this listing are competitive:\n
    {neighourhood_rankings}
    - Compared with listings in this zipcode {zipcode}, the following features of this listing are competitive:\n
    {zipcode_rankings}
    - Finally, You should explicitly highlight the listing features or attributes that stands out above or those ones that exactly matches with the user's preferences as a surprise factor.
    Make sure the description is persuasive while concise under one paragraph."
\end{lstlisting}
\subsection{User Simulation Prompt}
\label{app: simulation_prompt}
To avoid positional bias as demonstrated in \citep{zheng2023judging}, for each pairwise comparisons of descriptions generated by different models, we will prompt the GPT-4o-mini twice to generate separate scores as integers within $[0,100]$, and compare the final scores to decide which model wins. The prompt below shows an example of this prompt to obtain GPT-4o-mini judgement for the first description presented. ``Description 0'' and ``Description 1'' refers to descriptions generated by different models and are randomly shuffled. 
\begin{lstlisting}
You will be given a user profile, a listing and two descriptions of this listing. Optionally, you may also be given the user's history of preferences. Your task is to predict which description the user would prefer. \n\n
User Profile: {user_profile}
Listing: {listing}\n\n
Description 0: {description_0}\n\n
Description 1: {description_1}\n\n
Please first generate an analysis of the user's profile and history (if available), and then analyze why the user might prefer the first description. You can use the following format: 'The user might prefer the first description because...'
The score for the first description (an integer within [0, 100]): 
\end{lstlisting}

\subsection{Retriever Configuration}
\label{app: search_engine}

\begin{lstlisting}
    "mappings": {
            "properties": {
                "bedrooms": {"type": "float"},
                "bathrooms": {"type": "float"},
                "price": {"type": "float"},
                "description": {"type": "text"},
                "area": {"type": "float"},
                "street_address": {"type": "text"},
                "home_type": {"type": "keyword"},
                "state": {"type": "keyword"},
                "city": {"type": "keyword"},
                "page_view_count": {"type": "float"},
                "favorite_count": {"type": "float"},
                "home_insights": {"type": "keyword"},
                "neighborhood_region": {"type": "keyword"},
                "id": {"type": "keyword"}
            }
        }
\end{lstlisting}

%% file: past_papers/ai_realtor/statement.tex
\section{Artifact Evaluation}

We have open-sourced the code and part of anonimized data in \href{https://github.com/yangalan123/AI-Realtor-Codebase}{this github repo}.  

\section{Usage of Large Language Models}
In this work, we mainly use LLMs for the following purposes:
\begin{enumerate}
    \item Aid or Polish Writing (Gemini 2.5 Pro, ChatGPT 4/5)
    \item Literature Retrieval and Discovery (e.g., finding related work) (Gemini 2.5 Pro Deep Research, ChatGPT Deep Research)
    \item Assisting Code Writing and Debugging (Claude 3.5 Sonnet)
\end{enumerate}
We fully understand the responsibility of using LLMs in academic research. We carefully monitor any potential problems, such as plagiarism or scientific misconduct (e.g., fabrication of facts) when using LLMs. We make sure these problems do not occur in the paper. 

%% file: past_papers/addiction/sections/appendix.tex
\begin{table*}[!ht]
\centering
\resizebox{0.95\textwidth}{!}{%
\renewcommand{\arraystretch}{1.05}
\begin{tabular}{|c|l|}
\hline
Medical Use
& \begin{tabular}[c]{@{}l@{}}Medical use is defined as the use of prescription opioids that were prescribed by a medical\\ professional for the purpose of treating a medical condition\end{tabular}                                                              \\\hline
Misuse                                                 & \begin{tabular}[c]{@{}l@{}}Misuse is defined as the use of a substance that does not follow medical indications or prescribed \\ dosing. Substances are commonly used for nontherapeutic purposes to obtain psychotropic (eg, \\ euphoric, seditative, or anxiolytic) effects. Misuse is not restricted to prescription opioids.\end{tabular}                                \\ \hline
Addiction                                              & \begin{tabular}[c]{@{}l@{}}Addiction is defined as compulsive opioid use that occurs despite personal harm or negative \\ consequences. Addiction may involve impaired control and craving, neurobiologic dysfunction,\\ physical and psychological dependence, and withdrawal.\end{tabular}                                                                 \\ \hline
Recovery                                               & \begin{tabular}[c]{@{}l@{}}Recovery is a process of change through which individuals improve their health and wellness,\\ live a self-directed life, and strive to reach their full potential without using opioids.\end{tabular}                                                                 \\ \hline
Relapse                                                & Relapse is defined as the return to opioid use after an attempt to quit.
         \\ \hline
Not Using
& \begin{tabular}[c]{@{}l@{}}Posts should be labeled 'Not Using' which are about substances other than opioids \\ (e.g.,
marijuana), another person who uses opioids (e.g., family or friend), general questions \\ about opioids without evidence that the persons use opioids, and irrelevant information.
\end{tabular} \\ \hline
\end{tabular}
}
\caption{Expert guidelines on how to assign each post one of the six stages of the OUD continuum} 
\label{guidelines}
\end{table*}

\section{Annotation Guideline}
\label{app: annotation_guideline}
A brief annotation guideline created by experts is shown in \cref{guidelines}, which explains the definition for each OUD category. This guideline also comes with example posts picked by experts that help annotator under the definitions and we show them in \cref{tab:guideline_examples}. The full guideline is too large to put in this paper so we will release it in our GitHub project. 

Experts also help draft FAQs for clarification in the initial trial of annotations. Examples of FAQs are shown below:
\begin{dialogue}
    \speak{Question} What if the post described family, friend, or peer opioid use and there is no evidence that the person posting used opioids?
    \speak{Answer} This post should be labeled ‘not using’ because there is no evidence that the individual posting the comment used opioids.
    \speak{Question} What if the post discusses using stimulants, marijuana, or other drugs that are not opioids?
    \speak{Answer} This post should be labeled ‘not using’ because this study is specifically focused on understanding the development and advancement of opioid use disorder.
    \speak{Question} Is ‘misuse’ restricted to prescription opioids?
    \speak{Answer} We have decided for the purpose of this study that misuse will NOT be restricted to prescription opioids. Therefore, if someone describes trying a synthetic or semi-synthetic opioid (e.g., heroin) or using it infrequently, but does not display signs of being addicted, this post should be labeled ‘misuse.’
    \speak{Question} What if the post asks a question about opioid use, but does not provide evidence that the individual posting the comment used opioids?
    \speak{Answer} This post should be labelled ‘not using’ because there is no evidence that the individual posting the comment used opioids. They may just be curious.
    \speak{Question} If someone reports using drugs that are NOT opioids during a period of time when they are attempting to quit (i.e., when they are in recovery), should this be considered ‘relapse?’
    \speak{Answer} Because this study is focused on opioid use disorder, we have defined relapse as use of opioids after an attempt to quit. Thus, if the individual used other drugs that are not opioids during recovery, we will not consider this relapse.
\end{dialogue}

\begin{table*}[!ht]
\centering
\resizebox{0.95\textwidth}{!}{%
\renewcommand{\arraystretch}{1.05}
\begin{tabular}{|c|l|}
\hline
Medical Use
& \begin{tabular}[c]{@{}l@{}}I got pretty decent surgery on my feet and was prescribed 400 mg of oxy after takeing that\\ In about 10 days as needed due to pain ( never takeing more then prescribed ) \\but I have had minor withdrawal symptoms I took a 3 day break \\when do you think i can start taking it agian when my foot hurts and not withdrawal\end{tabular}                                                              \\\hline
Misuse                                                 & \begin{tabular}[c]{@{}l@{}}So I was given vicoprofen (7.5 hydrocodone to 200tylenol) for a severe toothache. \\ I have been using it as prescribed but dumb ass me decided to take quite a large dose last night after missing a few normal doses. \\ If I go back to using the normal doses now, after one large one, is it still going to be effective? \\Or should I wait and if so how long."\end{tabular}                                \\ \hline
Addiction                                              & \begin{tabular}[c]{@{}l@{}}I have been on opiates (oxycodone/contin) for like 5-6 years.\\ Started off really small, got really big, now at like medium use- compared to before. \\ I spent the last year or so very slowly tapering from my high of 330mg/day to now about 80mg/day. \\At this point is just maintenance to be able to function properly in my everyday life w out being sick or too tired.\end{tabular}                                                                 \\ \hline
Recovery                                               & \begin{tabular}[c]{@{}l@{}}"7 days clean from heroin today after having been IV'ing it on my daily basis since August, 2020"\end{tabular}                                                                 \\ \hline
Relapse                                                &\begin{tabular}[c]{@{}l@{}} "i made it 70 days clean. now i'm back to square one. \\ i wish i could stop but i can't. now i'm shooting 2 grams a day, plus 2-4 grams of coke a day. \\everytime i relapse i get more and more addicted. anyone else experience this ? that when you relapse it gets more out of control.\\ but godam i love it, i love the feeling, the lifestyle. "\end{tabular}
         \\ \hline
Not Using
& \begin{tabular}[c]{@{}l@{}}"How do you feel about Oxford houses/halfway houses/sober houses?"\\ "Dreary , rainy day here , thought about using , now binge watching Reno 911 instead . It’s so funny lol
\end{tabular} \\ \hline
\end{tabular}
}
\caption{Expert guidelines on example posts for each category} 
\label{tab:guideline_examples}
\end{table*}

\begin{table*}[htbp!]
\centering
\resizebox{0.90\textwidth}{!}{
\begin{tabular}{lcccc}
\toprule
\textbf{Category} & \textbf{T5-11B w/ Explanation} & \textbf{T5-11B w/o Explanation} & \textbf{GPT-4 w/ Explanation} & \textbf{GPT-4 w/o Explanation} \\
\midrule
Addiction & \(88\% / 98\%\) & \(84\% / 94\%\) & \(28\% / 62\%\) & \(20\% / 49\%\) \\
Medical Use & \(84\% / 87\%\) & \(76\% / 87\%\) & \(88\% / 87\%\) & \(84\% / 87\%\) \\
Misuse & \(88\% / 70\%\) & \(88\% / 75\%\) & \(76\% / 89\%\) & \(72\% / 82\%\) \\
Not Using & \(68\% / 66\%\) & \(68\% / 63\%\) & \(36\% / 30\%\) & \(32\% / 27\%\) \\
Recovery & \(92\% / 83\%\) & \(84\% / 67\%\) & \(88\% / 81\%\) & \(84\% / 70\%\) \\
Relapse & \(84\% / 81\%\) & \(88\% / 69\%\) & \(88\% / 81\%\) & \(88\% / 69\%\) \\
\bottomrule
\end{tabular}
}
\caption{Class-wise performance decomposition for different models. Results are presented in a format of ``Accuracy on Novice Test Set/Accuracy on Expert Test Set''.} 
\label{tab:class-wise-perf}
\end{table*}

\section{Heatmap for Worker-Expert labels over the Qualification Test}
\label{app:heatmap}
\begin{figure}
    \centering
        \centering
        \includegraphics[trim={1cm 2cm 1cm 2cm},clip,width=\columnwidth]{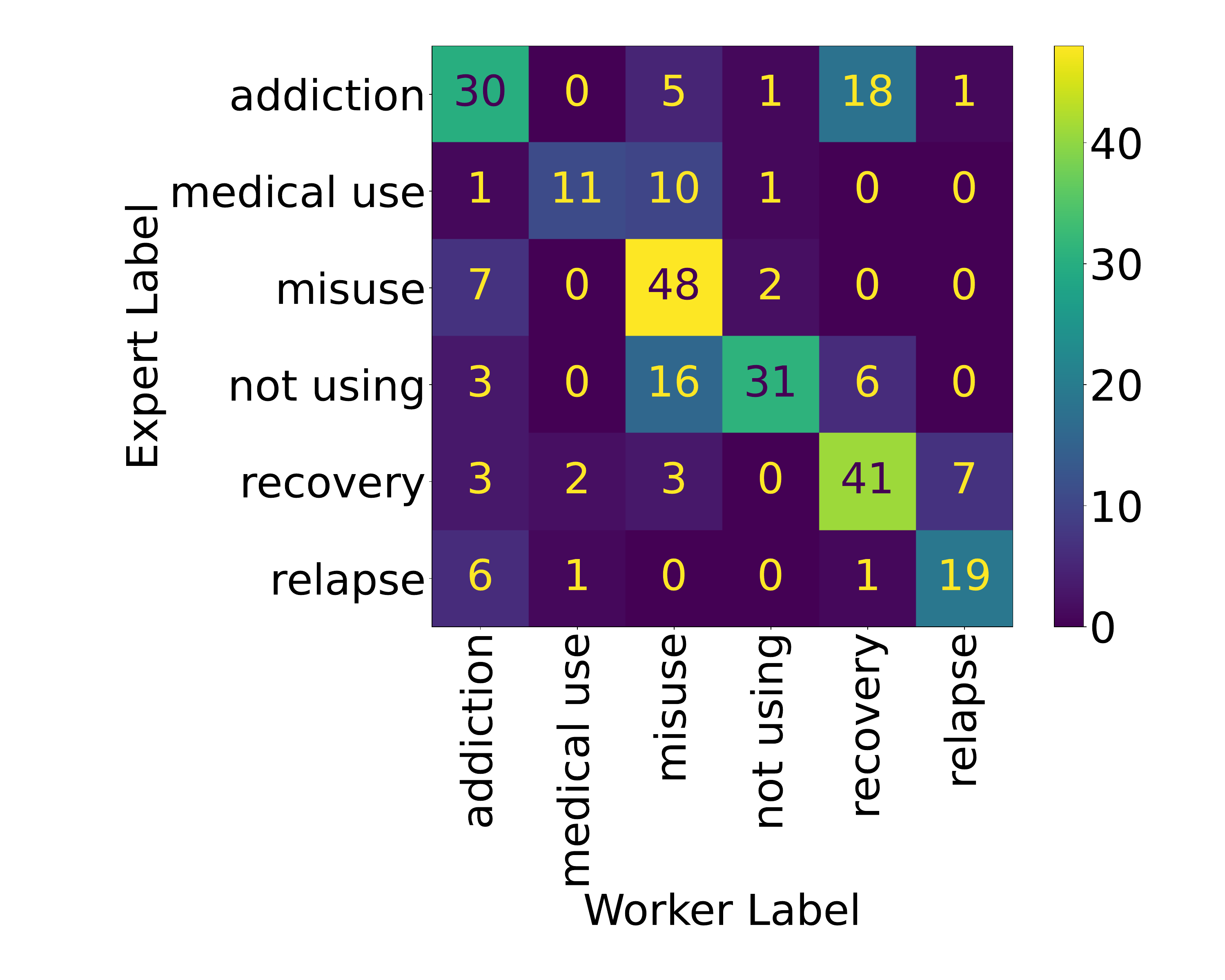}
        \caption{Heatmap for all worker labels and expert labels over the qualification test. }
        \label{fig: worker-expert-disagree}
        \vspace{-5mm}
\end{figure}

The heatmap summarizes the difference in annotations between workers and experts over the qualification test is shown in \cref{fig: worker-expert-disagree}.

\section{In-Context Samples in Few-Shot Learning Settings}
\label{app:in-context-samples}
The 13 in-context samples we used for prompting in the few-shot learning setting are shown in \cref{tab:in-context-samples}. These in-context samples are selected as the representative samples under each category after discussions with experts providing the annotations. The distribution of classes is decided based on preliminary experiments on held-out data. 

    \begin{table*}[h!]
    \centering
    \resizebox{0.98\textwidth}{!}{
    \begin{tabular}{|l|l|l|l|}
    \hline
    \textbf{Title} & \textbf{Text} & \textbf{Opioid Usage Label} & \textbf{Explanation} \\
    \hline
    Advice welcome & I am a 23yr old female, been addicted to H for 3 y... & relapse & I am a 23yr old female, been addicted to H for 3 y... \\ \hline
Nearly threw two months down the drain today. & Well everyone I've been clean from Heroin for the ... & recovery & Well everyone I've been clean from Heroin for the ... \\ \hline
2weeks clean from all opiates.i just want to vent a bit. & So two weeks ago I quit my job, opiate use, and go... & recovery & 2weeks clean from all opiates \\ \hline
Listening to Christmas music.... & I could have been someone ''Well so could anyone..... & not using & I could have been someone ''Well so could anyone..... \\ \hline
Heroin use & Hi, non user here, just curious as to what heroin ... & not using & Hi, non user here, just curious as to what heroin ... \\ \hline
Anyone here either a lawyer or have solid solid knowledge or experience on drug & Not really comfortable discussing this publicly, b... & not using & Anyone here either a lawyer or have solid solid kn... \\ \hline
Supeudol oxycodone, sniffable? & My doctor changed my oxy prescription to supeudol ... & medical use & My doctor changed my oxy prescription to supeudol ... \\ \hline
Back in the cycle.. & I started using more again (daily when I can) afte... & addiction & I started using more again (daily when I can) afte... \\ \hline
ROA 30mg roxis (blues) & I am currently on 7 blues. That i have done over t... & misuse & I am currently on 7 blues. That i have done over t... \\ \hline
Hey guys! Kinda worried! & Hey, I took abour 4 lines of heroin at 6pmAnother ... & misuse & Hey, I took abour 4 lines of heroin at 6pmAnother ... \\ \hline
Really wanting to try heroin :/ & I just wanna say before I start, Ik how bad it is ... & misuse & I’ve used weed, Xanax, coke, I’m off of 2 Kpins ri... \\ \hline

    \end{tabular}
    }

\caption{Thirteen in-context examples for each
Opioid Usage category. }
    \label{tab:in-context-samples}
    \end{table*}

\section{Post-processing needed for processing GPT-3 outputs}
\label{app:post_processing_gpt3}
In our experiments, we generally find that GPT-3 outputs cannot be taken as exact match as outputs and can contain some typos, we provide the following post-processing for it:

\begin{enumerate}
    \item We ignore any content after a newline symbol (i.e., ``$\backslash n$ '').
    \item If GPT3 responses are like ``1) ... 2) ...'', we only take the term between ``1)'' and ``2)''. 
    \item For morphological changes like predicting ``misuse'' as ``misusing'', we manually recover these changes. 
    \item For typos like ``misue'', we would manually correct it to be ``misuse''. 
\end{enumerate}

We tried to apply the same processing for GPT-4 as well, but we did not find significant changes. This may indicate GPT-4 has better instruction-following capability while GPT-3 does not.

\section{Fully Supervised Fine-Tuning Details}
\label{sec:fully_supervised_fine-tuning_detail}
In this section, we give details for fine-tuning language models under fully supervised setting.

For fine-tuning DeBERTa-v3-large~\citep{he2021debertav3}, we adopt the widely-used huggingface transformers fine-tuning implementation~\citep{wolf-etal-2020-transformers} with the learning rate of $2e-5$ and fine-tune the model for $10$ epochs. For optimizer, we use AdamW~\citep{loshchilov2018fixing}. 

For fine-tuning T5~\cite{raffel2020exploring}, we adopt the huggingface transformers implementation ~\citep{wolf-etal-2020-transformers} to fine-tune two versions of T5, the 3B model and the 11B model, respectively. We hold out 100 examples for validation from our training set to tune our models and find the best checkpoint. We use a batch size of 1024 for the 3B model and 512 for the 11B model. Further, we maintain a learning rate of 1e-4 and AdamW optimizer~\citep{loshchilov2018fixing} for both 3B and 11B models. We fine-tune all models on 4 A100 GPUs and use Deepspeed \cite{rasley2020deepspeed} integration for the 11B model. We fine-tune the 3B model for 20 epochs and the 11B model for eight epochs. During fine-tuning, we restrict the maximum sequence length of the source to 1024 (via truncation), while our target length is less than the default 128 tokens.

\section{Class-Wise Performance Decomposition}
\label{app:class-wise-main-exp}
In \cref{sec: exp}, we show the model average performance w/ and w/o explanations over all categories in \cref{tab:main_exp}. As there exist significant differences between OUD categories and their individual importance can vary depending on application purposes, we further show the class-wise performance decomposition in \cref{tab:class-wise-perf} for both expert and novice annotated test sets. 

\section{Scientific Artifacts}
In this paper, we use the following artifacts: 

\textit{cleantext}\footnote{\url{www.github.com/prasanthg3/cleantext}} (v1.1.4): is an open-source python package to clean raw text data. We use it to preprocess raw social media posts. This toolkit is released under an MIT license. 

\textit{Transformers} \cite{wolf-etal-2020-transformers}\footnote{https://github.com/huggingface/transformers} (v4.35.0): provides thousands of pretrained models to perform tasks on different modalities such as text, vision, and audio. We use it for model training and inference. This toolkit is released under an Apache-2.0 license. 

\textit{OpenAI-python}\footnote{\url{https://github.com/openai/openai-python}} (v1.0.0): provides convenient access to the OpenAI REST API from any Python 3.7+ application. The library includes type definitions for all request params and response fields, and offers both synchronous and asynchronous clients powered by httpx. We use it for prompting the GPT-series models. This toolkit is released under an Apache-2.0 license.

In addition, we plan to release our codebase and dataset under an MIT license in the formal version.